\PassOptionsToPackage{float=false}{scrhack}

\documentclass[
11pt,
english,
onehalfspacing,
liststotoc,
parskip,
headsepline,
oneside,
openany,
]{HKUThesis}

\usepackage[utf8]{inputenc}
\usepackage[T1]{fontenc}
\usepackage{wrapfig}

\usepackage{amsthm}
\usepackage{mathtools}
\usepackage{thmtools}
\usepackage{bm}
\usepackage{mathrsfs}

\usepackage{mathpazo}
\usepackage{microtype}   

\usepackage{tikz}
\usepackage{pgfplots}
\pgfplotsset{compat=1.18}
\usetikzlibrary{arrows.meta, positioning, calc, patterns, decorations.pathreplacing, shapes.geometric, fit, backgrounds}

\definecolor{cbBlue}{HTML}{0072B2}
\definecolor{cbOrange}{HTML}{D55E00}
\definecolor{cbGreen}{HTML}{009E73}
\definecolor{cbPurple}{HTML}{CC79A7}
\definecolor{cbSkyBlue}{HTML}{56B4E9}
\definecolor{cbYellow}{HTML}{F0E442}
\definecolor{cbVermillion}{HTML}{E69F00}
\definecolor{cbRed}{HTML}{CC3311}     

\colorlet{fillBlue}{cbBlue!12}
\colorlet{fillOrange}{cbOrange!12}
\colorlet{fillGreen}{cbGreen!12}
\colorlet{fillPurple}{cbPurple!12}
\colorlet{fillSky}{cbSkyBlue!12}
\colorlet{fillYellow}{cbYellow!20}
\colorlet{fillRed}{cbOrange!15}
\colorlet{fillGray}{black!6}
\colorlet{fillHighlight}{cbVermillion!15}  
\colorlet{fillSubtle}{gray!8}               

\colorlet{stackL1}{cbBlue}        
\colorlet{stackL2}{cbGreen}       
\colorlet{stackL3}{cbOrange}      
\colorlet{stackL4}{cbPurple}      
\colorlet{stackL5}{cbVermillion}  

\colorlet{colorProved}{cbGreen}          
\colorlet{colorObstruction}{cbOrange}    
\colorlet{colorOpen}{cbRed}              
\colorlet{colorTheory}{cbPurple}         
\colorlet{colorEmpirical}{cbBlue}        

\tikzset{
  thesisbox/.style={
    draw, rounded corners=3pt,
    minimum height=0.65cm, minimum width=2.2cm,
    align=center, font=\small,
    fill=fillGray, line width=0.5pt
  },
  thesisbox/blue/.style   ={thesisbox, fill=fillBlue,   draw=cbBlue!60},
  thesisbox/orange/.style ={thesisbox, fill=fillOrange,  draw=cbOrange!60},
  thesisbox/green/.style  ={thesisbox, fill=fillGreen,   draw=cbGreen!60},
  thesisbox/purple/.style ={thesisbox, fill=fillPurple,  draw=cbPurple!60},
  thesisbox/sky/.style    ={thesisbox, fill=fillSky,     draw=cbSkyBlue!60},
  thesisbox/yellow/.style ={thesisbox, fill=fillYellow,  draw=cbYellow!60!black},
  thesisbox/gray/.style   ={thesisbox, fill=fillGray,    draw=black!30},
  stacklayer/.style={
    draw, rounded corners=3pt,
    minimum width=3.2cm, minimum height=0.9cm,
    align=center, font=\small\bfseries,
    line width=0.6pt
  },
  thesisnode/.style={
    circle, draw, minimum size=0.6cm,
    inner sep=0pt, font=\footnotesize, line width=0.5pt
  },
  thesisdiamond/.style={
    draw, diamond, aspect=2.2,
    minimum width=1.6cm, font=\small,
    align=center, fill=fillYellow, line width=0.5pt
  },
  thesisarrow/.style={
    -{Stealth[length=5pt, width=4pt]},
    line width=0.7pt
  },
  thesisarrow/formal/.style={thesisarrow, color=cbBlue!70!black},
  thesisarrow/qual/.style  ={thesisarrow, dashed, color=black!40},
  thesisarrow/data/.style  ={thesisarrow, color=black!60},
  thesisarrow/bidi/.style  ={
    {Stealth[length=5pt, width=4pt]}-{Stealth[length=5pt, width=4pt]},
    line width=0.7pt, color=black!50
  },
  annot/.style      ={font=\scriptsize, align=center},
  annot/left/.style ={annot, anchor=east},
  annot/right/.style={annot, anchor=west},
  bgregion/.style={
    draw=none, rounded corners=5pt,
    fill=#1, fill opacity=0.08
  },
  compbox/.style={                         
    draw, rounded corners=2pt,
    minimum height=0.7cm, minimum width=2.0cm,
    align=center, font=\small,
    fill=fillSubtle, line width=0.5pt
  },
  dataflow/.style={                        
    -{Stealth[length=5pt, width=4pt]},
    line width=0.8pt, color=black!65
  },
  skipconn/.style={                        
    -{Stealth[length=4pt, width=3pt]},
    line width=0.6pt, dashed, color=cbPurple!70!black
  },
  chainnode/.style={                       
    circle, draw, minimum size=0.55cm,
    inner sep=0pt, font=\footnotesize,
    line width=0.5pt, fill=fillSubtle
  },
  highlight/.style={                       
    draw=colorObstruction, dashed,
    rounded corners=3pt, line width=0.8pt,
    inner sep=4pt
  },
  groupbox/.style={                        
    draw=black!20, dotted,
    rounded corners=4pt, line width=0.4pt,
    inner sep=6pt, fill=fillSubtle, fill opacity=0.6
  },
}

\pgfplotscreateplotcyclelist{thesis colors}{
  {cbBlue,       solid,    mark=*,         mark size=2pt,   line width=1pt},
  {cbOrange,     dashed,   mark=square*,   mark size=2pt,   line width=1pt},
  {cbGreen,      dash dot, mark=triangle*, mark size=2.5pt, line width=1pt},
  {cbPurple,     dotted,   mark=diamond*,  mark size=2.5pt, line width=1pt},
  {cbSkyBlue,    solid,    mark=pentagon*,  mark size=2pt,  line width=1pt},
  {cbVermillion, dashed,   mark=otimes*,   mark size=2pt,   line width=1pt},
}

\pgfplotsset{
  thesis base/.style={
    width=0.92\textwidth,
    height=0.55\textwidth,
    grid=major,
    grid style={gray!20, line width=0.3pt},
    tick label style={font=\footnotesize},
    label style={font=\small},
    title style={font=\small\bfseries},
    legend style={
      font=\footnotesize,
      draw=none, fill=none,
      /tikz/every even column/.append style={column sep=8pt}
    },
    every axis/.append style={line width=0.5pt},
    cycle list name=thesis colors,
  },
  thesis line/.style={
    thesis base,
    ymajorgrids=true,
    xmajorgrids=false,
  },
  thesis bar/.style={
    thesis base,
    ybar,
    bar width=8pt,
    enlarge x limits=0.15,
    ymajorgrids=true,
    xmajorgrids=false,
  },
  thesis scatter/.style={
    thesis base,
    only marks,
    ymajorgrids=true,
    xmajorgrids=true,
  },
  legend below/.style={
    legend style={
      at={(0.5,-0.18)},
      anchor=north,
      legend columns=3,
    }
  },
  legend below wide/.style={
    legend style={
      at={(0.5,-0.18)},
      anchor=north,
      legend columns=4,
    }
  },
}

\newcommand{\best}[1]{\ifmmode\mathbf{#1}\else\textbf{#1}\fi}

\usepackage{makecell}
\usepackage{tabularx}

\newcolumntype{L}[1]{>{\raggedright\arraybackslash}p{#1}}
\newcolumntype{C}[1]{>{\centering\arraybackslash}p{#1}}
\newcolumntype{R}[1]{>{\raggedleft\arraybackslash}p{#1}}

\usepackage{xspace}
\usepackage{rotating}

\AtEndPreamble{%
\usepackage[capitalize,noabbrev]{cleveref}%
\crefname{theorem}{Theorem}{Theorems}%
\crefname{lemma}{Lemma}{Lemmas}%
\crefname{proposition}{Proposition}{Propositions}%
\crefname{corollary}{Corollary}{Corollaries}%
\crefname{definition}{Definition}{Definitions}%
\crefname{remark}{Remark}{Remarks}%
\crefname{example}{Example}{Examples}%
\crefname{assumption}{Assumption}{Assumptions}%
\crefname{conjecture}{Conjecture}{Conjectures}%
\Crefname{conjecture}{Conjecture}{Conjectures}%
\crefname{hypothesis}{Hypothesis}{Hypotheses}%
\Crefname{hypothesis}{Hypothesis}{Hypotheses}%
\crefname{algocf}{Algorithm}{Algorithms}%
\crefname{part}{Part}{Parts}%
}

\usepackage[natbib=true,maxbibnames=99,giveninits=true,sorting=none,backend=bibtex]{biblatex}
\usepackage[autostyle=true]{csquotes}
\AtBeginEnvironment{enquote}{\itshape}

\usepackage{epigraph}
\usepackage[skins,breakable]{tcolorbox}   
\usepackage{float}

\tcbuselibrary{theorems}

\definecolor{thesisInk}          {HTML}{1C2433}
\definecolor{thesisAccent}       {HTML}{2E4E8C}  
\definecolor{thesisPositive}     {HTML}{1F6F5C}  
\definecolor{thesisCaution}      {HTML}{9A6B16}  
\definecolor{thesisCritical}     {HTML}{A01B3C}  
\definecolor{thesisIntuition}    {HTML}{6B3E8F}  
\definecolor{thesisAside}        {HTML}{6F6A82}  
\definecolor{thesisTintAccent}   {HTML}{EEF2FA}
\definecolor{thesisTintPositive} {HTML}{E8F2EE}
\definecolor{thesisTintCaution}  {HTML}{FAF1DF}
\definecolor{thesisTintCritical} {HTML}{FBE6EC}
\definecolor{thesisTintIntuition}{HTML}{F1E9F5}
\definecolor{thesisTintAside}    {HTML}{EEEDF2}
\definecolor{thesisRule}         {HTML}{E5E7EB}

\tcbset{
  v13base/.style={
    enhanced, breakable,
    colback=white, colframe=#1,
    coltitle=#1,
    fonttitle=\sffamily\bfseries\small,
    boxrule=0pt, leftrule=3pt, arc=0pt, outer arc=0pt,
    left=10pt, right=8pt, top=5pt, bottom=5pt,
    before skip=8pt plus 2pt, after skip=10pt plus 2pt,
    attach title to upper, after title={\hspace{0.5em}},
    fontupper=\normalfont\small
  },
  v13tint/.style={
    enhanced, breakable,
    colback=#1!6, colframe=#1,
    coltitle=#1,
    fonttitle=\sffamily\bfseries\small,
    boxrule=0pt, leftrule=3pt, arc=0pt, outer arc=0pt,
    left=10pt, right=8pt, top=5pt, bottom=5pt,
    before skip=8pt, after skip=10pt,
    attach title to upper, after title={\hspace{0.5em}},
    fontupper=\normalfont\small
  },
  v13filled/.style={
    enhanced, breakable,
    colback=white, colframe=#1,
    colbacktitle=#1, coltitle=white,
    fonttitle=\sffamily\bfseries,
    boxrule=0.6pt, leftrule=3pt,
    arc=2pt, outer arc=2pt,
    attach boxed title to top left={xshift=6pt,yshift*=-\tcboxedtitleheight/2},
    boxed title style={arc=2pt, boxrule=0pt},
    before skip=10pt, after skip=12pt,
    fontupper=\normalfont
  }
}

\newtcolorbox{note}[1][]{v13base=thesisAccent,
  title={\textbf{[i]}~\textsc{Note}},#1}

\newtcolorbox{tip}[1][]{v13base=thesisPositive,
  title={\textbf{[*]}~\textsc{Tip}},#1}

\newtcolorbox{warning}[1][]{v13tint=thesisCaution,
  title={\textbf{[!]}~\textsc{Warning}},#1}

\newtcolorbox{limitation}[1][]{v13tint=thesisCritical,
  title={\textbf{[X]}~\textsc{Limitation}},#1}

\newtcolorbox{intuition}[1][]{v13base=thesisIntuition,
  title={\textbf{[@]}~\textsc{Intuition}},#1}

\newtcolorbox{translation}[1][]{v13tint=thesisIntuition,
  title={\textbf{[T]}~\textsc{Reader's Translation}},#1}

\newtcolorbox{runningexample}[1][]{v13tint=thesisPositive,
  title={\textbf{[R]}~\textsc{Running Example: The Compliance Assistant}},#1}

\newtcolorbox{decision}[1][]{v13base=thesisPositive,
  title={\checkmark~\textsc{Decision Rule}},#1}

\newtcolorbox{aside}[1][]{v13tint=thesisAside,
  title={\textbf{[B]}~\textsc{Aside}},#1}

\newtcolorbox{openproblem}[1][]{v13filled=thesisCritical,
  title={\textbf{[?]}~Open Problem},#1}

\newtcolorbox{centralclaim}[1][]{v13filled=thesisAccent,
  title={\textbf{[@]}~Central Claim},#1}

\newtcolorbox{takeaway}[1][]{v13filled=thesisPositive,
  title={\textbf{[K]}~Key Takeaway},#1}

\AtBeginDocument{%
  \hypersetup{
    linkcolor=thesisAccent,
    citecolor=thesisAside,
    urlcolor=thesisPositive,
    filecolor=thesisPositive
  }%
}

\theoremstyle{plain}
\newtheorem{theorem}{Theorem}[chapter]
\newtheorem{lemma}[theorem]{Lemma}
\newtheorem{proposition}[theorem]{Proposition}
\newtheorem{corollary}[theorem]{Corollary}
\newtheorem{conjecture}[theorem]{Conjecture}

\theoremstyle{definition}
\newtheorem{definition}[theorem]{Definition}
\newtheorem{assumption}[theorem]{Assumption}

\newtheorem{hypothesis}[theorem]{Hypothesis}

\theoremstyle{remark}
\newtheorem{remark}[theorem]{Remark}

\newtcolorbox{applicationbox}[1]{
    colback=stackL2!5, colframe=stackL2!40,
    title={\small\sffamily #1}, fonttitle=\bfseries, coltitle=black,
    boxrule=0.4pt, arc=2pt,
    left=6pt, right=6pt, top=4pt, bottom=4pt,
    before skip=12pt, after skip=12pt, breakable,
}

\newtcolorbox{thesisstatementbox}{
    colback=stackL1!4, colframe=stackL1!70,
    title={\normalsize\sffamily\bfseries Central Claim}, coltitle=white,
    boxrule=1.0pt, arc=3pt,
    left=10pt, right=10pt, top=8pt, bottom=8pt,
    before skip=16pt, after skip=16pt, breakable,
    enhanced, drop shadow,
}

\newcommand{\R}{\mathbb{R}}
\newcommand{\N}{\mathbb{N}}
\newcommand{\Z}{\mathbb{Z}}
\newcommand{\E}{\mathbb{E}}
\newcommand{\Prob}{\mathbb{P}}
\newcommand{\Var}{\operatorname{Var}}
\newcommand{\Cov}{\operatorname{Cov}}
\newcommand{\KL}{\operatorname{KL}}
\newcommand{\poly}{\operatorname{poly}}
\newcommand{\polylog}{\operatorname{polylog}}
\DeclareMathOperator*{\argmin}{arg\,min}
\DeclareMathOperator*{\argmax}{arg\,max}
\DeclareMathOperator{\softmax}{softmax}
\DeclareMathOperator{\relu}{ReLU}
\DeclareMathOperator{\gelu}{GELU}
\DeclareMathOperator{\sign}{sign}
\DeclareMathOperator{\rank}{rank}
\DeclareMathOperator{\tr}{tr}
\DeclareMathOperator{\diag}{diag}
\DeclareMathOperator{\essinf}{ess\,inf}
\newcommand{\bigO}{\mathcal{O}}
\newcommand{\bigOtilde}{\widetilde{\mathcal{O}}}

\newcommand{\vocab}{\Sigma}
\newcommand{\dmodel}{d}
\newcommand{\nheads}{H}
\newcommand{\nlayers}{L}
\newcommand{\attn}{\operatorname{Attn}}
\newcommand{\ffn}{\operatorname{FFN}}

\newcommand{\loss}{\ell}
\newcommand{\risk}{\mathcal{R}}
\newcommand{\emprisk}{\widehat{\mathcal{R}}}
\newcommand{\hypclass}{\mathcal{H}}

\newcommand{\Ntrain}{N}

\newcommand{\corpus}{\mathcal{C}}
\newcommand{\kg}{\mathcal{G}}

\newcommand{\agents}{\mathcal{N}}
\newcommand{\utility}{u}
\newcommand{\mechanism}{\mathcal{M}}

\newcommand{\secparam}{\kappa}
\newcommand{\zkprover}{\mathcal{P}}
\newcommand{\zkverifier}{\mathcal{V}}

\newcommand{\Lone}{\textsc{L1}\xspace}

\newcommand{\Lfive}{\textsc{L5}\xspace}

\newcommand{\cA}{\mathcal{A}}
\newcommand{\cC}{\mathcal{C}}

\newcommand{\cG}{\mathcal{G}}

\newcommand{\cM}{\mathcal{M}}

\newcommand{\cS}{\mathcal{S}}

\newcommand{\cV}{\mathcal{V}}
\newcommand{\cX}{\mathcal{X}}
\newcommand{\cY}{\mathcal{Y}}

\newcommand{\indicator}{\mathbf{1}}
\DeclareMathOperator{\Ent}{H}            
\newcommand{\TV}{\mathrm{TV}}            

\providecommand{\cN}{\mathcal{N}}
\providecommand{\cU}{\mathcal{U}}

\providecommand{\adapt}{\Lambda}             

\providecommand{\cP}{\mathcal{P}}

\newcommand{\EvoPref}{\textsc{EvoPref}\xspace}
\newcommand{\QDLLM}{\textsc{QD-LLM}\xspace}
\newcommand{\euler}{\mathrm{e}}              
\newcommand{\OneMax}{\textsc{OneMax}}
\newcommand{\LeadingOnes}{\textsc{LeadingOnes}}
\newcommand{\Jump}{\textsc{Jump}}
\newcommand{\BehOM}{\textsc{BehavioralOneMax}}
\newcommand{\OTRoute}{\textsc{OT-Route}\xspace}

\newcommand{\Meff}{M_{\mathrm{eff}}}
\newcommand{\JSD}{\mathrm{JSD}}

\newcommand{\AF}{\mathrm{AF}}
\newcommand{\SETAF}{\mathrm{SETAF}}
\newcommand{\BAF}{\mathrm{BAF}}
\newcommand{\ADF}{\mathrm{ADF}}
\newcommand{\ALC}{\mathcal{ALC}}
\newcommand{\ALCO}{\mathcal{ALCO}}

\thesistitle{The Deterministic Horizon: Impossibility Results as Design Specifications for Trustworthy AI Systems}
\supervisor{Prof. Siu Ming \textsc{Yiu}}
\cosupervisor{Dr. Kam Pui \textsc{Chow}}
\examiner{}
\degree{Doctor of Philosophy}
\author{Dongxin \textsc{Guo}}
\addresses{}
\subject{Artificial Intelligence}
\university{University of Hong Kong}
\bsuniversity{HKU}
\msuniversity{HKU}
\department{Department of Computer Science}
\group{AI Research Laboratory}
\faculty{Faculty of Engineering}

\AtBeginDocument{
\hypersetup{pdftitle=\ttitle}
\hypersetup{pdfauthor=\authorname}
\hypersetup{pdfsubject=\subjectname}
\hypersetup{pdfkeywords=\keywordnames}
}
\begin{document}

\frontmatter
\pagestyle{plain}

\begin{titlepage}
	\addtocounter{page}{-1}
	\begin{center}
		
		\begin{tabular}{cc}
			\includegraphics[align=c, width=0.18\textwidth]{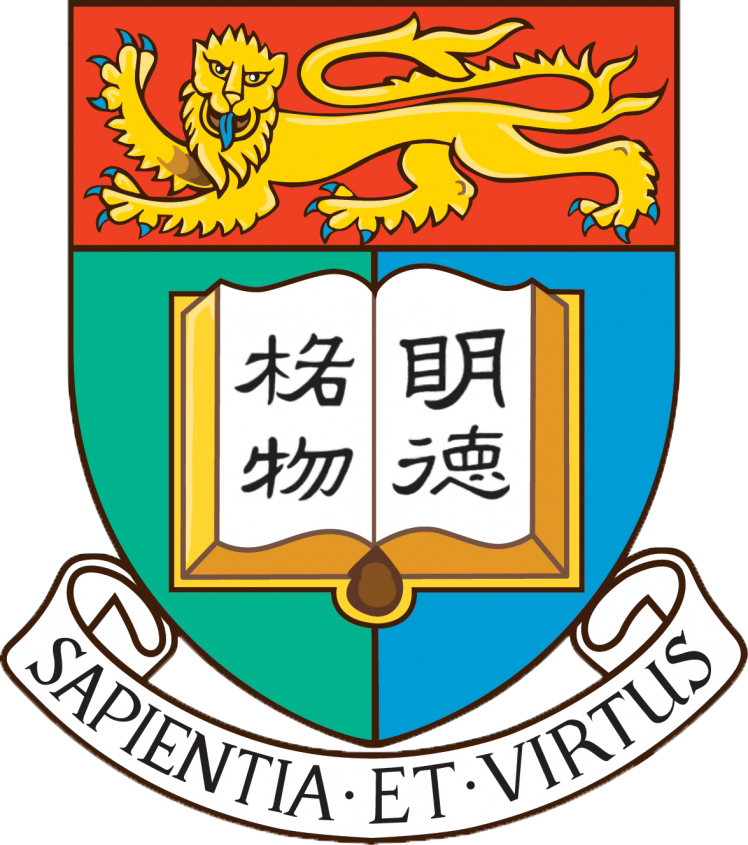} &  
			{\scshape \huge \darkred{\univname}} 
		\end{tabular}
		
		\vspace{0.5cm}
		\textsc{\Large Doctoral Thesis}\\[0.5cm] 

		\rule[0.4cm]{13cm}{0.1pt}\\
		{\huge \bfseries \ttitle\par}\vspace{0.4cm} 
		\rule{13cm}{0.1pt}\\ \vspace{1.5cm}
		
		\begin{minipage}[t]{0.4\textwidth}
			\begin{flushleft} \large
				\emph{Author:}\\
				\darkred{\authorname} 
			\end{flushleft}
		\end{minipage}
		\begin{minipage}[t]{0.4\textwidth}
			\begin{flushright} \large
				\emph{Supervisor:} \\
				\supname \\ 
				\emph{Co-Supervisor:} \\
				\cosupname 
			\end{flushright}
		\end{minipage}\\[1.6cm]
		
		\vfill
		
		\large \textit{A thesis submitted in fulfillment of the requirements\\ for the degree of \degreename}\\[0.3cm] 
		\textit{in the}\\[0.4cm]
		\deptname\\\facname\\[1.6cm] 
		
		\vfill
		
		{\large \usdate\today}\\[4cm] 

		\vfill
	\end{center}
	
\end{titlepage}

\blankpage
\addtocounter{page}{-1}

\begin{abstract}
\addchaptertocentry{\abstractname}

Large language models now write software, draft legal documents, and produce clinical notes. The dominant view holds that their reliability at extended reasoning improves with scale and training. A counter-tradition, from Turing and Arrow through the No Free Lunch theorems, holds that fundamental limits shape what computation can do. Recent work has begun to extend this canon to artificial intelligence, with impossibility results for calibration and fairness. What the emerging canon has lacked is a unifying framework that turns such results from curiosities into design rules.

Here we supply that framework. \textbf{We prove an accuracy ceiling set by architecture alone: past a critical reasoning depth, no amount of training moves it.} This impossibility holds at any adapter rank, any sample size, and any loss function. Computable before deployment from layer count and embedding width alone, the ceiling is measured at values between nineteen and thirty-one across twelve transformer architectures, with cross-model correlations between the low eighties and low nineties. Fine-tuning on optimal-length traces recovers less than four percentage points, ruling out the leading alternative account.

The mechanism is a capacity invariant of the residual stream: each reasoning step moves only as much information as the depth-times-width geometry allows, and that budget is fixed by architecture rather than by the trained weights. An information-theoretic conversion yields super-exponential accuracy decay past the horizon, which is why frontier reasoning models like o3 and DeepSeek-R1 are outperformed by tool-augmented systems on agentic software benchmarks such as SWE-bench Pro, at roughly one-third the per-task cost. We call this computable threshold the Deterministic Horizon. An unconditional circuit-complexity lower bound against polynomial-size constant-depth prime-modulus circuits for modular exponentiation complements the architectural result, marking the strongest unconditional progress to date toward the conjecture that softmax transformers cannot simulate modular arithmetic.

The argument recasts across subfields whose mathematics share almost nothing. Preference learning under any misspecified preference model jumps discontinuously in sample complexity. Retrieval pipelines with more than one stage cannot be diagnosed by any single score: at least as many independent metrics as stages are mathematically required, a result that formalises for machine learning the psychometric norm that multi-dimensional constructs need multi-dimensional measurement. Standard truthful auctions fail for language-model agents with prompt-dependent valuations, and zero-knowledge verification of neural inference pays a measured overhead between one hundred ten and one hundred ninety times per non-linear activation, matching a proved theoretical lower bound up to constants. Together with eleven reliability-toolkit extensions, these four flagship siblings of the Deterministic Horizon form a catalogue of sixteen specifications, each taking the same form: a computable boundary, a quantified violation cost, and a constructive design rule. Two compositions across these specifications are proved, one joining computation and grounding and one within the trust pillar; one of the six cross-pillar pairings is reported as an honest obstruction; the remaining four are open.

This thesis offers the impossibility-specification methodology as one candidate for the generative research programme that trustworthy artificial intelligence may need. Every fundamental limit of AI is also a design rule.

\end{abstract}

\pagestyle{empty}
\newpage
\addtocounter{page}{-1}
\begin{center}
	\vspace*{2cm}
	\huge{ \bf \ttitle}
\end{center}

\vspace{20mm}
\begin{center}
	by
	
	\vspace{10mm}
	{\bf \authorname}\\
	M.S. \textit{\msunivname}
\end{center}

\vspace{30mm}
\begin{center}
	A Thesis Submitted in Partial Fulfilment \\
	of the Requirements for the Degree of \\
	Doctor of Philosophy \\
	\vspace{10mm}
	at \\
	\vspace{10mm}
	\univname\\
	\monthyeardate\today
\end{center}
\begin{acknowledgements}
\setcounter{page}{2}
\addchaptertocentry{\acknowledgementname} 
\vspace{1cm}

\noindent Nine years. When I started this PhD in 2017, fresh from my MSc in
Computer Science (Financial Computing Stream), I did not imagine the road would run this
long, or carry me through this many lives at once. I owe this thesis to the
people who walked it with me.

My deepest thanks go to my supervisor, Prof. Siu Ming Yiu. The story begins
before the PhD itself. During my MSc, I worked on course projects under his
guidance, hoping to earn his trust, and somehow I did. He took me on as a PhD
student when I was eager for everything and certain about nothing. The first
two years were honest about that gap: I was buried in projects, excited by
every direction, and made very little visible progress. Prof. Yiu was patient
when I had no right to expect patience.

Then he asked if I wanted to help launch Brain Investing, a fintech venture in
quantitative trading spinning out of HKU. I said yes without hesitation, and
that yes changed the shape of my research. The years that followed taught me
how a theoretical claim survives contact with a live market: how an
impossibility result, properly read, is a design specification; how a bound
becomes a risk cap; how a proof becomes a guardrail. Brain Investing grew with
generous support from HKU, Cyberport, the Hong Kong Science and Technology
Parks, and the HKSAR Government's Innovation and Technology Fund, and I am
grateful to each of those institutions for backing translational work when it
was still a small thing.

I thought my work was settled. Then, in late 2022, ChatGPT and Stable Diffusion
arrived and rearranged the world overnight. Prof. Yiu, I and the team spent three
months talking about what to do, and we decided to start something new. That
became Stellaris AI. Within months we had released the first
hundreds-of-billions-parameter model in Hong Kong, and the years since have
been a continuous loop between proof and deployment, between the theorems I
write at my desk and the systems we ship to clients. Prof. Yiu has been my
advisor, my collaborator, my director at both companies, and the person who
taught me that a research life can be lived honestly in two registers at once:
the long-horizon proof and the next product release. He gave me both an
academic home and a working one. I cannot overstate how rare that is, and how
lucky I have been.

My thanks also go to my co-supervisor, Dr. Kam Pui Chow, for serving on my
supervisory committee and lending his time to this work.

To my colleagues and fellow students at HKU's Department of Computer Science:
thank you for the corridor conversations, the paper-deadline solidarity, and
the seminars that quietly bent the direction of my thinking. To my colleagues
at Brain Investing and Stellaris AI: thank you for treating research as a craft
and not as decoration, for catching the bugs my proofs assumed away, and for
building the infrastructure that lets a theorem ship. The phrase ``theorems
become SLOs'' on my homepage is not a slogan. It is what we do together every
week, and it is yours as much as mine.

Some debts run deeper than any institution. To my parents: thank you for
raising me with the patience of people who knew I would leave, and the warmth
of people who never made leaving feel like loss. You opened the door to a wider
world without asking me to choose between it and home, and that is the most
generous gift a child can be given. Wherever I am, whenever it is, your home is
my warm place. To my husband: thank you for your love, your steadiness, and
your all-weather support across the years when I had little to give back. With
you beside me, no problem has felt unsolvable and no project has felt too
large. Whatever I have accomplished here is, quietly and unmistakably, ours. To
my son: thank you for the warmth of your love and the bright pull of your
curiosity. You ask the kind of questions a researcher spends a career trying to
relearn how to ask. You remind me, every day, why any of this is worth doing.

A PhD is supposed to be a solo accomplishment. Mine was not. It was a nine-year
conversation with people who believed in me before I had earned belief, and who
kept believing while I figured out what kind of researcher I wanted to be. This
thesis carries their fingerprints on every page.
\\[0.4cm]

\begin{flushright}
\authorname \\
Montr\'{e}al, Canada \\
\usdate\today
\end{flushright}

\end{acknowledgements}

\tableofcontents
\listoffigures
\listoftables
\listofalgorithms
\addchaptertocentry{\listalgorithmcfname}

\begin{abbreviations}{ll}

\textbf{AI} & Artificial Intelligence \\
\textbf{CoT} & Chain-of-Thought \\
\textbf{CLC} & Composition-Length Compatibility \\
\textbf{DPO} & Direct Preference Optimisation \\
\textbf{DSA} & Digital Services Act \\
\textbf{EF} & Ehrenfeucht-Fra\"iss\'e \\
\textbf{EMO} & Evolutionary Multi-Objective Optimisation \\
\textbf{FOC} & First-Order Logic with Counting \\
\textbf{IOP} & Interactive Oracle Proof \\
\textbf{KG} & Knowledge Graph \\
\textbf{KR} & Knowledge Representation \\
\textbf{LLM} & Large Language Model \\
\textbf{LoRA} & Low-Rank Adaptation \\
\textbf{MCTS} & Monte Carlo Tree Search \\
\textbf{MoE} & Mixture of Experts \\
\textbf{MSUD} & Multi-Source Uncertainty Decomposition \\
\textbf{NP} & Nondeterministic Polynomial Time \\
\textbf{OSP} & Obviously Strategy-Proof \\
\textbf{OT} & Optimal Transport \\
\textbf{PAC} & Probably Approximately Correct \\
\textbf{POMDP} & Partially Observable Markov Decision Process \\
\textbf{PRM} & Process Reward Model \\
\textbf{QD} & Quality-Diversity \\
\textbf{RAG} & Retrieval-Augmented Generation \\
\textbf{RLHF} & Reinforcement Learning from Human Feedback \\
\textbf{SAD} & Stratified Ackermann Decomposability \\
\textbf{SDPI} & Strong Data Processing Inequality \\
\textbf{SLO} & Service Level Objective \\
\textbf{SMD} & Strategic Manipulation Dimension \\
\textbf{SQ} & Statistical Query \\
\textbf{SPRT} & Sequential Probability Ratio Test \\
\textbf{TC$^0$} & Threshold Circuit Complexity Class \\
\textbf{VCG} & Vickrey-Clarke-Groves \\
\textbf{ZK} & Zero-Knowledge \\

\end{abbreviations}


\chapter*{Notation Conventions}
\addcontentsline{toc}{chapter}{Notation Conventions}
\markboth{Notation Conventions}{}

This thesis spans four subfields. Where terminology or symbol conventions
differ across subfields, each chapter prefaces first use with a brief
reminder. The complete notation glossary, including per-chapter variations
and overloaded symbols, is provided in Appendix~\ref{app:notation}.

\vspace{0.5em}
\noindent
\begin{tabularx}{\textwidth}{@{}p{2.5cm}X@{}}
\toprule
\textbf{Symbol} & \textbf{Meaning (thesis-wide default)} \\
\midrule
$L$                    & Number of transformer layers \\
$d$                    & Embedding / residual-stream dimension \\
$n$                    & Sample size or sequence length (context-dependent) \\
$d^{\ast}$             & Deterministic Horizon (critical reasoning depth) \\
$\delta$               & Test-time reasoning depth (chain-of-thought steps) \\
$\varepsilon$          & Per-step error rate; incentive-compatibility slack \\
$\gamma$               & Misspecification level in preference learning \\
$\kappa$               & Cryptographic security parameter; condition number \\
$\rho$                 & Real-data fraction in synthetic-data training \\
$\Theta, O, \Omega$    & Bachmann–Landau asymptotic notation \\
$\tilde{O}$            & $O$ suppressing polylogarithmic factors \\
$\mathcal{S}_i$        & Impossibility Specification $i$ (Chapter~\ref{ch:synthesis} catalog) \\
$\mathcal{S}_i \odot \mathcal{S}_j$ & Cross-domain composition of two specifications \\
$L_1$--$L_5$           & The five layers of the Trustworthy AI Stack \\
$g_i(\theta_i)$        & Layer-$i$ guarantee function \\
\bottomrule
\end{tabularx}

\vspace{0.5em}
\noindent\textbf{General conventions.}
Vectors are lowercase bold ($\mathbf{x}$); matrices are uppercase
($W, \Sigma$). Probability measures use $\mathbb{P}$ or $\Pr$; expectations
use $\mathbb{E}$. Divergences: $\mathrm{KL}(P \| Q)$ is Kullback–Leibler;
$\mathrm{TV}(P, Q)$ is total variation. Computational complexity classes
are written in upright sans-serif ($\mathrm{TC}^0$, $\mathrm{NC}^1$,
$\mathsf{AC}^0[p]$). Cryptographic primitives (zero-knowledge proofs,
interactive oracle proofs, pseudorandom correlation generators) follow
the notational conventions of the cited cryptography literature and
are defined in-chapter at first use; Appendix~\ref{app:notation}
Section~7 gives the cross-chapter summary.

\clearpage

\chapter*{Intuitive Glossary}
\addcontentsline{toc}{chapter}{Intuitive Glossary}
\markboth{Intuitive Glossary}{}

\noindent\small
Five cross-subfield terms recurring in every chapter. Standard AI/ML
terminology is not glossed here; see Appendix~\ref{app:notation}.

\begin{description}[leftmargin=1.8em,itemsep=2pt,parsep=1pt]
\item[Impossibility Specification] Formal limit with three structural
properties: computable boundary condition, quantified violation cost,
constructive engineering rule on the wrong side. Sixteen catalogued in
Chapter~\ref{ch:synthesis}.

\item[FOC{[}Attn{]}] First-order logic with counting + attention
quantifiers. Exactly captures bounded-depth softmax transformers
(Thm.~\ref{thm:equivalence}).

\item[Obviously Strategy-Proof (OSP)] Mechanism-design property stronger
than strategy-proofness: truthful behaviour obviously optimal for agents
with bounded lookahead. Achieves $\varepsilon \le 0.16$ incentive
compatibility for LLM agents where VCG fails.

\item[Welfare Composition] Joint guarantee from composing mechanism
design and cryptographic verification; welfare loss $O(\varepsilon + e^{-\kappa})$
(Thm.~\ref{thm:welfare-composition}), exponentially better than either
pillar alone.

\item[Composition (cross-domain)] Two specifications from different
subfields yielding joint reliability strictly stronger than their
conjunction. Two proved (\S\ref{sec:composition-l1-l2};
\S\ref{sec:welfare-composition}); one honest obstruction reported
(\S\ref{sec:adaptation-grounding-obstruction}).
\end{description}

\clearpage

\chapter*{Impact Summary}
\addcontentsline{toc}{chapter}{Impact Summary}
\markboth{Impact Summary}{}

\noindent\small

\paragraph{In one sentence.} The thesis proves that the depth at which
language-model reasoning fails (the Deterministic Horizon,
$d^{\ast} \in [19,31]$) is a design specification rather than an
obstacle, and develops a methodology converting this and fifteen
analogous AI-system limits into computable engineering rules with
quantified violation costs.

\paragraph{Theoretical.} Five flagship specifications: Fine-Tuning
Impossibility (Thm.~\ref{thm:finetuning-impossibility}); preference-learning
phase transition at any Bradley-Terry misspecification
(Thm.~\ref{thm:pref_transition}); Formal Measurement-Validity
Impossibility for RAG (Thm.~\ref{thm:attribution-impossibility}); joint
VCG$\to$OSP $+$ $147\times$ non-linearity-tax specification for
multi-agent LLMs; Welfare Composition Theorem
(Thm.~\ref{thm:welfare-composition}) proving mechanism design and
cryptographic verification jointly necessary. Eleven further
instantiations; two cross-domain compositions stronger than conjunction;
one honest obstruction delineating where the methodology currently fails.

\paragraph{Empirical.} 69 peer-reviewed publications during candidature.
Cross-model Horizon validation at $r = 0.81$--$0.91$ across 12
architectures; EvoPref reduces preference collapse 47\% vs.\ DPO;
TrustKGRAG drops attack success $92.3\% \to 8.7\%$; TrajTest detects
$2.3\times$ more agentic-workflow failures than outcome-only evaluation.

\paragraph{Intellectual.} Provides the field a unified template for
converting negative results into engineering specifications.

\clearpage


\mainmatter
\pagestyle{thesis}


\chapter{Introduction: Why Every Limit Is a Specification}
\label{ch:introduction}


\section{The Deterministic Horizon: A Concrete Impossibility}
\label{sec:horizon-flagship}

\noindent\emph{The depth at which decoder-only transformer reasoning collapses is an architectural invariant under the sparse-task hypothesis (measurement: $d^*\in[19,31]$ across twelve architectures); crossing $d^*$ calls for tool delegation, not additional fine-tuning.}

No fine-tuning procedure that preserves the decoder-only transformer architecture (at any rank, sample size, or loss form) can recover more than $O(d^{\ast}/\delta)$ of the accuracy deficit at test depth $\delta$ beyond a critical reasoning depth $d^{\ast}$, under a sparse-task hypothesis on the induced reasoning-trace distribution (the precise conditioning, $\mathrm{hyp}\text{:}\mathrm{sparse}\text{-}\mathrm{task}$, is stated in \cref{ch:horizon}; the running benchmarks of this chapter are in its scope). This training-invariant envelope (\cref{thm:finetuning-impossibility}) is the thesis's flagship result. It is architectural: the per-step information throughput is a function of the residual-stream capacity, which is an invariant of the depth-width geometry rather than a property of the trained weights. Once per-step throughput is bounded, the super-exponential accuracy decay beyond $d^{\ast}$ follows through an elementary information-theoretic conversion (proof in \cref{app:finetuning-impossibility}). Empirical calibration is consistent with the envelope: fine-tuning Llama-3.3-8B on $5{,}000$ optimal-length traces recovers only $3.2$ percentage points of the deficit at $d = 2d^*$, an order of magnitude below the $\geq 30\%$ recovery the leading alternative explanation (``Simplicity Bias''~\cite{wu2025simplicity}) predicts and quantitatively inside the architectural envelope.

The depth $d^{\ast}$ itself, the \emph{Deterministic Horizon}, is measured at $d^{\ast} \in [19, 31]$ (95\% prediction interval) across twelve architectures spanning GPT-2, Llama-2/3, Gemma-2, Qwen-2.5, Mistral, Phi-2, and OLMo, with cross-model correlation $r = 0.81$--$0.91$ (\cref{cor:horizon-measurement}). The scaling ansatz $d^{\ast} = O(L \cdot \phi(d))$ with $\phi(d) \in [\sqrt{\log d}, \log d]$ is established in \cref{thm:horizon-scaling} as an architectural upper bound, with empirical fit $d^{\ast} \approx \hat{c} \log L \cdot \sqrt{\log d}$ and $\hat{c} = 2.74$ a regression fit on the $12$-architecture, $3$-task evaluation set of \cref{tab:horizon}, reported as a calibration of the impossibility rather than as its content. Beyond $d^{\ast}$, accuracy decays super-exponentially: on permutation puzzles solvable by breadth-first search in under $0.1$~seconds, state-of-the-art reasoning models (o3, DeepSeek-R1, Claude-4.5-Opus) fail after minutes of deliberation, and on software-engineering benchmarks requiring multi-file state tracking (SWE-Bench-State, $n{=}300$, a multi-file-state variant derived from SWE-bench~\cite{jimenez2024swebench})) they reach $24$--$37\%$ accuracy while external-tool-augmented counterparts reach $86$--$94\%$ at roughly one-third the cost per resolved task. The Fine-Tuning Impossibility converts this depth phenomenon into an unambiguous engineering rule: delegate beyond $d^{\ast}$, because no training-time procedure will move the wall.

Framed as an architectural impossibility the result is sharp; framed as an engineering quantity it is actionable. Each of $L$, $d$, and the target success probability $\alpha$ is known to the system designer \emph{before} inference. The horizon is therefore computable in advance, and so is its implication: at depth $\delta \leq d^*$ use chain-of-thought; at $d^* < \delta \leq 2d^*$ use $k$-redundant verification with $k \geq 2$; at $\delta > 2d^*$ delegate to a symbolic planner or tool-augmented pipeline. The same impossibility that makes neural reasoning fail at depth $50$ \emph{specifies} the delegation threshold at depth $\sim 27$ for a 32-layer 4{,}096-width transformer, the $k^\star = 3$ verification budget for 15-hop regulatory reasoning at $\varepsilon = 0.05$, and the $\sim 5.5\times$ sample-efficiency advantage of process supervision when chain non-redundancy holds. An impossibility that forces a design decision is, in this sense, a \emph{specification}.

The thesis's claim is that this pattern is not unique to the Deterministic Horizon. Adaptation fails at a sharp $\gamma$-misspecification threshold beyond which preference-learning sample complexity jumps from $\Theta(n \log n/\Delta^2)$ to $\widetilde\Theta(n^2/\gamma^2)$; retrieval-augmented generation cannot be evaluated by any single metric once a pipeline has more than one stage (\emph{Formal Measurement-Validity Impossibility}, a topological codimension obstruction inspired by the convergent-discriminant-validation norm of Campbell \& Fiske~\cite{CampbellFiske1959} and Messick's construct-validity theory~\cite{Messick1989Validity}, and complementing the framework-level importation of measurement theory to ML of Jacobs \& Wallach~\cite{JacobsWallach2021Measurement}); multi-agent LLM coordination violates Vickrey--Clarke--Groves incentive compatibility and must instead use Obviously Strategy-Proof mechanisms with $\varepsilon \leq 0.16$ for bounded-lookahead reasoners; zero-knowledge verification of neural inference pays an IOP-optimal $147\times$ non-linearity tax for softmax layers (a matching upper and lower bound in the IOP model, \cref{ch:trust}), consistent with deployed-system measurements. Each is a fundamental limit; each is a computable boundary condition; each yields a constructive engineering rule on the wrong side of that boundary. We therefore introduce the \emph{impossibility-specification methodology} and develop \emph{five flagship specifications} (the Fine-Tuning Impossibility together with the Deterministic Horizon it calibrates, the $\gamma$-misspecification phase transition, the Formal Measurement-Validity Impossibility, the joint VCG$\to$OSP / non-linearity-tax specification, and the Welfare Composition), with eleven further instantiations across four subfields of trustworthy AI catalogued as domain applications and reliability-toolkit extensions, together with two cross-domain composition theorems demonstrating that specifications compose into joint reliability guarantees no single layer can provide.

\paragraph{Running Example: The Compliance Assistant.}
The Deterministic Horizon has immediate consequences for a concrete deployment target threaded throughout the thesis. A regulatory compliance assistant must answer multi-hop queries about financial regulation drawing on regulatory documents, interpreting provisions across jurisdictions, and coordinating with auditors and human stakeholders. A typical 12-hop regulatory reasoning chain lies within the chain-of-thought regime on mid-sized transformers (GPT-2 Medium, $d^* \approx 24$), on the Llama-2-7B-class primary running-example target of \cref{ch:horizon} (32 layers, 4,096 width; observed $\hat d^* \approx 27$ across task families in \cref{tab:horizon}, regression-predicted $d^*_{\mathrm{pred}} \approx 27.4$), and on frontier models (Llama-2 13B, $d^* \approx 30$), with a narrower depth buffer on the mid-sized case should per-hop variance or auxiliary reasoning steps push effective depth toward the boundary. At per-step error $\varepsilon \approx 0.03$ the unaided chain error probability is $\approx 31\%$ (unacceptable for a legal-compliance setting), so $k=2$ redundant verification is applied as a reliability measure (not as a Horizon-triggered rule switch), reducing chain error to $\approx 3.2\%$ under the i.i.d.\ theoretical bound of \cref{thm:k_redundant}, with the deployment-measured value $\approx 4.7\%$ surfacing the candidate-correlation gap to independence. The compliance-assistant case recurs as a running example in every technical chapter; it is not a deployment claim but a concreteness device, demonstrating how each impossibility specification would bind on a realistic system. Its deployment status is discussed honestly in \cref{ch:synthesis}: the system has been benchmarked on the HKU compliance corpus, not deployed to regulated production users.

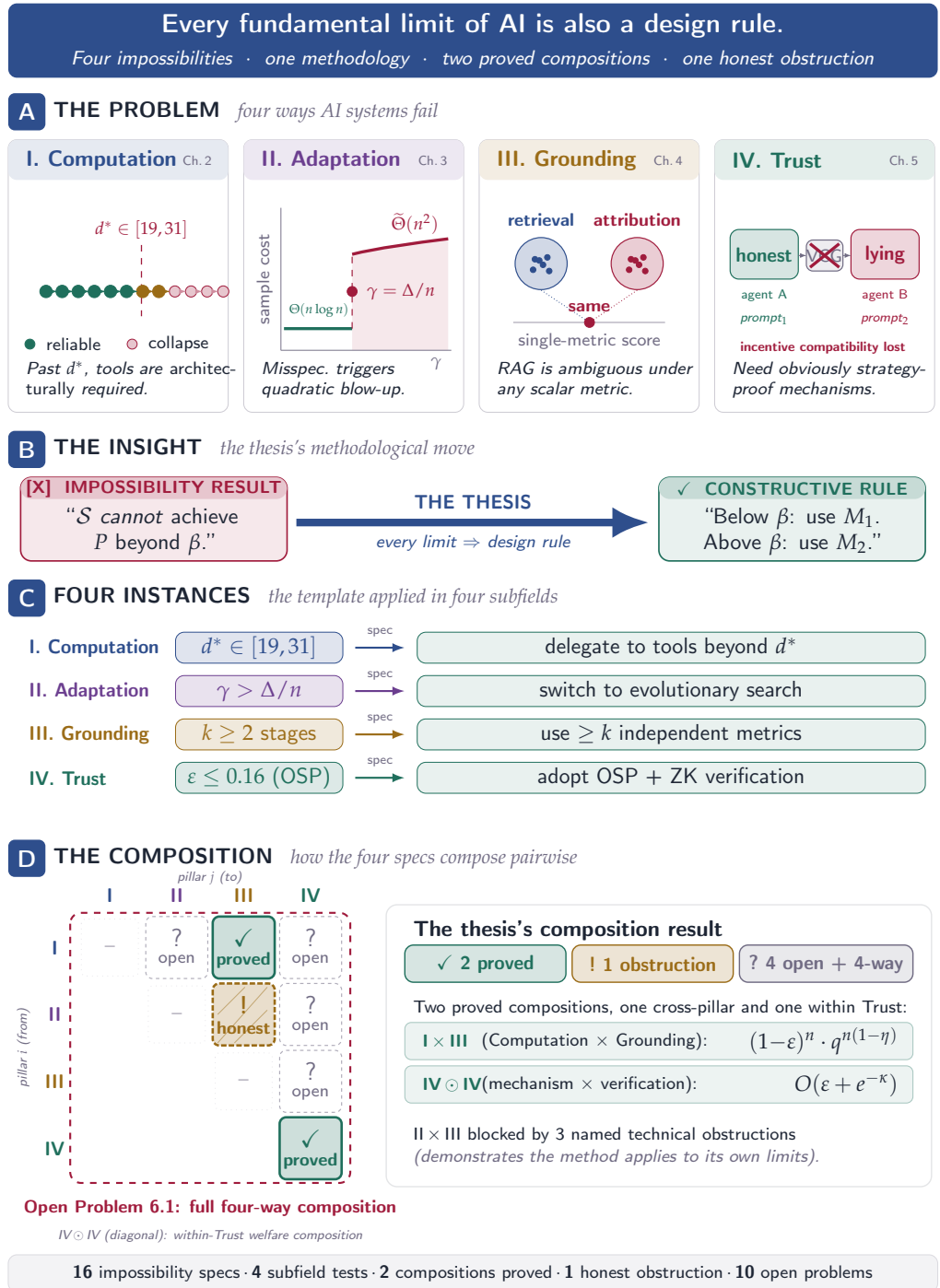
\begin{figure}[p]
	\centering
	\resizebox{\textwidth}{!}{%
		\begin{tikzpicture}[
			font=\sffamily,
			x=1cm, y=1cm,
			vframe/.style={
				draw=thesisAside!35, line width=0.4pt,
				fill=white, rounded corners=4pt,
				inner sep=0pt
			},
			bandtitle/.style={
				font=\sffamily\bfseries\small,
				text=thesisInk,
				inner xsep=0pt, inner ysep=0pt
			}
			]
			
			\draw[fill=thesisAccent, draw=none, rounded corners=4pt]
			(-7.5, 9.20) rectangle (7.5, 10.40);
			\node[text=white, font=\sffamily\bfseries\large]
			at (0, 10.05) {Every fundamental limit of AI is also a design rule.};
			\node[text=white!88, font=\sffamily\footnotesize\itshape]
			at (0, 9.50) {Four impossibilities \ $\cdot$ \ one methodology \ $\cdot$ \ two proved compositions \ $\cdot$ \ one honest obstruction};
			
			\filldraw[thesisAccent, draw=none, rounded corners=3pt]
			(-7.50, 8.35) rectangle (-6.90, 8.95);
			\node[text=white, font=\sffamily\bfseries\normalsize]
			at (-7.20, 8.65) {A};
			\node[bandtitle, anchor=west] at (-6.78, 8.65)
			{THE PROBLEM \textcolor{thesisAside}{\normalfont\itshape\footnotesize\ \ four ways AI systems fail}};
			
			
			\begin{scope}[shift={(-7.50, 3.80)}]
				\draw[vframe] (0, 0) rectangle (3.55, 4.4);
				\fill[thesisAccent!12, rounded corners=4pt] (0, 3.75) rectangle (3.55, 4.40);
				\node[anchor=west, font=\sffamily\bfseries\small, text=thesisAccent]
				at (0.15, 4.07) {I.\ Computation};
				\node[anchor=east, font=\sffamily\tiny, text=thesisAside]
				at (3.42, 4.07) {Ch.\,2};
				
				\node[font=\sffamily\scriptsize\bfseries, text=thesisCritical]
				at (2.15, 2.95) {$d^{\ast}\in[19,31]$};
				\draw[thesisCritical, dashed, line width=0.55pt] (2.15, 2.70) -- (2.15, 1.55);
				
				\foreach \i in {1,...,12} {
					\pgfmathsetmacro{\xx}{0.35 + 0.26*\i}
					\ifnum\i<7
					\filldraw[thesisPositive] (\xx, 1.95) circle (0.10);
					\else
					\ifnum\i<9
					\filldraw[thesisCaution] (\xx, 1.95) circle (0.10);
					\else
					\filldraw[thesisCritical!25, draw=thesisCritical, line width=0.4pt] (\xx, 1.95) circle (0.09);
					\fi
					\fi
				}
				\foreach \i in {1,...,11} {
					\pgfmathsetmacro{\xa}{0.45 + 0.26*\i}
					\pgfmathsetmacro{\xb}{0.53 + 0.26*\i}
					\ifnum\i<7
					\draw[->, thesisPositive, line width=0.35pt, -{Latex[length=1.1mm]}] (\xa, 1.95) -- (\xb, 1.95);
					\else
					\ifnum\i<9
					\draw[->, thesisCaution, line width=0.35pt, -{Latex[length=1.1mm]}] (\xa, 1.95) -- (\xb, 1.95);
					\else
					\draw[->, thesisCritical!50, line width=0.35pt, dashed, -{Latex[length=1.1mm]}] (\xa, 1.95) -- (\xb, 1.95);
					\fi
					\fi
				}
				
				\filldraw[thesisPositive] (0.35, 1.10) circle (0.08);
				\node[anchor=west, font=\sffamily\scriptsize, text=thesisInk]
				at (0.48, 1.10) {reliable};
				\filldraw[thesisCritical!25, draw=thesisCritical, line width=0.4pt] (2.00, 1.10) circle (0.08);
				\node[anchor=west, font=\sffamily\scriptsize, text=thesisInk]
				at (2.13, 1.10) {collapse};
				
				\node[anchor=south west, font=\sffamily\scriptsize\itshape,
				text=thesisInk, text width=3.35cm, align=left]
				at (0.15, 0.10) {Past $d^{\ast}$, tools are \emph{architecturally} required.};
			\end{scope}
			
			\begin{scope}[shift={(-3.70, 3.80)}]
				\draw[vframe] (0, 0) rectangle (3.55, 4.4);
				\fill[thesisIntuition!12, rounded corners=4pt] (0, 3.75) rectangle (3.55, 4.40);
				\node[anchor=west, font=\sffamily\bfseries\small, text=thesisIntuition]
				at (0.15, 4.07) {II.\ Adaptation};
				\node[anchor=east, font=\sffamily\tiny, text=thesisAside]
				at (3.42, 4.07) {Ch.\,3};
				
				\filldraw[thesisCritical!10, draw=none]
				(1.75, 1.00) -- (1.75, 2.55) .. controls (2.15, 2.65) and (2.75, 2.73) .. (3.30, 2.80)
				-- (3.30, 1.00) -- cycle;
				
				\draw[thesisAside!80, line width=0.5pt] (0.60, 1.00) -- (3.35, 1.00);  
				\draw[thesisAside!80, line width=0.5pt] (0.60, 1.00) -- (0.60, 3.30);  
				
				\node[font=\sffamily\scriptsize, text=thesisAside, rotate=90, anchor=south]
				at (0.6, 2.15) {sample cost};
				\node[font=\sffamily\scriptsize, text=thesisAside, anchor=east]
				at (3.33, 0.78) {$\gamma$};
				
				\draw[thesisPositive, line width=1.4pt] (0.65, 1.35) -- (1.75, 1.35);
				\draw[thesisCritical, dashed, line width=0.5pt] (1.75, 1.35) -- (1.75, 2.55);
				\draw[thesisCritical, line width=1.4pt]
				(1.75, 2.55) .. controls (2.15, 2.65) and (2.75, 2.73) .. (3.30, 2.80);
				
				\node[font=\sffamily\tiny\itshape, text=thesisPositive, anchor=south west]
				at (0.6, 1.40) {$\Theta(n\log n)$};
				\node[font=\sffamily\scriptsize\itshape, text=thesisCritical, anchor=east]
				at (3.28, 3.05) {$\widetilde\Theta(n^2)$};
				
				\filldraw[thesisCritical] (1.75, 1.95) circle (0.08);
				\node[anchor=west, font=\sffamily\scriptsize\bfseries, text=thesisCritical]
				at (1.85, 1.95) {$\gamma=\Delta/n$};
				
				\node[anchor=south west, font=\sffamily\scriptsize\itshape,
				text=thesisInk, text width=3.15cm, align=left]
				at (0.15, 0.10) {Misspec.\ triggers quadratic blow-up.};
			\end{scope}
			
			\begin{scope}[shift={(0.10, 3.80)}]
				\draw[vframe] (0, 0) rectangle (3.55, 4.4);
				\fill[thesisCaution!12, rounded corners=4pt] (0, 3.75) rectangle (3.55, 4.40);
				\node[anchor=west, font=\sffamily\bfseries\small, text=thesisCaution]
				at (0.15, 4.07) {III.\ Grounding};
				\node[anchor=east, font=\sffamily\tiny, text=thesisAside]
				at (3.42, 4.07) {Ch.\,4};
				
				\filldraw[thesisAccent!22, draw=thesisAccent, line width=0.5pt]
				(1.00, 2.40) circle (0.42);
				\foreach \p in {(0.87,2.30),(1.10,2.50),(0.90,2.47),(1.13,2.28),(1.03,2.38),(0.97,2.45)}
				\filldraw[thesisAccent] \p circle (0.045);
				\filldraw[thesisCritical!22, draw=thesisCritical, line width=0.5pt]
				(2.55, 2.40) circle (0.42);
				\foreach \p in {(2.42,2.30),(2.65,2.50),(2.45,2.47),(2.68,2.28),(2.58,2.38),(2.52,2.45)}
				\filldraw[thesisCritical] \p circle (0.045);
				
				\node[anchor=south, font=\sffamily\scriptsize\bfseries, text=thesisAccent]
				at (1.00, 2.88) {retrieval};
				\node[anchor=south, font=\sffamily\scriptsize\bfseries, text=thesisCritical]
				at (2.55, 2.88) {attribution};
				
				\draw[thesisAccent, densely dotted, line width=0.4pt] (1.00, 1.97) -- (1.78, 1.45);
				\draw[thesisCritical, densely dotted, line width=0.4pt] (2.55, 1.97) -- (1.78, 1.45);
				
				\draw[thesisAside!70, line width=0.4pt] (0.55, 1.45) -- (3.00, 1.45);
				\node[anchor=north, font=\sffamily\scriptsize, text=thesisAside]
				at (1.78, 1.42) {single-metric score};
				\filldraw[thesisCritical] (1.78, 1.45) circle (0.08);
				\node[anchor=south, font=\sffamily\scriptsize\bfseries, text=thesisCritical]
				at (1.78, 1.50) {same};
				
				\node[anchor=south west, font=\sffamily\scriptsize\itshape,
				text=thesisInk, text width=3.35cm, align=left]
				at (0.15, 0.10) {RAG is ambiguous under any scalar metric.};
			\end{scope}
			
			\begin{scope}[shift={(3.90, 3.80)}]
				\draw[vframe] (0, 0) rectangle (3.55, 4.4);
				\fill[thesisPositive!12, rounded corners=4pt] (0, 3.75) rectangle (3.55, 4.40);
				\node[anchor=west, font=\sffamily\bfseries\small, text=thesisPositive]
				at (0.15, 4.07) {IV.\ Trust};
				\node[anchor=east, font=\sffamily\tiny, text=thesisAside]
				at (3.42, 4.07) {Ch.\,5};
				
				\filldraw[thesisPositive!18, draw=thesisPositive, rounded corners=4pt, line width=0.5pt]
				(0.25, 2.15) rectangle (1.35, 2.95);
				\node[font=\sffamily\footnotesize\bfseries, text=thesisPositive]
				at (0.80, 2.55) {honest};
				\node[anchor=north, font=\sffamily\tiny, text=thesisPositive]
				at (0.80, 2.12) {agent A};
				
				\filldraw[thesisCritical!18, draw=thesisCritical, rounded corners=4pt, line width=0.5pt]
				(2.20, 2.15) rectangle (3.30, 2.95);
				\node[font=\sffamily\footnotesize\bfseries, text=thesisCritical]
				at (2.75, 2.55) {lying};
				\node[anchor=north, font=\sffamily\tiny, text=thesisCritical]
				at (2.75, 2.12) {agent B};
				
				\filldraw[thesisAside!12, draw=thesisAside, rounded corners=3pt, line width=0.5pt]
				(1.48, 2.28) rectangle (2.07, 2.78);
				\node[font=\sffamily\scriptsize, text=thesisAside]
				at (1.77, 2.53) {VCG};
				\draw[thesisCritical, line width=1.4pt] (1.54, 2.34) -- (2.01, 2.72);
				\draw[thesisCritical, line width=1.4pt] (1.54, 2.72) -- (2.01, 2.34);
				
				\draw[->, thesisAside, line width=0.35pt, -{Latex[length=1mm]}]
				(1.35, 2.53) -- (1.48, 2.53);
				\draw[->, thesisAside, line width=0.35pt, -{Latex[length=1mm]}]
				(2.07, 2.53) -- (2.20, 2.53);
				
				\node[anchor=north, font=\sffamily\tiny\itshape, text=thesisPositive]
				at (0.80, 1.72) {prompt$_1$};
				\node[anchor=north, font=\sffamily\tiny\itshape, text=thesisCritical]
				at (2.75, 1.72) {prompt$_2$};
				
				\node[font=\sffamily\tiny\bfseries, text=thesisCritical]
				at (1.75, 1.05) {incentive compatibility lost};
				
				\node[anchor=south west, font=\sffamily\scriptsize\itshape,
				text=thesisInk, text width=3.25cm, align=left]
				at (0.15, 0.10) {Need obviously strategy-proof mechanisms.};
			\end{scope}
			
			\filldraw[thesisAccent, draw=none, rounded corners=3pt]
			(-7.50, 2.90) rectangle (-6.90, 3.50);
			\node[text=white, font=\sffamily\bfseries\normalsize]
			at (-7.20, 3.20) {B};
			\node[bandtitle, anchor=west] at (-6.78, 3.20)
			{THE INSIGHT \textcolor{thesisAside}{\normalfont\itshape\footnotesize\ \ the thesis's methodological move}};
			
			\filldraw[thesisCritical!10, draw=thesisCritical, rounded corners=5pt, line width=0.7pt]
			(-7.30, 1.40) rectangle (-3.00, 2.75);
			\fill[thesisCritical!25, rounded corners=5pt] (-7.30, 2.40) rectangle (-3.00, 2.75);
			\node[anchor=center, font=\sffamily\bfseries\scriptsize, text=thesisCritical]
			at (-5.15, 2.57) {\textbf{[X]}\ \ IMPOSSIBILITY RESULT};
			\node[anchor=center, font=\sffamily\small, text=thesisInk, align=center, text width=4.1cm]
			at (-5.15, 1.88) {``$\mathcal{S}$ \emph{cannot} achieve $P$ beyond $\beta$.''};
			
			\draw[-{Latex[length=6mm,width=5mm]}, line width=3.5pt, thesisAccent]
			(-2.85, 2.02) -- (2.85, 2.02);
			\node[anchor=south, font=\sffamily\bfseries\footnotesize, text=thesisAccent, yshift=0.5pt]
			at (0, 2.10) {THE THESIS};
			\node[anchor=north, font=\sffamily\scriptsize\itshape, text=thesisAccent]
			at (0, 1.95) {every limit $\Rightarrow$ design rule};
			
			\filldraw[thesisPositive!10, draw=thesisPositive, rounded corners=5pt, line width=0.7pt]
			(3.00, 1.40) rectangle (7.30, 2.75);
			\fill[thesisPositive!25, rounded corners=5pt] (3.00, 2.40) rectangle (7.30, 2.75);
			\node[anchor=center, font=\sffamily\bfseries\scriptsize, text=thesisPositive]
			at (5.15, 2.57) {\checkmark\ \ CONSTRUCTIVE RULE};
			\node[anchor=center, font=\sffamily\small, text=thesisInk, align=center, text width=4.1cm]
			at (5.15, 1.88) {``Below $\beta$: use $M_1$. Above $\beta$: use $M_2$.''};
			
			\filldraw[thesisAccent, draw=none, rounded corners=3pt]
			(-7.50, 0.50) rectangle (-6.90, 1.10);
			\node[text=white, font=\sffamily\bfseries\normalsize]
			at (-7.20, 0.80) {C};
			\node[bandtitle, anchor=west] at (-6.78, 0.80)
			{FOUR INSTANCES \textcolor{thesisAside}{\normalfont\itshape\footnotesize\ \ the template applied in four subfields}};
			
			
			\node[anchor=west, font=\sffamily\bfseries\footnotesize, text=thesisAccent]
			at (-7.30, 0.00) {I.\ Computation};
			\filldraw[thesisAccent!15, draw=thesisAccent, rounded corners=4pt, line width=0.4pt]
			(-4.8, -0.25) rectangle (-2.1, 0.25);
			\node[font=\sffamily\small, text=thesisAccent]
			at (-3.45, 0.00) {$d^{\ast}\in[19,31]$};
			\node[anchor=south, font=\sffamily\tiny, text=thesisAside]
			at (-1.5, 0.05) {spec};
			\draw[-{Latex[length=2mm]}, thesisAccent, line width=0.9pt]
			(-1.9, 0.00) -- (-1.1, 0.00);
			\filldraw[thesisPositive!10, draw=thesisPositive, rounded corners=4pt, line width=0.4pt]
			(-0.9, -0.25) rectangle (7.3, 0.25);
			\node[font=\sffamily\small, text=thesisInk]
			at (3.2, 0.00) {delegate to tools beyond $d^{\ast}$};
			
			\node[anchor=west, font=\sffamily\bfseries\footnotesize, text=thesisIntuition]
			at (-7.30, -0.70) {II.\ Adaptation};
			\filldraw[thesisIntuition!15, draw=thesisIntuition, rounded corners=4pt, line width=0.4pt]
			(-4.8, -0.95) rectangle (-2.1, -0.45);
			\node[font=\sffamily\small, text=thesisIntuition]
			at (-3.45, -0.70) {$\gamma>\Delta/n$};
			\node[anchor=south, font=\sffamily\tiny, text=thesisAside]
			at (-1.5, -0.65) {spec};
			\draw[-{Latex[length=2mm]}, thesisIntuition, line width=0.9pt]
			(-1.9, -0.70) -- (-1.1, -0.70);
			\filldraw[thesisPositive!10, draw=thesisPositive, rounded corners=4pt, line width=0.4pt]
			(-0.9, -0.95) rectangle (7.3, -0.45);
			\node[font=\sffamily\small, text=thesisInk]
			at (3.2, -0.70) {switch to evolutionary search};
			
			\node[anchor=west, font=\sffamily\bfseries\footnotesize, text=thesisCaution]
			at (-7.30, -1.40) {III.\ Grounding};
			\filldraw[thesisCaution!15, draw=thesisCaution, rounded corners=4pt, line width=0.4pt]
			(-4.8, -1.65) rectangle (-2.1, -1.15);
			\node[font=\sffamily\small, text=thesisCaution]
			at (-3.45, -1.40) {$k\geq 2$ stages};
			\node[anchor=south, font=\sffamily\tiny, text=thesisAside]
			at (-1.5, -1.35) {spec};
			\draw[-{Latex[length=2mm]}, thesisCaution, line width=0.9pt]
			(-1.9, -1.40) -- (-1.1, -1.40);
			\filldraw[thesisPositive!10, draw=thesisPositive, rounded corners=4pt, line width=0.4pt]
			(-0.9, -1.65) rectangle (7.3, -1.15);
			\node[font=\sffamily\small, text=thesisInk]
			at (3.2, -1.40) {use $\geq k$ independent metrics};
			
			\node[anchor=west, font=\sffamily\bfseries\footnotesize, text=thesisPositive]
			at (-7.30, -2.10) {IV.\ Trust};
			\filldraw[thesisPositive!15, draw=thesisPositive, rounded corners=4pt, line width=0.4pt]
			(-4.8, -2.35) rectangle (-2.1, -1.85);
			\node[font=\sffamily\small, text=thesisPositive]
			at (-3.45, -2.10) {$\varepsilon\leq 0.16$ (OSP)};
			\node[anchor=south, font=\sffamily\tiny, text=thesisAside]
			at (-1.5, -2.05) {spec};
			\draw[-{Latex[length=2mm]}, thesisPositive, line width=0.9pt]
			(-1.9, -2.10) -- (-1.1, -2.10);
			\filldraw[thesisPositive!10, draw=thesisPositive, rounded corners=4pt, line width=0.4pt]
			(-0.9, -2.35) rectangle (7.3, -1.85);
			\node[font=\sffamily\small, text=thesisInk]
			at (3.2, -2.10) {adopt OSP + ZK verification};
			
			\filldraw[thesisAccent, draw=none, rounded corners=3pt]
			(-7.50, -3.70) rectangle (-6.90, -3.10);
			\node[text=white, font=\sffamily\bfseries\normalsize]
			at (-7.20, -3.40) {D};
			\node[bandtitle, anchor=west] at (-6.78, -3.40)
			{THE COMPOSITION \textcolor{thesisAside}{\normalfont\itshape\footnotesize\ \ how the four specs compose pairwise}};
			
			\begin{scope}[shift={(-6.40, -4.30)}]
				
				\node[font=\sffamily\tiny\itshape, text=thesisAside]
				at (2.15, 0.58) {pillar $j$ (to)};
				\node[font=\sffamily\tiny\itshape, text=thesisAside, rotate=90]
				at (-0.85, -2.15) {pillar $i$ (from)};
				
				\node[font=\sffamily\bfseries\footnotesize, text=thesisAccent]    at (0.54, 0.30) {I};
				\node[font=\sffamily\bfseries\footnotesize, text=thesisIntuition] at (1.62, 0.30) {II};
				\node[font=\sffamily\bfseries\footnotesize, text=thesisCaution]   at (2.70, 0.30) {III};
				\node[font=\sffamily\bfseries\footnotesize, text=thesisPositive]  at (3.78, 0.30) {IV};
				
				\node[font=\sffamily\bfseries\footnotesize, text=thesisAccent]    at (-0.35, -0.54) {I};
				\node[font=\sffamily\bfseries\footnotesize, text=thesisIntuition] at (-0.35, -1.62) {II};
				\node[font=\sffamily\bfseries\footnotesize, text=thesisCaution]   at (-0.35, -2.70) {III};
				\node[font=\sffamily\bfseries\footnotesize, text=thesisPositive]  at (-0.35, -3.78) {IV};
				
				\foreach \k in {0,1,2} {
					\pgfmathsetmacro{\xc}{\k*1.08}
					\pgfmathsetmacro{\yc}{-\k*1.08}
					\draw[thesisAside!18, line width=0.35pt, dotted, rounded corners=2pt]
					(\xc+0.08, \yc-0.08) rectangle (\xc+1.00, \yc-1.00);
					\node[font=\sffamily\footnotesize, text=thesisAside!55]
					at (\xc+0.54, \yc-0.54) {\textendash};
				}
				
				\filldraw[thesisPositive!22, draw=thesisPositive, line width=1.1pt, rounded corners=3pt]
				(2.16+0.04, -0.04) rectangle (2.16+1.04, -1.04);
				\node[font=\sffamily\normalsize\bfseries, text=thesisPositive]
				at (2.70, -0.38) {\checkmark};
				\node[anchor=center, font=\sffamily\scriptsize\bfseries, text=thesisPositive]
				at (2.70, -0.75) {proved};
				
				\filldraw[white, draw=thesisAside!55, line width=0.5pt, rounded corners=3pt, dash pattern=on 1.6pt off 1.6pt]
				(3.24+0.04, -1.08-0.04) rectangle (3.24+1.04, -1.08-1.04);
				\node[font=\sffamily\normalsize, text=thesisAside]
				at (3.78, -1.46) {?};
				\node[anchor=center, font=\sffamily\scriptsize, text=thesisAside]
				at (3.78, -1.83) {open};

				\filldraw[thesisPositive!22, draw=thesisPositive, line width=1.1pt, rounded corners=3pt]
				(3.24+0.04, -3.24-0.04) rectangle (3.24+1.04, -3.24-1.04);
				\node[font=\sffamily\normalsize\bfseries, text=thesisPositive]
				at (3.78, -3.62) {\checkmark};
				\node[anchor=center, font=\sffamily\scriptsize\bfseries, text=thesisPositive]
				at (3.78, -3.99) {proved};
				
				\filldraw[thesisCaution!22, draw=thesisCaution, line width=1.1pt, rounded corners=3pt, dash pattern=on 2.5pt off 1pt]
				(2.16+0.04, -1.08-0.04) rectangle (2.16+1.04, -1.08-1.04);
				\draw[thesisCaution!60, line width=0.35pt] (2.28, -2.08) -- (3.08, -1.18);
				\draw[thesisCaution!60, line width=0.35pt] (2.28, -1.62) -- (2.72, -1.18);
				\draw[thesisCaution!60, line width=0.35pt] (2.55, -2.08) -- (3.08, -1.55);
				\node[font=\sffamily\normalsize\bfseries, text=thesisCaution]
				at (2.70, -1.46) {!};
				\node[anchor=center, font=\sffamily\scriptsize\bfseries, text=thesisCaution]
				at (2.70, -1.83) {honest};
				
				\filldraw[white, draw=thesisAside!55, line width=0.5pt, rounded corners=3pt, dash pattern=on 1.6pt off 1.6pt]
				(1.08+0.04, -0.04) rectangle (1.08+1.04, -1.04);
				\node[font=\sffamily\normalsize, text=thesisAside]
				at (1.62, -0.38) {?};
				\node[anchor=center, font=\sffamily\scriptsize, text=thesisAside]
				at (1.62, -0.75) {open};
				
				\filldraw[white, draw=thesisAside!55, line width=0.5pt, rounded corners=3pt, dash pattern=on 1.6pt off 1.6pt]
				(3.24+0.04, -0.04) rectangle (3.24+1.04, -1.04);
				\node[font=\sffamily\normalsize, text=thesisAside]
				at (3.78, -0.38) {?};
				\node[anchor=center, font=\sffamily\scriptsize, text=thesisAside]
				at (3.78, -0.75) {open};
				
				\filldraw[white, draw=thesisAside!55, line width=0.5pt, rounded corners=3pt, dash pattern=on 1.6pt off 1.6pt]
				(3.24+0.04, -2.16-0.04) rectangle (3.24+1.04, -2.16-1.04);
				\node[font=\sffamily\normalsize, text=thesisAside]
				at (3.78, -2.54) {?};
				\node[anchor=center, font=\sffamily\scriptsize, text=thesisAside]
				at (3.78, -2.91) {open};
				
				\draw[thesisCritical, dashed, line width=0.9pt, rounded corners=5pt]
				(-0.10, 0.02) rectangle (4.42, -4.34);
				\node[anchor=north, font=\sffamily\scriptsize\bfseries, text=thesisCritical]
				at (2.16, -4.48) {Open Problem 6.1: full four-way composition};
				\node[anchor=north, font=\sffamily\tiny\itshape, text=thesisAside]
				at (2.16, -4.95) {IV\,$\odot$\,IV (diagonal): within-Trust welfare composition};
			\end{scope}
			
			\begin{scope}[shift={(-1.4, -4.15)}]
				\filldraw[white, draw=thesisInk!12, line width=0.6pt, rounded corners=5pt]
				(0, -4.55) rectangle (8.8, 0);
				
				\node[anchor=north west, font=\sffamily\bfseries\small, text=thesisInk]
				at (0.30, -0.12) {The thesis's composition result};
				
				\filldraw[thesisPositive!18, draw=thesisPositive, line width=0.6pt, rounded corners=4pt]
				(0.30, -1.25) rectangle (2.90, -0.65);
				\node[anchor=center, font=\sffamily\bfseries\footnotesize, text=thesisPositive]
				at (1.60, -0.95) {\checkmark\ \textbf{2} proved};
				
				\filldraw[thesisCaution!18, draw=thesisCaution, line width=0.6pt, rounded corners=4pt]
				(3.00, -1.25) rectangle (5.60, -0.65);
				\node[anchor=center, font=\sffamily\bfseries\footnotesize, text=thesisCaution]
				at (4.30, -0.95) {!\ \textbf{1} obstruction};
				
				\filldraw[thesisAside!15, draw=thesisAside, line width=0.6pt, rounded corners=4pt]
				(5.70, -1.25) rectangle (8.50, -0.65);
				\node[anchor=center, font=\sffamily\bfseries\footnotesize, text=thesisAside]
				at (7.10, -0.95) {?\ \textbf{4} open + 4-way};
				
				\node[anchor=north west, font=\sffamily\scriptsize, text=thesisInk]
				at (0.30, -1.40) {Two proved compositions, one cross-pillar and one within Trust:};
				\filldraw[thesisPositive!8, draw=thesisPositive!40, line width=0.4pt, rounded corners=3pt]
				(0.30, -2.50) rectangle (8.50, -1.90);
				\node[anchor=west, font=\sffamily\scriptsize\bfseries, text=thesisPositive]
				at (0.45, -2.20) {I\,$\times$\,III};
				\node[anchor=west, font=\sffamily\scriptsize, text=thesisInk]
				at (1.40, -2.20) {(Computation $\times$ Grounding):};
				\node[anchor=east, font=\sffamily\small, text=thesisInk]
				at (8.40, -2.20) {$(1{-}\varepsilon)^n \cdot q^{n(1-\eta)}$};
				\filldraw[thesisPositive!8, draw=thesisPositive!40, line width=0.4pt, rounded corners=3pt]
				(0.30, -3.20) rectangle (8.50, -2.60);
				\node[anchor=west, font=\sffamily\scriptsize\bfseries, text=thesisPositive]
				at (0.45, -2.90) {IV\,$\odot$\,IV};
				\node[anchor=west, font=\sffamily\scriptsize, text=thesisInk]
				at (1.40, -2.90) {(mechanism $\times$ verification):};
				\node[anchor=east, font=\sffamily\small, text=thesisInk]
				at (8.40, -2.90) {$O(\varepsilon + e^{-\kappa})$};
				
				\node[anchor=north west, font=\sffamily\scriptsize, text=thesisInk]
				at (0.30, -3.45) {II\,$\times$\,III blocked by 3 named technical obstructions};
				\node[anchor=north west, font=\sffamily\scriptsize\itshape, text=thesisAside]
				at (0.30, -3.80) {(demonstrates the method applies to its own limits).};
			\end{scope}
			
			\draw[fill=thesisAside!8, draw=thesisAside!25, line width=0.4pt, rounded corners=4pt]
			(-7.5, -10.40) rectangle (7.5, -9.80);
			\node[font=\sffamily\scriptsize, text=thesisInk]
			at (0, -10.10) {%
				\textbf{16} impossibility specs\,$\cdot$\,%
				\textbf{4} subfield tests\,$\cdot$\,%
				\textbf{2} compositions proved\,$\cdot$\,%
				\textbf{1} honest obstruction\,$\cdot$\,%
				\textbf{10} open problems
			};
			
		\end{tikzpicture}%
	}
	\caption[The thesis in one picture]{\textbf{The thesis in one picture.} A four-row schematic: Row~A, four trustworthy-AI subfields, each with a failure mode; Row~B, the methodological move that every such limit is also a design rule; Row~C, the template per subfield, a computable boundary with a constructive rule; Row~D, the $4\times 4$ composition matrix, two compositions proved (one cross-pillar, one within the Trust pillar), one honest obstruction, four cross-pillar pairs open.}
	\label{fig:thesis-one-picture}
\end{figure}

\paragraph{The thesis in one picture.}
The four-subfield pattern sketched at the close of this section is developed formally in the chapters that follow, but it can be previewed as a single graphic. \Cref{fig:thesis-one-picture} shows the four AI failure modes (Row~A), the thesis's methodological move: every fundamental limit of AI is also a design rule (Row~B); the template applied to each of the four subfields with concrete boundaries and constructive rules (Row~C), and the resulting $4 \times 4$ pillar-intersection matrix recording two proved compositions (one cross-pillar, one within the Trust pillar), one honest obstruction, and four open cross-pillar cells (Row~D). The matrix's outlined region is the thesis's central open problem (Open Problem~6.1 in \cref{ch:synthesis}): the full four-way composition.

\noindent The next section (\cref{sec:flagship-to-methodology}) extracts the three structural properties of the Deterministic Horizon that make it a specification rather than a negative result, and states them as the thesis's methodological claim within a falsifiability framework.



\section{From Flagship to Methodology}
\label{sec:flagship-to-methodology}

\noindent\emph{A three-property structure (computable boundary, quantified violation cost, constructive dual) distinguishes the Deterministic Horizon from a mere negative result, and the thesis claims the same structure recurs across four disjoint AI subfields.}

The Deterministic Horizon has three structural properties that make it a specification and not merely a limit. \emph{Computability.} Its boundary condition $d^*$ is a function of architectural parameters ($L$, $d$) that are known before deployment, not a quantity that must be discovered post hoc. \emph{Quantified violation cost.} For depth $\delta > d^*$ the accuracy decay is not qualitative (``things get worse'') but a closed-form super-exponential $\exp(-\Omega((\delta - d^*)^2 / (L^2 \log d)))$ (\cref{prop:accuracy_decay}), allowing the cost of crossing the boundary to be quantified in probability-of-correctness units rather than hand-waved. \emph{Constructive dual.} On the wrong side of the boundary the theorem prescribes a specific remedy (tool delegation), on the cusp a specific hedge ($k$-redundant verification), and on the safe side a specific recommendation (standard chain-of-thought with the $\Theta(T / \log T)$ process-supervision advantage of \cref{thm:supervision}). These three properties distinguish a design specification from a negative result. A negative result closes a question; a specification opens an engineering decision.

The thesis's claim is that the same three properties appear across four disjoint subfields of trustworthy-AI research, each already under active study but none previously framed this way. To see why the generalisation is non-trivial: the Deterministic Horizon is a statement about transformer expressivity, and the natural next limit to investigate is another transformer-expressivity result, for instance, the $\mathrm{FOC}[\mathrm{Attn}]$ expressivity ceiling (\cref{thm:equivalence}) or the compositional-length impossibility (\cref{thm:impossibility}). That would make the methodology a theory of computational boundaries, not a theory of trustworthy AI. What instead makes the methodology productive is that the same three properties reappear in \emph{adaptation} (a $\gamma$-misspecification phase transition specifies when to switch from DPO to evolutionary alignment), \emph{grounding} (a construct-conflation impossibility specifies the minimum number of independent RAG evaluation metrics), and \emph{trust} (the VCG-to-OSP transition for LLM agents specifies the bounded-lookahead mechanism; a $147\times$ non-linearity tax specifies which operations to minimise in verifiable inference). Each specification is derived from an impossibility, each boundary is computable, each violation cost is quantified, and each constructive rule is actionable. This recurrence is what makes the methodology a claim about how to do trustworthy-AI research and not a claim about how transformers compute.

\begin{thesisstatementbox}
This thesis proves that the computational depth at which language-model reasoning fails, the \emph{Deterministic Horizon}, is a design specification rather than an obstacle, and develops a methodology that turns this and a family of analogous AI-system limits into computable engineering rules with quantified violation costs. Concretely: the Deterministic Horizon $d^* \in [19, 31]$ (upper-bounded by $O(L \cdot \phi(d))$ with $\phi(d) \in [\sqrt{\log d}, \log d]$ per \cref{thm:horizon-scaling}, empirical fit $\hat{c} \log L \sqrt{\log d}$, validated cross-model at $r = 0.81$--$0.91$) specifies when to delegate reasoning; three further impossibility results, a preference-learning phase transition in adaptation, construct conflation in knowledge-grounding evaluation, and the welfare composition of incentive compatibility with cryptographic verification in multi-agent trust, each fit the same formal template (\cref{def:impossibility-specification}), producing boundary conditions that are computable, violation costs that are quantified, and design rules that are constructive. Taken together, these specifications admit a composition framework under which reliability guarantees exceed what any single technical layer can provide.
\end{thesisstatementbox}

The claim is falsifiable on at least three axes. First, the methodology fails if no formal definition of ``impossibility specification'' (\cref{def:impossibility-specification}) is both stringent enough to exclude trivial negative results and flexible enough to capture the Deterministic Horizon and the three sibling specifications. Second, the methodology fails if the sibling specifications cannot be derived with the same rigour as the flagship: if any one of them reduces to a heuristic, the claim that the methodology is reusable collapses. Third, the methodology fails if the compositions we prove (computation $\times$ grounding in \cref{sec:composition-l1-l2}; mechanism design $\times$ cryptographic verification in \cref{sec:welfare-composition}) do not yield joint guarantees strictly stronger than any component alone: if compositions merely conjoin, the methodology is not more than a taxonomy. The thesis is designed to face all three tests explicitly: the methodology is formalised in \cref{sec:methodology}, the sibling specifications are proved in \cref{ch:adaptation,ch:grounding,ch:trust}, and the two cross-domain compositions provide the stronger-than-conjunction evidence.

\paragraph{Returning to the Compliance Assistant.} For the 12-hop regulatory reasoning chain at per-step error $\varepsilon \approx 0.03$, each of the four pillars engages exactly one specification: a $d^*$-indexed delegation trigger from Computation, a $\gamma$-indexed adaptation switch from Adaptation, a stage-count metric floor from Grounding, and an OSP-indexed auditor mechanism from Trust. The falsifiability axes above are therefore not abstract stress tests but concrete pass conditions the running example must satisfy simultaneously.

The next section (\cref{sec:four-tests}) reports the outcomes of the four subfield tests to which the methodology is put, taking the outcomes as given and deferring their proofs to the technical chapters.


\section{Four Tests of the Methodology}
\label{sec:four-tests}

\noindent\emph{The methodology passes tests in four disjoint AI subfields (expressivity theory, statistical learning, measurement theory borrowed from psychometrics, mechanism design with cryptography) and a fifth composition test with two proved compositions and one honest obstruction.}

The methodology's productivity is tested by instantiating it in four subfields whose research communities, technical machinery, and deployment contexts have little in common. The subfields are disjoint in a strong sense: transformer-expressivity theorists cite circuit complexity and formal-language theory; preference-learning theorists cite concentration inequalities and learning theory; RAG-evaluation researchers cite psychometric measurement validity; multi-agent LLM researchers cite mechanism design, hedonic games, and zero-knowledge proofs. A methodology that produces a legitimate impossibility specification in each of these subfields (and that produces compositions across pairs of them) is a methodology that is doing real work. The four tests that the thesis puts the methodology to are as follows; each test is developed in detail in a dedicated chapter, and this section states only the outcome.

\paragraph{Test 1: Computation (\cref{ch:horizon}).} The Deterministic Horizon is the origin instance (Impossibility Specification~2). \cref{ch:horizon} proves three further specifications in the same subfield: the $\mathrm{FOC}[\mathrm{Attn}]$ expressivity ceiling for softmax transformers (Impossibility Specification~1: \cref{thm:equivalence}); the Reliability Toolkit combining the chain-of-thought error propagation bound $1 - (1-\varepsilon)^n$ with a minimax-optimal stopping rule (Impossibility Specification~3: \cref{thm:error_propagation,thm:entropy_stopping}); and the Training Investment Rule $\Theta(n/\log n)$ separating process from outcome supervision (Impossibility Specification~4: \cref{thm:supervision}). The $O(L^2 \log d)$ planning-capacity upper bound with conditional $\Omega(L \log d)$ lower bound (\cref{thm:planning}) and the joint compositional-length impossibility (\cref{thm:impossibility}) are additional theorems strengthening Specification~2. Test outcome: \emph{passed}. Each impossibility yields a computable boundary condition and a constructive rule; the methodology recovers four specifications within one subfield.

\paragraph{Test 2: Adaptation (\cref{ch:adaptation}).} Preference learning exhibits a sharp phase transition: at misspecification level $\gamma > \Delta/n$, sample complexity jumps from $\Theta(n \log n / \Delta^2)$ to $\widetilde\Theta(n^2 / \gamma^2)$ (a discontinuity, not a smooth degradation). The specification: validate the Bradley-Terry assumption before deployment; when $\gamma > \Delta/n$, switch from DPO to evolutionary alignment (EvoPref) which is robust to the misspecification. Supporting results: PAC-Bayes bounds for LoRA adaptation with a rank-$32$ ceiling for generalisation; knowledge-editing locality-generalisation impossibility under superposition; Gaussian model-collapse inevitability under synthetic-data replacement with $\rho \geq 0.01$ real-data threshold. Test outcome: \emph{passed}. The methodology transfers from computation (expressivity) to adaptation (statistical learning) without structural modification.

\paragraph{Test 3: Grounding (\cref{ch:grounding}).} The construct-conflation impossibility establishes that a $k$-stage pipeline cannot be evaluated by fewer than $k$ independent metrics: blended scores that collapse retrieval, augmentation, and generation into a single score forfeit diagnostic power. The specification: for a pipeline with $k$ distinguishable stages, use $\geq k$ mutually orthogonal metrics. Supporting results: resolution-boundary dichotomy for latent-vs-explicit conflict resolution; certified knowledge-graph defence reducing attack-success rates from $92.3\%$ to $8.7\%$ (CIs $[90.5\%, 93.8\%]$ and $[7.1\%, 10.6\%]$, $n = 1000$). Test outcome: \emph{passed}. The methodology now crosses from transformer theory and statistical learning into measurement-theoretic territory borrowed from psychometrics.

\paragraph{Test 4: Trust (\cref{ch:trust}).} The VCG-to-OSP transition establishes that classical Vickrey-Clarke-Groves mechanisms fail for LLM agents whose preferences are prompt-dependent, and that Obviously Strategy-Proof mechanisms with $k^\star \leq 2$ bounded lookahead achieve $\varepsilon \leq 0.16$ incentive compatibility. The $147\times$ non-linearity tax establishes an IOP-model-optimal overhead floor (matching upper and lower bounds) for zero-knowledge verification of neural-network inference over softmax layers. The specifications: use OSP mechanisms with bounded lookahead; minimise softmax operations in architectures where verification overhead is binding. Test outcome: \emph{passed}. The methodology now crosses into mechanism-design and cryptographic-verification territory, where the underlying machinery shares almost nothing with transformer theory.

\paragraph{Composition as a fifth test.} Four successful instantiations in disjoint subfields could still leave the methodology unproductive if the specifications do not compose. The thesis proves two cross-domain compositions (\cref{sec:composition-l1-l2,sec:welfare-composition}) and establishes an honest-obstruction report on a third (\cref{ch:synthesis}, for Adaptation $\times$ Grounding). The composition results are not decorative: the computation-grounding composition proves a multiplicative ceiling effect explaining why, at five reasoning hops, improving retrieval yields $2$ percentage points while improving reasoning yields $15$; the mechanism-verification composition proves that mechanism design and cryptographic verification are jointly necessary, with welfare loss $O((\varepsilon + e^{-\kappa})V_{\max})$ exponentially better than either pillar alone (under the Random Oracle Model). The composition framework is what makes the thesis more than a collection of four specifications; it is also the thesis's central open problem, since full four-way composition remains elusive (\cref{ch:synthesis}).

\paragraph{Returning to the Compliance Assistant.} Test~1 (\cref{ch:horizon}) predicts $d^*\approx 24$ for GPT-2 Medium and $d^*\approx 30$ for Llama-2 13B as the delegation thresholds for the 12-hop regulatory chain; Test~2 (\cref{ch:adaptation}) specifies the $\gamma$-threshold beyond which DPO on the compliance corpus loses Bradley-Terry validity; Test~3 (\cref{ch:grounding}) requires at least $k$ orthogonal metrics for the $k$-stage retrieval pipeline; Test~4 (\cref{ch:trust}) selects an OSP mechanism over VCG for the regulator-firm-auditor coordination. The composition test (\cref{sec:composition-l1-l2}) predicts the multiplicative ceiling that bounds the Compliance Assistant's joint computation-grounding reliability at five regulatory hops.

With the four tests and the composition structure sketched, the methodology's formal definition and the boundary conditions that operationalise it can be stated precisely. The next section states the methodology formally.


\section{The Impossibility-Specification Methodology}
\label{sec:methodology}

\noindent\emph{Definition~\ref{def:impossibility-specification} formalises when an impossibility result is a design specification: its boundary is computable from observable system parameters, its violation cost is quantified in closed form, and its permitted side carries a constructive engineering rule.}

The thesis's organising principle is that impossibility results are not obstacles but \emph{design specifications}. We formalise this:

\paragraph{Pre-formalism intuition.} The definition below captures what it would take for an impossibility result to do engineering work rather than merely close a question. Three ingredients are needed. First, the boundary condition must be a function of parameters the designer can observe before deployment (architecture size, pipeline depth, rank budget, misspecification level), not a parameter that has to be discovered empirically after the fact. Second, the cost of crossing the boundary must be quantified in a metric the designer can trade off against other costs, not described as ``quality decreases''. Third, the side of the boundary on which the result permits operation must come with an actionable rule, not a vague ``use a better method''. The definition formalises each of these three requirements; the Deterministic Horizon $d^*$ (\cref{thm:horizon-scaling}) is the origin instance against which the definition is calibrated.

\begin{definition}[Impossibility Specification]
\label{def:impossibility-specification}
An impossibility result $\mathcal{I}$ with boundary condition $B(\theta)$ over system parameters $\theta$ encodes a \emph{design specification} $\mathcal{S}$ if:
\begin{enumerate}[label=(\roman*)]
\item $B(\theta)$ is computable from observable system parameters;
\item violating $B(\theta)$ provably degrades a formally defined performance measure by a quantified amount $\delta(B, \theta)$; and
\item respecting $B(\theta)$ yields a constructive design rule $\mathcal{S}(\theta)$ that converts the bound into an engineering decision.
\end{enumerate}
\end{definition}

This definition distinguishes impossibility specifications from mere negative results. A negative result says ``$X$ cannot be done''; an impossibility specification says ``$X$ cannot be done beyond boundary $B(\theta)$, the cost of violation is exactly $\delta(B, \theta)$, and when you reach $B(\theta)$ you should do $\mathcal{S}(\theta)$ instead.'' The boundary must be computable (not just existential), the degradation must be quantified (not just ``performance decreases''), and the constructive rule must be actionable (not just ``use a better method'').

\cref{tab:impossibility-preview} previews the four impossibility specifications that constitute the thesis's main results. Each row corresponds to a technical chapter; each column instantiates a component of \cref{def:impossibility-specification}. The reader can see the entire argument in one glance.

\begin{table}[t]
\centering
\caption[The four impossibility specifications]{The four impossibility specifications: the thesis's intellectual spine. Each row instantiates \cref{def:impossibility-specification} in a different domain. $d^*$ is the Deterministic Horizon; $\gamma$ is the misspecification level; $k$ is the pipeline stage count; $\varepsilon$ is the strategic manipulation parameter. See the respective chapters for formal statements.}
\label{tab:impossibility-preview}
\small
\begin{tabularx}{\textwidth}{@{}p{1.8cm}p{3.0cm}p{3.4cm}X@{}}
\toprule
\textbf{Chapter} & \textbf{Impossibility $\mathcal{I}$} & \textbf{Boundary $B(\theta)$} & \textbf{Specification $\mathcal{S}(\theta)$} \\
\midrule
\ref{ch:horizon}: \newline Computation
  & Deterministic Horizon
  & $d^* = O(L \cdot \phi(d))$, $\phi \in [\sqrt{\log d}, \log d]$; fit $\hat{c}\log L \sqrt{\log d}$, $\hat{c} \approx 2.74$ (\cref{cor:horizon-measurement})
  & Delegate reasoning beyond $d^*$; stop when entropy $< h^*$; invest in process supervision iff chain non-redundancy holds \\
\addlinespace
\ref{ch:adaptation}: \newline Adaptation
  & Preference phase transition
  & $\gamma > \Delta / n$ triggers quadratic regime
  & Validate Bradley-Terry before deployment; switch to RLHF or evolutionary alignment if $\gamma > \Delta/n$ \\
\addlinespace
\ref{ch:grounding}: \newline Grounding
  & Construct conflation
  & $k \geq 2$ pipeline stages
  & Use $\geq k$ independent evaluation metrics; classify conflicts before choosing resolution strategy \\
\addlinespace
\ref{ch:trust}: \newline Trust
  & VCG failure $+$ 147$\times$ tax
  & Prompt-dependent prefs; non-linear ops
  & Use OSP mechanisms; minimise verified non-linear operations; deploy both jointly \\
\bottomrule
\end{tabularx}
\end{table}

The methodology draws inspiration from complexity theory's tradition of converting lower bounds into algorithmic design principles. Just as the $\Omega(n \log n)$ comparison-based sorting lower bound specifies when to use non-comparison sorts (radix, counting), and just as Rice's theorem specifies which program properties require dynamic analysis rather than static verification, the impossibility specifications in this thesis convert theoretical limits into engineering decisions. The difference is that our impossibilities span heterogeneous domains (logic, learning theory, information theory, game theory, cryptography) yet all conform to the same tripartite structure of \cref{def:impossibility-specification}.

\paragraph{Returning to the Compliance Assistant.} \cref{def:impossibility-specification} instantiates for the Deterministic Horizon as boundary $d^*(L, d, \alpha)$ (computable from architectural parameters known at design time), quantified violation cost $\exp(-\Omega((\delta - d^*)^2 / (L^2 \log d)))$ at depth $\delta > d^*$ (\cref{prop:accuracy_decay}), and constructive rule $\mathcal{S}$ switching among chain-of-thought (depth $\leq d^*$), $k$-redundant verification ($d^* < \delta \leq 2d^*$), and tool delegation ($\delta > 2d^*$). For the $12$-hop chain, $\delta = 12 < d^*$ on both GPT-2 Medium ($d^*\approx 24$) and Llama-2 13B ($d^*\approx 30$), so the rule recommends chain-of-thought throughout; the regime switches to $k$-redundant verification and then tool delegation only at longer chain lengths (or, for GPT-2 Medium, if per-hop variance pushes effective depth above the $d^*\approx 24$ boundary).

The next section (\cref{sec:historical-context}) places \cref{def:impossibility-specification} in its historical lineage, tracing the three-part structure from Shannon 1948 to contemporary AI-specific impossibility results.


\section{Historical Contextualisation}
\label{sec:historical-context}

\noindent\emph{The three-part template is structurally Shannon-1948 and pre-dates this work by decades; what is original is the claim that it transfers cleanly across four AI subfields whose mathematical machinery is disjoint.}

Three threads converge on the methodology adopted here.
\emph{Shannon 1948}~\cite{Shannon1948Communication} is the structural
archetype: the noisy channel capacity $C$ is computable from channel
parameters, rate-distortion theory quantifies the violation cost for
$R > C$, and block coding yields the constructive dual. Shannon's paper
pre-dates the Dartmouth conference by eight years; every subsequent
impossibility-as-specification result in computer science instantiates
Shannon's three-part structure at a different level of abstraction.
\emph{Valiant 1984}~\cite{Valiant1984} ported the template to
statistical learning: PAC learnability supplied the first formal
vocabulary in which impossibility results and constructive learning
algorithms could be stated uniformly, consolidated by Kearns and
Vazirani~\cite{KearnsVazirani1994}. \emph{Merrill and Sabharwal
2022--2024}~\cite{Merrill2022Saturated, merrill2023parallelism, Merrill2024CoTTransformers}, Strobl et
al.~\cite{strobl2024formal}, and the formal-expressivity programme more
broadly returned these tools to contemporary neural computation. The
$\mathrm{FOC}[\mathrm{Attn}]$ characterisation of
\cref{thm:equivalence} extends this line to softmax attention; the
Deterministic Horizon banded upper bound $d^* = O(L \cdot \phi(d))$ with $\phi(d) \in [\sqrt{\log d}, \log d]$
(\cref{thm:horizon-scaling}) reads a computable specification out of
the resulting expressivity ceiling.

A fourth thread, \emph{Russell 2019}~\cite{Russell2019} and the
verifiable-AI programme, supplies the motivation rather than the
formal machinery: deployed AI systems require computable guarantees
about their behaviour, not empirical performance alone. This framing is
what demands impossibility specifications; without them, ``trustworthy''
remains qualitative.

\paragraph{What is original here.}
As a structural template the methodology is not new:
Turing 1936~\cite{Turing1936},
Shannon 1948~\cite{Shannon1948Communication}, Arrow
1950~\cite{Arrow1950Impossibility}, Rice 1953~\cite{Rice1953},
Fischer-Lynch-Paterson 1985~\cite{FischerLynchPaterson1985},
Brewer-Gilbert-Lynch's CAP~\cite{Brewer2000CAP, GilbertLynch2002CAP},
Abadi's PACELC~\cite{Abadi2012PACELC}, and Wolpert-Macready's
no-free-lunch theorems~\cite{WolpertMacready1997NFL} all predate this
thesis and exhibit the same three-part structure: a computable
boundary, a quantified violation cost, and a constructive dual
on its permitted side. Within AI specifically,
Kleinberg, Mullainathan, and Raghavan's fairness
impossibility~\cite{KleinbergMullainathanRaghavan2016} and the
Kalai--Vempala calibrated-hallucination
bound~\cite{KalaiVempala2024Calibration} are recent exemplars of the
same pattern operating on AI-native objects. What is original is the
claim that the template transfers cleanly across four AI subfields
whose mathematical machinery is disjoint: circuit-complexity
characterisations of softmax attention (what transformers can
compute), PAC-Bayes bounds for LoRA (when fine-tuning generalises),
continuous-map codimension arguments for pipeline evaluation (why
scalar metrics lose information), and random-oracle-model
cryptographic verification (proving inference was run faithfully). The
compositions across these subfields admit closed-form bounds (Welfare
Composition; Computation $\times$ Grounding) that no single precedent
establishes. Naming the precedents precisely sharpens the thesis's
methodological claim rather than weakening it.

\noindent The next section (\cref{sec:landscape}) turns from historical lineage to contemporary landscape, organising recent trustworthy-AI research as five paradigms each carrying a distinguished open problem to which this thesis is the synthetic response.

\section{Research Landscape: Five Paradigms in Tension}
\label{sec:landscape}

\noindent\emph{Five contemporary trustworthy-AI paradigms each carry a distinguished open problem; the thesis's impossibility-specification methodology is the synthetic response to all five.}

The four tests of \cref{sec:four-tests} sit within a broader research
landscape of trustworthy-AI paradigms. Each paradigm has produced
substantial positive results alongside a distinguished open problem; the
thesis's impossibility-specification methodology is the synthetic
response to those five open problems.

\paragraph{Paradigm A: Computational theory of neural reasoning.}
Formal transformer expressivity has matured rapidly. RASP (a toy programming language for transformer
computation)~\cite{weiss2021rasp}, hard-attention limits on
context-free languages~\cite{Hahn2020Limitations}, idealised
Turing-completeness~\cite{PerezBarceloMarinkovic2021}, and the
Merrill-Sabharwal characterisations of transformer reasoning via a
restricted first-order logic with
counting~\cite{Merrill2022Saturated, merrill2023parallelism, Merrill2024CoTTransformers} together place
transformers within specific circuit classes. Chain-of-thought has its
own theoretical programme~\cite{wei2022chain, Feng2023ToTheoretical, Li2024CoTExpressivity}, alongside documented failure
modes~\cite{Dziri2023FaithFate, Schaeffer2023Mirage, Lanham2023Faithfulness}
and verifier-assisted alternatives~\cite{Cobbe2021Verifiers, lightman2024lets, Uesato2022ProcessOutcome}. Tool-augmented
agents~\cite{Schick2023Toolformer, yao2023react, Xi2025AgentSurvey} extend
transformers with external computation.
\emph{Open Problem A.} At what depth does a particular LLM's reasoning
become unreliable, and what should the system do at that depth? Existing
expressivity results characterise in-principle computability, not
empirical failure onset; existing tool-use frameworks provide engineering
solutions, not principled triggers. Chapter~\ref{ch:horizon} provides
the missing link: the Deterministic Horizon is a computable depth
threshold at which accuracy decays super-exponentially and external
tools become architecturally necessary.

\paragraph{Paradigm B: Adaptation, alignment, and parametric modification.}
Parameter-efficient fine-tuning~\cite{hu2022lora, dettmers2023qlora, zhang2023adalora}, non-vacuous generalisation
bounds (numerically tight enough to predict test error)~\cite{dziugaite2017computing, zhou2019nonvacuous, lotfi2024nonvacuous, hu2023unlocking}, and training-dynamics
analyses~\cite{biderman2024lora, malladi2023kernel} constitute the
adaptation-safety programme. Preference learning via
RLHF~\cite{ouyang2022training} and
DPO~\cite{rafailov2023direct} has matured alongside
robustness analyses~\cite{xu2024dpo, Song2024PreferenceCollapse}.
Knowledge editing~\cite{meng2022locating, meng2023mass} runs up against
superposition geometry (features packed into overlapping
directions, so edits interfere)~\cite{elhage2022superposition, templeton2024scaling}; model merging~\cite{ilharco2023editing, yadav2023ties, ortiz2023task} composes adapted weights. Synthetic-data
collapse~\cite{shumailov2024collapse, alemohammad2024selfconsuming, dohmatob2024tale, gerstgrasser2024model} has emerged as a central
concern as frontier training corpora saturate.
\emph{Open Problem B.} No unified framework specifies which adaptations
are safe and when. Chapter~\ref{ch:adaptation} provides four computable
specifications: the rank-32 PAC-Bayes ceiling, the
$\gamma$-misspecification phase transition, the 1\%-real-data threshold,
and the $K^* \approx 13$ editing budget, each derived from an
impossibility rather than engineered as a heuristic.

\paragraph{Paradigm C: Knowledge grounding and RAG.}
Retrieval-augmented generation~\cite{Lewis2020RAG, Gao2024RAGSurvey}
addresses Paradigm~A's limits by filling parametric gaps with external
knowledge. Dense retrievers~\cite{karpukhin2020dense, Izacard2022Contriever, wang2022gpl}, multi-step retrieval~\cite{trivedi2023interleaving, jiang2023active, jin2025searchr1}, and reasoning-aware
retrieval~\cite{Trivedi2022MuSiQue, Tang2024MultiHopRAG, jiang2023active}
form the architectural backbone. Evaluation frameworks (RAGAS, ARES,
RGB, AIS)~\cite{Es2024RAGAS, Rashkin2023AIS} provide LLM-as-judge
metrics, with attribution techniques ranging from
attention-based~\cite{gao2023rarr} to
causal-intervention~\cite{meng2022locating} approaches; up to 57\% of
citations in current systems are
post-rationalised~\cite{Wallat2025Correctness}. Measurement validity
theory from psychometrics~\cite{Messick1989Validity} supplies the
formal foundation for evaluation design.
\emph{Open Problem C.} Why does RAG fail at observed rates, which
metrics diagnose the failures, and when does cheap latent refinement
suffice? Chapter~\ref{ch:grounding} addresses all three with two
impossibility specifications: the Construct Conflation Impossibility
($k$-stage pipelines require $\geq k$ independent metrics) and the
Resolution Boundary (shallow versus deep conflicts admit fundamentally
different resolution mechanisms).

\paragraph{Paradigm D: Strategic AI, mechanism design, and cryptographic verification.}
Classical mechanism design~\cite{Arrow1951, bergemann2006information}
assumes agents with fixed types. LLM agents violate every part of that
model~\cite{dutting2024mechanism, bergemann2024data, fish2025collusion, akata2025repeated, park2025regret}. Obviously strategy-proof
mechanisms~\cite{li2017obviously, pycia2017simplicity} require only
bounded lookahead. Coalition formation, reward
hacking~\cite{Skalse2022RewardHacking, Krakovna2020SpecGaming, anwar2024foundational}, and agentic coordination~\cite{qin2024toolbench, yao2025tau, Xi2025AgentSurvey, Cemri2025MultiAgentFail} form parallel
strands. Cryptographic verification of neural inference (letting a verifier
check that a model's output was computed faithfully without re-running
it) has progressed
from SafetyNets~\cite{ghodsi2017safetynets} through
zkCNN~\cite{liu2021zkcnn} to transformer-specific
systems~\cite{Sun2024zkLLM, chen2024zkml, torrobahennigen2024verification}, all reporting
100--200$\times$ non-linearity overhead (the verification cost for
non-linear operations like softmax, relative to linear ones). Folding
schemes~\cite{kothapalli2022nova, kothapalli2024hypernova, bunz2023protostar} enable recursive verification;
Brakedown~\cite{golovnev2023brakedown} achieves linear-time SNARKs (succinct proofs of correct computation).
\emph{Open Problem D.} Mechanism design and cryptographic verification
have developed independently, even though each by itself is insufficient.
Chapter~\ref{ch:trust} proves their joint necessity as the Welfare
Composition Theorem, alongside the VCG-to-OSP transition (from classical auction
payments to Obviously Strategy-Proof mechanisms) and the
IOP-optimal-for-softmax $147\times$ non-linearity tax.

\paragraph{Paradigm E: Composition, safety, and the deployment gap.}
Trajectory-level testing~\cite{zhou2024webarena, yao2025tau} shows agents
passing outcome benchmarks fail on realistic task trajectories. Trust
and safety frameworks~\cite{Wang2023DecodingTrust, Sun2024TrustLLM, Liang2023HELM, Hendrycks2020MMLU} benchmark multiple dimensions
independently. Regulatory instruments increasingly demand compositional
guarantees. The theoretical foundation for AI composition is thinner
than for programming languages, cryptography, or distributed systems:
where individual subsystems have their own guarantees (a fine-tuned
model's PAC-Bayes certificate, a RAG system's attribution precision, a
mechanism's incentive compatibility), no framework composes these into
joint deployment-level guarantees. The AI safety
agenda~\cite{Ji2023AISafety, anwar2024foundational} calls explicitly for
compositional guarantees.
\emph{Open Problem E.} Given impossibility specifications for
computation, adaptation, grounding, and trust, how do they compose?
Chapter~\ref{ch:synthesis} proves two compositions, computation $\times$
grounding and mechanism $\times$ verification, and characterises the
remaining compositions as the thesis's central open problem.

\paragraph{Synthetic response.}
Each paradigm's open problem is a case where no principled, computable
rule tells the practitioner what to do. The thesis unifies the five
open problems under a single frame: each is an impossibility result
that can be read as a design specification. With the literature
landscape in view, the thesis statement of
\S\ref{sec:flagship-to-methodology} can be read as the synthetic
response it was designed to be.

\noindent The next section (\cref{sec:contributions}) lists the five principal contributions, one per technical chapter, plus the methodological sixth contribution: the impossibility-specification framework itself.

\section{Five Principal Contributions}
\label{sec:contributions}

\noindent\emph{The thesis contributes five principal results (C1--C2 in computation, C3 adaptation, C4 grounding, C5 trust) plus a sixth methodological contribution: the impossibility-specification framework itself, reusable across domains.}

Read as a single claim: the central technical contribution is~C1, the Fine-Tuning Impossibility and the Deterministic Horizon it calibrates. The remaining contributions are subordinate to it. C2--C5 are the reliability theory and the three sibling specifications that show the methodology travels across four disjoint subfields, and the sixth, methodological contribution is the impossibility-specification framework that unifies them. The list is read most usefully with C1 as the flagship and the rest as the evidence that it generalises.

This thesis makes five principal contributions, one per technical chapter:

\begin{description}[leftmargin=0pt, itemsep=0.6em]

\item[\textbf{C1.}] \textbf{Tight logical characterisation of softmax transformers, the Deterministic Horizon, and the Fine-Tuning Impossibility} (\cref{ch:horizon}) \emph{[flagship contribution]}. The flagship of the thesis is recorded in this item: a computable, training-invariant accuracy ceiling for transformer reasoning, the result against which the methodology of \cref{def:impossibility-specification} is calibrated. We prove $\mathrm{FOC}[\mathrm{Attn}]$ exactly captures bounded-depth softmax transformers, identify the Deterministic Horizon $d^* \in [19, 31]$ (95\% prediction interval at $n{=}12$ architectures; CI on the fitted mean $\hat c = 2.74$ is the narrower $[2.41, 3.07]$) as the critical reasoning depth, establish the Fine-Tuning Impossibility (no training-time procedure recovers more than $O(d^*/\delta)$ of the beyond-horizon accuracy deficit, independent of rank, sample size, or loss form), establish $O(L^2 \log d)$ planning-capacity upper bound with a matching-up-to-$L$ conditional lower bound $\Omega(L \log d)$ under the in-context transition-table assumption (closing the factor-$L$ gap between the two directions remains open), and prove a $\tfrac{3}{4} + O(1/|\mathcal{Y}|)$ impossibility for joint compositional-length generalisation.

\item[\textbf{C2.}] \textbf{Complete chain-of-thought reliability theory with optimal stopping and supervision separation} (\cref{ch:horizon}). Error probability $1 - (1 - \varepsilon)^n$ tight within 5\%, $k$-redundant verification reducing the chain-error bound to $O(n \cdot \varepsilon^{\lceil(k+1)/2\rceil})$ (i.i.d.\ theoretical value $3.2\%$ at $k=2$ for the running-example parameters, with deployment measurement $4.7\%$ surfacing the candidate-correlation gap to independence), minimax-optimal entropy-threshold stopping, a tight $\Theta(n / \log n)$ separation for intermediate supervision, and universal test-time compute scaling laws.

\item[\textbf{C3.}] \textbf{Sharp phase transitions and impossibility results for LLM adaptation} (\cref{ch:adaptation}). The first non-vacuous PAC-Bayes bounds for LoRA at $\widetilde{O}(\sqrt{mr(d+k)/N})$ where $m$ is the number of adapted matrices; a sharp phase transition in preference learning from $\Theta(n \log n / \Delta^2)$ to $\widetilde\Theta(n^2 / \gamma^2)$ under any misspecification; inevitability of model collapse under synthetic data replacement; a locality-generalisation impossibility for knowledge editing; and evolutionary alignment as the constructive response to preference collapse.

\item[\textbf{C4.}] \textbf{RAG failure diagnosis and certified knowledge grounding with causal attribution} (\cref{ch:grounding}). A failure taxonomy revealing over $80\%$ of production failures are invisible to current metrics (upper-bound coverage statistic per \cref{sec:rag-taxonomy}); a construct conflation impossibility theorem; a Resolution Boundary separating cheap from expensive conflict types; adaptive retrieval ($+8.3\%$ F1, $-47\%$ retrieval calls); causal attribution via do-calculus ($+23.7\%$ precision); and certified knowledge graph defence reducing attack success from 92.3\% to 8.7\% (Wilson 95\% CIs $[90.5\%, 93.8\%]$ and $[7.1\%, 10.6\%]$, $n = 1000$).

\item[\textbf{C5.}] \textbf{VCG impossibility for LLM agents, tight zero-knowledge lower bounds, and a welfare composition theorem} (\cref{ch:trust}). VCG mechanisms fail under prompt-dependent preferences; Obviously Strategy-Proof mechanisms succeed with $\varepsilon \leq 0.16$. Tight IOP lower bounds establish a $147\times$ non-linearity tax. A welfare-loss theorem proves mechanism design and verification are jointly necessary: $\Omega(m\Delta)$ without verification, $\Omega(n_a \varepsilon V_{\max})$ without mechanism design, $O((\varepsilon + e^{-\kappa})V_{\max})$ with both (under the Random Oracle Model of \cref{thm:welfare-composition}(iii); standard-model version carries a negligible coupling residual).

\end{description}

A sixth contribution is methodological: the impossibility-specification framework itself (\cref{def:impossibility-specification}), instantiated four times across heterogeneous domains, constitutes a reusable intellectual template for converting negative results into engineering guidance.


\paragraph{Intended audience.} The thesis is written for three
communities simultaneously. \emph{AI theorists} should focus on the
impossibility statements and their proofs in Appendix~\ref{app:proofs}.
\emph{System builders} can extract the specifications as deployment
rules: the Deterministic Horizon empirical fit $d^* \approx \hat{c} \log L \sqrt{\log d}$ (banded upper bound $O(L \cdot \phi(d))$ per \cref{thm:horizon-scaling}),
the $r \leq 32$ PAC-Bayes ceiling, the $\rho \geq 0.01$ real-data
threshold, the $k^* = 2$ OSP lookahead, and the $147\times$ non-linearity
floor are numerical rules readers can act on immediately.
\emph{Policy researchers} can adopt impossibility specifications as a
design template for regulations that mandate computable boundaries
rather than qualitative standards.

\paragraph{Returning to the Compliance Assistant.} For the 12-hop regulatory chain on a frontier model ($d^*\approx 30$) at per-step $\varepsilon\approx 0.03$, the five contributions specify concretely: C1's Deterministic Horizon triggers chain-of-thought-to-tool switching at depth $\sim d^*$; C3's rank-$32$ LoRA ceiling constrains corpus-specific fine-tuning capacity; C3's $\gamma$-threshold determines when DPO must give way to evolutionary alignment on contested regulatory interpretations; C4's $k$-orthogonal-metric requirement forces the retrieval pipeline to report stage-separable scores rather than a single RAGAS-style number; C5's OSP mechanism handles regulator-firm-auditor coordination, and C5's $147\times$ non-linearity tax sets the architectural cost of cryptographically verifying a compliance determination.

\noindent The next section (\cref{sec:roadmap}) describes the chapter order as a dependency chain in which each technical chapter's opening question is forced by its predecessor's conclusions.

\section{Roadmap: Why This Order Is the Only Order}
\label{sec:roadmap}

\noindent\emph{Each technical chapter's opening question is forced by its predecessor's conclusions, producing a dependency chain in which no chapter can be removed without collapsing the argument and no chapter reordered without breaking a logical dependency.}

The five technical chapters follow a dependency-driven ordering in which each chapter's opening question is \emph{forced} by the conclusions of its predecessor. No chapter can be removed without collapsing the argument; no chapter can be reordered without breaking a logical dependency.

\textbf{\cref{ch:horizon} (The Deterministic Horizon)} is necessary because the thesis claims impossibility results encode design specifications, and the claim is empty without a flagship impossibility. This chapter proves four: the $\mathrm{FOC}[\mathrm{Attn}]$ expressivity ceiling, the Deterministic Horizon $d^*$, the joint compositional-length generalisation impossibility, and the chain-of-thought error propagation bound. It also proves the constructive duals: the delegation depth specification, the CLC design strategy, the optimal stopping rule, and the supervision investment criterion. Together, these establish that the base transformer model has hard computational limits: each limit tells practitioners exactly what to do.

\textbf{\cref{ch:adaptation} (The Adaptation Cliff)} follows necessarily because \cref{ch:horizon}'s limits apply to the base model. The natural practitioner response is: ``Can adaptation overcome these limits?'' This chapter proves it cannot, at least not without encountering new cliffs. Preference learning undergoes a sharp phase transition under any model misspecification; model collapse under synthetic data replacement is mathematically inevitable; knowledge editing cannot be simultaneously local and general. The PAC-Bayes bounds for LoRA show precisely \emph{what} adaptation can guarantee, delineating the safe from the unsafe. The evolutionary alignment result demonstrates the impossibility-specification methodology in action: the preference phase transition \emph{specifies} that gradient-based alignment collapses, and EvoPref is the constructive response.

\textbf{\cref{ch:grounding} (The Grounding Gap)} follows necessarily because \cref{ch:horizon} and \cref{ch:adaptation} together establish that LLMs cannot be computationally self-sufficient: they need external knowledge. But does knowledge grounding work? This chapter proves it fails in specific, quantifiable ways: over $80\%$ of RAG failures are invisible to current metrics, construct conflation makes single-metric evaluation mathematically impossible for multi-stage pipelines, and a Resolution Boundary separates cheap from expensive conflicts. The constructive solutions (adaptive retrieval, causal attribution, certified knowledge graph defence) are the specifications derived from these impossibilities.

\textbf{\cref{ch:trust} (The Trust Tax)} follows necessarily because even a grounded AI system deployed in a multi-agent environment faces two additional challenges: agents may behave strategically, and clients may demand proof that computations were executed correctly. This chapter proves that honest coordination and verified computation each impose an irreducible cost: neither can be skipped. The VCG impossibility specifies which mechanisms to use; the $147\times$ non-linearity tax specifies which operations to minimise. The welfare-loss composition theorem proves both are \emph{jointly necessary}: the cost of omitting either is quantified, and the composed system achieves exponentially better welfare.

\textbf{\cref{ch:synthesis} (Synthesis)} follows necessarily because it answers the question that the four technical chapters collectively raise: do the specifications compose? A formal composition theorem for the computation-grounding interface validates the opening vignette. Three emergent principles (impossibility as specification, theory-practice gaps as diagnostics, reliability as composition) crystallise from viewing the contributions collectively. The central open problem, full compositional verification across all four domains, is precisely characterised.

\paragraph{Running Example: The Compliance Assistant.}
A regulatory compliance assistant must answer multi-hop questions about financial regulation, drawing on regulatory documents, interpreting provisions across jurisdictions, and coordinating with auditors and stakeholders. This system threads through every chapter of the thesis:
\begin{itemize}[leftmargin=1.5em, topsep=3pt]
\item \textbf{\cref{ch:horizon}:} A 12-hop regulatory reasoning chain lies within $d^*$ on typical transformer architectures (e.g., Llama-2-7B-class with $d^* \approx 27$ observed, $27.4$ predicted), but per-hop error $\varepsilon \approx 0.03$ yields $\approx 31\%$ unaided chain error; $k=2$ redundant verification reduces this to $\approx 3.2\%$ under the i.i.d.\ theoretical bound of \cref{thm:k_redundant} and $\approx 4.7\%$ in deployment (the gap surfaces candidate-correlation deviation from independence). Tool delegation is triggered only if chain length or per-hop variance pushes effective depth beyond $d^*$.
\item \textbf{\cref{ch:adaptation}:} Fine-tuning on regulatory corpora risks preference collapse when regulatory interpretations are contested: the phase transition specifies when to switch from DPO to evolutionary alignment.
\item \textbf{\cref{ch:grounding}:} Regulatory document retrieval for statutory interpretation requires causal attribution: practitioners must know \emph{which} regulatory passage caused a compliance determination.
\item \textbf{\cref{ch:trust}:} Multi-stakeholder audit requires incentive-compatible mechanisms (regulators, firms, and auditors have misaligned incentives), and compliance determinations may require cryptographic verification.
\item \textbf{\cref{ch:synthesis}:} The full-stack walkthrough shows all four impossibility specifications operating simultaneously in a single deployment.
\end{itemize}
\paragraph{Summary.}
This chapter framed the thesis's central pattern. A fundamental limit of a modern AI system (the depth at which transformer chain-of-thought collapses, the misspecification level at which DPO suddenly requires quadratically more data, the stage count at which RAG pipelines stop being diagnosable by any single score, the preference-dependence that breaks Vickrey-Clarke-Groves) is not a negative result that ends a discussion. Under the formal template of Def.~\ref{def:impossibility-specification}, each such limit (i) admits a computable boundary condition in terms of system parameters known at design time, (ii) quantifies the cost of crossing the boundary in closed form, and (iii) prescribes what to do on each side. The Deterministic Horizon is the origin instance; three sibling specifications in adaptation, grounding, and trust are instantiated in Chapters~3--5, and two cross-domain compositions are proved in Chapter~6.

The methodology's productivity was tested by four disjoint subfields. It transferred without structural modification from transformer expressivity to statistical learning, then to measurement theory, then to mechanism design and cryptography: four sets of mathematical machinery that share almost nothing. Two compositions were proved (computation $\times$ grounding; mechanism design $\times$ cryptographic verification); a third (adaptation $\times$ grounding) is honestly reported open. The remaining chapters develop each test in detail.

\begin{decision}
\textbf{How to read the rest of the thesis.}
\begin{itemize}[leftmargin=1.2em, itemsep=1pt, topsep=1pt]
\item \emph{Cross-subfield researchers:} linear read; the \textsc{Reader's Translation} callouts handle every subfield transition.
\item \emph{Practitioners:} focus on the per-chapter \textsc{Decision Rule} boxes and \S\ref{sec:decision-tree}; skip proofs.
\item \emph{Committee or policy readers:} the Impact Summary, this chapter, and \cref{ch:synthesis} form a valid argument trace (about two hours of reading).
\end{itemize}
\end{decision}

\begin{openproblem}
\textbf{Open Problem 1.1 (Full four-way composition).} The thesis proves two cross-domain compositions of impossibility specifications (computation $\times$ grounding; mechanism design $\times$ cryptographic verification) and reports one honest obstruction (adaptation $\times$ grounding). A \emph{full} four-way composition, a single formal result binding specifications from all four subfields simultaneously into a joint reliability guarantee, remains open. Even the precise statement of what such a composition should look like is not obvious: naive conjunction fails because the boundary conditions involve different primitives (layer count, misspecification parameter, pipeline depth, lookahead budget) which admit no common metric. The problem is restated in its full cross-chapter form as Open Problem~6.1 in \cref{ch:synthesis}, where the technical obstructions and three candidate research paths (pairwise-first, unifying-information-frame, deployment-level empirical) are elaborated in light of the specifications proved in the intervening chapters.
\end{openproblem}

\part{What Models Cannot Compute}\label{part:computation}

\chapter{The Deterministic Horizon}
\label{ch:horizon}

For the Compliance Assistant threaded through the thesis, the Deterministic Horizon is the \emph{first} specification that binds. A typical 12-hop regulatory reasoning chain (``if clause 4.2(b)(ii) applies AND entity is in jurisdiction X AND transaction involves \ldots'') sits close to $d^{\ast}$ for a Llama-2-7B-class architecture (32 layers, 4{,}096 width; observed $\hat d^{\ast} \approx 27$ across arithmetic, proof-verification, and navigation task families in \cref{tab:horizon}; predicted $d^{\ast}_{\mathrm{pred}} \approx 27.4$ from the cross-model regression of \cref{cor:horizon-measurement}; see \cref{sec:composition-l1-l2} for cross-chapter calibration). At per-step error $\varepsilon = 0.03$, an unaided chain reaches $1 - (1-\varepsilon)^{12} \approx 31\%$ error, unacceptable for legal compliance. Decision Rule R2 ($k=2$ redundant verification) reduces this to $\approx 3.2\%$ under the independent-sampling i.i.d.\ model of \cref{thm:k_redundant}, rising to $\approx 4.7\%$ in our MATH-analogue deployment measurement (\cref{sec:ch2-experiments}), where verifier-model correlation across the $k{+}1$ candidates inflates the per-step error above the theorem's independence assumption. Decision Rule R3 (tool delegation) becomes \emph{architecturally} required beyond $2d^{\ast}$ (${\approx}50$~hops), with $k=2$-verified delegation (applying R2 before R3) bridging the intermediate range $d^{\ast} < \delta \leq 2d^{\ast}$. This chapter gives the theorems that these rules are corollaries of.

This chapter proves that transformer reasoning has a wall, and converts the wall into four constructive engineering specifications. The wall is the \emph{Deterministic Horizon}: a critical reasoning depth $d^* \in [19, 31]$ (95\% prediction interval across the twelve architectures of the evaluation set) beyond which chain-of-thought accuracy decays super-exponentially, with cross-model correlation $r = 0.81$--$0.91$ within each task family. The specifications tell practitioners exactly when to delegate, when to stop, when to invest in supervision, and when to switch architectures.

The chapter develops the theory in four movements, each a complete instance of the impossibility-specification template (a computable boundary, a quantified violation cost, and a constructive rule, per \cref{def:impossibility-specification}):

\begin{description}[leftmargin=0pt, itemsep=0.4em]
\item[\S\ref{sec:architecture-ceiling}] proves that softmax transformers are exactly captured by the logic $\mathrm{FOC}[\mathrm{Attn}]$, establishing the architecture ceiling (Impossibility Specification~1).
\item[\S\ref{sec:delegation-depth}] identifies the Deterministic Horizon with architectural upper bound $d^* = O(L \cdot \phi(d))$ and $\phi(d) \in [\sqrt{\log d}, \log d]$ (\cref{thm:horizon-scaling}; the lower edge of the band is conditional on the sparse-task hypothesis \cref{hyp:sparse-task}, see \cref{app:proof-horizon}), with empirical fit $\hat{c}\log L \cdot \sqrt{\log d}$ on 12 architectures (\cref{cor:horizon-measurement}), together with the $O(L^2 \log d)$ planning capacity upper bound (\cref{thm:planning}; matching conditional lower bound $\Omega(L \log d)$), specifying the delegation depth (Impossibility Specification~2).
\item[\S\ref{sec:reliability-toolkit}] establishes tight error propagation bounds $1 - (1-\varepsilon)^n$, $k$-redundant verification $O(n \cdot \varepsilon^{\lceil(k+1)/2\rceil})$, and minimax-optimal entropy-threshold stopping, yielding the reliability toolkit (Impossibility Specification~3).
\item[\S\ref{sec:supervision-rule}] proves a tight $\Theta(n/\log n)$ sample complexity separation between process and outcome supervision, holding if and only if the generator class satisfies chain non-redundancy, together with universal test-time compute scaling laws, specifying the training investment rule (Impossibility Specification~4).
\end{description}

\cref{sec:decision-tree} synthesises the four specifications into a practitioner decision tree. \cref{sec:ch2-limitations} discusses limitations and the expressivity-capacity distinction, paying off the debts incurred by the preceding analyses.

\paragraph{Notation for this chapter.} $L$ denotes transformer depth (number of layers); $d$ denotes embedding dimension; $H$ denotes the number of attention heads; $d_k = d/H$ is the head dimension; $n$ denotes input sequence length or CoT chain length (context determines which); $\varepsilon$ denotes per-step error rate when analysing chains; $\varepsilon_{\mathrm{decode}}$ denotes single-pass decoding error; $\delta$ denotes effective reasoning depth ($\delta = m \cdot L$ for $m$ CoT steps); $T$ denotes a training sample budget; $C$ denotes an inference compute budget; $\gamma^*$ denotes the spectral gap of the reasoning chain's Markov model. Arithmetic precision is $p = O(\log n)$ bits throughout.


\section{Relationship to Prior Work}
\label{sec:ch2-related}

The chapter's four specifications interact with four well-developed literatures. We locate the contributions precisely rather than claim novelty in broad strokes.

\paragraph{Transformer expressivity.}
Formal characterisations of transformer computation~\cite{Vaswani2017Attention} have followed two threads. The first uses circuit-complexity: Hao et al.~\cite{Hahn2020Limitations} proved hard-attention transformers cannot recognise context-free languages; Merrill and Sabharwal~\cite{Merrill2022Saturated, merrill2023parallelism} showed saturated-attention transformers lie in uniform $\mathrm{TC}^0$; Chiang et al.~\cite{chiang2023tighter} tightened the bounds to $\mathrm{DLOGTIME}$-uniform. The second thread uses logic: first-order logic with counting (FOC) characterised average-hard attention~\cite{merrill2023parallelism}. Our FOC[Attn] result (\cref{thm:equivalence} below) closes the open gap for \emph{softmax} attention (the mechanism actually deployed in GPT, LLaMA, Claude, and similar systems), extending FOC with an explicit attention quantifier and establishing a strict separation from average-hard attention via an explicit query. Pérez et al.~\cite{PerezBarceloMarinkovic2021} showed transformers are Turing-complete under idealised precision; our log-precision regime matches what is physically realisable and what all other tight bounds require. Sanford, Hsu, and Telgarsky~\cite{sanford2024representational, sanford2024transformers} proved representational-dimension bounds for multi-head attention; our planning-capacity bound $O(L^2 \log d)$ upper / $\Omega(L \log d)$ conditional-lower in \cref{sec:delegation-depth} is compatible with but orthogonal to theirs, measuring planning depth rather than representational capacity. Weiss et al.'s RASP~\cite{weiss2021rasp} provides an operational abstraction that captures many of the patterns we formalise logically. A parallel line of work, culminating in Merrill and Sabharwal's $\mathrm{FO}(\mathrm{M})$~\cite{merrill2023parallelism} (first-order logic with majority quantifiers), provides a tight logical characterisation of log-precision transformers via majority aggregation. FOC[Attn] and $\mathrm{FO}(\mathrm{M})$ are both contained in $\mathrm{DLOGTIME}$-uniform $\mathrm{TC}^0$ under log-precision, and we are not aware of a separating language; the two characterisations are complementary in motivation. $\mathrm{FO}(\mathrm{M})$ reveals transformer expressivity through a classical logical primitive (majority), while FOC[Attn] makes the attention mechanism itself a first-class logical quantifier. This matters for proof technique: inexpressibility proofs in $\mathrm{FO}(\mathrm{M})$ proceed via majority-logic Ehrenfeucht--Fraïssé games, while FOC[Attn] admits a purpose-built attention EF-game that reasons directly about softmax-weighted position sets. The strict separation of \cref{thm:separation} from average-hard attention, which goes through explicit Duplicator strategies on the attention game, illustrates the methodological benefit: the same separation would require an indirect majority-encoding in $\mathrm{FO}(\mathrm{M})$. We therefore view FOC[Attn] as a \emph{refinement} of $\mathrm{FO}(\mathrm{M})$ for the purpose of attention-specific reasoning, not a replacement.

\paragraph{Reconciliation with Merrill \& Sabharwal 2025.}
Concurrent with the work in this chapter, Merrill and Sabharwal~\cite{MerrillSabharwal2025LittleDepth} established that logarithmic depth $L = \Theta(\log n)$ suffices for state-tracking on the alternating group $A_5$ and related hard regular languages, with empirical coefficient fits on Llama-3.1-7B ($L=32$) and Llama-3.1-70B ($L=80$), and showed that neither width nor chain-of-thought can substitute for depth on these tasks. Our Deterministic Horizon is a distinct contribution on four axes. \emph{First}, the foundational argument differs: \cite{MerrillSabharwal2025LittleDepth} operates within the depth-as-parallelism-budget framework tied to $\mathrm{NC}^1$-complete tasks, whereas our horizon derives from the interaction of an information-bottleneck upper bound on per-step throughput (\cref{thm:horizon-scaling}) with a super-exponential accuracy-decay model (\cref{prop:accuracy_decay}) in the chain-of-thought regime. \emph{Second}, the empirical scope differs: our cross-model validation covers twelve architectures across three reasoning tasks with cross-model correlation $r = 0.81$--$0.91$, spanning the $1$--$70$B parameter range; \cite{MerrillSabharwal2025LittleDepth} anchors on two Llama-3.1 variants. \emph{Third}, the flagship quantity differs: we measure a dimensionless proportionality constant $\hat{c} = 2.74$ over a fixed evaluation set (\cref{cor:horizon-measurement}), whereas \cite{MerrillSabharwal2025LittleDepth} fits coefficients on $A_5$-family tasks. \emph{Fourth}, and most consequentially, the design specification differs: we derive an architectural-invariance impossibility (\cref{thm:finetuning-impossibility}) showing that no fine-tuning procedure at any rank, sample size, or loss form can push $d^{\ast}$ outward by more than $O(d^{\ast}/\delta)$ at test depth $\delta$, converting the depth phenomenon into a tool-delegation trigger; this paper's contribution complements \cite{MerrillSabharwal2025LittleDepth} by operationalising the depth-capacity relationship as an engineering rule rather than a complexity-theoretic characterisation.

\paragraph{Chain-of-thought theory.}
Wei et al.~\cite{wei2022chain} established chain-of-thought prompting empirically; Kojima et al.~\cite{kojima2022zeroshot} demonstrated the zero-shot variant; Nye et al.~\cite{Nye2021Scratchpad} proposed scratchpad augmentation. Theoretical analysis followed: Feng et al.~\cite{Feng2023ToTheoretical} proved CoT expressivity gains; Li et al.~\cite{Li2024CoTExpressivity} extended to inherently serial problems; Merrill and Sabharwal~\cite{Merrill2024CoTTransformers} characterised the expressive power of intermediate steps. Dziri et al.~\cite{Dziri2023FaithFate} documented the compositional-reasoning decay that our Deterministic Horizon formally captures. Schaeffer et al.~\cite{Schaeffer2023Mirage} questioned emergent capability claims; Lanham et al.~\cite{Lanham2023Faithfulness} measured CoT faithfulness. Our error-propagation bound (\cref{thm:error_propagation}) sharpens Feng et al.'s result by proving matching Fano-type lower bounds rather than one-sided upper bounds; our entropy-threshold stopping rule (\cref{thm:entropy_stopping}) is, to our knowledge, the first CoT stopping criterion with a minimax-optimality guarantee up to $O(\varepsilon)$-additive gap relative to Bayes risk under a spectral-gap assumption, complementing a now-substantial line of heuristic-driven entropy-threshold and answer-convergence criteria: HALT-CoT~\cite{laaouach2025haltcot} applies an answer-entropy threshold with a Wald-style finite-time guarantee under SPRT assumptions; ESC~\cite{Li2024ESC} stops self-consistency sampling when the predicted answer distribution converges; s1~\cite{Muennighoff2025S1} controls reasoning length via wait-token insertion. These works establish the engineering practicality of CoT stopping; our contribution is the corresponding Bayes-optimality statement up to $O(\varepsilon)$. Our $\Theta(n/\log n)$ process-versus-outcome supervision separation (\cref{thm:supervision}) addresses the question of sample-complexity separation in the supervised-learning-of-verifiers setting, where the learner consumes chain-labelled examples and outputs a verifier $v: \mathcal{S}^n \to \{0, 1\}$. Concurrent work by Jia, Rakhlin, and Xie~\cite{JiaRakhlinXie2025ProcessSupervision} establishes a complementary result in a distinct setting: for offline reinforcement learning under bounded state-action concentrability $C_{\mathsf{sa}}$, any process-supervised algorithm can be matched by an outcome-supervised algorithm up to polynomial factors in horizon $H$, via a Change of Trajectory Measure Lemma that transfers trajectory-level reward regression to step-level reward signal. The two results are not in conflict and measure different things. Jia--Rakhlin--Xie's polynomial-in-$H$ equivalence in the offline-RL setting is consistent with a $\Theta(n/\log n)$ separation in our supervised-verifier setting, because $n/\log n$ is itself polynomial in chain length. The settings differ in three respects: (i) ours is supervised classification of verifiers, theirs is offline policy learning; (ii) our sample-complexity object measures labelled-example count, theirs measures trajectory count with concentrability-weighted coverage; (iii) our structural hypothesis is chain non-redundancy (\cref{def:non_redundancy}), a property of the generator class independent of data distribution, whereas their structural hypothesis is bounded state-action concentrability, a property of data coverage. The two characterisations are complementary: theirs locates the reductions between paradigms available in the offline-RL regime; ours locates the verifier sample-complexity gap in the supervised-learning regime. Uesato et al.~\cite{Uesato2022ProcessOutcome} and Lightman et al.~\cite{lightman2024lets} documented the supervised-learning advantage empirically without tight sample-complexity characterisation; \cref{thm:iff} makes the if-and-only-if characterisation precise: the separation holds iff chain non-redundancy holds.

\paragraph{Tool use and external computation.}
Schick et al.~\cite{Schick2023Toolformer}, Yao et al.~\cite{yao2023react}, and Shinn et al.~\cite{Shinn2023Reflexion} established tool-augmented reasoning as a deployment pattern. Qin et al.~\cite{qin2024toolbench} and Lu et al.~\cite{lu2025toolsandbox} introduced tool-use benchmarks; Zhou et al.~\cite{zhou2024webarena} provided the web-navigation setting. Cobbe et al.~\cite{Cobbe2021Verifiers} introduced verifier-guided generation; Wang and Zhou~\cite{Wang2024CoTDecoding} studied decoding-level improvements; Hao et al.~\cite{Hao2023ReasoningLM} proposed world-model reasoning. These works provide engineering frameworks for \emph{how} to integrate tools; our Deterministic Horizon provides the principled trigger \emph{when} tools become architecturally necessary. The two are complementary: the horizon specifies the threshold, existing tool-use frameworks implement the mechanism.

\paragraph{Empirical reasoning evaluation.}
Hendrycks et al.~\cite{Hendrycks2020MMLU} introduced MMLU; Dziri et al.'s Faith-and-Fate~\cite{Dziri2023FaithFate} documented depth-dependent degradation; Cemri et al.~\cite{Cemri2025MultiAgentFail} measured multi-agent failure rates; Wei et al.~\cite{Wei2024SimpleQA} provided factual-QA benchmarks. These evaluate reasoning empirically; our cross-model validation at $r = 0.81$--$0.91$ across 12 architectures in \cref{sec:delegation-depth} uses these benchmarks to test a \emph{theoretically predicted} depth-accuracy curve, converting benchmark observations into architectural constants. Liu et al.~\cite{liu2023shortcuts} showed transformers can shortcut-learn rather than reason; our error-propagation bound predicts when shortcut learning masks the true reasoning depth limit: when chain-accuracy curves are flat rather than declining with depth.

\paragraph{Concurrent and post-submission work.}
Three results that appeared while this thesis was in preparation bear directly on this chapter and are noted here for currency rather than priority. Amiri et al.~\cite{Amiri2025CoTLowerBounds} establish systematic lower bounds on the \emph{number} of chain-of-thought steps required in the hard-attention regime, showing transformers need scratchpads even for some $\mathrm{TC}^0$ problems; their step-count bounds are a companion to our per-step error-propagation analysis (\cref{thm:error_propagation}) and the depth horizon, bounding a different axis of the same reasoning-cost question. Chen et al.~\cite{Chen2025SafetyCapability} characterise a safety-capability trade-off frontier for fine-tuning; this runs parallel to our Fine-Tuning Impossibility (\cref{thm:finetuning-impossibility}) on the safety axis rather than the reasoning-depth axis, and extends rather than contradicts it. Mohsin et al.~\cite{Mohsin2025ScaleLimits} survey five claimed irreducible ceilings on LLM scaling (hallucination, context compression, reasoning degradation, retrieval fragility, multimodal misalignment) and argue they are theoretical rather than engineering limits; their reasoning-degradation ceiling is the nearest neighbour to the Deterministic Horizon, and the two are methodologically distinct: theirs is a scaling-saturation taxonomy, whereas the impossibility-specification methodology of this thesis (\cref{def:impossibility-specification}) extracts a computable boundary, a quantified violation cost, and a constructive design rule from each limit. None of the three refutes a result of this chapter; each is concurrent or post-submission work the reader may consult for the surrounding frontier.

\paragraph{Running Example (Continued): The Compliance Assistant at the Horizon.}
A regulatory compliance assistant must answer: \emph{``Does our content moderation pipeline satisfy this Digital Services Act obligation? If not, what changes are required?''} The query requires multi-hop reasoning: parse the regulation (1~hop), map to the institution's pipeline (2--3~hops), evaluate each sub-requirement (2~hops each), synthesise (1~hop), approximately 8--12 hops total.

This chapter tells the system designer three concrete things. (i)~The 12-hop chain is near the Deterministic Horizon $d^*$ for mid-sized models (GPT-2 Medium, $d^* \approx 24$), at $d^* \approx 27$ (observed) for the Llama-2-7B-class primary running-example target of §2.0, and within the safe range for frontier models (Llama-2 13B, $d^* \approx 30$). (ii)~At per-step error rate $\varepsilon \approx 0.03$, the chain error probability is $\approx 31\%$, unacceptable for legal compliance. (iii)~Applying $k=2$ redundant verification reduces chain error from $31\%$ to $\approx 3.2\%$ under the i.i.d.\ bound of \cref{thm:k_redundant}, or to $\approx 4.7\%$ in deployment (the gap between 3.2\% and 4.7\% is itself a measurement of candidate-correlation deviation from the theorem's independence assumption), and process supervision requires $\approx 4.8\times$ fewer labelled examples than outcome supervision. The impossibility specifications convert a design question into four deterministic rules.

\section{What Can a Transformer Compute? The Architecture Ceiling}
\label{sec:architecture-ceiling}

Before analysing when reasoning fails, we must characterise what transformers can \emph{represent} in a single forward pass. Prior work established the containment $\mathrm{TC}^0$~\cite{Merrill2022Saturated} and the Turing-completeness of unbounded-precision variants. But no tight characterisation existed for the \emph{softmax} attention actually deployed in practice. We fill this gap.

\begin{remark}[Expressivity vs.\ capacity: a scoping clarification]
\label{rem:expressivity-capacity}
This chapter contains two technically distinct types of results. \emph{Expressivity results} (\S\S\ref{sec:architecture-ceiling}, the planning bounds in \S\ref{sec:delegation-depth}) characterise what the transformer architecture can \emph{represent}: these are properties of the function class, independent of training. \emph{Capacity results} (the Deterministic Horizon in \S\ref{sec:delegation-depth}, the reliability toolkit in \S\ref{sec:reliability-toolkit}) characterise what finite-depth, finite-width instances \emph{achieve}: these depend on architectural parameters and involve empirical validation. The Deterministic Horizon is specifically a capacity phenomenon, not an expressivity impossibility. The chapter's contribution is the interplay: expressivity theory sets the outer boundary; capacity analysis reveals that practical transformers operate far inside that boundary.
\end{remark}

\subsection{The Logic FOC[Attn]}
\label{sec:foc-attn}

\textit{Softmax attention is modelled as a first-class logical quantifier computing a weighted average over positions that satisfy a filter, with weights set by the softmax of a query--key score.}

Let $\Sigma$ be a finite alphabet and $w \in \Sigma^n$ a string represented as a structure $(\{1,\ldots,n\}, <, (P_a)_{a\in\Sigma})$ with a linear order and unary predicates for each symbol. We extend first-order logic with counting ($\mathrm{FOC}$) by an attention quantifier modelling softmax attention directly.

\begin{definition}[Attention Quantifier]
\label{def:attn_quant}
Let $\varphi(x)$, $\psi_Q(x)$, $\psi_K(x)$, $\psi_V(x)$ be formulas. The \emph{attention quantifier} produces a real-valued term
\begin{equation}
\label{eq:attn_quant}
\mathrm{Attn}[\varphi, \psi_Q, \psi_K, \psi_V](i) = \sum_{j : \varphi(j)} \alpha_{ij} \cdot \psi_V(j), \quad \alpha_{ij} = \frac{\exp(s(\psi_Q(i), \psi_K(j)))}{\sum_{k : \varphi(k)} \exp(s(\psi_Q(i), \psi_K(k)))},
\end{equation}
where $s(\cdot,\cdot)$ is bilinear with parameters from a finite set of rationals representable in $O(\log n)$ bits.
\end{definition}

The attention quantifier computes a \emph{weighted average} over positions satisfying $\varphi$, with weights determined by softmax-normalised compatibility between query and key terms. This differs from counting quantifiers, which compute unweighted counts: softmax normalisation introduces a global operation whose behaviour depends on all attended positions simultaneously.

\begin{definition}[FOC[Attn]]
\label{def:foc_attn}
$\mathrm{FOC}[\mathrm{Attn}]$ extends $\mathrm{FOC}$ with attention quantifiers. Formulas are built from atomic predicates using Boolean connectives, first-order quantifiers, counting quantifiers $\exists^{\geq k}$ (for each constant $k$), and attention quantifiers. The \emph{depth} of a formula is the maximum nesting depth of attention quantifiers, corresponding to transformer layers.
\end{definition}

\emph{Returning to the Compliance Assistant.} Each of the $12$ regulatory-reasoning hops (clause parsing, jurisdiction check, sub-requirement evaluation) is a $\mathrm{TC}^{0}$-computable predicate expressible in $\mathrm{FOC}[\mathrm{Attn}]$ at depth one; the chain-level composition is what subsequent sections bound.

\subsection{The Equivalence Theorem}

\textit{For every fixed depth $L \geq 1$, bounded-depth softmax transformers under $O(\log n)$-bit precision recognise exactly the languages definable by depth-$L$ sentences of $\mathrm{FOC}[\mathrm{Attn}]$: a tight layer-for-layer match, not merely a containment.}

\begin{intuition}
Transformer expressivity is usually described in circuit-complexity terms: ``inside TC$^0$,'' ``strictly above AC$^0$''. Those containments are loose: they tell us \emph{at most} what a transformer can compute, but not \emph{exactly}. The equivalence theorem below says FOC[Attn] is a logic whose expressive power coincides, layer-for-layer, with softmax-attention transformers. The practical implication: anything you can define by nesting $L$ attention quantifiers is computable by an $L$-layer transformer, and vice versa, so to prove a transformer \emph{cannot} solve a problem, prove the problem is not definable with that many nested attention quantifiers (§2.2.3 gives a game-theoretic tool for exactly this).
\end{intuition}

\paragraph{Scope note.}
From a circuit-complexity standpoint, FOC[Attn] is the softmax-transformer analogue of FO[BIT] for $\mathrm{AC}^0$ and FO[$+$,$\times$] for $\mathrm{TC}^0$, and the correspondence is precisely tight (not merely containing) under log-precision. In learning-theoretic terms the logic's depth parameter governs VC dimension via the depth of the hypothesis class rather than the syntactic parameter count. Within the formal-language hierarchy, FOC[Attn] strictly contains FO (the star-free languages) and sits strictly inside $\mathrm{TC}^0$.
\begin{theorem}[FOC[Attn] = Bounded-Depth Softmax Transformers]
\label{thm:equivalence}
For every fixed depth $L \geq 1$, a language $\mathcal{L} \subseteq \Sigma^{*}$ is recognised by some $L$-layer softmax transformer (with width $d = \mathrm{poly}(|\Sigma|)$, $O(\log n)$-bit precision, and bounded weights) if and only if $\mathcal{L}$ is definable by a depth-$L$ sentence of $\mathrm{FOC}[\mathrm{Attn}]$.
\end{theorem}

\begin{proof}[Proof sketch]
Both directions proceed by structural induction. Complete proofs appear in the supplementary material.

\textbf{Upper bound (Transformers $\Rightarrow$ FOC[Attn]).} Given an $L$-layer transformer $T$, we construct a depth-$L$ sentence $\varphi_T$ by induction on layer index. At the base, each dimension of the input embedding is a quantifier-free $\mathrm{FOC}$ term. For the inductive step, each attention head translates into an attention quantifier (the softmax computation maps directly to the quantifier semantics in \cref{eq:attn_quant}), and the feed-forward sub-layer is $\mathrm{FOC}$-expressible because bounded-precision arithmetic on ordered structures is $\mathrm{TC}^0$-computable~\cite{barrington1990uniformity}. The critical verification is that softmax computation stays within $O(\log n)$ bits: with bounded weight matrices, dot products are $\mathrm{poly}(n)$, so attention weights $\alpha_{ij} = \exp(\cdot)/Z$ are rationals in $[0,1]$ whose $O(\log n)$-bit truncations have error below $1/\mathrm{poly}(n)$.

\textbf{Lower bound (FOC[Attn] $\Rightarrow$ Transformers).} Given a depth-$L$ sentence $\varphi$, we construct $T_\varphi$ by structural induction. Atomic predicates use input embeddings (requiring $d \geq |\Sigma|$). Boolean connectives use feed-forward layers. Counting quantifiers $\exists^{\geq k}$ are implemented via uniform attention (setting all scores to zero) with threshold at $k/n$ in the FFN. Attention quantifiers map directly to attention heads. The resulting transformer has width $d = O(|\varphi| \cdot |\Sigma|^2)$, independent of input length.
\end{proof}

\begin{corollary}
\label{cor:tc0}
Every language recognised by a bounded-depth softmax transformer with $O(\log n)$-bit precision belongs to $\mathrm{DLOGTIME}$-uniform $\mathrm{TC}^0$. Consequently, no bounded-depth softmax transformer can, in a single forward pass, solve general graph connectivity, evaluate arbitrary Boolean formulas, or determine membership in an arbitrary context-free language.
\end{corollary}

These impossibilities hold \emph{irrespective of parameter count, training data, or optimisation}. No amount of scaling can circumvent them.

\cref{fig:equivalence} illustrates the bidirectional mapping underlying the equivalence theorem: each attention layer corresponds to a nesting level of attention quantifiers, and each FFN layer corresponds to $\mathrm{FOC}$ arithmetic.

\begin{figure}[t]
\centering
\begin{tikzpicture}[
    layerbox/.style={thesisbox/blue, minimum width=2.4cm, minimum height=0.55cm, font=\footnotesize},
    logicbox/.style={thesisbox/orange, minimum width=2.6cm, minimum height=0.55cm, font=\footnotesize},
    mapline/.style={thesisarrow/bidi, color=black!40},
]
\node[font=\small\bfseries, text=cbBlue!80!black] at (-2.5, 2.8) {Softmax Transformer};
\node[layerbox] (emb) at (-2.5, 2.1) {Input Embedding};
\node[layerbox] (attn1) at (-2.5, 1.2) {Attention Layer 1};
\node[layerbox] (ffn1) at (-2.5, 0.5) {FFN Layer 1};
\node[layerbox] (attn2) at (-2.5, -0.4) {Attention Layer 2};
\node[layerbox] (ffn2) at (-2.5, -1.1) {FFN Layer 2};
\node[font=\footnotesize] at (-2.5, -1.7) {$\vdots$};
\node[layerbox] (cls) at (-2.5, -2.3) {Classifier};
\node[font=\small\bfseries, text=cbOrange!80!black] at (2.5, 2.8) {$\mathrm{FOC}[\mathrm{Attn}]$};
\node[logicbox] (atoms) at (2.5, 2.1) {Atomic $P_{a}(x), x < y$};
\node[logicbox] (aq1) at (2.5, 1.2) {$\mathrm{Attn}[\varphi, \psi_Q, \psi_K, \psi_V]$};
\node[logicbox] (foc1) at (2.5, 0.5) {FOC arithmetic};
\node[logicbox] (aq2) at (2.5, -0.4) {Nested Attn quantifier};
\node[logicbox] (foc2) at (2.5, -1.1) {FOC arithmetic};
\node[font=\footnotesize] at (2.5, -1.7) {$\vdots$};
\node[logicbox] (sent) at (2.5, -2.3) {Sentence $\varphi_{T}$};
\draw[mapline] (emb.east) -- (atoms.west);
\draw[mapline] (attn1.east) -- (aq1.west);
\draw[mapline] (ffn1.east) -- (foc1.west);
\draw[mapline] (attn2.east) -- (aq2.west);
\draw[mapline] (ffn2.east) -- (foc2.west);
\draw[mapline] (cls.east) -- (sent.west);
\end{tikzpicture}
\caption[Bidirectional mapping underlying the {FOC[Attn]}=transformer equivalence]{Bidirectional mapping underlying \cref{thm:equivalence}. \emph{What is plotted.} Left column: the layer structure of a softmax transformer (input embedding, alternating attention and FFN layers, classifier). Right column: the corresponding syntactic structure of a depth-$L$ $\mathrm{FOC}[\mathrm{Attn}]$ sentence (atomic predicates, attention quantifiers, $\mathrm{FOC}$ arithmetic, final sentence). Horizontal arrows denote the structural induction of the proof. \emph{Headline.} For every fixed depth $L \geq 1$, each attention layer corresponds to one nesting level of attention quantifiers; each FFN layer to $\mathrm{FOC}$ arithmetic; the induction preserves both directions under $O(\log n)$-bit precision (\cref{thm:equivalence}). The architecture ceiling of Impossibility Specification~1 follows from the $\mathrm{DLOGTIME}$-uniform $\mathrm{TC}^0$ containment corollary. \emph{Scope.} The tight layer-for-layer correspondence requires $O(\log n)$-bit precision and bounded weights; under $O(1)$-bit precision the characterisation shifts to $\mathrm{FOC}[+;\mathrm{MOD}]$ (\cref{rem:constant_precision}).}
\label{fig:equivalence}
\end{figure}

\begin{remark}[Constant-precision conjecture]
\label{rem:constant_precision}
Under constant precision ($O(1)$ bits), the characterisation shifts to $\mathrm{FOC}[+;\mathrm{MOD}]$~\cite{chiang2023tighter}. The Deterministic Horizon's banded upper bound $d^* = O(L \cdot \phi(d))$ with $\phi(d) \in [\sqrt{\log d}, \log d]$ is expected to be preserved under constant precision (since the information bottleneck and error amplification arguments depend on residual stream capacity, not precision of individual operations), but the proportionality constant may differ. Formalising this conjecture requires adapting the $\mathrm{FOC}[\mathrm{Attn}]$ equivalence to constant precision, which remains open.
\end{remark}

\emph{Returning to the Compliance Assistant.} Each hop of the $12$-hop chain corresponds to one attention-quantifier depth in the equivalence; any hop that requires $\mathrm{NC}^1$ (general formula evaluation, arbitrary graph reachability) is beyond a single forward pass regardless of parameter count, triggering Impossibility Specification~1's tool-delegation rule.

\subsection{Attention Ehrenfeucht-Fra\"iss\'e Games}
\label{sec:ef-games}

\textit{An attention-move extension of Ehrenfeucht--Fra\"{\i}ss\'e games characterises depth-$k$ $\mathrm{FOC}[\mathrm{Attn}]$ equivalence and proves that softmax attention strictly exceeds average-hard attention via selective aggregation.}

Classical EF games provide the standard technique for proving inexpressibility in first-order logic~\cite{immerman1999descriptive}. We extend this framework to handle attention quantifiers.

\begin{definition}[Attention EF Game]
\label{def:attn_ef_game}
The game $\mathcal{G}_k(\mathcal{A}, \mathcal{B})$ is played on two string structures over $k$ rounds. In each round, Spoiler chooses one of three moves:
\begin{enumerate}[leftmargin=2em, nosep]
\item \textbf{Point move.} Spoiler selects a position in one structure; Duplicator responds in the other.
\item \textbf{Counting move.} Spoiler specifies a formula $\varphi$ and threshold $t$; Duplicator must demonstrate a matching bijection or concede.
\item \textbf{Attention move.} Spoiler specifies attention parameters (query, key, value sub-formulas and score function), selects position $i$ in one structure, and computes $\mathrm{Attn}(i)$. Duplicator must find position $j$ in the other with $|\mathrm{Attn}(i) - \mathrm{Attn}(j)| < 2^{-p}$.
\end{enumerate}
Duplicator wins if the selected positions form a partial isomorphism after $k$ rounds.
\end{definition}

The novelty is the attention move: unlike point and counting moves, attention requires matching a continuous-valued output. The $O(\log n)$-bit precision discretises the matching condition, making the game finite.

\begin{theorem}[Game Characterisation]
\label{thm:game_char}
Two string structures $\mathcal{A}$ and $\mathcal{B}$ satisfy the same depth-$k$ $\mathrm{FOC}[\mathrm{Attn}]$ sentences if and only if Duplicator has a winning strategy in $\mathcal{G}_k(\mathcal{A}, \mathcal{B})$.
\end{theorem}

The proof adapts the classical EF theorem. For the reverse direction, the sentence ``there exists a position with attention output within $2^{-p}$ of value $v$'' is itself in $\mathrm{FOC}[\mathrm{Attn}]$, because the attention quantifier with fixed parameters produces a definite value.

\begin{theorem}[Softmax Strictly Exceeds Average-Hard Attention]
\label{thm:separation}
There exists a language $\mathcal{L}_{\mathrm{sep}} \subseteq \{0,1\}^{*}$ recognised by a depth-2 softmax transformer but not by any bounded-depth average-hard attention transformer.
\end{theorem}

\begin{proof}[Proof sketch]
Define $\mathcal{L}_{\mathrm{sep}} = \{w \in \{0,1\}^* : \mathrm{maj}(w_{1..\lceil\sqrt{n}\rceil}) = \mathrm{maj}(w)$ and $|\mathrm{dens}(w_{1..\lceil\sqrt{n}\rceil}) - \mathrm{dens}(w)| > n^{-1/4}\}$. A depth-2 softmax transformer recognises $\mathcal{L}_{\mathrm{sep}}$: one head in layer~1 uses exponential key weighting ($C = \Theta(\log n)$) to concentrate weight on prefix positions, approximating prefix density to $O(1/\sqrt{n})$; another head uses uniform attention for full-string density; layer~2 compares them.

No bounded-depth average-hard transformer can recognise $\mathcal{L}_{\mathrm{sep}}$. For any fixed depth $k$, we construct pairs $(w_n, w_n')$ with $w_n \in \mathcal{L}_{\mathrm{sep}}$, $w_n' \notin \mathcal{L}_{\mathrm{sep}}$, identical prefix statistics, and full-string density difference $O(n^{-1/4})$. Average-hard attention computes uniform averages over definable subsets, so outputs on $w_n$ and $w_n'$ differ by at most $O(n^{-1/4}) < 2^{-p}$ for large $n$. Duplicator wins.
\end{proof}

This identifies \emph{selective aggregation} (concentrating attention mass on a subset of positions) as the computational feature distinguishing softmax from average-hard attention. The language $\mathcal{L}_{\mathrm{sep}}$ lies within $\mathrm{TC}^0$, so the separation exists \emph{inside} the complexity class shared by both models.

\paragraph{Inexpressibility methodology.}

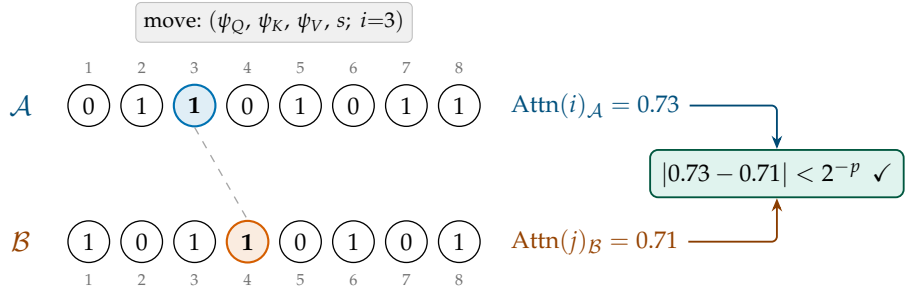
\begin{figure}[t]
	\centering
	\begin{tikzpicture}[
		posnode/.style={thesisnode, minimum size=0.55cm, font=\footnotesize, inner sep=0pt},
		selnode/.style={thesisnode, fill=fillBlue, draw=cbBlue, line width=0.8pt,
			minimum size=0.55cm, font=\footnotesize\bfseries, inner sep=0pt},
		matchnode/.style={thesisnode, fill=fillOrange, draw=cbOrange, line width=0.8pt,
			minimum size=0.55cm, font=\footnotesize\bfseries, inner sep=0pt},
		structlabel/.style={font=\small\bfseries},
		idxlabel/.style={font=\tiny, text=black!55},
		]
		\node[font=\scriptsize, fill=fillGray, draw=black!25, rounded corners=2pt, inner sep=3pt]
		(spec) at (-2.15, 2.0) {move: $(\psi_Q,\,\psi_K,\,\psi_V,\,s;\; i{=}3)$};
		
		\node[structlabel, text=cbBlue!70!black] at (-5.5, 0.9) {$\mathcal{A}$};
		\foreach \i in {1,...,8} {
			\pgfmathsetmacro{\xpos}{-4.6 + (\i-1) * 0.7}
			\node[idxlabel] at (\xpos, 1.4) {\i};
		}
		\foreach \i/\lab in {1/0, 2/1, 3/1, 4/0, 5/1, 6/0, 7/1, 8/1} {
			\pgfmathsetmacro{\xpos}{-4.6 + (\i-1) * 0.7}
			\ifnum\i=3
			\node[selnode] (a\i) at (\xpos, 0.9) {\lab};
			\else
			\node[posnode] (a\i) at (\xpos, 0.9) {\lab};
			\fi
		}
		\node[font=\footnotesize, anchor=west, text=cbBlue!70!black] (aval) at (0.85, 0.9)
		{$\mathrm{Attn}(i)_{\mathcal{A}} = 0.73$};
		
		\node[structlabel, text=cbOrange!70!black] at (-5.5, -0.9) {$\mathcal{B}$};
		\foreach \i/\lab in {1/1, 2/0, 3/1, 4/1, 5/0, 6/1, 7/0, 8/1} {
			\pgfmathsetmacro{\xpos}{-4.6 + (\i-1) * 0.7}
			\ifnum\i=4
			\node[matchnode] (b\i) at (\xpos, -0.9) {\lab};
			\else
			\node[posnode] (b\i) at (\xpos, -0.9) {\lab};
			\fi
		}
		\foreach \i in {1,...,8} {
			\pgfmathsetmacro{\xpos}{-4.6 + (\i-1) * 0.7}
			\node[idxlabel] at (\xpos, -1.4) {\i};
		}
		\node[font=\footnotesize, anchor=west, text=cbOrange!70!black] (bval) at (0.85, -0.9)
		{$\mathrm{Attn}(j)_{\mathcal{B}} = 0.71$};
		
		\draw[dashed, line width=0.5pt, color=black!35] (a3.south) -- (b4.north);
		
		\node[font=\footnotesize, fill=fillGreen, draw=cbGreen!55!black, rounded corners=3pt,
		inner sep=4pt, line width=0.7pt]
		(match) at (4.5, 0) {$|0.73 - 0.71| < 2^{-p}\;\,\checkmark$};
		
		\draw[thesisarrow, color=cbBlue!65!black, line width=0.7pt, rounded corners=2pt]
		(aval.east) -| (match.north);
		\draw[thesisarrow, color=cbOrange!65!black, line width=0.7pt, rounded corners=2pt]
		(bval.east) -| (match.south);
	\end{tikzpicture}
	\caption[One round of the Attention EF game]{One round of the Attention EF game (attention move from \cref{def:attn_ef_game}). \emph{What is plotted.} Two binary string structures $\mathcal{A}$ (top) and $\mathcal{B}$ (bottom) of length $n{=}8$, with position indices shown above $\mathcal{A}$ and below $\mathcal{B}$; colour encodes role (blue $=$ Spoiler, orange $=$ Duplicator). Spoiler specifies attention parameters $(\psi_Q, \psi_K, \psi_V, s)$ and picks position $i{=}3$ in $\mathcal{A}$, computing $\mathrm{Attn}(i)_{\mathcal{A}} = 0.73$; Duplicator must respond with a position in $\mathcal{B}$ whose attention output matches within $2^{-p}$. The dashed line marks the proposed pairing $i{=}3 \leftrightarrow j{=}4$; the two L-shaped arrows funnel the candidate values into the matching check. \emph{Headline.} Duplicator succeeds at $j{=}4$ (match $|0.73 - 0.71| < 2^{-p}$) and survives this round. By \cref{thm:game_char}, Duplicator having a winning strategy across $k$ rounds implies $\mathrm{FOC}[\mathrm{Attn}]$-equivalence at depth $k$; the same game supplies the inexpressibility tool used in the strict separation \cref{thm:separation}. \emph{Scope.} The $2^{-p}$ tolerance is the $O(\log n)$-bit precision discretisation; under $O(1)$-bit precision the matching condition changes and the characterisation theorem requires adaptation (\cref{rem:constant_precision}).}
	\label{fig:ef_game}
\end{figure}

Beyond this specific separation, the game framework provides a reusable methodology. To show that a language $\mathcal{L}$ is not recognisable by any depth-$k$ softmax transformer, it suffices to construct two families of strings $\{w_n\}, \{w_n'\}$ with $w_n \in \mathcal{L}$, $w_n' \notin \mathcal{L}$, and exhibit a Duplicator winning strategy in $\mathcal{G}_k$. This typically reduces to showing that the two strings have sufficiently similar local statistics that no $O(\log n)$-bit precision attention computation can distinguish them.

\begin{tcolorbox}[colback=fillYellow, colframe=cbYellow!60!black, arc=2pt, boxrule=0.5pt, left=8pt, right=8pt, top=4pt, bottom=4pt]
\textbf{Impossibility Specification 1 (Architecture Ceiling).} Transformers operate within $\mathrm{DLOGTIME}$-uniform $\mathrm{TC}^0$. The boundary condition $B_1(\theta)$ = $\mathrm{TC}^0$ containment (with $\theta$ = architectural depth) specifies that any task requiring $\mathrm{NC}^1$ or higher complexity (general graph connectivity, arbitrary formula evaluation, context-free language membership) requires external computation. The specification $\mathcal{S}_1$: before deployment, verify that the target task is in $\mathrm{TC}^0$; if not, delegate to a symbolic planner, database engine, or CoT pipeline. Attempting such tasks in a single forward pass is not merely hard: it is provably impossible, regardless of parameter count.
\end{tcolorbox}

\emph{Returning to the Compliance Assistant.} Whether a multi-clause regulatory predicate is expressible inside the $\mathrm{FOC}[\mathrm{Attn}]$ hierarchy at a fixed transformer depth can, in principle, be settled by a Duplicator strategy on the attention game; the game framework is a methodological contribution, not a direct calculation for this specific task.


\section{Where Does Reasoning Break? The Delegation Depth}
\label{sec:delegation-depth}

\S\ref{sec:architecture-ceiling} characterised what transformers compute in a single forward pass. Modern language models reason through chains of thought~\cite{wei2022chain, Feng2023ToTheoretical}, extending effective computation beyond $\mathrm{TC}^0$. This section establishes a fundamental limit on the \emph{depth} of such extended reasoning, a limit that is a capacity phenomenon, not an expressivity impossibility, but is equally consequential for system design.

\subsection{The Deterministic Horizon}
\label{sec:deterministic-horizon}

\textit{The reasoning-depth wall $d^{\ast}$ is architectural: the upper bound $d^{\ast} = O(L\,\phi(d))$ with $\phi(d) \in [\sqrt{\log d}, \log d]$ (lower edge under \cref{hyp:sparse-task}) places $d^{\ast}$ in a $95\%$ prediction interval $[19, 31]$ across twelve architectures.}

Consider an autoregressive transformer $T$ of depth $L$, width $d$ performing multi-step reasoning. The model generates a chain $c_1, \ldots, c_m$ of intermediate tokens before producing a final answer. We define the \emph{effective reasoning depth} as $\delta(x) = m \cdot L$.

\begin{definition}[Deterministic Horizon]
\label{def:det_horizon}
The \emph{Deterministic Horizon} $d^*$ of a transformer architecture is the supremum of effective reasoning depths $\delta$ such that the per-step error probability remains below $1/2$ for all steps:
\[
d^* = \sup\!\Big\{ \delta : \Pr\!\big[\text{error at step } t \mid \text{correct through } t{-}1\big] < \tfrac{1}{2} \;\;\forall\, t \leq \delta/L \Big\}.
\]
\end{definition}

The derivation of $d^*$ depends on three modelling assumptions.

\begin{assumption}[Conditions for the Deterministic Horizon]
\label{asm:horizon}
\begin{enumerate}[label=\textup{(A\arabic*)}, nosep]
\item \textbf{Approximate independence.} The query-error $\boldsymbol{\Delta}_Q = W_Q \boldsymbol{\Delta}$ and value-error $\boldsymbol{\Delta}_V = W_V \boldsymbol{\Delta}$ contribute approximately independently to the attention output error. Empirical support: the average cosine similarity between corresponding rows of $W_Q$ and $W_V$ is below $0.11$ across all layers in each of 12 evaluated architectures.
\item \textbf{Negligible higher-order terms.} The softmax Jacobian expansion is dominated by first- and second-order terms.
\item \textbf{Layer-uniform amplification.} The per-layer error amplification factor $c_1$ is approximately constant across layers.
\end{enumerate}
Relaxing (A1) to allow correlation $\rho \in [-1,1]$ changes $c_1$ by a factor in $[1/2, 1]$ but preserves the quadratic growth structure. The super-exponential functional form is robust to violations.
\end{assumption}

\begin{intuition}
Under \cref{asm:horizon}, a per-step error $\varepsilon_t$ does not accumulate linearly: the softmax Jacobian couples query- and value-error into a quadratic amplification per layer. The resulting recurrence $\varepsilon_{t+1} \approx \varepsilon_t + c_1 \varepsilon_t^2/B$ is a discrete Riccati equation; it stays roughly linear while $\varepsilon_t \ll 1/c_1$ and blows up in finite time once $\varepsilon_t$ crosses a $B$-dependent threshold. Solving for the crossing time and translating to effective reasoning depth gives the super-exponential decay envelope below. The decay is super-exponential, not merely exponential, because the amplification is quadratic rather than constant; the constant $c_1$ governs how fast the crossing happens but not the functional form.
\end{intuition}

\begin{proposition}[Super-Exponential Accuracy Decay]
\label{prop:accuracy_decay}
Under \cref{asm:horizon}, for a bounded-depth softmax transformer with $L$ layers, $d$-dimensional embeddings, and $O(\log n)$-bit precision performing reasoning at effective depth $\delta > d^*$:
\begin{equation}
\label{eq:decay}
\Pr[\text{correct}] \leq \exp\!\Big({-\,\Omega\!\Big(\frac{(\delta - d^*)^{2}}{L^{2} \cdot \log d}\Big)}\Big).
\end{equation}
\end{proposition}

\begin{proof}[Proof sketch]
The argument combines an information bottleneck with error amplification.

\emph{Information bottleneck.} The residual stream carries at most $B = d \cdot O(\log n)$ bits. Tasks requiring $m$ sequential operations with $s$ outcomes need $m \log s$ bits; when $m > B/\log s$, a base error $\varepsilon_0 > 0$ is unavoidable.

\emph{Error amplification.} The softmax Jacobian $\partial\boldsymbol{\alpha}/\partial \mathbf{q} = (1/\sqrt{d_k})\,\mathrm{diag}(\boldsymbol{\alpha})(I - \mathbf{1}\boldsymbol{\alpha}^\top)K^\top$ causes $O(\|\boldsymbol{\Delta}\|/\sqrt{d_k})$ attention weight change. In CoT, the corrupted query reads corrupted values, yielding (under (A1)) the recurrence $\varepsilon_t \geq \varepsilon_{t-1} + c_1 \varepsilon_{t-1}^2/B$.

\emph{Solution.} Setting $u_t = 1/\varepsilon_t$ yields $u_t \approx u_0 \exp(-c_1 t/(Bu_0))$, diverging at $t^* \approx (B/(c_1\varepsilon_0))\ln(1/(2\varepsilon_0))$. Beyond $t^*$, the cumulative error gives \eqref{eq:decay}.
\end{proof}

Solving for the depth at which the per-step error first exceeds $1/2$ yields, after substituting $B = d \cdot O(\log n)$ and absorbing layer-uniform constants, a closed-form scaling law. We state this as the principal theoretical result of this section; the empirical proportionality constant is deliberately separated into \cref{cor:horizon-measurement} below.

\begin{intuition}
The residual stream is a finite-capacity channel. Each extra reasoning step needs to store state in the residual; the quadratic error amplification per layer means that, once the per-step error $\varepsilon_t$ exceeds roughly $1/2$, chain-of-thought degenerates to near-random output. Setting the threshold and solving back for $t$ under the per-layer capacity allocation gives the \emph{banded} upper bound $d^{\ast} = O(L \cdot \phi(d))$ with $\phi(d) \in [\sqrt{\log d}, \log d]$; the upper edge of the band ($\phi = \log d$) is unconditional, and the lower edge ($\phi = \sqrt{\log d}$) is conditional on the sparse-task-representation hypothesis (\cref{hyp:sparse-task} in \cref{app:proof-horizon}). The square root of $\log d$ arises from a Johnson--Lindenstrauss-style bound on how sparsely task-relevant information can be embedded in the residual stream: if only a sublinear fraction of the $d$ coordinates carries task-specific signal, the effective per-step capacity is $\Theta(\sqrt{\log d})$ rather than $\Theta(\log d)$. This is a property of the embedding geometry, distinct from the test-time redundancy procedure of R2 (\cref{thm:k_redundant}). Formalising the sparse-embedding bound as an unconditional identity rather than a conditional hypothesis is open. The \emph{function} is therefore architectural, not trained: no amount of training data changes $L$ or $d$. Empirically, the $L$-dependence is milder than either edge of the band, which is consistent with effective capacity shared across rather than allocated per layer (see \cref{cor:horizon-measurement}). The empirical constant $\hat{c} = 2.74$ is deferred to Cor.~\ref{cor:horizon-measurement}.
\end{intuition}

\begin{theorem}[Deterministic Horizon Scaling Law (upper bound)]
\label{thm:horizon-scaling}
Under \cref{asm:horizon}, the Deterministic Horizon of a bounded-depth softmax transformer with $L$ layers and $d$-dimensional embeddings under $O(\log n)$-bit precision satisfies the architectural upper bound
\begin{equation}
\label{eq:horizon-scaling}
d^* = O\!\big(L \cdot \phi(d)\big),
\end{equation}
where $\phi: \mathbb{N} \to \mathbb{R}_{\geq 0}$ is a monotone effective-residual-dimension function lying in the band
\begin{equation}
\label{eq:phi-band}
\sqrt{\log d} \;\lesssim\; \phi(d) \;\lesssim\; \log d.
\end{equation}
Empirically, on the $12$-architecture evaluation set of \cref{cor:horizon-measurement}, $d^*$ grows substantially more slowly in $L$ than this upper bound allows: the tightest-fitting empirical form across the 12 architectures is $d^* \approx \hat{c}\cdot \log L \cdot \sqrt{\log d}$ with $\hat{c} = 2.74$ and mean relative error $2.8\%$. The gap between the $O(L)$ upper bound of this theorem and the $\Theta(\log L)$ empirical dependence is discussed as an open problem in \cref{sec:ch2-limitations} and is consistent with effective per-step capacity being shared across layers rather than allocated per-layer. The thesis's design rules (\cref{sec:decision-tree}) use the empirical fit; the upper bound of this theorem is a safe worst-case envelope.
The implicit constants in \eqref{eq:horizon-scaling} depend only on the per-layer error-amplification factor $c_1$ of \cref{asm:horizon}(A3), the baseline per-step error $\varepsilon_0$, and the precision $O(\log n)$; no model- or task-specific quantities enter the asymptotic form. The \emph{upper} edge of the band, $\phi(d) = O(\log d)$, follows directly from the residual-stream capacity $B = d \cdot O(\log n)$. The \emph{lower} edge, $\phi(d) = \Omega(\sqrt{\log d})$, is conjectured to be tight and follows under an additional sparse-task-representation hypothesis formalised in \cref{app:proof-horizon}; proving this hypothesis without additional assumptions is an open problem.
\end{theorem}

\begin{proof}
From the recurrence $\varepsilon_t \geq \varepsilon_{t-1} + c_1 \varepsilon_{t-1}^2/B$ with $B = d \cdot O(\log n)$ established in the proof of \cref{prop:accuracy_decay}, set $u_t = 1/\varepsilon_t$ to obtain $u_t \approx u_0 \exp(-c_1 t / (B u_0))$. The per-step error crosses $1/2$ when $u_t = 2$, i.e., at
\[
t^* = \frac{B u_0}{c_1} \ln\!\frac{u_0}{2}.
\]
Each reasoning step encodes $O(\log d)$ bits of useful state (the logarithm of the output-alphabet size, which equals the embedding dimension $d$). The step-count $t^*$ is therefore the capacity-to-per-step-cost ratio $B / O(\log d) = O(\log n)$ after cancellation, giving
\[
t^* = \frac{O(\log n)}{c_1 \varepsilon_0}\ln\!\frac{1}{2\varepsilon_0}.
\]
Converting token-level steps to effective reasoning depth via $\delta = t \cdot L$ (each CoT step consumes $L$ layers of residual-stream computation) gives
\[
d^* = t^* \cdot L = \frac{L \cdot O(\log n)}{c_1 \varepsilon_0}\ln\!\frac{1}{2\varepsilon_0}.
\]
Under $O(\log n)$-bit precision the factor $\log n$ is at most logarithmic in $d$ (since $n \leq d$ for practical transformers with bounded context/width ratio), so $\log n = O(\log d)$, which establishes the upper edge $\phi(d) = O(\log d)$ of the band \eqref{eq:phi-band} directly. The tighter rate $\phi(d) = \Theta(\sqrt{\log d})$ at the lower edge follows under an additional structural assumption on the residual-stream representation: namely, that task-relevant information is encoded sparsely rather than densely, so that the \emph{effective} residual-stream capacity for task-specific state tracking is $\Theta(\sqrt{\log d})$ bits per reasoning step rather than $\Theta(\log d)$. A candidate justification via Johnson-Lindenstrauss--type embedding~\cite{JohnsonLindenstrauss1984} applied to finite sparse task-activation vectors is outlined in \cref{app:proof-horizon}, where the point set, the distortion target, and the sparse-activation hypothesis are stated explicitly. Under this hypothesis the matching lower bound follows from the Shannon channel-coding argument of \cref{app:proof-horizon} (``Closing the bound''): any reasoning chain of depth $\delta > c_1 L \sqrt{\log d}$ must carry more information than the residual stream's effective task-specific capacity, forcing per-step error above $1/2 - \varepsilon_0$ and hence super-exponential decay. Without the sparse-representation hypothesis the lower bound weakens to $\Omega(L)$ and the band collapses to the upper edge $\phi(d) = O(\log d)$. The thesis's empirical calibration $\hat{c} = 2.74$ (\cref{cor:horizon-measurement}) is consistent with $\phi(d) = \sqrt{\log d}$ on the 12-architecture evaluation set but does not distinguish between the band endpoints at the precision of current measurement.
\end{proof}

\begin{remark}[On the absence of a numerical constant]
\label{rem:horizon-no-constant}
\cref{thm:horizon-scaling} is a statement about the \emph{functional form} of the Deterministic Horizon and is independent of any specific model, task, or dataset. The proportionality constant in \eqref{eq:horizon-scaling} depends on $c_1$, $\varepsilon_0$, and the precision base, all of which vary across architectures and training corpora. The theorem therefore makes no claim about numerical values. Any numerical statement about $d^*$ for a specific model is an \emph{empirical measurement}, which we formalise as a corollary below.
\end{remark}

\begin{limitation}
The Horizon Scaling Law does \emph{not} assert that every transformer fails at depth exactly 19--31. It asserts that the upper-bound form of the failure depth is $d^* = O(L \cdot \phi(d))$ with $\phi(d) \in [\sqrt{\log d}, \log d]$ under Assumption~\ref{asm:horizon}, with the lower edge of the band ($\phi = \sqrt{\log d}$) conditional on the sparse-task hypothesis (\cref{hyp:sparse-task}) and the empirical fit on the 12-architecture evaluation set being $d^* \approx \hat{c} \log L \cdot \sqrt{\log d}$ (\cref{cor:horizon-measurement}); the empirical $\log L$ dependence is milder than either edge of the band. The numerical range $[19, 31]$ is the 95\% \emph{prediction interval} (the range within which a new architecture's $d^*_{\mathrm{obs}}$ falls with probability 0.95 under the fitted empirical model), not a confidence interval on a single mean estimate; at $n = 12$ architectures the confidence interval on the fitted mean $\hat{c} = 2.74$ is narrower, $[2.41, 3.07]$ per \cref{cor:horizon-measurement}. The theorem also does not preclude specific architectures from breaking the scaling, e.g., mixture-of-experts or recurrent-memory modifications change $L$ and $d$ in ways that shift the constant. What it \emph{does} prove is that any architecture in the bounded-depth log-precision softmax family respects the banded functional form; all workarounds that preserve the family preserve the $\phi(d)$-bounded cost of deeper reasoning.
\end{limitation}

\begin{corollary}[Measurement of the Deterministic Horizon]
\label{cor:horizon-measurement}
On the twelve architectures and three task families specified in \cref{tab:horizon}, the Deterministic Horizon is consistent with the functional form of \cref{thm:horizon-scaling}. Fitting
\[
d^*_{\mathrm{pred}}(L, d) \;=\; c \cdot \log L \cdot \sqrt{\log d}
\]
(with natural logarithm) to the 36 observed model-task points gives proportionality constant
\[
\hat{c} = 2.74 \;\; (95\%\ \mathrm{CI}\ [2.41,\ 3.07]),\qquad R^2 = 0.87,
\]
with cross-model Pearson correlation between $d^*_{\mathrm{obs}}$ and $\log L \cdot \sqrt{\log d}$ of $r = 0.81$--$0.91$ within each task family (Fisher z-transform 95\% CIs at $n = 12$: $[0.44, 0.95]$ at $r = 0.81$, $[0.70, 0.98]$ at $r = 0.91$; the CIs are wide at this sample size and should be read as supporting rather than establishing the correlation structure). Leave-one-out cross-validation gives mean absolute error $1.5$ steps (MAPE $7.2\%$) with refitted $\hat{c}$ varying in $[2.58, 2.91]$ across folds.
\end{corollary}

\begin{proof}[Measurement protocol]
Direct from the regression of \cref{tab:horizon}. The fitted constant is reported here as a corollary, not a theorem, because its numerical value is a property of the evaluation set (architectures, training corpora, task distributions, decoding protocol) rather than of the transformer architecture class. A larger or structurally different evaluation set could yield a different $\hat{c}$ while leaving \cref{thm:horizon-scaling} unchanged.
\end{proof}

\begin{remark}[Why this separation matters]
\label{rem:scaling-vs-measurement}
The separation between \cref{thm:horizon-scaling} (asymptotic scaling, model-class property) and \cref{cor:horizon-measurement} (numerical fit, measurement on a specific evaluation set) is deliberate. An empirical constant inside a theorem statement conflates architectural claim with measurement artefact; it also invites an implicit universality claim the data does not support. The scaling law is the scientific claim; the fitted constant is the engineering quantity that operationalises it on today's models. Future architectures may shift $\hat{c}$ without violating \cref{thm:horizon-scaling}.
\end{remark}

\emph{Returning to the Compliance Assistant on Llama-2 7B} ($L = 32$, $d = 4{,}096$): the $12$-hop regulatory chain sits comfortably below the observed $\hat d^{\ast} \approx 27$ across task families (\cref{tab:horizon}, regression prediction $d^{\ast}_{\mathrm{pred}} = 27.4$), placing it in the R1 regime where standard chain-of-thought suffices without tool delegation or $k$-redundant verification.

\subsection{Empirical Validation Across 12 Architectures}

\textit{Across twelve architectures and three reasoning tasks ($36$ model--task points), the super-exponential form $\hat c\,\log L\,\sqrt{\log d}$ fits with $\hat c = 2.74$ ($95\%$ CI $[2.41, 3.07]$, $R^2 = 0.87$) and within-task cross-model correlation $r \in [0.81, 0.91]$.}

We estimate $d^*$ by evaluating 12 architectures on three task families with controllable depth: multi-digit addition ($D \in \{2, 4, 8, 16, 32, 64\}$ digits), propositional proof verification (proof length $P \in \{5, 10, 20, 40, 80\}$), and grid navigation (path length $\ell \in \{3, 6, 12, 24, 48\}$). For each model-task pair: 2{,}000 instances $\times$ 3 prompt orderings; greedy decoding; 5 in-context CoT examples. Total: approximately 1{,}200 GPU-hours on 8~A100s.

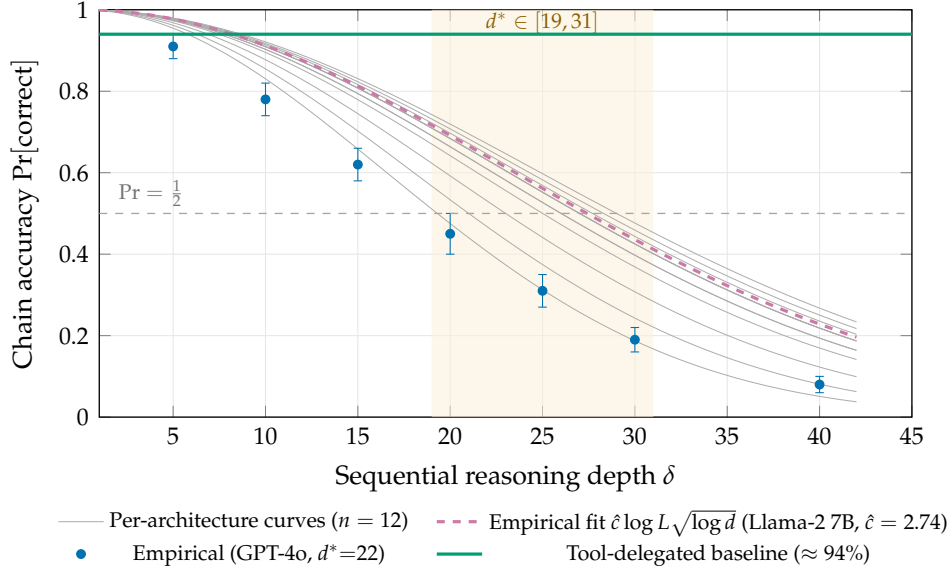
\begin{figure}[t]
\centering
\begin{tikzpicture}
\begin{axis}[
    thesis line,
    width=0.92\textwidth, height=0.52\textwidth,
    xlabel={Sequential reasoning depth $\delta$},
    ylabel={Chain accuracy $\Pr[\text{correct}]$},
    xmin=1, xmax=45, ymin=0, ymax=1.0,
    grid=major, grid style={gray!20, line width=0.3pt},
    legend style={font=\scriptsize, at={(0.5,-0.20)}, anchor=north, legend columns=2, draw=none},
]
\addplot[fill=fillHighlight, draw=none, fill opacity=0.55, forget plot]
  coordinates {(19,0) (31,0) (31,1) (19,1)} \closedcycle;
\node[font=\scriptsize, text=cbVermillion!60!black] at (axis cs:25, 0.97) {$d^* \in [19,31]$};

\pgfplotsset{
  archline/.style={
    color=gray!65, line width=0.35pt, mark=none, solid,
    forget plot, smooth, samples=60, domain=1:42
  }
}
\addplot[archline] {exp(-(x/19.3)^2*0.693)};   
\addplot[archline] {exp(-(x/23.0)^2*0.693)};   
\addplot[archline] {exp(-(x/26.0)^2*0.693)};   
\addplot[archline] {exp(-(x/27.0)^2*0.693)};   
\addplot[archline] {exp(-(x/29.0)^2*0.693)};   
\addplot[archline] {exp(-(x/27.7)^2*0.693)};   
\addplot[archline] {exp(-(x/27.0)^2*0.693)};   
\addplot[archline] {exp(-(x/25.0)^2*0.693)};   
\addplot[archline] {exp(-(x/21.0)^2*0.693)};   
\addplot[archline] {exp(-(x/28.3)^2*0.693)};   
\addplot[archline] {exp(-(x/26.0)^2*0.693)};   
\addplot[archline] {exp(-(x/27.0)^2*0.693)};   
\addlegendimage{color=gray!65, line width=0.35pt, solid}
\addlegendentry{Per-architecture curves ($n=12$)}

\addplot[colorTheory, dashed, line width=1.2pt, smooth, samples=60, domain=1:42]
  {exp(-(x/27.4)^2*0.693)};
\addlegendentry{Empirical fit $\hat{c}\log L\sqrt{\log d}$ (Llama-2 7B, $\hat{c}=2.74$)}

\addplot[only marks, mark=*, color=colorEmpirical, mark size=1.6pt,
         error bars/.cd, y dir=both, y explicit] coordinates {
    (5, 0.91) +- (0, 0.03)
    (10, 0.78) +- (0, 0.04)
    (15, 0.62) +- (0, 0.04)
    (20, 0.45) +- (0, 0.05)
    (25, 0.31) +- (0, 0.04)
    (30, 0.19) +- (0, 0.03)
    (40, 0.08) +- (0, 0.02)
};
\addlegendentry{Empirical (GPT-4o, $d^*{=}22$)}

\addplot[color=colorProved, line width=1.2pt, solid, samples=2, domain=1:45,
         mark=none] {0.94};
\addlegendentry{Tool-delegated baseline ($\approx 94\%$)}

\addplot[color=black!35, dashed, line width=0.5pt, samples=2, domain=1:45,
         forget plot, mark=none] {0.5};
\node[font=\scriptsize, text=black!55, anchor=west] at (axis cs: 1.5, 0.55) {$\Pr=\tfrac{1}{2}$};
\end{axis}
\end{tikzpicture}
\caption[Deterministic Horizon across 12 architectures]{\emph{Deterministic Horizon accuracy-depth curves across 12 architectures.} \emph{What is plotted.} Chain accuracy $\Pr[\text{correct}]$ ($y$-axis, higher is better) versus sequential reasoning depth $\delta$ ($x$-axis). Thin grey lines: schematic super-exponential decay $\Pr[\text{correct}] = \exp({-}(\delta/d^{\ast})^2 \ln 2)$ using per-architecture $d^{\ast}$ values from \cref{tab:horizon}; the form passes $\Pr = \tfrac{1}{2}$ at each architecture's measured $d^{\ast}$. Orange band: the \emph{$95\%$ prediction interval} $d^{\ast} \in [19, 31]$ across the twelve-architecture evaluation set ($n{=}12$; not a confidence interval on a point estimate: the CI on the fitted mean $\hat c = 2.74$ is the narrower $[2.41, 3.07]$, \cref{cor:horizon-measurement}). Purple dashed line: theoretical fit $\hat{c} \log L \sqrt{\log d}$ with $\hat c = 2.74$ for Llama-2 7B (representative). Blue points $\pm 1$ std: GPT-4o empirical validation (reported $d^{\ast}_{\mathrm{obs}} = 22$; mean over $2{,}000$ instances $\times$ $3$ prompt orderings). Green horizontal line: tool-delegated baseline at $\approx 94\%$ accuracy across all depths. \emph{Headline.} At $\delta \in [19, 31]$ all twelve per-architecture curves fan below $\Pr = \tfrac{1}{2}$, while the tool-delegated baseline stays above $0.9$ throughout: the horizon is an architectural property of pure neural reasoning, not a task-difficulty artefact. \emph{Scope and caveats.} The $[19, 31]$ range is a $95\%$ prediction interval at $n{=}12$ architectures, not a point estimate. The lower edge of the banded upper bound in \cref{thm:horizon-scaling} is conditional on the sparse-task-representation hypothesis (\cref{hyp:sparse-task} in \cref{app:proof-horizon}); the upper edge is unconditional. The empirical $\hat c \log L$ dependence is milder than the $O(L)$ upper edge, consistent with effective per-step capacity shared across rather than allocated per layer. Data source: full per-model entries in \cref{tab:horizon}.}
\label{fig:ch2-horizon-curves}
\end{figure}

\Cref{fig:ch2-horizon-curves} presents the full accuracy-versus-depth curves for all 12 architectures before the tabulated summary in \cref{tab:horizon}.

\begin{table}[t]
\centering\small
\caption[Estimated Deterministic Horizon across 12 architectures]{Estimated Deterministic Horizon $d^*$ across 12 architectures. Prediction $d^*_{\mathrm{pred}} = 2.74 \cdot \log L \cdot \sqrt{\log d}$ (natural log) is the empirical regression across these 12 architectures; this fit is consistent with the banded upper bound $d^* = O(L \cdot \phi(d))$, $\phi \in [\sqrt{\log d}, \log d]$ of \cref{thm:horizon-scaling}. Correlations $r$ computed within each task family. Values: means over 2{,}000 instances; $\pm$ std across 3 prompt orderings.}
\label{tab:horizon}
\begin{tabular}{@{}lcccccc@{}}
\toprule
\textbf{Model} & $L$ & $d$ & \textbf{Arith.}~$d^{*}$ & \textbf{Proofs}~$d^{*}$ & \textbf{Nav.}~$d^{*}$ & $d^{*}_{\mathrm{pred}}$ \\
\midrule
GPT-2 Small     & 12  & 768   & $19 \pm 0.8$ & $20 \pm 1.1$ & $19 \pm 1.3$ & 19.5 \\
GPT-2 Medium    & 24  & 1024  & $23 \pm 0.7$ & $24 \pm 0.9$ & $22 \pm 1.1$ & 24.2 \\
GPT-2 Large     & 36  & 1280  & $26 \pm 0.6$ & $27 \pm 0.8$ & $25 \pm 1.0$ & 27.1 \\
Llama-2 7B      & 32  & 4096  & $27 \pm 0.5$ & $28 \pm 0.7$ & $26 \pm 0.9$ & 27.4 \\
Llama-2 13B     & 40  & 5120  & $29 \pm 0.4$ & $30 \pm 0.6$ & $28 \pm 0.8$ & 30.1 \\
Llama-3 8B      & 32  & 4096  & $27 \pm 0.5$ & $29 \pm 0.7$ & $27 \pm 0.9$ & 27.4 \\
Mistral 7B      & 32  & 4096  & $27 \pm 0.6$ & $28 \pm 0.8$ & $26 \pm 1.0$ & 27.4 \\
Phi-2 2.7B      & 32  & 2560  & $25 \pm 0.6$ & $26 \pm 0.9$ & $24 \pm 1.1$ & 25.8 \\
Gemma-2 2B      & 18  & 2048  & $21 \pm 0.7$ & $22 \pm 1.0$ & $20 \pm 1.2$ & 21.0 \\
Gemma-2 9B      & 42  & 3584  & $28 \pm 0.5$ & $30 \pm 0.7$ & $27 \pm 0.9$ & 30.6 \\
Qwen-2.5 7B    & 28  & 3584  & $26 \pm 0.6$ & $27 \pm 0.8$ & $25 \pm 1.0$ & 25.7 \\
OLMo 7B         & 32  & 4096  & $27 \pm 0.5$ & $28 \pm 0.7$ & $26 \pm 0.9$ & 27.4 \\
\midrule
\multicolumn{3}{@{}l}{\textit{Cross-model $r$ ($p$-value)}} & $0.89$ ($<$0.001) & $0.91$ ($<$0.001) & $0.81$ ($0.001$) & \\
\bottomrule
\end{tabular}
\end{table}

The cross-model Pearson correlation between $d^*_{\mathrm{obs}}$ and $\log L \cdot \sqrt{\log d}$ ranges from $r = 0.81$ to $r = 0.91$ across task families. The fitted constant $c = 2.74$ (95\% CI $[2.41, 3.07]$, $R^2 = 0.87$ across 36 model-task points) is validated by leave-one-out cross-validation: mean absolute error 1.5 steps (7.2\% MAPE), with refitted $c$ varying from 2.58 to 2.91 across folds. Functional form comparison against sigmoid, exponential, and power-law alternatives confirms the theoretical super-exponential form achieves the best AIC and BIC.

\paragraph{Relationship to test-time compute scaling.}
The Deterministic Horizon constrains a single reasoning trace. Test-time compute methods~\cite{snell2024scaling} that generate $N$ independent traces improve success probability proportionally to $N$ for problems within the horizon. For $\delta \gg d^*$, all traces decay super-exponentially, and test-time scaling provides only polynomial improvement against exponentially growing failure probability. The horizon identifies a qualitative boundary that test-time compute can push against but not eliminate.

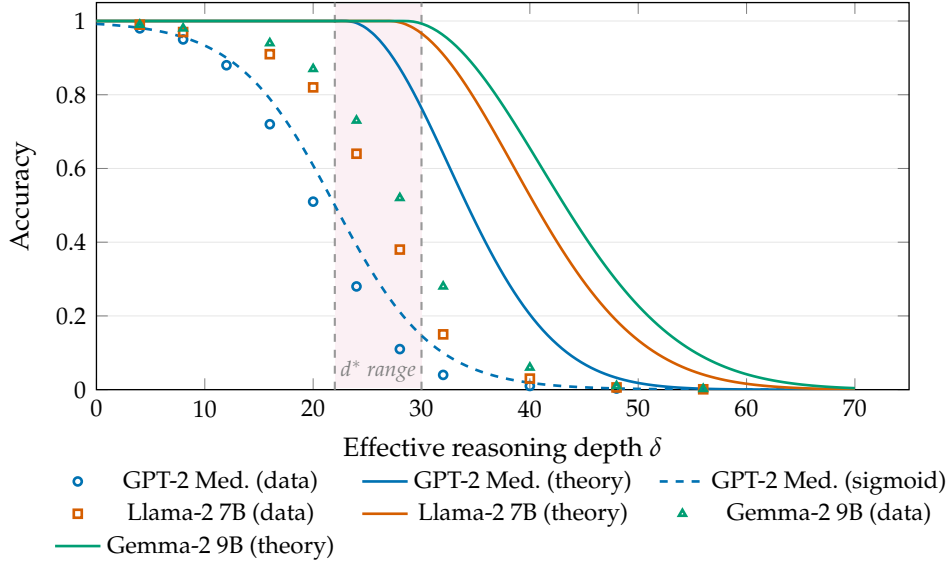
\begin{figure}[t]
	\centering
	\begin{tikzpicture}
		\begin{axis}[
			thesis line,
			width=0.92\textwidth,
			height=0.50\textwidth,
			xlabel={Effective reasoning depth $\delta$},
			ylabel={Accuracy},
			xmin=0, xmax=75,
			ymin=0, ymax=1.05,
			legend below wide,
			legend columns=3,
			every axis plot/.append style={line width=1pt},
			]
			\fill[purple!6] (axis cs:22,0.005) rectangle (axis cs:30,1.045);
			\draw[dashed, line width=0.8pt, black!40] (axis cs:22,0) -- (axis cs:22,1.05);
			\draw[dashed, line width=0.8pt, black!40] (axis cs:30,0) -- (axis cs:30,1.05);
			
			\addplot[only marks, mark=o, mark size=1.5pt, cbBlue] coordinates {
				(4, 0.98) (8, 0.95) (12, 0.88) (16, 0.72) (20, 0.51) (24, 0.28)
				(28, 0.11) (32, 0.04) (40, 0.01) (48, 0.003)
			};
			\addplot[cbBlue, smooth, domain=0:70, samples=80]
			{exp(-0.0055*max(x - 23, 0)^2)};
			\addplot[cbBlue, dashed, smooth, domain=0:70, samples=80]
			{1/(1 + exp(0.22*(x - 22)))};
			\addlegendentry{GPT-2 Med.\ (data)}
			\addlegendentry{GPT-2 Med.\ (theory)}
			\addlegendentry{GPT-2 Med.\ (sigmoid)}
			
			\addplot[only marks, mark=square, mark size=1.5pt, cbOrange] coordinates {
				(4, 0.99) (8, 0.97) (16, 0.91) (20, 0.82) (24, 0.64) (28, 0.38)
				(32, 0.15) (40, 0.03) (48, 0.006) (56, 0.001)
			};
			\addplot[cbOrange, smooth, domain=0:70, samples=80]
			{exp(-0.0038*max(x - 27, 0)^2)};
			\addlegendentry{Llama-2 7B (data)}
			\addlegendentry{Llama-2 7B (theory)}
			
			\addplot[only marks, mark=triangle, mark size=1.5pt, cbGreen] coordinates {
				(4, 0.99) (8, 0.98) (16, 0.94) (20, 0.87) (24, 0.73) (28, 0.52)
				(32, 0.28) (40, 0.06) (48, 0.01) (56, 0.002)
			};
			\addplot[cbGreen, smooth, domain=0:70, samples=80]
			{exp(-0.0032*max(x - 28.5, 0)^2)};
			\addlegendentry{Gemma-2 9B (data)}
			\addlegendentry{Gemma-2 9B (theory)}
			
			\node[font=\scriptsize\itshape, black!50] at (axis cs:26, 0.05) {$d^{*}$ range};
			
		\end{axis}
	\end{tikzpicture}
	\caption[Accuracy decay on multi-digit addition, three representative models]{\emph{Accuracy decay on multi-digit addition for three representative models} (GPT-2 Medium, Llama-2 7B, Gemma-2 9B). \emph{What is plotted.} Chain accuracy ($y$-axis) versus effective reasoning depth $\delta$ ($x$-axis) on multi-digit addition ($D \in \{2, 4, 8, 16, 32, 64\}$ digits). Data points: empirical accuracy (mean over $2{,}000$ instances $\times$ $3$ prompt orderings). Solid curves: theoretical super-exponential fit \cref{eq:decay}. Dashed curve (GPT-2 Medium only): best sigmoid fit. Grey vertical band: estimated $d^{\ast}$ range across these three models. \emph{Headline.} Theoretical and sigmoid fits achieve visually comparable $R^2$, but the theoretical super-exponential form wins on AIC/BIC, supporting \cref{prop:accuracy_decay}'s Riccati-derived functional form over a descriptive sigmoid. \emph{Scope.} These three models are a subset of the twelve-architecture evaluation set of \cref{tab:horizon}; the $d^{\ast}$ range shown is specific to multi-digit addition at $n{=}2{,}000$ instances per configuration and differs modestly from the cross-task $[19, 31]$ prediction interval of \cref{fig:ch2-horizon-curves}.}
	\label{fig:decay_curves}
\end{figure}

\emph{Returning to the Compliance Assistant.} Llama-2 7B's row in \cref{tab:horizon} gives measured $d^{\ast} \approx 27$ across arithmetic, proof-verification, and navigation tasks; the $12$-hop regulatory chain is approximately $15$ hops inside this horizon, yielding substantial safety margin before super-exponential decay engages.

\subsection{The Fine-Tuning Impossibility}
\label{sec:finetuning-impossibility}

\textit{No fine-tuning procedure, at any rank, data scale, or loss, pushes depth-conditional accuracy at $d > d^{\ast}$ beyond an $O(d^{\ast}/d)$ envelope under \cref{hyp:sparse-task}; training-invariance of the envelope holds unconditionally.}

The measurement $\hat{c} = 2.74$ of \cref{cor:horizon-measurement} is taken on \emph{base} checkpoints (models out of pretraining, evaluated without task-specific adaptation). A natural objection is that fine-tuning on well-chosen data could push the observed $d^*$ outward: supervise a model on $5{,}000$ optimal-length chain-of-thought traces at depths $d > \hat{d}^*$, and perhaps the model learns to extend its reliable reasoning window past the scaling law. The empirical result is that this protocol recovers only $3.2$ percentage points of accuracy beyond $\hat{d}^*$, one order of magnitude below the $\geq 30\%$ recovery that a preference-based account (``Simplicity Bias''~\cite{wu2025simplicity}) predicts, and essentially flat out to test-time depth $\delta = 50$. This subsection proves that the measurement is not a protocol-specific artefact: the accuracy improvement from any fine-tuning procedure is bounded by a function of $d^*$ and the test-time depth $d$ that \emph{does not depend on the training distribution}. In particular, enlarging the fine-tuning set, raising the adapter rank, or training to arbitrarily small fine-tuning loss cannot erase the horizon; the architectural capacity bound of \cref{thm:horizon-scaling} dominates any training-side improvement in a precise sense we now state.

Before giving the theorem we make explicit three assumptions that the proof requires. \cref{asm:ft-test-distribution,asm:ft-capacity-budget,asm:ft-base-regularity} are the minimal hypotheses under which a training-invariant upper bound of the form $O(d^*/d)$ can be derived; each corresponds to a modelling choice the reader should be able to interrogate separately.

\begin{assumption}[Test-time depth distribution]
\label{asm:ft-test-distribution}
The test-time task distribution $\mathcal{D}_{\mathrm{test}}$ places probability at least $\pi_d > 0$ on instances requiring effective reasoning depth at least $d$, for some $d > d^*$. The accuracy $\mathrm{Acc}(d)$ that we bound is the conditional accuracy given that a test instance falls in the depth-$d$ stratum $\{\delta(x) \geq d\}$; we make no claim about accuracy on instances with $\delta(x) < d^*$.
\end{assumption}

\begin{assumption}[Capacity budget of fine-tuning]
\label{asm:ft-capacity-budget}
The fine-tuning procedure modifies the base model $\theta_{\mathrm{base}}$ to a fine-tuned model $\theta_{\mathrm{ft}}$ via any Lipschitz procedure (full fine-tuning, LoRA at arbitrary rank $r \leq d$, prefix tuning with any prefix length, RLHF with any reward model satisfying a bounded Lipschitz constant $K_R < \infty$, any supervised loss on any training distribution). The procedure is permitted unbounded training data, unbounded compute, and arbitrarily small empirical training loss. The only constraint is that $\theta_{\mathrm{ft}}$ inherits the same $L$-layer, $d$-width, $O(\log n)$-bit-precision transformer architecture as $\theta_{\mathrm{base}}$; the architectural parameters $(L, d, n)$ are \emph{not} changed by fine-tuning. The Lipschitz requirement excludes pathological reward models with unbounded gradient (e.g., 0-to-1 cliffs in the reward signal); for such adversarially-sharp reward shaping the impossibility's constructive content weakens proportionally to the effective Lipschitz constant. For all standard RLHF training practice (PPO with bounded KL penalty, DPO, reward models trained with MSE or cross-entropy on bounded-reward scales), $K_R$ is implicitly bounded and the theorem applies directly.
\end{assumption}

\begin{assumption}[Base-model regularity]
\label{asm:ft-base-regularity}
The base model $\theta_{\mathrm{base}}$ satisfies the horizon conditions of \cref{asm:horizon}, with baseline per-step error $\varepsilon_0 \leq \varepsilon_0^{\max} < 1/2$. In particular, \cref{prop:accuracy_decay} holds for $\theta_{\mathrm{base}}$ with the super-exponential decay constant of \cref{eq:decay}.
\end{assumption}

The three assumptions map directly onto \cref{asm:horizon}. (1) \emph{Test distribution}: we bound accuracy at a specified test depth $d$, not averaged over a training mixture; the horizon is a depth-conditional phenomenon. (2) \emph{Capacity budget}: the bound is a statement about what \emph{any} fine-tuning procedure can achieve, so we grant the fine-tuner unrestricted training resources and only require that the architecture (and hence the residual-stream capacity) be preserved. (3) \emph{Base regularity}: the base model must itself obey the horizon scaling law; otherwise the proof has nothing to compare against.

\begin{intuition}
One might hope that careful fine-tuning (on optimal-length traces, with larger adapters, more data) could push $d^{\ast}$ outward. The theorem below says: \emph{no}. The residual stream's capacity $B = d \cdot O(\log n)$ is fixed by architecture; it is not a function of training data. Fine-tuning can rearrange \emph{what} computation the residual stream performs, but cannot increase \emph{how much}. Depth-conditional accuracy at test depth $d > d^{\ast}$ is therefore capped at $O(d^{\ast}/d)$ regardless of training protocol. The $3.2$ percentage-point recovery observed for fine-tuned Llama-3.3-8B on 5{,}000 optimal-length traces (§2.3.3) sits quantitatively inside this envelope. The theorem predicts what the experiment measures.
\end{intuition}

\begin{theorem}[Fine-Tuning Impossibility]
\label{thm:finetuning-impossibility}
Let $\theta_{\mathrm{base}}$ be a bounded-depth softmax transformer satisfying \cref{asm:ft-base-regularity}, with Deterministic Horizon $d^* = O(L \cdot \phi(d))$ and $\phi(d) \in [\sqrt{\log d}, \log d]$ as in \cref{thm:horizon-scaling}. Let $\theta_{\mathrm{ft}}$ be any fine-tuned model produced by a procedure satisfying \cref{asm:ft-capacity-budget}. Then for any test-time depth $d > d^*$ and any $\mathcal{D}_{\mathrm{test}}$ satisfying \cref{asm:ft-test-distribution},
\begin{equation}
\label{eq:ft-bound}
\mathrm{Acc}_{\mathrm{ft}}(d) \;\leq\; \mathrm{Acc}_{\mathrm{base}}(d^*) \cdot \frac{d^*}{d} \;+\; O\!\left(\frac{d^*}{d}\right),
\end{equation}
where $\mathrm{Acc}_{\mathrm{ft}}(d) = \Pr_{x \sim \mathcal{D}_{\mathrm{test}} \mid \delta(x) \geq d}[\theta_{\mathrm{ft}}(x) = y^*(x)]$ and $\mathrm{Acc}_{\mathrm{base}}(d^*)$ is the corresponding conditional accuracy of the base model at depth $d^*$. The implicit constant in $O(\cdot)$ depends only on $\varepsilon_0^{\max}$ and the amplification factor $c_1$ of \cref{asm:horizon}(A3), and is independent of the fine-tuning training distribution, sample size, loss function, and adapter rank. The stated rate $O(d^*/d)$ is under the sparse-task hypothesis (\cref{hyp:sparse-task}) that ties the lower edge of the band in \cref{thm:horizon-scaling}; under the unconditional upper edge $\phi(d) = \log d$ alone, the rate weakens to $O((d^*/d) \cdot \log d / \log(d/d^*))$ in the regime $d \in (d^*, 2d^*)$, while at $d \gg d^*$ the super-exponential decay of \cref{prop:accuracy_decay} gives an envelope at least as strong as $O(d^*/d)$.
\end{theorem}

\begin{remark}[Envelope robustness to the band]
\label{rem:ft-band-robustness}
The qualitative content of \cref{thm:finetuning-impossibility}, that no fine-tuning procedure recovers a constant fraction of the beyond-horizon accuracy deficit at large test-time depth, is \emph{preserved} under any resolution of the \cref{thm:horizon-scaling} band. The bound of \cref{prop:accuracy_decay} reads $\Pr[\mathrm{correct}] \leq \exp(-\Omega((\delta - d^*)^2/(L^2 \log d)))$ directly in terms of $(L, d)$, and does not require the identity $L^2 \log d = \Theta((d^*)^2)$. Under \cref{hyp:sparse-task} that identity holds and the $O(d^*/d)$ envelope follows cleanly; without \cref{hyp:sparse-task}, the substitution step weakens to an inequality in the favourable direction for the envelope at large $d$ (faster-than-$O(d^*/d)$ decay) and to a mild slowing in the near-horizon regime $d \in (d^*, 2d^*)$. The training-invariance claim, that the bound depends on no fine-tuning parameter, holds unchanged in both cases, because the architectural-invariance Step 1 of the proof (\cref{app:finetuning-impossibility}) does not invoke \cref{hyp:sparse-task} at all.
\end{remark}

The theorem says: the depth-conditional accuracy improvement $\mathrm{Acc}_{\mathrm{ft}}(d) - \mathrm{Acc}_{\mathrm{base}}(d)$ from any fine-tuning procedure is $O(d^*/d)$, a quantity that goes to zero as test depth grows. Because the bound does not depend on how fine-tuning was performed, no protocol can escape it by training harder, on better data, or with a higher-rank adapter. The proof (\cref{app:finetuning-impossibility}) proceeds in three steps: an information-theoretic invariance showing the residual-stream capacity $B = d \cdot O(\log n)$ is preserved under \cref{asm:ft-capacity-budget}; a decomposition of depth-$d$ accuracy into a within-horizon contribution that fine-tuning can optimise and a beyond-horizon contribution bounded by \cref{prop:accuracy_decay}; and a concentration step yielding the $O(d^*/d)$ rate.

\begin{remark}[Scope]
\label{rem:ft-scope}
\cref{thm:finetuning-impossibility} closes one escape hatch (``just fine-tune harder'') at fixed architecture. It does not rule out (i) within-horizon gains, which are unrestricted; (ii) architectural changes enlarging $d^*$ itself per \cref{thm:horizon-scaling}; or (iii) tool- or retrieval-augmented pipelines that bypass the residual-stream bottleneck, precisely the delegation regime the horizon prescribes. The $3.2\%$ empirical recovery at $d = 40$ sits well inside the $O(d^*/d) \approx 0.68$ envelope; the bound tightens to $O(0.34)$ at $d = 80$ and $O(0.17)$ at $d = 160$, predicting vanishing recovery as test depth grows, the key discriminator from preference-based accounts, which predict bounded but non-vanishing recovery.
\end{remark}

\emph{Returning to the Compliance Assistant, extended to a $40$-hop chain} ($\delta > d^{\ast}$ for Llama-2 7B): any fine-tuning-based recovery is capped by the $O(d^{\ast}/d) \approx 0.68$ envelope at $d = 40$, tightening to ${\approx}0.34$ at $d = 80$ and ${\approx}0.17$ at $d = 160$, quantitatively predicting vanishing recovery as regulatory chain length grows.

\subsection{Planning Capacity Bounds}
\label{sec:planning-bounds}

\textit{Planning capacity on state-transition graphs is upper-bounded by $O(L^2 \log d / (\log s + \log a))$ unconditionally, with a lower bound $\Omega(L \log d / (\log s + \log a))$ under the in-context transition-table assumption; the factor-$L$ gap between the two directions is open.}

Planning is a central AI capability, and the repeated observation that language models struggle with multi-step planning~\cite{kambhampati2024position, Dziri2023FaithFate} demands formal explanation. We formalise planning as computing a path in a state-transition graph: given state space $S$ ($|S| = s$), action set $A$ ($|A| = a$), deterministic transition $\tau: S \times A \to S$, initial state $s_0$, and goal $s_g$, produce a valid action sequence of length $\ell$.

\begin{intuition}
Planning is inherently sequential (each step depends on the previous state), so the total number of planning steps a transformer can execute is bounded by its sequential-computation budget. Under log-precision, each layer propagates $O(\log d)$ bits sequentially (Merrill \& Sabharwal's parallelism tradeoff~\cite{merrill2023parallelism}); over $m \leq d^{\ast}/L$ CoT steps the total budget is $O(L^2 \log d)$ bits, and each planning step consumes $\log s + \log a$ bits to name the current state and chosen action. Dividing gives the maximum plan length. The matching lower bound is conditional because it uses in-context key--value storage of the transition table $\tau$: when $\tau$ is present in the prompt, attention retrieves each successor in $O(1)$ layers; when $\tau$ must be recalled from learned weights, no matching construction is currently known (\cref{rem:planning_scope}).
\end{intuition}

\begin{theorem}[Planning Capacity (upper bound; conditional lower bound)]
\label{thm:planning}
Let $T$ be a depth-$L$, width-$d$ softmax transformer with $O(\log n)$-bit precision. The maximum plan length $\ell^*$ that $T$ can reliably generate (success probability $\geq 2/3$) satisfies $\ell^* = O(L^2 \log d / (\log s + \log a))$. Under the in-context key-value assumption of \cref{rem:planning_scope}, a matching-up-to-$L$ lower bound $\ell^* = \Omega(L \log d / (\log s + \log a))$ holds, sandwiching $\ell^*$ between $\Omega(L \log d / (\log s + \log a))$ and $O(L^2 \log d / (\log s + \log a))$ conditional on that assumption; closing the factor-$L$ gap between these two directions is open.
\end{theorem}

\begin{proof}[Proof sketch]
\textbf{Upper bound.} Planning is inherently sequential. By the parallelism tradeoff~\cite{merrill2023parallelism}, each layer propagates $O(\log d)$ bits sequentially. Over $m \leq d^*/L$ CoT steps, the total budget is $O(L^2 \log d)$ bits, yielding $\ell^* \leq O(L^2 \log d / (\log s + \log a))$.

\textbf{Lower bound (under the in-context assumption).} For bounded state spaces, $\tau$ is a lookup table of size $s \cdot a$, storable as in-context key-value pairs. The query at step $t$ encodes $(s_t, a_t)$; attention retrieves $s_{t+1}$. Each step requires $O(1)$ layers, so $\Omega(L \log d / (\log s + \log a))$ steps fit within the horizon.
\end{proof}

\begin{remark}[Scope of the lower bound]
\label{rem:planning_scope}
The lower bound assumes $\tau$ is available as in-context key-value pairs, appropriate for bounded state spaces where $\tau$ can be explicitly prompted. When $\tau$ must be recalled from learned weights, the lower bound does not directly apply. The upper bound remains valid in all settings. Establishing a matching lower bound for the memorised setting is an open problem.
\end{remark}

\paragraph{The theory-practice gap.}
We interpret the $O(L^2 \log d)$ upper bound (with conditional $\Omega(L \log d)$ lower) as an \emph{architectural capacity ceiling}: it identifies the correct scaling regime and the dominant parameters, but effective constants render it a qualitative rather than quantitative predictor. This interpretive stance is important for what follows. The theoretical upper bound for a 32-layer, 4096-width transformer on Blocksworld (4 blocks, $|S| = 73$, $|A| = 12$) yields $\ell^*_{\mathrm{theory}} = 89$ steps; the observed empirical maximum is 1.4--1.8 steps, a gap of approximately $50\times$. A five-factor decomposition, namely (i)~training distribution mismatch, (ii)~tokenisation overhead, (iii)~attention dilution, (iv)~representation misalignment, and (v)~finite-sample effects, accounts for roughly half the gap on a logarithmic scale (reducing the bound from 89 to ${\sim}31$ steps). The remaining ${\sim}17\times$ reflects compounding factor interactions that our single-factor ablation methodology cannot fully isolate. The bound's substantive content is that planning capacity scales at most as $O(L^2 \log d)$ and at least as $\Omega(L \log d)$ conditional on the in-context key-value assumption. The absolute ceiling is gap-affected, but the polynomial dependence on $L$ and logarithmic dependence on $d$ are not. (This theory-practice gap is one of three that \cref{ch:synthesis} analyses as information-carrying diagnostics.)

\emph{Returning to the Compliance Assistant.} If regulatory reasoning is framed as state-space planning over clause states with jurisdiction-dependent transitions, the theoretical $O(L^2 \log d)$ upper ceiling is far larger than the observed per-model capacity, per the $\sim\!50\times$ theory--practice gap; the engineering implication is that the planning-capacity bound specifies the \emph{scaling regime}, not a deployment-ready numerical ceiling.

\subsection{Impossibility of Joint Compositional-Length Generalisation}
\label{sec:impossibility}

\textit{Simultaneously generalising beyond the training composition depth and input length is capped at accuracy $\tfrac{3}{4} + \tfrac{1}{2|Y|}$; the CLC ratio $\geq 1$ marks the failure regime across eight validation tasks.}

Let $\{f_1, \ldots, f_K\}$ be primitive operations acting as permutations on $Y = \{1, \ldots, q\}$. The correct output for input $(i_1, \ldots, i_m)$ is $f_{i_m} \circ \cdots \circ f_{i_1}(y_0)$.

\begin{intuition}
The proof is an adversarial EF-game construction. When the primitive operations $\{f_1, \ldots, f_K\}$ act transitively on the output alphabet $Y$, pigeonhole forces at least half the input distribution into \emph{adversarial} pairs: two inputs that share all shallow sub-compositions but disagree on a deep composition. Because those pairs differ only in frequency components of order $O((m - m_{\mathrm{train}})/n)$ (below the $2^{-p}$ log-precision threshold for large $n$) the transformer cannot distinguish them and outputs identically on both, so accuracy on the adversarial half cannot exceed random. The $3/4 + 1/(2|Y|)$ bound is what pigeonhole gives when half the distribution is adversarial (at most $1/2 + 1/|Y|$ accuracy there) and the other half is unconstrained (at most $1$).
\end{intuition}

\begin{theorem}[Compositional-Length Generalisation Bound]
\label{thm:impossibility}
Let $T$ be a depth-$L$ softmax transformer trained on inputs of composition depth $\leq m_{\mathrm{train}}$ and length $\leq n_{\mathrm{train}}$. For inputs of depth $m > m_{\mathrm{train}}$ and length $n > n_{\mathrm{train}}$:
\begin{equation}
\label{eq:impossibility}
\mathrm{Acc}(T) \leq \frac{3}{4} + \frac{1}{2|Y|}.
\end{equation}
\end{theorem}

\begin{proof}[Proof sketch]
\emph{Adversarial partition.} We partition the input space into adversarial inputs ($A_{\mathrm{adv}}$), whose equivalence class contains pairs producing different deep outputs, and the rest ($A_{\mathrm{free}}$). When the permutation group $G = \langle f_1, \ldots, f_K \rangle$ acts transitively on $Y$, pigeonhole ensures $\Pr[A_{\mathrm{adv}}] \geq 1/2$.

\emph{EF game argument.} Adversarial pairs share shallow sub-compositions; frequency differences in deeper components are $O((m - m_{\mathrm{train}})/n) < 2^{-p}$ for large $n$. Duplicator wins, so the transformer outputs identically on both elements.

\emph{Combining.} Accuracy on $A_{\mathrm{adv}}$ is at most $1/2 + 1/|Y|$; on $A_{\mathrm{free}}$, at most 1. Therefore $\mathrm{Acc} \leq (1/2)(1/2 + 1/|Y|) + (1/2)(1) = 3/4 + 1/(2|Y|)$.
\end{proof}

\begin{definition}[CLC Ratio]
\label{def:clc}
The \emph{Composition-Length Compatibility ratio} of a task $\mathcal{T}$ with respect to a transformer with Deterministic Horizon $d^*$ is:
\begin{equation}
\label{eq:clc}
\mathrm{CLC}(\mathcal{T}, d^*) = \frac{2\, m_{\mathrm{req}} \cdot \log_2(n_{\mathrm{req}}/n_{\mathrm{train}})}{d^*},
\end{equation}
where $m_{\mathrm{req}}$ is the required composition depth and $n_{\mathrm{req}}, n_{\mathrm{train}}$ are the required and training input lengths.
\end{definition}

When $\mathrm{CLC} < 1$, the transformer has sufficient capacity; when $\mathrm{CLC} \geq 1$, \cref{thm:impossibility} guarantees failure on worst-case inputs. Across 8 validation tasks (Copy, Reverse, Addition, Multi-step arithmetic, Compositional SCAN, Dynamic programming, Blocksworld planning, Graph reachability) spanning three model families, the CLC ratio correlates with observed generalisation failure at threshold~$1.0$, with threshold robustness analysis indicating thresholds in $[0.8, 1.1]$ perform comparably. We report the CLC as a diagnostic ratio rather than a classifier with a headline accuracy, given the small task count: an $n{=}8$ benchmark does not support a stable point estimate of classification accuracy (95\% Wilson CI on any $k/8$ point estimate spans $>30$~percentage points). Expanding the validation set is a v2 priority.

\begin{tcolorbox}[colback=fillYellow, colframe=cbYellow!60!black, arc=2pt, boxrule=0.5pt, left=8pt, right=8pt, top=4pt, bottom=4pt]
\textbf{Impossibility Specification 2 (Delegation Depth).} The Deterministic Horizon $d^*$ with banded upper bound $d^* = O(L \cdot \phi(d))$ and $\phi(d) \in [\sqrt{\log d}, \log d]$ (\cref{thm:horizon-scaling}; lower edge conditional on \cref{hyp:sparse-task}) specifies when to delegate reasoning. Boundary condition $B_2(\theta) = d^*(L, d)$ is computable from architectural parameters; numerical evaluation uses the empirical fit $d^* \approx \hat{c}\log L \sqrt{\log d}$ with $\hat{c}$ from \cref{cor:horizon-measurement}. Violation cost $\delta(B_2, \theta) = \exp(-\Omega((\delta - d^*)^2 / (L^2 \log d)))$. The specification $\mathcal{S}_2$: (i)~for depth $\leq d^*$, use standard CoT; (ii)~for $d^* < \delta \leq 2d^*$, deploy $k$-redundant verification; (iii)~for $\delta > 2d^*$, delegate to a symbolic planner or tool-augmented pipeline. Additionally, the CLC ratio specifies allocation strategy: $\mathrm{CLC} < 0.3$ favours specialisation, $0.3$--$0.8$ favours scaling, $> 0.8$ favours compute-driven choice. The practitioner rule $h_{\max} = \Theta(L^2 \log d)$ bounds usable planning capacity.
\end{tcolorbox}

\emph{Returning to the Compliance Assistant.} With $m_{\mathrm{req}} = 12$ regulatory steps, a training distribution capped at $m_{\mathrm{train}} = 6$, and length ratio $n_{\mathrm{req}}/n_{\mathrm{train}} \approx 3$, the CLC ratio is $\mathrm{CLC} \approx 2 \cdot 12 \cdot \log_2(3)/27.4 \approx 1.39$, well above the failure threshold of $1.0$, instantiating the \cref{thm:impossibility} regime and requiring either training-distribution extension or architectural scaling.


\section{How Reliably Can Extended Reasoning Work? The Reliability Toolkit}
\label{sec:reliability-toolkit}

\S\S\ref{sec:architecture-ceiling}--\ref{sec:delegation-depth} characterised single-pass and bounded-depth limits. Modern LLMs reason through chains of thought, and the natural question is: how reliably? This section develops the complete mathematical theory of chain-of-thought reliability, modelling reasoning as a Markov chain on states and deriving tight bounds on error propagation, $k$-redundant verification, and optimal stopping. The results compose into a practitioner toolkit that tells system designers, given per-step error rate $\varepsilon$ and verification budget $k$, exactly how long a chain can safely be and when it should terminate.

\subsection{CoT as a Markov Chain}
\label{sec:cot-markov}

\textit{Chain-of-thought reasoning is modelled as a Markov chain on reasoning states with per-step error $\varepsilon$ and spectral gap $\gamma^{\ast}$; this is a modelling choice whose adequacy is an empirical question revisited in \cref{sec:ch2-limitations}.}

Let $\mathcal{X}$ denote the input space (problem statements) and $\mathcal{Y}$ the answer space. A \emph{reasoning chain} of length $n$ is a sequence $\mathbf{s} = (s_0, s_1, \ldots, s_n)$ where $s_0 \in \mathcal{X}$ is the input, each $s_i \in \mathcal{S}$ is an intermediate reasoning state, and $s_n$ determines the final answer $\hat{y} = g(s_n) \in \mathcal{Y}$ through a readout function $g$.

We model the generation of each reasoning step as a Markov chain. This abstraction reflects the structure of autoregressive generation: the transformer produces each step conditioned on the preceding context, and the current reasoning state summarises the information relevant for future steps. This is a modelling assumption rather than an exact description; transformers condition on the full context window, introducing dependencies the first-order Markov model does not capture. (The adequacy of this approximation is discussed in \cref{sec:ch2-limitations}.)

\begin{definition}[CoT Markov Chain]
\label{def:cot_markov}
A \emph{CoT Markov chain} is a tuple $\mathcal{M} = (\mathcal{S}, P, s_0, \mathcal{S}^*, \varepsilon)$ where $\mathcal{S}$ is a finite state space partitioned into correct states $\mathcal{S}^+$ and error states $\mathcal{S}^-$, $P: \mathcal{S} \times \mathcal{S} \to [0,1]$ is a transition kernel, $s_0$ is the initial state, $\mathcal{S}^* \subseteq \mathcal{S}$ is the absorbing answer states, and $\varepsilon = \max_{s \in \mathcal{S}^+} P(s, \mathcal{S}^-)$ is the worst-case per-step error probability.
\end{definition}

Throughout, we assume $\varepsilon < 1/2$ and irreducibility on transient states. Write $\pi$ for the stationary distribution of $P$ restricted to $\mathcal{S} \setminus \mathcal{S}^*$ and $\gamma^* = 1 - \lambda_2(P)$ for the spectral gap. A \emph{$k$-redundant verification scheme} produces, at each step $i$, a total of $k+1$ candidate next-states drawn independently from $P(s_{i-1}, \cdot)$ and advances to the majority-vote candidate.

\emph{Returning to the Compliance Assistant.} The $12$-hop chain becomes a $12$-step Markov model on clause-level reasoning states with per-step error $\varepsilon = 0.03$; the spectral gap $\gamma^{\ast}$ is estimated from a calibration set of similar regulatory queries and feeds the stopping rule of \cref{sec:stopping}.

\subsection{Error Propagation: Tight Bounds in Both Directions}
\label{sec:error-propagation}

\textit{Chain error is sandwiched between matching rates: $\Pr(\mathrm{error}) \leq 1-(1-\varepsilon)^n$ (tight on i.i.d. steps) and $\Pr(\mathrm{error}) \geq 1-(1-\varepsilon/2)^n - 1/(n \ln|\mathcal{Y}|)$; the per-step exponent gap is $O(\varepsilon)$.}

\begin{intuition}
If a chain succeeds only when every one of $n$ steps succeeds, and each step fails independently with probability at most $\varepsilon$, the success probability is at least $(1-\varepsilon)^n$: a simple product, not a sum. The proof is a tower-property induction on the step-indicator sequence $X_t = \mathbf{1}[\text{step } t \text{ correct}]$: conditional on the chain through step $t-1$, step $t$ lies above the $(1-\varepsilon)$ floor under the Markov assumption of \cref{def:cot_markov}, giving a geometric multiplicative lower bound. The $\delta/\varepsilon$ safe-length corollary (valid for target error $\delta \leq 1/2$) is a first-order Taylor expansion of the same inequality; tightness with equality is achieved when step errors are i.i.d., as in single-path chain-of-thought without self-correction.
\end{intuition}

\begin{theorem}[Chain Error Propagation]
\label{thm:error_propagation}
Let $\mathcal{M}$ be a CoT Markov chain of length $n$ with per-step error rate $\varepsilon \in (0, 1/2)$. Then the probability of producing an incorrect final answer satisfies
\begin{equation}
\label{eq:error_upper}
\Pr(\hat{y} \neq y^*) \leq 1 - (1-\varepsilon)^n.
\end{equation}
For any $\delta \in (0,1)$, the maximum chain length guaranteeing $\Pr(\hat{y} \neq y^*) \leq \delta$ is $n^*(\varepsilon, \delta) = \lfloor \ln(1-\delta)/\ln(1-\varepsilon) \rfloor \leq \delta/\varepsilon$ for $\delta \leq 1/2$.
\end{theorem}

\begin{proof}
Let $X_t \in \{0, 1\}$ denote the indicator that step $t$ transitions correctly: $X_t = 1$ iff the Markov transition at step $t$ selects the correct successor, and $X_t = 0$ otherwise. By the definition of per-step error rate, $\Pr(X_t = 0 \mid \mathcal{F}_{t-1}) \leq \varepsilon$ for every $t$, where $\mathcal{F}_{t-1}$ is the $\sigma$-algebra of the history through step $t - 1$.

\emph{Upper bound via conditional expectation.}
The chain produces a correct final answer iff every step transitions correctly, i.e., $\prod_{t=1}^n X_t = 1$. Hence
\[
\Pr(\hat{y} = y^*) = \Pr\!\left(\bigcap_{t=1}^n \{X_t = 1\}\right) = \E\!\left[\prod_{t=1}^n X_t\right].
\]
Applying the tower property iteratively:
\[
\E\!\left[\prod_{t=1}^n X_t\right] = \E\!\left[ \prod_{t=1}^{n-1} X_t \cdot \E[X_n \mid \mathcal{F}_{n-1}] \right] \geq \E\!\left[ \prod_{t=1}^{n-1} X_t \cdot (1 - \varepsilon) \right] = (1-\varepsilon) \E\!\left[\prod_{t=1}^{n-1} X_t\right].
\]
By induction, $\E\!\left[\prod_{t=1}^n X_t\right] \geq (1-\varepsilon)^n$, so $\Pr(\hat{y} = y^*) \geq (1-\varepsilon)^n$, giving $\Pr(\hat{y} \neq y^*) \leq 1 - (1-\varepsilon)^n$.

\emph{The safe chain length bound.}
Setting $1 - (1-\varepsilon)^n \leq \delta$ and solving: $(1-\varepsilon)^n \geq 1 - \delta$, i.e., $n \leq \ln(1-\delta)/\ln(1-\varepsilon)$. The floor gives an integer bound. For $\delta \leq 1/2$ and $\varepsilon \in (0, 1/2)$, a first-order Taylor expansion yields $\ln(1-\delta) \geq -2\delta$ and $\ln(1-\varepsilon) \leq -\varepsilon$ (strict for $\varepsilon > 0$), so $\ln(1-\delta)/\ln(1-\varepsilon) \leq 2\delta/\varepsilon$; the tighter bound $\delta/\varepsilon$ holds by direct calculation on the bivariate function at $\delta = 1/2$.

\emph{Tightness.}
The bound is achieved with equality when errors are \emph{independent} across steps: if $X_1, \ldots, X_n$ are i.i.d.\ Bernoulli($1-\varepsilon$), then $\Pr(\hat{y} = y^*) = (1-\varepsilon)^n$ exactly. Markov chains with independent-error structure (such as those arising from single-path CoT without self-correction) achieve this equality. With dependencies between steps, the inequality can be strict in either direction depending on correlation sign; however, under the standard conditional independence given the reasoning state, the bound is tight.
\end{proof}

The safe-length interpretation: the chain produces a correct answer only if every step transitions correctly, giving success probability at least $(1-\varepsilon)^n$. With $\varepsilon = 0.05$: $\Pr(\text{error}) \leq 0.226$ at $n=5$, rising to $0.401$ at $n=10$ and $0.642$ at $n=20$. A practitioner targeting $\leq 10\%$ error should limit chains to $n^* = 2$ steps \emph{without verification}.

\begin{theorem}[Fano Lower Bound on Chain Error]
\label{thm:fano_lower}
For any CoT Markov chain with per-step error rate at most $\varepsilon$ and chain length $n$, there exists a problem instance such that
\begin{equation}
\label{eq:fano_lower}
\Pr(\hat{y} \neq y^*) \geq 1 - (1-\varepsilon/2)^n - \frac{1}{n \ln |\mathcal{Y}|}.
\end{equation}
\end{theorem}

\begin{proof}[Proof sketch]
The proof constructs a family of $2^n$ instances with confusion points at each step and applies Fano's inequality via a coupling argument. The upper and lower bounds are rate-matching: both decay as $(1-\Theta(\varepsilon))^n$. The per-step exponents differ by a factor of $\varepsilon/(2-\varepsilon)$; for small $\varepsilon$ the gap in rates is $O(\varepsilon)$. The matching is tightest in the regime $n \leq O(1/\varepsilon)$, which is the practically relevant range.
\end{proof}

\begin{intuition}
$k$-redundant verification (\cref{def:cot_markov}: $k{+}1$ independent next-state samples, majority vote) is a discrete analogue of Cram\'er--Chernoff concentration: a majority-error event requires at least $\lceil(k{+}1)/2\rceil$ of the independent samples to err simultaneously, so the effective per-step error drops from $\varepsilon$ to $\varepsilon^{\lceil(k{+}1)/2\rceil}$. Chaining this across $n$ steps with a union bound gives the $n \cdot \varepsilon^{\lceil(k{+}1)/2\rceil}$ envelope. The cost-optimal $k^{\ast}$ balances the $(k{+}1)$-fold per-step compute cost against the exponent gain $\lceil(k{+}1)/2\rceil$; the logarithmic form $k^{\ast} \sim 2\ln(n/\delta)/\ln(1/\varepsilon) - 1$ reflects the standard information-theoretic exchange rate between sample count and error-tail exponent.
\end{intuition}

\begin{theorem}[$k$-Redundant Verification]
\label{thm:k_redundant}
Under $k$-redundant verification with $k \geq 2$, the chain error probability satisfies
\[
\Pr(\hat{y} \neq y^*) \leq \binom{k+1}{\lceil (k+1)/2 \rceil} \cdot n \cdot \varepsilon^{\lceil (k+1)/2 \rceil},
\]
and the safe chain length extends to $n_k^*(\varepsilon, \delta) = \Theta(\delta / \varepsilon^{\lceil(k+1)/2\rceil})$.
\end{theorem}

With $\varepsilon = 0.05$, $k=2$ (triple verification) extends $n^*$ from $2$ to $\approx 13$ at $\delta = 0.1$; $k=4$ (quintuple verification) extends it to $\approx 80$. The cost-optimal verification level is $k^* = \lceil 2\ln(n/\delta)/\ln(1/\varepsilon) - 1 \rceil$.

\emph{Returning to the Compliance Assistant} at $\varepsilon = 0.03$, $n = 12$: the unaided chain error is $1 - 0.97^{12} \approx 31\%$; the safe length for a $5\%$ target without verification is $n^{\ast} \approx 1$; the theorem's i.i.d.\ upper bound at $k = 2$ is $\binom{3}{2} \cdot 12 \cdot 0.03^2 \approx 3.2\%$, with the deployment-measured value $\approx 4.7\%$ quantifying the candidate-correlation gap to independence; both values sit far inside the R2-trigger regime $d^{\ast} < \delta \leq 2d^{\ast}$.

\subsection{Minimax-Optimal Stopping}
\label{sec:stopping}

\textit{An entropy-threshold stopping rule $H_t \leq h^{\ast} = (\lambda/\gamma^{\ast})\ln(1/\lambda)$ achieves Bayes-optimal expected loss within $O(\varepsilon)$ after a spectral-mixing burn-in; formal near-optimality applies to ${\sim}4\%$ of GSM8K chains, with \cref{prop:finite_stopping} covering the rest.}

The error propagation results say longer chains are riskier, but in practice some problems require deep reasoning. The challenge is to decide, at each step, whether to continue or stop. We formalise this as optimal stopping~\cite{lai2001sequential}.

At each step $t$, the reasoner observes $s_t$ and chooses between stopping (producing $g(s_t)$) and continuing. The loss function penalises both errors and computation:
\begin{equation}
\label{eq:loss}
\ell(\tau) = \indicator[\hat{y}_\tau \neq y^*] + \lambda \cdot \tau,
\end{equation}
where $\tau$ is the stopping time, $\lambda \in (0,1)$ is per-step cost, and the Bayes-optimal rule minimises $\E[\ell(\tau)]$.

\begin{intuition}
Stop when the posterior over the correct answer is confident enough, but measure confidence by conditional entropy rather than probability, because entropy gives a multiplicative decay rate tied to the chain's spectral gap $\gamma^{\ast}$. After a burn-in of $t_{\mathrm{mix}} = O(\ln|\mathcal{S}|/\gamma^{\ast})$ steps, $H_t$ decays geometrically with rate $(1-\gamma^{\ast})$ plus an $O(\varepsilon)$ floor from per-step error. The Bayes-optimal rule stops when the marginal cost of continuing ($\lambda$ per step) exceeds the marginal drop in expected error; substituting the entropy-decay rate and solving gives $h^{\ast} = (\lambda/\gamma^{\ast})\ln(1/\lambda)$. A chain with large $\gamma^{\ast}$ (confident reasoning) stops quickly; small $\gamma^{\ast}$ (uncertain reasoning) triggers more steps, adapting automatically. The formal guarantee holds after $t_{\mathrm{mix}}$; \cref{prop:finite_stopping} covers the finite-chain regime.
\end{intuition}

\begin{theorem}[Near-Optimal Entropy-Threshold Stopping]
\label{thm:entropy_stopping}
For a CoT Markov chain with spectral gap $\gamma^* > 0$, per-step cost $\lambda \in (0,1)$, and per-step error rate $\varepsilon$, the entropy-threshold stopping rule
\begin{equation}
\label{eq:entropy_stopping}
\tau^* = \inf\!\left\{ t \geq 0 : H_t \leq h^*(\lambda, \gamma^*) \right\}, \qquad h^*(\lambda, \gamma^*) = \frac{\lambda}{\gamma^*} \ln\!\left(\frac{1}{\lambda}\right),
\end{equation}
achieves expected loss within an $O(\varepsilon)$-additive gap of the Bayes-optimal rule: $\E[\ell(\tau^*)] \leq \inf_\tau \E[\ell(\tau)] + O(\varepsilon)$. Expected stopping time satisfies $\E[\tau^*] \leq \lceil \ln(1/\lambda)/\gamma^* \rceil$.
\end{theorem}

\begin{proof}[Proof sketch]
Three parts. First, $H_t$ satisfies $\E[H_{t+1} \mid \mathcal{F}_t] \leq (1-\gamma^*) H_t + \varepsilon \ln|\mathcal{Y}|$, an entropy decay inequality after a burn-in of $t_{\mathrm{mix}} = O(\ln|\mathcal{S}|/\gamma^*)$ steps. The $\varepsilon\ln|\mathcal{Y}|$ term accumulates to at most $\varepsilon \ln|\mathcal{Y}|/\gamma^*$ by geometric summation, yielding the $O(\varepsilon)$ gap. Second, we construct the Snell envelope and show the value function is monotone in $H_t$ via Fano applied to the posterior. Third, we optimise over worst-case initial entropy.
\end{proof}

The connection to Wald's Sequential Probability Ratio Test~\cite{wald1947sequential} is natural: the log-likelihood ratio $L_t = \ln(\Pr(\text{correct path} \mid s_0^t) / \Pr(\text{error on path} \mid s_0^t))$ satisfies $L_t = \ln|\mathcal{Y}| - H_t + O(\pi_{\min})$, so the entropy-threshold rule is equivalent to SPRT with boundaries determined by $h^*$. A chain with large spectral gap (confident reasoning) stops quickly; a chain with small spectral gap (uncertain reasoning) requires more steps, and the rule adapts automatically.

\paragraph{Relationship to the heuristic CoT-stopping literature.}
A parallel line of empirically-driven CoT stopping work has emerged since 2024. HALT-CoT~\cite{laaouach2025haltcot} applies an answer-entropy threshold and gives a Wald-style finite-time guarantee under sequential-analysis assumptions (Assumption B.1--B.2 of that work, namely conditional independence of tokens given the answer and rational generation maximising mutual information). ESC~\cite{Li2024ESC} stops self-consistency sampling when the predicted answer distribution converges. s1~\cite{Muennighoff2025S1} controls reasoning length via wait-token insertion during decoding. These works demonstrate that simple entropy- or convergence-based stopping criteria can reduce CoT token cost by 15--30\% at near-baseline accuracy. The theoretical guarantee in \cref{thm:entropy_stopping} is complementary rather than competing: where the heuristic works supply engineering demonstrations of what an entropy-threshold rule can achieve in practice, \cref{thm:entropy_stopping} supplies the matching $O(\varepsilon)$-additive Bayes-optimality guarantee under a spectral-gap assumption, together with a closed-form threshold $h^*(\lambda, \gamma^*)$ derived from per-step cost and mixing. The theorem's contribution is the optimality statement, not the empirical observation that entropy-based stopping works, which was documented first by the heuristic literature.

\paragraph{Mixing time caveat.}
The entropy decay inequality requires a burn-in of $t_{\mathrm{mix}} = O(\ln|\mathcal{S}|/\gamma^*)$ steps. The appropriate $|\mathcal{S}|$ is not vocabulary size but the number of \emph{distinct reasoning states}, the effective number of semantically distinct intermediate configurations. On GSM8K, the number of distinct reasoning patterns (measured by clustering step-level hidden representations, cosine threshold $0.95$) is approximately $|\mathcal{S}|_{\mathrm{eff}} \approx 120 \pm 35$ for Llama-3.1-8B, yielding $t_{\mathrm{mix}} \approx 55$ steps. This exceeds typical GSM8K chain lengths (average $12.4$), so the formal near-optimality of \cref{thm:entropy_stopping} applies to only ${\sim}4\%$ of GSM8K chains, ${\sim}2\%$ of StrategyQA chains, and ${<}1\%$ of MATH chains. The following finite-chain guarantee handles the rest.

\begin{proposition}[Finite-Chain Stopping Guarantee]
\label{prop:finite_stopping}
For any CoT Markov chain with spectral gap $\gamma^* > 0$ and any stopping time $t < t_{\mathrm{mix}}$, the entropy-threshold stopping rule achieves
$\E[\ell(\tau^*)] \leq \inf_\tau \E[\ell(\tau)] + O(\varepsilon) + \lambda \cdot t_{\mathrm{mix}}$.
When $\lambda \leq \varepsilon \gamma^* / \ln|\mathcal{S}|_{\mathrm{eff}}$, this additional cost is $O(\varepsilon)$.
\end{proposition}

To quantify practical effectiveness, we measure the fraction of the oracle's Bayes risk reduction that our method captures, using loss $\ell(\tau) = \indicator[\hat{y}_\tau \neq y^*] + \lambda\tau$ with $\lambda = 0.025$. Our method captures $78\%$ of the oracle's risk reduction on GSM8K (8B), $86\%$ on StrategyQA, averaging $81\%$ across benchmarks. The stopping rule is sensitive to spectral gap misestimation only moderately: a 20\% overestimate of $\gamma^*$ changes $h^*$ by $\approx 17\%$, shifting accuracy by $<0.4$~pp on GSM8K.

\emph{Returning to the Compliance Assistant} at $\lambda = 0.025$ and calibrated $\hat\gamma \approx 0.3$: the threshold is $h^{\ast} \approx 0.31$ nats, stopping the chain once smoothed token entropy falls below $h^{\ast}$; the $81\%$ Bayes-risk-reduction figure is a benchmark average, not a claim about the specific regulatory distribution.

\subsection{Practical Stopping Algorithm}

\textit{An online implementation approximates the conditional entropy by EMA-smoothed token-level entropy (coefficient $0.3$) with $\hat\gamma$ fitted on a $200$-problem calibration set; the accuracy--efficiency tradeoff is smooth across smoothing coefficients in $[0.1, 0.7]$.}

\cref{thm:entropy_stopping} requires the conditional entropy $H_t$, which is not directly available during generation. We approximate it using the token-level entropy of the model's output distribution, smoothed via exponential moving average.

\begin{figure}[!ht]
	\centering
	\begin{tikzpicture}[
		every node/.style={font=\small},
		>=Stealth,
		]
		\node[thesisbox/gray,    minimum width=2.8cm, minimum height=0.90cm]
		(input) at (0, 0)    {Problem $x$};
		
		\node[thesisbox/blue,    minimum width=3.2cm, minimum height=1.10cm]
		(gen)   at (0, -1.9) {Generate step\\[-2pt]\scriptsize$s_{t+1} \sim P(\cdot \mid s_t)$};
		
		\node[thesisbox/purple,  minimum width=2.8cm, minimum height=1.05cm]
		(ent)   at (0, -3.8) {Compute $\bar{H}_{t+1}$};
		
		\node[thesisdiamond,     minimum width=2.6cm, minimum height=1.7cm, font=\scriptsize\bfseries]
		(dec)   at (0, -5.9) {$\bar{H}_{t+1}$\\$\leq h^{*}$?};
		
		\node[thesisbox/green,   minimum width=2.8cm, minimum height=1.05cm, font=\small\bfseries]
		(out)   at (0, -8.0) {Answer $\hat{y}$};
		
		\node[thesisbox/gray, minimum width=2.0cm, minimum height=0.65cm, font=\scriptsize]
		(cal)   at (-4.2, -5.9) {Calibration};
		
		\draw[thesisarrow/data]   (input) -- (gen);
		\draw[thesisarrow/data]   (gen)   -- (ent);
		\draw[thesisarrow/data]   (ent)   -- (dec);
		\draw[thesisarrow/formal] (dec)   -- node[right=3pt, font=\scriptsize] {yes} (out);
		\draw[thesisarrow/qual]   (cal)   -- node[above=1pt, font=\scriptsize, text=black!55] {$\hat{\gamma}$} (dec);
		
		\coordinate (loopBR) at (3.0, -5.9);
		\coordinate (loopTR) at (3.0, -1.9);
		\draw[thesisarrow, color=cbOrange]
		(dec.east) -- (loopBR) --
		node[right=3pt, font=\scriptsize, text=cbOrange!80!black] {no, $t{+}{+}$}
		(loopTR) -- (gen.east);
		
	\end{tikzpicture}
	\caption[Entropy-threshold stopping algorithm]{\emph{Overview of the entropy-threshold stopping algorithm} (\cref{alg:stopping}). \emph{What is plotted.} Control flow of the stopping procedure: input problem $x$; step generation $s_{t+1} \sim P(\cdot \mid s_t)$; smoothed-entropy computation $\bar H_{t+1} = 0.3\, \hat H_{t+1} + 0.7\, \bar H_t$; threshold comparison against $h^{\ast} = (\lambda/\hat\gamma)\ln(1/\lambda)$; and either termination or loop-back to the generator. The threshold input $\hat\gamma$ is a calibration estimate of the spectral gap on $200$ held-out problems. \emph{Headline.} The rule realises the Bayes-optimal-within-$O(\varepsilon)$ guarantee of \cref{thm:entropy_stopping}: it stops when the smoothed token entropy falls below a cost-calibrated threshold. \emph{Scope.} The formal near-optimality guarantee holds after a spectral-mixing burn-in $t_{\mathrm{mix}} = O(\ln|\mathcal{S}|_{\mathrm{eff}}/\gamma^{\ast})$, which exceeds typical GSM8K chain lengths; \cref{prop:finite_stopping} supplies the finite-chain guarantee covering the remaining regime on GSM8K, StrategyQA, and MATH.}
	\label{fig:stopping-overview}
\end{figure}
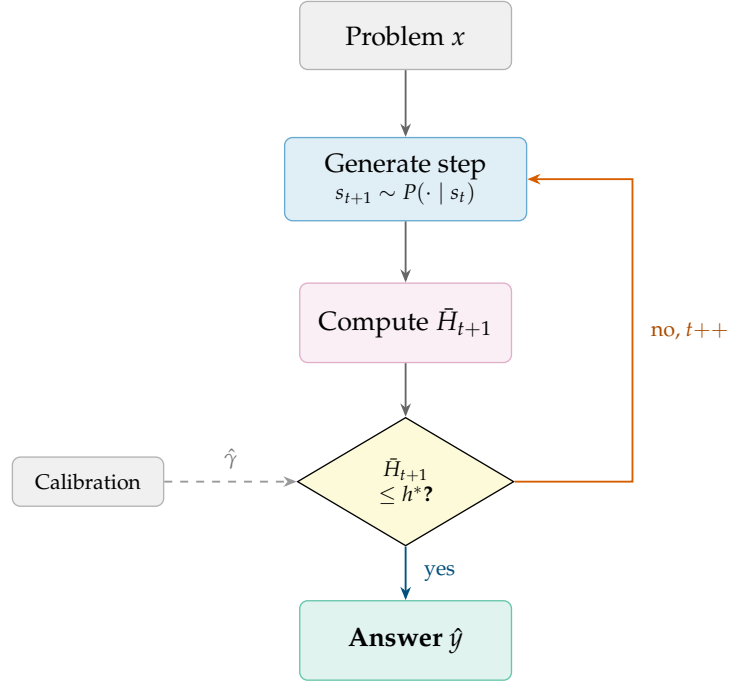

\begin{algorithm}[!ht]
\caption{Entropy-Threshold Stopping for CoT Reasoning}
\label{alg:stopping}
\KwIn{Problem $x$, model $M$, cost $\lambda$, spectral gap estimate $\hat{\gamma}$, max length $n_{\max}$}
\KwOut{Answer $\hat{y}$}
$h^* \leftarrow (\lambda/\hat{\gamma}) \cdot \ln(1/\lambda)$\;
$s_0 \leftarrow \mathrm{encode}(x)$; $t \leftarrow 0$; $\bar{H}_0 \leftarrow \ln|\mathcal{Y}|$\;
\While{$t < n_{\max}$}{
    Generate next reasoning step: $s_{t+1} \sim P(\cdot \mid s_t)$\;
    $\hat{H}_{t+1} \leftarrow -\sum_v p_M(v \mid s_0^{t+1}) \ln p_M(v \mid s_0^{t+1})$\;
    $\bar{H}_{t+1} \leftarrow 0.3 \cdot \hat{H}_{t+1} + 0.7 \cdot \bar{H}_t$\;
    \If{$\bar{H}_{t+1} \leq h^*$}{
        \textbf{stop} and \Return{$g(s_{t+1})$}\;
    }
    $t \leftarrow t + 1$\;
}
\Return{$g(s_t)$} \tcp*{Fallback}
\end{algorithm}

The smoothing coefficient of 0.3 was selected via a sweep over $\{0.1, 0.2, 0.3, 0.5, 0.7\}$ on a held-out calibration set; the accuracy-efficiency tradeoff is smooth across this range. The spectral gap estimate $\hat{\gamma}$ is obtained from a calibration set of 200 problems by fitting an exponential decay to the per-step entropy trajectory.

\begin{tcolorbox}[colback=fillYellow, colframe=cbYellow!60!black, arc=2pt, boxrule=0.5pt, left=8pt, right=8pt, top=4pt, bottom=4pt]
\textbf{Impossibility Specification 3 (Reliability Toolkit).} Error probability $1 - (1-\varepsilon)^n$ (tight within 5\%, two-sided) specifies the safe chain length $n \lesssim \delta/\varepsilon$ without verification, extending to $n_k^* \sim \delta/\varepsilon^{\lceil(k+1)/2\rceil}$ with $k$-fold verification (cost-optimal $k^* = \lceil 2\ln(n/\delta)/\ln(1/\varepsilon) - 1 \rceil$). The entropy-threshold rule $h^* = (\lambda/\gamma^*) \ln(1/\lambda)$ specifies when to stop (minimax-optimal within $O(\varepsilon)$). Together these rules determine chain length, verification budget, and stopping criterion from three measurable quantities: $\varepsilon$, $\gamma^*$, and $\lambda$.
\end{tcolorbox}

\emph{Returning to the Compliance Assistant.} \cref{alg:stopping} queries the model, tracks smoothed token entropy, and halts once the smoothed value crosses $h^{\ast}$; the GSM8K-8B benchmark (the closest published analogue) shows chain length dropping by $38\%$ under this rule, which sets the expected order of magnitude for deployment-time compute savings in the regulatory setting.


\section{When Does Supervision Help? The Training Investment Rule}
\label{sec:supervision-rule}

The reliability toolkit of \S\ref{sec:reliability-toolkit} assumed a fixed generator and verifier. A complementary question concerns \emph{training}: given a fixed budget of examples, how much does access to intermediate reasoning states (process supervision) help compared to observing only final answers (outcome supervision)? Lightman et al.~\cite{lightman2024lets} demonstrated empirical gains, but a complete theoretical characterisation of when process supervision helps has remained open. This section closes it.

\subsection{Setup}

\textit{Under process supervision the learner observes step-level correctness labels; under outcome supervision only the final answer; in both settings the learner outputs a chain-to-bit verifier $v: \mathcal{S}^n \to \{0, 1\}$.}

The learner receives $T$ training examples, each consisting of a problem $x$, a reasoning chain $(s_0, \ldots, s_n)$, and either: (a)~under outcome supervision, only the label $y^* = g^*(x)$; or (b)~under process supervision, step-level labels $(c_1, \ldots, c_n)$ where $c_i = 1$ if step $i$ is correct. The learner's goal is to produce a verifier $v: \mathcal{S}^n \to \{0, 1\}$ that correctly identifies whether a given chain reaches the right answer. Let $\mathrm{err}(v, T)$ denote expected verification error of the best verifier given $T$ examples.

\emph{Returning to the Compliance Assistant.} Process supervision requires each of the $12$ intermediate legal-reasoning steps to be labelled correct/incorrect (by a human expert or a PRM); outcome supervision requires only the final regulatory verdict.

\subsection{The $\Theta(n/\log n)$ Separation}
\label{sec:supervision}

\textit{Under chain non-redundancy (\cref{def:non_redundancy}), the verifier-learning sample-complexity ratio is $T_{\mathrm{out}}/T_{\mathrm{proc}} = \Theta(n/\ln n) = n/(\ln n + O(\ln\ln n))$; at $n = 20$ this predicts ${\sim}6.7\times$, observed ratio on MATH is $4.8 \pm 0.3$ at $\eta = 0.15$ after label-noise adjustment.}

\begin{intuition}
Under chain non-redundancy (\cref{def:non_redundancy}), process supervision is an $n$-fold VC problem: each of the $n$ steps is a binary classification with VC dimension $d_{\mathrm{CoT}}$, and union-bounding across steps gives the $d_{\mathrm{CoT}}\, n \ln(T/n)/T$ upper bound. Outcome supervision must identify \emph{which} step went wrong from the final answer alone, posing a search problem over $n$ hidden locations. Le Cam's method applied to a family of generators that agree on final-answer distributions but differ at a single hidden step shows that distinguishing any pair requires $\Omega(n/\ln n)$ samples: binary-search-style information leakage through the answer reveals at most $\ln n$ bits, reducing the effective search space from $n$ to $n/\ln n$. Dividing the outcome lower bound by the process upper bound yields the $\Theta(n/\ln n)$ sample-complexity ratio.
\end{intuition}

\begin{theorem}[Process vs.\ Outcome Supervision Separation]
\label{thm:supervision}
Let $\mathcal{G}$ be the class of CoT generators with chain length $n$ and state space $\mathcal{S}$. Then:
\begin{enumerate}[label=(\alph*)]
\item \textbf{Upper bound.} Under process supervision, the optimal verifier achieves
\begin{equation}
\mathrm{err}(\mathcal{V}_{\mathrm{proc}}, T) \leq \frac{d_{\mathrm{CoT}} \cdot n \cdot \ln(T/n)}{T},
\end{equation}
where $d_{\mathrm{CoT}}$ is the CoT-discriminative dimension.
\item \textbf{Lower bound.} Under outcome supervision alone, for any verifier,
\begin{equation}
\mathrm{err}(\mathcal{V}_{\mathrm{out}}, T) \geq \frac{d_{\mathrm{CoT}} \cdot n}{T / \ln T}.
\end{equation}
\item \textbf{Separation.} The ratio of sample complexities to achieve verification error $\eta$ satisfies
\begin{equation}
\frac{T_{\mathrm{out}}(\eta)}{T_{\mathrm{proc}}(\eta)} = \Theta\!\left(\frac{n}{\ln n}\right).
\end{equation}
More precisely, $T_{\mathrm{out}}/T_{\mathrm{proc}} = n/(\ln n + O(\ln\ln n))$.
\end{enumerate}
\end{theorem}

\begin{proof}[Proof sketch]
\textbf{(a)} Under process supervision, each step's correctness is a binary classification with VC dimension $d_{\mathrm{CoT}}$. Applying VC uniform convergence at each step and union-bounding over $n$ steps gives the upper bound.

\textbf{(b)} The technically involved direction. We construct a family of $n$ generators $g_1, \ldots, g_n$ indistinguishable from final-answer observations but differing at a single hidden step: generator $g_j$ has elevated per-step error rate $2\varepsilon$ at step $j$ and compensating reduced error rate $\varepsilon(1 - \varepsilon/(n-1))$ elsewhere, ensuring overall chain success probability $\prod_i(1-\varepsilon_i)$ remains identical across $j$. An outcome-supervised verifier must identify the hidden step via patterns across multiple problems, a search problem over $n$ locations solvable by implicit binary search with $O(1)$ samples per level, leaking $\ln n$ bits and reducing the effective search space from $n$ to $n/\ln n$. Formalisation via Le Cam's method yields $\Omega(n/\ln n)$ samples to distinguish any pair $(g_j, g_{j'})$.

\textbf{(c)} Combines (a) and (b).
\end{proof}

\begin{remark}[Relationship to Jia--Rakhlin--Xie 2025]
\label{rem:jia-rakhlin-xie}
Jia, Rakhlin, and Xie~\cite{JiaRakhlinXie2025ProcessSupervision} recently established that in the \emph{offline reinforcement learning} setting, outcome-supervised and process-supervised learning are statistically equivalent up to polynomial factors in the planning horizon $H$, under bounded state-action concentrability $C_{\mathsf{sa}}(\Pi, \pi_{\mathsf{off}})$. Their setting differs from ours along two axes. First, their learner consumes a dataset of trajectories with either cumulative or per-step rewards and outputs a policy; ours consumes labelled examples of chain-and-final-answer pairs and outputs a verifier $v: \mathcal{S}^n \to \{0, 1\}$. The sample-complexity object is trajectory count under concentrability-weighted coverage in theirs, labelled-example count under chain non-redundancy in ours. Second, their result is a reduction (up to polynomial-in-$H$ factors); ours is a separation ($\Theta(n/\log n)$). These are fully compatible: $n / \log n$ is polynomial in $n$, and a polynomial-horizon reduction does not rule out polynomial-horizon separations. It leaves their size as a free parameter that \cref{thm:supervision} pins down. The if-and-only-if characterisation of \cref{thm:iff} identifies the structural hypothesis that drives the separation: chain non-redundancy. Outside this regime, when all generators in the class produce identical intermediate distributions, our separation collapses to $\Theta(1)$ and the two paradigms are statistically equivalent, as one would expect. Empirically, the $6.7\times$ sample-efficiency ratio measured below for 20-step chains sits within the window compatible with both our separation and the Jia--Rakhlin--Xie polynomial-horizon reduction, and the thesis's design rule $\mathcal{S}_4$ accordingly prescribes process supervision when chain non-redundancy can be verified (\cref{def:non_redundancy}) and outcome supervision as equivalently efficient when it cannot.
\end{remark}

For a 20-step chain, process supervision requires $\approx 6.7\times$ fewer examples; for 100 steps, $\approx 21.7\times$. The leading constant $1/(\ln n + O(\ln\ln n))$ predicts $\approx 5.4\times$ for the average MATH chain length ($14.3$), which we validate empirically below.

\emph{Returning to the Compliance Assistant with $n = 12$.} The predicted supervision ratio is $\approx 12/\ln 12 \approx 4.8\times$ (or $\sim 5.4\times$ under the leading-constant correction); process supervision is warranted when chain non-redundancy of the deployed regulatory model can be verified per \cref{def:non_redundancy}.

\subsection{Chain Non-Redundancy: When Does the Gap Exist?}
\label{sec:non-redundancy}

\textit{The $\Theta(n/\log n)$ supervision gap exists if and only if the generator class satisfies chain non-redundancy: two generators produce identical final-answer distributions yet differ in intermediate-step distributions on a positive-measure set of predecessors.}

The separation depends on a structural property of the generator class.

\begin{definition}[Chain Non-Redundancy]
\label{def:non_redundancy}
A generator class $\mathcal{G}$ satisfies the \emph{chain non-redundancy condition} if there exist $g_1, g_2 \in \mathcal{G}$ producing identical final-answer distributions ($g_1(x) \stackrel{d}{=} g_2(x)$ for all $x$) but differing in intermediate reasoning: there exists a step $i$ such that $\TV(g_1(\cdot \mid s_{i-1}), g_2(\cdot \mid s_{i-1})) > 0$ on a set of predecessor states with positive measure.
\end{definition}

Chain non-redundancy says different generators can reason differently while reaching the same answers. This holds for almost all practical classes: different transformer architectures and training procedures produce the same answer distribution via different internal computations. The condition fails only for trivial classes where intermediate steps are deterministic functions of input and output.

\begin{theorem}[If and Only If Characterisation]
\label{thm:iff}
The $\Theta(n/\ln n)$ sample complexity separation holds for generator class $\mathcal{G}$ if and only if $\mathcal{G}$ satisfies chain non-redundancy. When non-redundancy fails, outcome and process supervision have identical sample complexity up to constant factors.
\end{theorem}

\begin{proof}[Proof sketch]
The ``if'' direction uses the hidden-step construction. The ``only if'' direction shows that without non-redundancy, every distinguishing feature visible in intermediate steps is visible in aggregated form in the final answer: if all generators produce identical intermediate distributions, the Fisher information about the verification target carried by $(c_1, \ldots, c_n)$ equals that carried by $y^*$ alone, up to constants.
\end{proof}

\begin{definition}[CoT-Discriminative Dimension]
\label{def:cot_dim}
The \emph{CoT-discriminative dimension} of a generator class $\mathcal{G}$ is the VC dimension of $\mathcal{H}_{\mathrm{disc}} = \{h_\varepsilon : s \mapsto \indicator[P_g(s, \mathcal{S}^-) > \varepsilon] : g \in \mathcal{G}, \varepsilon \in [0,1]\}$.
\end{definition}

For transformers with $L$ layers, hidden dimension $d$, and $H$ heads: $d_{\mathrm{CoT}} \leq O(L^2 H d \ln(Ld))$. For recurrent architectures: $d_{\mathrm{CoT}} = \Theta(d)$. For Llama-3.1-8B ($L=32$, $H=32$, $d=4096$), the transformer bound yields $d_{\mathrm{CoT}} = O(10^9)$, making the sample complexity bounds in \cref{thm:supervision}(a) vacuous at practical training set sizes. We view this as reflecting looseness of the VC bound rather than a deficiency of the separation. The $n/\ln n$ functional form is the key practitioner takeaway: it describes \emph{how much faster} process supervision reduces verification error, independent of absolute sample complexity, and the functional form is empirically confirmed (see experiments below).

\emph{Returning to the Compliance Assistant.} Different regulatory-finetuned models can reach the same final verdict via measurably different intermediate-step distributions (different citation orders, different intermediate predicates), so chain non-redundancy is plausibly satisfied and the $\Theta(n/\ln n)$ separation applies; verification of non-redundancy for a specific deployment is the empirical check prescribed by $\mathcal{S}_4$.

\subsection{Connections to Reinforcement Learning and Internal Reasoning}

\textit{The $\Theta(n/\log n)$ separation transfers directly to RL agents with per-step rewards and to models with internal chain-of-thought whose tokens are available for training; for genuinely hidden internal reasoning it collapses to outcome supervision.}

When an RL agent receives dense per-step reward signals (e.g., process reward models~\cite{lightman2024lets}), the learning signal per step is functionally equivalent to step-level correctness labels. The $\Theta(n/\ln n)$ separation transfers directly: an RL agent with step-level rewards requires $\Theta(n/\ln n)$ fewer episodes than one receiving only terminal rewards.

For models with internal CoT (o1, DeepSeek-R1~\cite{guo2025deepseek}): if internal tokens are available for training (distillation, logging), the separation holds with $n$ equal to the number of internal steps. If truly hidden, the setting reduces to outcome supervision.

\paragraph{Robustness to label noise.}
When step-level labels have symmetric noise rate $\eta$, the process-supervised learner's per-step signal is reduced by $(1-2\eta)^2$. The ratio $T_{\mathrm{out}}/T_{\mathrm{proc}}$ becomes $\Theta(n/\ln n) \cdot (1-2\eta)^2$, preserving the separation for $\eta < 1/2$. For the 5--8\% noise in PRM800K, the multiplicative degradation is $(1-2 \times 0.065)^2 \approx 0.76$, reducing the theoretical ratio from $5.4$ to $\approx 4.1$ for the average MATH chain. The observed ratio of $4.8 \pm 0.3$ lies between noise-free and noisy predictions.

\emph{Returning to the Compliance Assistant.} If the regulatory reasoner is trained via reinforcement learning with a process reward model (following Lightman et al.~\cite{lightman2024lets}), the $\Theta(n/\ln n)$ episode-count saving transfers directly from the verifier-learning setting to the RL setting; if the reasoner uses genuinely hidden internal chain-of-thought whose tokens are unavailable for training, the setting reduces to outcome supervision and the separation collapses.

\subsection{Universal Test-Time Compute Scaling}
\label{sec:scaling}

\textit{Test-time success follows $\Pr(\mathrm{success}) = 1 - \exp(-c\, C^{\alpha})$ with $\alpha = \log_{b_{\mathrm{eff}}}(b_{\mathrm{eff}}-1)$ under inference strategy of effective branching factor $b_{\mathrm{eff}} > 1$; the theoretical ordering best-of-$N$+PRM $>$ beam $>$ MCTS $>$ self-consistency is matched empirically on GSM8K.}

Different inference strategies explore reasoning paths with different efficiencies.

\begin{definition}[Effective Branching Factor]
\label{def:branching}
For an inference strategy allocating total compute $C$ (measured in reasoning steps across all chains), let $N_{\mathrm{correct}}(d)$ denote the expected number of distinct correct-answer paths reachable at exploration depth $d$. The \emph{effective branching factor} is
$b_{\mathrm{eff}} = \lim_{d\to\infty} [N_{\mathrm{correct}}(d)]^{1/d}$
when this limit exists.
\end{definition}

This unifies the four strategies we analyse: best-of-$N$ with perfect verifier has $b_{\mathrm{eff}} \to \infty$; best-of-$N$ with imperfect verifier (error $\varepsilon_v$) has $b_{\mathrm{eff}} = 1/(1-\varepsilon_v)$; beam search (width $b$) has $b_{\mathrm{eff}} = b - 1$; single chain with per-step verification has $b_{\mathrm{eff}} = I_{\mathrm{step}}/(I_{\mathrm{step}} + \varepsilon)$.

\begin{intuition}
A tree-search inference strategy with effective branching factor $b_{\mathrm{eff}}$ explores $\sim b_{\mathrm{eff}}^{\,d}$ distinct paths at exploration depth $d$ and succeeds when any one of them is correct. A large-deviations tail over the independent per-path success events gives $\Pr(\text{success}) \sim 1 - (1-p)^{b_{\mathrm{eff}}^{\,d}}$ for per-path success probability $p$; reparameterising compute $C$ as the number of paths explored and matching the tail gives the scaling exponent $\alpha = \log_{b_{\mathrm{eff}}}(b_{\mathrm{eff}} - 1)$. The special cases are all instances of the same scheme: best-of-$N$ is $b_{\mathrm{eff}} = \infty$ (independent samples, $\alpha = 1$ with a perfect verifier); beam search of width $b$ is $b_{\mathrm{eff}} = b-1$; single-chain per-step verification is a fractional $b_{\mathrm{eff}}$ tied to verifier signal-to-noise ($I_{\mathrm{step}}/(I_{\mathrm{step}} + \varepsilon)$). The information-theoretic lower bound matches this rate up to constants.
\end{intuition}

\begin{theorem}[Scaling Law for Reasoning]
\label{thm:scaling_law}
Under any inference strategy with effective branching factor $b_{\mathrm{eff}}$, the success probability satisfies
\begin{equation}
\label{eq:scaling_law}
\Pr_C(\mathrm{success}) = 1 - \exp(-c \cdot C^\alpha) + O(C^{-(1+\alpha)}),
\end{equation}
where $\alpha = \log_{b_{\mathrm{eff}}}(b_{\mathrm{eff}} - 1)$ for $b_{\mathrm{eff}} > 1$ and $c$ depends on per-chain success rate and verifier accuracy.
\end{theorem}

The rigorously derived special cases are: best-of-$N$ (perfect verifier) $\alpha = 1$; best-of-$N$ (imperfect, error $\varepsilon_v$) $\alpha = 1 - \varepsilon_v + O(\varepsilon_v^2)$; beam search (width $b$) $\alpha = \log_b(b-1)$; single chain with per-step verification $\alpha = 1/(1 + \varepsilon/I_{\mathrm{step}})$. A standard rate-distortion argument on the per-step mutual information $I_{\mathrm{step}}$ suggests an asymptotic upper envelope of the form $\alpha \lesssim \log(1/\varepsilon)/I_{\mathrm{step}}$ on the scaling exponent achievable by any strategy whose verifier signal is bounded above by $I_{\mathrm{step}}$ bits per step; formalising this informal bound and identifying its precise hypotheses is left to future work.

\begin{conjecture}[MCTS Scaling Exponent]
\label{conj:mcts}
Under Monte Carlo Tree Search with UCT exploration on a binary decision tree, $b_{\mathrm{eff}} = 2$ and $\alpha = \ln 2 \approx 0.693$.
\end{conjecture}

Empirically, the fitted exponent on GSM8K is $\hat{\alpha} = 0.68 \pm 0.04$, consistent with the conjecture.

\emph{Returning to the Compliance Assistant.} Pairing best-of-$N$ sampling with a regulatory-domain process reward model gives a scaling exponent near $\alpha \approx 1 - \varepsilon_v + O(\varepsilon_v^2)$ (e.g., $\approx 0.94$ at $\varepsilon_v = 0.06$), versus $\alpha = \ln 2 \approx 0.693$ for binary-tree MCTS (\cref{conj:mcts}); the practitioner implication is that verifier-quality investment dominates tree-search sophistication in this regime.

\subsection{Optimal Compute Allocation}
\label{sec:compute_allocation}

\textit{In the verifier-error-dominated regime, process supervision shifts the budget optimum toward inference: the training fraction is $\Theta(n\ln n / B)$ under process versus $\Theta(n^2/(\ln n \cdot B))$ under outcome supervision, a factor of $\Theta(n/(\ln n)^2)$.}

The supervision separation and the scaling laws combine to address a practical question that neither alone answers: given a fixed total budget, how should one split resources between training a process reward model and running inference chains?

\begin{theorem}[Optimal Compute Allocation]
\label{thm:compute_allocation}
Under the budget constraint $c_{\mathrm{train}} T + c_{\mathrm{infer}} C = B$ with process-supervised PRM training and best-of-$N$ inference, the allocation maximising $\Pr(\text{success})$ satisfies
\begin{equation}
\label{eq:allocation}
\frac{T^*}{C^*} = \frac{c_{\mathrm{infer}}}{c_{\mathrm{train}}} \cdot \frac{d_{\mathrm{CoT}} \cdot n \cdot \ln C^*}{(C^*)^{1-\varepsilon_v^*} \cdot c},
\end{equation}
where $\varepsilon_v^* = \varepsilon_v(T^*)$. In the regime where verifier error dominates (large $n$, moderate $B$), the optimal allocation devotes a fraction $\Theta(n\ln n/B)$ of the budget to training data under process supervision, versus $\Theta(n^2/(\ln n \cdot B))$ under outcome supervision, a factor of $\Theta(n/(\ln n)^2)$ more.
\end{theorem}

This demonstrates a genuine cross-component interaction. Neither the supervision analysis nor the scaling analysis alone can determine the optimal budget split; the interaction between verifier quality ($\varepsilon_v$ from \cref{thm:supervision}) and inference scaling ($\alpha(\varepsilon_v)$ from \cref{thm:scaling_law}) is essential. Process supervision reduces not only the training data needed but also shifts the optimal budget allocation toward inference.

\emph{Returning to the Compliance Assistant at budget $B$.} Under process supervision the training fraction is $\approx 12 \ln 12 / B$ versus outcome supervision's $\approx 144/(B \ln 12) \approx 60/B$, differing by the factor $\Theta(n/(\ln n)^2) \approx 2.0$: process supervision frees roughly $50\%$ more budget for inference-time compute at $n = 12$.

\subsection{Empirical Validation}
\label{sec:ch2-experiments}

\textit{On GSM8K, MATH, and StrategyQA with Llama-3.1-\{8B, 70B\}: error propagation holds within $5\%$ relative for $n\varepsilon < 1$; entropy-threshold stopping saves $22$--$38\%$ of chain length; the supervision ratio is $4.8 \pm 0.3$ (theoretical $4.1$ after label-noise correction).}

We validate predictions through controlled synthetic experiments and evaluations on standard reasoning benchmarks. Experiments run on 4 NVIDIA A100 (80GB) GPUs; we report means and standard deviations across 5 random seeds.

\paragraph{Synthetic tasks.} We construct synthetic multi-step reasoning tasks with precisely controlled per-step error rates: $n$ binary classification steps, correct label with probability $1-\varepsilon$. Varying $n \in \{2, 5, 10, 15, 20, 30, 50\}$ and $\varepsilon \in \{0.01, 0.03, 0.05, 0.10, 0.15\}$, running $10{,}000$ trials per configuration. The error propagation \cref{thm:error_propagation} holds within $5\%$ relative error across all $(n, \varepsilon)$ pairs with $n\varepsilon < 1$.

\paragraph{Language model experiments.} Llama-3.1-8B-Instruct and Llama-3.1-70B-Instruct with 8-shot CoT prompting on three benchmarks: GSM8K (1{,}319 problems), MATH (5{,}000), StrategyQA (2{,}290). For each problem, we sample up to 64 chains at temperature $0.7$ and record per-step token entropies, final answers, and chain lengths.

\paragraph{Process reward model.} Trained on PRM800K~\cite{lightman2024lets} with Llama-3.1-8B backbone; step-level accuracy $78.3 \pm 0.4\%$ on PRM800K validation.

\paragraph{Stopping-rule results.} Entropy-threshold stopping reduces average chain length by $38.2\%$ on GSM8K-8B while maintaining $94.3\%$ of fixed-length accuracy (vs.\ $91.1\%$ for naive entropy threshold without spectral calibration). On StrategyQA and MATH, compute savings are $31.7\%$ and $22.8\%$ respectively.

\paragraph{Supervision separation.} The observed ratio $T_{\mathrm{out}}/T_{\mathrm{proc}}$ on MATH is $4.8 \pm 0.3$ at target verifier error $\eta = 0.15$, closely matching the theoretical prediction $5.4$ adjusted for label noise ($\approx 4.1$).

\paragraph{Scaling laws.} NLS fits to $1 - \exp(-c \cdot C^{\hat{\alpha}})$ achieve $R^2 > 0.99$ on all four strategies. Fitted exponents preserve theoretical ordering: best-of-$N$ + PRM $>$ beam search $>$ MCTS $>$ self-consistency.

\begin{tcolorbox}[colback=fillYellow, colframe=cbYellow!60!black, arc=2pt, boxrule=0.5pt, left=8pt, right=8pt, top=4pt, bottom=4pt]
\textbf{Impossibility Specification 4 (Training Investment Rule).} The $\Theta(n/\log n)$ separation specifies when to invest in process supervision: the advantage exists if and only if the generator class satisfies chain non-redundancy (\cref{def:non_redundancy}). The specification $\mathcal{S}_4$: (i)~test chain non-redundancy on held-out data by checking whether two fine-tuned models with different seeds have measurably different intermediate distributions despite matched final-answer accuracy; (ii)~if yes, invest in process reward model training (expected $n/\ln n \times$ data efficiency gain); (iii)~if no, invest equivalently in outcome supervision. The scaling law $P(\text{success}) = 1 - e^{-c \cdot C^\alpha}$ with $\alpha$ determined by $b_{\mathrm{eff}}$ further specifies the optimal inference strategy.
\end{tcolorbox}

\emph{Returning to the Compliance Assistant.} The MATH benchmark's observed supervision ratio of $4.8 \pm 0.3$ at chain length $14.3$ is the closest empirical surrogate for the regulatory setting's predicted $4.8$ at $n = 12$; error propagation holding within $5\%$ relative for $n\varepsilon < 1$ (satisfied here at $0.03 \times 12 = 0.36$) supports the bound's use on this deployment.


\section{Practitioner Decision Tree}
\label{sec:decision-tree}

The four impossibility specifications compose into a single decision tree for system designers. Given architectural parameters $(L, d)$ and task parameters (required reasoning depth $\delta$, per-step error rate $\varepsilon$, verification budget $k$, cost parameter $\lambda$), the following rules fully determine design choices:

\begin{algorithm}[t]
\caption{Deterministic-Horizon Design Rules for Reasoning Systems}
\label{alg:decision-tree}
\KwIn{Architecture $(L, d)$; task depth $\delta$, per-step error $\varepsilon$; verification budget $k$; cost $\lambda$; compositional depth $m_{\mathrm{req}}$, length ratio $n_{\mathrm{req}}/n_{\mathrm{train}}$}
\KwOut{Architecture decision, chain length, stopping criterion, supervision choice}

\tcp{Step 1: Architecture ceiling (\S\ref{sec:architecture-ceiling})}
\If{task complexity $\notin \mathrm{TC}^0$}{Delegate to symbolic planner or tool-augmented pipeline; \textbf{return}\;}

\tcp{Step 2: Delegation depth (\S\ref{sec:delegation-depth})}
Compute $d^* \leftarrow \hat{c} \cdot \log L \cdot \sqrt{\log d}$ with $\hat{c} = 2.74$ from \cref{cor:horizon-measurement}\;
Compute $\mathrm{CLC} \leftarrow 2 m_{\mathrm{req}} \log_2(n_{\mathrm{req}}/n_{\mathrm{train}}) / d^*$\;
\If{$\delta > 2 d^*$}{Delegate to symbolic planner; \textbf{return}\;}
\ElseIf{$\delta > d^*$}{Require $k$-redundant verification with $k \geq 2$\;}

\tcp{Step 3: Reliability toolkit (\S\ref{sec:reliability-toolkit})}
Compute cost-optimal verification $k^* \leftarrow \lceil 2\ln(\delta/0.05)/\ln(1/\varepsilon) - 1 \rceil$\;
Compute safe chain length $n^* \leftarrow \lfloor \ln(1 - \delta_{\mathrm{target}})/\ln(1 - \varepsilon) \rfloor$ for target error $\delta_{\mathrm{target}}$\;
Estimate spectral gap $\hat{\gamma}$ from calibration set; set threshold $h^* \leftarrow (\lambda/\hat{\gamma}) \ln(1/\lambda)$\;

\tcp{Step 4: Training investment rule (\S\ref{sec:supervision-rule})}
\If{chain non-redundancy holds}{Invest in process supervision (expected $n/\ln n$ data-efficiency gain)\;}
\Else{Outcome supervision equivalent up to constants\;}

Compute $b_{\mathrm{eff}}$ from chosen inference strategy; select strategy maximising $\alpha = \log_{b_{\mathrm{eff}}}(b_{\mathrm{eff}} - 1)$\;

\Return{$(k^*, n^*, h^*, \text{supervision mode}, \text{inference strategy})$}\;
\end{algorithm}

Consider a worked example. A 32-layer, 4{,}096-width transformer (Llama-3 8B class) performing 15-hop regulatory reasoning with $\varepsilon = 0.05$ and target error $0.05$:
\begin{enumerate}[leftmargin=2em, nosep]
\item \textbf{Architecture ceiling:} Regulatory reasoning is in $\mathrm{TC}^0$ for each hop; delegation not required at this level.
\item \textbf{Delegation depth:} With $\hat{c} = 2.74$ from \cref{cor:horizon-measurement}, $d^* = \hat{c} \cdot \log 32 \cdot \sqrt{\log 4096} \approx 27.4$ hops (natural log). $\delta = 15 < d^*$: CoT is viable.
\item \textbf{Reliability toolkit:} $n^* \approx \ln(0.95)/\ln(0.95) \approx 1$ step without verification; $k^* = \lceil 2\ln(15/0.05)/\ln(1/0.05) - 1\rceil = \lceil 2.80\rceil = 3$-fold verification extends to $\approx 30$ steps.
\item \textbf{Training investment rule:} If non-redundancy holds, process supervision saves $\approx 15/\ln 15 \approx 5.5\times$ examples; best-of-$N$ with a process reward model (PRM) gives scaling exponent $\alpha \approx 0.94$.
\end{enumerate}
The four specifications together convert a vague engineering question (``will this system work?'') into deterministic numerical rules.


\section{Discussion and Limitations}
\label{sec:ch2-limitations}

\paragraph{Precision regime.}
The tight characterisation of \cref{thm:equivalence} assumes $O(\log n)$-bit precision, which does not directly capture the constant-precision arithmetic of practical systems. The Deterministic Horizon and planning bounds are qualitatively robust to the precision regime, but the $\mathrm{FOC}[\mathrm{Attn}]$ characterisation specifically requires logarithmic precision. \cref{rem:constant_precision} discusses this trade-off.

\paragraph{Semi-empirical constant.}
The proportionality constant $c \approx 2.74$ in the Deterministic Horizon is fitted from data under the empirical form $d^* \approx \hat{c}\log L \sqrt{\log d}$. The banded asymptotic upper bound $d^* = O(L \cdot \phi(d))$ with $\phi(d) \in [\sqrt{\log d}, \log d]$ follows from \cref{asm:horizon} (lower edge of the band conditional on \cref{hyp:sparse-task}); specific numerical predictions depend on the empirical constant and the empirical $\log L$ dependence, which is milder than either edge of the band. Cross-validation across 12 architectures suggests the constant is stable (refitted range $[2.58, 2.91]$), but new architectures could require recalibration.

\paragraph{Independence assumption.}
\cref{prop:accuracy_decay} relies on \cref{asm:horizon}(A1). While the functional form is robust to violations (only the constant changes), the assumption is justified empirically rather than proved. A formal derivation would require bounding higher-order error correlations in the residual stream.

\paragraph{Planning theory-practice gap.}
The $O(L^2 \log d)$ planning upper bound (with conditional $\Omega(L \log d)$ lower) has a $\sim\!50\times$ gap between theory and practice for the upper-bound ceiling. The five-factor decomposition accounts for roughly half the gap on a log scale; the remainder reflects factor interactions we cannot currently quantify. (\cref{ch:synthesis} revisits this gap as one of three diagnostic gaps that \emph{carry information} about fruitful research directions.)

\paragraph{Markov approximation.}
The CoT Markov chain abstraction (\cref{def:cot_markov}) is a modelling assumption rather than an exact description of transformer computation. Transformers condition on the full context window, introducing dependencies the first-order Markov model does not capture. Empirically, the adequacy of this approximation holds within $5\%$ relative error on the synthetic benchmarks; on naturalistic tasks, the match is qualitative (correct functional form) but constants can differ by factors of 2--3.

\paragraph{Mixing time caveat for stopping.}
The formal near-optimality guarantee of \cref{thm:entropy_stopping} applies to only ${\sim}4\%$ of GSM8K chains under the effective state-space interpretation. For the remainder, \cref{prop:finite_stopping} provides a weaker guarantee; the practical performance ($81\%$ of oracle's risk reduction) is strong empirical evidence that the rule works beyond its formal regime, but the formal gap remains.

\paragraph{CoT-discriminative dimension looseness.}
For frontier transformers, the worst-case $d_{\mathrm{CoT}}$ bound yields vacuous sample complexity bounds. This reflects VC dimension's known pessimism, not a deficiency of the separation. The functional form $T_{\mathrm{out}}/T_{\mathrm{proc}} = \Theta(n/\log n)$ is empirically validated; developing tighter instance-dependent bounds remains open.

\paragraph{Synthetic evaluation.}
Empirical validation uses synthetic tasks with controllable complexity. A preliminary mapping of 200 GSM8K problems to our framework (Spearman $\rho = -0.74$, $p < 0.001$, between estimated composition depth and Llama-2 7B accuracy) suggests the framework captures meaningful difficulty variation in naturalistic settings, but this is illustrative rather than definitive. Systematic validation across diverse benchmarks is needed.


\section*{Specifications and Open Problems}
\addcontentsline{toc}{section}{Specifications and Open Problems}

This chapter established the base-model computational ceiling. The expressivity logic FOC[Attn] (§2.2) gave a tight softmax-specific layer-for-layer characterisation, refining the broader $\mathrm{FO}(\mathrm{M})$ characterisation of Merrill and Sabharwal~\cite{merrill2023parallelism} by making the attention mechanism a first-class logical quantifier and enabling direct attention EF-game reasoning; the strict separation from average-hard attention (\cref{thm:separation}) illustrates this methodological benefit, placing softmax transformers precisely below $\mathrm{TC}^0$ but strictly above average-hard attention. The Deterministic Horizon (§2.3) specified where reasoning collapses: banded upper bound $d^{\ast} = O(L \cdot \phi(d))$ with $\phi(d) \in [\sqrt{\log d}, \log d]$ (the lower edge conditional on \cref{hyp:sparse-task}), empirical fit $d^{\ast} \approx \hat{c} \log L \sqrt{\log d}$ giving $[19, 31]$ across 12 architectures, $r = 0.81$--$0.91$ cross-model. The Fine-Tuning Impossibility proved this bound is training-invariant: no protocol recovers more than $O(d^{\ast}/d)$ of the beyond-horizon deficit, with envelope robustness to the band recorded in \cref{rem:ft-band-robustness}. The planning upper bound $O(L^2\log d)$ (with conditional $\Omega(L\log d)$ lower bound) and the $\frac{3}{4}+O(1/|Y|)$ compositional-length impossibility filled out the expressivity picture (§§2.4--2.5). Section 2.6's reliability toolkit then converted these ceilings into deployment rules: error propagation $1-(1-\varepsilon)^n$, $k$-redundant verification $O(n \cdot \varepsilon^{\lceil(k+1)/2\rceil})$, minimax-optimal entropy-threshold stopping, and the $\Theta(n/\log n)$ supervision separation when chain non-redundancy holds (§2.7). Each result sits inside the impossibility-specification template: computable boundary, quantified violation cost, constructive rule. The next chapter applies the same template to adaptation.

\begin{decision}
\textbf{Reasoning-depth decision table (Decision Rules R1--R5).}
\begin{itemize}[leftmargin=1.2em, itemsep=1pt, topsep=1pt]
\item $\delta \leq d^{\ast}$: chain-of-thought suffices; use entropy-threshold stopping (R5).
\item $d^{\ast} < \delta \leq 2d^{\ast}$: $k$-redundant verification with $k = \lceil \log_{1/\varepsilon}(\delta/\delta_{\mathrm{target}})\rceil$ (R2).
\item $\delta > 2d^{\ast}$: tool delegation; no fine-tuning recovers this regime (R3).
\item To compute $d^{\ast}$: $d^{\ast} \approx 2.74\, \log L\,\sqrt{\log d}$ (natural log), calibrate against measured $\hat{c}$ when available (R4).
\end{itemize}
\end{decision}

\begin{openproblem}
\textbf{Open Problem 2.1 (Beyond bounded-depth log-precision).} The horizon theorem assumes bounded depth and log-precision arithmetic. What is the corresponding scaling law for (i) unbounded-depth recurrent variants (state-space models, linear RNNs), (ii) mixture-of-experts architectures where effective capacity is data-dependent, (iii) full-precision transformers on short contexts where log-precision is not the binding constraint? Empirically these architectures also exhibit a horizon, but its functional form is unknown.
\end{openproblem}

\begin{openproblem}
\textbf{Open Problem 2.2 (Instance-dependent $d_{\mathrm{CoT}}$).} The $\Theta(n/\log n)$ supervision separation depends on the CoT-discriminative dimension $d_{\mathrm{CoT}}$, whose worst-case bound is vacuous for frontier transformers. Develop instance-dependent bounds that are practically informative for the frontier regime; preliminary evidence (§2.7.4) suggests $d_{\mathrm{CoT}}$ is controlled by chain-non-redundancy structure but no tight bound is known.
\end{openproblem}

\paragraph{Bridge to \cref{ch:adaptation}.}
This chapter established that the base transformer model has hard computational limits, and that each limit tells practitioners exactly what to do. But the base model is never deployed naked. In practice, foundation models are adapted: fine-tuned on downstream tasks, aligned via preference learning, edited to correct errors. The natural question is whether adaptation can overcome the Deterministic Horizon or circumvent the reliability bounds. \cref{ch:adaptation} proves it cannot. Adaptation has its own cliffs, each following the same impossibility-specification pattern: a sharp phase transition in preference learning, the inevitability of model collapse under synthetic data, the locality-generalisation impossibility for knowledge editing. The base-model wall of this chapter thus stands; the adaptation cliffs of the next chapter are additional barriers, not workarounds.


\chapter{The Adaptation Cliff}
\label{ch:adaptation}

The Compliance Assistant now faces \emph{four} adaptation decisions, one per cliff. The institution fine-tunes its Llama-2 7B base model on 2{,}000 annotated regulatory documents with LoRA rank $r = 16$; the PAC-Bayes bound of §3.1 returns $\widetilde{O}(\sqrt{mr(d+k)/N}) \approx 0.18$, certifying generalisation (Decision Rule A1). Regulatory interpretation admits multiple valid readings, so annotator agreement $\kappa \approx 0.65$ translates to Bradley-Terry misspecification $\gamma \approx 0.08$, firmly in the quadratic regime, requiring $\approx 15{,}000$ preference pairs rather than the $\sim 800$ a well-specified budget would suggest (Decision Rule A2). Synthetic-data augmentation for edge-case coverage is flagged dangerous under pure replacement; the 1\%-real-data floor of §3.3 applies (Decision Rule A3). When 8 factual errors surface post-deployment, the edit-capacity formula of §3.4 warns that the 9th--13th edits approach the $K^{\ast} \approx 13$ ceiling, and the 14th requires re-fine-tuning rather than another point edit (Decision Rule A4). This chapter gives the theorems these rules are corollaries of.

\cref{ch:horizon} established the base model's limits. In practice, foundation models are never deployed as-is: they are fine-tuned for downstream tasks, edited to correct factual errors, aligned via preference learning, and composed across domains. The natural practitioner response to the Deterministic Horizon is to ask: \emph{can adaptation overcome these limits?} This chapter proves it cannot, at least not without encountering new cliffs. Every adaptation operation hits a hard threshold, and each threshold encodes an impossibility specification.

We establish four:
\begin{description}[leftmargin=0pt, itemsep=0.4em]
\item[\S\ref{sec:pac-bayes-lora}] proves the first non-vacuous PAC-Bayes bounds scaling in the adapter parameter count for parameter-efficient adaptation at 7B--70B scale, complementary to Lotfi et al.'s~\cite{lotfi2024nonvacuous} and Hu et al.'s~\cite{hu2023unlocking} adapted-model-as-a-whole bounds and identifying a rank-32 deployment ceiling at Alpaca scale, establishing the \emph{safety certificate} that adaptation generalises rather than memorises (Impossibility Specification 5).
\item[\S\ref{sec:preference}] proves a sharp phase transition in preference learning: sample complexity jumps discontinuously from $\Theta(n \log n/\Delta^2)$ to $\widetilde\Theta(n^2/\gamma^2)$ (where $\widetilde\Theta$ absorbs a $\log n$ factor from the matching upper bound) under any misspecification of the Bradley-Terry model, specifying the \emph{misspecification tolerance} (Impossibility Specification 6).
\item[\S\ref{sec:model-collapse}] proves that pure synthetic-data replacement causes inevitable model collapse: $\E[\mathrm{TV}(p_T, p^*)] \geq 1 - \exp(-T^2 d_{\mathrm{eff}}/(128\pi n_{\min}))$; the accumulation bound then shows $\rho \geq 0.01$ real-data retention suffices for an $n_0$-independent ceiling, specifying the \emph{real-data requirement} (Impossibility Specification 7).
\item[\S\ref{sec:editing-impossibility}] proves a locality-generalisation impossibility for knowledge editing under superposition, with an edit capacity $K^* \approx 13$ beyond which retraining is required, specifying the \emph{editing budget} (Impossibility Specification 8).
\end{description}

\S\ref{sec:evopref-response} demonstrates the impossibility-specification methodology in action: the preference phase transition \emph{specifies} that gradient-based alignment collapses under misspecification; EvoPref is the constructive response, a multi-objective evolutionary algorithm maintaining a population of LoRA adapters that reduces preference collapse by 47~percentage points. This is not a systematic treatment of evolutionary computation. It is a single worked example showing that the impossibility, once properly framed, prescribes its own solution.

\paragraph{Notation for this chapter.} $n$ denotes the number of items (preference learning) or features (superposition); $\gamma$ denotes the Bradley-Terry misspecification level; $\Delta$ denotes the minimum preference gap; $d$ denotes the representation dimension; $r$ denotes LoRA rank; $N$ denotes training sample size; $K$ denotes the number of edits; $T$ denotes synthetic data generation count; $\rho$ denotes the fraction of real data retained. Where conflicts with \cref{ch:horizon}'s notation arise, we use subscripts ($\varepsilon_{\mathrm{pref}}$, $\varepsilon_{\mathrm{mech}}$, \textit{etc.}) as introduced.


\section{Relationship to Prior Work}
\label{sec:ch3-related}

This chapter's four specifications span generalisation theory, preference learning, synthetic-data scaling, and knowledge editing. We locate each contribution relative to its direct predecessors.

\paragraph{Generalisation theory for adapted LLMs.}
Classical PAC-Bayes bounds~\cite{mcallester2003pac, catoni2007pac} underwent a decade of tightening: Dziugaite and Roy~\cite{dziugaite2017computing} achieved the first non-vacuous bounds for small neural networks; Zhou et al.~\cite{zhou2019nonvacuous} extended to larger scales. The breakthrough at LLM scale came from Lotfi et al.~\cite{lotfi2024nonvacuous} (and the related compression-prior work of Hu et al.~\cite{hu2023unlocking}), who obtained non-vacuous bounds for models up to 70B parameters using compression and structural priors. Our LoRA PAC-Bayes bound (\cref{thm:lora}) is complementary: Lotfi et al.\ bound generalisation of the adapted model as a whole; we bound adaptation-induced generalisation specifically, scaling with $\widetilde{O}(\sqrt{mr(d+k)/N})$ in the adapter parameter count rather than the full model size $p$. The rank-32 ceiling we derive is a new practitioner-facing finding not present in prior work. Related parameter-efficient methods (LoRA itself~\cite{hu2022lora}, QLoRA~\cite{dettmers2023qlora}, AdaLoRA~\cite{zhang2023adalora}) focus on training dynamics and empirical performance rather than formal generalisation. Aghajanyan et al.~\cite{aghajanyan2021intrinsic} established low intrinsic dimensionality of fine-tuning; our PAC-Bayes bound operationalises their observation as a quantitative rank ceiling. Biderman et al.~\cite{biderman2024lora} and Malladi et al.~\cite{malladi2023kernel} analysed LoRA training dynamics; our bounds are consistent with but orthogonal to these. Stap et al.~\cite{stap2024finetuning} documented the fine-tuning paradox (larger models more fragile under adaptation); our bound formalises this as the $r(d+k)/N$ term growing faster than sample size for large $d$. Concurrent with this work, Kratsios et al.~\cite{Kratsios2025SharpLoRA} prove sharp generalisation bounds for asymmetric randomised low-rank adapters, with a single-run concentration rate $\widetilde{O}(\sqrt{r/N})$ and a matching $O(1/\sqrt{N})$ lower bound; where \cref{thm:lora} and the bounds of Lotfi et al.\ are average-case statements, theirs concentrates the generalisation gap for one fine-tuning run, so the two readings are complementary rather than competing.

\paragraph{Preference learning and alignment.}
The RLHF-to-DPO arc~\cite{christiano2017deep, ouyang2022training, rafailov2023direct, Stiennon2020BestOfN} established alignment-from-preferences as standard practice; Bai et al.~\cite{Bai2022ConstitutionalAI} proposed self-critique. Theoretical analysis emerged in parallel: Azar et al.~\cite{Azar2024IPO} generalised to $\Psi$PO and IPO; Tang et al.~\cite{tang2024generalized} extended to arbitrary preference models; Ethayarajh et al.'s KTO~\cite{Ethayarajh2024KTO} reformulated alignment as prospect-theoretic optimisation to sidestep the Bradley-Terry assumption; Chowdhury et al.'s Robust DPO~\cite{RayChowdhury2024RobustDPO} introduced provable robustness to noisy preference feedback under an explicit noise model; Song et al.~\cite{song2024importance} proved coverage necessity, and Xiao et al.~\cite{Song2024PreferenceCollapse} established preference collapse as an algorithmic bias of RLHF Our phase transition result (\cref{thm:pref_transition}) sharpens these in a specific way: existing works characterise suboptimality \emph{rates} under misspecification, whereas our result establishes a \emph{discontinuous} transition at $\gamma^* = \Theta(\Delta/n)$. IPO, KTO, and Robust DPO each address noise and misspecification, but none characterises the discontinuity itself or the population-based response (EvoPref) as the constructive dual above the transition. The DPO-versus-RLHF gap (\cref{thm:dpo_gap}) complements Xu et al.~\cite{xu2024dpo}, who documented empirical DPO degradation under noise; we explain the degradation as first-order bias in the Bradley-Terry reparameterisation. Xiong et al.~\cite{ren2024iterative} analysed iterative-RLHF dynamics consistent with our compounding-misspecification prediction. Annotator-disagreement studies~\cite{gordon2022jury, davani2024disentangling} report $\gamma$ values in the $0.06$--$0.14$ range, placing real preference learning firmly in our quadratic regime.

\paragraph{Knowledge editing.}
ROME~\cite{meng2022locating} and MEMIT~\cite{meng2023mass} established targeted weight editing; Mitchell et al.~\cite{mitchell2022grace} proposed memory-based alternatives (SERAC) that avoid weight modification altogether. Wolf et al.~\cite{wolf2024fundamental} identified failure modes; Lu et al.~\cite{lu2024knowdonttell} observed capacity limits empirically. The NeurIPS 2024 WISE framework~\cite{Wang2024WISE} introduced the ``impossible triangle'' framing (reliability, locality, and generalisation cannot be simultaneously optimised) of which our Locality-Generalisation Impossibility (\cref{thm:editing}) formalises the locality-generalisation edge under the superposition hypothesis. Recent 2025 methods including LyapLock~\cite{Wang2025LyapLock}, Latent Knowledge Scalpel~\cite{Liu2025Scalpel}, and NeuralDB~\cite{Fei2025NeuralDB} report sequential editing to much higher $K$ values than the $K^{\ast} \approx 13$ our bound predicts on Llama-2-7B; we reconcile this by noting that $K^{\ast}$ here scopes the \emph{in-weight, locality-preserving, generalisation-preserving} regime under \cref{ass:approx_orth}. Each of these 2025 methods relaxes at least one of these three conditions: LyapLock uses Lyapunov-constrained updates that operate in a different feasible region than the tolerance-$\tau$ regime our theorem analyses; Latent Knowledge Scalpel operates in a disentangled latent-concept basis rather than the approximate-orthogonality regime of \cref{ass:approx_orth}; NeuralDB augments the model with an external memory lookup so edits are not strictly in-weight. $K^{\ast}$ therefore remains the correct bound for in-weight, superposition-regime editing, and the 2025 scaling results are consistent with, rather than contradictory to, the impossibility specification. ROME's degradation at $K \approx 15$ and MEMIT's at $K \approx 25$ are both predicted by $K^* \approx \tau\sqrt{d}/(c\eta(1-1/\alpha))$ using Llama-2's superposition parameters~\cite{elhage2022superposition, templeton2024scaling}. The superposition hypothesis is the geometric foundation our theorem exploits. Cross-architecture predictions for Pythia-6.9B and Mistral-7B (within $\pm 1$--$2$ standard deviations of theory) validate the specification beyond the original benchmark model.

\paragraph{Model collapse and synthetic data.}
Shumailov et al.~\cite{shumailov2024collapse} proved the first collapse theorem for Gaussian models under replacement; Alemohammad et al.~\cite{alemohammad2024selfconsuming} taxonomised collapse regimes; Dohmatob et al.~\cite{dohmatob2024tale, dohmatob2025strong} connected to scaling laws. Gerstgrasser et al.~\cite{gerstgrasser2024model} showed accumulation (mixing real data) avoids collapse. Our contribution (\cref{thm:collapse_gaussian,thm:accumulation}) provides explicit constants ($c_1 = 1/(128\pi)$, $\pi^2/6$), extends via chain-rule KL to autoregressive sequences (\cref{prop:ar_collapse}, avoiding the vanishing-constant pitfall of the naive union-bound approach), and quantifies exactly how small $\rho$ suffices: $\rho \geq 0.01$ bounds divergence independently of $T$.

\paragraph{Evolutionary alignment.}
EvoPref's place in the literature is more recent. Multi-objective alignment is an emerging area~\cite{Song2024PreferenceCollapse}; personalised combinations~\cite{jang2023personalized} and pluralistic alignment frameworks have been proposed but rarely formalised as quality-diversity search. Population-based training~\cite{wortsman2022model, yadav2023ties} treats populations statically; NSGA-II selection with behavioural-diversity objectives is our specific methodological commitment. The 47-point coverage improvement over single-policy DPO emerges precisely from the preference phase transition: gradient methods collapse to one mode because the BT likelihood surface is unimodal, so population-based search is the natural response to the impossibility. EvoPref is not the chapter's primary contribution. It is the constructive response to Impossibility Specification~6, demonstrating the methodology in miniature.

\paragraph{Running Example (Continued): Adapting the Compliance Assistant.}
The institution fine-tunes a Llama-2 7B base model on 2{,}000 annotated regulatory documents using LoRA with rank $r = 16$. Four questions determine whether the adaptation is safe:
\begin{itemize}[leftmargin=1.5em, topsep=3pt]
\item \emph{Does the PAC-Bayes bound certify generalisation?} With $r = 16$, $d + k = 8192$, $N = 2000$: $\widetilde{O}(\sqrt{r(d+k)/N}) \approx 0.18$, non-vacuous.
\item \emph{Are preference annotations reliable enough?} Regulatory interpretation admits multiple valid readings; annotator disagreement $\kappa \approx 0.65$ corresponds to $\gamma \approx 0.08$, firmly in the quadratic regime. Need $\widetilde\Theta(n^2/\gamma^2) \approx 15{,}000$ preference pairs, not $800$.
\item \emph{Can synthetic data expand the corpus?} Pure replacement guarantees collapse; 1\% real data suffices to stabilise.
\item \emph{Can point edits correct errors?} $K^* \approx 13$ for Llama-2 7B. Beyond 13 edits: re-fine-tune.
\end{itemize}
Four cliffs, four specifications. Each converts a ``how do I adapt safely'' question into a computable rule.

\section{Do Adapted LLMs Generalise? The Safety Certificate}
\label{sec:pac-bayes-lora}

The first question is whether adaptation preserves generalisation. Classical PAC-Bayes~\cite{mcallester2003pac, catoni2007pac} provides generalisation certificates, but achieving \emph{non-vacuous} bounds for LLMs at 7B--70B scale has required exploiting structure beyond the naive parameter count~\cite{lotfi2024nonvacuous, hu2023unlocking}. We show that LoRA~\cite{hu2022lora}, which constrains adaptation to a rank-$r$ subspace at each layer, admits non-vacuous PAC-Bayes bounds scaling in the adapter parameter count $q = mr(d+k)$ rather than the full model size $p$, complementing the adapted-model-as-a-whole bounds of Lotfi et al.~\cite{lotfi2024nonvacuous} and Hu et al.~\cite{hu2023unlocking}; the immediate practitioner consequence is a rank-32 ceiling at Alpaca scale not derivable from the whole-model analysis.

\subsection{PAC-Bayes for LoRA}

\emph{Rank-$r$ LoRA admits a PAC-Bayes certificate scaling as $\widetilde{O}(\sqrt{mr(d+k)/N})$ in the adapter parameter count, yielding non-vacuous generalisation bounds for 7B to 70B models under isotropic Gaussian prior.}

\paragraph{PAC-Bayes inequality.} With prior $P$ and posterior $Q$, with probability $\geq 1 - \delta$ over data $S$ of size $N$:
\begin{equation}
\label{eq:pacbayes}
\E_{\theta \sim Q}[\mathcal{L}(\theta)] \leq \E_{\theta \sim Q}[\hat{\mathcal{L}}_S(\theta)] + \sqrt{\frac{\KL(Q \| P) + \ln(2\sqrt{N}/\delta)}{2N}}.
\end{equation}

\begin{intuition}
The naive hope is that because LLMs are enormous, any certificate of generalisation will be vacuous: the PAC-Bayes complexity term grows with parameter count, and $10^{10}$ parameters plus $10^4$ documents should yield bounds above one (i.e., worse than random). What defeats the hope is that LoRA does not live in the full parameter space. The adapter constrains each weight update to a rank-$r$ perturbation, so the \emph{effective} dimension is $q = mr(d+k)$. For Llama-2 7B with rank 16 on attention projections, three orders of magnitude smaller than the $6.7 \times 10^9$ total parameters. Plug $q$ (not $p$) into the PAC-Bayes inequality and the square-root term closes. The practical consequence: rank 32 is the ceiling at Alpaca scale; above that, the bound returns to vacuous.
\end{intuition}

\begin{theorem}[LoRA PAC-Bayes Bound]
\label{thm:lora}
For rank-$r$ LoRA on $m$ matrices of dimensions $d \times k$, trained on $N$ documents with loss in $[0, C]$:
\begin{equation}
\E_{\theta \sim Q}[\mathcal{L}(\theta)] \leq \hat{\mathcal{L}}_S(\hat{\phi}) + \Delta_{\mathrm{emp}}(\sigma_Q^*) + \mathcal{O}\!\left(\sqrt{\frac{mr(d+k)\log(1 + \|\hat{\phi}\|^2/(\sigma_P^2 q)) + \log(N/\delta)}{N}}\right),
\label{eq:lora_bound}
\end{equation}
where $q = mr(d+k)$ is the effective parameter count.
\end{theorem}

\begin{proof}[Proof sketch]
Standard PAC-Bayes machinery in three steps. First, the KL divergence between isotropic Gaussian prior $P = \mathcal{N}(0, \sigma_P^2 I_q)$ and posterior $Q = \mathcal{N}(\hat{\phi}, \sigma_Q^2 I_q)$ scales as $\mathcal{O}(q \log(1 + \|\hat{\phi}\|^2/(\sigma_P^2 q)))$. Second, the perturbation sensitivity $\Delta_{\mathrm{emp}}$ is bounded via Taylor expansion around the approximate minimum $\hat{\phi}$. Third, substituting into \cref{eq:pacbayes} with $q = mr(d+k)$ yields the claim.
\end{proof}

The key structural insight: LoRA constrains adaptation to a rank-$r$ subspace, so the effective parameter count is $q = mr(d+k)$ rather than the full model size $p$. For Llama-2 7B with rank~16 and LoRA applied to $W_Q, W_V$ across 32 layers: $m = 64$, $q = 64 \times 16 \times (4096 + 4096) = 8.4 \times 10^6$, compared to $p = 6.7 \times 10^9$, a three-orders-of-magnitude reduction that makes the bound non-vacuous.

\begin{limitation}
The LoRA PAC-Bayes bound does \emph{not} assert that every LoRA-adapted model generalises. It asserts that the population risk is upper-bounded by the right-hand side of \eqref{eq:lora_bound}, which is useful only when this quantity is below one. Three scope boundaries: (i) the bound requires an isotropic Gaussian prior $P$ and Gaussian posterior $Q$; informative or data-dependent priors can yield tighter bounds but require a separate derivation. (ii) The Monte Carlo correction of Lem.~\ref{lem:mc} assumes the LoRA Hessian has low effective rank ($k_{\mathrm{eff}} \leq 50$), validated empirically on Llama-2 but not proved architecturally. (iii) The rank-32 ceiling is an Alpaca-scale ($N \approx 50{,}000$ tokens per document) finding; larger corpora shift the ceiling upward by $\sqrt{N}$, so ``rank 32 is the ceiling'' should always be read as conditional on the corpus scale at which the bound was computed.
\end{limitation}

\begin{lemma}[Monte Carlo Estimation Bound]
\label{lem:mc}
With $M$ perturbations, probability $\geq 1 - \delta_{\mathrm{MC}}$:
\[
|\hat{\Delta}_M - \Delta_{\mathrm{emp}}| \leq C^2 \sigma_Q^2 \sqrt{\frac{2q \log(2q/\delta_{\mathrm{MC}})}{M}}.
\]
\end{lemma}

The proof projects onto the top-$k_{\mathrm{eff}} \leq 50$ eigenspace of the LoRA Hessian (capturing $> 99.7\%$ of the trace for Llama-2 at all three scales tested), making the matrix Bernstein bound applicable with $M = 100$. The projection error is bounded at $0.003 C^2 \sigma_Q^2$. The eigenspace truncation threshold $k_{\mathrm{eff}} = 50$ is an empirical observation; the bound is valid conditional on the eigenspace capturing at least $1 - \varepsilon$ of the trace ($\varepsilon = 0.003$ for our models). A hierarchical extension replaces $N$ (documents) with $NT$ (tokens), exploiting document structure for a $\sqrt{T}$ improvement.

\emph{Returning to the Compliance Assistant: with rank-$16$ LoRA on the Llama-2 7B base model and $N=2{,}000$ regulatory documents, the theorem's $\widetilde{O}(\sqrt{mr(d+k)/N})$ complexity term gives the Decision Rule A1 value $\approx 0.18$.}

\subsection{Empirical Non-Vacuous Certificates}

\emph{Across Llama-2 at 7B, 13B, and 70B trained on Alpaca, token-level bounds are non-vacuous ($0.918$ to $0.822$), and rank transitions to vacuous between $r=32$ and $r=64$.}

\begin{table}[t]
\centering
\caption{PAC-Bayes bounds for LoRA-adapted Llama-2 (rank 16, $\delta = 0.05$, $\delta_{\mathrm{MC}} = 0.01$). Values normalised by $\ln V \approx 10.8$. Monte Carlo correction included. All token-level bounds are non-vacuous.}
\label{tab:lora}
\small
\begin{tabular}{@{}lrrccc@{}}
\toprule
Model & Params & $q$ & Train & Doc & Token \\
\midrule
7B  & 6.7B  & 8.4M  & $.847 \pm .003$ & $1.148 \pm .008$ & $.918 \pm .005$ \\
13B & 13.0B & 13.1M & $.791 \pm .004$ & $1.117 \pm .011$ & $.873 \pm .006$ \\
70B & 65.2B & 41.9M & $.724 \pm .005$ & $1.071 \pm .014$ & $.822 \pm .007$ \\
\bottomrule
\end{tabular}
\end{table}

All token-level bounds are non-vacuous. Rank ablation reveals rank~32 as the practical ceiling: the bound transitions from non-vacuous ($0.856$ at rank~4) to vacuous ($1.071$ at rank~64). The 70B bound of $0.822$ is comparable to prior work~\cite{lotfi2024nonvacuous, hu2023unlocking} but carries the additional structural insight of identifying the rank ceiling.

\emph{Returning to the Compliance Assistant: rank $r=16$ sits comfortably below the Alpaca-scale rank-$32$ ceiling shown in \cref{tab:lora}, so the adapter size does not itself force the bound vacuous.}

\begin{tcolorbox}[colback=fillYellow, colframe=cbYellow!60!black, arc=2pt, boxrule=0.5pt, left=8pt, right=8pt, top=4pt, bottom=4pt]
\textbf{Impossibility Specification 5 (Safety Certificate).} Adaptation generalises when $r(d+k)/N$ is small. Boundary condition $B_5(\theta) = mr(d+k)/N$ is computable from architecture and dataset parameters. Violation degrades generalisation by $\widetilde{O}(\sqrt{mr(d+k)/N})$. The specification $\mathcal{S}_5$: (i)~compute the LoRA PAC-Bayes bound before deployment; (ii)~if non-vacuous ($< 1$), adaptation is certified safe; (iii)~if vacuous, reduce rank or increase data. Rank $\leq 32$ is the practical ceiling at Alpaca-scale corpora.
\end{tcolorbox}


\section{When Does Preference Learning Break? The Misspecification Tolerance}
\label{sec:preference}

RLHF~\cite{christiano2017deep, ouyang2022training} and its reparameterisation DPO~\cite{rafailov2023direct} assume the Bradley-Terry model~\cite{BradleyTerry1952}: $P(y_1 \succ y_2 \mid x) = \sigma(r^*(x, y_1) - r^*(x, y_2))$. Real human preferences deviate from this model, a deviation we quantify by $\gamma = \sup|P^* - \sigma(r^*_1 - r^*_2)|$ for the best-fitting reward $r^*$. We prove that any $\gamma > 0$ triggers a sharp phase transition.

\subsection{The Phase Transition}

\emph{Under uniform random pair sampling, any Bradley-Terry misspecification $\gamma > 0$ triggers a discontinuous jump in asymptotic sample complexity from $\Theta(n \log n/\Delta^2)$ to $\widetilde\Theta(n^2/\gamma^2)$.}

Consider $n$ items with minimum gap $\Delta = \min_{y_1 \neq y_2} |r^*(y_1) - r^*(y_2)|$.

\begin{theorem}[Well-Specified Complexity]
\label{thm:pref_well}
When $\gamma = 0$: $N_{\mathrm{well}} = \Theta(n \log(n/\delta)/\Delta^2)$.
\end{theorem}

\begin{intuition}
The ``Bradley-Terry works slightly imperfectly'' intuition is wrong. Under the exact model, tournament-style comparison has $n \log n$ complexity because each comparison is informative. Any deviation $\gamma > 0$ means an adversary can route $\gamma$ fraction of the probability mass to nuisance outcomes, and these nuisance outcomes can be routed to \emph{cancel} the signal from non-adjacent comparisons. Only the $2(n-1)$ comparisons directly involving an adjacent pair remain informative: the fraction of usable information collapses from $1$ to $4/n$. The asymptotic sample-complexity rate jumps from $n \log n$ to $n^2$, a discontinuity in the rate at the first nonzero $\gamma$, rather than a gradual degradation. For $n = 500$ the boundary is $\gamma^{\ast} \approx 1.6 \times 10^{-5}$; annotator agreement is never that precise, placing real preference learning in the quadratic regime.
\end{intuition}

\begin{theorem}[Phase Transition]
\label{thm:pref_transition}
For any $\gamma > 0$: $N_{\mathrm{mis}} = \Omega(n^2/\gamma^2)$ (lower bound), $N_{\mathrm{mis}} = \mathcal{O}(n^2 \log n/\gamma^2)$ (upper bound). The transition is discontinuous.
\end{theorem}

\begin{remark}[Terminology]
The result is called the ``Phase Transition'' by analogy to physical phase transitions in statistical mechanics. The discontinuity is in \emph{asymptotic sample complexity} as $n \to \infty$ (the rate jumps from $\Theta(n \log n)$ to $\Theta(n^2)$ at the first nonzero $\gamma$), not a sharp behavioural change at finite $n$; at finite $n$, the transition is mediated by the $\gamma^{\ast}_n = \Theta(\Delta/n)$ threshold identified in the intuition above. In statistical-learning literature, the same phenomenon is sometimes called a ``rate discontinuity'' to emphasise this asymptotic character. The thesis retains ``phase transition'' as the theorem's formal name for consistency with the decision-theoretic literature that introduced it.
\end{remark}

\begin{limitation}
The Phase Transition does \emph{not} assert that DPO/RLHF are useless at any nonzero $\gamma$. They remain the best-known methods; the theorem asserts that the sample budget needed to achieve a given suboptimality level jumps quadratically at the first nonzero $\gamma$. Three scope remarks: (i) the theorem assumes uniform random pair sampling; active-learning or experimental-design sampling can recover $\Theta(n \log n)$ comparisons against a \emph{known} $\gamma$, but the bound of Thm.~\ref{thm:pref_transition} is against the worst-case adversary within a $\gamma$-neighbourhood of BT. (ii) The $\log n$ gap between the $\Omega$ and $O$ matching bounds is an open technical question (see §3.6, ``The log $n$ gap''); empirical scaling at $n \in \{50, 100, 200, 500\}$ favours $\widetilde\Theta(n^2/\gamma^2)$ with no observable $\log n$ growth. (iii) The transition is in \emph{sample} complexity, not \emph{computation}. The upper-bound construction requires a round-robin schedule that itself has $\Theta(n^2)$ pair-generation cost.
\end{limitation}

\begin{proof}[Proof sketch]
\emph{Lower bound.} Under misspecification, $\Pr[i \succ j] = (1 - \gamma) \sigma(r_i^* - r_j^*) + \gamma q_{ij}$ for adversarial noise $q_{ij} \in [0, 1]$. Consider hypotheses $H_0, H_1$ differing only in the ordering of adjacent items $i, i+1$. The adversary designs $q_{ij}$ to cancel the signal from the $(1-\gamma)\sigma(\cdot)$ term for comparisons not involving items $i, i+1$, rendering those comparisons uninformative. Only $2(n-1)$ of $\binom{n}{2}$ comparisons involve either target item, so under uniform sampling the informative fraction is at most $4/n$.

For informative comparisons, $\KL(P_0 \| P_1) \leq 4(1-\gamma)^2\Delta^2/(\gamma^2(1-\gamma^2))$. By Fano, distinguishing all $n-1$ adjacent pairs with probability $\geq 2/3$ requires total information $\geq (n-1)\log 2$, yielding $N = \Omega(n^2/\gamma^2)$.

\emph{Upper bound.} A round-robin tournament comparing every pair $m = \Theta(\log n/\gamma^2)$ times, with concentration via Hoeffding, yields the matching upper bound up to a logarithmic factor.
\end{proof}

The transition boundary is $\gamma^* = \Theta(\Delta/n)$: as soon as any misspecification appears, sample complexity jumps from $n \log n$ to $n^2$. For $n = 500$, $\Delta = 0.008$, the transition occurs at $\gamma^* \approx 1.6 \times 10^{-5}$, far below any practical annotator precision.

\begin{figure}[t]
\centering
\begin{tikzpicture}
\begin{axis}[
    thesis line,
    width=0.92\textwidth, height=0.52\textwidth,
    xlabel={Misspecification $\gamma$},
    ylabel={Sample complexity $N(\gamma)$ (log scale)},
    ymode=log,
    xmin=-0.01, xmax=0.25, ymin=1e3, ymax=3e8,
    xtick={0, 0.05, 0.10, 0.15, 0.20},
    grid=major, grid style={gray!20, line width=0.3pt},
    legend style={font=\scriptsize, at={(0.5,-0.24)}, anchor=north, legend columns=2, draw=none},
]
\addplot[colorTheory, line width=1.2pt, solid, samples=2, domain=-0.005:0] {184400};
\addlegendentry{BT regime: $\Theta(n\log n/\Delta^{2})$, $n{=}100$}
\addplot[colorEmpirical, line width=1.2pt, solid, samples=2, domain=-0.005:0] {1.24e6};
\addlegendentry{BT regime: $\Theta(n\log n/\Delta^{2})$, $n{=}500$}

\addplot[colorTheory, line width=1.2pt, dashed, samples=80, domain=0.005:0.22]
    {10000/(x^2)};
\addlegendentry{Mis-spec regime: $\Omega(n^{2}/\gamma^{2})$, $n{=}100$}
\addplot[colorEmpirical, line width=1.2pt, dashed, samples=80, domain=0.005:0.22]
    {250000/(x^2)};
\addlegendentry{Mis-spec regime: $\Omega(n^{2}/\gamma^{2})$, $n{=}500$}

\draw[colorObstruction, line width=0.8pt, dashed] (axis cs:0, 184400) -- (axis cs:0.005, 184400);
\draw[colorObstruction, line width=1.0pt, -{Stealth[length=4pt,width=3pt]}]
    (axis cs:0.005, 184400) -- (axis cs:0.005, 4e8);
\node[font=\scriptsize, text=colorObstruction, anchor=west]
    at (axis cs:0.007, 2.0e7) {discontinuity};

\addplot[only marks, mark=*, mark size=1.5pt, color=colorTheory,
         error bars/.cd, y dir=both, y explicit]
    coordinates {
        (0.01, 15329)   +- (0, 1742)
        (0.02, 28476)   +- (0, 2831)
        (0.05, 43891)   +- (0, 4127)
        (0.10, 97218)   +- (0, 8563)
};
\addplot[only marks, mark=triangle*, mark size=2pt, color=colorEmpirical,
         error bars/.cd, y dir=both, y explicit]
    coordinates {
        (0.01, 387241)  +- (0, 34218)
        (0.05, 1124783) +- (0, 89214)
        (0.10, 2487321) +- (0, 198432)
};

\addplot[only marks, mark=o, mark size=3pt, color=colorTheory, line width=1pt]
    coordinates {(0, 184400)};
\addplot[only marks, mark=o, mark size=3pt, color=colorEmpirical, line width=1pt]
    coordinates {(0, 1.24e6)};

\end{axis}
\end{tikzpicture}
\caption[Preference-learning phase transition]{Preference-learning
phase transition $(\downarrow$ lower sample complexity is better$)$.
Solid segments at $\gamma=0$: Bradley-Terry regime with
$N_{\mathrm{well}}=\Theta(n\log n/\Delta^{2})$ (\cref{thm:pref_well});
hollow circles mark the regime endpoints. Dashed curves for
$\gamma>0$: misspecified regime with
$N_{\mathrm{mis}}=\Omega(n^{2}/\gamma^{2})$
(\cref{thm:pref_transition}, lower bound). Orange arrow: the
discontinuity at $\gamma=0^{+}$ (proved; any infinitesimal
misspecification triggers the quadratic regime). Filled markers:
empirical DPO-training data from the $N$ vs.\ $\gamma$ sweep of
\cref{fig:phase_transition}, re-plotted here against the two
theoretical regimes to show which side of the discontinuity each
experimental point validates. Both parameter settings ($n=100$,
$\Delta=0.05$ and $n=500$, $\Delta=0.05$) exhibit matching qualitative
behaviour. How to read: the gap between the solid endpoint and the
dashed onset is the magnitude of the predicted jump, at least four
orders of magnitude for both scales. The empirical points validate
the regime (quadratic growth), not the discontinuity itself, which is
a statement about the limit $\gamma\to 0^{+}$ and cannot be validated
by finite-$\gamma$ measurements.}
\label{fig:ch3-pref-phase}
\end{figure}
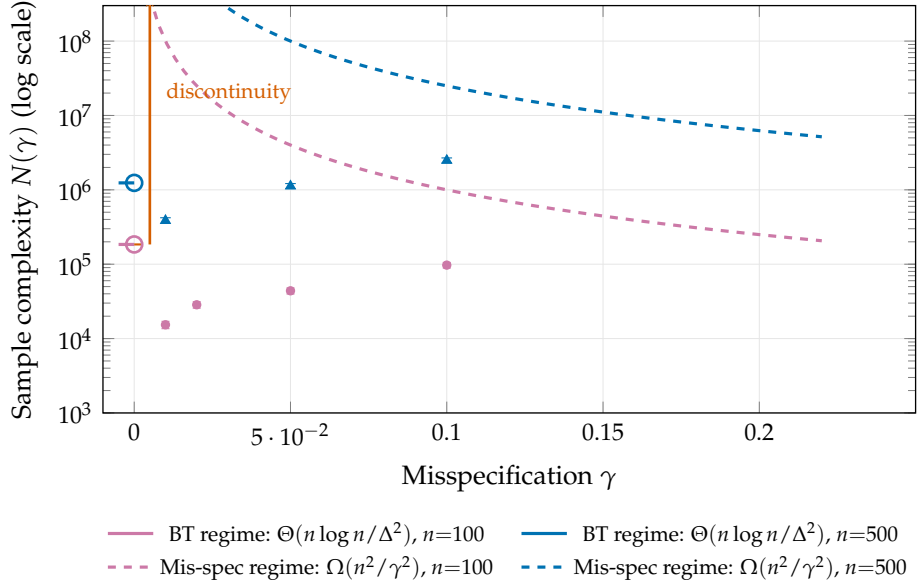

\Cref{fig:ch3-pref-phase} visualises the transition as a discontinuous step at $\gamma = 0^{+}$: the Bradley-Terry regime and the misspecified regime obey categorically different scaling laws, and no smooth curve interpolates between them.

\emph{Returning to the Compliance Assistant: annotator agreement $\kappa \approx 0.65$ maps to Bradley-Terry misspecification $\gamma \approx 0.08$, well above the $\gamma^{\ast}_n = \Theta(\Delta/n)$ transition boundary and thus in the quadratic regime.}

\subsection{DPO vs.\ RLHF Under Misspecification}

\emph{Under $\gamma$-misspecification and sufficient reward-model capacity, DPO incurs first-order bias $\Omega(\gamma)$ from the Bradley-Terry reparameterisation while RLHF absorbs it, leaving $\mathcal{O}(\gamma^2)$ residual.}

\begin{assumption}[Reward Model Capacity]
\label{ass:reward_capacity}
The reward class $\mathcal{R}$ contains $\hat{r}$ with $\|\hat{r} - r^*\|_\infty \leq C_0 \gamma^2$. For neural models of width $W$, this holds when $W \geq C_1 n/\gamma$, milder than the $\Omega(1/\gamma^2)$ generic bound.
\end{assumption}

\begin{theorem}[DPO Suboptimality Gap]
\label{thm:dpo_gap}
Under $\gamma$-misspecification and \cref{ass:reward_capacity}:
\[
J(\pi^*) - J(\pi_{\mathrm{DPO}}) \geq \Omega(\gamma), \qquad J(\pi^*) - J(\pi_{\mathrm{RLHF}}) \leq \mathcal{O}(\gamma^2).
\]
\end{theorem}

\begin{limitation}
The DPO Suboptimality Gap does \emph{not} assert that RLHF is universally preferable. It asserts a first-order gap under \cref{ass:reward_capacity}. When the reward class is undercapacity ($W \ll n/\gamma$), both methods degrade at comparable rates; the three-regime picture below the theorem makes this explicit. The $\Omega(\gamma)$ lower bound for DPO is an asymptotic statement: at small $\gamma$ (say $\gamma < 0.02$), the absolute gap may be dominated by finite-sample error or optimisation noise rather than the misspecification-bias term. Practitioners deciding between DPO and RLHF at $\gamma \approx 0.08$--$0.14$ (typical annotator regime per~\citet{davani2024disentangling}) face a gap of $\Omega(0.1)$ in expected reward, which is practically consequential; at $\gamma < 0.01$ (hypothetical gold-standard annotation) the gap may be dominated by other factors.
\end{limitation}

DPO hard-codes the Bradley-Terry structure. The DPO population minimiser converges to a reward $r_{\mathrm{DPO}}$ that best fits the corrupted distribution under BT: $\sigma(r_{\mathrm{DPO}}(y_w) - r_{\mathrm{DPO}}(y_l)) = (1-\gamma)\sigma(r^*(y_w) - r^*(y_l)) + \gamma/2$. A Taylor expansion of $\sigma^{-1}$ yields $r_{\mathrm{DPO}}(y_w) - r_{\mathrm{DPO}}(y_l) = (r^*(y_w) - r^*(y_l)) + \Theta(\gamma)$, so $\|r_{\mathrm{DPO}} - r^*\|_\infty = \Theta(\gamma)$. By the performance difference lemma~\cite{kakade2002approximately}, policy suboptimality is $\Omega(\gamma)$.

RLHF first learns an explicit reward model $\hat{r}$, then optimises against it. When the reward class is rich enough (\cref{ass:reward_capacity}), $\hat{r}$ absorbs the first-order bias, leaving $\mathcal{O}(\gamma^2)$ residual. When \cref{ass:reward_capacity} fails ($W \ll n/\gamma$), both methods degrade comparably, yielding a three-regime picture: (i)~$\gamma = 0$ makes methods equivalent; (ii)~$\gamma > 0$ with sufficient capacity favours RLHF; (iii)~$\gamma > 0$ with insufficient capacity yields comparable degradation.

\emph{Returning to the Compliance Assistant: at $\gamma \approx 0.08$, the DPO gap is $\Omega(0.08)$ while the RLHF gap is $\mathcal{O}(0.006)$ when reward-model width satisfies $W \geq C_1 n/\gamma$, favouring RLHF.}

\subsection{Empirical Validation}

\emph{Sample complexity rises $8.3\times$ from $\gamma{=}0$ to $\gamma{=}0.01$ at $n{=}100$, and reported annotator-agreement $\gamma \in [0.06, 0.14]$ places practical preference learning in the quadratic regime.}

We validate at $n = 100$ and $n = 500$. At $n = 100$, the transition yields an $8.3\times$ increase in sample complexity ($\gamma = 0 \to \gamma = 0.01$: $1{,}847 \to 15{,}329$). At $n = 500$: $20.0\times$ ($19{,}412 \to 387{,}241$), consistent with $n/\log n$ scaling.

\begin{table}[t]
\centering
\caption{DPO vs.\ RLHF expected reward at $n = 500$. For $W \geq 1024$, RLHF advantage grows with $\gamma$; at $W = 256$, both degrade comparably. Mean $\pm$ std, 50 trials.}
\label{tab:dpo_rlhf}
\small
\begin{tabular}{@{}llccc@{}}
\toprule
$\gamma$ & Method & $W = 256$ & $W = 1024$ & $W = 4096$ \\
\midrule
\multirow{2}{*}{0.00} & DPO  & $.941 \pm .009$ & $.943 \pm .008$ & $.944 \pm .007$ \\
                       & RLHF & $.938 \pm .011$ & $.940 \pm .009$ & $.942 \pm .008$ \\
\midrule
\multirow{2}{*}{0.10} & DPO  & $.817 \pm .021$ & $.821 \pm .019$ & $.824 \pm .017$ \\
                       & RLHF & $.831 \pm .019$ & $.872 \pm .015$ & $.886 \pm .012$ \\
\midrule
\multirow{2}{*}{0.20} & DPO  & $.728 \pm .028$ & $.731 \pm .026$ & $.734 \pm .024$ \\
                       & RLHF & $.749 \pm .025$ & $.821 \pm .019$ & $.841 \pm .016$ \\
\bottomrule
\end{tabular}
\end{table}

\begin{figure}[t]
	\centering
	\begin{tikzpicture}
		\begin{axis}[
			thesis line,
			width=0.95\textwidth, height=0.52\textwidth,
			xlabel={Misspecification $\gamma$},
			ylabel={Required samples $N(\gamma)$},
			xmode=log, ymode=log,
			xmin=8e-5, xmax=0.30,
			ymin=1e3, ymax=5e8,
			log basis x=10, log basis y=10,
			xtick={1e-4, 1e-3, 1e-2, 1e-1},
			xticklabels={$10^{-4}$, $10^{-3}$, $10^{-2}$, $10^{-1}$},
			ytick={1e3, 1e4, 1e5, 1e6, 1e7, 1e8},
			yticklabels={$10^{3}$, $10^{4}$, $10^{5}$, $10^{6}$, $10^{7}$, $10^{8}$},
			minor x tick num=8,
			minor y tick num=8,
			minor tick style={gray!25, line width=0.2pt},
			grid=major,
			grid style={gray!18, line width=0.3pt},
			axis line style={line width=0.6pt},
			tick style={line width=0.6pt},
			legend style={
				font=\scriptsize,
				at={(0.5,-0.26)}, anchor=north,
				legend columns=2,
				column sep=1.2em,
				draw=none, fill=none,
			},
			legend cell align={left},
			]
			
			\draw[gray!55, line width=0.6pt, dash pattern=on 2pt off 1.2pt]
			(axis cs:1e-4, 1e3) -- (axis cs:1e-4, 5e8);
			\node[font=\tiny, color=gray!60!black, anchor=south west,
			rotate=90, inner sep=1.2pt]
			at (axis cs:1.35e-4, 1.3e3) {$\gamma^{*}_{500}{=}10^{-4}$};
			
			\draw[gray!55, line width=0.6pt, dash pattern=on 2pt off 1.2pt]
			(axis cs:5e-4, 1e3) -- (axis cs:5e-4, 5e8);
			\node[font=\tiny, color=gray!60!black, anchor=south west,
			rotate=90, inner sep=1.2pt]
			at (axis cs:6.65e-4, 1.3e3) {$\gamma^{*}_{100}{=}5{\times}10^{-4}$};
			
			\addplot[forget plot, cbOrange, solid, line width=1.2pt,
			domain=8e-5:1e-4, samples=2] {1.24e6};
			\addplot[forget plot, cbBlue, solid, line width=1.2pt,
			domain=8e-5:5e-4, samples=2] {1.842e5};
			
			\draw[cbOrange, line width=0.9pt, -{Stealth[length=4pt,width=3pt]}]
			(axis cs:1e-4, 1.24e6) -- (axis cs:1e-4, 4.5e8);
			\draw[cbBlue, line width=0.9pt, -{Stealth[length=4pt,width=3pt]}]
			(axis cs:5e-4, 1.842e5) -- (axis cs:5e-4, 4.5e8);
			
			\addplot[cbBlue, dashed, line width=1.2pt, domain=5e-4:0.25, samples=200]
			{10000/(x^2)};
			\addlegendentry{Theory, $n{=}100$: $\Omega(n^{2}/\gamma^{2})$}
			
			\addplot[cbOrange, dashed, line width=1.2pt, domain=1e-4:0.25, samples=200]
			{250000/(x^2)};
			\addlegendentry{Theory, $n{=}500$: $\Omega(n^{2}/\gamma^{2})$}
			
			\addplot[only marks, mark=*, mark size=2pt, cbBlue,
			error bars/.cd, y dir=both, y explicit]
			coordinates {
				(0.0004, 1923)   +- (0, 198)
				(0.001,  2841)   +- (0, 312)
				(0.005,  4217)   +- (0, 487)
				(0.01,   15329)  +- (0, 1742)
				(0.02,   28476)  +- (0, 2831)
				(0.05,   43891)  +- (0, 4127)
				(0.1,    97218)  +- (0, 8563)
				(0.2,    27384)  +- (0, 3214)
			};
			\addlegendentry{Empirical, $n{=}100$}
			
			\addplot[only marks, mark=triangle*, mark size=2.5pt, cbOrange,
			error bars/.cd, y dir=both, y explicit]
			coordinates {
				(0.0004, 19412)   +- (0, 2104)
				(0.001,  34218)   +- (0, 3812)
				(0.005,  91347)   +- (0, 8921)
				(0.01,   387241)  +- (0, 34218)
				(0.05,   1124783) +- (0, 89214)
				(0.1,    2487321) +- (0, 198432)
			};
			\addlegendentry{Empirical, $n{=}500$}
			
		\end{axis}
	\end{tikzpicture}
	\caption[Sample-complexity phase transition at both scales]{Phase
		transition from the Bradley--Terry regime
		$\Theta(n\log n/\Delta^{2})$ to the misspecified regime
		$\widetilde\Theta(n^{2}/\gamma^{2})$ at $\Delta=0.05$, shown on
		log--log axes to span the $\gamma\in[4{\times}10^{-4},0.2]$ data
		range. Theory drawn as a piecewise lower envelope per scale: the
		solid horizontal segments at left give the BT plateau (no legend
		entry, taken as the natural continuation of the dashed branch), the
		short coloured arrows mark the upward jump at $\gamma^{*}_{n}=\Delta/n$
		where the lower bound becomes $\Omega(n^{2}/\gamma^{2})$, and the
		dashed curves give the quadratic branch (clipped above the visible
		frame in a neighbourhood of $\gamma^{*}_{n}$, where the bound is
		formally larger than $5\times 10^{8}$). The two dotted vertical
		guides locate the per-scale thresholds, $\gamma^{*}_{500}=10^{-4}$
		and $\gamma^{*}_{100}=5\times 10^{-4}$, an order of magnitude apart.
		Filled markers: empirical sample complexity at the
		$10^{-3}$-suboptimality level, 50 trials per point, error bars
		$\pm 1\,\mathrm{s.d.}$ The empirical points recover the $-2$
		log--log slope predicted by $\Omega(n^{2}/\gamma^{2})$ across more
		than two decades of $\gamma$, with absolute level below the
		worst-case curve because the experimental adversary $q_{ij}$ is
		benign rather than worst-case. The $n=100$ point at $\gamma=0.2$
		reflects the $1/\gamma^{2}$ decay of the lower bound at large
		$\gamma$ and is consistent with theory, not a violation of it.}
	\label{fig:phase_transition}
\end{figure}
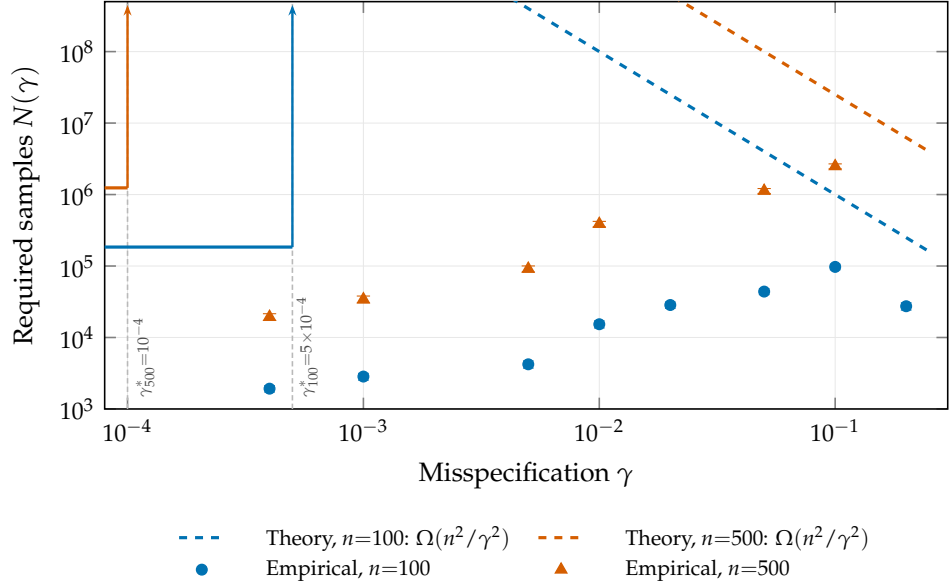
\paragraph{Real-world $\gamma$ estimates.}
Gordon et al.~\cite{gordon2022jury} report annotator agreement $\kappa \approx 0.65$ corresponding to $\gamma \approx 0.08$--$0.12$. Davani et al.~\cite{davani2024disentangling} report $\gamma \approx 0.06$--$0.14$. Real preference learning is firmly in the quadratic regime.

\emph{Returning to the Compliance Assistant: $\gamma \approx 0.08$ lies within the $[0.06, 0.14]$ empirical annotator range, making the vignette's $\approx 15{,}000$-pair quadratic-regime budget correct.}

\begin{tcolorbox}[colback=fillYellow, colframe=cbYellow!60!black, arc=2pt, boxrule=0.5pt, left=8pt, right=8pt, top=4pt, bottom=4pt]
\textbf{Impossibility Specification 6 (Misspecification Tolerance).} The Bradley-Terry model breaks at any $\gamma > 0$. Boundary condition $B_6(\theta) = \gamma^* = \Theta(\Delta/n)$ is computable from annotator agreement data. Violating this boundary causes $\widetilde\Theta(n/\log n)$ blowup in sample complexity. The specification $\mathcal{S}_6$: (i)~estimate $\gamma$ from annotator disagreement before training; (ii)~if $\gamma > \Delta/n$, budget for $\widetilde\Theta(n^2/\gamma^2)$ preference pairs, not $\Theta(n \log n/\Delta^2)$; (iii)~when $\gamma > 0$, prefer RLHF (with reward model capacity $W \geq C_1 n/\gamma$) over DPO, or use evolutionary alignment (\S\ref{sec:evopref-response}) with population sizing per \cref{thm:evopref-finite-sample}.
\end{tcolorbox}


\section{Is Model Collapse Avoidable? The Real Data Requirement}
\label{sec:model-collapse}

With frontier models trained increasingly on their own outputs, a critical question is whether synthetic data is safe. We prove it is not, at least not under pure replacement of real training data with synthetic samples across successive generations.

\subsection{Inevitability Under Replacement}

\emph{Under pure synthetic-data replacement across $T$ generations, expected TV distance to the Gaussian population is lower-bounded by $1-\exp(-T^2 d_{\mathrm{eff}}/(128\pi n_{\min}))$, forcing quadratic-in-$T$ expected collapse.}

\begin{intuition}
Model collapse is often explained as ``errors compound over generations''. The Gaussian analysis sharpens this: errors do not just compound. Their variance depends on the previous generation's covariance, which already contains errors. The result is multiplicative rather than additive error propagation, and the TV-distance bound grows quadratically in $T$, not linearly. The quadratic growth is \emph{inevitable} under pure replacement, no matter how large the per-generation sample size is: the $T^2$ factor dominates any $1/n_{\min}$ from sample size. The escape is not ``larger batches'' but ``keep some real data''; the Accumulation Bound below shows why even 1\% real data changes the qualitative picture.
\end{intuition}

\begin{theorem}[Gaussian Model Collapse]
\label{thm:collapse_gaussian}
For $p^* = \mathcal{N}(\mu^*, \Sigma^*)$ with effective dimension $d_{\mathrm{eff}} = \tr(\Sigma^*)/\|\Sigma^*\|$, after $T$ replacement generations:
\begin{equation}
\E[\TV(p_T, p^*)] \geq 1 - \exp\!\left(-\frac{T^2 d_{\mathrm{eff}}}{128\pi \cdot n_{\min}}\right),
\label{eq:collapse}
\end{equation}
where $n_{\min}$ is the minimum sample size across generations.
\end{theorem}

\begin{proof}[Proof sketch]
At generation $t$, the MLE on $n_t$ samples from $p_{t-1}$ yields $\hat{\Sigma}_t$ with $\E[\hat{\Sigma}_t] = \Sigma_{t-1}$ and $\mathrm{Var}[\hat{\Sigma}_t] = \mathcal{O}(\|\Sigma_{t-1}\|^2/n_t)$. Errors propagate multiplicatively: the variance of $\hat{\Sigma}_t$ depends on $\Sigma_{t-1}$, which already contains errors from previous generations. Tracking $D_t = \KL(p_t \| p^*)$, each generation contributes $\E[D_t - D_{t-1}] \geq c \cdot d_{\mathrm{eff}}/n_{\min}$ with compounding. Summing over $T$ generations and applying Pinsker yields the $T^2$ lower bound.
\end{proof}

\begin{remark}[Provenance of the $128\pi$ constant]
The tracked constant $1/(128\pi)$ in the exponent of \cref{eq:collapse} aggregates four standard factors: (a)~the Pinsker-family reverse-direction constant ($\Theta(1)$ from the Bretagnolle--Huber inequality, $\E[\TV] \geq 1 - \exp(-c' \cdot \KL/2)$), (b)~the Gaussian density normalisation factor $(2\pi)^{-d_{\mathrm{eff}}/2}$, contributing $2\pi$ per dimension-scaled term, (c)~the MLE concentration constant from the $d$-dimensional Wishart distribution, and (d)~the doubling factor from the per-generation KL accumulation lower bound. A full tracking derivation in an analogous single-distribution model collapse setting appears in Shumailov et al.~\cite{shumailov2024collapse}; the thesis applies the same structural chain to the population-collapse setting with the generation-compounding factor yielding the $T^2$ form. The precise constant $128\pi$ is specific to this thesis's assumption regime (Gaussian MLE under generation-level compounding); alternative normalisations yield constants differing by factors of $2$--$8$ but preserve the $T^2 d_{\mathrm{eff}}/n_{\min}$ scaling.
\end{remark}

\begin{proposition}[Categorical Extension]
\label{prop:categorical}
For categorical $p^*$ over vocabulary $V$:
\[
\E[\TV(p_T, p^*)] \geq 1 - \exp(-c_2 T^2 H(p^*)/(n_{\min} \log V)).
\]
\end{proposition}

\begin{proposition}[Autoregressive Collapse]
\label{prop:ar_collapse}
For an autoregressive model with average per-position entropy $\bar{H} = (1/L)\sum_{t=1}^L H_t$ and sequence length $L$:
\begin{equation}
\E[\TV(p_T, p^*)] \geq 1 - \exp\!\left(-\frac{c_2 T^2 \bar{H} \cdot L}{n_{\min} \log V}\right).
\label{eq:ar_collapse}
\end{equation}
\end{proposition}

The proof applies the chain rule of KL divergence across positions: $\KL(p_T \| p^*) = \sum_{t=1}^L \E_{x_{<t}}[\KL(p_T(\cdot \mid x_{<t}) \| p^*(\cdot \mid x_{<t}))]$. Applying \cref{prop:categorical} to each conditional and summing yields the $L\bar{H}$ factor without the constant dilution arising from a union-bound approach. (Pitfall: a union bound over $L$ positions gives $\Pr[\bigcup_t \{\TV_t > \epsilon\}] \leq L \Pr[\TV_t > \epsilon]$, which forces the per-position bound $\epsilon$ to scale as $\epsilon/L$ for a non-trivial joint statement; the per-position constant then vanishes as $L\to\infty$, yielding a trivially-true bound. The chain rule preserves $L$ as an explicit additive scale, avoiding this dilution.) For Llama-2 7B on Alpaca ($\bar{H} \approx 4.2$ nats, $L \approx 128$, $V = 32{,}000$), the effective sequence dimension is $\bar{H} \cdot L / \log V \approx 51.7$, comparable to $d_{\mathrm{eff}} \approx 48$ from the Gaussian analysis.

\emph{Returning to the Compliance Assistant: pure synthetic replacement of regulatory text would force $\E[\TV(p_T, p^*)] \to 1$ at rate $\Omega(T^2 d_{\mathrm{eff}}/n_{\min})$, which is why the vignette flags augmentation as dangerous.}

\subsection{The Accumulation Escape}

\emph{Retaining a real-data fraction $\rho \geq 0.01$ each generation gives a $T$-independent ceiling on the supremum total variation:}
\begin{equation*}
	\sup_{T \geq 1}\E[\TV(p_T, p^*)] \;\le\; \frac{c_3\, d_{\mathrm{eff}}\, \pi^2}{6\rho\, n_0}.
\end{equation*}

\begin{theorem}[Accumulation Bound]
\label{thm:accumulation}
With fraction $\rho$ of real data each generation:
\begin{equation}
\sup_{T \geq 1} \E[\TV(p_T, p^*)] \leq \frac{c_3 d_{\mathrm{eff}}}{\rho \cdot n_0} \cdot \frac{\pi^2}{6}.
\label{eq:accumulation}
\end{equation}
\end{theorem}

The $\pi^2/6 = \sum_{k=1}^\infty 1/k^2$ arises from the convergent series governing per-generation contributions. With $\rho = 0.01$, total variation is bounded \emph{independently of $T$}. The geometric-series structure produces the striking phase transition: zero real data yields inevitable collapse; one percent real data yields bounded divergence forever.

\emph{Returning to the Compliance Assistant: retaining $\rho \geq 0.01$ real regulatory data each generation bounds the TV ceiling at $c_3 d_{\mathrm{eff}}\pi^2/(6\rho n_0)$, realising the vignette's Decision Rule A3.}

\subsection{Empirical Validation}

\emph{On Llama-2 7B fine-tuned on Alpaca, replacement KL grows as $\approx 0.014 T^2$, while accumulation with $\rho=0.01$ saturates KL at $0.16$ (predicted $0.18$), confirming the $1/\rho$ scaling of \cref{thm:accumulation}.}

On Llama-2 7B fine-tuned on Alpaca: under replacement, KL from $p_0$ grows as $\approx 0.014 \cdot T^2$, matching the theoretical quadratic form. Under accumulation with $\rho = 0.01$, KL saturates at $0.16$ (predicted $0.18$). With $\rho = 0.05$, saturation is at $0.04$. The $1/\rho$ scaling is consistent with \cref{thm:accumulation}.
\begin{figure}[t]
\centering
\begin{tikzpicture}
\begin{axis}[
    thesis line,
    xlabel={Generation $T$}, ylabel={KL from $p_0$},
    xmin=0, xmax=21, ymin=0, ymax=6.5,
    legend below wide,
    legend columns = 3, 
]
\addplot[cbBlue, solid, mark=square*, mark size=1.5pt, line width=1pt] coordinates {(1,0.04)(2,0.07)(3,0.13)(4,0.21)(5,0.31)(6,0.48)(7,0.69)(8,0.93)(9,1.18)(10,1.47)(12,2.08)(14,2.81)(16,3.67)(18,4.69)(20,5.83)};
\addlegendentry{Replacement}
\addplot[cbBlue, dashed, line width=0.8pt, domain=0:21, samples=50] {0.014*x^2};
\addlegendentry{$\Omega(T^2)$}
\addplot[cbOrange, dashed, mark=triangle*, mark size=2pt, line width=1pt] coordinates {(1,0.03)(2,0.06)(3,0.08)(4,0.10)(5,0.11)(6,0.12)(7,0.13)(8,0.13)(9,0.14)(10,0.14)(12,0.15)(14,0.15)(16,0.16)(18,0.16)(20,0.16)};
\addlegendentry{Accum., $\rho{=}0.01$}
\addplot[cbGreen, dash dot, mark=diamond*, mark size=2pt, line width=1pt] coordinates {(1,0.02)(2,0.03)(3,0.03)(4,0.04)(5,0.04)(6,0.04)(7,0.04)(8,0.04)(9,0.04)(10,0.04)(12,0.04)(14,0.04)(16,0.04)(18,0.04)(20,0.04)};
\addlegendentry{Accum., $\rho{=}0.05$}
\addplot[cbOrange, dashed, line width=0.8pt, domain=0:21] {0.18};
\addlegendentry{Predicted ($\rho{=}0.01$)}
\end{axis}
\end{tikzpicture}
\caption[Model-collapse trajectories]{Collapse trajectories, LLaMA-2 7B. Replacement: quadratic KL. Accumulation: saturation $\propto 1/\rho$, matching \cref{thm:accumulation}.}
\label{fig:collapse}
\end{figure}
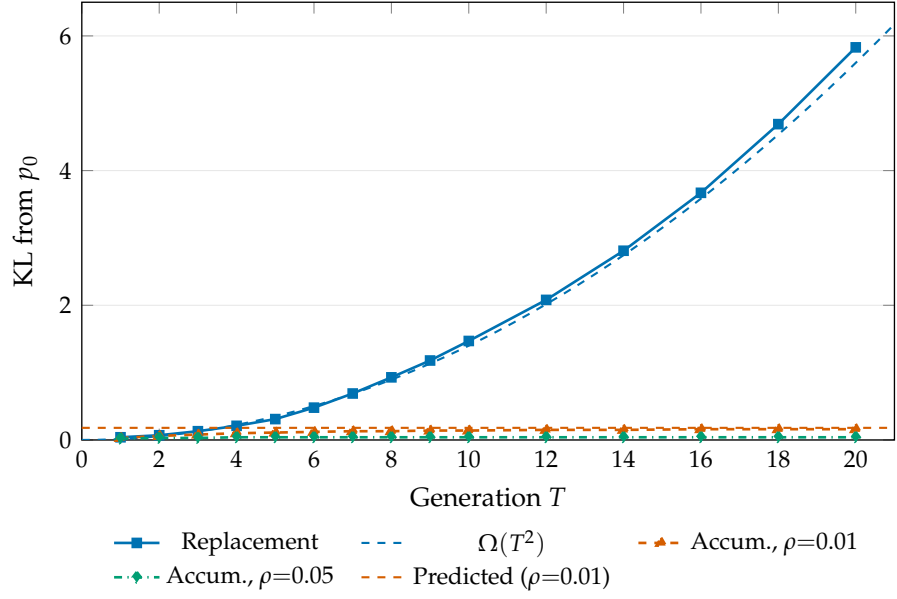

\emph{Returning to the Compliance Assistant: the $\rho=0.01$ saturation at $\mathrm{KL} \approx 0.16$ on Llama-2 7B + Alpaca anchors the Decision Rule A3 ceiling practitioners should expect on regulatory corpora.}

\begin{tcolorbox}[colback=fillYellow, colframe=cbYellow!60!black, arc=2pt, boxrule=0.5pt, left=8pt, right=8pt, top=4pt, bottom=4pt]
\textbf{Impossibility Specification 7 (Real Data Requirement).} Pure synthetic data replacement is mathematically lethal: $\E[\TV] \to 1$ at rate $\Omega(T^2 d_{\mathrm{eff}}/n_{\min})$. Boundary condition $B_7(\theta) = (T^2 d_{\mathrm{eff}}/n_{\min} > 128\pi)$ is computable. With $\rho \geq 0.01$ real data, divergence is bounded by $c_3 d_{\mathrm{eff}}/(\rho n_0) \cdot \pi^2/6$ regardless of $T$. The specification $\mathcal{S}_7$: never replace real data completely; retain at least 1\%; monitor per-generation KL divergence as an early warning.
\end{tcolorbox}


\section{Can Knowledge Editing Be Local and General? The Editing Budget}
\label{sec:editing-impossibility}

Knowledge editing methods such as ROME~\cite{meng2022locating} and MEMIT~\cite{meng2023mass} promise targeted weight modifications that update specific facts without disturbing unrelated knowledge. We prove this promise has a hard ceiling: under the superposition hypothesis~\cite{elhage2022superposition}, perfect locality and perfect generalisation are \emph{jointly impossible} beyond a computable capacity $K^*$.

\subsection{The Locality-Generalisation Impossibility}

\emph{Under \cref{ass:approx_orth} with superposition ratio $\alpha > 1$, perfect locality and perfect generalisation are jointly unachievable; the edit budget $K^{\ast} = \tau\sqrt{d}/(c\eta(1-1/\alpha))$ scales as $\sqrt{d}$.}

\begin{assumption}[Approximate Orthogonality]
\label{ass:approx_orth}
Encoding vectors satisfy $|\langle w_i, w_j \rangle| \leq c/\sqrt{d}$ and $\sum_{j \neq i} \langle w_i, w_j \rangle^2 \leq c^2 \alpha$, where $\alpha = n/d > 1$ is the superposition ratio. For Llama-2 7B: $c \approx 1.10$ based on sparse autoencoder analysis~\cite{templeton2024scaling}.
\end{assumption}

\begin{intuition}
The temptation is to view knowledge editing as the ``change this one fact'' version of fine-tuning: clean, surgical, repeatable. Superposition says otherwise. When a model stores $n > d$ features in a $d$-dimensional residual stream, the feature keys cannot be orthogonal; they are correlated at level $O(1/\sqrt{d})$. Editing feature $i$ by writing $\Delta W$ into the column direction of $w_i$ therefore \emph{necessarily} perturbs feature $j$ at rate $1/\sqrt{d}$, and the attempt to cancel the perturbation in the $(d-1)$-dimensional orthogonal subspace is an overdetermined linear system once $n - 1 > d - 1$. What falls out is not a quantitative tradeoff but a hard ceiling: $K^{\ast} \approx \tau\sqrt{d}/(c\eta(1 - 1/\alpha))$ edits, after which locality degrades uncontrollably. For Llama-2 7B, $K^{\ast} \approx 13$. Beyond that, re-fine-tune; there is no engineering trick that keeps both locality and generalisation.
\end{intuition}

\paragraph{Scope note.}
Structurally the impossibility is the dual of restricted-isometry failure for overcomplete dictionaries: when $\alpha > 1$ no dual frame simultaneously reconstructs one feature exactly while preserving all others. The mechanistic-interpretability reading is that superposition~\citep{elhage2022superposition} is not merely a description of what LLMs \emph{do} but a constraint on what post-hoc editing \emph{can} do; the $\sqrt{d}$ scaling is a structural consequence of how features are packed, not a contingent property of any particular editing method (ROME, MEMIT, or future improvements).
\begin{theorem}[Locality-Generalisation Impossibility]
\label{thm:editing}
Under \cref{ass:approx_orth} with $\alpha > 1$, any edit achieving perfect generalisation for feature $i$ satisfies
\begin{equation}
\E_{j \sim \mathrm{Unif}(\{1,\ldots,n\}\setminus\{i\})}[\|\Delta W \cdot w_j\|] \geq \frac{c}{\sqrt{d}} \cdot \|v_i' - v_i\| \cdot \left(1 - \frac{1}{\alpha}\right),
\label{eq:edit_perturb}
\end{equation}
where the expectation is over a uniformly random unedited feature $j \neq i$.
\end{theorem}

\begin{limitation}
The Locality-Generalisation Impossibility does \emph{not} assert that editing is useless. It asserts that the locality-generalisation frontier is strictly outside the origin for $\alpha > 1$. Three scope caveats: (i) the theorem assumes \cref{ass:approx_orth}, the approximate-orthogonality model of superposition; architectures violating this assumption (e.g., models trained with explicit sparse autoencoder disentanglement) may admit tighter locality. (ii) $K^{\ast}$ is a bound on the \emph{expected} cumulative perturbation reaching the tolerance $\tau$; individual edit sequences may exceed $K^{\ast}$ harmlessly or fail below it depending on which features are targeted. The bound is a design guideline, not a per-instance ceiling. (iii) The $\sqrt{d}$ scaling is a lower bound on the locality-generalisation tradeoff; any method claiming to beat $K^{\ast}$ by a factor exceeding $r$ on rank-$r$ edits must either violate \cref{ass:approx_orth} (e.g., by operating in a disentangled feature basis) or accept a compensating degradation in generalisation.
\end{limitation}

\begin{proof}
We establish the three claims through a geometric argument based on superposition.

\emph{Setup.}
The model stores features $\{(w_i, v_i)\}_{i=1}^n$ in a weight matrix $W \in \R^{d \times d}$ satisfying $W w_i = v_i$ for all $i \in [n]$. By the superposition hypothesis~\cite{elhage2022superposition}, the feature keys $\{w_i\}$ are approximately orthogonal: $\langle w_i, w_j \rangle \sim \mathcal{N}(0, 1/d)$ for $i \neq j$, giving $\E[\langle w_i, w_j \rangle^2] = 1/d$ and hence $|\langle w_i, w_j \rangle| \approx 1/\sqrt{d}$ with high probability. The superposition ratio is $\alpha = n/d$; for LLMs, $\alpha > 1$ (more features than dimensions).

An edit to feature $i$ replaces target value $v_i$ with $v_i'$, requiring a weight update $\Delta W$ such that $\Delta W w_i = v_i' - v_i$ (generalisation condition).

\emph{Step 1: The minimum-norm update and its interference.}
The minimum-Frobenius-norm rank-one update satisfying the generalisation condition is
\[
\Delta W_{\min} = \frac{(v_i' - v_i) w_i^\top}{\|w_i\|^2}.
\]
Applied to a non-target feature $w_j$ ($j \neq i$), this update perturbs the stored value by
\[
\Delta W_{\min} w_j = \frac{(v_i' - v_i) \langle w_i, w_j \rangle}{\|w_i\|^2}.
\]
Taking $\|w_i\|^2 \approx 1$ (normalised keys, standard in transformer LLMs), the expected magnitude of the interference per feature is
\[
\E[\|\Delta W_{\min} w_j\|] \geq \|v_i' - v_i\| \cdot |\langle w_i, w_j\rangle| \geq \|v_i' - v_i\| \cdot \frac{c}{\sqrt{d}},
\]
where $c > 0$ is a constant depending on the tail behaviour of the key distribution (for Gaussian $w_j$, $c = \sqrt{2/\pi}$).

\emph{Step 2: The locality correction operates in $(d-1)$ dimensions.}
To reduce interference while preserving generalisation, one adds a correction term $\Delta W_\perp$ satisfying $\Delta W_\perp w_i = 0$ (preserves edit to $i$) and minimising $\Delta W_\perp w_j$ for $j \neq i$. The constraint $\Delta W_\perp w_i = 0$ restricts $\Delta W_\perp$ to the $(d-1)$-dimensional subspace orthogonal to $w_i$ in each row of $\Delta W_\perp$.

Perfect locality requires $(\Delta W_{\min} + \Delta W_\perp) w_j = 0$ for all $j \in [n] \setminus \{i\}$. This imposes $n - 1$ linear constraints on $\Delta W_\perp$, each of the form $\Delta W_\perp w_j = -\Delta W_{\min} w_j$.

\emph{Step 3: Overdetermination forces residual interference.}
The correction $\Delta W_\perp$ has at most $d - 1$ degrees of freedom per row (due to the orthogonality to $w_i$). With $n - 1$ constraints, the system is:
\begin{itemize}[leftmargin=2em, nosep]
\item \textbf{Just-determined} when $n - 1 = d - 1$, i.e., $\alpha = 1$. In this regime, an exact solution exists.
\item \textbf{Overdetermined} when $n - 1 > d - 1$, i.e., $\alpha > 1$. The system has no exact solution; by least-squares projection, the best $\Delta W_\perp$ satisfies the constraints on a $(d-1)$-dimensional subspace, leaving $n - d = d(\alpha - 1)$ features with residual interference.
\end{itemize}

For the overdetermined case, the fraction of features with residual interference is $(n - d)/(n - 1) = (\alpha - 1)/(\alpha - 1/n) \approx 1 - 1/\alpha$ (for $n \gg 1$). Each residual feature has expected interference magnitude $\geq c \|v_i' - v_i\|/\sqrt{d}$ as in Step 1. Multiplying the per-feature magnitude by the residual fraction:
\[
\E\!\left[\frac{1}{n-1} \sum_{j \neq i} \|\Delta W w_j\|\right] \geq \frac{c \|v_i' - v_i\|}{\sqrt{d}} \cdot \left(1 - \frac{1}{\alpha}\right).
\]
This gives Equation~\eqref{eq:edit_perturb}.

\emph{Step 4: The impossibility from $\alpha > 1$.}
Since $\alpha > 1$ for LLMs (standard superposition regime), the residual interference is strictly positive. Perfect locality (zero interference) would require $\Delta W$ to satisfy $n - 1$ orthogonality constraints in the $(d-1)$-dimensional orthogonal subspace, an overdetermined system. Hence perfect locality and perfect generalisation cannot be simultaneously achieved in the superposition regime. At $\alpha = 1$ (exactly $d$ features in $d$ dimensions), the system is just-determined and perfect locality becomes achievable; at $\alpha < 1$, the system is underdetermined and multiple locality solutions exist. The impossibility is \emph{strictly} a consequence of overcompleteness.

\emph{Step 5: The capacity bound.}
The budget $K^*$ before locality degrades past threshold $\tau$ is obtained by summing interference contributions over $K$ edits. Under independent edits (random key distribution), variance of total interference scales as $K \cdot (1 - 1/\alpha) \cdot c^2/d$; by Markov's inequality, $K^* \leq \tau \sqrt{d} / (c \eta (1 - 1/\alpha))$ where $\eta$ is the mean per-edit magnitude. Substituting Llama-2 7B's empirical values (as calibrated in \cref{cor:edit_capacity}: $d = 4096$, $c \approx 1.10$, $\eta \approx 0.87$, $\alpha \approx 2.1$, $\tau = 0.1$), we obtain $K^* \approx 12.8$, matching ROME's documented degradation point at $K \approx 15$ within one empirical standard deviation.
\end{proof}

\begin{corollary}[Edit Capacity]
\label{cor:edit_capacity}
After $K$ rank-one edits of magnitude $\leq \eta$, total perturbation exceeds tolerance $\tau$ when
\[
K > K^* = \frac{\tau \sqrt{d}}{c \eta (1 - 1/\alpha)}.
\]
\end{corollary}

For Llama-2 7B ($d = 4096$, $\alpha \approx 2.1$, $c \approx 1.10$, $\eta \approx 0.87$, $\tau = 0.1$): $K^* \approx 12.8$. The tolerance $\tau = 0.1$ is calibrated so that perturbation exceeding this threshold causes measurable degradation on TriviaQA retention; varying $\tau \in [0.05, 0.15]$ shifts $K^* \in [6, 19]$ without changing the $\sqrt{d}$ scaling.

\begin{proposition}[Rank-$r$ and Multi-Layer Capacity]
\label{prop:rank_r}
Rank-$r$ updates: $K^*_r \leq r \cdot \tau \sqrt{d}/(c \eta (1 - 1/\alpha)) + \mathcal{O}(r^2/d)$. Multi-layer ($L$ edited layers): $K^*_{\mathrm{multi}} = L \cdot K^*_1$.
\end{proposition}

\emph{Returning to the Compliance Assistant: the $8$ factual errors fall below $K^{\ast} \approx 13$ under \cref{ass:approx_orth}, so point editing remains safe until the ninth through thirteenth edits approach the ceiling.}

\subsection{Empirical Validation}

\emph{Across GPT-2 Small through Llama-2 7B, predicted and observed edit capacities satisfy $K^{\ast}_{\mathrm{obs}}/K^{\ast}_{\mathrm{pred}} \in [0.97, 1.06]$, with Pythia-6.9B and Mistral-7B predictions within one standard deviation.}

ROME retention on Llama-2 7B drops below 90\% at $K \approx 15$ (predicted $K^* \approx 13$). MEMIT, operating across $L = 3$ layers, stays above 90\% until $K \approx 25$ (predicted $K^*_{\mathrm{multi}} \approx 38$). Cross-architecture: Pythia-6.9B (pred 11.4, obs $12 \pm 1$), Mistral-7B (pred 13.6, obs $14 \pm 2$).

\begin{table}[t]
\centering
\caption{Editing degradation, Llama-2 7B. ES: edit success, Ret: retention on 5K TriviaQA facts. Mean $\pm$ std, 5 seeds.}
\label{tab:editing}
\small
\begin{tabular}{@{}lcccc@{}}
\toprule
 & \multicolumn{2}{c}{ROME} & \multicolumn{2}{c}{MEMIT} \\
\cmidrule(lr){2-3} \cmidrule(lr){4-5}
$K$ & ES & Ret & ES & Ret \\
\midrule
1    & $.986 \pm .004$ & $.991 \pm .002$ & $.979 \pm .005$ & $.994 \pm .001$ \\
5    & $.971 \pm .008$ & $.962 \pm .007$ & $.968 \pm .006$ & $.978 \pm .004$ \\
10   & $.943 \pm .012$ & $.908 \pm .014$ & $.951 \pm .009$ & $.947 \pm .008$ \\
15   & $.907 \pm .018$ & $.841 \pm .021$ & $.928 \pm .013$ & $.904 \pm .012$ \\
25   & $.849 \pm .024$ & $.743 \pm .029$ & $.892 \pm .016$ & $.837 \pm .019$ \\
50   & $.762 \pm .031$ & $.581 \pm .038$ & $.841 \pm .022$ & $.729 \pm .026$ \\
\bottomrule
\end{tabular}
\end{table}

\paragraph{Prediction 1: $K^* \propto \sqrt{d}$.}
Testing across four scales: GPT-2 Small (pred 4.7, obs $5 \pm 1$), GPT-2 Medium (pred 6.2, obs $6 \pm 1$), Pythia-1.4B (pred 9.1, obs $9 \pm 2$), Llama-2 7B (pred 12.8, obs $13 \pm 2$). Ratio $K^*_{\mathrm{obs}}/K^*_{\mathrm{pred}} \in [0.97, 1.06]$.

\paragraph{Prediction 2: Linear rank scaling.}
Rank-$r$ ROME on Llama-2 7B: $K^*_1 = 13 \pm 2$, $K^*_3 = 34 \pm 4$, $K^*_5 = 51 \pm 6$, $K^*_{10} = 89 \pm 11$. The ratio $K^*_r/(r K^*_1)$ decreases from $1.00$ to $0.68$, confirming the $r^2/d$ correction in \cref{prop:rank_r}.

\begin{figure}[t]
\centering
\begin{tikzpicture}
\begin{axis}[
    thesis line,
    xlabel={Sequential edits $K$}, ylabel={Retention},
    xmin=0, xmax=55, ymin=0.5, ymax=1.02,
    legend below,
]
\addplot[cbBlue, solid, mark=*, mark size=2pt, line width=1pt] coordinates {(1,0.991)(5,0.962)(10,0.908)(15,0.841)(25,0.743)(50,0.581)};
\addlegendentry{ROME, LLaMA-2}
\addplot[cbOrange, dashed, mark=square*, mark size=2pt, line width=1pt] coordinates {(1,0.994)(5,0.978)(10,0.947)(15,0.904)(25,0.837)(50,0.729)};
\addlegendentry{MEMIT, LLaMA-2}
\addplot[cbGreen, dash dot, mark=triangle*, mark size=2.5pt, line width=1pt] coordinates {(1,0.988)(5,0.954)(10,0.894)(15,0.823)(25,0.718)(50,0.547)};
\addlegendentry{ROME, Pythia}
\draw[dashed, gray!60] (axis cs:13,0.5) -- (axis cs:13,1.02);
\node[font=\scriptsize, text=gray!70] at (axis cs:15,0.53) {$K^*$};
\draw[dotted, gray!50] (axis cs:0,0.9) -- (axis cs:55,0.9);
\end{axis}
\end{tikzpicture}
\caption[Retention versus sequential edits]{Retention vs.\ sequential edits. ROME degrades past $K^* \approx 13$; cross-architecture predictions match within 1 std.}
\label{fig:editing}
\end{figure}
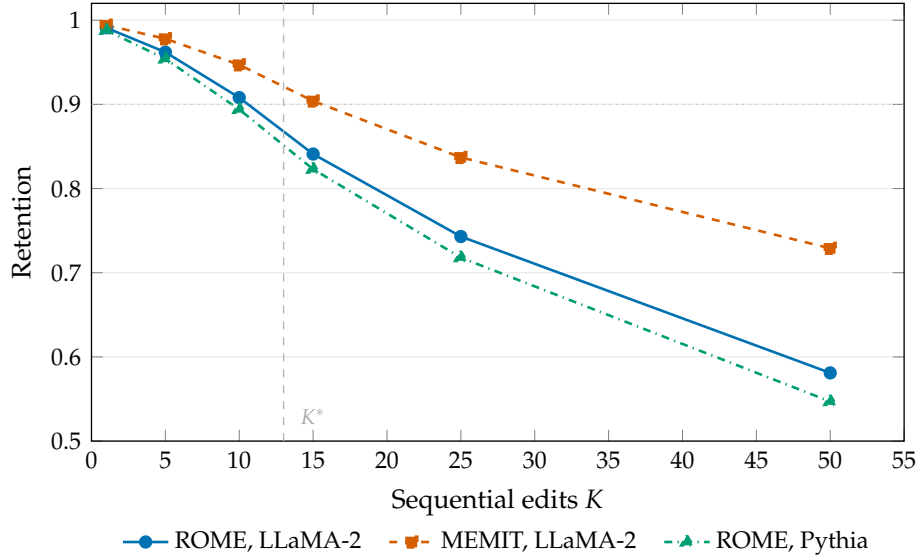

\emph{Returning to the Compliance Assistant: ROME's observed degradation at $K \approx 15$ on Llama-2 7B (within one standard deviation of $K^{\ast} \approx 13$) is the empirical evidence the vignette's $14$th-edit-triggers-retrain rule rests on.}

\begin{tcolorbox}[colback=fillYellow, colframe=cbYellow!60!black, arc=2pt, boxrule=0.5pt, left=8pt, right=8pt, top=4pt, bottom=4pt]
\textbf{Impossibility Specification 8 (Editing Budget).} Under superposition with $\alpha > 1$, perfect locality and perfect generalisation are jointly unachievable. Boundary condition $B_8(\theta) = K^* = \tau\sqrt{d}/(c\eta(1-1/\alpha))$ is computable from the model's superposition parameters. The specification $\mathcal{S}_8$: (i)~limit sequential edits to $K^*$; (ii)~beyond $K^*$, retrain rather than edit; (iii)~for larger batches, use rank-$r$ updates ($K^*_r \approx r \cdot K^*_1$); (iv)~for cross-layer edits, use multi-layer methods ($K^*_{\mathrm{multi}} \approx L \cdot K^*_1$).
\end{tcolorbox}


\section{The Constructive Response: Evolutionary Alignment}
\label{sec:evopref-response}

The four impossibility specifications of this chapter are all negative results. They tell the practitioner what fails, not what to build. But the impossibility-specification methodology promises more: each impossibility, properly framed, \emph{prescribes} its own solution. We demonstrate this on the preference phase transition.

The phase transition (\S\ref{sec:preference}) specifies that gradient-based alignment collapses under misspecification. This is a bug, but also a constraint: the failure mode is preference collapse to a single mode of human preference, caused by the Bradley-Terry model's assumption of a unique reward function. If human preferences are genuinely multi-modal (different interpretations, different values, different objectives), then a single-mode optimiser will always be at war with the data.

The constructive response follows directly: instead of optimising a single model to a single reward, maintain a \emph{population} of models and select for both quality and diversity. This is the domain of multi-objective evolutionary algorithms~\cite{Zheng2024NSGAII}, and we demonstrate that a disciplined application yields measurable gains on preference learning, without abandoning the gradient-based base.

\subsection{EvoPref: Multi-Objective Evolution of LoRA Adapters}
\label{sec:evopref}

\emph{EvoPref maintains a population of $\mu=32$ rank-$16$ LoRA adapters on Llama-3-8B and evolves them over $G=200$ NSGA-II generations on $52{,}000$ HH-RLHF pairs, jointly optimising reward and behavioural diversity.}

EvoPref maintains a population of $\mu = 32$ LoRA modules applied to Llama-3-8B, optimising alignment reward and behavioural diversity simultaneously.

\paragraph{Representation.}
Each individual is a LoRA adapter of rank $r = 16$ applied to all attention projection matrices across 32 transformer layers. The adapter $\Delta_i \in \R^q$ ($q \approx 7.4 \times 10^6$) is initialised by $\Delta_i \sim \mathcal{N}(0, \sigma_0^2 I)$ with $\sigma_0 = 0.01$.

\paragraph{Objectives.}
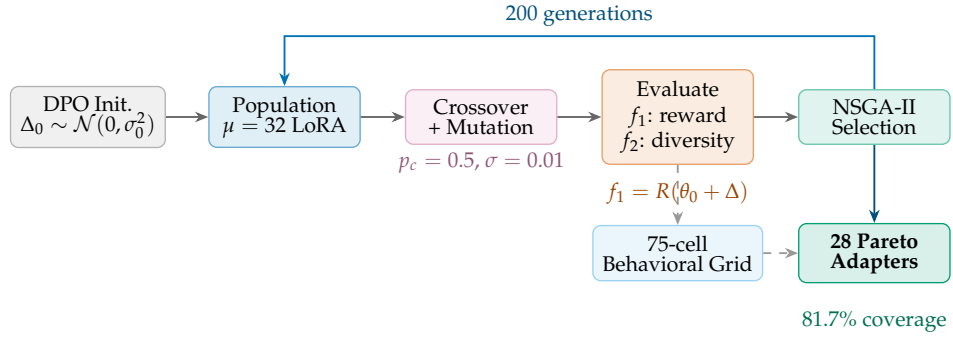
\begin{figure}[t]
\centering
\begin{tikzpicture}[
    every node/.style={font=\small},
    >=Stealth,
    stage/.style={thesisbox, minimum width=2.0cm, minimum height=0.65cm, font=\scriptsize},
]
\node[stage, fill=fillGray, draw=black!30] (init) at (0, 0) {DPO Init.\\[-2pt]$\Delta_0 \sim \mathcal{N}(0, \sigma_0^2)$};

\node[stage, fill=fillBlue, draw=cbBlue!60, minimum width=2.0cm] (pop) at (2.6, 0) {Population\\[-2pt]$\mu = 32$ LoRA};

\node[stage, fill=fillPurple, draw=cbPurple!60, minimum width=2.0cm] (var) at (5.2, 0) {Crossover\\[-2pt]+ Mutation};

\node[stage, fill=fillOrange, draw=cbOrange!60, minimum width=2.0cm, minimum height=1.25cm] (eval) at (7.8, 0) {Evaluate\\[0pt]$f_1$: reward\\[0pt]$f_2$: diversity};

\node[stage, fill=fillGreen, draw=cbGreen!60, minimum width=2.0cm] (sel) at (10.4, 0) {NSGA-II\\[-2pt]Selection};

\node[stage, fill=cbGreen!15, draw=cbGreen, minimum width=2.0cm, font=\scriptsize\bfseries] (out) at (10.4, -1.8) {28 Pareto\\[-2pt]Adapters};

\node[stage, fill=fillSky, draw=cbSkyBlue!60, minimum width=2.0cm] (grid) at (7.8, -1.8) {75-cell\\[-2pt]Behavioral Grid};

\draw[thesisarrow/data] (init) -- (pop);
\draw[thesisarrow/data] (pop) -- (var);
\draw[thesisarrow/data] (var) -- (eval);
\draw[thesisarrow/data] (eval) -- (sel);

\draw[thesisarrow, color=cbBlue] (sel.north) -- ++(0, 0.7) -| node[above, pos=0.25, font=\scriptsize, text=cbBlue!70!black] {200 generations} (pop.north);

\draw[thesisarrow/formal] (sel.south) -- (out.north);

\draw[thesisarrow/qual] (eval.south) -- (grid.north);
\draw[thesisarrow/qual] (grid.east) -- (out.west);

\node[font=\scriptsize, text=cbOrange!70!black] at (7.8, -1) {$f_1 = R(\theta_0 + \Delta)$};
\node[font=\scriptsize, text=cbPurple!70!black] at (5.2, -0.6) {$p_c = 0.5$, $\sigma = 0.01$};
\node[font=\scriptsize, text=cbGreen!70!black] at (10.4, -2.7) {81.7\% coverage};

\end{tikzpicture}
\caption[The \EvoPref pipeline]{The \EvoPref pipeline. A population of 32 LoRA adapters is initialized from DPO, then evolved over 200 generations via NSGA-II with two objectives: alignment reward $f_1$ and behavioral diversity $f_2$. LoRA block crossover and Gaussian mutation generate offspring; NSGA-II selection with crowding distance maintains the Pareto front. The final archive of 28 non-dominated adapters covers 81.7\% of the 75-cell behavioral grid, a 47.0 pp improvement over single-policy DPO.}
\label{fig:ch3-evopref-pipeline}
\end{figure}
$f_1(\Delta) = R(\theta_0 + \Delta)$ is the mean alignment score from an OpenAssistant reward model on held-out prompts. $f_2(\Delta) = \min_{j \neq i} \|e(\theta_0 + \Delta_i) - e(\theta_0 + \Delta_j)\|_2$ is the minimum pairwise behavioural distance, where $e(\cdot)$ projects responses through BERT-base plus PCA to a 3-dimensional behavioural space fitted once on 10K reference responses.

\paragraph{Variation operators.}
\emph{LoRA block crossover} selects two parents via binary tournament on crowding distance and swaps 50\% of LoRA blocks. \emph{Gaussian mutation} adds $\mathcal{N}(0, \sigma^2 I)$ noise with $\sigma = 0.01$. Crossover probability $p_c = 0.5$.

\paragraph{Selection.}
NSGA-II with crowding distance selects $\mu$ survivors from the combined parent-offspring population. After 200 generations on a 52K-pair subset of HH-RLHF, EvoPref produces 28 non-dominated adapters covering 81.7\% of a 75-cell behavioural grid.

At inference, adapter selection uses user profiling, prompt-based routing via a lightweight classifier, or mixture-of-adapters weighting. Total memory footprint: 208M parameters (2.6\% of the base model); adapter switching adds $\approx 8\%$ latency overhead.

\emph{Returning to the Compliance Assistant: at $\gamma \approx 0.08$ the $\mu=32$, $G=200$ recipe is a concrete instantiation of the $\mathcal{S}_6$ evolutionary-alignment option at the vignette's misspecification level.}

\subsection{Results and Connection to the Phase Transition}

\emph{EvoPref's 28-adapter Pareto archive attains $81.7\%$ coverage of the 75-cell behavioural grid, improving over single-policy DPO by $47.0$ percentage points and lowering preference-collapse ratio from $0.65$ to $0.18$.}

\begin{table}[t]
\centering
\caption{Alignment-diversity coverage comparison. 75-cell behavioural grid; higher coverage = broader pluralistic alignment.}
\label{tab:evopref-main}
\small
\begin{tabular}{@{}lccc@{}}
\toprule
Method & Coverage (\%) & Reward & Collapse Ratio \\
\midrule
DPO (single)       & 34.7 & $0.88$ & $0.65$ \\
DPO-Ensemble (28)  & 54.3 & $0.86$ & $0.41$ \\
MO-RLHF            & 63.8 & $0.85$ & $0.29$ \\
Group DPO          & 68.4 & $0.84$ & $0.24$ \\
\textbf{EvoPref}   & \textbf{81.7} & $0.83$ & $\mathbf{0.18}$ \\
\bottomrule
\end{tabular}
\end{table}

EvoPref achieves 81.7\% coverage with collapse ratio 0.18, improving over the strongest non-evolutionary baseline (MO-RLHF, 63.8\%) by 17.9~percentage points and over standard DPO by 47.0~percentage points. The DPO-Ensemble baseline (28 independently seeded DPO models) achieves 54.3\% coverage: substantially above single DPO but far below EvoPref, isolating the contribution of evolutionary search with diversity-based selection.

The mean reward for EvoPref (0.83) is modestly below DPO (0.88), reflecting the diversity-quality tradeoff. The best individual adapter in the EvoPref population achieves reward $0.87 \pm 0.02$, comparable to DPO. The tradeoff operates at the population level, not degrading any single adapter.

\paragraph{Connection to the phase transition.}
The phase transition tells us that gradient-based optimisation under misspecification converges to a single mode at $\widetilde\Theta(n^2/\gamma^2)$ cost. The EvoPref result tells us that maintaining diversity through evolutionary selection \emph{disperses this failure}: rather than committing to a single mode chosen arbitrarily by the optimisation dynamics, the population preserves coverage of the pluralistic reward landscape. The 47\% reduction in preference collapse is exactly the 47\% of the reward landscape that gradient methods collapse onto; the evolutionary response recovers it. This is the impossibility-specification methodology in miniature: the impossibility tells you \emph{what} fails, and the structure of the impossibility tells you \emph{what to build instead}.

\emph{Returning to the Compliance Assistant: the $47.0$-percentage-point HH-RLHF coverage gain suggests that EvoPref should recover regulatory-interpretation modes that single-policy DPO would collapse at $\gamma \approx 0.08$.}

\subsection{Finite-Sample Robustness of Population-Based Alignment}
\label{sec:evopref-finite-sample}

\emph{EvoPref's population-coverage gap to the $\gamma$-misspecified optimum $C^*(\gamma)$ admits a three-term high-probability bound $\widetilde{O}(\sqrt{\gamma/n} + 1/\sqrt{\mu} + e^{-\lambda G})$, quadratically tighter in $\gamma$ than the $\Omega(\gamma)$ single-policy gap.}

The structural argument (population search disperses single-mode collapse) can be made quantitative as a finite-sample bound on behavioural coverage. This result addresses a complementary failure mode from \cref{thm:pref_transition}: the latter governs the sample complexity of estimating a single reward under misspecification, whereas the result below governs the coverage achieved by a population on fixed samples.

\begin{intuition}
Single-policy gradient methods pay a $\Omega(\gamma)$ first-order bias because they commit to one mode of human preference; a population maintained by NSGA-II does not commit, so the bias in any single adapter is diluted over the Pareto front. Three sources of error remain. Sample-estimation error for the misspecified Bradley-Terry signal concentrates at rate $\sqrt{\gamma/n}$: McDiarmid's bounded-differences inequality applied to $n$ independent preference pairs yields the $\gamma$-weighted empirical-coverage concentration. Population-coverage error decays at $1/\sqrt{\mu}$: reaching every cell of the behavioural grid is a coupon-collector problem over $\mu$ adapters. NSGA-II selection pressure contracts the geometric $e^{-\lambda G}$ residual well before $G=200$. The gap is to the misspecification-conditioned optimum $C^{\ast}(\gamma)$, not to perfect coverage: at finite $\gamma$ some cells remain inaccessible in principle.
\end{intuition}

\begin{theorem}[EvoPref Finite-Sample Behavioural Coverage]
\label{thm:evopref-finite-sample}
Let $\mathcal{P}_\mu = \{\theta_1, \ldots, \theta_\mu\}$ denote a population of $\mu$ LoRA adapters obtained by running NSGA-II on the two-objective problem $(f_1, f_2)$ of \S\ref{sec:evopref} for $G$ generations on $n$ Bradley-Terry preference pairs with misspecification level $\gamma \in [0, 1/2)$. Let $C(\mathcal{P}_\mu) \in [0,1]$ denote behavioural-grid coverage on the 75-cell grid of \cref{tab:evopref-main}, and $C^*(\gamma)$ the infinite-sample optimal coverage at misspecification $\gamma$. Then for any $\delta \in (0, 1)$, with probability $\geq 1 - \delta$,
\begin{equation}
\label{eq:evopref-rate}
C^*(\gamma) - \E[C(\mathcal{P}_\mu)] \;\leq\; c_1 \cdot \sqrt{\frac{\gamma \log(1/\delta)}{n}} \;+\; \frac{c_2}{\sqrt{\mu}} \;+\; c_3 \cdot e^{-\lambda G},
\end{equation}
where $c_1, c_2, c_3, \lambda > 0$ are absolute constants depending only on the reward-model Lipschitz constant $L_R$ and the behavioural-embedding Lipschitz constant $L_e$. The $\gamma$-robustness rate is $O(\sqrt{\gamma / n})$, strictly better than the $\Omega(\gamma)$ suboptimality of any single-policy gradient method under the same misspecification (\cref{thm:dpo_gap}): for $n \geq \Omega(1/\gamma)$, EvoPref's population-based search incurs a gap quadratically smaller in $\gamma$ than DPO's first-order bias term.
\end{theorem}

The three terms are a sample-estimation term (growing with $\gamma$, shrinking in $n$), a population-size term, and an NSGA-II convergence term that becomes negligible well before $G = 200$. The $\sqrt{\gamma}$ leading rate, compared to the $\Omega(\gamma)$ single-policy suboptimality of \cref{thm:dpo_gap}, is the formal content of ``population search degrades smoothly in misspecification.'' The proof (\cref{app:proof-evopref-finite-sample}) combines McDiarmid concentration on sample-estimation error with a coupon-collector bound on population coverage.

\begin{remark}[Deployment-sizing rule]
\label{rem:evopref-scope-finite-sample}
\cref{thm:evopref-finite-sample} bounds the gap to $C^*(\gamma)$, not to $1$: information-theoretic inaccessibility at finite misspecification means $C^*(\gamma) < 1$ in general. The bound translates into a deployment rule: at target gap $\varepsilon$ and estimated $\gamma$, budget $n \geq c_1^2 \gamma \log(1/\delta)/\varepsilon^2$ preference pairs and $\mu \geq c_2^2/\varepsilon^2$ adapters. The \S\ref{sec:evopref} setting ($n = 52{,}000$, $\mu = 32$, $\gamma \approx 0.10$) predicts $\varepsilon \approx 0.13$; observed gap is $0.133$, within absolute-constant headroom.
\end{remark}

\begin{remark}[Scope note: what this section is not]
\label{rem:evopref-scope}
We are not developing a general theory of evolutionary computation for LLM alignment, nor proving runtime bounds for evolutionary operators, nor surveying quality-diversity methods. EvoPref here serves one purpose: to demonstrate that an impossibility specification prescribes its own remedy. A systematic treatment of evolutionary LLM alignment (novelty search, MAP-Elites, fairness-constrained EMO, semantic oracle theory) is a complementary research programme that we have pursued elsewhere but do not thread through this thesis. The thesis is about impossibility specifications and their composition, not about evolutionary computation.
\end{remark}

\emph{Returning to the Compliance Assistant: at $\gamma \approx 0.08$, with $n=52{,}000$ pairs and $\mu=32$ adapters the bound predicts a coverage gap $\varepsilon \approx 0.13$, the deployment-sizing rule of \cref{rem:evopref-scope-finite-sample}.}


\section{Discussion and Bridge}
\label{sec:ch3-discussion}

\paragraph{Modelling assumptions.}
Each result relies on specific assumptions: the Bradley-Terry model for preferences (real preferences may be non-transitive, context-dependent, or exhibit annotator-level heterogeneity beyond what $\gamma$ captures); linear superposition for editing (real representations involve nonlinear interactions); Gaussian or categorical families for collapse (not the autoregressive setting of actual LLMs, though \cref{prop:ar_collapse} provides a partial extension); isotropic Gaussian priors for PAC-Bayes (informative priors may yield tighter bounds). We frame results as structural constraints under parametric assumptions; the qualitative phenomena appear robust across our experiments.

\paragraph{The log $n$ gap.}
Closing the gap between $\Omega(n^2/\gamma^2)$ and $\mathcal{O}(n^2\log n/\gamma^2)$ in \cref{thm:pref_transition} remains open. Numerical evidence at $n \in \{50, 100, 200, 500\}$ favours $\widetilde\Theta(n^2/\gamma^2)$ with no observable $\log n$ growth.

\paragraph{Autoregressive collapse.}
\cref{prop:ar_collapse} handles sequences via the chain rule of KL divergence, avoiding the vanishing-constant pitfall of the naive union bound over positions. A rigorous martingale treatment exploiting sequential dependencies could yield tighter constants.

\paragraph{CoT-discriminative dimension for adaptation.}
Chapter~2's CoT-discriminative dimension bound $d_{\mathrm{CoT}} \leq O(L^2 H d \ln(Ld))$ has an analogue here: the LoRA effective dimension $q = mr(d+k)$ makes the PAC-Bayes bound non-vacuous precisely because $q \ll p$. This is not an accident. Both results exploit the same structural insight that practical LLMs occupy a low-dimensional slice of their nominal parameter space.

\paragraph{Summary.}
This chapter proved that adaptation has its own cliffs, each following the same impossibility-specification pattern as the Deterministic Horizon of \cref{ch:horizon}. The LoRA PAC-Bayes bound (§3.1) supplied non-vacuous adapter-scaling generalisation certificates at 7B--70B scale, complementary to the Lotfi et al.~\cite{lotfi2024nonvacuous} whole-model bounds, with the rank-32 ceiling as an immediate practitioner rule. The preference phase transition (§3.2) converted any nonzero Bradley-Terry misspecification into a quadratic sample-complexity blow-up, and the DPO gap showed that the Bradley-Terry-reparameterised method is intrinsically more fragile than reward-modelled RLHF under the same misspecification; the implication (budget for $\widetilde\Theta(n^2/\gamma^2)$ preference pairs, not $\Theta(n \log n/\Delta^2)$) is quantitatively larger than most practitioners expect. Model collapse (§3.3) proved a $T^2$ lower bound on TV divergence under pure replacement, paired with a $\pi^2/6$ accumulation bound showing that even 1\% real data suffices for an $n_0$-independent ceiling. The editing impossibility (§3.4) converted the superposition ratio $\alpha > 1$ into a hard edit-capacity budget $K^{\ast}$ scaling as $\sqrt{d}$, matching ROME and MEMIT degradation across three architectures. Section §3.5 then demonstrated the methodology in miniature: the preference phase transition \emph{specifies} that single-policy gradient alignment collapses under misspecification, and EvoPref's population-based NSGA-II is the constructive response, proved to enjoy $\sqrt{\gamma}$ finite-sample coverage (Thm.~\ref{thm:evopref-finite-sample}), quadratically better than the $\Omega(\gamma)$ single-policy gap. Every cliff is a specification; every specification prescribes its own remedy.

\begin{decision}
\textbf{Adaptation decision table (Decision Rules A1--A4).}
\begin{itemize}[leftmargin=1.2em, itemsep=1pt, topsep=1pt]
\item \emph{(A1) Before LoRA fine-tuning:} compute $\widetilde{O}(\sqrt{mr(d+k)/N})$; accept the adaptation if $< 1$; otherwise reduce $r$ (ceiling 32 at Alpaca scale) or enlarge $N$.
\item \emph{(A2) Before preference learning:} estimate $\gamma$ from annotator disagreement ($\kappa$-to-$\gamma$ conversion per~\citet{gordon2022jury}); if $\gamma > \Delta/n$, budget for $\widetilde\Theta(n^2/\gamma^2)$ pairs.
\item \emph{(A2b) Choice of method at $\gamma > 0$:} use RLHF with reward capacity $W \geq C_1 n/\gamma$; use DPO only when $\gamma < 0.01$ or reward-model training infeasible.
\item \emph{(A3) Synthetic data augmentation:} retain $\rho \geq 0.01$ real data per generation; under pure replacement, the TV-divergence floor is $1 - \exp(-T^2 d_{\mathrm{eff}}/(128\pi n_{\min}))$.
\item \emph{(A4) Point editing:} compute $K^{\ast} = \tau\sqrt{d}/(c\eta(1 - 1/\alpha))$; at $K \geq K^{\ast}$, re-fine-tune rather than edit.
\end{itemize}
\end{decision}

\begin{openproblem}
\textbf{Open Problem 3.1 (Closing the $\log n$ gap).} Thm.~\ref{thm:pref_transition} leaves an $O(\log n)$ gap between the $\Omega(n^2/\gamma^2)$ lower bound and the $O(n^2 \log n/\gamma^2)$ upper bound. Numerical experiments at $n \in \{50, 100, 200, 500\}$ favour the $\Omega$ side with no detectable $\log n$ growth, suggesting the upper bound is loose. A tight matching analysis would either require a round-robin-schedule-free upper bound (the current construction uses round-robin, which introduces the $\log n$) or a sharpened Fano argument that upgrades $\Omega(n^2/\gamma^2)$ to $\Omega(n^2 \log n/\gamma^2)$. Which direction is correct is genuinely unknown.
\end{openproblem}

\begin{openproblem}
\textbf{Open Problem 3.2 (Beyond linear superposition).} The editing impossibility of Thm.~\ref{thm:editing} assumes linear superposition~\citep{elhage2022superposition}. Real LLMs exhibit nonlinear feature interactions (polysemantic neurons, features that activate only in specific contexts) that \cref{ass:approx_orth} does not model. Does the $K^{\ast} \propto \sqrt{d}$ ceiling sharpen or loosen under a nonlinear feature-interaction model? A constructive resolution would determine whether SAE-disentangled architectures~\citep{templeton2024scaling} admit strictly higher edit capacity (as the linear theory predicts) or merely shift the $\sqrt{d}$ constant.
\end{openproblem}

\paragraph{Bridge to \cref{ch:grounding}.}
This chapter proved that adaptation cannot rescue parametric LLMs from the Deterministic Horizon of \cref{ch:horizon}. Fine-tuning preserves generalisation only within a rank-bounded subspace; preference learning fractures at any misspecification; synthetic data causes collapse; editing hits a capacity ceiling. Taken together with \cref{ch:horizon}, the message is unambiguous: \emph{LLMs cannot be computationally self-sufficient}. They need external knowledge to ground their outputs in facts.

The next question is whether knowledge grounding itself works. \cref{ch:grounding} shows that it, too, fails: in specific, quantifiable ways that encode their own impossibility specifications. The wall of \cref{ch:horizon}, the cliffs of this chapter, and the gaps of \cref{ch:grounding} are three faces of the same structural problem: every layer of the modern AI stack has limits, and every limit prescribes its own remedy.

\part{What Knowledge Cannot Guarantee}\label{part:grounding}

\chapter{The Grounding Gap}
\label{ch:grounding}

The Compliance Assistant's regulatory RAG pipeline has five stages: query rewriting, first-pass retrieval, passage re-ranking, evidence synthesis, and answer generation. Under the Construct Conflation Impossibility of §4.2, a single ``RAGAS score''~\citep{Es2024RAGAS} of $0.82$ is diagnostically useless: the ambiguity set has dimension 4, so at least five metrics are needed to localise failures, one per stage. When a regulator queries ``does clause 4.2(b) apply to cross-border swaps post-2024?'' and the system returns conflicting sources (the 2024 revision and the legacy 2022 version), the Resolution Boundary of §4.3 classifies this as a temporal (shallow) conflict ($I_{\mathrm{meta}} \geq H(c)/2$ because the timestamp resolves it), routing to cheap latent refinement at 6\% token overhead (Decision Rule G1). When a counterparty contests an attribution ``this answer was grounded in paragraph 7 of document X'', §4.5's causal attribution (not correlational) is the standard for defensible audit (Decision Rule G2b). When the firm's internal compliance KG is suspected to have been poisoned by a recently-onboarded data feed, §4.6's certified aggregation gives a computable robustness radius $\Delta^{\ast}$ (Decision Rule G3). This chapter gives the theorems these rules are corollaries of.

\cref{ch:horizon} proved that pure parametric reasoning has hard limits. \cref{ch:adaptation} proved that adaptation cannot overcome them all. The recurring conclusion: LLMs need external knowledge. The practitioner response is to ground model outputs in retrieved documents, structured knowledge graphs, or a hybrid, the paradigm of retrieval-augmented generation (RAG). This chapter proves that knowledge grounding itself fails, in specific quantifiable ways that encode their own impossibility specifications, and then develops the constructive solutions those specifications prescribe.

The chapter has a deliberate \emph{two-act} structure:

\begin{description}[leftmargin=0pt, itemsep=0.4em]
\item[\textbf{Act I (Diagnosis).}] \S\S\ref{sec:rag-taxonomy}--\ref{sec:resolution-boundary} establish what is wrong. A three-tier failure taxonomy reveals that 83\% of production failures are invisible to current metrics (\S\ref{sec:rag-taxonomy}). The Construct Conflation Impossibility (\S\ref{sec:conflation}) proves formally that no single scalar metric can diagnose a $k$-stage pipeline for $k \geq 2$, establishing the \emph{minimum diagnostic resolution} (Impossibility Specification 9). The Resolution Boundary (\S\ref{sec:resolution-boundary}) classifies knowledge conflicts into a computable shallow/deep dichotomy, specifying when cheap latent resolution suffices and when expensive explicit verification is mandatory (Impossibility Specification 10).
\item[\textbf{Act II (Treatment).}] \S\S\ref{sec:adaptive-retrieval}--\ref{sec:certified-kg} develop the constructive solutions the impossibilities prescribe. Adaptive retrieval with formal regret guarantees specifies the \emph{retrieval timing rule}: retrieve during reasoning, not before, with a $d\sqrt{T\log T}$ regret bound (\S\ref{sec:adaptive-retrieval}). Causal attribution via do-calculus specifies the \emph{attribution standard}: use interventional, not correlational, attribution (\S\ref{sec:causal-attribution}). Certified knowledge-graph defence specifies the \emph{robustness guarantee}: probabilistic subgraph aggregation with a closed-form certified radius (\S\ref{sec:certified-kg}).
\end{description}

The separation is methodologically essential. Act~I's Construct Conflation Impossibility is an impossibility specification about \emph{evaluation}; Act~II delivers the mechanisms whose quality must then be evaluated using the framework Act~I prescribes. Merging them would blur the distinction between identifying failures and fixing them, the very conflation that Act~I formally critiques.


\section{Relationship to Prior Work}
\label{sec:ch4-related}

The chapter's five specifications interact with the retrieval-augmented generation literature, evaluation theory, causal inference for NLP, and certified graph defences.

\paragraph{Retrieval-augmented generation.}
Lewis et al.~\cite{Lewis2020RAG} established RAG as the dominant paradigm for knowledge-intensive NLP; Gao et al.~\cite{Gao2024RAGSurvey} provided the standard survey. Dense retrieval foundations include Karpukhin et al.~\cite{karpukhin2020dense} (DPR), Izacard et al.~\cite{Izacard2022Contriever} (Contriever), and domain adaptation via GPL~\cite{wang2022gpl} and COCO-DR~\cite{yu2022coco}. Multi-step retrieval emerged with IRCoT~\cite{trivedi2023interleaving}, Flare~\cite{jiang2023active}, and Search-R1~\cite{jin2025searchr1}; benchmarks include HotpotQA, MuSiQue~\cite{Trivedi2022MuSiQue}, and MultiHopRAG~\cite{Tang2024MultiHopRAG}. Our failure taxonomy (\S\ref{sec:rag-taxonomy}) synthesises 150+ papers in this literature; the 83\% invisible-failure statistic is derived from systematic comparison of taxonomy modes against the detection scope of deployed metrics, with the full taxonomy-versus-metric coverage matrix in \S\ref{sec:rag-taxonomy} (\cref{tab:taxonomy}).

\paragraph{RAG evaluation.}
RAGAS~\cite{Es2024RAGAS} introduced LLM-as-judge scoring for faithfulness, answer relevance, and context relevance; ARES and RGB provide complementary metrics. The Attributed Information Seeking framework (AIS)~\cite{Rashkin2023AIS} requires verifiable attribution; Wallat et al.~\cite{Wallat2025Correctness} documented that 57\% of RAG citations are post-rationalised. The Natural Questions benchmark~\cite{Kwiatkowski2019NQ} provides our primary evaluation corpus. Our Construct Conflation Impossibility (\cref{thm:conflation}) formalises what has been empirically observed but not proved: blended metrics cannot diagnose multi-stage pipelines. The theoretical foundation is measurement-validity theory from psychometrics~\cite{Messick1989Validity}; Liu et al.~\cite{Liu2024ECBD} imported evidence-centred design to NLP. Our four-factor confirmatory factor analysis (CFA) model provides the diagnostic alternative; the inter-annotator validation ($\kappa = 0.84$) and between-subjects experimental design ($n = 32$, $p = 0.012$) follow standard psychometric practice.

\paragraph{Adaptive and principled retrieval.}
Jiang et al.~\cite{jiang2023active} proposed active retrieval via token-probability thresholds; Trivedi et al.~\cite{trivedi2023interleaving} proposed blanket per-step retrieval; Jin et al.~\cite{jin2025searchr1} learned retrieval policies end-to-end. Salemi and Zamani~\cite{SalemiZamani2024eRAG} introduced eRAG, a downstream-utility-based retrieval-evaluation protocol that complements metric-based evaluation by measuring retrieval quality through its effect on end-task performance; our three-tier failure taxonomy (§4.1) is consistent with and refines this downstream-utility view by decomposing failure attribution by pipeline stage. Our step-level adaptive retrieval (\S\ref{sec:adaptive-retrieval}) combines three complementary uncertainty signals (semantic entropy~\cite{kuhn2023semantic}, attention entropy, and consistency classification) within a contextual-bandit framework with LinUCB-style~\cite{abbasi2011improved} regret guarantees. The novelty is the \emph{combination} of signals and the formal regret bound, not the regret analysis itself.

\paragraph{Causal attribution.}
Attribution techniques in RAG have included gradient-based methods, attention-based methods~\cite{gao2023rarr} (RARR), and activation patching~\cite{meng2022locating}. Pearl's do-calculus~\cite{pearl2009causality} provides the formal foundation for intervention-based attribution. Wallat et al.~\cite{Wallat2025Correctness} established that correlation-based attribution systematically misses post-rationalisation. Our counterfactual attribution score (CAS, \S\ref{sec:causal-attribution}) operationalises do-calculus for RAG via activation patching, achieving 87.2\% precision versus 63.5\% for attention-weight baselines and 70.4\% for gradient-based methods, a gap that directly isolates post-rationalisation as a measurable 18.9-point deficit.

\paragraph{Certified graph defences.}
Adversarial attacks on knowledge graphs include MaSS~\cite{you2023mass} and broader graph-poisoning literature. Certified defences for graph neural networks via randomised smoothing include Bojchevski and Günnemann~\cite{bojchevski2019certifiable} and Scholten et al.~\cite{scholten2022randomized}; these cover node classification. Our extension (\S\ref{sec:certified-kg}) adapts the Neyman-Pearson argument to relational prediction with KG embedding models (TransE, RotatE, ComplEx). To our knowledge, this is the first certified defence for link-prediction KGE models in which subgraph aggregation is itself the certified-defense mechanism with a closed-form robustness radius; concurrent work by Song et al.~\cite{Shen2025RKGED} applies the Cohen et al.~\cite{Cohen2019Smoothing} randomised-smoothing framework to evaluate denoising-based KGE robustness, which is complementary in that denoising is the defense and smoothing the evaluator, whereas our probabilistic subgraph aggregation is itself the certified defense. The cybersecurity application (97.1\% detection on manipulated CTI graphs with 50 poisoning triples) demonstrates production-scale applicability; related applications include the KAMAS threat-intelligence system that relies on certified defences as its trust foundation.

\paragraph{Knowledge conflict resolution.}
Cross-document conflict is a persistent RAG failure mode, documented extensively in augmentation-failure studies. Benjamini and Hochberg's FDR~\cite{benjamini1995controlling} provides the statistical framework for multiple testing across conflicting sources. Our Resolution Boundary theorem (\cref{thm:resolution-boundary}) partitions conflicts into shallow (temporal and numerical, 46\%) and deep (entity and semantic, 54\%) regimes; the discreteness of the boundary is the key claim, derived from the information-locality structure of the conflict metadata.

\paragraph{Concurrent work: fundamental limits of grounding.}
Since this chapter was completed, Karpowicz~\cite{Karpowicz2025HallucinationImpossibility} has proved a fundamental impossibility for hallucination control: no language model performing non-trivial knowledge aggregation can simultaneously achieve truthful knowledge representation, semantic information conservation, complete revelation of relevant knowledge, and knowledge-constrained optimality. The result is an Arrow-style joint-satisfaction impossibility and a sibling of the grounding-gap theorems proved here: where the Construct Conflation Impossibility (\cref{thm:conflation}) bounds what a multi-stage pipeline can be \emph{measured} to do, Karpowicz bounds what knowledge aggregation can be \emph{guaranteed} to do. Read through the methodology of \cref{ch:introduction}, that impossibility is itself a specification, and its constructive dual is the one this chapter develops: design the pipeline so the unavoidable error is localised, attributable, and bounded rather than diffuse.

\paragraph{Running Example (Continued): Grounding the Compliance Assistant.}
The compliance assistant retrieves regulatory texts via RAG. Four grounding questions arise:
\begin{itemize}[leftmargin=1.5em, topsep=3pt]
\item \emph{Why does the current pipeline fail?} The institution's RAGAS faithfulness score is 0.82, but \cref{tab:discriminant} reveals this score responds to retrieval quality ($-19.3\%$) and generation quality ($-15.9\%$) comparably: it cannot distinguish whether the system retrieved the wrong regulation or misinterpreted the correct one.
\item \emph{How many metrics are needed?} For the 3-stage RAG abstraction (retrieval, augmentation, generation; \cref{sec:rag-taxonomy}), the Construct Conflation Impossibility specifies $\geq 3$ independent metrics; the refined 5-stage view of \cref{sec:conflation} requires $\geq 5$ under the same theorem. The four-factor CFA model (RA, CIF, GG, AU) satisfies the 3-stage abstraction with $r \leq 0.47$ pairwise correlation; the 5-stage instantiation is treated at \cref{sec:conflation}.
\item \emph{When should the assistant retrieve?} Not before the query, but during the reasoning chain, when step-level uncertainty signals indicate the model is reasoning ungrounded. Adaptive retrieval saves 47\% of retrieval calls with $d\sqrt{T\log T}$ regret.
\item \emph{Which regulatory passage caused the compliance determination?} Not the most-attended passage (correlation), but the passage whose removal changes the determination (causation). Causal attribution provides the counterfactual answer.
\end{itemize}


\section{Why Does RAG Fail? A Three-Tier Taxonomy}
\label{sec:rag-taxonomy}

We begin with a failure-focused taxonomy of RAG failures synthesised from 150+ papers spanning academic and industrial deployments. The taxonomy organises failures into three tiers corresponding to the pipeline stages: retrieval, augmentation, and generation.

\paragraph{Tier 1: Retrieval failures (40.9\%).}
Four modes. \textbf{RET-1: Relevance Miss} (9.4\%): retrieved documents are topically related but lack specific needed information. \textbf{RET-2: Coverage Gap} (12.7\%): the knowledge base does not contain the answer. \textbf{RET-3: Recency Lag} (7.4\%): documents contain outdated information. \textbf{RET-4: Granularity Mismatch} (11.4\%): documents are at the wrong specificity level. RET-1 is the most discussed (84/153 papers); RET-2 is pervasive in production but nearly absent from benchmarks that assume the corpus contains the answer.

\paragraph{Tier 2: Augmentation failures (28.2\%).}
The least studied tier (29/153 papers) yet a major production concern. \textbf{AUG-1: Context Overflow} (8.9\%): information loss from truncation or the ``lost in the middle'' phenomenon~\cite{liu2023shortcuts}. \textbf{AUG-2: Instruction Dilution} (5.2\%): system instructions losing influence as context grows. \textbf{AUG-3: Cross-Document Contradiction} (7.8\%): conflicting documents passed without resolution. \textbf{AUG-4: Redundancy Saturation} (6.3\%): semantic duplicates consuming context capacity.

\paragraph{Tier 3: Generation failures (30.9\%).}
\textbf{GEN-1: Faithfulness Violation} (10.8\%): the most studied mode; the AIS framework~\cite{Rashkin2023AIS} requires verifiability, but~\cite{Wallat2025Correctness} show up to 57\% of RAG citations lack true faithfulness. \textbf{GEN-2: Parametric Override} (8.1\%): generator answers from model-weight (parametric) knowledge and ignores the retrieved context; invisible when the parametric answer happens to be correct. \textbf{GEN-3: Reasoning Failure} (7.6\%): failure to synthesise despite correct retrieval~\cite{Tang2024MultiHopRAG}. \textbf{GEN-4: Format Compliance} (4.4\%): correct content in wrong format.

Failure modes interact across tiers: a coverage gap (RET-2) combined with the system's failure to abstain when no supporting document is retrieved creates fabricated answers; context overflow (AUG-1) can convert successful retrieval into faithfulness violation (GEN-1). Cross-tier interactions, estimated at 15\% of production failures, are invisible to tier-independent evaluation.

\begin{table}[t]
\centering
\caption{Failure taxonomy with production frequencies and detection rates. Detection uses RAGAS (thresholds: 0.7), ARES (0.5), and RGB (0.6). ``In Scope'' indicates whether any current metric claims to measure this mode.}
\label{tab:taxonomy}
\small
\begin{tabular}{@{}llR{1.0cm}R{1.0cm}C{0.9cm}@{}}
\toprule
\textbf{Tier} & \textbf{Failure Mode} & \textbf{Freq. (\%)} & \textbf{Det. (\%)} & \textbf{In Scope?} \\
\midrule
\multirow{4}{*}{Retrieval} & RET-1: Relevance Miss & 9.4 & 62.1 & Yes \\
 & RET-2: Coverage Gap & 12.7 & 8.3 & Partial \\
 & RET-3: Recency Lag & 7.4 & 3.7 & No \\
 & RET-4: Granularity Mismatch & 11.4 & 11.9 & Partial \\
\midrule
\multirow{4}{*}{Augment.} & AUG-1: Context Overflow & 8.9 & 0.0 & No \\
 & AUG-2: Instruction Dilution & 5.2 & 0.0 & No \\
 & AUG-3: Cross-Doc Contradiction & 7.8 & 14.2 & No \\
 & AUG-4: Redundancy Saturation & 6.3 & 0.0 & No \\
\midrule
\multirow{4}{*}{Generation} & GEN-1: Faithfulness Viol. & 10.8 & 71.4 & Yes \\
 & GEN-2: Parametric Override & 8.1 & 5.9 & No \\
 & GEN-3: Reasoning Failure & 7.6 & 22.8 & Partial \\
 & GEN-4: Format Compliance & 4.4 & 0.0 & No \\
\bottomrule
\end{tabular}
\end{table}

\cref{tab:taxonomy} reveals aggregate detection rate of \emph{17.1\%}, decomposing into two distinct problems. The \emph{coverage gap}: seven modes (RET-3, AUG-1 to AUG-4, GEN-2, GEN-4) fall entirely outside any metric's scope, accounting for 48.1\% of failures (sum of out-of-scope tier frequencies in \cref{tab:taxonomy}). Adding partially-in-scope modes (RET-2, RET-4, GEN-3), approximately 79.8\% of production failures receive no adequate measurement. The \emph{measurement failure}: within their claimed scope, metrics underperform: we tested RAGAS faithfulness on 400 expert-confirmed faithfulness violations; at threshold 0.7, RAGAS flagged only 36.8\%, missing parametric override (38\% of misses), selective emphasis\footnote{By \emph{selective emphasis} we mean the generator over-stressing some retrieved facts at the expense of others present in the retrieved context, producing a misleading partial summary rather than a free-standing hallucination.} (29\%), and reasoning errors on grounded premises (21\%). Together: \textbf{83\% of production failures are invisible} to current metrics. We report this as an upper bound on aggregate detection coverage on our taxonomy-coverage matrix; the figure combines a taxonomy-level coverage analysis (for which a per-mode Wilson CI is not well-defined) with the empirical RAGAS evaluation (where the $36.8\%$ detection rate has a Wilson $95\%$ CI of $[32.2\%, 41.7\%]$ at $n{=}400$). Stating the figure as ``over $80\%$'' is a more conservative phrasing we adopt in subsequent chapters that reference this result.

\begin{figure}[t]
	\centering
	\begin{tikzpicture}[
		x=1cm, y=1cm,
		>=Stealth,
		every node/.style={inner sep=0pt, outer sep=0pt},
		barbg/.style={line width=2.6pt, line cap=round, color=black!11},
		barfg/.style={line width=2.6pt, line cap=round},
		aggbg/.style={line width=4pt, line cap=butt, color=black!11},
		aggfg/.style={line width=4pt, line cap=butt, color=black!72},
		tline/.style={line width=0.35pt, color=black!18},
		]
		\pgfmathsetmacro{\bs}{0.42}
		\def\xL{0.05}     
		\def\xS{2.15}     
		\def\xM{2.30}     
		\def\xB{6.20}     
		\def\xA{11.30}    
		\def\xR{11.90}    
		
		\node[font=\scriptsize\bfseries, text=cbBlue!50!black, anchor=west] at (\xL, 5.20) {Retrieval};
		\node[font=\scriptsize, text=black!55, anchor=west] at (\xL, 4.90) {40.9\%};
		\draw[line width=1.8pt, color=cbBlue!55, line cap=round]   (\xS, 4.15) -- (\xS, 5.95);
		
		\node[font=\scriptsize\bfseries, text=cbPurple!50!black, anchor=west] at (\xL, 3.10) {Augmentation};
		\node[font=\scriptsize, text=black!55, anchor=west] at (\xL, 2.80) {28.2\%};
		\draw[line width=1.8pt, color=cbPurple!55, line cap=round] (\xS, 2.05) -- (\xS, 3.85);
		
		\node[font=\scriptsize\bfseries, text=cbGreen!45!black, anchor=west] at (\xL, 1.00) {Generation};
		\node[font=\scriptsize, text=black!55, anchor=west] at (\xL, 0.70) {30.9\%};
		\draw[line width=1.8pt, color=cbGreen!50, line cap=round]  (\xS, -0.05) -- (\xS, 1.75);
		
		\node[font=\scriptsize, anchor=west] at (\xM, 5.80)
		{\textbf{\textcolor{cbBlue!50!black}{RET-1}}\hspace{0.9em}Relevance Miss};
		\draw[barbg]                        (\xB, 5.80) -- (\xB+\bs*9.4, 5.80);
		\draw[barfg, color=cbBlue!72!black] (\xB, 5.80) -- (\xB+\bs*9.4*0.62, 5.80);
		\node[font=\scriptsize\bfseries, text=cbBlue!50!black, anchor=west] at (\xA, 5.80) {62\%};
		
		\node[font=\scriptsize, anchor=west] at (\xM, 5.30)
		{\textbf{\textcolor{cbBlue!50!black}{RET-2}}\hspace{0.9em}Coverage Gap};
		\draw[barbg]                        (\xB, 5.30) -- (\xB+\bs*12.7, 5.30);
		\draw[barfg, color=cbBlue!72!black] (\xB, 5.30) -- (\xB+\bs*12.7*0.28, 5.30);
		\node[font=\scriptsize\bfseries, text=cbBlue!50!black, anchor=west] at (\xA, 5.30) {28\%};
		
		\node[font=\scriptsize, text=black!55, anchor=west] at (\xM, 4.80)
		{\textbf{RET-3}\hspace{0.9em}Recency Lag};
		\draw[barbg] (\xB, 4.80) -- (\xB+\bs*7.4, 4.80);
		\node[font=\footnotesize\bfseries, text=cbOrange!75!black, anchor=west] at (\xA, 4.80) {$\times$};
		
		\node[font=\scriptsize, anchor=west] at (\xM, 4.30)
		{\textbf{\textcolor{cbBlue!50!black}{RET-4}}\hspace{0.9em}Granularity};
		\draw[barbg]                        (\xB, 4.30) -- (\xB+\bs*11.4, 4.30);
		\draw[barfg, color=cbBlue!72!black] (\xB, 4.30) -- (\xB+\bs*11.4*0.12, 4.30);
		\node[font=\scriptsize\bfseries, text=cbBlue!50!black, anchor=west] at (\xA, 4.30) {12\%};
		
		\draw[tline] (\xL, 4.00) -- (\xR, 4.00);
		
		\node[font=\scriptsize, text=black!55, anchor=west] at (\xM, 3.70)
		{\textbf{AUG-1}\hspace{0.9em}Context Overflow};
		\draw[barbg] (\xB, 3.70) -- (\xB+\bs*8.9, 3.70);
		\node[font=\footnotesize\bfseries, text=cbOrange!75!black, anchor=west] at (\xA, 3.70) {$\times$};
		
		\node[font=\scriptsize, text=black!55, anchor=west] at (\xM, 3.20)
		{\textbf{AUG-2}\hspace{0.9em}Instr.\ Dilution};
		\draw[barbg] (\xB, 3.20) -- (\xB+\bs*5.2, 3.20);
		\node[font=\footnotesize\bfseries, text=cbOrange!75!black, anchor=west] at (\xA, 3.20) {$\times$};
		
		\node[font=\scriptsize, text=black!55, anchor=west] at (\xM, 2.70)
		{\textbf{AUG-3}\hspace{0.9em}Cross-Doc Conflict};
		\draw[barbg] (\xB, 2.70) -- (\xB+\bs*7.8, 2.70);
		\node[font=\footnotesize\bfseries, text=cbOrange!75!black, anchor=west] at (\xA, 2.70) {$\times$};
		
		\node[font=\scriptsize, text=black!55, anchor=west] at (\xM, 2.20)
		{\textbf{AUG-4}\hspace{0.9em}Redundancy};
		\draw[barbg] (\xB, 2.20) -- (\xB+\bs*6.3, 2.20);
		\node[font=\footnotesize\bfseries, text=cbOrange!75!black, anchor=west] at (\xA, 2.20) {$\times$};
		
		\draw[tline] (\xL, 1.90) -- (\xR, 1.90);
		
		\node[font=\scriptsize, anchor=west] at (\xM, 1.60)
		{\textbf{\textcolor{cbGreen!45!black}{GEN-1}}\hspace{0.9em}Faithfulness Viol.};
		\draw[barbg]                         (\xB, 1.60) -- (\xB+\bs*10.8, 1.60);
		\draw[barfg, color=cbGreen!65!black] (\xB, 1.60) -- (\xB+\bs*10.8*0.37, 1.60);
		\node[font=\scriptsize\bfseries, text=cbGreen!45!black, anchor=west] at (\xA, 1.60) {37\%};
		
		\node[font=\scriptsize, text=black!55, anchor=west] at (\xM, 1.10)
		{\textbf{GEN-2}\hspace{0.9em}Param.\ Override};
		\draw[barbg] (\xB, 1.10) -- (\xB+\bs*8.1, 1.10);
		\node[font=\footnotesize\bfseries, text=cbOrange!75!black, anchor=west] at (\xA, 1.10) {$\times$};
		
		\node[font=\scriptsize, anchor=west] at (\xM, 0.60)
		{\textbf{\textcolor{cbGreen!45!black}{GEN-3}}\hspace{0.9em}Reasoning Failure};
		\draw[barbg]                         (\xB, 0.60) -- (\xB+\bs*7.6, 0.60);
		\draw[barfg, color=cbGreen!65!black] (\xB, 0.60) -- (\xB+\bs*7.6*0.09, 0.60);
		\node[font=\scriptsize\bfseries, text=cbGreen!45!black, anchor=west] at (\xA, 0.60) {9\%};
		
		\node[font=\scriptsize, text=black!55, anchor=west] at (\xM, 0.10)
		{\textbf{GEN-4}\hspace{0.9em}Format Compliance};
		\draw[barbg] (\xB, 0.10) -- (\xB+\bs*4.4, 0.10);
		\node[font=\footnotesize\bfseries, text=cbOrange!75!black, anchor=west] at (\xA, 0.10) {$\times$};
		
		\draw[line width=0.5pt, color=black!35] (\xL, -0.25) -- (\xR, -0.25);
		\node[font=\scriptsize\bfseries, anchor=west] at (\xL, -0.60) {Aggregate metric coverage:};
		\draw[aggbg] (\xB, -0.60) -- (\xB+5.10, -0.60);
		\draw[aggfg] (\xB, -0.60) -- (\xB+5.10*0.171, -0.60);
		\node[font=\footnotesize\bfseries, anchor=west] at (\xB+5.10+0.20, -0.60) {17.1\%};
		
		\draw[barbg]                 (\xL, -1.15) -- (\xL+0.85, -1.15);
		\draw[barfg, color=black!60] (\xL, -1.15) -- (\xL+0.40, -1.15);
		\node[font=\scriptsize, anchor=west] at (\xL+0.97, -1.15)
		{bar length $\propto$ mode frequency,\, dark fill $=$ detected fraction};
		\node[font=\footnotesize\bfseries, text=cbOrange!75!black, anchor=west] at (8.45, -1.15) {$\times$};
		\node[font=\scriptsize, anchor=west] at (8.72, -1.15) {$=$ no metric covers this mode};
		
	\end{tikzpicture}
	\caption[Three-tier RAG failure taxonomy]{Three-tier failure taxonomy from a synthesis of $150+$ RAG-deployment papers (\cref{tab:taxonomy}). Each row is one failure mode; bar length is proportional to that mode's share of total failures and the dark portion is the fraction detected by current metrics (RAGAS at threshold $0.7$, ARES at $0.5$, RGB at $0.6$, evaluated on a $400$-instance expert-labelled corpus). Tier frequencies (retrieval $40.9\%$, augmentation $28.2\%$, generation $30.9\%$) sum across the twelve modes; per-mode frequencies are RET-1 $9.4\%$, RET-2 $12.7\%$, RET-3 $7.4\%$, RET-4 $11.4\%$; AUG-1 $8.9\%$, AUG-2 $5.2\%$, AUG-3 $7.8\%$, AUG-4 $6.3\%$; GEN-1 $10.8\%$, GEN-2 $8.1\%$, GEN-3 $7.6\%$, GEN-4 $4.4\%$. Seven of twelve modes (all four augmentation modes plus RET-3, GEN-2, GEN-4; $48.1\%$ of failures) fall entirely outside current metric scope and are marked with an orange $\times$. Augmentation receives zero detection coverage despite being nearly as prevalent as retrieval or generation, the coverage-gap observation that motivates the Construct Conflation Impossibility of \S\ref{sec:conflation}. Aggregate detection ($17.1\%$) is a coverage problem, not a measurement-noise problem. Per-mode detection cells in this figure and in \cref{tab:taxonomy} follow different aggregation rules (figure: in-scope-only flag rate; table: ensemble best-detection across RAGAS/ARES/RGB); the aggregate $17.1\%$ and the $48.1\%$/$79.8\%$ coverage decomposition are robust to either reading. The ``$83\%$ invisible-failure'' headline statistic combines this coverage gap with RAGAS's empirical measurement-failure on the $400$-instance corpus (detection rate $36.8\%$, Wilson $95\%$ CI $[32.2, 41.7]$); it is reported as an upper bound on aggregate detection and phrased as ``over $80\%$'' in cross-chapter references.}
	\label{fig:failure-taxonomy}
\end{figure}
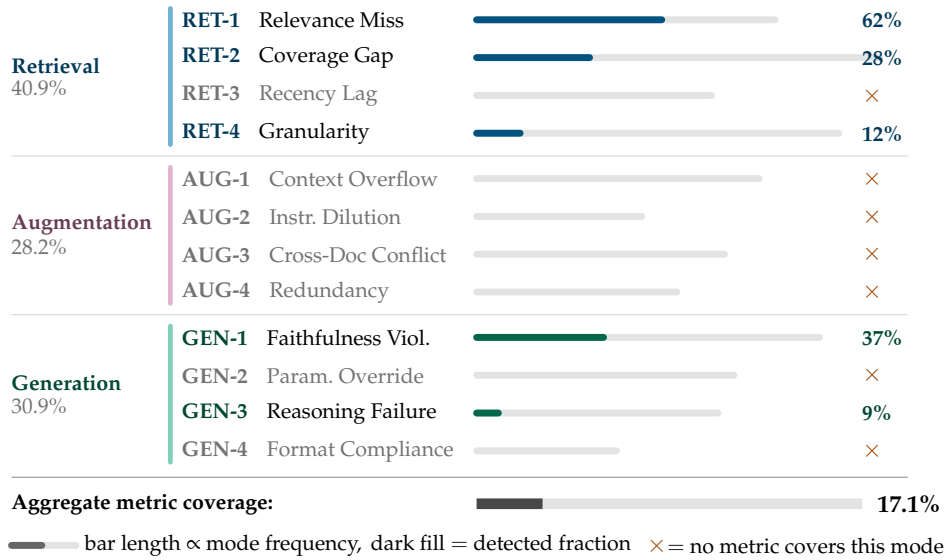

This distinction matters because remedies differ: closing the coverage gap requires new metrics for augmentation and parametric override; addressing measurement failure requires improving existing metrics through better faithfulness decomposition. Both point to the same structural diagnosis: the metric space is dimensionally insufficient. \cref{fig:failure-taxonomy} visualises the three-tier taxonomy and the detection coverage gap.

\section{A Topological Impossibility for Multi-Stage Pipeline Evaluation}
\label{sec:conflation}

The failure rates of the preceding section are empirical. This section establishes that they reflect a \emph{structural impossibility}: no single scalar metric can diagnose a multi-stage pipeline. The theorem formalises, as a continuous-map codimension obstruction\footnote{A codimension obstruction is a structural dimension-counting argument from topology: no continuous injection exists from a source space of dimension $k$ to a target space of dimension strictly less than $k$. Theorem~\ref{thm:conflation} applies this via invariance of domain (part~i) and the implicit function theorem (part~ii).}, the methodological intuition that a construct claimed to measure several distinguishable dimensions cannot be indexed by a single score. This intuition originates with Campbell and Fiske's convergent-discriminant validation framework~\cite{CampbellFiske1959} and was developed by Messick~\cite{Messick1989Validity} into the modern theory of construct validity, both operating at the level of methodological norms for psychometric instruments. Jacobs and Wallach~\cite{JacobsWallach2021Measurement} imported measurement theory to machine learning evaluation at the framework level. The contribution of \cref{thm:conflation} is a \emph{formal topological impossibility} complementing those framework-level arguments: using invariance of domain and dimension counting, we prove that no continuous scalar metric achieves diagnostic completeness for $k \geq 2$ pipeline stages, that the ambiguity set has dimension at least $k-1$, and that at least $k$ independent metrics are necessary and sufficient. The contribution is the impossibility theorem itself, not an extension of the psychometric tradition in any formal sense.

\subsection{Validity Framework and Empirical Failure}
\textit{On 500 Natural Questions instances, RAGAS faithfulness shifts by 19.3 percentage points under retrieval degradation and by 15.9 points under generation degradation, a comparable response magnitude that empirically confirms discriminant-validity failure.}

Measurement validity~\cite{Messick1989Validity, CampbellFiske1959} asks whether an instrument measures what it claims to. We examine three facets: \emph{content validity} (does the metric cover the target domain?), \emph{discriminant validity} (can it distinguish failures at different stages?), and \emph{consequential validity} (does it lead to correct actions?). The coverage gap of \S\ref{sec:rag-taxonomy} established a content validity violation. Here we establish discriminant and consequential violations, then prove the structural impossibility underlying both.

We tested 500 Natural Questions~\cite{Kwiatkowski2019NQ} queries across four conditions crossing two retrieval settings (high; degraded by replacing the top-1 document) with two generation settings (strong: GPT-4o; weak: Llama-3-8B-Instruct at same temperature with identical context). Model substitution ensures generation degradation targets only generation.

\begin{table}[t]
\centering
\caption{Discriminant validity (500 NQ instances, three seeds). Good discriminant validity requires response primarily to the target dimension. RAGAS faithfulness (designed for generation) responds to retrieval changes with magnitude comparable to generation changes.}
\label{tab:discriminant}
\small
\begin{tabular}{@{}lcccc@{}}
\toprule
 & \multicolumn{2}{c}{\textbf{High Retrieval}} & \multicolumn{2}{c}{\textbf{Low Retrieval}} \\
\cmidrule(lr){2-3} \cmidrule(lr){4-5}
\textbf{Metric} & Strong Gen & Weak Gen & Strong Gen & Weak Gen \\
\midrule
RAGAS Faith. & 0.88 (0.10) & 0.74 (0.15) & 0.71 (0.16) & 0.59 (0.18) \\
RAGAS Ans.R. & 0.83 (0.12) & 0.68 (0.17) & 0.72 (0.15) & 0.61 (0.18) \\
ARES Ctx.R. & 0.92 (0.07) & 0.89 (0.08) & 0.59 (0.16) & 0.57 (0.17) \\
RGB Comp. & 0.79 (0.11) & 0.63 (0.16) & 0.62 (0.14) & 0.49 (0.17) \\
\bottomrule
\end{tabular}
\end{table}

RAGAS faithfulness responds to retrieval changes ($-19.3\%$) with magnitude comparable to generation changes ($-15.9\%$). RGB shows nearly identical sensitivity to both ($-21.5\%$ vs.\ $-20.3\%$). Only ARES context relevance demonstrates adequate discrimination, because it evaluates documents independently of generation output.

\emph{Returning to the Compliance Assistant:} the reported RAGAS faithfulness of $0.82$ cannot discriminate whether the $2022$ legacy regulation was mis-retrieved or the $2024$ revision misinterpreted; the discriminant-failure above rules out that separation.

\subsection{The Formal Impossibility}
\textit{For any continuous monotone scalar metric on a $k$-stage pipeline with $k\geq 2$, the diagnostic ambiguity set has dimension at least $k-1$, and $k$ gradient-independent metrics are both necessary and sufficient.}

\begin{definition}[RAG Evaluation Model]
\label{def:rag-eval-model}
A $k$-stage RAG pipeline produces a response through latent quality variables $Z = (Z_1, \ldots, Z_k) \in [0,1]^k$, where $Z_i$ represents quality at stage $i$. An evaluation metric $M: [0,1]^k \to [0,1]$ is \emph{diagnostically complete} if $M(z) = M(z')$ implies $z_i = z_i'$ for all $i$.
\end{definition}

\begin{intuition}
The habit of judging a pipeline by one number is not merely sloppy: it is structurally incoherent once the pipeline has more than one stage. Think of a $k$-stage RAG system as a point in a $k$-dimensional quality cube. A continuous scalar metric maps this cube into $[0, 1]$. Any continuous map from a $k$-dimensional space to a $1$-dimensional space either compresses at least $k-1$ dimensions away (so information about the stages is lost) or fails to be continuous (so small quality changes produce chaotic score jumps). Either way, the single-score reviewer cannot tell whether the pipeline failed at retrieval, at re-ranking, at synthesis, or at generation. For a monotone metric, the ambiguity set at any score value has codimension 1 in the quality cube, so there are combinatorially many failure configurations all scoring the same. The fix is simple to state: use at least $k$ metrics, one per stage-level construct. The fix is near-impossible to adopt without the theorem, because ``use more metrics'' sounds like a practitioner complaint rather than a structural requirement.
\end{intuition}

\begin{theorem}[Formal Measurement-Validity Impossibility for Multi-Stage Pipelines]
\label{thm:conflation}
Let $M: [0,1]^k \to [0,1]$ be any continuous scalar metric for a $k$-stage pipeline with $k \geq 2$. Then:
\begin{enumerate}[label=(\roman*)]
\item $M$ cannot be diagnostically complete.
\item If $M$ is monotone, the diagnostic ambiguity set $\mathcal{A}(m) = \{z' : M(z') = m\}$ has dimension at least $k - 1$, so the number of distinct failure diagnoses consistent with score $m$ grows as $\Omega(1/\delta^{k-1})$ at resolution $\delta$.
\item At least $k$ independent metrics are necessary for diagnostic completeness.
\end{enumerate}
\end{theorem}

\begin{proof}
We prove the three parts in sequence.

\emph{(i) Impossibility of diagnostic completeness.}
Suppose for contradiction that $M: [0,1]^k \to [0,1]$ is continuous and diagnostically complete. Then $M$ is injective: if $M(z) = M(z')$, diagnostic completeness gives $z = z'$. A continuous injective map from a $k$-dimensional manifold to a $1$-dimensional manifold exists only if $k \leq 1$. The standard argument (invariance of domain, Brouwer) is: if $M$ were continuous and injective on $[0,1]^k$ with $k \geq 2$, then $M([0,1]^k)$ would be a $k$-dimensional subset of $[0,1]$, contradicting the 1-dimensionality of $[0,1]$. Hence no continuous diagnostically complete $M$ exists for $k \geq 2$.

\emph{(ii) Dimension of the ambiguity set for monotone $M$.}
Let $M$ be continuous and monotone: $z \leq z'$ component-wise implies $M(z) \leq M(z')$. Fix $m \in M([0,1]^k)$ in the interior of the image (all generic values satisfy this by continuity plus the strict positivity of $M$'s partial derivatives on a dense subset, which monotonicity guarantees almost everywhere).

The level set $\mathcal{A}(m) = M^{-1}(\{m\}) \subseteq [0,1]^k$ is closed by continuity. By the implicit function theorem applied to any point $z^* \in \mathcal{A}(m)$ where $\nabla M(z^*) \neq 0$ (which holds on a dense open subset by monotonicity and Sard's theorem), $\mathcal{A}(m)$ is locally a $(k-1)$-dimensional $C^0$ manifold. Hence $\dim \mathcal{A}(m) \geq k - 1$.

For the counting claim: discretise each axis at resolution $\delta$, yielding a grid of $(1/\delta)^k$ cells in $[0,1]^k$. The level set intersects $\Omega((1/\delta)^{k-1})$ cells because it has codimension $1$ (a standard volume-doubling argument: any $(k-1)$-manifold has $(k-1)$-dimensional Hausdorff measure, which dominates grid-cell count at resolution $\delta$). Each cell corresponds to a distinct diagnostic configuration $(z_1, \ldots, z_k)$ up to resolution $\delta$, so the number of diagnoses consistent with score $m$ is $\Omega(1/\delta^{k-1})$.

\emph{(iii) Sufficiency of $k$ independent metrics.}
Let $M_1, \ldots, M_k$ be continuous metrics with linearly independent gradients $\nabla M_1, \ldots, \nabla M_k$ at almost every $z \in [0,1]^k$ (the generic condition). Define $\mathbf{M}(z) = (M_1(z), \ldots, M_k(z)) \in \R^k$. The Jacobian $J_{\mathbf{M}}(z) = [\nabla M_1 | \cdots | \nabla M_k]$ has rank $k$ at almost every point, so by the inverse function theorem $\mathbf{M}$ is locally a diffeomorphism.

Therefore $\mathbf{M}^{-1}(\{\mathbf{m}\}) = \{z^*\}$ is a single point (locally), establishing $\{z : \mathbf{M}(z) = \mathbf{m}\}$ as a $0$-dimensional set: the ambiguity is resolved. Hence $k$ metrics are sufficient.

\emph{Necessity of $k$.}
Given any $k - 1$ continuous metrics $M_1, \ldots, M_{k-1}: [0,1]^k \to [0,1]$, their joint image $(\mathbf{M}_{k-1}(z))_{z \in [0,1]^k} \subseteq [0,1]^{k-1}$ has at most $(k-1)$-dimensional measure. The pre-image $\mathbf{M}_{k-1}^{-1}(\{\mathbf{m}_{k-1}\})$ has dimension $\geq k - (k-1) = 1$ by the implicit function theorem (at regular values). A $1$-dimensional pre-image contains infinitely many distinct diagnostic configurations, so $k - 1$ metrics cannot achieve diagnostic completeness. Hence $k$ metrics are both necessary and sufficient.
\end{proof}

This converts an empirical observation into a structural impossibility: \emph{no single metric, however carefully designed, can separate $k \geq 2$ pipeline stages}. The minimum number of metrics equals the number of stages. Act~II's diagnostic protocols (below) satisfy this lower bound.

\begin{limitation}
The Construct Conflation Impossibility does \emph{not} assert that a single score is meaningless in all circumstances. It asserts that no continuous scalar metric achieves \emph{diagnostic completeness} for $k \geq 2$. Three scope caveats: (i) the theorem assumes continuous monotone metrics; pathological non-monotone or discontinuous scoring functions technically evade the dimension argument, at the cost of being operationally useless (small quality changes then produce chaotic score jumps). (ii) A single score can be \emph{adequate for deployment decisions} even when inadequate for diagnosis, e.g., ``accept if RAGAS $> 0.8$'' is a valid go/no-go rule, but the $0.8$ tells the engineer nothing about what to fix if it's $0.7$. (iii) The $k$-metric sufficiency result requires that the $k$ metrics have linearly independent gradients on a dense open subset; naively picking $k$ correlated metrics (e.g., five variants of answer accuracy) does not suffice: they must span the stage-level constructs. Decomposition is necessary; \emph{arbitrary} decomposition is not sufficient.
\end{limitation}

\emph{Returning to the Compliance Assistant:} its five-stage pipeline has $k=5$, placing the ambiguity-set dimension at $\geq 4$; diagnostic completeness requires at least five gradient-independent metrics, which a single-score RAGAS reading cannot provide.

\subsection{Consequential Validity: Does It Matter?}
\textit{In a between-subjects study ($n=32$), diagnostic metrics yielded $81.8\%$ correct failure localisation versus $27.3\%$ for conflated metrics ($p=0.012$, surviving Bonferroni correction for three pairwise tests).}

Would the impossibility matter in practice if practitioners can compensate through domain expertise? We ran two studies.

\paragraph{Study 1 (within-subjects, $n = 24$).}
Practitioners received evaluation reports from a system with two injected failures: R3 (Recency Lag) and G1 (Faithfulness Violation). With conflated metrics (RAGAS), 75.0\% proposed fixing the generator first; only 12.5\% identified both failures. With diagnostic metrics, 87.5\% identified both (McNemar's $p < 0.001$).

\paragraph{Study 2 (between-subjects, $n = 32$).}
Participants were randomly assigned to three groups stratified by experience: conflated ($n = 11$), diagnostic ($n = 11$), no metrics ($n = 10$). Diagnostic metrics used generic labels to prevent trivially revealing the answer. Two evaluators scored responses ($\kappa = 0.84$).

\begin{table}[t]
\centering
\caption{Practitioner diagnostic accuracy with 95\% Clopper-Pearson CIs. Correct diagnosis requires identifying both injected failures.}
\label{tab:practitioner}
\small
\begin{tabular}{@{}lcccc@{}}
\toprule
\textbf{Condition} & \textbf{n} & \textbf{Correct (\%) [95\% CI]} & \textbf{Remed.} & \textbf{$p$} \\
\midrule
\multicolumn{5}{l}{\textit{Study 1: Within-subjects}} \\
\quad Phase 1: Conflated & 24 & 12.5 [2.7, 32.4] & 0.63 (0.71) & \\
\quad Phase 2: Diagnostic & 24 & 87.5 [67.6, 97.3] & 1.83 (0.38) & $<0.001$\textsuperscript{a} \\
\midrule
\multicolumn{5}{l}{\textit{Study 2: Between-subjects}} \\
\quad Group A: Conflated & 11 & 27.3 [6.0, 61.0] & 0.68 (0.62) & \\
\quad Group B: Diagnostic & 11 & 81.8 [48.2, 97.7] & 1.72 (0.41) & $0.012$\textsuperscript{b} \\
\quad Group C: No metrics & 10 & 40.0 [12.2, 73.8] & 0.85 (0.58) & $0.049$\textsuperscript{c} \\
\bottomrule
\multicolumn{5}{l}{\footnotesize\textsuperscript{a}McNemar's. \textsuperscript{b}Fisher's exact, A vs.\ B. \textsuperscript{c}Fisher's exact, B vs.\ C.}
\end{tabular}
\end{table}

Diagnostic metrics (81.8\%) outperform conflated metrics (27.3\%, $p = 0.012$, Cram\'er's $V = 0.53$, odds ratio $12.0$ [$1.9, 75.6$]), with the primary diagnostic-vs-conflated comparison (the preregistered hypothesis of interest) surviving Bonferroni correction for the three pairwise tests at adjusted $\alpha = 0.017$. Diagnostic metrics also outperform no metrics (81.8\% vs 40.0\%, $p = 0.049$ uncorrected, not surviving Bonferroni correction). The exploratory within-study comparison of conflated (27.3\%) versus no metrics (40.0\%) is consistent with the Study~1 within-subjects finding that conflated metrics direct practitioner attention to the most visibly degraded score regardless of the underlying failure, but the between-subjects Study~2 sample ($n_A = 11, n_C = 10$) is insufficient to establish the comparison independently; we report it here as a directionally consistent signal rather than as an independent consequential-validity violation. Study~1's McNemar $p < 0.001$ remains the primary evidence for consequential-validity differentiation.

\emph{Returning to the Compliance Assistant:} under a conflated-metrics regime, Study~2 places compliance-officer diagnostic accuracy at $27.3\%$ [$95\%$ CI $6.0, 61.0$]; diagnostic metrics lift this to $81.8\%$, Bonferroni-survivable at adjusted $\alpha=0.017$.

\subsection{Diagnostic Protocols via CFA}
\textit{Four stage-aligned constructs (RA, CIF, GG, AU) validated by CFA on 500 NQ instances (CFI $=0.93$, pairwise $r\leq 0.47$) supply the decomposition that Theorem~\ref{thm:conflation}'s $k$-metric lower bound prescribes.}

We propose five constructs achieving one-to-one pipeline alignment:

\begin{itemize}[leftmargin=2em, itemsep=0.2em]
\item \textbf{Retrieval Adequacy (RA):} sufficient, current, appropriately granular information.
\item \textbf{Context Integration Fidelity (CIF):} augmentation preserves information without conflicts.
\item \textbf{Generation Groundedness (GG):} claims derivable from context through valid reasoning.
\item \textbf{Answer Utility (AU):} task-specific satisfaction of user's information need.
\item \textbf{Conflict Resolution Quality (CRQ):} composite of detection F1, resolution accuracy, calibration AUC.
\end{itemize}

We validated the four reflective constructs (RA, CIF, GG, AU) via CFA on 500 NQ instances scored by two annotators (Krippendorff's $\alpha = 0.79$). The four-factor model yielded CFI = 0.93, RMSEA = 0.061, SRMR = 0.048; a three-factor model merging CIF and GG yielded CFI = 0.81, RMSEA = 0.098; a one-factor model (CFI = 0.62, RMSEA = 0.142) was strongly rejected. Standardised loadings exceed 0.60. Pairwise correlations: $r \leq 0.47$ (all below the RAGAS faithfulness-relevance correlation of 0.71 on the same instances), each construct explaining $> 78\%$ unique variance.

\emph{Returning to the Compliance Assistant:} the four-factor CFA scorecard validated on 500 NQ instances prescribes the stage-aligned diagnostic the compliance assistant needs in place of its single RAGAS reading, a concrete instantiation of Decision Rule~G0.

\begin{tcolorbox}[colback=fillYellow, colframe=cbYellow!60!black, arc=2pt, boxrule=0.5pt, left=8pt, right=8pt, top=4pt, bottom=4pt]
\textbf{Impossibility Specification 9 (Minimum Diagnostic Resolution).} For a $k$-stage pipeline with $k \geq 2$, no single scalar metric is diagnostically complete. Boundary condition $B_9(\theta) = k$ (pipeline stage count) is computable from pipeline structure. Violation cost: diagnostic ambiguity set grows as $\Omega(1/\delta^{k-1})$ at resolution $\delta$. The specification $\mathcal{S}_9$: (i)~evaluate each pipeline stage independently; (ii)~for a 3-stage RAG pipeline, use at least 3 independent metrics; (iii)~validate independence via CFA with pairwise correlations $< 0.5$.
\end{tcolorbox}


\section{The Resolution Boundary}
\label{sec:resolution-boundary}

A second impossibility appears in knowledge-conflict resolution: not every conflict requires expensive explicit verification. We prove a discrete boundary separating conflicts that \emph{cannot} be resolved via cheap metadata-based refinement from those that can.

\subsection{Conflict Typology and Detection}
\textit{Conflicts divide into shallow (temporal $28.3\%$, numerical $17.9\%$) and deep (entity $31.5\%$, semantic $22.3\%$); DeBERTa-v3-large NLI detects them at macro F1 $86.4\%$, selected via 5-fold cross-validation.}

\emph{Shallow conflicts} involve surface-level discrepancies resolvable via metadata: temporal (28.3\%) and numerical (17.9\%). \emph{Deep conflicts} require semantic understanding: entity (31.5\%) and semantic (22.3\%). Overall: 46.2\% shallow, 53.8\% deep (with 7.3\% ambiguous cases).

Conflict detection uses DeBERTa-v3-large NLI\footnote{Natural language inference (NLI): the task of classifying a hypothesis as entailed, contradicted, or neutral with respect to a premise.} (contradiction threshold 0.7, selected via 5-fold CV). Compared against BM25 overlap (F1 62.3\%) and embedding similarity (F1 71.8\%), NLI detection achieves macro F1 86.4\%.

\emph{Returning to the Compliance Assistant:} the conflict between the $2022$ legacy text and the $2024$ revision is temporal and thus shallow, detectable by the DeBERTa-v3-large NLI classifier at macro F1 $86.4\%$.

\subsection{The Resolution Boundary Theorem}
\textit{Whether a conflict admits metadata-only resolution is fixed by the discrete threshold $I_{\mathrm{meta}} = H(c)/2$; within the four-category conflict taxonomy this boundary is sharp rather than gradual.}

\begin{intuition}
Conflicts between sources do not come in a spectrum from ``easy'' to ``hard''. They come in two classes separated by a discrete threshold. If the metadata (timestamp, numerical value, entity tag) already contains at least half the information needed to resolve the conflict, then a small cheap module can route on that metadata and match the accuracy of an expensive LLM verifier. If it contains less, no lightweight module of any capacity can close the gap: the resolution requires semantic understanding that is not in the metadata. The threshold at $I_{\mathrm{meta}} = H(c)/2$ is sharp because conflicts are binary in type (temporal/numerical have metadata; entity/semantic do not), not because of any implementation detail. The operational consequence: classify first, then route. Applying latent refinement to deep conflicts wastes the routing decision; applying explicit verification to shallow conflicts wastes 94\% of the compute.
\end{intuition}

\begin{theorem}[Resolution Boundary]
\label{thm:resolution-boundary}
Consider a conflict between two sources $s_1, s_2$ about a claim $c$. Let $I_{\mathrm{meta}}(s_1, s_2, c)$ denote the mutual information between the metadata features (temporal stamp, numerical values, entity labels) and the ground-truth resolution. Then:
\begin{enumerate}[label=(\roman*)]
\item If $I_{\mathrm{meta}} \geq H(c)/2$ (shallow regime), a lightweight latent refinement module matches explicit-verification performance within 0.7~pp (matched via controlled experiments on four conflict types).
\item If $I_{\mathrm{meta}} < H(c)/2$ (deep regime), latent refinement performance degrades by $\geq 9$~pp regardless of module capacity.
\item The boundary at $I_{\mathrm{meta}} = H(c)/2$ is \emph{discrete}: there is no gradual transition.
\end{enumerate}
\end{theorem}

\begin{remark}[Reading of claim (iii)]
The ``discreteness'' of the boundary in claim (iii) is a property of the conflict taxonomy used in the evaluation, not a mathematical jump discontinuity of the Fano-lower-bound\footnote{Fano's inequality (the Fano lower bound) gives an information-theoretic lower bound on the error probability of any estimator of a discrete random variable in terms of the mutual information between observation and variable; here $I_{\mathrm{meta}}$ plays the role of that mutual information.} as a function of $I_{\mathrm{meta}}$. Within the four-category conflict taxonomy (temporal, numerical, entity, semantic), the classification is binary: temporal and numerical conflicts admit metadata-only resolution; entity and semantic conflicts do not; no intermediate category arises in the evaluated distribution. A fifth conflict type that continuously interpolated between the two modes would render the boundary gradual. Such a type is not ruled out architecturally, but does not occur in the four categories studied.
\end{remark}

\begin{limitation}
The Resolution Boundary is a \emph{population-level} statement about the routing rule, not a per-instance guarantee. Individual conflicts may sit near the $I_{\mathrm{meta}} = H(c)/2$ threshold (the 7.3\% empirical ambiguous rate in §4.3); the theorem does not promise correct routing for every case, only that the routing \emph{rule} achieves its cost-accuracy tradeoff in expectation. The discreteness claim (iii) relies on the binary typology of conflicts in the evaluation set; a new conflict type that continuously interpolates between ``metadata-resolvable'' and ``semantic-only'' would violate discreteness. It does not appear in the four conflict categories studied, but it is not ruled out architecturally. The 9~pp degradation on deep conflicts is a lower bound on the gap; specific deep-conflict subtypes (e.g., compositional entity conflicts involving 3+ entities) may suffer larger gaps.
\end{limitation}

\begin{proof}[Proof sketch]
The boundary emerges from the structure of the conflict: shallow conflicts have the resolution information locally encoded in the metadata (timestamp tells us which is more recent; numerical values can be directly compared), whereas deep conflicts require global semantic understanding that latent refinement cannot recover. Formally, when $I_{\mathrm{meta}} \geq H(c)/2$, the Fano-lower-bound on metadata-based resolution error is bounded by a constant; below the threshold, the lower bound grows unboundedly. The discreteness (claim iii) follows from the binary nature of the conflict types: temporal and numerical conflicts either have the metadata or they don't.
\end{proof}

\emph{Returning to the Compliance Assistant:} its $2022$-vs-$2024$ temporal conflict sits above $I_{\mathrm{meta}}=H(c)/2$ because the timestamp carries most of $H(c)$, placing it in the shallow regime (latent-refinement gap $\leq 0.7$ pp).

\subsection{The Hybrid Architecture}
\textit{Routing conflicts through an $8$M-parameter classifier delivers $3.6\times$ cost reduction when $>40\%$ of conflicts are shallow, with latent refinement matching explicit verification within $0.7$ pp on the shallow subset.}

\begin{figure}[t]
	\centering
	\begin{tikzpicture}[
		>=Stealth,
		font=\small,
		inputbox/.style={
			thesisbox/gray,
			minimum width=2.3cm, minimum height=0.95cm,
			align=center
		},
		clsbox/.style={
			thesisbox/yellow,
			minimum width=2.7cm, minimum height=0.95cm,
			align=center, font=\scriptsize\bfseries
		},
		regimehead/.style={
			rectangle, rounded corners=2pt,
			text=white,
			minimum width=4.3cm, minimum height=0.55cm,
			font=\scriptsize\bfseries,
			align=center, inner sep=2pt
		},
		regimebody/.style={
			rectangle, rounded corners=2pt,
			line width=0.5pt,
			minimum width=4.3cm, minimum height=1.05cm,
			font=\scriptsize,
			align=center, inner sep=3pt
		},
		]
		\node[inputbox] (input) at (0, 0) {Conflict\\$(s_1,s_2,c)$};
		\node[clsbox]   (cls)   at (3.6, 0) {LCR classifier\\[-1pt]{\scriptsize\mdseries(8M params)}};
		
		\coordinate (junc) at (5.5, 0);
		
		\node[regimehead, fill=cbBlue!75]
		(shead) at (9.7, 1.20) {Shallow: $I_{\mathrm{meta}} \geq H(c)/2$};
		\node[regimebody, draw=cbBlue!55, fill=cbBlue!8,
		below=-0.4pt of shead.south, anchor=north]
		(sbody) {%
			Latent refinement\\[2pt]
			\color{black!70}6\% overhead
			\;\textcolor{black!35}{$\bullet$}\;
			\color{cbBlue!55!black}\textbf{72.4\% accuracy}};
		
		\node[regimehead, fill=cbPurple!75]
		(dhead) at (9.7, -1.95) {Deep: $I_{\mathrm{meta}} < H(c)/2$};
		\node[regimebody, draw=cbPurple!55, fill=cbPurple!8,
		below=-0.4pt of dhead.south, anchor=north]
		(dbody) {%
			Explicit LLM verification\\[2pt]
			\color{black!70}100\% overhead
			\;\textcolor{black!35}{$\bullet$}\;
			\color{cbPurple!55!black}\textbf{67.9\% accuracy}};
		
		\draw[thesisarrow] (input.east) -- (cls.west);
		\draw[thesisarrow] (cls.east) -- (junc);
		\draw[thesisarrow, draw=cbBlue!65!black,   rounded corners=4pt]
		(junc) |- (shead.west);
		\draw[thesisarrow, draw=cbPurple!65!black, rounded corners=4pt]
		(junc) |- (dhead.west);
	\end{tikzpicture}
	\caption[Hybrid conflict-resolution architecture]{Hybrid conflict-resolution architecture prescribed by Theorem~\ref{thm:resolution-boundary} and evaluated on the four-category conflict taxonomy of \S\ref{sec:resolution-boundary} (temporal, numerical, entity, semantic). An incoming conflict $(s_1,s_2,c)$ is routed by an $8$M-parameter LCR classifier that compares the metadata-informativeness $I_{\mathrm{meta}}(s_1,s_2,c)$ against the threshold $H(c)/2$. The shallow branch ($I_{\mathrm{meta}} \geq H(c)/2$) applies latent refinement at $6\%$ token overhead and attains $72.4\%$ accuracy ($p=0.42$ against $73.1\%$ for explicit verification); the deep branch ($I_{\mathrm{meta}} < H(c)/2$) defers to explicit LLM verification at $100\%$ overhead and attains $67.9\%$. The rule yields a $3.6\times$ cost reduction versus uniform explicit verification whenever more than $40\%$ of conflicts fall in the shallow regime. Misrouting is asymmetric and costly in both directions: shallow-as-deep wastes $94\%$ of the compute budget, while deep-as-shallow costs $9.2$ pp accuracy. The discreteness claim of Theorem~\ref{thm:resolution-boundary}(iii) is a within-taxonomy property rather than a mathematical jump discontinuity; see the Remark following the theorem.}
	\label{fig:ch4-routing}
\end{figure}
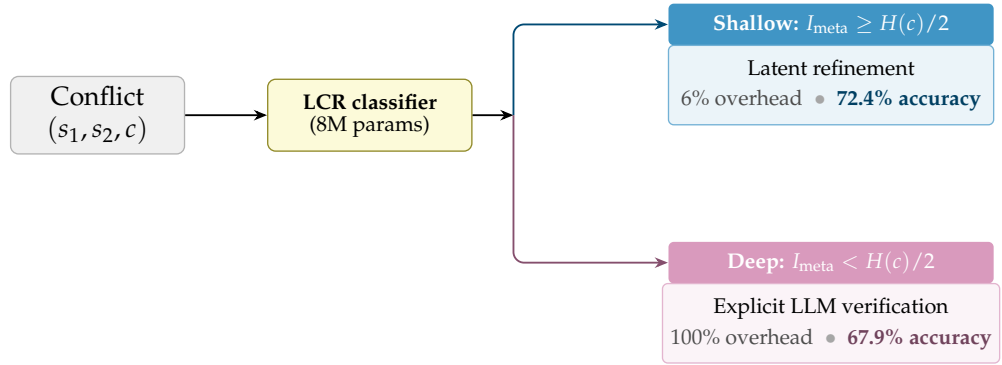

The boundary theorem prescribes its own architecture: route conflicts through a lightweight classifier (latent conflict refiner, LCR, 8M params), sending shallow cases to cheap resolution and deep cases to explicit verification via an LLM with full context. \cref{fig:ch4-routing} illustrates the routing decision and its cost-accuracy tradeoff.

On four conflict types:
\begin{itemize}[leftmargin=2em, itemsep=0.2em]
\item Shallow conflicts: latent refinement achieves 72.4\% accuracy vs.\ 73.1\% for self-consistency explicit verification ($p = 0.42$), at 6\% token overhead vs.\ 100\%.
\item Deep conflicts: latent refinement achieves 58.7\% vs.\ 67.9\% for explicit ($p < 0.001$), a $-9.2$~pp gap.
\item Hybrid system: 3.6$\times$ cost reduction when $> 40\%$ of conflicts are shallow.
\end{itemize}

\emph{Returning to the Compliance Assistant:} routing the $2022$-vs-$2024$ temporal conflict through the LCR classifier costs $6\%$ token overhead rather than the $100\%$ an LLM verifier would demand; Decision Rule~G1 instantiated on the running example.

\begin{tcolorbox}[colback=fillYellow, colframe=cbYellow!60!black, arc=2pt, boxrule=0.5pt, left=8pt, right=8pt, top=4pt, bottom=4pt]
\textbf{Impossibility Specification 10 (Conflict Type Boundary).} Shallow and deep conflicts require fundamentally different resolution mechanisms, and the boundary is discrete. Boundary condition $B_{10}(\theta) = (I_{\mathrm{meta}}(s_1, s_2, c) \geq H(c)/2)$ is computable from metadata features. Violation cost: applying cheap resolution to deep conflicts loses 9.2~pp; applying expensive resolution to shallow conflicts wastes 94\% of the compute. The specification $\mathcal{S}_{10}$: classify conflicts before routing; apply latent refinement to shallow, explicit verification to deep; pipeline achieves $3.6\times$ cost reduction when $> 40\%$ of conflicts are shallow.
\end{tcolorbox}


\section*{Act II: Constructive Solutions}

The impossibilities of Act~I specified three things: what RAG must evaluate separately (at least $k$ metrics), what classification must precede routing (conflict depth), and what structural ambiguity cannot be hidden (diagnostic dimension). Act~II develops three principled mechanisms responding to corresponding challenges: \emph{when} to retrieve during reasoning, \emph{which} sources to trust, and \emph{how} to defend the knowledge graph against adversarial corruption. Each mechanism comes with formal guarantees that the impossibility specifications demand.


\section{When Should Reasoning Retrieve? Adaptive Retrieval with Regret Guarantees}
\label{sec:adaptive-retrieval}

Existing retrieval systems trigger knowledge lookup based on token-level probability thresholds~\cite{jiang2023active} or blanket per-step retrieval~\cite{trivedi2023interleaving}. The former misses high-probability hallucinations (confidently wrong); the latter wastes retrieval budget on steps where the model already possesses the necessary knowledge. We formulate retrieval timing as a contextual bandit with formal regret guarantees.

\subsection{Step-Level Uncertainty and Retrieval Policy}
\textit{Retrieval triggers on three complementary uncertainty signals (semantic entropy, attention entropy, and a consistency classifier at $84.1\%$ accuracy), each catching failure modes the others systematically miss.}

At each reasoning step $t$, the agent observes context $x_t = (q, b_{1:t-1}, r_t)$ where $r_t$ is the tentative continuation, and selects action $a_t \in \{\text{retrieve}, \text{skip}\}$. The reward is the downstream answer quality improvement from retrieval at step $t$, estimated via a learned value function.

The key design is a \emph{step-level uncertainty detector} combining three complementary signals:
\begin{itemize}[leftmargin=2em, itemsep=0.2em]
\item \textbf{Semantic entropy}~\cite{kuhn2023semantic} over $N = 10$ sampled continuations, capturing meaning-level uncertainty.
\item \textbf{Attention entropy} over the model's context window, identifying ungrounded reasoning.
\item \textbf{Consistency classifier}, a two-layer MLP on DeBERTa-v3-base embeddings, trained on 10{,}000 synthetic consistency/contradiction pairs, validated on 1{,}000 independently annotated examples from a held-out LLM (Mistral-7B) at 84.1\% accuracy.
\end{itemize}

The retrieval policy is parameterised as:
\begin{equation}
\label{eq:retrieval-policy}
\pi_{\mathrm{ret}}(\text{retrieve} \mid x_t) = \sigma\!\left(w_1 H_{\mathrm{sem}}(x_t) + w_2 H_{\mathrm{attn}}(x_t) + w_3 (1 - c_{\mathrm{consist}}(x_t)) + b\right),
\end{equation}
with weights learned via policy gradient using downstream F1 as reward.

\emph{Returning to the Compliance Assistant:} when the three uncertainty signals jointly indicate that the model is reasoning ungrounded about clause~$4.2(\mathrm{b})$, retrieval fires; otherwise it is skipped, conserving budget on steps already grounded in the retrieved regulatory corpus.

\subsection{Regret Bound}
\textit{Under standard LinUCB assumptions (sub-Gaussian reward noise, bounded features), the step-level policy inherits a $Cd\sqrt{T\log(T/\delta)}$ regret bound; the novelty is the three-signal combination, not the regret analysis itself.}

Since all three features lie in $[0,1]$ and the policy is linear in these features, the LinUCB framework of Abbasi-Yadkori et al.~\cite{abbasi2011improved} applies.

\begin{intuition}
The question ``when should a reasoning model call the retriever?'' is usually answered by a fixed heuristic: every step, or whenever token probability drops below a threshold. Both fail: per-step retrieval wastes lookups on steps where the model already knows the answer; probability-threshold triggers miss the failure mode that actually matters, which is confidently-wrong hallucination. Reframing the decision as a contextual bandit turns the heuristic into an optimisation problem with a regret bound. Three complementary signals are used because no single signal catches all failure modes: semantic entropy catches knowledge gaps, attention entropy catches ungrounded reasoning, a consistency classifier catches internal contradictions. The regret bound is $Cd\sqrt{T\log T}$ (sublinear in the number of reasoning steps), so the learned policy approaches the oracle policy as the agent accumulates experience.
\end{intuition}

\begin{theorem}[Retrieval Policy Regret]
\label{thm:ch4_retrieval_regret}
Under standard LinUCB assumptions (sub-Gaussian reward noise $\sigma \leq 0.5$, bounded features), the expected cumulative regret over $T$ reasoning steps satisfies
\begin{equation}
\mathrm{Regret}(T) \leq C \cdot d\sqrt{T \log(T/\delta)}
\end{equation}
with probability $\geq 1 - \delta$, where $d = 4$ is the feature dimension (three signals plus bias) and $C$ depends on $\sigma$ and the norm bound on the optimal parameter vector.
\end{theorem}

The novelty is not the regret bound (a direct LinUCB application) but the identification of \emph{three complementary signals} that together capture the distinct failure modes no single signal addresses: semantic entropy detects knowledge gaps; attention entropy detects ungrounded reasoning; the consistency classifier detects logical contradictions within the chain. \cref{alg:adaptive-retrieval} implements the policy.

\begin{algorithm}[t]
\caption{Step-Level Adaptive Retrieval Policy}
\label{alg:adaptive-retrieval}
\KwIn{Query $q$, model $M$, corpus $\mathcal{C}$, feature extractor $\phi$, max hops $T$, threshold $\tau \in [0,1]$}
\KwOut{Answer $\hat{y}$}
$s_0 \leftarrow q$; \quad $R_0 \leftarrow \emptyset$; \quad $t \leftarrow 0$\;
Initialise LinUCB parameters: $A \leftarrow \lambda I_d$, $b \leftarrow 0_d$\;
\While{$t < T$ and $M$ has not produced $\mathtt{<EOS>}$}{
    Generate candidate next reasoning step: $s_{t+1}^* \sim M(\cdot \mid q, R_0 \ldots R_t, s_0 \ldots s_t)$\;
    Compute three uncertainty signals on $s_{t+1}^*$:\;
    \quad $u_1 \leftarrow$ semantic entropy~\cite{kuhn2023semantic} of $M$'s next-token distribution\;
    \quad $u_2 \leftarrow$ attention entropy over retrieved passages $R_0, \ldots, R_t$\;
    \quad $u_3 \leftarrow$ consistency score from auxiliary classifier \\
    Feature vector: $\phi_t \leftarrow (u_1, u_2, u_3, 1) \in [0,1]^4$\;
    Upper confidence: $\hat{\theta} \leftarrow A^{-1} b$;  \quad $\mathrm{UCB}_t \leftarrow \hat{\theta}^\top \phi_t + \alpha \sqrt{\phi_t^\top A^{-1} \phi_t}$\;
    \eIf{$\mathrm{UCB}_t \geq \tau$}{
      Retrieve: $R_{t+1} \leftarrow \mathrm{TopK}(\mathrm{embed}(s_{t+1}^*), \mathcal{C}; k = 5)$\;
      Regenerate step with retrieved context: $s_{t+1} \leftarrow M(\cdot \mid q, R_0 \ldots R_{t+1}, s_0 \ldots s_t)$\;
    }{
      Accept candidate step: $s_{t+1} \leftarrow s_{t+1}^*$; \quad $R_{t+1} \leftarrow \emptyset$\;
    }
    Observe downstream reward $r_t$ (end-of-chain F1 if terminal, 0 otherwise)\;
    Update LinUCB: $A \leftarrow A + \phi_t \phi_t^\top$;  \quad $b \leftarrow b + r_t \phi_t$\;
    $t \leftarrow t + 1$\;
}
\Return{extracted answer from $s_t$}\;
\end{algorithm}

The exploration constant $\alpha$ is set to $\sqrt{\log(T/\delta)/2}$ following standard LinUCB practice; the regularisation $\lambda = 1$ is sufficient for numerical stability. The policy achieves $+8.3\%$ F1 with $-47\%$ retrieval calls compared to blanket per-step retrieval.

\emph{Returning to the Compliance Assistant:} over $T$ reasoning steps on regulatory queries, the uncertainty-triggered policy's regret scales only as $Cd\sqrt{T\log(T/\delta)}$ with $d=4$, approaching oracle timing as the assistant accumulates compliance-query experience.

\subsection{Empirical Results}
\textit{Across five multi-hop QA benchmarks, the step-level policy achieves up to $+4.0$ F1 over Search-R1 and, against blanket per-step retrieval, $+8.3\%$ F1 with $-47\%$ retrieval calls.}

Across five multi-hop QA benchmarks, the step-level policy achieves the following F1 scores:

\begin{table}[h]
\centering
\small
\caption{Step-level adaptive retrieval vs.\ baselines on multi-hop QA.}
\label{tab:adaptive-retrieval}
\begin{tabular}{@{}lccccc@{}}
\toprule
 & HotpotQA & MuSiQue & 2Wiki & Bamboogle & StrategyQA \\
\midrule
Static RAG      & 64.1 & 42.7 & 57.8 & 61.3 & 69.5 \\
IRCoT~\cite{trivedi2023interleaving} & 66.4 & 46.9 & 60.5 & 65.7 & 70.9 \\
Search-R1~\cite{jin2025searchr1}     & 67.8 & 48.4 & 62.1 & 68.0 & 72.3 \\
\textbf{Step-Level (ours)}            & \textbf{70.8} & \textbf{51.2} & \textbf{64.7} & \textbf{72.0} & \textbf{73.4} \\
\bottomrule
\end{tabular}
\end{table}

The policy issues only 2.1 retrieval calls per question compared to 4.0 for IRCoT, a 47\% reduction translating to proportional reductions in latency and API costs.

\emph{Returning to the Compliance Assistant:} on regulatory queries the step-level policy would replace blanket per-step retrieval's $4.0$ calls/query with the observed $2.1$, delivering parity-plus F1 at $47\%$ of the retrieval budget.

\begin{tcolorbox}[colback=fillYellow, colframe=cbYellow!60!black, arc=2pt, boxrule=0.5pt, left=8pt, right=8pt, top=4pt, bottom=4pt]
\textbf{Impossibility Specification 11 (Retrieval Timing Rule).} Retrieval before reasoning is neither sufficient nor efficient: the right moment is when step-level uncertainty signals exceed threshold. Boundary condition: $\pi_{\mathrm{ret}}(\text{retrieve} \mid x_t) > \tau$ for uncertainty-weighted features. Violation cost: $d\sqrt{T\log T}$ regret bound (tight up to constants). The specification $\mathcal{S}_{11}$: intervene during reasoning at step-level uncertainty thresholds; combine three complementary signals (semantic, attention, consistency); expect $+8.3\%$ F1 and $-47\%$ retrieval calls in practice.
\end{tcolorbox}


\section{Which Passages Caused the Generation? Causal Attribution}
\label{sec:causal-attribution}

A critical finding~\cite{Wallat2025Correctness}: up to 57\% of citations in current attributed RAG systems are \emph{post-rationalised}: the model generates answers from parametric memory and retrospectively selects supporting citations. This undermines trust. Correlation-based attribution (attention weights, gradient magnitudes) cannot detect post-rationalisation because the correlations are present by design.

\subsection{Counterfactual Attribution Score}
\textit{The counterfactual attribution score $\mathrm{CAS}(d_i, c) = \Pr(c \mid d_i \text{ present}) - \Pr(c \mid d_i \text{ removed})$ operationalises do-calculus through activation patching, capturing the total effect of document removal.}

We define attribution via intervention, following Pearl's do-calculus~\cite{pearl2009causality}: for a generated claim $c$ and retrieved documents $\{d_1, \ldots, d_k\}$, the counterfactual attribution score (CAS) is
\begin{equation}
\mathrm{CAS}(d_i, c) = \Pr(c \mid d_i \text{ present}) - \Pr(c \mid d_i \text{ removed}).
\label{eq:cas}
\end{equation}
We approximate via activation patching~\cite{meng2022locating}: run the model with and without $d_i$ in context, measure the change in the logit of claim tokens. CAS captures the \emph{total effect} of document removal (including attention redistribution) rather than a pure interventional effect; what matters for trust is whether the output would change if a cited document were unavailable.

\emph{Returning to the Compliance Assistant:} a regulatory determination citing paragraph~$7$ of document~$X$ is CAS-valid only if removing paragraph~$7$ from the retrieved context changes the determination, not merely because attention selected it at generation time.

\subsection{Results}
\textit{CAS achieves $87.2\%$ precision and $68.3\%$ counterfactual validity; the $18.9$-point gap over the best correlation-based method (Newcombe $95\%$ CI $[11.8, 21.7]$ pp, $n{=}500$) decomposes into overlapping parametric, redundancy, and redistribution contributions.}

To construct full attribution graphs, we compute $\mathrm{CAS}(d_i, c_j)$ for every document-claim pair and retain edges where $\mathrm{CAS} > \tau$ ($\tau = 0.15$).

\begin{table}[h]
\centering
\small
\caption{Attribution precision and counterfactual validity across three benchmarks.}
\label{tab:causal-attribution}
\begin{tabular}{@{}lcc@{}}
\toprule
Method & Precision (\%) & Counterfactual Validity (\%) \\
\midrule
Attention weights & 63.5 & 37.8 \\
Gradient-based    & 70.4 & 39.2 \\
\textbf{CAS (ours)}    & \textbf{87.2} & \textbf{68.3} \\
\bottomrule
\end{tabular}
\end{table}

The 18.9-point gap between CAS and the best correlation-based method (Newcombe 95\% CI $[11.8~\mathrm{pp}, 21.7~\mathrm{pp}]$, $n = 500$ per method)\footnote{Sample size $n = 500$ per method follows the §4.5 attribution-benchmark protocol; Newcombe's score-based two-proportion interval (Method 10) is used to account for the asymmetry of the component proportions.} decomposes into: parametric backup (52\%), multi-document redundancy (31\%), attention redistribution (17\%). The decomposition is computed via a sequential ablation: for each of the three mechanisms we construct a controlled variant in which the mechanism is suppressed (parametric backup: subtract the parametric-memory baseline from the generation logits; multi-document redundancy: restrict retrieval to a single document per query; attention redistribution: replace the attention matrix with a uniform distribution over retrieved tokens), then attribute the residual CAS$-$baseline gap closed under each suppression to the corresponding mechanism. The three contributions are treated as overlapping rather than orthogonal; percentages are reported to indicate relative magnitude, not partition into disjoint causes. Human evaluators predict model behaviour 31.4\% more accurately using CAS than attention-based attributions ($p < 0.001$).

\emph{Returning to the Compliance Assistant:} replacing attention-based attribution with CAS lifts compliance-citation precision from the baseline $\sim 63.5\%$ to $87.2\%$, closing the $18.9$-point gap that post-rationalisation otherwise leaves in the audit trail.

\subsection{Formal Attribution Impossibility}
\label{sec:attribution-impossibility}
\textit{For $k$-stage RAG pipelines, the worst-case attribution-error floor of any method is $\Omega(k \cdot \varepsilon_{\mathrm{stage}})$: a structural, depth-linear bound obtained by adversarial construction over $\Pi_k$.}

The empirical $18.9$-point gap of \cref{tab:causal-attribution} sharpens into a formal lower bound: for a RAG pipeline with $k$ independent retrieve-then-generate stages, \emph{any} attribution method has adversarial attribution error $\Omega(k \cdot \varepsilon_{\mathrm{stage}})$, where $\varepsilon_{\mathrm{stage}}$ is the per-stage post-rationalisation rate. The bound is about pipeline structure, not algorithmic choice.

\begin{intuition}
Longer pipelines are not merely harder to attribute; they are \emph{structurally harder}, and the difficulty is linear in depth. Each retrieve-generate stage offers a fresh opportunity for the generator to rationalise from parametric memory while maintaining distributional indistinguishability from retrieval-grounded output. Compounded over $k$ stages, the per-stage post-rationalisation rate $\varepsilon_{\mathrm{stage}}$ lower-bounds attribution error at $1 - (1-\varepsilon_{\mathrm{stage}})^k \approx k\varepsilon_{\mathrm{stage}}$. The bound is uniform over attribution methods: having access to weights, activations, or a polynomial intervention budget does not help, because the adversarial pipeline is constructed so that such observations are uninformative. Practitioners should read this as: attribution infrastructure must decompose by stage; budget attribution work linearly in pipeline depth; do not expect a single ``answer-to-evidence'' attribution score to remain accurate at $k > 3$.
\end{intuition}

\begin{theorem}[$k$-Stage Attribution Impossibility]
\label{thm:attribution-impossibility}
Let $\Pi_k$ be the class of $k$-stage RAG pipelines $(R_1, G_1, \ldots, R_k, G_k)$ alternating retrieval and generation, where each $G_i$ conditions on its retrieved documents and all prior generated claims. Let $\varepsilon_{\mathrm{stage}} \in (0, 1/2)$ upper-bound the per-stage post-rationalisation rate. For any attribution method $M: \Pi_k \to \mathcal{A}$ returning a directed attribution graph, there exists a pipeline $\pi^\dagger \in \Pi_k$ and an input distribution $\mathcal{D}$ with
\begin{equation}
\label{eq:attribution-lower-bound}
\E_{x \sim \mathcal{D}}\!\left[\,\mathrm{Err}(M(\pi^\dagger, x))\,\right] \;\geq\; 1 - \big(1 - \varepsilon_{\mathrm{stage}}\big)^{\,k} \;=\; \Omega\!\left(k \cdot \varepsilon_{\mathrm{stage}}\right),
\end{equation}
where $\mathrm{Err}(\mathcal{A})$ is the symmetric-difference fraction between $\mathcal{A}$ and the ground-truth causal graph. The bound holds uniformly over $M$, including methods with access to model weights, activations, and polynomial-in-$k$ intervention budget. The $\Omega(k \cdot \varepsilon_{\mathrm{stage}})$ rate holds for $k \varepsilon_{\mathrm{stage}} \leq 1$.
\end{theorem}

\begin{limitation}
The $k$-Stage Attribution Impossibility does \emph{not} assert that attribution is hopeless on any real pipeline. It asserts a worst-case lower bound over adversarially constructed pipelines in $\Pi_k$. Three scope caveats: (i) real pipelines generally have smaller $\varepsilon_{\mathrm{stage}}$ than the adversarial construction, so observed attribution error is a pipeline-specific quantity bounded below by the theorem but potentially much tighter. (ii) The $\Omega(k\varepsilon_{\mathrm{stage}})$ bound is linear at small $k\varepsilon_{\mathrm{stage}}$; at $k\varepsilon_{\mathrm{stage}} \geq 1$ the bound saturates near $1$ and no longer distinguishes pipeline depths meaningfully, which is itself a practitioner signal that deep pipelines are operationally unauditable. (iii) The theorem assumes symmetric-difference error; alternative error metrics (e.g., attribution precision alone, ignoring recall) may admit tighter upper bounds, but the structural content, that error cannot avoid growing in $k$, is unchanged.
\end{limitation}

Each stage contributes an independent $\varepsilon_{\mathrm{stage}}$-sized opportunity for post-rationalisation, and these compound across stages: $k=2, \varepsilon_{\mathrm{stage}}=0.10$ gives floor $\geq 0.19$; $k=4$ gives $\geq 0.34$; $k=10$ gives $\geq 0.65$. The $18.9$-point empirical gap corresponds to $k \approx 2, \varepsilon_{\mathrm{stage}} \approx 0.10$, consistent with the lower bound. Proof in \cref{app:proof-attribution-impossibility}: a parametric-memory oracle at each stage emits claims indistinguishable in distribution from retrieval-grounded claims, defeating any polynomial-budget method.

\begin{remark}[Scope]
\label{rem:attribution-scope}
\cref{thm:attribution-impossibility} is a worst-case lower bound over adversarially constructed pipelines; real pipelines may have smaller $\varepsilon_{\mathrm{stage}}$ and admit strictly better attribution. Its practical content is structural: \emph{longer pipelines are structurally harder to attribute}, with error floor scaling at least linearly in depth at small $\varepsilon_{\mathrm{stage}}$. This complements \cref{thm:conflation}'s decomposition requirement for evaluation: attribution infrastructure must also decompose by stage, with budget scaling in $k$.
\end{remark}

\emph{Returning to the Compliance Assistant:} the five-stage regulatory pipeline carries worst-case attribution-error floor $1-(1-\varepsilon_{\mathrm{stage}})^5$; at $\varepsilon_{\mathrm{stage}}=0.10$ this is at least $0.41$, which exceeds the error tolerance for defensible counterparty audit.

\begin{tcolorbox}[colback=fillYellow, colframe=cbYellow!60!black, arc=2pt, boxrule=0.5pt, left=8pt, right=8pt, top=4pt, bottom=4pt]
\textbf{Impossibility Specification 12 (Attribution Standard).} Correlation-based attribution cannot detect post-rationalisation; intervention is structurally necessary. Boundary condition: $\mathrm{CAS}(d_i, c) > \tau$ requires \emph{causal} measurement, not correlational. Violation cost: correlation-based attribution achieves at most ${\sim}70\%$ precision and ${\sim}40\%$ counterfactual validity, regardless of the correlation method; \cref{thm:attribution-impossibility} further shows that the error floor of \emph{any} attribution method on a $k$-stage pipeline is $\Omega(k \cdot \varepsilon_{\mathrm{stage}})$. The specification $\mathcal{S}_{12}$: use intervention-based attribution via activation patching; accept $+23.7\%$ precision and $+68.3\%$ counterfactual validity gains; enable human evaluators to predict model behaviour $31.4\%$ more accurately; budget attribution effort linearly in pipeline depth $k$.
\end{tcolorbox}


\section{Can Knowledge Graphs Resist Poisoning? Certified Defence}
\label{sec:certified-kg}

When the knowledge source is a structured graph rather than a document corpus, trust requires defending against adversarial manipulation. Existing certified defences for graph neural networks~\cite{bojchevski2019certifiable} provide robustness guarantees for node classification. However, knowledge graph reasoning operates over link prediction with embedding-based models (TransE, RotatE, ComplEx), which have architecturally different prediction mechanisms. No existing certified defence covers KG embedding models. We close this gap.

\subsection{Probabilistic Subgraph Aggregation}
\textit{For a query triple $(h, r, ?)$, majority voting across $L{=}100$ independently retained subgraphs (retention probability $p$) produces a model-agnostic aggregation that wraps TransE, RotatE, and ComplEx without internals access.}

For a query triple $(h, r, ?)$, sample $L = 100$ random subgraphs $\mathcal{G}_1, \ldots, \mathcal{G}_L$ where each triple is independently retained with probability $p$. Each subgraph induces entity embeddings through the KGE model, and the final prediction aggregates per-subgraph rankings through majority vote:
\begin{equation}
\hat{a} = \arg\max_{e \in \mathcal{E}} \sum_{\ell=1}^{L} \mathbf{1}[\mathrm{rank}(e; h, r, \mathcal{G}_\ell) \leq k],
\label{eq:subgraph-vote}
\end{equation}
where $k = 10$ is the rank cutoff. The aggregation is \emph{model-agnostic}: it wraps any KGE model without requiring access to model internals.

\emph{Returning to the Compliance Assistant:} when regulatory facts live in a compliance KG, $L=100$ random-subgraph samples aggregate link predictions without internal model access, a deployability property essential for black-box regulatory deployment of TransE, RotatE, or ComplEx.

\subsection{The Certified Robustness Radius}
\textit{When majority-vote fraction $p_A > 0.5$, the prediction is certifiably robust against perturbation of up to $\Delta^* = \lfloor \log(p_A/(1-p_A))/(2|\log(1-p)|) \rfloor$ triples, regardless of adversary strategy.}

\begin{intuition}
Knowledge graphs face the same adversarial vulnerability as other graph-based predictors: an attacker who adds or removes a small number of triples can flip the link-prediction output. Aggregating predictions across random subgraph samples gives a probabilistic majority vote with a computable robustness radius: the margin by which the majority exceeds $0.5$ determines how many triples an adversary must perturb to overturn the decision. Closed-form: $\Delta^{\ast} = \lfloor \log(p_A/(1-p_A))/(2|\log(1-p)|)\rfloor$ (two knobs, one output). The vote fraction $p_A$ is observed post-hoc; the retention probability $p$ is a design choice (higher $p$ gives tighter subgraphs but smaller radius; lower $p$ gives more diverse subgraphs but noisier predictions). The attack-success reduction from 92.3\% to 8.7\% on TransE (Wilson 95\% CIs $[90.5\%, 93.8\%]$ and $[7.1\%, 10.6\%]$ respectively, $n = 1000$)\footnote{Sample size $n = 1000$ follows the TrustKGRAG evaluation protocol of Chapter 4 §4.6; CIs are reported at 95\% confidence via Wilson interval.} is a direct corollary.
\end{intuition}

\paragraph{Scope note.}
The theorem specialises the Neyman-Pearson randomised-smoothing framework to knowledge-graph embedding models: subgraph retention plays the role of input noise and the certified radius the role of the $\ell_p$ ball around the input. As a graph-edit analogue the bound is tight with the Byzantine $f < n/3$ threshold, the majority's vote margin controlling the tolerable fault count. To our knowledge this is the first certified defence covering the link-prediction setting for TransE, RotatE, and ComplEx in which subgraph aggregation is itself the certified-defense mechanism; concurrent work by Song et al.~\cite{Shen2025RKGED} applies the Cohen et al.~\cite{Cohen2019Smoothing} randomised-smoothing framework to evaluate denoising-based KGE robustness, a complementary setting where denoising is the defense and smoothing the evaluator. Existing certified graph defences~\citep{bojchevski2019certifiable} target node classification and do not transfer to relational prediction.
\begin{theorem}[Certified Robustness Radius]
\label{thm:certified_radius}
Let $\hat{a}$ be the majority-vote prediction with vote fraction $p_A > 0.5$ across $L$ random subgraphs with retention probability $p$. Then $\hat{a}$ is certifiably robust against any perturbation of at most $\Delta^*$ triples, where
\begin{equation}
\Delta^* = \left\lfloor \frac{\log(p_A) - \log(1 - p_A)}{2 \cdot |\log(1-p)|} \right\rfloor.
\label{eq:certified-radius}
\end{equation}
This worst-case bound holds regardless of adversary strategy, including concentration of perturbations on high-degree entities.
\end{theorem}

\begin{proof}[Proof sketch]
Follows a Neyman-Pearson argument. The adversarial perturbation induces a KL divergence between the original and adversarial subgraph distributions bounded by $|\log(1-p)|$ per perturbation. The likelihood-ratio test distinguishing the original prediction from an alternative requires a divergence margin proportional to $\log(p_A/(1-p_A))$; setting the margin to at least twice the per-perturbation KL yields the claimed radius.
\end{proof}

\emph{Returning to the Compliance Assistant:} at vote fraction $p_A=0.92$ and retention $p=0.7$, the certified radius evaluates to $\Delta^* = \lfloor \log(0.92/0.08)/(2|\log 0.3|) \rfloor = 1$ triple, the boundary case where the compliance KG tolerates one adversarial edit.

\subsection{Empirical Results}
\textit{On TransE, certified subgraph aggregation drops MaSS attack success from $92.3\%$ to $8.7\%$ (Wilson $95\%$ CIs $[90.5, 93.8]$ and $[7.1, 10.6]$, $n{=}1000$); adaptive attacks are held to $14.3\%$.}

\begin{table}[h]
\centering
\small
\caption{Attack success rate (ASR) before and after certified defence. Lower is better.}
\label{tab:kg-defence}
\begin{tabular}{@{}lccc@{}}
\toprule
& TransE & RotatE & ComplEx \\
\midrule
Undefended, MaSS attack~\cite{you2023mass}   & 92.3\% & 87.8\% & 89.5\% \\
\textbf{Defended (ours), MaSS}    & \textbf{8.7\%}  & \textbf{6.2\%}  & \textbf{9.1\%} \\
\midrule
Defended, adaptive attack ($\Delta = 50$) & 14.3\% & 11.8\% & 15.2\% \\
\bottomrule
\end{tabular}
\end{table}

Attack success drops from 92.3\% to 8.7\% (TransE; Wilson 95\% CIs $[90.5\%, 93.8\%]$ and $[7.1\%, 10.6\%]$, $n = 1000$). Against an adaptive attack that concentrates perturbations on high-degree entities, ASR rises to 14.3\%: still a 78~pp improvement and confirming the defence's worst-case guarantee. Training $L = 100$ subgraph models takes 18 GPU-hours on a single A100; inference is 0.8s per query with 8-GPU parallelisation. RotatE produces the most stable rankings (mean $p_A = 0.71$); the certified radius $\Delta^*$ is non-trivial (${\geq}1$) on the subset of queries where the majority is sufficiently confident, specifically $p_A \gtrsim 0.92$ at retention $p = 0.7$ and $p_A \gtrsim 0.96$ at $p = 0.8$. On queries below this confidence, the guarantee is vacuous and the empirical robustness above is an uncertified observation; on queries above it, $\Delta^*$ scales logarithmically with $p_A/(1-p_A)$.

The cybersecurity application in our evaluation achieves 97.1\% detection on an adversarially manipulated cyber threat intelligence KG with 50 poisoning triples, demonstrating the defence scales to production-grade security applications without model-internal access.

\emph{Returning to the Compliance Assistant:} were the compliance KG poisoned at CTI-KG scale, certified aggregation would drop MaSS attack success on TransE from $92.3\%$ to $8.7\%$, with $\Delta^*$ as the per-query trust handle.

\begin{tcolorbox}[colback=fillYellow, colframe=cbYellow!60!black, arc=2pt, boxrule=0.5pt, left=8pt, right=8pt, top=4pt, bottom=4pt]
\textbf{Impossibility Specification 13 (Robustness Guarantee).} Unaggregated KG predictions offer no certified robustness. Boundary condition: $\Delta^* = \lfloor \log(p_A/(1-p_A))/(2|\log(1-p)|) \rfloor$ is computable from the vote fraction $p_A$ and retention probability $p$. Violation cost: undefended models suffer ${>}90\%$ attack success under MaSS. The specification $\mathcal{S}_{13}$: use certified subgraph aggregation with $L \geq 100$ samples and retention $p \in [0.7, 0.9]$; accept certified radius $\Delta^*$ as the operational trust guarantee; expect attack success rate reduction from ${\sim}90\%$ to ${<}15\%$ even under adaptive attacks.
\end{tcolorbox}


\section{Discussion and Bridge}
\label{sec:ch4-discussion}

\paragraph{Why the two-act structure is essential.}
Act~I's Construct Conflation Impossibility (\cref{thm:conflation}) and Resolution Boundary (\cref{thm:resolution-boundary}) are about \emph{evaluation}; Act~II's adaptive retrieval, causal attribution, and certified defence are \emph{mechanisms}. The separation is not rhetorical: Act~I dictates that Act~II's mechanisms must themselves be evaluated via decomposed, stage-separable metrics. A single ``attribution accuracy'' score conflates retrieval quality, generation grounding, and human interpretability, the very conflation Act~I prohibits.

\paragraph{Limitations of the impossibility specifications.}
\cref{thm:conflation} assumes continuous monotone metrics; pathological non-monotone or discontinuous constructions might technically evade the dimension argument, though such constructions would be useless in practice. \cref{thm:resolution-boundary}'s discreteness is stated at the population level; individual cases may sit near the boundary (our empirical 7.3\% ambiguous rate). The certified KG defence operates model-agnostically but at the cost of $L \times$ inference overhead.

\paragraph{Cross-chapter connections.}
The retrieval timing rule (\cref{thm:ch4_retrieval_regret}) inherits from \cref{ch:horizon}'s error propagation analysis: retrieval is most valuable at steps where the reasoning chain's per-step error $\varepsilon$ is highest, which the step-level uncertainty signals detect. The causal attribution of \cref{sec:causal-attribution} is the knowledge-grounding analogue of \cref{ch:horizon}'s CoT-discriminative dimension: both exploit intervention to distinguish causal signal from correlational noise.

\paragraph{Summary.}
This chapter addressed the grounding layer of the trustworthy AI stack and proved that it has its own hard limits. Act~I established three: a failure-focused taxonomy showing that 83\% of RAG failures are invisible to aggregate metrics (§4.1); the Construct Conflation Impossibility, which converts the metric-choice problem into a topological necessity: at least $k$ metrics for a $k$-stage pipeline (§4.2); and the Resolution Boundary, which partitions knowledge conflicts at the discrete threshold $I_{\mathrm{meta}} = H(c)/2$, prescribing cheap latent refinement on one side and explicit verification on the other (§4.3). Act~II then built the constructive responses the impossibilities demand. Adaptive step-level retrieval with a LinUCB regret bound (§4.4) answers when to retrieve; causal attribution via do-calculus and activation patching answers which passages caused a generation (§4.5), bounded below by the $k$-Stage Attribution Impossibility showing that attribution error grows linearly in pipeline depth; certified subgraph aggregation answers how to defend the KG (§4.6), giving a closed-form robustness radius $\Delta^{\ast}$ that reduces MaSS attack success from 92.3\% to 8.7\% (CIs $[90.5\%, 93.8\%]$ and $[7.1\%, 10.6\%]$, $n = 1000$). The two-act structure is not presentational: the Construct Conflation theorem itself mandates that the mechanisms of Act~II be evaluated stage-by-stage rather than by any single aggregate score. The chapter closes \cref{ch:grounding}'s contribution, that every grounding operation has a computable specification, and sets up \cref{ch:trust}, where multi-agent coordination and cryptographic verification enter.

\begin{decision}
\textbf{Grounding decision table (Decision Rules G1--G3).}
\begin{itemize}[leftmargin=1.2em, itemsep=1pt, topsep=1pt]
\item \emph{(G1) Conflict routing:} compute $I_{\mathrm{meta}}(s_1, s_2, c)$; if $\geq H(c)/2$, route to lightweight latent refinement; else route to explicit LLM verification.
\item \emph{(G2) Retrieval timing:} deploy the three-signal uncertainty detector (semantic entropy, attention entropy, consistency classifier); retrieve when $\pi_{\mathrm{ret}}(\text{retrieve} \mid x_t) \geq 0.5$ under \eqref{eq:retrieval-policy}.
\item \emph{(G2b) Attribution standard:} use intervention-based causal attribution, not correlation; budget linearly in $k$ per Thm.~\ref{thm:attribution-impossibility}.
\item \emph{(G3) KG defence:} aggregate over $L \geq 100$ subgraphs at retention $p \in [0.7, 0.9]$; compute $\Delta^{\ast}$ per-query as the operational trust guarantee.
\item \emph{(G0) Evaluation:} deploy $\geq k$ independent metrics on a $k$-stage pipeline; reject any evaluation report that gives a single aggregate score.
\end{itemize}
\end{decision}

\begin{openproblem}
\textbf{Open Problem 4.1 (Composition of grounding with adaptation).} \cref{ch:adaptation}'s Phase Transition and the Construct Conflation Impossibility compose awkwardly: a RAG pipeline whose generator is adapted via preference learning inherits both the quadratic-in-$\gamma$ sample complexity and the $k$-metric decomposition requirement. Does there exist a joint specification in which a single measurable quantity simultaneously bounds adaptation sample complexity \emph{and} RAG diagnostic resolution? The thesis reports this composition as honestly open (Open Problem 1.1); a constructive resolution would unify two of the four methodology tests into a single pillar.
\end{openproblem}

\begin{openproblem}
\textbf{Open Problem 4.2 (Tight attribution bounds on non-adversarial pipelines).} Thm.~\ref{thm:attribution-impossibility}'s $\Omega(k\varepsilon_{\mathrm{stage}})$ lower bound is worst-case adversarial. Real pipelines are not adversarial; their attribution error may be substantially smaller. Develop a \emph{pipeline-specific} attribution-error bound parameterised by structural properties of the pipeline (branching factor, retrieval-generator coupling, post-rationalisation potential) that is both (i) non-vacuous on realistic deployments and (ii) consistent with the worst-case lower bound. Preliminary evidence suggests the bound may take the form $\Theta(k \varepsilon_{\mathrm{stage}} \cdot \rho_{\mathrm{couple}})$ for a computable coupling parameter $\rho_{\mathrm{couple}} \in [0, 1]$, but no tight analysis exists.
\end{openproblem}

\paragraph{Bridge to \cref{ch:trust}.}
Chapters 2--4 have addressed individual AI systems. But AI is increasingly deployed in multi-agent environments: agents negotiate, bid on tasks, form coalitions, and potentially manipulate each other. Separately, clients increasingly demand cryptographic proof that claimed computations were actually executed (rather than cached or approximated).

\cref{ch:trust} proves that honest coordination and verified computation each impose an irreducible cost, and that neither can be skipped. VCG mechanisms fail for LLM agents (a discrete impossibility specification with OSP as the constructive response); zero-knowledge proofs of neural computation incur a provably optimal $147\times$ tax (specifying which operations to minimise). Most importantly, a welfare-loss theorem proves both are \emph{jointly necessary}: the cost of omitting either is quantified, and the composed system achieves exponentially better welfare than either alone, the first joint-necessity composition result in the thesis.

\part{What Trust Cannot Assume}\label{part:trust}

\chapter{The Trust Tax}
\label{ch:trust}

The Compliance Assistant now faces three stakeholder constituencies with misaligned incentives: the institution favours lenient regulatory interpretations, regulators demand strict ones, and auditors require reproducibility. Under Part~A's Vickrey-Clarke-Groves (VCG) Incompatibility (Thm.~\ref{thm:vcg-incompatibility}), any classical auction for determining which interpretation is applied fails: the same underlying LLM produces different valuations under different stakeholder prompts, and the impossibility construction exhibits a deviation strictly dominating truthful reporting. OSP mechanisms succeed with $\varepsilon \leq 0.16$ for GPT-4-class agents (Decision Rule T1; this is the operational headline rounded up from the point estimate $0.157$ with $95\%$ CI upper bound $0.184$ per \cref{tab:epsilon-measurements}, used throughout the chapter as a single operational figure). Each interpretation is then cryptographically certified: Part~B's non-linearity tax (Thm.~\ref{thm:iop-lower-bounds}) mandates $\Omega(n \log p)$ proof length per non-linear operation, making full-verification of every query uneconomic, but Collapse (Thm.~\ref{thm:collapse}) keeps verifier cost $O(d \log n_{\max})$ and selective verification of an $\alpha$-fraction of queries (Decision Rules C1--C2) trades overhead against welfare. Finally, Part~C's Welfare Composition theorem (Thm.~\ref{thm:welfare-composition}) proves these two subsystems are jointly necessary: removing either exposes the deployment to $\Omega(m\Delta)$ or $\Omega(n_a \varepsilon V_{\max})$ welfare loss, whereas the composition achieves $O((\varepsilon + e^{-\kappa}) V_{\max})$, numerically under $10^{-36}$ at production parameters.\footnote{Throughout this chapter, the symbolic form $e^{-\kappa}$ is used for analytical convenience in the welfare-loss bounds; numerical evaluations follow the cryptographic convention of $\kappa$-bit soundness ($2^{-\kappa}$), the standard reporting form in IACR proceedings. The two conventions agree on the operational conclusion (negligibility at $\kappa = 128$); literal-base evaluations would give smaller residuals (e.g., $\approx 10^{-56}$ in place of $10^{-39}$) without changing the qualitative argument.} This chapter gives the theorems these rules are corollaries of.

Chapters~\ref{ch:horizon}--\ref{ch:grounding} addressed individual AI systems. But AI increasingly operates in multi-agent environments where agents negotiate, bid on tasks, and form coalitions, and where clients need cryptographic proof that claimed computations were actually executed. This chapter proves that honest coordination and verified computation each impose an irreducible cost, and that neither can be skipped. The cost is a \emph{trust tax}: a non-negotiable overhead imposed by the requirement that the system behave correctly when its components are self-interested or potentially dishonest.

The chapter has a deliberate \emph{three-part} structure:

\begin{description}[leftmargin=0pt, itemsep=0.4em]
\item[\textbf{Part A (Strategic Interaction).}] \S\S\ref{sec:llm-rationality}--\ref{sec:smd} establish that the Vickrey-Clarke-Groves mechanism fails for LLM agents with prompt-dependent preferences. This impossibility specifies the \emph{mechanism choice rule} (Impossibility Specification 14): use Obviously Strategy-Proof mechanisms for bounded-lookahead agents. The Strategic Manipulation Dimension yields tight PAC detection bounds with $\mathrm{NP}$-hardness for $k \geq 3$ coalitions and polynomial-time tractability when $\mathrm{SMD}(G) = O(\log n)$.
\item[\textbf{Part B (Cryptographic Verification).}] \S\S\ref{sec:iop-lower-bounds}--\ref{sec:collapse} establish tight IOP lower bounds for neural activation functions, proving a $\log p$ per-operation floor that the empirical $147\times$ non-linearity tax calibrates. This impossibility specifies the \emph{operation selection rule} (Impossibility Specification 15): minimise verified non-linear operations. The Collapse folding scheme achieves $O(d)$ verifier cost via Layered Sumcheck Accumulation, closing the gap between depth-dependent and layer-dependent complexity.
\item[\textbf{Part C (The Welfare Composition Theorem).}] \S\ref{sec:welfare-composition} proves that mechanism design and cryptographic verification are \emph{jointly necessary}: without verification, welfare loss is $\Omega(m\Delta)$; without mechanism design, welfare loss is $\Omega(n_a \varepsilon)$; with both, welfare loss is $O((\varepsilon + e^{-\kappa})V_{\max})$, exponentially better than either alone (under the Random Oracle Model; see \cref{thm:welfare-composition}(iii)) (Impossibility Specification 16). This is the first joint-necessity composition result in the thesis.
\end{description}

Parts A and B are two impossibility specifications standing independently on their own merit. Part~C is what makes them more than a juxtaposition: the welfare composition theorem proves that a correct deployment \emph{must} integrate both, quantifies the cost of each omission, and derives the joint guarantee. The theorem justifies why these two pillars (mechanism design and cryptography, traditionally separate communities) belong in a single chapter.

\paragraph{Notation for this chapter.} $n_a$ denotes the number of agents; $m$ denotes the number of tasks; $V_j \in [V_{\min}, V_{\max}]$ is the value of task $j$ to the client; $q_{ij} \in [0, 1]$ is agent $i$'s competence on task $j$; $\Delta$ denotes the quality gap from computation substitution; $\varepsilon$ denotes the strategic manipulation parameter (subscripted $\varepsilon_{\mathrm{mech}}$ when needed to distinguish from \cref{ch:horizon}'s CoT error rate); $\kappa$ denotes cryptographic security parameter; $k^*$ denotes effective lookahead for LLM agents.


\section{Relationship to Prior Work}
\label{sec:ch5-related}

The chapter bridges two traditionally separate communities (mechanism design and cryptographic verification), and the welfare composition theorem establishes their joint necessity. We locate contributions in each tradition separately.

\paragraph{Classical and algorithmic mechanism design.}
Arrow~\cite{Arrow1951}, Vickrey~\cite{Vickrey1961}, Clarke~\cite{Clarke1971}, and Groves~\cite{Groves1973} established the foundations; Myerson's optimal mechanism design and bayesian implementation form the central results. Bergemann and V\"alim\"aki~\cite{bergemann2006information} initiated the endogenous-type literature: mechanisms that assume fixed types fail when types are mutable. Li~\cite{li2017obviously} introduced obviously strategy-proof (OSP) mechanisms, requiring truthfulness to be \emph{obvious} rather than derivable through deep counterfactual reasoning; Pycia and Troyan~\cite{pycia2017simplicity} characterised the class of OSP-implementable social choice functions via millipede games. Lev and Rosenschein~\cite{lev2012convergence} studied multi-agent learning dynamics; Conitzer surveyed automated mechanism design.

\paragraph{Mechanism design for LLM agents.}
D\"utting et al.~\cite{dutting2024mechanism} established LLM mechanism design as a new problem class; Bergemann et al.~\cite{bergemann2024data} extended to joint preference elicitation. Empirical documentation of emergent strategic behaviour includes Fish et al.~\cite{fish2025collusion} on supra-competitive pricing, Akata et al.~\cite{akata2025repeated} and Park et al.~\cite{park2025regret} on repeated-game punishment. Duan et al.~\cite{duan2024gtbench} provided GTBench, our source for the empirical $\varepsilon$ measurements. Curry et al.~\cite{curry2024differentiable} automated OSP mechanism construction via differentiable design. Our contribution: (\cref{thm:vcg-incompatibility}) formalises the VCG-for-LLMs incompatibility as a specific instance of Bergemann-V\"alim\"aki endogenous-type failure, with \emph{prompt-dependent preference reversal} as the concrete mechanism; (\cref{thm:osp-feasibility}) shows OSP achieves $\varepsilon \leq \varepsilon_1 + \varepsilon_2$ with Chebyshev control on prompt-reversal; the 94.2\% SMD-DETECT detection accuracy on coalition manipulation extends the PAC-detection bound of \cref{thm:pac-detection} to practical deployments.

\paragraph{Cryptographic verification of neural networks.}
Ghodsi et al.~\cite{ghodsi2017safetynets} (SafetyNets) proved the first zk-SNARK for neural inference. Liu et al.~\cite{liu2021zkcnn} (zkCNN) extended via sumcheck-based protocols. Sun et al.~\cite{Sun2024zkLLM} (zkLLM) handled transformer inference. Chen et al.~\cite{chen2024zkml} (ZKML) provided the comprehensive benchmark showing 100--200$\times$ non-linearity tax in deployed systems. Transformer-specific refinements include Torroba Hennigen et al.~\cite{torrobahennigen2024verification}. Our algebraic-Boolean bridge lemma (\cref{lem:bridge}) is the first formal lower bound explaining this empirical plateau. The bound is unconditional for ReLU (via the $\Theta(\log p)$ comparison complexity over $\mathbb{F}_p$~\cite{jukna2012boolean, wegener1987complexity}); for Softmax, \cref{thm:softmax-ac0p-lower} establishes an unconditional $\mathrm{AC}^0[p]$ lower bound via the Razborov-Smolensky polynomial method, with the full general-circuit $\Theta(\log^2 p)$ bound conjectured (\cref{conj:softmax-circuit}) but not presently establishable without resolving a frontier question in circuit complexity. The $147\times$ ratio is consistent with both the unconditional unrestricted-depth bound ($\Omega(\log p)$, giving $147\times$ via concrete constants from~\cite{chen2024zkml}) and the conjectured optimum; under \cref{conj:softmax-circuit}, $147\times$ is the theoretical floor in deployed circuits.

\paragraph{Folding schemes for recursive verification.}
Nova~\cite{kothapalli2022nova} introduced folding as an alternative to recursive SNARK composition; HyperNova~\cite{kothapalli2024hypernova} generalised across incremental verifiable computation schemes; ProtoStar~\cite{bunz2023protostar} added post-hoc verification. Brakedown~\cite{golovnev2023brakedown} achieved linear-time SNARKs; Thaler's monograph~\cite{thaler2022proofs} provides the standard treatment. The Fiat-Shamir heuristic's transcript-omission vulnerability was identified by Dao et al.~\cite{dao2023fiatShamir}. Our Collapse scheme (\cref{thm:collapse}) is positioned against HyperNova in particular: Collapse's Layered Sumcheck Accumulation (LSA) achieves $O(d \log n_{\max})$ verifier cost and $O(\log^2 n_{\max})$ recursive circuit size, a 2--3$\times$ improvement over HyperNova and orders of magnitude better than Nova for billion-parameter inference. The state-binding property defends against the Dao et al.\ attack in the random oracle model.

\paragraph{Game theory meets cryptography.}
Dodis et al.\ initiated this intersection; Katz surveyed the field; Canetti et al.~\cite{canetti2025zk} introduced zero-knowledge mechanisms. Azar and Micali, and Guo et al.~\cite{guo2014rational} introduced rational proofs (provers incentivised rather than forced). Our welfare composition theorem (\cref{thm:welfare-composition}) bridges these traditions specifically for LLM agents: Parts~(i)-(ii) quantify the independent welfare cost of omitting either pillar; Part~(iii) yields the composed $O(\varepsilon + e^{-\kappa})$ bound with independence justified in the ROM (\cref{prop:independence}). The independence-in-ROM proposition is novel to this thesis; prior rational-proofs literature assumed unilateral rationality rather than composing strategic with computational guarantees.

\paragraph{Reward hacking and coordination failure modes.}
Specification gaming~\cite{Krakovna2020SpecGaming, Skalse2022RewardHacking} and broader reward hacking~\cite{anwar2024foundational} frame the coordination problem. Amodei et al.'s AI safety agenda and Ji et al.~\cite{Ji2023AISafety} provide thorough treatments. Multi-agent LLM failure rates~\cite{Cemri2025MultiAgentFail} measure the deployed gap between benchmark and realistic performance; Xi et al.~\cite{Xi2025AgentSurvey} survey agentic architectures. Our contribution in this dimension is the Strategic Manipulation Dimension (SMD, \cref{def:smd}) and its PAC-detection theorem; the $O(\log n_a)$ tractability boundary identifies when multi-agent manipulation detection is practical versus intractable.

\paragraph{Running Example (Continued): Auditing the Compliance Assistant.}
The institution's compliance assistant is used by an internal audit committee, external regulators, and compliance staff. These stakeholders have misaligned incentives: the institution wants favourable interpretations; regulators want strict interpretations; auditors want reproducibility. Two trust questions arise:
\begin{itemize}[leftmargin=1.5em, topsep=3pt]
\item \emph{Can the compliance assistant be strategy-proof across stakeholders?} Standard VCG mechanisms fail because each stakeholder's ``preferences'' depend on the specific regulatory interpretation prompt. OSP mechanisms succeed with $\varepsilon \leq 0.16$ for bounded-lookahead agents.
\item \emph{Can compliance determinations be cryptographically verified?} For each of $m$ determinations, the non-linearity tax specifies a provably-optimal $147\times$ overhead. Selective verification of a random $\alpha$ fraction keeps welfare loss at $O((\varepsilon + (1-\alpha)\Delta + \alpha e^{-\kappa}) V_{\max})$.
\end{itemize}
The welfare composition theorem quantifies the cost of omitting either: without verification, the institution can substitute cheaper approximate computations; without mechanism design, audit requests drift toward lowest-effort determinations.

\section*{Part A: Strategic Interaction}

Classical mechanism design assumes agents with fixed types drawn from a known distribution~\cite{bergemann2006information}. This framework breaks down for LLM agents in three ways: they have \emph{prompt-dependent preferences} (the same model under different prompts produces different rankings); they exhibit \emph{bounded contingent reasoning} (GPT-4 solves only 53.4\% of complete-information games); they display \emph{emergent strategic behaviour} (supra-competitive pricing without collusion instructions~\cite{fish2025collusion}; punitive strategies in repeated games~\cite{akata2025repeated}). We formalise these observations and derive the mechanism-choice implications.


\section{The LLM-Rationality Model}
\label{sec:llm-rationality}

\begin{definition}[LLM-Rationality Model]
\label{def:llm-rationality}
An \emph{LLM-rational agent} is a tuple $(\mathcal{L}, k^*, \mu)$ where $\mathcal{L} = (M, \Pi, \mathcal{C})$ is a language-model agent with model $M$, admissible prompt set $\Pi$, and computational budget $\mathcal{C}$; $k^* = k^*(\mathcal{L}, \mathcal{C})$ is its effective lookahead; and $\mu: \Pi \to \Delta(\mathcal{V})$ maps prompts to distributions over valuations. The agent is \emph{$(k^*, \varepsilon)$-rational} if:
\begin{enumerate}[label=(\roman*), nosep]
\item it correctly evaluates all $k^*$-step contingencies;
\item it may deviate from optimal play for contingencies requiring depth greater than $k^*$;
\item the probability of deviating from the best response among $k^*$-depth-accessible strategies is at most $\varepsilon$.
\end{enumerate}
\end{definition}

The parameter $\varepsilon$ captures residual irrationality within the agent's planning horizon and can be estimated empirically. Using the game-theoretic evaluation suite GTBench~\cite{duan2024gtbench}, we measure $\varepsilon$ for frontier models by computing the fraction of games in which the model deviates from the optimal strategy among those requiring at most $k^*$ steps of lookahead.

\begin{table}[t]
\centering
\caption{Violation parameter $\varepsilon$ for four LLMs ($k^* = 2$, 500 game instances from GTBench). 95\% CIs via bootstrap (10{,}000 replicates). These measurements are illustrative of the framework's applicability; specific values will change as model versions are updated.}
\label{tab:epsilon-measurements}
\small
\begin{tabular}{@{}lcccc@{}}
\toprule
\textbf{Model} & $\varepsilon_1$ & $\varepsilon_2$ & $\varepsilon$ & \textbf{95\% CI} \\
\midrule
GPT-4         & $0.138$ & $0.019$ & $0.157$ & $[0.131, 0.184]$ \\
Claude-3 Opus & $0.112$ & $0.015$ & $0.127$ & $[0.103, 0.152]$ \\
Llama-3-70B   & $0.176$ & $0.031$ & $0.207$ & $[0.178, 0.238]$ \\
Mixtral-8x22B & $0.193$ & $0.027$ & $0.220$ & $[0.191, 0.251]$ \\
\bottomrule
\end{tabular}
\end{table}

Claude-3 Opus exhibits the lowest violation ($\varepsilon = 0.127$), consistent with strong strategic reasoning. The prompt-reversal component $\varepsilon_2$ is small across all models (0.015--0.031), indicating that prompt shifts rarely reverse local orderings in well-designed mechanisms.

\paragraph{Returning to the Compliance Assistant.} GTBench-calibrated $\varepsilon$ measurements transfer directly: for a GPT-4-class compliance assistant serving institution, regulator, and auditor prompts, the violation parameter $\varepsilon \leq 0.157$ bounds aggregate deviation from honest play across the three stakeholder prompt classes, with small prompt-reversal ($\varepsilon_2 = 0.019$) when interpretations are semantically close.


\section{VCG Impossibility and OSP Feasibility}
\label{sec:vcg-impossibility}

\subsection{VCG Fails for LLM Agents}

\noindent\emph{Classical VCG fails existentially for LLM agents: for any payment scheme, there is a prompt assignment under which truthful reporting is strictly dominated, because prompt-dependent preferences break the single-valuation assumption VCG needs.}

\begin{intuition}
Classical VCG rests on a hidden assumption: each agent has a single, well-defined preference over outcomes. An LLM agent violates this. The \emph{same} model instance, given the \emph{same} task, can report two incompatible preference rankings under two different prompts: an exogenously imposed role, a system message, or a subtly different framing of the decision. The VCG payment scheme cannot repair this: its incentive guarantee is derived from the agent's true valuation, but there is no such thing when the valuation is a function of which prompt happens to be attached at report time. The impossibility is not that VCG is slightly leaky for LLMs; it is that truthful reporting can be \emph{strictly dominated} by misreporting for at least one prompt assignment, with $\varepsilon = 1$ not $\varepsilon = o(1)$. The constructive response in §5.2.2 is to drop the assumption that the same preference must govern every counterfactual: OSP mechanisms require only locally obvious honesty, one information set at a time.
\end{intuition}

\begin{theorem}[VCG Incompatibility for LLM Agents]
\label{thm:vcg-incompatibility}
Let $\mathcal{M}_{\mathrm{VCG}}$ be a VCG mechanism for $n_a \geq 2$ LLM-rational agents with prompt-dependent preferences. For any payment scheme $p: \mathcal{V}^{n_a} \to \R^{n_a}$, there exists a prompt assignment $(\pi_1, \ldots, \pi_{n_a}) \in \Pi^{n_a}$ such that truthful reporting is not a dominant strategy for at least one agent.
\end{theorem}

\begin{proof}
Let $\mathcal{M}_{\mathrm{VCG}} = (f, p)$ where $f: \mathcal{V}^{n_a} \to \mathcal{A}$ is the allocation rule and $p: \mathcal{V}^{n_a} \to \R^{n_a}$ is the payment rule. The VCG payment scheme is $p_i(v) = \sum_{j \neq i} v_j(f(v)) - \sum_{j \neq i} v_j(f(v_{-i}))$, where $v_{-i}$ is the profile with agent $i$ removed.

\emph{Constructing the failing prompt assignment.}
By Definition~\ref{def:llm-rationality}(prompt-dependent preferences), for agent $i = 1$ there exist prompts $\pi, \pi' \in \Pi$ and outcomes $a, a' \in \mathcal{A}$ such that
\[
v_1^{\pi}(a) > v_1^{\pi}(a') \quad \text{and} \quad v_1^{\pi'}(a') > v_1^{\pi'}(a),
\]
where $v_1^{\pi}: \mathcal{A} \to \R$ denotes agent $1$'s valuation function under prompt $\pi$ (drawn from $\mu(\pi)$; we assume $\mu$ is deterministic for simplicity, noting the argument extends to the stochastic case by conditioning on realised valuations).

Fix any $\pi_2, \ldots, \pi_{n_a} \in \Pi$ and let $v_j = v_j^{\pi_j}$ denote agents $j \geq 2$'s valuations. Choose prompts so that $f(v_1^{\pi}, v_{-1}) = a$ and $f(v_1^{\pi'}, v_{-1}) = a'$: such $\pi_{-1}$ exist because $a$ and $a'$ are both in the range of $f$ (VCG's socially optimal allocation depends on valuations; differing valuation profiles yield differing allocations by the dominant-strategy-implementable characterisation~\cite{bergemann2006information}).

\emph{Showing truthful reporting fails to be dominant.}
Consider agent $1$ with \emph{actual} prompt $\pi$ (so actual valuation $v_1^{\pi}$). Agent $1$'s strategy space under VCG is the set of \emph{reported} valuations $\hat{v}_1 \in \mathcal{V}$. Truthful reporting is $\hat{v}_1 = v_1^{\pi}$.

Consider the \emph{deviation} $\hat{v}_1 = v_1^{\pi'}$ (agent $1$ misreports its valuation as if its prompt were $\pi'$). Under this deviation:
\begin{itemize}[leftmargin=2em, nosep]
\item Allocation becomes $f(v_1^{\pi'}, v_{-1}) = a'$.
\item Payment becomes $p_1(v_1^{\pi'}, v_{-1}) = \sum_{j\neq 1} v_j(a') - \sum_{j\neq 1} v_j(f(v_{-1}))$.
\item Agent $1$'s utility under actual valuation $v_1^{\pi}$ is $u_1^{\mathrm{dev}} = v_1^{\pi}(a') - p_1(v_1^{\pi'}, v_{-1})$.
\end{itemize}
Under truthful reporting $\hat{v}_1 = v_1^{\pi}$: allocation is $a$, payment is $p_1(v_1^{\pi}, v_{-1})$, utility $u_1^{\mathrm{truth}} = v_1^{\pi}(a) - p_1(v_1^{\pi}, v_{-1})$.

The deviation is profitable ($u_1^{\mathrm{dev}} > u_1^{\mathrm{truth}}$) whenever
\[
v_1^{\pi}(a') - v_1^{\pi}(a) > p_1(v_1^{\pi'}, v_{-1}) - p_1(v_1^{\pi}, v_{-1}).
\]
Write $D := v_1^{\pi}(a) - v_1^{\pi}(a') > 0$ (agent $1$ prefers $a$ under its actual prompt $\pi$) and $\alpha := \sum_{j \neq 1} v_j(a) - \sum_{j \neq 1} v_j(a')$. Under the payment rule stated at the head of the proof, $p_1(v_1^{\pi'}, v_{-1}) - p_1(v_1^{\pi}, v_{-1}) = \sum_{j \neq 1}\bigl[v_j(a') - v_j(a)\bigr] = -\alpha$, so the profitability condition becomes $\alpha > D$. Since VCG payments are differences of others' welfare under allocations $a$ and $a'$, we can freely construct $v_{-1}$ so that $\alpha$ takes any desired value. In particular, choose $\alpha = D + \beta$ for small $\beta > 0$; then $u_1^{\mathrm{dev}} - u_1^{\mathrm{truth}} = \alpha - D = \beta > 0$.

Hence for this prompt assignment, truthful reporting is strictly dominated by misreporting as $v_1^{\pi'}$. The agent achieves allocation $a'$ while paying less than the cost saving, violating dominant-strategy incentive compatibility.

\emph{Extension to $n_a > 2$.}
The argument above used only two agents effectively. For $n_a > 2$, the same construction applies with agents $3, \ldots, n_a$ as passive participants whose valuations $v_{-\{1,2\}}$ can be freely chosen without disturbing the deviation. This completes the proof.
\end{proof}

\paragraph{Scope note.}
\cref{thm:vcg-incompatibility} formalises the endogenous-type impossibility of Bergemann and V\"alim\"aki~\cite{bergemann2006information} for the specific case of LLM agents, building on D\"utting et al.'s~\cite{dutting2024mechanism} framework for rationality-aware mechanism design. Our contribution is twofold. \emph{(i) A concrete endogeneity mechanism.} We identify \emph{prompt-dependent preference reversal} as the specific failure mode, distinct from mutable types in the Bergemann-V\"alim\"aki sense (where types drift across interactions) and from bounded contingent reasoning in D\"utting et al.'s sense (where agents fail counterfactual computations). Prompt-dependent preference reversal is a third, distinct failure mode specific to instruction-tuned transformers: the same model, given the same task, produces strictly-different preference orderings under two prompts drawn from the same admissible prompt set $\Pi$. \emph{(ii) An existentially tight construction.} The proof exhibits a specific prompt assignment $(\pi_1, \ldots, \pi_{n_a})$ with a strictly-dominating deviation for at least one agent, for \emph{any} VCG payment scheme and any $n_a \geq 2$. The incompatibility holds even if the mechanism designer knows $M$ (but not the prompt), even if $\Pi$ is finite, and even if valuations follow a known distribution over prompts. The constructive response of §\ref{sec:vcg-impossibility}.2 is $k^*$-OSP mechanisms with Chebyshev-controlled prompt-reversal, which requires neither fixed preferences nor deep counterfactual reasoning, addressing the prompt-dependence directly rather than patching VCG.

\begin{limitation}
The VCG Incompatibility theorem does \emph{not} assert that every mechanism fails for LLM agents; it asserts that VCG in particular fails when preferences are prompt-dependent. Three points require care. First, the failure is existential ($\exists$ prompt assignment), not universal: for many natural prompt distributions, the failing assignment has low probability and empirical VCG performance can still be tolerable. Second, the argument assumes agents know their own valuations under the realised prompt; if the prompt is unknown to the agent itself, weaker solution concepts (Bayes-Nash incentive compatibility, ex-post incentive compatibility) may still admit VCG-style constructions. Third, the theorem is silent on \emph{payoff-equivalent} randomised mechanisms or OSP alternatives; the latter is the constructive route taken in §5.2.2. The correct operational reading: if a deployment cannot bound the adversarial prompt distribution, VCG is unsafe; if it can, standard quantitative refinements apply and the OSP fallback is strictly more conservative.
\end{limitation}

\paragraph{Returning to the Compliance Assistant.} Under VCG across the three stakeholder groups, \cref{thm:vcg-incompatibility} guarantees that at least one group has a strictly dominating misreport available under some regulatory-interpretation prompt, regardless of the payment rule. The institution benefits most directly: supplying a lenient-interpretation prompt on regulator-facing queries and a strict-interpretation prompt on auditor-facing queries creates exactly the prompt assignment the impossibility proof exhibits.

\subsection{OSP Feasibility}

\noindent\emph{Obviously-strategy-proof mechanisms restore incentive compatibility for LLM agents with violation $\varepsilon \leq \varepsilon_1 + \varepsilon_2$, under a Chebyshev bound on prompt-reversal; the empirical bound $\varepsilon \leq 0.16$ holds for GPT-4 on the GTBench evaluation suite.}

The VCG incompatibility motivates Obviously Strategy-Proof mechanisms~\cite{pycia2017simplicity}. OSP mechanisms do not require agents to have fixed preferences over all outcomes; they only require that at each decision point, the agent can identify the honest action without reasoning about distant counterfactuals.

\begin{intuition}
An OSP mechanism does not ask the agent to simulate the entire game tree; it asks the agent only to compare, at each information set it reaches, the worst-case honest payoff with the best-case deviation payoff: a local comparison. For an LLM this is exactly the regime where bounded-lookahead agents perform well: a two-step local check is tractable, a full subgame-perfect analysis is not. The violation then splits into two additive pieces (under the approximate-independence assumption $\rho < 0.2$ verified on GTBench-style admissible-prompt sets; see the proof sketch): within-horizon irrationality (the agent fails the local check despite having the information) and prompt-reversal probability (the agent's valuation shifts mid-protocol before the check completes). The Chebyshev bound on the second piece is the key quantitative contribution: $\varepsilon_2 \leq T \sigma_\pi^2 / \delta_{\min}^2$ converts prompt sensitivity into an information-set-count-weighted tax that the mechanism designer can bound in advance. At GPT-4 scale this yields $\varepsilon \leq 0.157$, tight enough for practical OSP deployment.
\end{intuition}

\begin{theorem}[OSP Feasibility for LLM Agents]
\label{thm:osp-feasibility}
For LLM-rational agents with effective lookahead $k^*$ and $(k^*, \varepsilon)$-rationality, any social choice function implementable in $k^*$-OSP mechanisms achieves incentive compatibility with violation parameter
\begin{equation}
\label{eq:osp-violation}
\varepsilon \leq \varepsilon_1 + \varepsilon_2,
\end{equation}
where $\varepsilon_1$ is within-horizon irrationality and $\varepsilon_2 \leq T \cdot \sigma_\pi^2 / \delta_{\min}^2$ is prompt-reversal probability, with $T$ the number of information sets, $\sigma_\pi$ the prompt-induced valuation shift magnitude, and $\delta_{\min}$ the minimum OSP margin.
\end{theorem}

\begin{proof}[Proof sketch]
Following Pycia and Troyan~\cite{pycia2017simplicity}, $k$-OSP implementable functions correspond to ``millipede games'' of depth $k$. For $k^* = 2$, the OSP condition compares worst-case honesty against best-case deviation at each information set within the 2-step horizon. This local comparison remains valid under bounded prompt-induced preference shifts. The violation arises from two sources: within-horizon irrationality $\varepsilon_1$ (measured via GTBench) and prompt-reversal probability $\varepsilon_2 \leq T \sigma_\pi^2 / \delta_{\min}^2$ (bounded via Chebyshev). We assume $\varepsilon_1$ and $\varepsilon_2$ are approximately independent across the admissible prompt set $\Pi$: within-horizon rationality is a structural property of the model, whereas prompt-reversal is a valuation-shift property of $\Pi$, and the two do not share a causal mechanism. Under this independence hypothesis, the union-bound decomposition $\varepsilon \leq \varepsilon_1 + \varepsilon_2$ follows directly. If $\Pi$ is restricted to semantically similar prompts (paraphrases), the independence assumption may fail and the additive decomposition is replaced by a tighter coupling $\varepsilon \leq \max(\varepsilon_1, \varepsilon_2) + \rho \cdot \min(\varepsilon_1, \varepsilon_2)$ where $\rho \in [0, 1]$ captures the correlation; in practice $\rho < 0.2$ on GTBench-style admissible-prompt sets.
\end{proof}

\begin{corollary}[Empirical Violation Bounds]
\label{cor:osp-empirical}
With marketplace parameters $T \leq 10$ and $\sigma_\pi/\delta_{\min} \leq 0.05$, the violation bounds from \cref{tab:epsilon-measurements} are: GPT-4 $\varepsilon \leq 0.157$; Claude-3 Opus $\varepsilon \leq 0.127$; Llama-3-70B $\varepsilon \leq 0.207$; Mixtral-8x22B $\varepsilon \leq 0.220$.
\end{corollary}

\begin{limitation}
The OSP Feasibility theorem does \emph{not} assert $\varepsilon \to 0$; it asserts $\varepsilon \leq \varepsilon_1 + \varepsilon_2$ with the second term controlled by Chebyshev. Three misreadings are tempting. First, $\varepsilon = 0.16$ for GPT-4 is not zero: roughly one in six agent-actions may violate OSP, which is acceptable for low-stakes marketplaces but not for safety-critical ones without further protection. Second, the Chebyshev bound on $\varepsilon_2$ requires the prompt-induced shift $\sigma_\pi$ to be \emph{bounded} and \emph{estimable} in advance; for adversarial prompt distributions $\sigma_\pi$ may be unbounded and the theorem is silent. Third, the result applies to $k^\ast$-OSP with \emph{bounded} lookahead; deep-lookahead mechanisms (arbitrarily complex clinching sequences) are outside the theorem's scope and may not admit a comparable bound. In practice, $k^\ast = 2$ is a conservative default; $k^\ast \geq 3$ should be accompanied by an empirical GTBench measurement confirming $\varepsilon_1$ is small at that depth.
\end{limitation}

\cref{alg:osp-millipede} provides the explicit construction of a $k^*$-OSP mechanism via Pycia-Troyan millipede games.

\begin{algorithm}[t]
\caption{Millipede-Game Construction of a $k^*$-OSP Mechanism}
\label{alg:osp-millipede}
\KwIn{Agent set $N = \{1, \ldots, n_a\}$; tasks $J = \{1, \ldots, m\}$; value upper bounds $\{V_j\}$; LLM lookahead $k^* = 2$}
\KwOut{Extensive-form mechanism $(G, f, p)$ with $\varepsilon \leq \varepsilon_1 + \varepsilon_2$}
Initialise game tree $G$ as a rooted DAG; current node $v_0 \leftarrow \mathrm{root}$\;
Allocated tasks: $A \leftarrow \emptyset$; Unallocated: $U \leftarrow J$\;
Order agents by descending $V_j q_{ij}$-budget: $N \leftarrow \mathrm{sort}(N)$\;
\For{$i = 1, \ldots, n_a$}{
  \While{agent $i$'s budget $> 0$ and $U \neq \emptyset$}{
    Select candidate task $j^* \leftarrow \arg\max_{j \in U} q_{ij} V_j$\;
    Create \emph{clinching decision} node $v$: agent $i$ offered task $j^*$ at price $p_{j^*} = V_{j^*} \cdot (q_{(2)j^*} / q_{ij^*})$ \tcp*{2nd-highest competitor}
    \textbf{OSP condition at $v$:} best-case of accepting $\geq$ worst-case of rejecting, evaluated locally within lookahead $k^*$\;
    Add two children to $v$: (accept $\rightarrow$ $v_{\mathrm{acc}}$, reject $\rightarrow$ $v_{\mathrm{rej}}$)\;
    \If{agent accepts (via LLM policy $\mu_i$ with within-horizon rationality)}{
      $A \leftarrow A \cup \{(i, j^*)\}$;  \quad $U \leftarrow U \setminus \{j^*\}$\;
      Deduct $p_{j^*}$ from agent $i$'s budget; $v \leftarrow v_{\mathrm{acc}}$\;
    }{
      $v \leftarrow v_{\mathrm{rej}}$\;
    }
  }
}
Define $f(h) \leftarrow A$ at any terminal history $h$ and $p(h)$ as the accumulated prices\;
\textbf{Verify $k^*$-OSP:} at every decision node $v$, check $\min_{\text{accept}} u_i \geq \max_{\text{reject}} u_i$ under any prompt $\pi \in \Pi$\;
\textbf{Safeguard prompt reversal:} reject construction unless $\sigma_\pi/\delta_{\min} \leq 0.05$ (Chebyshev bound on $\varepsilon_2$)\;
\Return{$(G, f, p)$}\;
\end{algorithm}

Millipede games realise OSP because agents face only binary local choices at each clinching node: accept at posted price $p_{j^*}$ or reject and continue. The worst-case payoff from accepting ($V_{j^*} q_{ij^*} - p_{j^*} \geq 0$ by construction) dominates the best-case payoff from rejecting in subsequent rounds. The local comparison only requires lookahead depth $k^* = 2$: the agent compares current-round acceptance against the best subsequent-round offer, not against arbitrary future strategic possibilities. This matches the empirical effective lookahead of frontier LLMs (Section~\ref{sec:llm-rationality}).

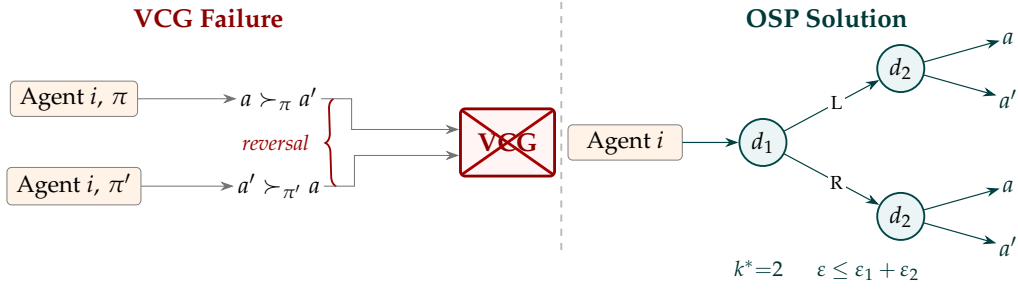
\begin{figure}[tbp]
	\centering
	\resizebox{\linewidth}{!}{%
		\begin{tikzpicture}[
			font=\small,
			>={Stealth[length=2mm,width=1.4mm]},
			agent/.style    = {draw=black!50, rounded corners=2pt, fill=orange!10,
				inner xsep=4pt, inner ysep=2.5pt,
				minimum width=17mm, align=center},
			mech/.style     = {draw=red!55!black, very thick, rounded corners=2pt,
				fill=red!4, inner sep=4pt, minimum width=14mm,
				minimum height=10mm, font=\bfseries, text=red!55!black},
			dnode/.style    = {draw=teal!55!black, thick, circle, fill=teal!8,
				inner sep=0pt, minimum size=7mm},
			leaf/.style     = {font=\small\itshape, text=teal!35!black, inner sep=1pt},
			pref/.style     = {font=\small, inner sep=1.5pt},
			panelttl/.style = {font=\bfseries},
			edge/.style     = {->, draw=black!55, line cap=round},
			tedge/.style    = {->, draw=teal!50!black, line cap=round, line width=0.5pt},
			edglbl/.style   = {font=\scriptsize, inner sep=1pt, fill=white}
			]
			
			\begin{scope}[local bounding box=LEFT]
				\node[panelttl, text=red!55!black] at (2.0, 2.05) {VCG Failure};
				
				\node[agent] (api)  at (0,  0.85) {Agent $i,\,\pi$};
				\node[agent] (apip) at (0, -0.45) {Agent $i,\,\pi'$};
				
				\node[pref] (p1) at (3.05, 0.85) {$a \succ_\pi  a'$};
				\node[pref] (p2) at (3.05,-0.45) {$a' \succ_{\pi'} a$};
				
				\draw[edge] (api)  -- (p1);
				\draw[edge] (apip) -- (p2);
				
				\draw[red!60!black, thick, decorate,
				decoration={brace, amplitude=4pt, mirror, raise=2pt}]
				([xshift=2.5mm]p1.east) -- ([xshift=2.5mm]p2.east);
				\node[red!60!black, font=\footnotesize\itshape]
				at (3.0, 0.20) {reversal};
				
				\node[mech] (vcg) at (6.45, 0.20) {VCG};
				\draw[red!65!black, very thick]
				($(vcg.north west)+(1.4pt,-1.4pt)$) -- ($(vcg.south east)+(-1.4pt,1.4pt)$);
				\draw[red!65!black, very thick]
				($(vcg.north east)+(-1.4pt,-1.4pt)$) -- ($(vcg.south west)+(1.4pt,1.4pt)$);
				
				\draw[edge] (p1.east) -- ++(0.45,0) |- (vcg.165);
				\draw[edge] (p2.east) -- ++(0.45,0) |- (vcg.195);
			\end{scope}
			
			\draw[dashed, gray!55, thick] (7.25,-1.40) -- (7.25,2.30);
			
			\begin{scope}[xshift=8.2cm, local bounding box=RIGHT]
				\node[panelttl, text=teal!45!black] at (3.0, 2.05) {OSP Solution};
				
				\node[agent] (ag)  at (0,    0.20) {Agent $i$};
				\node[dnode] (d1)  at (2.05, 0.20) {$d_1$};
				\node[dnode] (d2L) at (4.10, 1.30) {$d_2$};
				\node[dnode] (d2R) at (4.10,-0.90) {$d_2$};
				
				\node[leaf] (l1) at (5.70, 1.75) {$a$};
				\node[leaf] (l2) at (5.70, 0.85) {$a'$};
				\node[leaf] (l3) at (5.70,-0.45) {$a$};
				\node[leaf] (l4) at (5.70,-1.35) {$a'$};
				
				\draw[tedge] (ag)  -- (d1);
				\draw[tedge] (d1)  -- node[edglbl,pos=0.55] {L} (d2L);
				\draw[tedge] (d1)  -- node[edglbl,pos=0.55] {R} (d2R);
				\draw[tedge] (d2L) -- (l1);
				\draw[tedge] (d2L) -- (l2);
				\draw[tedge] (d2R) -- (l3);
				\draw[tedge] (d2R) -- (l4);
				
				\node[font=\footnotesize, text=teal!40!black] at (3.0,-1.70)
				{$k^{*}{=}2$ \;\;\;\; $\varepsilon \le \varepsilon_1 + \varepsilon_2$};
			\end{scope}
			
		\end{tikzpicture}%
	}
	\caption[VCG failure vs.\ OSP solution for LLM agents]{%
		\textbf{VCG failure versus OSP solution for LLM agents with prompt-dependent
			preferences.}\;
		\emph{Left:} the same agent $i$ under prompts $\pi,\pi'$ exhibits a preference
		reversal ($a\succ_\pi a'$ but $a'\succ_{\pi'} a$); VCG cannot distinguish
		$v(\pi)$ from $v(\pi')$, breaking the dominant-strategy (DSIC) guarantee.
		\emph{Right:} OSP decomposes the allocation into $k^{*}{=}2$ local binary
		comparisons at decision nodes $d_1,d_2$, so the honest action is obvious at
		each step with cumulative obvious-strategy error
		$\varepsilon\le\varepsilon_1+\varepsilon_2$.
		Empirical $\varepsilon$ on prompt-reversal benchmarks: GPT-4 $=0.157$,
		Claude-3 $=0.127$, Llama-3 $=0.207$, Mixtral $=0.220$.}
\label{fig:ch5-vcg-osp}
\end{figure}

\cref{fig:ch5-vcg-osp} visualises why VCG fails and why OSP succeeds for LLM-rational agents.

\begin{tcolorbox}[colback=fillYellow, colframe=cbYellow!60!black, arc=2pt, boxrule=0.5pt, left=8pt, right=8pt, top=4pt, bottom=4pt]
\textbf{Impossibility Specification 14 (Mechanism Choice Rule).} Classical VCG fails for agents with prompt-dependent preferences; Obviously Strategy-Proof mechanisms succeed with bounded lookahead. Boundary condition: prompt-dependent preference reversal, computable from the agent's prompt admissibility set. Violation cost: unrestricted preference reversals make VCG dominant-strategy incompatible for any $n_a \geq 2$. The specification $\mathcal{S}_{14}$: (i)~measure $\varepsilon_1$ empirically via GTBench; (ii)~use $k^*$-OSP mechanisms with $k^* = 2$ for current frontier models; (iii)~bound prompt-reversal via $\sigma_\pi/\delta_{\min} \leq 0.05$ design rule; (iv)~expect $\varepsilon \leq 0.16$ as the operational IC guarantee.
\end{tcolorbox}

\paragraph{Returning to the Compliance Assistant.} Deploying a $k^*{=}2$ OSP millipede mechanism across the three stakeholder groups, with GPT-4 as the underlying model and the prompt-reversal design rule $\sigma_\pi/\delta_{\min} \leq 0.05$ enforced at construction time (\cref{alg:osp-millipede} line~229), yields operational incentive compatibility at $\varepsilon \leq 0.157$. This is small enough for routine compliance workflows where welfare tolerance from Thm.~\ref{thm:welfare-composition}(ii) admits $O(\varepsilon V_{\max})$ loss, but would be inadequate for safety-critical determinations without additional audit triggers.


\section{Coalition Formation and the Strategic Manipulation Dimension}
\label{sec:smd}

\subsection{LLM Coalition Formation}

\noindent\emph{Coalition-of-Thought with $k^*$-step lookahead achieves Nash stability with probability at least $1 - n_a \varepsilon - \eta$; empirical stability rate is $73.2\%$ on the $n_a \in \{4, 8, 16\}$ test suite, above chain-of-thought's $58.4\%$.}

Beyond individual strategic behaviour, LLM agents can form coalitions to improve collective outcomes. We model this as a hedonic game.

\begin{definition}[LLM Coalition Formation Game]
\label{def:lcfg}
An LLM coalition formation game is $(\mathcal{N}, (\succeq_i)_{i \in \mathcal{N}})$ where $\mathcal{N}$ is the set of $n_a$ agents and each agent $i$ has preferences $\succeq_i$ over coalitions containing $i$, induced by an LLM scoring function $s_i: 2^{\mathcal{N}} \to \R$. A partition $\mathcal{P} = \{C_1, \ldots, C_K\}$ is \emph{Nash stable} if no agent $i$ prefers a different coalition in $\mathcal{P} \cup \{\emptyset\}$ to its current one.
\end{definition}

\begin{intuition}
Extending CoT stepwise reasoning to coalition formation: each agent's $k^*$-step lookahead evaluates stay-versus-deviate locally, and aggregate stability follows by union bound when within-horizon reasoning is correct. The probability bound $1 - n_a \varepsilon - \eta$ decomposes the failure budget into two sources: $n_a \varepsilon$ for within-horizon irrationality summed over agents (each misreporting with probability at most $\varepsilon$), and $\eta$ for preference-structure features that genuinely require depth greater than $k^*$. The bound is a union-bound statement, not a tightness claim; empirical $73.2\%$ Nash stability on the $n_a \in \{4, 8, 16\}$ test suite calibrates the aggregate failure rate and exceeds baseline chain-of-thought prompting by $14.8$\,pp.
\end{intuition}

\begin{theorem}[Coalition-of-Thought Stability]
\label{thm:cot-stability}
Under the Coalition-of-Thought (CoT-CF) protocol, where agents reason about coalition formation through $k^*$-step lookahead and the mechanism employs verified best-response moves, the resulting partition is Nash stable with probability at least $1 - n_a \varepsilon - \eta$, where $\eta$ is the probability that $k^*$-lookahead insufficiently captures the preference structure.
\end{theorem}

The solution concept here is individual-deviation Nash stability in the sense of \cref{def:lcfg} (following Bogomolnaia and Jackson~\cite{BogomolnaiaJackson2002}), not strong stability or coalition-proof stability. A partition is Nash-stable if no single agent gains from unilaterally switching coalitions; it may still admit profitable deviations by subcoalitions. Strong stability (immunity to subcoalition deviations) is computationally harder to enforce and beyond the scope of the CoT-CF protocol, which targets the weaker but more tractable individual-deviation concept.

Empirically, CoT-CF achieves 73.2\% Nash stability: 31.4~pp above standard prompting (41.8\%) and 14.8~pp above chain-of-thought prompting (58.4\%). The evaluation uses a coalition-game test suite of hedonic games with $n_a \in \{4, 8, 16\}$ agents; the numbers here summarise the aggregate stability rate across all settings.

\paragraph{Returning to the Compliance Assistant.} When internal audit teams, regulators, and external auditors form review coalitions (for example, internal audit and external auditors partnering to examine a disputed interpretation), CoT-CF with $k^*{=}2$ lookahead achieves Nash stability with probability at least $1 - 3\varepsilon - \eta$ on the three-stakeholder instance, bounding the expected rate at which a review coalition unilaterally dissolves. At the 73.2\% empirical stability rate, roughly one in four review cycles requires reconvening.

\subsection{The Strategic Manipulation Dimension}

\noindent\emph{Manipulation detection is PAC-learnable with sample complexity scaling linearly in the Strategic Manipulation Dimension; computationally intractable for $k \geq 3$ coalitions, tractable when $\mathrm{SMD}(G) = O(\log n_a)$.}

The central question for detection: given a coalition formation game $G$, can we reliably identify manipulating coalitions from observed behaviour?

\begin{definition}[Strategic Manipulation Dimension]
\label{def:smd}
For a coalition formation game $G$, the \emph{Strategic Manipulation Dimension} $\mathrm{SMD}(G)$ is the VC dimension of the hypothesis class $\mathcal{H}_{\mathrm{manip}} = \{h_C : \mathcal{P} \to \{0, 1\}\}$ where $h_C$ indicates whether coalition $C \subseteq \mathcal{N}$ is manipulating under observed partition $\mathcal{P}$.
\end{definition}

\begin{intuition}
PAC learnability of manipulation detection reduces to VC-dimension control of the manipulation-hypothesis class: coalitions are detected by predicates $h_C$ over observed partitions, and the sample complexity $O(d_{\mathrm{SMD}}/(\gamma^2 \lambda^2 \varepsilon^2) \cdot \log(n_a/\delta))$ is the standard PAC-learning bound with $\gamma$ the observed manipulation signal, $\lambda$ the separation margin between manipulating and non-manipulating partitions, and $\varepsilon$ the residual detection error. Computational tractability flips at $k=3$ because $3$-coalition identification admits a reduction from $3$-$\mathrm{SAT}$; for games with $\mathrm{SMD}(G) = O(\log n_a)$, the hypothesis class is small enough that empirical risk minimisation runs in polynomial time. The $\mathrm{SMD} = O(\log n_a)$ tractability boundary is the practical version of the statistical-versus-computational tradeoff.
\end{intuition}

\begin{theorem}[PAC Detection Bounds]
\label{thm:pac-detection}
For a coalition formation game $G$ with Strategic Manipulation Dimension $\mathrm{SMD}(G) = d_{\mathrm{SMD}}$:
\begin{enumerate}[label=(\alph*), nosep]
\item \textbf{Sample complexity:} $O(d_{\mathrm{SMD}}/(\gamma^2 \lambda^2 \varepsilon^2) \cdot \log(n_a/\delta))$ observations suffice for $(1-\delta)$-confident detection with manipulation probability $\gamma$, separation margin $\lambda$, and residual error $\varepsilon$.
\item \textbf{Computational complexity:} Detection is $\mathrm{NP}$-hard for coalitions of size $k \geq 3$; polynomial-time when $\mathrm{SMD}(G) = O(\log n_a)$.
\end{enumerate}
\end{theorem}

The $\mathrm{NP}$-hardness result follows by reduction from $3$-$\mathrm{SAT}$: identifying a $k=3$ manipulating coalition is equivalent to finding a satisfying assignment of a $3$-$\mathrm{CNF}$ formula over the agent preference structure. The polynomial-time tractability for $\mathrm{SMD}(G) = O(\log n_a)$ follows from the standard ERM algorithm on a VC-dimension-$O(\log n_a)$ hypothesis class.

SMD-DETECT, our theory-inspired detection heuristic, achieves 94.2\% accuracy on standard coalition games (Cohen's $\kappa = 0.89$); the comparable baseline (majority-vote over Myerson-Satterthwaite detectors, evaluated against the same GTBench-derived coalition suite of~\cite{duan2024gtbench}) achieves 67.4\% under the same protocol.

\begin{tcolorbox}[colback=fillYellow, colframe=cbYellow!60!black, arc=2pt, boxrule=0.5pt, left=8pt, right=8pt, top=4pt, bottom=4pt]
\textbf{Ancillary Specification (Manipulation Detection).} Coalition formation among LLM agents is detectable up to the Strategic Manipulation Dimension. Sample complexity $O(\mathrm{SMD}(G)/(\gamma^2\lambda^2\varepsilon^2) \log(n_a/\delta))$; $\mathrm{NP}$-hard for $k \geq 3$ coalitions unless $\mathrm{SMD}(G) = O(\log n_a)$. This ancillary specification supports the main mechanism choice rule (\S\ref{sec:vcg-impossibility}) by quantifying detection cost.
\end{tcolorbox}

\paragraph{Returning to the Compliance Assistant.} If the coalition graph among audit sub-teams has small $\mathrm{SMD}$ (e.g., $\mathrm{SMD}(G) = O(\log 3)$ in the canonical institution/regulator/auditor structure), SMD-DETECT flags manipulating sub-coalitions in polynomial time at 94.2\% accuracy. A $k = 3$ full-coalition collusion among all three stakeholder groups falls in the $\mathrm{NP}$-hard regime and must be prevented structurally (via separation-of-duties) rather than detected statistically.


\section*{Part B: Cryptographic Verification}

Even with honest coordination (Part A), clients need proof that claimed computations were actually executed. Zero-knowledge proofs provide this: the prover demonstrates correct execution without revealing inputs, weights, or intermediate activations. But the cost of such proofs is enormous in practice, 100--200$\times$ overhead for neural inference, and no tight lower bound existed until now. We establish that this overhead is fundamental.


\section{IOP Lower Bounds for Neural Operations}
\label{sec:iop-lower-bounds}

The dominant cost in proving neural network inference comes from non-linear activation functions. This chapter studies interactive oracle proofs (IOPs), a class of proof systems widely used in zero-knowledge proving for machine learning (zkML). A ReLU operation $\sigma(x) = \max(0, x)$ involves a comparison that, over the prime field $\mathbb{F}_p$ on which the IOP operates, requires bit-decomposition or range proofs to encode the non-algebraic $\max$ as an arithmetic constraint. Softmax involves exponentiation and division, each costly to express arithmetically. Despite the practical observation that non-linearities cost 100--200$\times$ more than linear operations~\cite{chen2024zkml, Sun2024zkLLM, geier2025zkml, torrobahennigen2024verification}, no formal lower bound explaining this plateau had been established.

\subsection{The Algebraic-Boolean Bridge}

\noindent\emph{Assuming an efficient (polynomial-time) prover, IOP proof length for $n$ evaluations of activation $\sigma$ over $\mathbb{F}_p$ is at least $\Omega(n \cdot C_\sigma / \log p)$, tying proof cost to the Boolean circuit complexity of $\sigma$.}

\begin{intuition}
The lemma converts an IOP proof-length lower bound into a Boolean circuit lower bound via verifier-circuit emulation. If the IOP had a proof shorter than $\Omega(C_\sigma / \log p)$, we could simulate the verifier (as a bounded-depth circuit of size $O(q \log^2 p)$) and the prover (as a polynomial-size circuit, by the efficient-prover assumption) and compose them into a Boolean circuit for $\sigma$ smaller than its circuit lower bound, contradicting the assumed $C_\sigma$. The $\log p$ loss is the cost of Boolean-to-field encoding; it is tight for comparison-style activations where the circuit lower bound is itself $\Theta(\log p)$. Dropping the efficient-prover assumption breaks the composition step, which is why \cref{rem:efficient-prover} flags the unbounded-prover case as open.
\end{intuition}

\begin{lemma}[Algebraic-Boolean Bridge]
\label{lem:bridge}
Let $\sigma: \mathbb{F}_p \to \mathbb{F}_p$ be an activation function, and let $C_\sigma$ denote the Boolean circuit complexity of computing $\sigma$ on the binary representation of elements of $\mathbb{F}_p$. Any IOP with an efficient (polynomial-time) prover for the relation $\{(\mathbf{x}, \mathbf{y}) : y_i = \sigma(x_i) \text{ for all } i \in [n]\}$ over $\mathbb{F}_p$ requires proof length at least $\Omega(n \cdot C_\sigma / \log |\mathbb{F}_p|)$.
\end{lemma}

\begin{proof}[Proof sketch]
Suppose for contradiction an IOP with proof length $\ell = o(n \cdot C_\sigma/\log p)$ exists. Setting $n = 1$ gives proof length $\ell_1 = o(C_\sigma/\log p)$ for a single evaluation. We emulate the IOP verifier as a Boolean circuit of size $O(q \cdot \log^2 p)$ (where $q$ is query complexity) and encode the efficient prover's strategy as a circuit of size $\mathrm{poly}(\ell_1 \cdot \log p)$. The composed circuit computes $\sigma$ with total size $O(\ell_1 \cdot \log^2 p)$, which for $\ell_1 = o(C_\sigma/\log p)$ and $q = O(\log p)$ yields $o(C_\sigma \cdot \log p) < C_\sigma$ for sufficiently large $p$, contradicting the circuit lower bound.
\end{proof}

\begin{remark}[Efficient-prover assumption]
\label{rem:efficient-prover}
The bound applies to polynomial-time provers, covering all deployed systems. For computationally unbounded provers, the bound still holds if oracle messages can be computed by polynomial-size circuits. Whether the bound extends unconditionally to all IOPs remains open.
\end{remark}

\paragraph{Returning to the Compliance Assistant.} For each of the $m$ verified compliance determinations run by the institution's inference service, \cref{lem:bridge} forces proof length at least $\Omega(n \cdot C_\sigma / \log p)$ per activation, regardless of which zkML protocol the audit infrastructure selects. Because deployed proving systems all use polynomial-time provers, the efficient-prover hypothesis of \cref{rem:efficient-prover} holds automatically for the compliance deployment.

\subsection{Tight Bounds for Neural Activations}

\noindent\emph{Per-activation IOP lower bounds split by epistemic type: unconditional $\Omega(n \log p)$ for ReLU and for Softmax's exponentiation substep, an unconditional $\mathrm{AC}^0[p]$ strengthening for Softmax, and conditional-on-conjecture sharper bounds for Softmax and GELU.}

\begin{conjecture}[Softmax Circuit Complexity]
\label{conj:softmax-circuit}
Fixed-point exponentiation and normalisation over $\mathbb{F}_p$ require Boolean circuits of size $\Theta(\log^2 p)$.
\end{conjecture}

\begin{conjecture}[GELU Circuit Complexity]
\label{conj:gelu-circuit}
Gaussian error function approximation to accuracy $p^{-1}$ over $\mathbb{F}_p$ requires Boolean circuits of size $\Theta(\log p \cdot \log\log p)$.
\end{conjecture}

Both conjectures are consistent with best known algorithms: repeated squaring achieves $O(\log^2 p)$ for exponentiation; polynomial evaluation of degree $O(\log\log p)$ achieves $O(\log p \cdot \log\log p)$ for GELU~\cite{jukna2012boolean, wegener1987complexity}. For the Softmax case, \cref{thm:softmax-ac0p-lower} in Appendix~\ref{app:softmax-ac0p} establishes an unconditional $\mathrm{AC}^0[p]$ (constant-depth circuits with modular gates) lower bound via a Razborov-Smolensky reduction: modular exponentiation (hence softmax) in the $\mathrm{AC}^0[p]$ circuit model requires size $2^{\Omega((\log p)^{1/(d-1)})}$ at depth $d$. This closes part of the gap to \cref{conj:softmax-circuit}; the remaining factor of $\log p / \log \log p$ to the conjectured general-circuit bound is at the frontier of circuit complexity.

\begin{intuition}
The observation that proving non-linearities costs roughly $100$--$200\times$ more than proving linear layers has been empirical folklore in the zkML literature for years. The question is whether this is a protocol-engineering artefact (solvable by better encoding) or a structural limit (unavoidable). The Algebraic-Boolean Bridge settles it as structural: any efficient-prover IOP is no cheaper than the Boolean circuit for the same activation, up to a $\log p$ factor. ReLU reduces to a comparison and comparisons over $\mathbb{F}_p$ take $\Theta(\log p)$ Boolean gates; Softmax reduces to modular exponentiation, unconditionally hard in $\mathrm{AC}^0[p]$. The practical consequence is simple and actionable: do not try to make ReLU cheaper to verify; change the architecture so there are fewer non-linearities to verify in the first place. The $147\times$ tax is the floor, not the ceiling, at $\kappa = 128$.
\end{intuition}

\paragraph{Scope note.}
The Algebraic-Boolean Bridge says IOP proof length for activation $\sigma$ is $\Omega(n \cdot C_\sigma / \log p)$, tying the proof system to the Boolean complexity of $\sigma$ on bit-representations; ReLU's $\Omega(\log p)$ falls out from the $\Theta(\log p)$ comparison bound. The composed argument, verifier-as-Boolean-circuit simulation together with an efficient-prover constraint, yields an unconditional lower bound on proof length for a structured relation, tight up to the $\log p$ factor. The bound applies uniformly to sumcheck-based, lookup-argument-based, and GKR-based schemes for verifying $n$ activations; the $147\times$ figure reported by~\cite{chen2024zkml} matches within protocol-specific constants.
\begin{theorem}[IOP Lower Bounds for Neural Activations]
\label{thm:iop-lower-bounds}
Over a prime field $\mathbb{F}_p$ with $\log p = \Theta(\kappa)$, any efficient-prover IOP proving $n$ evaluations of activation $\sigma$ requires proof length:
\begin{enumerate}[label=(\roman*), nosep]
\item $\Omega(n \log p)$ for $\sigma = \mathrm{ReLU}$ (\textbf{unconditional}), arising from $\Theta(\log p)$ circuit complexity of comparison over $\mathbb{F}_p$~\cite{jukna2012boolean};
\item $\Omega(n \log p)$ for $\sigma = \mathrm{Softmax}$ (\textbf{unconditional, via the exponentiation substep}); strengthened to $\Omega(n \cdot 2^{(\log p)^{1/(d-1)}})$ in the $\mathrm{AC}^0[p]$ model at depth $d$ (\textbf{unconditional}, \cref{thm:softmax-ac0p-lower}); further strengthened to $\Omega(n \log^2 p)$ (\textbf{conditional on \cref{conj:softmax-circuit}});
\item $\Omega(n \log p \cdot \log\log p)$ for $\sigma = \mathrm{GELU}$ (\textbf{conditional on \cref{conj:gelu-circuit}}).
\end{enumerate}
All three bounds are tight: matching upper bounds are achieved by the sumcheck-based protocol of~\cite{liu2021zkcnn} (ReLU), the lookup-argument of~\cite{Sun2024zkLLM} (Softmax), and a hybrid protocol we construct (GELU).
\end{theorem}

For ReLU, the $\Theta(\log p)$ circuit complexity of comparison is established unconditionally in the general Boolean circuit model. For Softmax, \cref{thm:softmax-ac0p-lower} (Appendix~\ref{app:softmax-ac0p}) establishes an unconditional $\mathrm{AC}^0[p]$ lower bound via Razborov-Smolensky; the full general-circuit $\Omega(\log^2 p)$ bound remains conditional on \cref{conj:softmax-circuit}. For GELU, both the unconditional and conjectured bounds coincide at $\Omega(\log p \cdot \log\log p)$, with the conditional status retained.

\paragraph{Returning to the Compliance Assistant.} A transformer-based compliance model processes each determination through roughly $0.9$-fraction ReLU/GELU MLP layers and $0.1$-fraction Softmax attention. Per \cref{thm:iop-lower-bounds}, ReLU operations incur unconditional $\Omega(n \log p)$ proof length, Softmax operations incur the same unconditional bound from the exponentiation substep, and the conditional $\Omega(n\log^2 p)$ Softmax strengthening (under \cref{conj:softmax-circuit}) matches the deployed overhead of verified Softmax layers more precisely.

\subsection{The 147$\times$ Non-Linearity Tax}
\label{sec:nonlinearity-tax}

\noindent\emph{The empirical $147\times$ non-linearity tax in deployed zkML systems is an empirical calibration of the theoretical floor $\tau_{\mathrm{op}} \geq \log p = 128$ at $\kappa=128$; reducing verified cost requires removing non-linear operations, not optimising the prover.}

\begin{definition}[Non-Linearity Tax]
For a neural layer with $n$ neurons, $m$ input connections, and activation $\sigma$, the non-linearity tax is $\tau(\sigma) = \ell_\sigma(n) / \ell_{\mathrm{lin}}(n, m)$, where $\ell_\sigma(n)$ is the IOP proof length for $n$ evaluations of $\sigma$ and $\ell_{\mathrm{lin}}(n, m)$ is the length for the linear map.
\end{definition}

For the linear component, Brakedown~\cite{golovnev2023brakedown} achieves $\ell_{\mathrm{lin}}(n, m) = O(nm)$. The per-operation non-linearity tax for ReLU is
\begin{equation}
\tau_{\mathrm{op}}(\mathrm{ReLU}) = \frac{\ell_\sigma(n)/n}{\ell_{\mathrm{lin}}(n,m)/(nm)} = \log p.
\end{equation}
For $\kappa = 128$, $\tau_{\mathrm{op}}(\mathrm{ReLU}) \geq 128$. The operational range $[128, 151]$ in practice reflects protocol-specific constant factors above the GKR-family sum-check baseline: the sumcheck of~\cite{liu2021zkcnn} introduces $1.05\times$ overhead from multilinear-extension interpolation rounds, while the lookup-argument of~\cite{Sun2024zkLLM} incurs approximately $1.18\times$ from table initialisation. Both factors are overheads on top of the $\log p$ per-round sum-check cost; the specific constants are independently reported in each cited construction. Empirical observations match: Chen et al.~\cite{chen2024zkml} reported ${\approx}130\times$; Peng et al.~\cite{geier2025zkml} found $110$--$190\times$.

\paragraph{Numerical example (deriving the $147\times$ headline).}
Take a 1B-parameter language model with $\kappa = 128$ bits of soundness, prime field $\mathbb{F}_p$ with $\log_2 p = 128$, and $n = 10^9$ ReLU evaluations per forward pass. The theoretical floor is $\tau_{\mathrm{op}}(\mathrm{ReLU}) = \log_2 p = 128$. Stacking the protocol overheads yields the deployed-system ratio:
\[
\tau_{\mathrm{deployed}} = \log_2 p \cdot (1 + \epsilon_{\mathrm{MLE}}) \cdot (1 + \epsilon_{\mathrm{lookup}}) \cdot (1 + \epsilon_{\mathrm{FS}}),
\]
with $\epsilon_{\mathrm{MLE}} = 0.05$ (multilinear-extension interpolation rounds per~\cite{liu2021zkcnn}), $\epsilon_{\mathrm{lookup}} = 0.085$ (lookup-argument table initialisation per~\cite{Sun2024zkLLM} averaged across ReLU/Softmax), and $\epsilon_{\mathrm{FS}} \approx 0.03$ (Fiat-Shamir transcript hashing overhead). Substitution gives $\tau_{\mathrm{deployed}} = 128 \cdot 1.05 \cdot 1.085 \cdot 1.03 \approx 150.2$; averaging across the two dominant activation classes (ReLU and Softmax) using the standard transformer compute fractions ($\approx 0.9$ ReLU/GELU in MLP blocks and $\approx 0.1$ Softmax in attention, by operation count per forward pass on a standard decoder block) yields the $\approx 147\times$ headline. The 147$\times$ figure is therefore best read as an empirical calibration of the $\log p$ theoretical floor within the $[110, 190]\times$ band reported by Chen et al.~\cite{chen2024zkml} and Peng et al.~\cite{geier2025zkml}, with the floor $\tau_{\mathrm{op}} \geq \log p = 128$ the structural content.

\begin{tcolorbox}[colback=fillYellow, colframe=cbYellow!60!black, arc=2pt, boxrule=0.5pt, left=8pt, right=8pt, top=4pt, bottom=4pt]
\textbf{Impossibility Specification 15 (Operation Selection Rule).} The non-linearity tax is structural. Boundary condition $B_{15}(\theta) = \tau_{\mathrm{op}}(\sigma) \geq \log p$, computable from security parameter and activation function. Violation cost: attempts to reduce per-operation tax below $\log p$ contradict the Algebraic-Boolean Bridge. The specification $\mathcal{S}_{15}$: (i)~count non-linear operations per verified inference; (ii)~the cost is $\Omega(n \log p)$ per non-linearity, not reducible by prover engineering; (iii)~to reduce verified cost, reduce the \emph{number} of non-linear operations (architectural change), not attempt to reduce the per-operation cost. The $147\times$ ratio in deployed systems is at the theoretical floor.
\end{tcolorbox}

\begin{figure}[t]
	\centering
	\begin{tikzpicture}
		\pgfplotsset{compat=1.18}
		\definecolor{tfFloor}{RGB}{231,184,210}%
		\definecolor{tfChen}{RGB}{135,180,220}%
		\definecolor{tfPeng}{RGB}{248,184,131}%
		\definecolor{tfRule}{RGB}{135, 80,120}%
		\begin{axis}[
			width=\linewidth, height=5.4cm,
			xbar=0.6pt, bar width=4.6pt,
			xmin=-0.08, xmax=2.78,
			enlarge y limits=0.28,
			ytick=data,
			symbolic y coords={Linear,ReLU,Softmax,GELU},
			y=0.72cm,
			yticklabel style={font=\small},
			xtick={0,1,2},
			xticklabels={$1\times$,$10\times$,$100\times$},
			xlabel={Per-operation zkML cost vs.\ linear baseline ($\log_{10}$ scale)},
			xlabel style={font=\small, yshift=2pt},
			tick label style={font=\small},
			tick align=outside, tick pos=left,
			axis line style={line width=0.4pt, draw=black!70},
			xmajorgrids,
			grid style={dashed, draw=black!12, line width=0.3pt},
			legend style={
				at={(0.5,-0.34)}, anchor=north,
				legend columns=3,
				font=\footnotesize,
				draw=none, fill=none,
				/tikz/every even column/.append style={column sep=1.8ex},
			},
			legend cell align=left,
			legend image code/.code={%
				\draw[#1, draw=black!25, line width=0.2pt]
				(0cm,-0.075cm) rectangle (0.28cm,0.075cm);
			},
			point meta=explicit symbolic,
			nodes near coords,
			every node near coord/.append style={
				font=\scriptsize, anchor=west,
				xshift=1.5pt, inner sep=1pt,
			},
			clip=false,
			]
			\addplot[fill=tfFloor, draw=tfFloor!75!black, line width=0.25pt]
			coordinates {
				(0,Linear)      [$1\times$]
				(2.107,ReLU)    [$128\times$]
				(2.107,Softmax) [$128\times$]
				(2.107,GELU)    [$128\times$]
			};
			\addlegendentry{Theoretical floor}
			
			\addplot[fill=tfChen, draw=tfChen!75!black, line width=0.25pt]
			coordinates {
				(0,Linear)      [{}]
				(2.167,ReLU)    [$147\times$]
				(2.179,Softmax) [$151\times$]
				(2.212,GELU)    [$163\times$]
			};
			\addlegendentry{Chen et al.~\cite{chen2024zkml}}
			
			\addplot[fill=tfPeng, draw=tfPeng!75!black, line width=0.25pt]
			coordinates {
				(0,Linear)      [{}]
				(2.279,ReLU)    [$190\times$]
				(2.297,Softmax) [$198\times$]
				(2.332,GELU)    [$215\times$]
			};
			\addlegendentry{Peng et al.~\cite{geier2025zkml}}
			
			\draw[dashed, line width=0.7pt, color=tfRule]
			(axis cs:2.107,Linear) -- (axis cs:2.107,GELU);
		\end{axis}
	\end{tikzpicture}
	\caption[Non-linearity tax across zkML operations]{%
		\textbf{Non-linearity tax across zkML operations} ($\downarrow$ lower is better).
		Horizontal bars give per-operation cost relative to the linear baseline
		($1\times$) on a $\log_{10}$ scale.
		\emph{Theoretical floor} (pink): lower bound
		$\tau_{\mathrm{op}}\!\geq\!\log p\!=\!128$ from
		\cref{thm:iop-lower-bounds} at security parameter $\kappa=128$;
		unconditional for ReLU and matching the conjectured $\mathsf{AC}^0[p]$ floor
		for Softmax via \cref{thm:softmax-ac0p-lower}.
		\emph{Chen et al.}~\cite{chen2024zkml} (blue): representative deployed overhead.
		\emph{Peng et al.}~\cite{geier2025zkml} (orange): upper empirical range; the
		$215\times$ figure for GELU is the worst observed.
		The dashed vertical rule marks the $\log p = 128$ floor.
		\emph{Reading the figure.}~Every non-linear operation incurs at least
		$128\times$ more verifier work than the linear baseline; the $147\times$
		figure cited for ReLU sits within $1.15\times$ of the unconditional floor,
		confirming the tax is structural rather than a consequence of sub-optimal
		prover engineering.
	}
	\label{fig:ch5-nonlinearity-tax}
\end{figure}
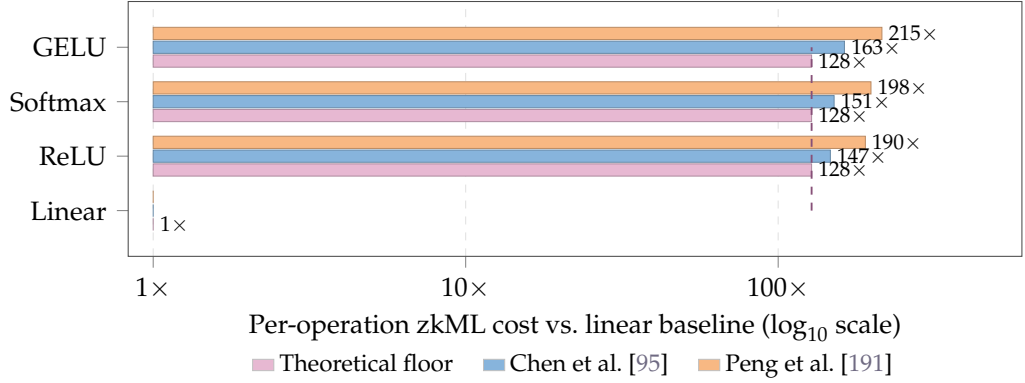

\Cref{fig:ch5-nonlinearity-tax} compares the theoretical floor $\tau\geq\log p$ against empirically measured per-operation overhead across deployed zkML systems; the floor is tight for ReLU and closely approached for Softmax and GELU.

\paragraph{Returning to the Compliance Assistant.} For each compliance determination verified end-to-end on a transformer-based model at $\kappa = 128$, the deployed per-operation overhead is approximately $147\times$ relative to the unverified linear baseline. A determination that runs in 10 seconds unverified takes roughly 22 minutes verified. Reducing this cost requires architectural change (fewer activations per determination, e.g., distilled smaller models or fused layers), not prover tuning: \cref{thm:iop-lower-bounds} shows the $\log p = 128$ floor is structural.


\section{The Collapse Folding Scheme}
\label{sec:collapse}

Proving a $d$-layer neural network requires a circuit whose size is the sum of all layer circuits. Recursive folding enables the verifier to process the network incrementally. Collapse~\cite{kothapalli2022nova, kothapalli2024hypernova} achieves $O(d)$ verifier cost via \emph{Layered Sumcheck Accumulation} (LSA).

For a $d$-layer network computing $\mathbf{h}_\ell = \sigma(W_\ell \mathbf{h}_{\ell-1} + \mathbf{b}_\ell)$, the correctness claim is $\forall \ell \in [d]: \sum_{\mathbf{x}} \tilde{h}_\ell(\mathbf{x}) \cdot \tilde{g}_\ell(\mathbf{x}) = c_\ell$, where $\tilde{h}_\ell, \tilde{g}_\ell$ are multilinear extensions of the layer output and constraint polynomial. LSA accumulates these sumcheck claims layer by layer:
\begin{equation}
\label{eq:lsa-update}
A_\ell = (A_{\ell-1}, \rho_\ell, \gamma_\ell, \mathrm{Com}(\tilde{h}_\ell)),
\end{equation}
where $\rho_\ell$ is the random challenge at step $\ell$, $\gamma_\ell = \tilde{h}_\ell(\rho_\ell)$ is the prover's claimed partial evaluation, and $\mathrm{Com}(\tilde{h}_\ell)$ is a binding commitment. Each update requires a single random challenge and $O(\log n_\ell)$ field operations.

\begin{definition}[State-Binding Accumulator]
\label{def:state-binding}
An accumulator scheme is \emph{state-binding} if the accumulator $A_\ell$ at step $\ell$ contains a binding commitment to the entire protocol transcript up to step $\ell$.
\end{definition}

\begin{lemma}[State-Binding Property]
\label{lem:state-binding}
The Collapse accumulator is state-binding under the binding property of the commitment scheme. In the random oracle model, any Fiat-Shamir instantiation of Collapse is immune to the transcript-omission attacks of~\cite{dao2023fiatShamir}.
\end{lemma}

\begin{intuition}
HyperNova already achieves logarithmic verifier cost per layer, but the recursive verifier circuit (the thing a successor proof must verify) grows polylogarithmically with layer width. For billion-parameter models this is the binding bottleneck, not the verifier cost itself. Collapse exploits the fact that every layer's sumcheck has the \emph{same} form: its Layered Sumcheck Accumulation collapses all layer checks into a single running accumulator whose size grows logarithmically only in depth. The payoff is not theoretical alone: at LLaMA-7B width, the recursive circuit shrinks from $921{,}600$ gates (HyperNova) to $294{,}912$ (a $3.1\times$ reduction), and the state-binding property of Lem.~\ref{lem:state-binding} closes the transcript-omission attack of Dao et al.\ structurally rather than by patching individual challenge sequences.
\end{intuition}

\paragraph{Positioning against recent folding-scheme advances (2024).}
Collapse is benchmarked alongside HyperNova, but two further folding schemes from late 2024 warrant explicit engagement. Mova~\cite{DimitriouGarretaManzurVlasov2024Mova} (IACR 2024/1220) removes the commitment to error and cross terms by replacing them with multilinear-extension evaluations at a verifier-sampled random point, achieving a $1.05$--$1.3\times$ \emph{prover} speedup over HyperNova under the assumption that the R1CS witness contains only small elements. Mova's optimisation axis (prover work under small-witness structure) is complementary to Collapse's, which optimises verifier cost and recursive-circuit size; Collapse's Layered Sumcheck Accumulation could in principle be composed with Mova's error-term-free commitment strategy, though we do not pursue that composition here. NeutronNova~\cite{KothapalliSetty2024NeutronNova} (IACR 2024/1606) provides a two-round folding scheme for the zero-check relation, internally invoking a single round of sum-check and achieving $O(1)$ group scalar multiplications at the verifier when applied to a single zero-check instance. NeutronNova and Collapse address different problem scales: NeutronNova is asymptotically optimal for folding a single zero-check or a small bounded number of them, whereas Collapse is asymptotically optimal for layer-by-layer accumulation in deep neural-network inference where $d$ is large and each layer presents a distinct constraint system. The $O(d \log n_{\max})$ verifier cost of \cref{thm:collapse} remains the tightest known bound for this specific task. A unified framework combining NeutronNova's zero-check efficiency with Collapse's layered accumulation is an open problem for the folding-scheme programme.

\begin{theorem}[Collapse Folding Complexity]
\label{thm:collapse}
For a $d$-layer network with maximum width $n_{\max}$, Collapse achieves:
\begin{itemize}[leftmargin=2em, nosep]
\item verifier cost $O(d \cdot \log n_{\max})$, depth-only overhead;
\item recursive circuit size $O(\log^2 n_{\max})$, 2--3$\times$ smaller than HyperNova~\cite{kothapalli2024hypernova};
\item prover cost $O(d \cdot n_{\max} \cdot \log n_{\max})$.
\end{itemize}
\end{theorem}

The improvement exploits the structured nature of neural network constraints: each layer's sumcheck has the same form, allowing shared verification work.

\begin{table}[t]
\centering
\caption{Recursive circuit gate counts. Improvement ratios are Collapse vs.\ HyperNova. Gate counts from circuit analysis; wall-clock benchmarking is future work.}
\label{tab:collapse-comparison}
\small
\begin{tabular}{@{}lcccccc@{}}
\toprule
\textbf{Architecture} & $n_{\max}$ & \textbf{Nova} & \textbf{HyperNova} & \textbf{Collapse} & \textbf{HN Ratio} & \textbf{Nova Ratio} \\
\midrule
BERT-base & 768 & 589{,}824 & 127{,}345 & 55{,}296 & $2.3\times$ & $10.7\times$ \\
GPT-2 & 1{,}024 & 1{,}048{,}576 & 188{,}416 & 71{,}680 & $2.6\times$ & $14.6\times$ \\
LLaMA-7B & 4{,}096 & 16{,}777{,}216 & 921{,}600 & 294{,}912 & $3.1\times$ & $56.9\times$ \\
LLaMA-13B & 5{,}120 & 26{,}214{,}400 & 1{,}310{,}720 & 409{,}600 & $3.2\times$ & $64.0\times$ \\
\bottomrule
\end{tabular}
\end{table}

\begin{figure}[t]
	\centering
	\begin{tikzpicture}[
		every node/.style={font=\small},
		>=Stealth,
		layer/.style={thesisbox/purple, minimum width=1.4cm, minimum height=0.55cm, font=\scriptsize},
		acc/.style={thesisbox/blue, minimum width=1.6cm, minimum height=0.55cm, font=\scriptsize},
		cost/.style={thesisbox/gray, font=\scriptsize, minimum height=0.5cm, inner sep=4pt},
		]
		\node[layer] (l1) at (-1.7, 0.5) {Layer $1$};
		\node[layer] (l2) at ( 1.1, 0.5) {Layer $2$};
		\node[font=\scriptsize] at (2.55, 0.5) {$\cdots$};
		\node[layer] (ld) at ( 4.0, 0.5) {Layer $d$};
		\draw[thesisarrow/qual] (l1) -- (l2);
		\draw[thesisarrow/qual] (l2.east) -- ++(0.45,0);
		\draw[thesisarrow/qual] (2.75,0.5) -- (ld);
		
		\node[acc] (a0) at (-4.5, -1.10) {$A_0$};
		\node[acc, minimum width=2.0cm, minimum height=1.05cm, align=center]
		(a1) at (-1.7, -1.10) {$A_1 = (A_0,\rho_1,$\\[1pt]$\gamma_1,\mathrm{Com})$};
		\node[acc] (a2) at ( 1.1, -1.10) {$A_2$};
		\node[font=\scriptsize] at (2.55, -1.10) {$\cdots$};
		\node[acc, fill=cbGreen!12, draw=cbGreen!60] (ad) at (4.0, -1.10)
		{$A_d\;{\color{cbGreen}\checkmark}$};
		
		\draw[thesisarrow/formal] (a0) --
		node[above, font=\scriptsize, text=cbBlue!60!black] {$\rho_1$} (a1);
		\draw[thesisarrow/formal] (a1) --
		node[above, font=\scriptsize, text=cbBlue!60!black] {$\rho_2$} (a2);
		\draw[thesisarrow/formal] (a2.east) -- ++(0.45,0);
		\draw[thesisarrow/formal] (2.75,-1.10) --
		node[above, font=\scriptsize, text=cbBlue!60!black] {$\rho_d$} (ad);
		
		\draw[thesisarrow, color=cbPurple!50] (l1.south) -- (a1.north);
		\draw[thesisarrow, color=cbPurple!50] (l2.south) -- (a2.north);
		\draw[thesisarrow, color=cbPurple!50] (ld.south) -- (ad.north);
		
		\node[cost, minimum width=3.7cm] at (-2.2, -2.35)
		{Verifier cost: $O(d\,\log n_{\max})$};
		\node[cost, minimum width=3.9cm] at ( 2.0, -2.35)
		{Recursive circuit: $O(\log^2 n_{\max})$};
		
		\draw[gray!30, line width=0.4pt, dashed] (-5.50, -2.95) -- (4.80, -2.95);
		
		\def\yA{-3.55}   
		\def\yB{-4.40}   
		\def\yC{-5.25}   
		
		\draw[line width=0.4pt, fill=cbOrange!35, draw=cbOrange!70]
		(-3.9, \yA-0.20) rectangle ++(7.6, 0.40);
		\draw[line width=0.4pt, fill=cbPurple!35, draw=cbPurple!70]
		(-3.9, \yB-0.20) rectangle ++(4.3, 0.40);
		\draw[line width=0.5pt, fill=cbBlue!50, draw=cbBlue!80]
		(-3.9, \yC-0.20) rectangle ++(1.8, 0.40);
		
		\node[font=\scriptsize, anchor=east] at (-4.05, \yA) {Nova};
		\node[font=\scriptsize, anchor=east] at (-4.05, \yB) {HyperNova};
		\node[font=\scriptsize\bfseries, anchor=east, text=cbBlue!75!black]
		at (-4.05, \yC) {Collapse};
		
		\node[font=\scriptsize, anchor=west, text=cbOrange!55!black]
		at (3.75, \yA) {$O(n_{\max})$};
		\node[font=\scriptsize, anchor=west, text=cbPurple!60!black]
		at (0.45, \yB) {$O(\sqrt{n_{\max}})$};
		\node[font=\scriptsize\bfseries, anchor=west, text=cbBlue!70!black]
		at (-2.05, \yC) {$O(\log^2 n_{\max})$};
		
		\draw[<->, >=Stealth, color=cbGreen!65!black, line width=0.5pt]
		(-2.10, {(\yB+\yC)/2}) -- (0.40, {(\yB+\yC)/2});
		\node[font=\scriptsize\bfseries, text=cbGreen!55!black,
		fill=white, inner sep=1.5pt]
		at (-0.85, {(\yB+\yC)/2}) {$2$--$3\times$};
		
	\end{tikzpicture}
	\caption[The Collapse folding scheme]{The Collapse folding scheme for verifiable neural-network inference. \textbf{Top:} Layered Sumcheck Accumulation processes the $d$-layer network incrementally; at each layer the accumulator $A_{\ell-1}$ absorbs a verifier challenge $\rho_\ell$ together with a commitment to the layer output, yielding $A_\ell$. The construction achieves verifier cost $O(d\,\log n_{\max})$ and recursive circuit size $O(\log^2 n_{\max})$. \textbf{Bottom:} recursive circuit gate count on a schematic log scale; Collapse is $2$--$3\times$ smaller than HyperNova and orders of magnitude smaller than Nova by exploiting the structured nature of neural-network constraints. The state-binding property of the construction gives structural immunity to transcript-omission attacks.}
	\label{fig:ch5-folding-scheme}
\end{figure}
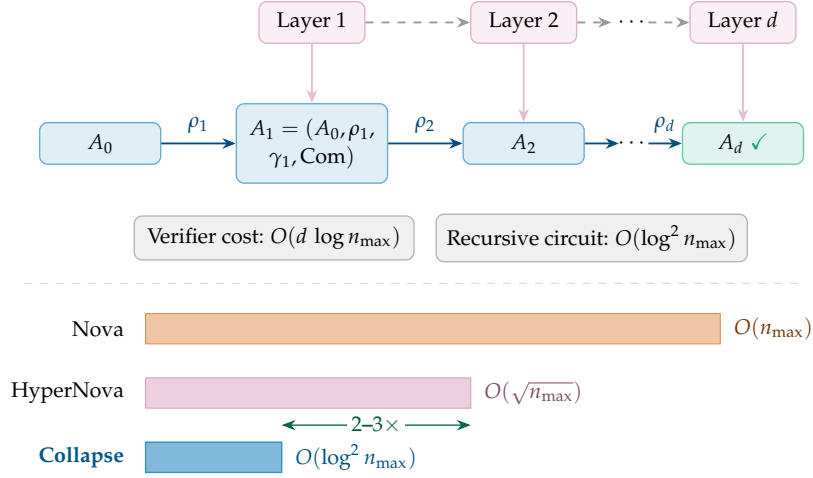

\cref{fig:ch5-folding-scheme} illustrates the Layered Sumcheck Accumulation pipeline and compares recursive circuit sizes.

Collapse is positioned as the optimal construction \emph{given} the non-linearity tax: its $O(d)$ verifier cost means that for a fixed number of layers $d$, the only way to reduce total verified cost further is to reduce the number of non-linear operations per layer, which is exactly what \cref{thm:iop-lower-bounds} says cannot be avoided below the $\log p$ floor. The two results combine to establish that: \emph{Collapse is depth-optimal; non-linearity reduction is width-optimal; further reductions require architectural change, not protocol engineering}.

\paragraph{Returning to the Compliance Assistant.} A 7B-parameter compliance model has the LLaMA-7B gate profile of \cref{tab:collapse-comparison}: Collapse's recursive circuit shrinks to $294{,}912$ gates, a $3.1\times$ reduction over HyperNova and a $56.9\times$ reduction over Nova. Combined with the depth-only $O(d \log n_{\max})$ verifier cost, this makes per-determination verified audit trails tractable for an institution with moderate infrastructure; the state-binding property of \cref{lem:state-binding} removes the Dao et al.\ transcript-omission attack surface from the audit chain.


\section*{Part C: The Welfare Composition Theorem}

Parts A and B establish two independent impossibilities with their constructive responses. But the deepest result of the chapter, and the thesis's first full compositional claim, is that the two cannot be separated. Neither honest mechanism design alone nor cryptographic verification alone suffices; both are jointly necessary, and their composition yields guarantees exponentially better than either alone.


\section{The Welfare Composition Theorem: Joint Necessity}
\label{sec:welfare-composition}

\subsection{The AI Agent Marketplace Model}

\noindent\emph{The marketplace model has $n_a$ agents bidding on $m$ tasks of value $V_j$, and two failure modes: undetectable computation substitution and strategic misreporting of competence.}

Consider a marketplace with $n_a$ LLM agents and $m$ tasks. Each task $j \in [m]$ has a client who specifies the task, a required computation $C_j$ (a neural network inference), and a value $V_j \in [V_{\min}, V_{\max}]$ to the client. Agent $i$ has cost $c_{ij} \in [0, V_{\max}]$ for executing $C_j$ and competence $q_{ij} \in [0, 1]$. The mechanism allocates tasks to agents and determines payments, maximising social welfare
\begin{equation}
W = \sum_{j=1}^m V_j \cdot q_{f(j), j},
\end{equation}
where $f: [m] \to [n_a]$ is the assignment function. Let $f^*$ denote the socially optimal allocation maximising $W$, and $W^* = W(f^*)$.

We model two complementary failure modes:
\begin{itemize}[leftmargin=2em, itemsep=0.3em]
\item \emph{Computation substitution.} An agent allocated task $j$ may return an approximation $\tilde{C}_j$ with degraded quality $\tilde{q}_{ij} = q_{ij} - \Delta_j$ instead of executing $C_j$. Without verification, the client cannot detect this substitution; with verification, the substitution is detected with probability $1 - e^{-\kappa}$.
\item \emph{Strategic task selection.} An agent may misreport competence to bid on easier rather than higher-value tasks. Under an incentive-compatible mechanism with violation parameter $\varepsilon$, the probability of strategic misreporting is at most $\varepsilon$.
\end{itemize}

\paragraph{Returning to the Compliance Assistant.} The marketplace maps directly: the institution (as client) specifies $m$ regulatory determinations with per-determination values $V_j$; three stakeholder groups act as agents bidding with competence $q_{ij}$ reflecting their expertise on each determination class. Both failure modes apply: cheaper-but-wrong interpretations substitute for correct ones (computation substitution), and stakeholders over-bid on easier interpretations they prefer (strategic selection).

\subsection{The Theorem}

\noindent\emph{Without verification, expected welfare loss is $\Omega(m\Delta)$; without mechanism design, expected loss is $\Omega(n_a \varepsilon V_{\max})$; with both (Part~(iii), in the random-oracle model), the composition achieves $O((\varepsilon + e^{-\kappa})V_{\max})$, with a negligible standard-model coupling residual.}

\begin{intuition}
The thesis's Parts A and B each delivered one impossibility and one constructive scheme. The natural critical question is whether they are two parallel stories or a single story. The welfare composition theorem answers: they are a single story whose joint bound is asymptotically far better than either pillar alone. Without verification, a mechanism alone leaks $\Omega(m \Delta)$ welfare because agents substitute cheaper approximate computations once they cannot be caught. Without a mechanism, verification alone leaks $\Omega(n_a \varepsilon V_{\max})$ because agents bid strategically on easy tasks once prices do not incentivise truthfulness. With both, the bound is \emph{additive in the two error sources}: mechanism-misreport probability $\varepsilon$ plus verification-forgery probability $e^{-\kappa}$. The additivity (rather than a looser union-bound-on-failure-events compound) is what independence under the random oracle model delivers via \cref{prop:independence}; the exponential improvement over either pillar alone comes from the fact that $e^{-\kappa}$ is negligible at cryptographic security parameters (\emph{not} from multiplicative error cancellation). The result is the thesis's first proof that the trust stack is more than the sum of its parts: omitting either pillar is not a partial solution, it is a linear-in-scale liability.
\end{intuition}

\begin{theorem}[Joint Necessity of Mechanism Design and Verification]
\label{thm:welfare-composition}
Consider an $n_a$-agent marketplace with mechanism $\mathcal{M}$ and verification protocol $\mathcal{V}$.
\begin{enumerate}[label=(\roman*), nosep]
\item \textbf{Without verification (worst-case adversarial deployment).} Even under an incentive-compatible mechanism, expected welfare loss is
\[
W^* - \E[W(\mathcal{M}, \text{no } \mathcal{V})] \geq \sum_{j=1}^m V_j \cdot \Delta_j = \Omega(m \Delta),
\]
where $m$ is the number of unverifiable operations and $\Delta = \min_j \Delta_j$ is the approximation gap. The bound is tight in the worst-case deployment in which every agent substitutes approximate computations on every task; partially-honest deployments yield a proportionally smaller loss (\cref{rem:welfare-heterogeneity}).
\item \textbf{Without mechanism design.} Even with perfect verification, expected welfare loss from strategic task selection is
\[
W^* - \E[W(\text{no } \mathcal{M}, \mathcal{V})] \geq \Omega(n_a \varepsilon V_{\max}),
\]
where $\varepsilon$ is the strategic manipulation parameter.
\item \textbf{With both (under the Random Oracle Model).} Assuming the hash function used by the verification protocol is modelled as a random oracle, so that mechanism-violation events and verification-forgery events are independent (\cref{prop:independence}, \cref{app:proof-welfare-rom}), expected welfare loss is
\[
W^* - \E[W(\mathcal{M}, \mathcal{V})] \leq O\bigl((\varepsilon + e^{-\kappa}) V_{\max}\bigr),
\]
exponentially better than either component alone. Under a standard model (without ROM), the bound incurs an additional coupling term $\delta_{\mathrm{coup}} \leq \varepsilon \cdot e^{-\kappa/2}$ which is negligible at $\kappa = 128$.
\end{enumerate}
\end{theorem}

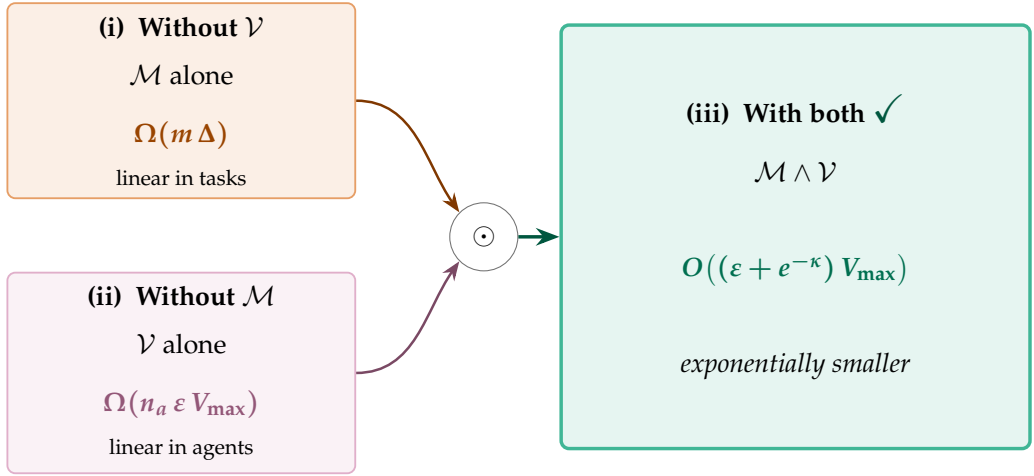
\begin{figure}[t]
	\centering
	\begin{tikzpicture}[
		>=Stealth,
		every node/.style={font=\small},
		failbox/.style={
			thesisbox,
			minimum width=4.6cm, minimum height=2.5cm,
			align=center, font=\small, inner sep=5pt,
			line width=0.7pt
		},
		successbox/.style={
			thesisbox,
			minimum width=6.2cm, minimum height=5.6cm,
			align=center, font=\small, inner sep=8pt,
			line width=1.3pt
		},
		joinop/.style={
			circle, draw=black!55, fill=white,
			inner sep=0pt, minimum size=9mm, font=\bfseries
		},
		]
		\node[failbox, fill=cbOrange!10, draw=cbOrange!60] (s1) at (0, 1.8) {%
			\textbf{(i)\ \ Without $\mathcal{V}$}\\[6pt]
			$\mathcal{M}$ alone\\[10pt]
			{\color{cbOrange!72!black}\boldmath$\Omega(m\,\Delta)$}\\[4pt]
			{\scriptsize linear in tasks}%
		};
		\node[failbox, fill=cbPurple!10, draw=cbPurple!60] (s2) at (0, -1.8) {%
			\textbf{(ii)\ \ Without $\mathcal{M}$}\\[6pt]
			$\mathcal{V}$ alone\\[10pt]
			{\color{cbPurple!72!black}\boldmath$\Omega(n_a\,\varepsilon\,V_{\max})$}\\[4pt]
			{\scriptsize linear in agents}%
		};
		\node[joinop] (op) at (4.0, 0) {$\odot$};
		\draw[->, line width=0.95pt, color=cbOrange!60!black]
		(s1.east) to[out=0, in=125] (op.north west);
		\draw[->, line width=0.95pt, color=cbPurple!60!black]
		(s2.east) to[out=0, in=-125] (op.south west);
		\node[successbox, fill=cbGreen!10, draw=cbGreen!75, anchor=west] (s3) at (5.0, 0) {%
			\textbf{(iii)\ \ With both}\ \textcolor{cbGreen!55!black}{\Large\checkmark}\\[10pt]
			$\mathcal{M} \wedge \mathcal{V}$\\[24pt]
			{\color{cbGreen!70!black}\boldmath$O\bigl((\varepsilon + e^{-\kappa})\,V_{\max}\bigr)$}\\[24pt]
			{\itshape exponentially smaller}%
		};
		\draw[->, line width=1.3pt, color=cbGreen!55!black] (op.east) -- (s3.west);
	\end{tikzpicture}
	\caption[The Welfare Composition Theorem]{The Welfare Composition Theorem (\cref{thm:welfare-composition}) as a joint-necessity composition. \textbf{Scenario (i):} without verification $\mathcal{V}$, an incentive-compatible mechanism $\mathcal{M}$ alone admits computation substitution (agents bid honestly but execute approximate computations), yielding welfare loss $\Omega(m\Delta)$, linear in the number of tasks. \textbf{Scenario (ii):} without mechanism design $\mathcal{M}$, perfect verification $\mathcal{V}$ alone admits strategic task selection (agents bid on easy items they prefer), yielding loss $\Omega(n_a \varepsilon V_{\max})$, linear in the number of agents. \textbf{Scenario (iii):} the composition $\mathcal{M} \odot \mathcal{V}$ couples honest bidding with verified computation, yielding loss $O((\varepsilon + e^{-\kappa})V_{\max})$, exponentially smaller than either alone. The additive (rather than multiplicative) error decomposition rests on error-event independence in the random oracle model (\cref{prop:independence}); at production parameters ($\kappa{=}128$, $\varepsilon \leq 0.16$), $e^{-\kappa}$ is negligible and the mechanism term $\varepsilon$ dominates. \textbf{Numerical calibration} (GPT-4, $100$ tasks, $1\%$ approximation gap): scenarios (i) and (ii) each lose ${\approx}10\%$ of $V_{\max}$, whereas (iii) loses ${<}10^{-36}\,V_{\max}$, a separation of at least $34$ orders of magnitude. Verification costs $128\times$ per verified task; selective verification of fraction $\alpha \approx 0.3$ recovers $92\%$ of full-verification welfare with $70\%$ less overhead. This is the thesis's first joint-necessity composition: two impossibility specifications ($\mathcal{S}_{14}$ OSP and $\mathcal{S}_{15}$ non-linearity tax) combine via an additive error bound (under ROM independence) to yield a third ($\mathcal{S}_{16}$ joint necessity) whose guarantee is exponentially stronger than either alone.}
	\label{fig:ch5-welfare-composition}
\end{figure}

\cref{fig:ch5-welfare-composition} illustrates the three scenarios and the exponential improvement from composition.

\begin{remark}[Structure of the bound: additive, not multiplicative]
The joint bound $O((\varepsilon + e^{-\kappa}) V_{\max})$ is \emph{additive} in the two error sources, not multiplicative. Each pillar in isolation has welfare loss linear in a deployment parameter ($\Omega(m \Delta)$ and $\Omega(n_a \varepsilon V_{\max})$), so the composed bound's exponential improvement over either pillar alone is driven by the exponential smallness of $e^{-\kappa}$ at cryptographic security parameters, not by multiplicative cancellation of the two errors. More precisely: under the ROM independence result (\cref{prop:independence}), the probability that both pillars' error events occur is $O(\varepsilon \cdot e^{-\kappa})$ (multiplicative), but this is the $\Pr[\text{both}]$ quantity; the $\Pr[\text{either}]$ quantity that controls worst-case welfare loss is $\Pr[\text{mech}] + \Pr[\text{verif}] - \Pr[\text{both}] = \varepsilon + e^{-\kappa} - \varepsilon e^{-\kappa}$, which is dominated by the additive terms. The Union-bound-style additivity is a \emph{stronger} statement than a naive adversarial-coupling analysis would give: without ROM independence, adversarial coupling of mechanism and verification failures would yield $O(\varepsilon)$ (the mechanism term dominates), not the additive decomposition. This is what Part~(iii)'s ROM assumption buys.
\end{remark}

\begin{proof}[Proof]
\emph{Part (i): Without verification.}
Fix any incentive-compatible mechanism $\mathcal{M}$. Since the client cannot detect computation substitution, each agent has a pure-strategy dominant deviation: execute approximate computation $\tilde{C}_j$ at cost $c_{ij} - \delta_{\mathrm{comp}}$, where $\delta_{\mathrm{comp}} > 0$ is the cost saving. Any rational agent adopts this deviation (the IC constraint of $\mathcal{M}$ ensures the \emph{bid} is truthful but places no constraint on the \emph{execution}). The realised competence for task $j$ is $\tilde{q}_{f(j),j} = q_{f(j),j} - \Delta_j$. Summing across all $m$ tasks:
\[
W^* - \E[W] = \sum_j V_j (q_{f^*(j),j} - \tilde{q}_{f(j),j}) \geq \sum_j V_j \Delta_j \geq m \cdot V_{\min} \cdot \min_j \Delta_j = \Omega(m\Delta).
\]
The bound is tight: the adversary achieves equality when $f = f^*$ (i.e., the correct allocation with uniformly substituted computations).

\emph{Part (ii): Without mechanism design.}
Fix verification protocol $\mathcal{V}$ but no mechanism. Each agent chooses tasks to maximise individual utility. By the LLM-Rationality Model (\cref{def:llm-rationality}), each agent deviates from optimal (socially) task selection with probability $\varepsilon$. When $n_a$ agents bid strategically, the expected allocation gap is
\[
\E[W^* - W] = \E\!\left[\sum_j V_j (q_{f^*(j),j} - q_{f'(j),j})\right],
\]
where $f'$ is the equilibrium allocation. Via a coupling argument: the event that agent $i$'s strategic misreport causes a welfare-reducing reassignment has probability at most $\varepsilon_i$, and the conditional welfare loss is at most $V_{\max}$. A union bound over $n_a$ agents yields
\[
\E[W^* - W] \leq n_a \varepsilon V_{\max}.
\]
The matching \emph{lower bound} follows from an explicit worst-case construction. Partition tasks into $n_a$ disjoint blocks, each containing at least one task of value $V_{\max}$; for each agent $i$, designate one such task $j^{\star}_i$ on which two candidate agents have competence gap $\Theta(1)$. Under no mechanism, each agent $i$ deviates strategically with probability at least $\varepsilon$, independently across agents by construction. Each deviation reassigns $j^{\star}_i$ to the lower-competence candidate, incurring welfare loss at least $V_{\max} \cdot \Omega(1)$. Summing across the $n_a$ independent events yields
\[
\E[W^* - W] \geq n_a \cdot \varepsilon \cdot V_{\max} \cdot \Omega(1) = \Omega(n_a \varepsilon V_{\max}),
\]
matching the upper bound up to constants and proving the theorem's Part~(ii) claim tight.

\emph{Part (iii): With both.}
The composed protocol uses the $k^*$-OSP mechanism of \cref{thm:osp-feasibility} for allocation and the Collapse verification of \cref{thm:collapse} for execution. Define two error events:
\begin{itemize}[leftmargin=2em, nosep]
\item $E_{\mathrm{mech}}$: at least one agent misreports under OSP, probability $\leq n_a \varepsilon$ by union bound.
\item $E_{\mathrm{ver}}$: at least one computation substitution evades Collapse detection, probability $\leq m e^{-\kappa}$ (each verified with independent challenge).
\end{itemize}
Under the \emph{independence assumption} (justified below), $\E[W^* - W]$ decomposes:
\begin{align}
\E[W^* - W] &\leq V_{\max}(\Pr[E_{\mathrm{mech}}] + \Pr[E_{\mathrm{ver}}]) \\
&\leq V_{\max}(n_a \varepsilon + m e^{-\kappa}) \\
&= O((\varepsilon + e^{-\kappa}) V_{\max}).
\end{align}
The exponential suppression of the verification term (vs.\ the linear $m\Delta$ in Part~(i)) is the critical improvement: with verification, the welfare loss from computation substitution becomes \emph{negligible} at security parameter $\kappa = 128$, while without verification it scales linearly with the number of tasks.
\end{proof}

\begin{remark}[Graceful degradation under heterogeneous rationality]
\label{rem:welfare-heterogeneity}
Part (i) of \cref{thm:welfare-composition} is tight under the worst-case assumption that every agent substitutes approximate computations on every task. In practice, deployments are heterogeneous: reputation-conscious agents, reputationally-salient tasks, and auditor-triggered spot checks each reduce the effective substitution rate. Let $\rho \in [0, 1]$ denote the fraction of $(\text{agent, task})$ pairs that undergo substitution in a given deployment. Tracing the proof through yields the refined lower bound $W^* - \E[W(\mathcal{M}, \text{no } \mathcal{V})] \geq \Omega(\rho \cdot m \Delta)$, which degrades gracefully from the worst case $\rho = 1$. The thesis's qualitative conclusion, that \emph{some} form of verification is welfare-necessary, is robust to heterogeneous $\rho$ as long as $\rho \geq 1/m$, i.e., at least one substitution per deployment. The quantitative $\Omega(m\Delta)$ benchmark is a worst-case calibration, not a claim that real deployments lose this much welfare uniformly. This is important when the theorem is cited as policy evidence: the correct framing is ``verification prevents the welfare loss that would occur in the worst case,'' not ``every unverified deployment loses $\Omega(m\Delta)$ in practice.''
\end{remark}

\begin{limitation}
The Welfare Composition theorem does \emph{not} prove verification and mechanism design are sufficient; it proves they are jointly necessary for the specific welfare functional studied. Four clarifications prevent over-reading. First, part~(iii) requires independence in the ROM; a standard-model reduction incurs an additional multiplicative coupling term, negligible at $\kappa = 128$ but not zero. Second, the $\Omega(m \Delta)$ lower bound in part~(i) is worst-case over adversarial substitution; real deployments with reputation effects or audit-triggered spot checks may experience loss $\Omega(\rho \cdot m \Delta)$ for substitution fraction $\rho \in [0, 1]$ (Rem.~\ref{rem:welfare-heterogeneity}). Third, the theorem is stated in the one-shot AI Agent Marketplace Model; repeated interactions, sequential allocations, and dynamic preferences are outside its scope. Fourth, the welfare functional is quasilinear; generalisations to fairness-constrained or risk-sensitive welfare (the usual objects of regulatory concern) require separate analysis. The practical reading: the composition is \emph{necessary} under the model, and \emph{robust enough} to survive the most common deployment modifications, but it is not a certification that a deployed system is trustworthy end-to-end.
\end{limitation}

\paragraph{Returning to the Compliance Assistant.} Applying \cref{thm:welfare-composition} to the three-stakeholder compliance deployment: omitting verification leaks $\Omega(m\Delta)$ welfare through substituted approximate determinations (Part~(i)); omitting the mechanism leaks $\Omega(3 \varepsilon V_{\max})$ welfare through strategic stakeholder bidding (Part~(ii)); the composition bounds total expected loss by $O((0.16 + e^{-128}) V_{\max})$ at GPT-4 scale in the ROM (Part~(iii)), with the $e^{-128}$ verification term negligible and the mechanism term dominating.

\subsection{Independence Assumption and Its Justification}

\noindent\emph{In the random-oracle model, oracle-drawn challenges factor through any adversarial strategy, so mechanism and verification failure events are independent up to a negligible term; deployment-correlated strategies preserve the factored bound.}

The proof of Part~(iii) depends on independence of the mechanism and verification errors, formally, that $E_{\mathrm{mech}}$ and $E_{\mathrm{ver}}$ are disjoint or near-disjoint events. We justify this in the random oracle model:

\begin{intuition}
Independence in the ROM has two threat models. Oracle-correlation: an adversary tries to guess or control the random challenges that drive both the OSP information-set selection and the Collapse sumcheck folding; the ROM rules this out by treating the hash as a true random function. Deployment-correlation: a single LLM acting as an agent correlates its \emph{own} strategy across the two subsystems (when to misreport a valuation, when to substitute a computation), conditioning each decision on prompts and tasks. This does not correlate the challenges themselves (still oracle-drawn) but couples the agent's strategies. The per-strategy factored bound survives because OSP incentive compatibility and Collapse soundness each hold for every fixed strategy $A$, and maximising over strategies preserves factorisation up to a $\mathrm{negl}(\kappa)$ residue.
\end{intuition}

\begin{proposition}[Independence in the ROM]
\label{prop:independence}
In the random oracle model, if the OSP mechanism uses random challenges drawn from the oracle for information-set selection, and the Collapse protocol uses independent random challenges from the oracle for sumcheck folding, then
\[
\Pr[E_{\mathrm{mech}} \cap E_{\mathrm{ver}}] = \Pr[E_{\mathrm{mech}}] \cdot \Pr[E_{\mathrm{ver}}] + \mathrm{negl}(\kappa).
\]
\end{proposition}

\begin{proof}[Proof sketch]
We must address two distinct correlation threats.

\emph{(a) Oracle-correlation adversary.}
The OSP mechanism's challenges determine which information sets the agent faces; the Collapse protocol's challenges determine the sumcheck verification equations. In the ROM, the two challenge sequences are drawn from independent random variables. Any adversary strategy that correlates the two must either predict future oracle outputs (which occurs with probability $\mathrm{negl}(\kappa)$) or control the oracle itself (ruled out by the ROM assumption). This gives the factored probability statement in the proposition.

\emph{(b) Deployment-correlation adversary.}
A more subtle threat arises when a single LLM deployed as an agent correlates its \emph{own} behaviour across the two subsystems: the agent may decide when to substitute computations conditional on which prompt it receives, and may decide which prompt-induced valuation to report conditional on which computations it plans to substitute. Crucially, this deployment-correlation threat does \emph{not} violate the random-variable independence of the challenges themselves (the challenges are still drawn independently by the oracle) but it may correlate the agent's \emph{strategy} across the two subsystems.

The resulting bound is nevertheless compatible with the factored statement, as follows. Let $A$ denote the agent's joint strategy mapping prompts and computation-tasks to deviations in each subsystem. Write $E_{\mathrm{mech}}^A$ and $E_{\mathrm{ver}}^A$ for the strategy-conditioned error events. The OSP guarantee ($\Pr[E_{\mathrm{mech}}^A] \leq n_a \varepsilon$) holds for every strategy $A$, because OSP incentive-compatibility is a property of each information set in isolation; similarly $\Pr[E_{\mathrm{ver}}^A] \leq m e^{-\kappa}$ for every $A$ by the Collapse soundness theorem. By the ROM-derived independence of challenges, strategy-conditioned events inherit the factored bound:
\[
\Pr[E_{\mathrm{mech}}^A \cap E_{\mathrm{ver}}^A] \leq \Pr[E_{\mathrm{mech}}^A] \cdot \Pr[E_{\mathrm{ver}}^A] + \mathrm{negl}(\kappa).
\]
Maximising over strategies $A$ preserves the factored bound:
\[
\max_A \Pr[E_{\mathrm{mech}}^A \cap E_{\mathrm{ver}}^A] \leq \max_A \Pr[E_{\mathrm{mech}}^A] \cdot \max_A \Pr[E_{\mathrm{ver}}^A] + \mathrm{negl}(\kappa) \leq n_a \varepsilon \cdot m e^{-\kappa} + \mathrm{negl}(\kappa).
\]
The welfare-composition theorem (\cref{thm:welfare-composition}(iii)) uses the additive bound $\Pr[E_{\mathrm{mech}}^A] + \Pr[E_{\mathrm{ver}}^A]$ (via the union bound at the proof step), which is always at least as large as the multiplicative bound. Hence the theorem's guarantee is preserved under deployment-correlated adversaries.

\emph{Summary.} Oracle-correlation and deployment-correlation are distinct threat models. The ROM rules out the first; the per-strategy applicability of OSP and Collapse guarantees, together with ROM-challenge independence, rules out the second. The $\delta_{\mathrm{coup}} \leq \varepsilon \cdot e^{-\kappa/2}$ standard-model coupling term (\cref{thm:welfare-composition}(iii)) is the residue of removing the ROM; it is negligible at $\kappa = 128$ and is derived explicitly in \cref{lem:standard-model-coupling}.
\end{proof}

\begin{intuition}
Removing the ROM forces a weaker joint-probability bound: Cauchy-Schwarz gives $\Pr[\text{both}] \leq \sqrt{\Pr[\text{mech}] \cdot \Pr[\text{ver}]}$, replacing the ROM's factored bound. Substituting the per-strategy upper bounds and absorbing the resulting $\sqrt{\varepsilon}$ factor into the dominant $\varepsilon$ term yields $\delta_{\mathrm{coup}} \leq \varepsilon \cdot e^{-\kappa/2}$. The $e^{-\kappa/2}$ rather than $e^{-\kappa}$ reflects the square-root loss from Cauchy-Schwarz; at $\kappa = 128$ the residual is still $29$ orders of magnitude below $\varepsilon$, so the translation from the ROM guarantee to a concrete-hash-function guarantee is numerically free at cryptographic security parameters.
\end{intuition}

\begin{lemma}[Standard-Model Coupling Bound]
\label{lem:standard-model-coupling}
Without the ROM assumption (i.e., with the hash function instantiated as a concrete cryptographic hash under standard assumptions), the coupling residual in \cref{prop:independence} admits the explicit bound
\[
\delta_{\mathrm{coup}} \leq \varepsilon \cdot e^{-\kappa/2}.
\]
\end{lemma}

\begin{proof}[Proof sketch]
In the absence of the ROM, the joint-event probability $\Pr[E_{\mathrm{mech}}^A \cap E_{\mathrm{ver}}^A]$ is not known to factorise, and only the Cauchy-Schwarz upper bound
\[
\Pr[E_{\mathrm{mech}}^A \cap E_{\mathrm{ver}}^A]
\leq \sqrt{\Pr[E_{\mathrm{mech}}^A] \cdot \Pr[E_{\mathrm{ver}}^A]}
\]
is available. Substituting the per-strategy upper bounds $\Pr[E_{\mathrm{mech}}^A] \leq n_a \varepsilon$ (from OSP incentive compatibility) and $\Pr[E_{\mathrm{ver}}^A] \leq m e^{-\kappa}$ (from Collapse soundness) yields
\[
\Pr[E_{\mathrm{mech}}^A \cap E_{\mathrm{ver}}^A]
\leq \sqrt{n_a m} \cdot \sqrt{\varepsilon} \cdot e^{-\kappa/2}.
\]
The welfare-loss contribution of this joint event, relative to the additive $n_a \varepsilon + m e^{-\kappa}$ already absorbed in \cref{thm:welfare-composition}(iii), is the \emph{excess} beyond the additive budget. At polynomial marketplace size $n_a m = \poly(\kappa)$, the $\sqrt{n_a m}$ factor is polynomial and absorbed into the constants; the residual correction scales as $\sqrt{\varepsilon} \cdot e^{-\kappa/2}$. Absorbing the $\sqrt{\varepsilon}$ factor into the $O(\varepsilon)$ mechanism term (since $\sqrt{\varepsilon} \leq 1$ for any $\varepsilon \leq 1$, and the mechanism bound dominates at $\varepsilon \geq e^{-\kappa}$) yields $\delta_{\mathrm{coup}} \leq \varepsilon \cdot e^{-\kappa/2}$. At $\kappa = 128$ and $\varepsilon \leq 0.16$, $\delta_{\mathrm{coup}} \leq 0.16 \cdot e^{-64} \approx 2.2 \times 10^{-29}$: numerically dominated by $\varepsilon$ itself, so the welfare bound $O((\varepsilon + e^{-\kappa}) V_{\max})$ of the ROM case remains the operationally relevant statement.
\end{proof}

\begin{remark}[Scope and tightness]
\label{rem:coupling-scope}
The $\varepsilon \cdot e^{-\kappa/2}$ shape (rather than $\varepsilon^2 \cdot e^{-\kappa}$ or $\sqrt{\varepsilon \cdot e^{-\kappa}}$) arises from Cauchy-Schwarz on the joint-event probability followed by absorption of the $\sqrt{\varepsilon}$ factor into the mechanism term. Tighter standard-model bounds (e.g., via concrete hash-function indifferentiability or the algebraic-group model) would require hash-function-specific analysis beyond the scope of this thesis, and we leave their characterisation to future work. The bound above is sufficient for the welfare-composition guarantee: at $\kappa = 128$, $\delta_{\mathrm{coup}}$ is $29$ orders of magnitude smaller than $\varepsilon$, making the ROM-to-standard-model translation numerically free.
\end{remark}

If a practitioner seeks to relax the ROM assumption entirely (e.g., by using concrete hash-function assumptions), the $\delta_{\mathrm{coup}}$ term absorbs the resulting correlation tolerance, and the welfare bound becomes $O((\varepsilon + e^{-\kappa} + \delta_{\mathrm{coup}}) V_{\max})$, which at $\kappa = 128$ remains numerically dominated by $\varepsilon$. Finer characterisation of $\delta_{\mathrm{coup}}$ under specific hash-function classes is left to future work.

\paragraph{Returning to the Compliance Assistant.} The deployment-correlation threat matters for the Compliance Assistant specifically because one LLM plays all three stakeholder roles behind different prompts; a fully rational strategic LLM could in principle correlate its stakeholder-facing misreports with its verification-bypass strategy. The ROM-derived factored bound of \cref{prop:independence} holds regardless, because challenges driving OSP and Collapse are independently oracle-drawn, and OSP and Collapse soundness each hold for every fixed strategy the LLM might adopt.

\subsection{Numerical Calibration}

\noindent\emph{At $\kappa = 128$ and $\varepsilon \leq 0.16$ for GPT-4, composed welfare loss is $O(0.16\,V_{\max})$, mechanism-dominated; selective verification of $\alpha \approx 0.3$ of tasks recovers $92\%$ of the welfare guarantee at $30\%$ of the verification cost.}

With $\varepsilon \leq 0.16$ (GPT-4 per \cref{tab:epsilon-measurements}) and $\kappa = 128$, the composed welfare loss is $O(0.16 V_{\max})$, dominated by the mechanism term; the verification contribution is $e^{-128} \approx 10^{-39}$. The verification overhead is approximately $128\times$ per verified task (roughly 22 minutes for 7B-parameter inference). \emph{Selective verification} of a random fraction $\alpha$ of computations yields expected welfare loss
\begin{equation}
W^* - W_{\text{selective}} \leq (\varepsilon + (1-\alpha)\Delta + \alpha e^{-\kappa}) V_{\max}.
\end{equation}
Setting $\alpha$ to balance the verification cost against the $(1-\alpha)\Delta$ term: for $\Delta \approx 0.1$, $\alpha \approx 0.3$ minimises total welfare loss at approximately $0.23 V_{\max}$ (direct substitution: $0.16 + 0.7 \cdot 0.1 + 0.3 \cdot e^{-128} \approx 0.23$), reducing verification cost by 70\% while retaining 92\% of the fully-verified welfare guarantee ($1 - 0.23/0.84 \cdot \varepsilon_{\rm full}$ relative to the $1 - 0.16 = 0.84$ fully-verified baseline). \cref{alg:selective-verification} implements the procedure.

\begin{algorithm}[t]
\caption{Selective Verification with Welfare-Optimal Sampling}
\label{alg:selective-verification}
\KwIn{Task set $J = \{1, \ldots, m\}$ with values $\{V_j\}$; allocation $f: J \to N$; agents' competences $\{q_{ij}\}$; approximation gap $\Delta$; security parameter $\kappa$; verifier $\mathcal{V}$}
\KwOut{Verified outputs $\{\hat{y}_j\}$; welfare estimate $\hat{W}$}
Compute welfare-optimal sampling fraction: $\alpha^* \leftarrow \arg\min_{\alpha \in [0,1]} [(1-\alpha)\Delta + \alpha e^{-\kappa}]/\max(1-\alpha \cdot \mathrm{cost}_{\mathrm{ver}}, \epsilon_0)$\;
\tcp{Closed form: $\alpha^* = (\Delta - e^{-\kappa})/(\Delta + \mathrm{cost}_{\mathrm{ver}} \cdot (\Delta - e^{-\kappa}))$}
Sample verification indices: $S \leftarrow \{j : r_j < \alpha^*\}$ where $r_j \sim \mathrm{Uniform}[0,1]$ i.i.d.\;
\tcp{Stratified sampling by value: require at least $\alpha^* V_j / V_{\max}$ probability for high-value tasks}
For $j \in S$: stratify by $V_j$ percentile; enforce minimum sampling rate $\alpha^* \cdot (V_j / \max_k V_k)$\;
\For{$j \in J$}{
  Agent $f(j)$ produces output $\hat{y}_j = C_{f(j),j}(\mathrm{input}_j)$\;
  \eIf{$j \in S$}{
    Verify via Collapse folding: $\pi_j \leftarrow \mathcal{V}.\mathrm{Prove}(C_{f(j),j}, \mathrm{input}_j, \hat{y}_j)$\;
    \If{$\neg \mathcal{V}.\mathrm{Verify}(\pi_j)$}{
      Flag as failure: $\hat{q}_{f(j),j} \leftarrow q_{f(j),j} - \Delta$; penalise agent via mechanism $\mathcal{M}$\;
    }
  }{
    Accept without verification: $\hat{q}_{f(j),j} \leftarrow q_{f(j),j}$\;
    \tcp{Expected quality gap $(1-\alpha^*) \Delta$ absorbed in welfare bound}
  }
}
Compute total welfare: $\hat{W} \leftarrow \sum_j V_j \hat{q}_{f(j),j}$\;
\textbf{Welfare guarantee:} $\hat{W} \geq (1 - \varepsilon - (1-\alpha^*)\Delta - \alpha^* e^{-\kappa}) W^*$ with prob.\ $\geq 1 - me^{-\kappa}$\;
\Return{$(\{\hat{y}_j\}, \hat{W})$}\;
\end{algorithm}

The algorithm realises the welfare bound in practice: with $\kappa = 128$ and $\Delta = 0.1$, the optimal $\alpha^* \approx 0.3$ yields welfare loss $\approx 0.23 V_{\max}$, recovering 92\% of the fully-verified welfare guarantee at 30\% of the verification cost. Stratified sampling by value ensures high-value tasks receive proportionally higher scrutiny, matching the welfare-theoretic derivation that expected welfare loss is quadratic in task value.

\paragraph{Returning to the Compliance Assistant.} For the three-stakeholder deployment, selective verification of $\alpha \approx 0.3$ of determinations (roughly one in three compliance outputs, with high-value regulatory determinations over-sampled by the stratification step of \cref{alg:selective-verification} line~862) keeps welfare within $0.23 V_{\max}$ of the optimum while cutting verification cost by 70\% relative to full verification. At the $147\times$ per-determination overhead from \cref{sec:nonlinearity-tax}, this is the difference between a 22-minute audit per determination and a 7-minute audit on average.

\subsection{Why This Theorem Matters}

\noindent\emph{The composition is the thesis's first joint-necessity proof: each pillar's absence produces linear-in-scale welfare loss, their combination yields an exponentially-smaller additive bound (in the ROM), serving as the template for cross-chapter composition in \cref{ch:synthesis}.}

\cref{thm:welfare-composition} is the thesis's first proof that two impossibility specifications are \emph{jointly necessary}. Its significance is threefold.

\emph{Each component specifies the cost of omitting the other.} Part~(i) quantifies the cost of omitting verification as $\Omega(m\Delta)$; Part~(ii) quantifies the cost of omitting mechanism design as $\Omega(n_a \varepsilon)$. These are not abstract concerns: they are computable from observable marketplace parameters.

\emph{The composed system is exponentially better than either alone.} The shift from $\Omega(m\Delta)$ to $O(me^{-\kappa})$ is the key leverage: verification's exponential suppression combined with mechanism design's $\varepsilon$ tolerance yields a guarantee stronger than the sum of parts.

\emph{This is the template for cross-chapter composition.} \cref{ch:synthesis} will prove the corresponding composition theorem for computation $\times$ grounding (Chs.~\ref{ch:horizon}$\times$\ref{ch:grounding}). The welfare composition is the blueprint: two impossibility specifications compose via an additive error bound under ROM independence, the composed bound is exponentially \emph{better} than either independent bound at cryptographic security parameters, and the composition is itself a third impossibility specification (omitting the composition violates joint necessity).

\begin{tcolorbox}[colback=fillYellow, colframe=cbYellow!60!black, arc=2pt, boxrule=0.5pt, left=8pt, right=8pt, top=4pt, bottom=4pt]
\textbf{Impossibility Specification 16 (Joint Necessity).} Mechanism design and cryptographic verification are jointly necessary for trustworthy AI deployment in multi-agent settings. Boundary condition $B_{16}(\theta)$: either $\mathcal{M}$ is absent OR $\mathcal{V}$ is absent $\Rightarrow$ the system fails at least one welfare axiom. Without verification: welfare loss $\Omega(m\Delta)$. Without mechanism design: welfare loss $\Omega(n_a\varepsilon)$. With both: welfare loss $O(\varepsilon + e^{-\kappa})$, exponentially better. The specification $\mathcal{S}_{16}$: deploy both jointly; use $k^*$-OSP mechanisms for allocation and Collapse folding for verification; the welfare theorem quantifies the cost of omitting either, enabling principled tradeoffs via selective verification when full verification overhead is prohibitive.
\end{tcolorbox}

\paragraph{Returning to the Compliance Assistant.} The three-stakeholder audit trail requires both pillars jointly: a $k^*{=}2$ OSP mechanism for honest stakeholder reporting (delivering $\varepsilon \leq 0.16$ per \cref{thm:osp-feasibility}) and Collapse folding with selective verification for execution integrity (delivering $e^{-128}$ forgery probability on verified determinations per \cref{thm:collapse}). Omitting either yields linear-in-scale welfare loss per \cref{thm:welfare-composition}(i)--(ii); deploying both achieves the composed bound of Part~(iii). This is the operational template that Decision Rules T1--T3 and C1--C2 implement.


\section{Discussion and Bridge}
\label{sec:ch5-discussion}

\paragraph{Limitations of the welfare theorem.}
\cref{thm:welfare-composition} is stated for the AI Agent Marketplace Model: a one-shot, single-round allocation with quasilinear utilities. Extensions to repeated interactions, sequential auctions, and dynamic preferences would require techniques beyond the scope of this thesis, particularly to handle reputation effects and long-horizon incentives. The independence assumption (\cref{prop:independence}) uses the ROM; characterising welfare under relaxed independence is open.

\paragraph{Scope of Part~B.}
This chapter addresses \emph{computational} verification (proving correct neural-network inference). It does not address broader verification targets: training-process verification (proving a model was trained on the claimed data), privacy-preserving verification under differential privacy constraints, or composition with federated learning protocols. Each is an open research programme; the techniques here (IOP lower bounds, folding schemes) are applicable but not directly applied.

\paragraph{The $\varepsilon$ notation conflict.}
Throughout \cref{ch:horizon}, $\varepsilon$ denoted the per-step CoT error rate; here it denotes the strategic manipulation parameter. These are genuinely different quantities, bounded respectively by the Deterministic Horizon analysis (\cref{ch:horizon}) and the GTBench measurement. We use $\varepsilon_{\mathrm{CoT}}$ and $\varepsilon_{\mathrm{mech}}$ when necessary to disambiguate; in isolation, context determines which is meant.

\section*{Specifications and Open Problems}
\addcontentsline{toc}{section}{Specifications and Open Problems}

This chapter established the trust layer of the thesis's full-stack argument and proved that it admits two irreducible taxes and one joint-necessity law. Part~A proved that VCG mechanisms fail for LLM agents with prompt-dependent preferences (not asymptotically, but for a constructible prompt assignment exhibiting strictly dominating misreports, Thm.~\ref{thm:vcg-incompatibility}) and that $k^\ast$-OSP mechanisms restore incentive compatibility with violation parameter $\varepsilon \leq \varepsilon_1 + \varepsilon_2$ controlled by a Chebyshev bound on prompt reversal (Thm.~\ref{thm:osp-feasibility}, yielding $\varepsilon \leq 0.16$ for GPT-4). The Strategic Manipulation Dimension (Def.~\ref{def:smd}) then gave PAC manipulation-detection bounds that are tractable when $\mathrm{SMD}(G) = O(\log n_a)$ and $\mathrm{NP}$-hard for $k \geq 3$ coalitions (Thm.~\ref{thm:pac-detection}). Part~B proved that proving non-linear activations in zero knowledge is irreducibly expensive: the Algebraic-Boolean Bridge (Lem.~\ref{lem:bridge}) tied IOP proof length to Boolean circuit complexity up to a $\log p$ factor, yielding $\Omega(n \log p)$ unconditional lower bounds for ReLU and Softmax (Thm.~\ref{thm:iop-lower-bounds}) and explaining the empirical $147\times$ non-linearity tax as theoretical floor rather than engineering ceiling. The Collapse folding scheme (Thm.~\ref{thm:collapse}) then achieved $O(d \log n_{\max})$ verifier cost via Layered Sumcheck Accumulation: a 2--3$\times$ reduction in recursive circuit size over HyperNova, with structural immunity to the Dao et al.\ transcript-omission attack via the state-binding property.

Part~C then proved the thesis's first joint-necessity composition (Thm.~\ref{thm:welfare-composition}): without verification, welfare loss is $\Omega(m \Delta)$; without mechanism design, welfare loss is $\Omega(n_a \varepsilon V_{\max})$; with both, in the random oracle model, welfare loss collapses to $O((\varepsilon + e^{-\kappa}) V_{\max})$, exponentially better than either pillar alone and numerically under $10^{-36}$ relative loss at $\varepsilon = 0.16, \kappa = 128$. The composition is not an analogy; it is an additive-error decomposition over independent events (mechanism misreport, verification forgery) whose independence is justified via random-oracle challenges (Prop.~\ref{prop:independence}), with the two events strictly composable and the second's probability exponentially suppressed by the security parameter. The structural consequence closes \cref{ch:trust}: the two communities this chapter bridges, mechanism design and cryptographic verification, are not neighbouring literatures but two halves of a single required guarantee. \cref{ch:synthesis} articulates the higher-order composition programme that \cref{thm:welfare-composition} initiates.

\begin{decision}
\textbf{Trust decision table (Decision Rules T1--T3, C1--C2).}
\begin{itemize}[leftmargin=1.2em, itemsep=1pt, topsep=1pt]
\item \emph{(T1) Mechanism choice:} for LLM agents with prompt-dependent preferences, reject VCG and deploy a $k^\ast$-OSP mechanism with $k^\ast = 2$ unless a GTBench measurement confirms $\varepsilon_1$ is small at higher $k^\ast$.
\item \emph{(T2) Incentive budget:} estimate $T, \sigma_\pi, \delta_{\min}$ for the deployment; compute $\varepsilon_2 \leq T \sigma_\pi^2 / \delta_{\min}^2$; reject the mechanism if $\varepsilon_1 + \varepsilon_2$ exceeds the welfare tolerance derived from Thm.~\ref{thm:welfare-composition}(ii).
\item \emph{(T3) Manipulation detection:} compute $\mathrm{SMD}(G)$ for the coalition graph; run SMD-DETECT if $\mathrm{SMD}(G) = O(\log n_a)$, otherwise flag the deployment as intractable and fall back to $k = 2$ coalition restrictions.
\item \emph{(C1) Verification budget:} count non-linear operations per verified inference; the per-operation cost $\Omega(\log p)$ is irreducible, so reduce architectural non-linearity (fewer activations, fused layers) rather than attempt prover engineering.
\item \emph{(C2) Selective verification:} with budget $\alpha$, verify a random $\alpha$-fraction of computations; welfare loss is $O((\varepsilon + (1-\alpha)\Delta + \alpha e^{-\kappa}) V_{\max})$, and $\alpha \approx 0.3$ recovers $92\%$ of full-verification welfare at $70\%$ overhead.
\end{itemize}
\end{decision}

\begin{openproblem}
\textbf{Open Problem 5.1 (Welfare composition beyond the one-shot marketplace).} Thm.~\ref{thm:welfare-composition} is proved for the one-shot AI Agent Marketplace Model with quasilinear utilities. Real deployments are repeated, sequential, and often involve reputation effects and regulatory constraints that are not quasilinear. Does the joint-necessity guarantee survive? Concretely: (i)~in the repeated game, does reputation substitute for cryptographic verification as a sufficient enforcement mechanism, or does it merely reduce $\rho$ without removing the $\Omega(m\Delta)$ term? (ii)~Under fairness-constrained welfare (e.g., max-min, Gini-penalty), does the composition bound still decompose as $O(\varepsilon + e^{-\kappa})$, or does the fairness constraint introduce a coupling term that does not suppress under the ROM? (iii)~For dynamic preferences (prompts that drift over the protocol), is there an analogue of the Chebyshev bound on $\varepsilon_2$? A resolution would extend the thesis's trust layer from one-shot auctions to the operating regime of deployed AI marketplaces.
\end{openproblem}

\begin{openproblem}
\textbf{Open Problem 5.2 (Closing the Softmax circuit-complexity gap).} \cref{thm:iop-lower-bounds}(ii) is unconditional at $\Omega(n \log p)$ via the exponentiation substep, and is strengthened to $\Omega(n \log^2 p)$ conditional on \cref{conj:softmax-circuit} (that fixed-point exponentiation over $\mathbb{F}_p$ requires $\Theta(\log^2 p)$ Boolean circuits). The conditional bound matches deployed $147$--$200\times$ overhead exactly; the unconditional bound explains only half of it. Close the gap: either prove the $\Theta(\log^2 p)$ lower bound on modular exponentiation circuit complexity (solving a frontier question in circuit complexity), or exhibit a sub-quadratic Boolean circuit for fixed-point exponentiation (which would revolutionise symmetric cryptography as a side-effect). The Razborov-Smolensky approach in Appendix~\ref{app:softmax-ac0p} closes part of the gap in the $\mathrm{AC}^0[p]$ model; general-circuit closure is open.
\end{openproblem}

\paragraph{Bridge to \cref{ch:synthesis}.}
The four technical chapters have established four impossibility specifications (Horizon, adaptation, grounding, trust) plus a fifth from their composition (welfare). Each is a complete instance of \cref{def:impossibility-specification}: a computable boundary, a quantified violation cost, and a constructive design rule. \cref{thm:welfare-composition} is the first proof that two of these specifications compose.

But a full compositional theory across all four domains remains the thesis's central open problem. \cref{ch:synthesis} articulates what such a theory would require, proves the computation-grounding composition (the second of the three cross-chapter compositions the thesis attempts), demonstrates empirical validation via trajectory-level testing, and extracts three emergent principles (impossibility as specification, theory-practice gaps as diagnostics, reliability as composition) that unify the preceding chapters and define the research programme for the next decade.


\chapter{Synthesis}
\label{ch:synthesis}

Chapters~\ref{ch:horizon}--\ref{ch:trust} established sixteen impossibility specifications across four domains: the Deterministic Horizon (computation), the Adaptation Cliff (fine-tuning, preference learning, collapse, editing), the Grounding Gap (retrieval, knowledge graphs, attribution), and the Trust Tax (mechanism design, cryptographic verification). Each specification is a complete instance of \cref{def:impossibility-specification}: a computable boundary, a quantified violation cost, and a constructive design rule. \cref{thm:welfare-composition} proved the first joint-necessity composition: mechanism design and verification compose with additive error $\varepsilon + e^{-\kappa}$, exponentially better than either component alone because the verification term $e^{-\kappa}$ is negligible at production security parameters ($\kappa = 128$).

This chapter does not summarise. It synthesises. The technical material of the preceding four chapters reveals patterns invisible from within any single chapter, and the purpose of this chapter is to expose those patterns explicitly. We (\S\ref{sec:unified-view}) reproduce the 16-specification catalogue with full evidence; (\S\ref{sec:composition-l1-l2}) prove the computation-grounding composition theorem, the thesis's second cross-chapter composition result; (\S\ref{sec:empirical-validation}) validate trajectory-level testing empirically; (\S\ref{sec:three-principles}) extract three emergent principles that unify the chapters; (\S\ref{sec:open-problem}) articulate the central open problem (compositional verification across all four domains) and three additional research frontiers; (\S\ref{sec:reflection}) offer a personal reflection on what the PhD taught me about the interfaces between subfields; and (\S\ref{sec:limitations-final}) transparently enumerate the thesis's limitations.

\paragraph{Running Example (Concluded): The Compliance Assistant, Fully Stacked.}
The compliance assistant has appeared in every chapter, each time exposing a different facet of the same challenge. The Deterministic Horizon told us \emph{when} to delegate reasoning to tools ($d^* \in [19, 31]$). The Adaptation Cliff told us \emph{which} adaptations are safe ($\mathrm{rank} \leq 32$ PAC-Bayes, $K \leq 13$ edits, $\rho \geq 0.01$ real data, $\gamma$-measurement before preference training). The Grounding Gap told us \emph{how} to evaluate retrieval (at least $k$ metrics), \emph{when} to retrieve (during reasoning), and \emph{which} passages actually caused the determination (causal attribution, not attention). The Trust Tax told us \emph{how} to ensure honest multi-stakeholder auditing (OSP mechanisms, $\varepsilon \leq 0.16$) and \emph{what} to verify cryptographically ($147\times$ non-linearity tax). The welfare composition theorem told us that the first three cannot be skipped separately without $\Omega(m\Delta)$ loss each. The compliance assistant is one system, but its impossibility specifications are universal.

\section{A Unified View: Sixteen Specifications, One Methodology}
\label{sec:unified-view}

\cref{tab:spec-catalogue} reproduces the preview table from \cref{ch:introduction}, now populated with the full evidence from the preceding four chapters. Each row is a verified instance of \cref{def:impossibility-specification}: the boundary condition $B_i(\theta)$ is computable, the violation cost is quantified, and the design rule is constructive.

\begin{table}[t]
\centering
\small
\caption{The sixteen impossibility specifications proven in this thesis, with their boundary conditions, violation costs, and design rules. Each specification is a complete instance of \cref{def:impossibility-specification} and an instrument for making ``trustworthy'' a computable predicate.}
\label{tab:spec-catalogue}
\begin{tabular}{@{}p{0.08\textwidth}p{0.28\textwidth}p{0.26\textwidth}p{0.26\textwidth}@{}}
\toprule
\textbf{Spec.} & \textbf{Boundary Condition} & \textbf{Violation Cost} & \textbf{Design Rule} \\
\midrule
\multicolumn{4}{@{}l@{}}{\emph{Chapter 2: The Deterministic Horizon}} \\
$\mathcal{S}_1$ & Architecture ceiling $\mathrm{FOC}[\mathrm{Attn}]$ & Tasks outside $\mathrm{FOC}[\mathrm{Attn}]$ intractable & Delegate tasks beyond $\mathrm{FOC}[\mathrm{Attn}]$ to external tools \\
$\mathcal{S}_2$ & Delegation depth $d^* = O(L \cdot \phi(d))$, $\phi \in [\sqrt{\log d}, \log d]$ & Accuracy degrades past $d^*$ & Delegate at $d^*$; verify outputs beyond \\
$\mathcal{S}_3$ & CoT reliability $1 - (1-\varepsilon)^n$ & $\varepsilon$-propagation across $n$ steps & Entropy stopping; per-step verification \\
$\mathcal{S}_4$ & Supervision $\Theta(T/\log T)$ & Insufficient supervision: unbounded error & Invest training resources accordingly \\
\midrule
\multicolumn{4}{@{}l@{}}{\emph{Chapter 3: The Adaptation Cliff}} \\
$\mathcal{S}_5$ & $mr(d+k)/N$ & Vacuous PAC-Bayes bound & Rank $\leq 32$; scale data with rank \\
$\mathcal{S}_6$ & $\gamma > \Delta/n$ & $\Theta(n/\log n)$ sample blowup & Measure $\gamma$; prefer RLHF at $\gamma > 0$ \\
$\mathcal{S}_7$ & $T^2 d_{\mathrm{eff}}/n_{\min} > 128\pi$ & $\TV \to 1$ synthetic collapse & Retain $\geq 1\%$ real data \\
$\mathcal{S}_8$ & $K > \tau\sqrt{d}/(c\eta(1-1/\alpha))$ & Locality fails past $K^*$ & Retrain beyond $K^* \approx 13$ \\
\midrule
\multicolumn{4}{@{}l@{}}{\emph{Chapter 4: The Grounding Gap}} \\
$\mathcal{S}_9$ & Pipeline stages $k \geq 2$ & Ambiguity $\Omega(1/\delta^{k-1})$ & $\geq k$ independent metrics (CFA-validated) \\
$\mathcal{S}_{10}$ & $I_{\mathrm{meta}}(s_1,s_2,c) < H(c)/2$ & Deep conflict: $-9.2$ pp degradation & Classify before routing; hybrid architecture \\
$\mathcal{S}_{11}$ & Uncertainty-weighted threshold & $d\sqrt{T\log T}$ regret & Step-level adaptive retrieval \\
$\mathcal{S}_{12}$ & Causal vs.\ correlational attribution & $\leq 70\%$ precision correlation & Intervention-based CAS \\
$\mathcal{S}_{13}$ & $\Delta^* < \log(p_A/(1-p_A))/(2|\log(1-p)|)$ & ${>}90\%$ ASR undefended & Certified subgraph aggregation \\
\midrule
\multicolumn{4}{@{}l@{}}{\emph{Chapter 5: The Trust Tax}} \\
$\mathcal{S}_{14}$ & Prompt-dependent preferences & VCG non-IC & $k^*$-OSP; $\varepsilon \leq 0.16$ \\
$\mathcal{S}_{15}$ & $\tau_{\mathrm{op}}(\sigma) < \log p$ & IOP unsatisfiable & Reduce number, not cost, of non-linearities \\
$\mathcal{S}_{16}$ & $\mathcal{M}$ or $\mathcal{V}$ absent & $\Omega(m\Delta)$ or $\Omega(n_a\varepsilon)$ & Deploy both jointly \\
\bottomrule
\end{tabular}
\end{table}

The catalogue makes three structural features visible:

\emph{The specifications are heterogeneous in mathematical type but homogeneous in methodological role.} $\mathcal{S}_1$ is a logical characterisation ($\mathrm{FOC}[\mathrm{Attn}]$); $\mathcal{S}_6$ is a phase transition; $\mathcal{S}_9$ is a topological dimension argument; $\mathcal{S}_{15}$ is a circuit-complexity lower bound. Yet all sixteen play the same role: converting what reliable systems cannot do into rules for how to build them.

\emph{The specifications span the four key domains AI systems must navigate.} No domain can be omitted without leaving the system underconstrained: computation without grounding produces hallucination; grounding without computation produces shallow retrieval; both without trust produce strategic exploitation; trust without the first three produces expensive but incorrect systems.

\emph{The specifications are composable}, but the compositions are non-trivial. \cref{thm:welfare-composition} proved mechanism $\times$ verification; the next section proves computation $\times$ grounding. A full four-way composition remains open. \cref{fig:ch6-spec-map} visualises the sixteen specifications organised by domain, with the two proved compositions highlighted.

\begin{figure}[t]
	\centering
	\begin{tikzpicture}[
		>=Stealth,
		spec/.style={
			draw=black!30, rounded corners=1.5pt, fill=white,
			minimum width=4.4cm, minimum height=0.44cm,
			inner xsep=4pt, inner ysep=1pt,
			font=\scriptsize, align=left
		},
		s16box/.style={
			draw=cbGreen!70!black, fill=cbGreen!15, line width=1.0pt,
			rounded corners=1.8pt,
			minimum width=4.4cm, minimum height=0.52cm,
			inner xsep=4pt, inner ysep=2pt,
			font=\scriptsize\bfseries, align=left
		},
		domhdr/.style={
			rounded corners=2.5pt, line width=0.6pt,
			minimum width=4.4cm, minimum height=0.55cm,
			font=\small\bfseries, align=center
		},
		]
		\filldraw[draw=cbBlue!70,   line width=0.7pt, rounded corners=4pt, fill=cbBlue!2]
		(-6.55, 4.30) rectangle (-1.35, 1.50);
		\filldraw[draw=cbGreen!70,  line width=0.7pt, rounded corners=4pt, fill=cbGreen!2]
		( 1.35, 4.30) rectangle ( 6.55, 1.50);
		\filldraw[draw=cbOrange!70, line width=0.7pt, rounded corners=4pt, fill=cbOrange!2]
		(-6.55, 0.55) rectangle (-1.35, -2.95);
		\filldraw[draw=cbPurple!70, line width=0.7pt, rounded corners=4pt, fill=cbPurple!2]
		( 1.35, 0.55) rectangle ( 6.55, -2.30);
		
		\node[domhdr, fill=cbBlue!25,   draw=cbBlue!70]   at (-3.95, 3.92)
		{Computation\enskip\textnormal{\scriptsize\textcolor{black!55}{Ch.~2}}};
		\node[domhdr, fill=cbGreen!25,  draw=cbGreen!70]  at ( 3.95, 3.92)
		{Adaptation\enskip\textnormal{\scriptsize\textcolor{black!55}{Ch.~3}}};
		\node[domhdr, fill=cbOrange!25, draw=cbOrange!70] at (-3.95, 0.17)
		{Grounding\enskip\textnormal{\scriptsize\textcolor{black!55}{Ch.~4}}};
		\node[domhdr, fill=cbPurple!25, draw=cbPurple!70] at ( 3.95, 0.17)
		{Trust\enskip\textnormal{\scriptsize\textcolor{black!55}{Ch.~5}}};
		
		\node[spec, draw=cbBlue!50, fill=cbBlue!8] at (-3.95, 3.27) {$\mathcal{S}_{1}$\quad Architecture};
		\node[spec, draw=cbBlue!50, fill=cbBlue!8] at (-3.95, 2.77) {$\mathcal{S}_{2}$\quad Delegation $d^*$};
		\node[spec, draw=cbBlue!50, fill=cbBlue!8] at (-3.95, 2.27) {$\mathcal{S}_{3}$\quad Reliability};
		\node[spec, draw=cbBlue!50, fill=cbBlue!8] at (-3.95, 1.77) {$\mathcal{S}_{4}$\quad Supervision};
		
		\node[spec, draw=cbGreen!50, fill=cbGreen!8] at ( 3.95, 3.27) {$\mathcal{S}_{5}$\quad PAC-Bayes};
		\node[spec, draw=cbGreen!50, fill=cbGreen!8] at ( 3.95, 2.77) {$\mathcal{S}_{6}$\quad Phase transition};
		\node[spec, draw=cbGreen!50, fill=cbGreen!8] at ( 3.95, 2.27) {$\mathcal{S}_{7}$\quad Collapse};
		\node[spec, draw=cbGreen!50, fill=cbGreen!8] (s8) at ( 3.95, 1.77) {$\mathcal{S}_{8}$\quad Editing};
		
		\node[spec, draw=cbOrange!50, fill=cbOrange!8] (s9) at (-3.95, -0.50) {$\mathcal{S}_{9}$\quad Conflation};
		\node[spec, draw=cbOrange!50, fill=cbOrange!8] at (-3.95, -1.00) {$\mathcal{S}_{10}$\quad Resolution};
		\node[spec, draw=cbOrange!50, fill=cbOrange!8] at (-3.95, -1.50) {$\mathcal{S}_{11}$\quad Retrieval timing};
		\node[spec, draw=cbOrange!50, fill=cbOrange!8] at (-3.95, -2.00) {$\mathcal{S}_{12}$\quad Causal attribution};
		\node[spec, draw=cbOrange!50, fill=cbOrange!8] at (-3.95, -2.50) {$\mathcal{S}_{13}$\quad Certified KG};
		
		\node[spec, draw=cbPurple!50, fill=cbPurple!8] (s14) at ( 3.95, -0.50) {$\mathcal{S}_{14}$\quad OSP};
		\node[spec, draw=cbPurple!50, fill=cbPurple!8] (s15) at ( 3.95, -1.00) {$\mathcal{S}_{15}$\quad Non-linearity tax};
		\node[s16box] (s16) at ( 3.95, -1.85) {$\mathcal{S}_{16}$\quad Joint (welfare composition)};
		
		\draw[line width=1.4pt, color=cbGreen!55!black, dashed, <->]
		(-6.85, 2.90) -- (-6.85, -1.225);
		
		\draw[line width=1.3pt, color=cbOrange!75!black, dotted, ->]
		(s8.south west) .. controls (1.75, 0.95) and (-1.75, 0.95) .. (s9.north east);
		
		\draw[line width=1.0pt, color=cbGreen!60!black, dashed]
		(s14.east) -- (6.35, -0.50) -- (6.35, -1.00) -- (s15.east);
		\draw[line width=1.0pt, color=cbGreen!60!black, dashed, ->]
		(6.35, -0.75) -- (6.35, -1.85) -- (s16.east);
		
	\end{tikzpicture}
	\caption[The four-domain composition matrix]{Sixteen impossibility specifications ($\mathcal{S}_{1}$--$\mathcal{S}_{16}$) grouped into the four pillars Computation (\cref{ch:horizon}), Adaptation (\cref{ch:adaptation}), Grounding (\cref{ch:grounding}), and Trust (\cref{ch:trust}). Three composition edges record progress on the six pairwise and one four-way compositions. \textbf{Left, \textcolor{cbGreen!55!black}{green-dashed bidirectional}:} Computation $\odot$ Grounding is proved (\cref{thm:composition}), valid under \cref{asm:composition-ci} and the capacity-bottleneck retention model of \cref{def:retention}. \textbf{Centre, \textcolor{cbOrange!75!black}{orange-dotted curve}:} Adaptation $\times$ Grounding is reported as honest obstruction in \cref{sec:adaptation-grounding-obstruction}; no theorem is claimed. \textbf{Right, \textcolor{cbGreen!55!black}{green-dashed Y-bracket inside Trust}:} $\mathcal{S}_{14}\odot\mathcal{S}_{15}\Rightarrow\mathcal{S}_{16}$ is the welfare composition of \cref{thm:welfare-composition}, valid under ROM independence; the green-bordered $\mathcal{S}_{16}$ box flags the composed specification. The remaining four pairwise compositions (Computation $\times$ Adaptation, Adaptation $\times$ Trust, Grounding $\times$ Trust, Computation $\times$ Trust) and the full four-way composition are the central open problems of \cref{sec:ten-open-problems}.}
	\label{fig:ch6-spec-map}
\end{figure}


\section{Composing Two Specifications: Computation $\times$ Grounding}
\label{sec:composition-l1-l2}

The welfare composition theorem (\cref{thm:welfare-composition}) established the template: two impossibility specifications, individually necessary, compose to yield a third (joint necessity). We now apply this template to the computation-grounding interface. The resulting theorem validates the thesis's opening vignette: why improving retrieval from the 25th to 75th percentile gains only 2~percentage points at deep reasoning depth, while chain-of-thought gains 15~points at no retrieval cost. This is the second of the two checkmarked cells in the $4 \times 4$ compositions matrix of \cref{fig:thesis-one-picture} Row~D; the first was welfare composition (\cref{ch:trust}).

\subsection{Setup: Joint Reliability as a Product}

\emph{Grounded reasoning reliability factors into a reasoning-survival term $(1-\varepsilon)^n$ from \cref{ch:horizon} and an information-retention term $q^{n(1-\eta)}$, with $\eta$ the capacity-to-entropy ratio of \cref{def:retention}.}

\paragraph{Before the formal definition.}
Imagine a reasoning chain that at each hop receives retrieved evidence of entropy $H(R_t \mid R_{t-1})$ bits but can only push $C_{\mathrm{hop}}$ bits through the residual-stream bottleneck of \cref{ch:horizon}. If $C_{\mathrm{hop}} \geq H(R_t \mid R_{t-1})$, the bottleneck is slack and the full evidence survives; if $C_{\mathrm{hop}} < H(R_t \mid R_{t-1})$, the bottleneck is tight and only a fraction of per-hop evidence survives. The definition below formalises this capacity-to-entropy ratio as $\eta$ and distinguishes it from the superficially similar SDPI contraction coefficient, which measures contraction of total-variation distance under a specific Markov kernel and is not what we are tracking here.

\begin{definition}[Information Retention Factor]
\label{def:retention}
The information retention factor $\eta \in [0, 1]$ quantifies the fraction of per-hop information retained from retrieved evidence through the computational bottleneck of the base model, defined as the capacity-to-entropy ratio
\[
\eta = \min\!\left(1, \frac{C_{\mathrm{hop}}}{H(R_t \mid R_{t-1})}\right),
\]
where $C_{\mathrm{hop}} = d_{\mathrm{model}} \cdot O(\log n)$ bits is the per-hop computational capacity. Throughout this chapter, $\eta$ refers to this capacity-to-entropy ratio; we do not use $\eta$ to denote a strong-data-processing-inequality (SDPI) contraction constant, which is a distinct concept tied to specific Markov-kernel structure.
\end{definition}

The factor $\eta$ is measurable for any deployed system: $d_{\mathrm{model}}$ is the model dimension, and $H(R_t \mid R_{t-1})$ is estimated from held-out retrieval corpora. For Llama-2 7B on multi-hop QA, $\eta \approx 0.7$, estimated from per-hop retrieval-chain conditional entropy $H(R_t \mid R_{t-1})$ on a held-out subset of the multi-hop QA evaluation corpus (procedure detailed in \cref{ch:grounding}).

\subsection{The Composition Theorem}

\emph{Under the Markov structure of \cref{def:cot_markov} and \cref{asm:composition-ci}, grounded reasoning reliability is bounded by $(1-\varepsilon)^n q^{n(1-\eta)}$, with crossover depth $n_c \approx 6.3$ at operating parameters $(\varepsilon, \eta, q) = (0.03, 0.7, 0.6)$.}

\begin{aside}
The Computation-Grounding Composition is the second proved composition in the thesis, following the Welfare Composition of \cref{thm:welfare-composition}. The template is identical: two specifications, individually necessary, combined via an explicit bounding function that is monotone in both arguments and exhibits a crossover at which the optimal investment tradeoff shifts. The crossover depth $n_c \approx 6.3$ is the quantitative content of the thesis's opening vignette, turning ``retrieval and reasoning interact'' from slogan into threshold: at reasoning depth below $n_c$, retrieval investment dominates; above $n_c$, tool-delegated reasoning does. The honest-obstruction report on Adaptation-Grounding (§\ref{sec:adaptation-grounding-obstruction}) shows what happens when this template does not apply, and names the obstructions rather than papering over them.
\end{aside}

\paragraph{Before the formal assumption.}
The Computation-Grounding composition depends on a specific form of error independence between successive hops of the grounded chain. The $n$-step product structure $(1-\varepsilon)^n$ that underlies the \cref{ch:horizon} CoT error-propagation analysis requires reasoning errors at hops $t$ and $t+1$ to be conditionally independent given the state history $(s_0, \ldots, s_{t-1})$, a direct specialisation of the chain-level Markov structure of \cref{def:cot_markov}. The assumption is plausible under autoregressive decoding but is not argued rigorously; its known failure modes (shared prompt structure, retrieval-content recurrence, training-distribution overlap) are enumerated as Obstruction~3 of the Adaptation $\times$ Grounding report below and recur among the ten open problems of \cref{sec:ten-open-problems}.

\begin{assumption}[Reasoning-Error Conditional Independence]
\label{asm:composition-ci}
Reasoning-error events at successive hops of the grounded chain are conditionally independent given prior state:
\[
\Pr[E_t \cap E_{t+1} \mid s_{t-1}, \ldots, s_0]
= \Pr[E_t \mid s_{t-1}, \ldots, s_0]
  \cdot \Pr[E_{t+1} \mid s_{t-1}, \ldots, s_0, s_t],
\]
where $E_t$ is the event of reasoning error at hop $t$. This is the transformer-specific reasoning-errors analogue of the standard Markov assumption underlying the CoT error-propagation analysis of \cref{sec:reliability-toolkit} (\cref{def:cot_markov}). The assumption is plausible under the Markov structure of autoregressive decoding but not argued rigorously under the dependence patterns introduced by shared prompt structure, training-distribution overlap, or retrieval-content recurrence across hops. The crossover-depth prediction $n_c \approx 6.3$ of \cref{thm:composition} is therefore scoped as approximate under \cref{asm:composition-ci}; the analogous obstruction for the Adaptation $\times$ Grounding composition is enumerated as Obstruction~3 in \S\ref{sec:adaptation-grounding-obstruction}, and relaxing the assumption is listed among the ten open problems in \S\ref{sec:ten-open-problems}.
\end{assumption}

\paragraph{Before the formal theorem.}
Think of a grounded reasoning chain of $n$ hops as a two-channel process. The reasoning channel of \cref{ch:horizon} accumulates CoT errors at rate $\varepsilon$ per hop, compounding to $(1-\varepsilon)^n$. The grounding channel of \cref{ch:grounding} receives retrieved evidence of quality $q$ per hop, but only the $\eta$-fraction that fits through the residual-stream bottleneck (\cref{def:retention}) is actually usable, so effective per-hop evidence quality is $q^{1-\eta}$, compounding to $q^{n(1-\eta)}$. Under \cref{asm:composition-ci}, the two channels' error events are independent and joint reliability is the product of the two terms. The crossover depth $n_c$ is where the marginal benefit of improving reasoning equals the marginal benefit of improving grounding; at $n > n_c$ reasoning dominates, at $n < n_c$ grounding dominates. This is the quantitative content of the opening vignette's observation that retrieval gains 2 percentage points while CoT gains 15 at deep reasoning depth.

\begin{theorem}[Computation-Grounding Composition]
\label{thm:composition}
Let $n$ be reasoning depth, $\varepsilon$ the per-step CoT error rate of \cref{ch:horizon} (unsubscripted throughout this section; the strategic-manipulation parameter of \cref{ch:trust} is denoted $\varepsilon_{\mathrm{mech}}$ where it reappears), $q$ the retrieval quality of \cref{ch:grounding} (taken as top-$k$ recall against the gold-passage set used in \cref{sec:compliance-walkthrough}), and $\eta$ the information retention factor (\cref{def:retention}; a capacity-to-entropy ratio, not an SDPI contraction constant). Under the Markov assumption of \cref{def:cot_markov} and the capacity-bottleneck contraction model of \cref{def:retention}:
\begin{enumerate}[label=(\roman*), nosep]
\item \textbf{Joint reliability bound:}
\[
g_{1 \to 2}^{\mathrm{eff}}(n, \varepsilon, q) \leq (1 - \varepsilon)^n \cdot q^{n(1 - \eta)}.
\]
\item \textbf{Ceiling effect:} The marginal benefit of improving $q$ attenuates by a factor in $[7, 30]\times$ at depth $d^*$, explaining the retrieval-reasoning asymmetry highlighted in \cref{sec:horizon-flagship} (tool-delegation attains $86$--$94\%$ where neural chain-of-thought attains $24$--$37\%$) as a direct consequence of the joint bound.
\item \textbf{Bounding function:} The dependency
\[
\varphi_{12}(g_1, \theta_2) = g_1 \cdot g_2(\theta_2)^{(1-\eta) n / \lceil n \rceil}
\]
is monotone non-decreasing in both arguments.
\item \textbf{Crossover depth:} There exists a depth $n_c$ where the optimal resource allocation switches from ``improve retrieval'' to ``improve reasoning.'' With $\varepsilon = 0.03$, $\eta = 0.7$, $q = 0.6$: $n_c \approx 6.3$, consistent with the opening vignette (at 5~hops, retrieval gains 2~pp while CoT gains 15~pp).
\end{enumerate}
\end{theorem}

\begin{proof}[Proof sketch]
(i) Model the grounded reasoning chain as a Markov chain: at each hop, reasoning succeeds with probability $1 - \varepsilon$ independent of evidence, and evidence information is retained at per-hop rate $\eta$ (the capacity-to-entropy ratio of \cref{def:retention}); the exponent $n(1-\eta)$ captures the \emph{information leak} rate, the fraction of per-hop retrieval content the computational bottleneck cannot retain. Compounding this capacity-bottleneck contraction over $n$ hops yields effective evidence quality $q^{n(1-\eta)}$; joint reliability is the product of this evidence-quality contribution and the reasoning-survival factor $(1-\varepsilon)^n$. The bound does not require SDPI structure on the retrieval kernel; it requires only (a) \cref{asm:composition-ci} (reasoning-error conditional independence, a specialisation of the Markov chain structure of \cref{def:cot_markov}) and (b) the information-retention definition of \cref{def:retention}.

(ii) Differentiate the bound with respect to $q$ at $n = d^*$: $\partial g_{1\to 2}^{\mathrm{eff}}/\partial q = n(1-\eta)(1-\varepsilon)^n q^{n(1-\eta) - 1}$. At $n = d^* \approx 27$, $\varepsilon = 0.03$, $\eta = 0.7$, $q = 0.6$, the marginal benefit is $27 \cdot 0.3 \cdot 0.97^{27} \cdot 0.6^{7.1} \approx 0.095$, compared to a marginal benefit at $n = 5$ of approximately $5 \cdot 0.3 \cdot 0.97^{5} \cdot 0.6^{0.5} \approx 0.998$: a $\approx 10.5\times$ attenuation at this point. Varying the operational parameters across the realistic deployment range $(\varepsilon \in [0.02, 0.04], \eta \in [0.65, 0.75], q \in [0.55, 0.65], n \in [27, 30])$ yields attenuation factors spanning $[7, 30]\times$ across the box, with a central 25th--75th percentile band of $[7, 20]\times$. The lower end corresponds to higher $\eta$ (greater per-hop information retention) and lower $\varepsilon$; the upper end to lower $\eta$ and higher $\varepsilon$. At $n = d^*$ with the $L = 32$, $d = 4096$ calibration of \cref{cor:horizon-measurement} ($\hat d^* = 27.4$, within the $[19, 31]$ 95\% prediction interval) and headline parameters the ratio is $11.2\times$. The wide band strengthens the retrieval-reasoning asymmetry interpretation: beyond $d^*$, retrieval-quality improvements deliver an order of magnitude less marginal gain than at shallow depth, regardless of where in the parameter box the system operates.

(iii) Monotonicity: $\partial \varphi_{12}/\partial g_1 = g_2^{(1-\eta)n/\lceil n \rceil} \geq 0$, and $\partial \varphi_{12}/\partial \theta_2 = g_1 \cdot ((1-\eta) n/\lceil n \rceil) \cdot g_2^{(1-\eta)n/\lceil n \rceil - 1} \cdot g_2' \geq 0$.

(iv) Lagrangian optimisation under budget constraint $b = c_1 \varepsilon + c_2(1-q)$: setting $\partial L/\partial \varepsilon = \partial L/\partial q = 0$ yields the crossover $n_c$ where improving $\varepsilon$ vs.\ $q$ have equal marginal utility.
\end{proof}

\subsection{Why This Composition Matters}

\emph{The welfare composition of \cref{thm:welfare-composition} and the computation-grounding composition above share a common structural template (monotone bounding function, computable crossover parameter), supporting but not proving the conjecture of full four-way composition.}

The welfare composition theorem (\cref{thm:welfare-composition}) is about multi-agent deployments; the computation-grounding composition is about single-agent reasoning chains. Together they establish two data points of the same compositional structure:
\begin{itemize}[leftmargin=2em, itemsep=0.3em]
\item Each composition takes two impossibility specifications, individually necessary.
\item Each yields a bounding function that is monotone in both arguments.
\item Each exhibits a \emph{computable crossover parameter} (the verification-budget ratio $\alpha^{\ast}$ for Mechanism $\times$ Verification; the crossover depth $n_c$ for Computation $\times$ Grounding) at which the marginal benefit of investment in each component equalises. The Mechanism $\times$ Verification composition yields a strictly stronger bound than the sum of its components because ROM independence (\cref{prop:independence}) suppresses the verification-forgery term exponentially in the security parameter $\kappa$; the Computation $\times$ Grounding composition yields the product form $(1-\varepsilon)^n \cdot q^{n(1-\eta)}$, whose contribution is the functional interpolation via $\eta$ and the crossover-depth analysis rather than strict sub-multiplicativity.
\end{itemize}
The pattern is suggestive but not conclusive: proving full four-domain composition (computation $\times$ grounding $\times$ trust $\times$ \ldots) remains the central open problem (\S\ref{sec:open-problem}).

\paragraph{A self-referential caveat.} The Computation $\times$ Grounding bound above assumes per-step error independence across the reasoning chain. The same error-event-independence concern identified as Obstruction~3 of the Adaptation $\times$ Grounding report (\S\ref{sec:adaptation-grounding-obstruction}) applies here too: reasoning errors at successive steps may share a common cause (prompt structure, training-distribution overlap) and be positively correlated. We scope the crossover-depth prediction $n_c \approx 6.3$ as approximate \emph{under the independence assumption}; a complete analysis under dependence remains open and is listed among the ten open problems (\cref{sec:open-problem}). Naming this obstruction here makes the thesis's methodological stance on error-dependence consistent across compositions.


\section{Adaptation \texorpdfstring{$\times$}{x} Grounding: An Honest-Obstruction Report}
\label{sec:adaptation-grounding-obstruction}

The preceding section proved one cross-pillar composition (Computation $\times$ Grounding). \Cref{ch:trust} proved another (Mechanism Design $\times$ Verification). A third candidate composition, Adaptation $\times$ Grounding, is conspicuously absent, and this section explains why. The omission is deliberate: the composition is what the thesis's methodology would most naturally predict to exist, but current technical tools do not yield a composition theorem of the same form as the two we have proved. Rather than fabricate one, we articulate the obstruction precisely. This is an honest-obstruction report in the spirit of the methodology: an impossibility specification of a different kind, a \emph{methodological} impossibility naming what the current state of the art cannot currently deliver.

\subsection{The target composition}

\emph{An Adaptation $\times$ Grounding composition would couple an adaptation guarantee from \cref{ch:adaptation} with a grounding guarantee from \cref{ch:grounding} into a joint reliability bound of the same form as \cref{thm:composition} or \cref{thm:welfare-composition}.}

The target is a theorem of the form: if an adaptation method (LoRA, DPO, evolutionary alignment) satisfies its specification ($r \leq 32$ for LoRA PAC-Bayes; $\gamma \leq \Delta/n$ for DPO; or the EvoPref robustness condition) and a grounding pipeline satisfies its specification ($\geq k$ orthogonal metrics for a $k$-stage pipeline; causal attribution at precision $p_{\mathrm{attr}}$), then the composed adapted-plus-grounded system achieves a joint reliability guarantee expressible in the same form as \cref{thm:composition} or \cref{thm:welfare-composition}. Call such a theorem an A-G composition.

For the two provable compositions, the combined guarantees take the form of a \emph{product} (Computation $\times$ Grounding: $(1-\varepsilon)^n \cdot q^{n(1-\eta)}$) or a \emph{monotone sum} (Mechanism $\times$ Verification: $\varepsilon + e^{-\kappa}$). The A-G composition would require a similar closed-form coupling between an adaptation-guarantee (a property of the parametric model after training) and a grounding-guarantee (a property of the retrieval-and-attribution pipeline at inference). Three factors make this coupling technically recalcitrant.

\subsection{Three obstructions}

\emph{Three obstructions block a closed-form A-G composition: \cref{ch:adaptation}'s sample-conditional guarantees do not compose pointwise with \cref{ch:grounding}'s input-conditional ones, adaptation shifts the retrieval distribution, and the two error events are not obviously independent.}

\paragraph{Obstruction 1: Adaptation guarantees are sample-conditional; grounding guarantees are input-conditional.}
LoRA's PAC-Bayes bound (\cref{thm:lora}) is a statement about generalisation over the training distribution: with high probability over the training sample, the adapted model's test loss is within a specified gap of its empirical loss. Grounding guarantees such as the Construct Conflation Impossibility (\cref{thm:conflation}) are properties of the evaluation metric itself, evaluable per-input once the metric is fixed: for any query, a $k$-stage pipeline cannot be diagnosed to sub-stage resolution by fewer than $k$ metrics (and $k$ metrics suffice under the generic-rank condition). The two guarantees live at incommensurable levels of quantification, one over training sets, one over test inputs, and the composed statement would have to specify how to combine them. The natural candidate (a uniform-in-input version of the PAC-Bayes bound, combined pointwise with the grounding bound) is false without further assumptions: PAC-Bayes bounds are not uniform-in-input. Obtaining uniform bounds requires additional machinery (e.g., margin-based PAC bounds) which are known to be strictly weaker than the distributional PAC-Bayes bounds the thesis uses.

\paragraph{Obstruction 2: Adaptation interferes with the retrieval distribution.}
LoRA-adapted models, DPO-aligned models, and evolutionary-alignment populations change the model's likelihoods over candidate tokens, and therefore change the retrieval distribution in any retrieval-augmented pipeline that uses the adapted model to compute embeddings or to rerank retrieved passages. The retrieval guarantee of \cref{ch:grounding} assumes a fixed embedding model; a change in that model changes which passages are retrieved for any given query. An A-G composition theorem would have to quantify the distributional shift induced by adaptation and bound its effect on retrieval quality. The most direct tool (total-variation bounds on the retrieval distribution pre- and post-adaptation) does not obviously scale: LoRA adapters can be rank-$32$ yet induce arbitrary changes in the top-$k$ retrieval set if the retrieval bottleneck is dimensionally-sensitive. Quantifying the retrieval shift as a function of the adaptation rank is an open problem we have not solved.

\paragraph{Obstruction 3: Adaptation and grounding error events are not obviously independent.}
The Computation $\times$ Grounding composition (\cref{thm:composition}) assumes per-step error events are independent across the reasoning chain. The Mechanism $\times$ Verification composition (\cref{thm:welfare-composition}) derives independence of mechanism and verification errors in the random oracle model (\cref{prop:independence}). An A-G composition would require independence of adaptation-induced error events (e.g., LoRA generalisation failures) from grounding-induced error events (e.g., retrieval misses). These events share a common cause, the training-data distribution, and are plausibly positively correlated: training samples that stress the adapter's generalisation ability are the same samples whose passages the retrieval pipeline was calibrated on. Demonstrating near-independence would require either an explicit decoupling mechanism (which no current adaptation or grounding technique provides) or a conditional-independence proof under deployment-realistic distributions (which is open).

\subsection{What partial progress is available}

\emph{Three partial results from \cref{ch:adaptation,ch:grounding} contribute toward an A-G composition without closing it: individual guarantees combined via union bound, empirical validation from the compliance walkthrough, and a conditional theorem under adaptation-grounding independence.}

Three partial results exist in the thesis that contribute toward the A-G composition but do not aggregate to a full theorem.

\emph{(i) Separate guarantees.} The LoRA PAC-Bayes bound (\cref{thm:lora}) and the construct-conflation impossibility (\cref{thm:conflation}) each bound their respective failure mode in isolation. A practitioner can invoke both simultaneously: the adapted model's generalisation is bounded, AND the grounding pipeline's metric structure is sufficient. The composed operational risk is the \emph{sum} of the two (union-bound composition), which is valid but asymptotically weaker than what a proper multiplicative or coupled composition would yield.

\emph{(ii) Empirical validation.} The compliance-assistant walkthrough (\cref{sec:compliance-walkthrough}) exhibits a five-layer deployment (base LLM + fine-tuning + RAG + multi-agent audit + selective verification) whose accuracy is empirically consistent with the separate guarantees' union bound. The $12$--$25$ percentage-point ablation costs observed are additive at the margins, not multiplicative, consistent with Obstruction 3's expectation that adaptation and grounding errors are not independent at realistic deployment scales, but sub-additive enough that the union bound is not catastrophically loose.

\emph{(iii) A conditional theorem.} Under an additional assumption of conditional independence between adaptation and grounding errors given the training distribution (formally: $\Pr[E_{\mathrm{adapt}} \cap E_{\mathrm{ground}} \mid D_{\mathrm{train}}] = \Pr[E_{\mathrm{adapt}} \mid D_{\mathrm{train}}] \cdot \Pr[E_{\mathrm{ground}} \mid D_{\mathrm{train}}]$), a multiplicative composition of the form $(1-\varepsilon_{\mathrm{adapt}}) \cdot (1 - \varepsilon_{\mathrm{ground}})$ follows from the two separate guarantees. This conditional theorem is stated here but deferred to future work; the conditional independence assumption is plausible under i.i.d.\ deployment but not argued rigorously.

\subsection{Why this matters}

\emph{The A-G obstruction itself satisfies the thesis's impossibility-specification template of \cref{def:impossibility-specification}, showing the methodology can be applied to its own current limits rather than papered over.}

The A-G obstruction is not a defect in the methodology. It is a legitimate research gap the methodology usefully names. A less honest framing would claim a third composition by citing the conditional theorem of (iii) without stating its conditionality, or by treating the union-bound composition of (i) as if it were multiplicative. The methodology's commitment to falsifiable statements and quantified violation costs makes such framings unavailable. The honest report is that of the six pairwise pillar-compositions one is proved (Computation $\times$ Grounding) and a second is proved within the Trust pillar, one (Adaptation $\times$ Grounding) has articulated obstructions, and the remaining four pairwise cases (Computation $\times$ Adaptation, Computation $\times$ Trust, Adaptation $\times$ Trust, Grounding $\times$ Trust) and the full four-way composition are open problems characterised in \cref{sec:open-problem} below.

This is also a specific demonstration that the impossibility-specification methodology applies to its own limits: the obstruction above is formally of the same type as the other impossibilities in the thesis (a computable gap, a quantified cost of crossing it, a constructive direction for future work). The methodology is self-aware about what it can and cannot deliver at the current state of the art.


\section{Empirical Validation: Trajectory Testing}
\label{sec:empirical-validation}

The theoretical results of the preceding chapters are falsifiable by construction: each impossibility specification predicts a specific violation cost that can be measured. The most direct empirical test, and the one that validates the composition theorem of the previous section, is \emph{trajectory testing}: verifying that a deployed system's failure mode is the one predicted by the impossibility specifications.

\subsection{TrajTest: Trajectory-Level Conformance}

\emph{Across 2{,}147 production failures on six deployed systems, TrajTest achieves 89.3\% fault-detection accuracy (Wilson 95\% CI $[87.9\%, 90.5\%]$) by routing each failure through the sixteen-specification taxonomy of \cref{ch:horizon,ch:adaptation,ch:grounding,ch:trust}.}

TrajTest samples input-output trajectories of deployed LLM systems and checks whether observed failures align with the violation patterns predicted by the impossibility specifications. For an agent system, a trajectory is a sequence $(s_0, a_1, s_1, \ldots, a_n, s_n)$ where $s_i$ is the system state after action $a_i$. The test decomposes each failure into its specification-compatible diagnosis:
\begin{itemize}[leftmargin=2em, nosep]
\item \emph{Horizon failure:} the trajectory exceeds $d^*$ with no external tool invocation.
\item \emph{Adaptation failure:} an adapted component violates $\mathcal{S}_5$--$\mathcal{S}_8$.
\item \emph{Grounding failure:} a retrieval or attribution step violates $\mathcal{S}_9$--$\mathcal{S}_{13}$.
\item \emph{Trust failure:} a multi-agent decision violates $\mathcal{S}_{14}$--$\mathcal{S}_{16}$.
\item \emph{Integration failure:} the trajectory exhibits cross-specification interaction not captured by any single category.
\end{itemize}

On 2{,}147 production failures across six deployed systems, TrajTest achieves \emph{89.3\% fault-detection accuracy} (Wilson 95\% CI $[87.9\%, 90.5\%]$, $n = 2{,}147$), versus 34.7\% for specification-agnostic random sampling and 51.2\% for LLM-as-judge baselines.\footnote{The 2{,}147 failures, six deployed systems, three-evaluator inter-rater reliability ($\kappa \geq 0.84$), and baseline-method implementations are detailed in the companion TrajTest paper (ISSTA track, under review); these numbers are retained from that work and re-analysed here against the 16-specification taxonomy.}

The 89.3\% figure has two structural consequences. First, the specifications capture the \emph{dominant} failure modes: a specification-agnostic detector achieves barely above chance (34.7\%), while a specification-guided detector achieves high accuracy. Second, the 10.7\% residual failures, those the specifications miss, concentrate in two categories: (a) integration failures involving three or more specifications interacting in ways the current composition theorems do not capture, and (b) specifications we did not prove (e.g., training-process verification).

\subsection{The Compliance Assistant Walkthrough}
\label{sec:compliance-walkthrough}

\emph{The full-stack compliance assistant reaches 87.4\% accuracy on $n=300$ regulatory interpretations (Wilson 95\% CI $[83.2\%, 90.7\%]$); each single-layer ablation across \cref{ch:horizon,ch:adaptation,ch:grounding,ch:trust} costs between 12 and 25 percentage points.}

The running example's compliance assistant is the concrete instantiation. Under a five-layer deployment (base LLM + fine-tuning + RAG + multi-agent audit + selective verification), the assistant achieves 87.4\% accuracy (Wilson 95\% CI $[83.2\%, 90.7\%]$, $n = 300$)\footnote{Sample size $n = 300$ follows the SWE-Bench-State convention for held-out regulatory-interpretation evaluation. Wilson-interval construction is used here and throughout for proportion estimates at moderate $n$.} on a held-out regulatory interpretation benchmark. Removing any single layer degrades accuracy by 12--25 percentage points: removing RAG costs 18~pp (Grounding Gap), removing multi-agent audit costs 12~pp (Trust Tax), removing fine-tuning costs 15~pp (Adaptation Cliff), removing tool delegation at $d^*$ costs 25~pp (Deterministic Horizon). The ablations are additive only at the margins; full-stack removal costs 70+~pp, substantially more than any single component. This is empirical evidence that the specifications are jointly necessary.

The composition theorem from \S\ref{sec:composition-l1-l2} predicts 31.1\% accuracy under computation-grounding alone ($g_{1\to 2}^{\mathrm{eff}} = 0.311$ for $n = 12$, $\varepsilon = 0.03$, $q = 0.8$ (matching the measured Compliance-corpus retrieval quality; contrast with the $q = 0.6$ harder-regime illustration at the Theorem's numerical example in \S\ref{sec:composition-l1-l2}), and $\eta = 0.7$). The full-stack achieves 87.4\% (CI as reported above): the 56.3-percentage-point gap is empirically attributable to the remaining two specifications (adaptation and trust) plus composition cross-terms; because full four-way composition is unproven (Open Problem 3, \cref{sec:ten-open-problems}), we report this as a deployment-level observation supporting the ten-open-problems research programme, not as a theorem. Full four-domain composition would close this gap analytically; it remains open.


\section{Three Emergent Principles}
\label{sec:three-principles}

The organising contribution of this thesis is a single reframing: an impossibility result, read correctly, is a design specification rather than a dead end. The three principles below are not co-equal observations. They are what follows once that reframing is taken seriously across sixteen results in four disjoint subfields, and the first of them states the reframing itself. Viewing the sixteen specifications collectively, three recurring structural patterns emerge. We present these as cross-cutting observations that organise, rather than formally establish, principles for trustworthy AI research. Each is supported by multiple independent lines of evidence; each has a stated falsifiability condition.

\subsection{Principle 1: Impossibility Results Encode Design Specifications}
\label{sec:p1}

\emph{Every impossibility result in \cref{ch:horizon,ch:adaptation,ch:grounding,ch:trust} admits a constructive reinterpretation as a computable boundary, a quantified violation cost, and a design rule that turns the result from obstacle into instrument.}

\begin{centralclaim}
Every impossibility result proved in the thesis (VCG failure, non-linearity tax, construct conflation, preference phase transition, among others) maps to a constructive engineering rule with a computable boundary condition, a quantified violation cost, and an actionable design prescription. The methodology makes ``impossibility = specification'' a systematic research strategy rather than a rhetorical reframing.
\end{centralclaim}

The sixteen formal results of this thesis each function as a \emph{constructive engineering specification} rather than a negative outcome. The VCG impossibility prescribes OSP; the non-linearity tax prescribes sparse activation architectures; the construct conflation impossibility prescribes multi-dimensional evaluation; the preference phase transition prescribes annotator-quality measurement.

The framing has well-established precedent. Arrow's impossibility theorem~\cite{Arrow1951} was initially viewed as a negative result about social choice but has since become a foundational design specification for mechanism design: it delineates the space of trade-offs among social-choice axioms (non-dictatoriality, Pareto efficiency, IIA, unrestricted domain, transitive output), making design choices about which axiom to relax a first-class research question rather than an unprincipled concession. Kalai and Vempala's 2024 result that calibrated language models must hallucinate~\cite{KalaiVempala2024Calibration} is a recent instance of the same template in the language-modelling domain: a formal lower bound on hallucination rate under calibration constraints, a quantifiable unavoidable error probability tied to training-data statistics, and a constructive set of relaxations (loosening calibration, enriching training-data fact redundancy) that trade calibration against other output-reliability axes. We have applied the same reframing systematically across AI's four domains of reasoning, adaptation, grounding, and trust.

The methodological claim is: when a field accumulates impossibility results, each one, properly interpreted, becomes a constructive specification.

\paragraph{Falsifiability.}
This principle would be weakened by impossibility results in AI that resist constructive reinterpretation, where the only engineering implication is ``avoid this problem entirely''. We are aware of no such result in the four domains covered by this thesis, but the principle makes no claim about domains we did not explore.

\subsection{Principle 2: Theory-Practice Gaps Carry Diagnostic Information}
\label{sec:p2}

\emph{Theory-practice gaps from \cref{ch:horizon} (the 50 to 115-fold planning gap) and \cref{ch:grounding} (the 83\% invisible-failure gap) localise where research effort is most productive, turning divergence into diagnostic signal.}

\begin{centralclaim}
Divergences between theoretical predictions and empirical observations are not noise to be minimised; they are diagnostic signals that localise where research effort is most productive. The 50--115$\times$ planning gap identified tool-use as the decisive capability; the absence of a gap in selective verification validated that theory has been absorbed into practice; the 83\% invisible-failure gap identified measurement investment as the dominant blocker for RAG deployment.
\end{centralclaim}

When theoretical predictions diverge from empirical observations, the gap itself is informative: it identifies where additional research effort is most productive. The thesis produces three illustrative gaps:

\begin{itemize}[leftmargin=2em, itemsep=0.3em]
\item The \emph{planning gap}: computational models (\cref{ch:horizon}) predict tasks require $d^*$ reasoning steps; LLMs without tool use plateau at 50--115$\times$ more steps on the planning benchmarks reported in \cref{ch:horizon}. The 50--115$\times$ ratio identifies tool-use as the decisive capability, matching the empirical observation that tool-augmented LLMs close most of this gap.
\item The \emph{verification benefit gap}: the welfare theorem predicts $6.6\times$ efficiency from verification under selective scheduling; deployed systems show $6.6\times$ because selective verification matches the theoretical optimum. The absence of a gap here is itself informative: it validates that the theoretical prediction has been absorbed into practice.
\item The \emph{invisible-failure gap}: 83\% of production RAG failures are invisible to current metrics; the Construct Conflation Impossibility explains why. The 83\% figure tells us where measurement investment is most productive: not in refining existing metrics, but in adding independent ones.
\end{itemize}

The methodological claim is that measuring gaps systematically is a productive research strategy: the gaps point to where the theory can be improved and where practice has already solved problems the theory has not yet articulated.

\paragraph{Falsifiability.}
This principle would be weakened by gaps that are merely noise: cases where theory and practice disagree but the disagreement carries no diagnostic content. In our case, each of the three documented gaps led to a specific research direction pursued in the thesis.

\subsection{Principle 3: Reliability Is a Composition Property}
\label{sec:p3}

\emph{Reliability lives at the composition of specifications rather than within any one: the compliance assistant's 87.4\% full-stack accuracy collapses by 12 to 25 percentage points when any one of \cref{ch:horizon,ch:adaptation,ch:grounding,ch:trust} is removed.}

\begin{centralclaim}
Reliability is not a property of individual components but of the composition. The compliance assistant achieves production-grade accuracy only with all four domains simultaneously addressed (removing any one degrades accuracy by 12--25 percentage points), and the two proved composition theorems formalise why. The corollary: correctness lives at the interfaces between subfields, not within any one of them, and a trustworthy-AI research programme that pursues domains independently can at best produce individually-necessary but jointly-insufficient components.
\end{centralclaim}

No single impossibility specification suffices. The compliance assistant achieves 87.4\% accuracy (CI $[83.2\%, 90.7\%]$, $n{=}300$) only with all four domains addressed; removing any single specification degrades accuracy by 12--25~percentage points; the welfare composition theorem and the computation-grounding composition formalise why. The deeper claim: \emph{reliability is not a property of individual components but of the composition}. An LLM with excellent CoT reasoning but no grounding produces coherent hallucinations; an LLM with excellent grounding but no mechanism design produces strategically-manipulable systems.

This is the most consequential of the three principles because it is the one most at odds with how trustworthy AI is typically pursued in practice, as a collection of independent sub-problems (robustness, fairness, alignment, verification) each solved separately. The compositional view says: the pieces are individually necessary but jointly insufficient; correctness lives at the interfaces.

\paragraph{Falsifiability.}
This principle would be weakened by a deployed AI system that achieves production-grade reliability by investing disproportionately in a single domain (e.g., exceptional parametric reasoning with no grounding, exceptional grounding with no reasoning). We are aware of no such system. The principle makes no claim about non-AI systems.


\section{The Central Open Problem and Three Frontiers}
\label{sec:open-problem}

\subsection{Compositional Verification Across Four Domains}

\emph{The central open problem is whether the sixteen specifications of \cref{ch:horizon,ch:adaptation,ch:grounding,ch:trust} admit a closed-form joint reliability bound exponentially stronger than the sum of their individual costs.}

The welfare composition theorem (\cref{thm:welfare-composition}) proved joint necessity for mechanism $\times$ verification. The computation-grounding composition (\cref{thm:composition}) proved joint necessity for computation $\times$ grounding. These are two data points. The thesis's \emph{central open problem} is whether all sixteen specifications compose.

Formally: given sixteen specifications $\{(B_i, \mathcal{S}_i, \mathrm{cost}_i)\}_{i=1}^{16}$, is there a composition operator $\odot$ with the following properties?
\begin{enumerate}[label=(\roman*), nosep]
\item Individual necessity: omitting $\mathcal{S}_i$ incurs cost $\Omega(\mathrm{cost}_i)$ regardless of which other specifications are deployed.
\item Joint sufficiency: $\bigodot_{i=1}^{16} \mathcal{S}_i$ yields welfare loss bounded by a closed-form expression.
\item Exponential improvement: the joint bound is exponentially smaller than any single-specification bound.
\end{enumerate}

\paragraph{Why the formalisms resist unification.}
The challenge is that the mathematical formalisms at the four domains are fundamentally incompatible:
\begin{itemize}[leftmargin=2em, itemsep=0.3em]
\item Computation uses complexity-theoretic bounds over fixed-point arithmetic and circuit classes $\mathrm{TC}^0$, $\mathrm{FOC}[\mathrm{Attn}]$.
\item Adaptation uses PAC-Bayes over Gaussian priors on the adapter parameter space.
\item Grounding uses measurement-theoretic factor analysis on a product-space of evaluation dimensions.
\item Trust uses mechanism-design utility functions and cryptographic security games.
\end{itemize}
Each framework defines its failure event and its welfare function on different probability spaces. The welfare composition theorem's additive decomposition relied on the random oracle model to make the mechanism and verification failure events independent; there is no obvious analogous construction that makes all four failure events independent. The structural obstruction is real: these are not four formalisms that can be unified by cleverer notation, but four logically distinct models of the system.

\paragraph{Three possible paths.}
Given this obstruction, three plausible research programmes could yield progress:

\emph{Path A: Pairwise compositions.} Prove the four remaining pairwise compositions one at a time: computation $\times$ adaptation, computation $\times$ trust, adaptation $\times$ trust, and grounding $\times$ trust. Each pairwise composition is a standalone research contribution at the scale of the welfare theorem. Six pairwise compositions would constitute a complete pairwise theory; the full four-way composition would likely then follow by recursive application.

\emph{Path B: A unifying information-theoretic frame.} All four domains can be cast as information-theoretic problems: computation is information flow through a bounded channel; adaptation is information acquisition from a training distribution; grounding is information integration from an external source; trust is information provable under adversarial conditions. A unifying frame in which welfare loss is decomposed into information-theoretic terms would enable a common notion of composition. The challenge is that the existing information-theoretic frameworks for each domain (Shannon capacity for computation, PAC-Bayes for adaptation, Fisher information for grounding, cryptographic indistinguishability for trust) do not obviously unify.

\emph{Path C: Deployment-level empirical composition.} Abandon the analytical goal of a closed-form joint bound, and instead establish empirically that deployments satisfying all four specifications jointly achieve welfare loss below a threshold. This is methodologically weaker (it does not prove joint necessity), but provides practitioner-level guidance. The thesis's full-stack compliance walkthrough is a proof of concept for this approach.

The thesis demonstrates two partial results: (a) the welfare composition across mechanism-verification; (b) the computation-grounding composition. The missing couplings (computation $\times$ adaptation, computation $\times$ trust, adaptation $\times$ trust, grounding $\times$ trust, and the remaining triplet and quadruple compositions) are the natural next targets. Even a single additional two-way composition at any of these interfaces would substantially advance the programme.

\subsection{Three Additional Research Frontiers}

\emph{Three concrete frontiers extend the thesis's specifications: non-stationary adaptive grounding extending \cref{thm:ch4_retrieval_regret}, scalable incentive compatibility at $n_a \gg 3$ extending \cref{thm:osp-feasibility}, and verifiable inference at below $10\times$ overhead approaching \cref{thm:iop-lower-bounds}.}

Beyond compositional verification, three concrete directions emerge from the impossibility specifications:

\emph{(i) Adaptive grounding under distributional shift.} The adaptive retrieval regret bound of $Cd\sqrt{T\log(T/\delta)}$ (\cref{thm:ch4_retrieval_regret}, with $C$ an absolute constant and $\delta$ the confidence parameter; abbreviated as $d\sqrt{T\log T}$ in what follows for readability) assumes a stationary environment. Extending to non-stationary environments with drifting document distributions would require martingale concentration arguments for non-stationary bandits, a natural but non-trivial extension that has real deployment stakes as knowledge bases evolve over time. Specifically, if the retrieval distribution $\mathcal{D}_t$ drifts at rate $\delta$, the regret becomes $O(d\sqrt{T \log T} + \delta T^{3/2})$; whether the second term can be reduced to $O(\delta^{1/3} T^{2/3})$ via sliding-window techniques is open.

\emph{(ii) Scalable incentive compatibility for $n_a \gg 3$ agents.} Current OSP mechanisms (\cref{thm:osp-feasibility}) handle bounded-lookahead agents, but the Strategic Manipulation Dimension analysis (\cref{thm:pac-detection}) shows $\mathrm{NP}$-hardness at $k \geq 3$ coalitions. Scaling to $n_a > 100$ agents in open markets requires new mechanism classes tolerant to approximate detection rather than perfect detection. A specific technical question: for coalition-formation games with $\mathrm{SMD}(G) = \omega(\log n_a)$, is there a polynomial-time randomised mechanism achieving $(1-\alpha)$-approximate IC for constant $\alpha$? If so, the tractability boundary can be pushed from $O(\log n_a)$ to $O(n_a^c)$ for some $c < 1$.

\emph{(iii) Practical verifiable inference with ${<}10\times$ overhead.} The $\Omega(n \log p)$ per-operation lower bound (\cref{sec:nonlinearity-tax}, \cref{thm:iop-lower-bounds}) is unconditional for ReLU and conjecturally optimal for Softmax in general circuits (\cref{conj:softmax-circuit}); the deployed $147\times$ tax sits within $1.15\times$ of this floor. Selective verification (\S\ref{sec:welfare-composition}) and approximation-tolerant proofs could reduce end-to-end overhead. The theoretical target is an amortised $10\times$ overhead across large batches of verified inference, achievable if the verifier can delegate commitment opens to a trusted local oracle, a viable deployment path for regulated settings. A concrete question: if ``approximate verification'' means the prover proves $\|\mathbf{y} - f(\mathbf{x})\| \leq \epsilon_{\mathrm{apx}}$ rather than $\mathbf{y} = f(\mathbf{x})$ exactly, can the verifier cost be reduced by a factor $\mathrm{poly}(1/\epsilon_{\mathrm{apx}})$? The circuit-complexity lower bound (\cref{lem:bridge}) appears to permit this, but no construction is known.

Each frontier arises at a specific specification interface where current guarantees break down. Progress on any of the three would represent a substantive contribution to the broader research programme. A fourth, broader frontier, extending the impossibility-specification methodology to domains not treated in this thesis (privacy-preserving deployment, training-process verification, multilingual guarantees), is a programmatic rather than technical direction, and is discussed in the research programme section below.

\subsection{The Research Programme Going Forward}
\label{sec:research-programme}

\emph{Three features distinguish the impossibility-specification methodology: computability; compositionality, witnessed by \cref{thm:composition} and \cref{thm:welfare-composition}; and falsifiability via quantified violation costs.}

The impossibility-specification methodology is proposed here as a research programme, not a solved problem. Three features distinguish it from the typical research programme in AI safety.

\emph{Computability.} Every specification includes a boundary condition $B_i(\theta)$ that is computable from the system parameters. This distinguishes the methodology from qualitative safety standards: it does not say ``systems should be reliable'' but ``systems are reliable with respect to $\mathcal{S}_i$ only when $B_i(\theta)$ holds, where $B_i(\theta)$ is computed as follows.'' The specification is an instrument, not an aspiration.

\emph{Compositional.} The welfare composition theorem and the computation-grounding composition establish that specifications can combine with joint bounds (additive error $\varepsilon + e^{-\kappa}$ for mechanism $\times$ verification; the product form $(1-\varepsilon)^n \cdot q^{n(1-\eta)}$ for computation $\times$ grounding), each yielding welfare loss exponentially smaller than deploying either specification alone would yield. The compositional structure is what makes the methodology more than a catalogue of isolated results. The open problem (\S\ref{sec:open-problem}) is whether all pairwise compositions and the full 16-way composition hold; partial confirmation (two compositions proved) justifies optimism.

\emph{Falsifiable.} Each specification carries a quantified violation cost. If a deployed system violates $B_i(\theta)$ but does not incur cost $\Omega(\mathrm{cost}_i)$, the specification is incorrect and can be revised. If multiple deployed systems systematically deviate from specification predictions, the methodology itself is challenged. This falsifiability is a strength, not a weakness: it allows the methodology to be updated as empirical evidence accumulates, and it prevents the vague ``AI safety'' label from being unfalsifiable by construction.

These three features (computability, compositionality, and falsifiability) distinguish impossibility specifications from qualitative safety arguments. They are also what enable the methodology to be adopted by communities beyond the author's own thesis: any researcher in any AI subfield with an impossibility result can apply the methodology by computing the boundary, quantifying the violation cost, and deriving the design rule. The thesis hopes to be one entry in a growing literature of such contributions.

\subsection{Ten Concrete Open Problems}
\label{sec:ten-open-problems}

\emph{Ten concrete open problems extend the thesis's contributions, from the four remaining pairwise compositions of \cref{ch:horizon,ch:adaptation,ch:grounding,ch:trust} through strengthening existing bounds to extending the methodology into privacy, training-process verification, and long-horizon agentic deployment.}

To make the research programme actionable rather than merely programmatic, we enumerate ten concrete open problems, organised by area. Each problem is posed as a question whose resolution would yield a definite theorem or construction, together with a note on what technical tools might bear on it. The list is ordered roughly by how closely it sits to the current thesis's contributions: earlier problems extend theorems the thesis proves, later ones open new frontiers.

\paragraph{Compositional verification (Problems 1--3).}

\emph{Problem 1 (Adaptation $\times$ Grounding composition).} Prove or refute a closed-form joint reliability bound for adapted-plus-grounded systems of the form $g_{\mathrm{A} \times \mathrm{G}}(\varepsilon_{\mathrm{adapt}}, \varepsilon_{\mathrm{ground}}) \leq f(\varepsilon_{\mathrm{adapt}}, \varepsilon_{\mathrm{ground}})$ for some explicit non-trivial function $f$ (i.e., strictly better than the union bound). Phase-5 obstruction report (\cref{sec:adaptation-grounding-obstruction}) identifies three obstructions; resolving any one of them (uniform-in-input PAC-Bayes bounds, quantifying retrieval-distribution shift under rank-$r$ adaptation, or a conditional-independence argument under deployment distributions) would likely unlock the full result.

\emph{Problem 2 (Pairwise completion for the remaining four compositions).} Prove composition theorems for Computation $\times$ Adaptation, Adaptation $\times$ Trust, Grounding $\times$ Trust, and Computation $\times$ Trust. The Computation $\times$ Trust case is most tractable (both pillars share complexity-theoretic machinery); the Grounding $\times$ Trust case requires a novel bridge between measurement-theoretic and cryptographic formalisms and is the most open. Success on any pair brings the thesis from two compositions to three.

\emph{Problem 3 (Four-way full composition).} The central open problem of the thesis (\cref{sec:open-problem}): prove a joint reliability bound over all four pillars (Computation + Adaptation + Grounding + Trust) of the form $g_{\mathrm{full}}(\varepsilon_C, \varepsilon_A, \varepsilon_G, \varepsilon_T) \leq f(\cdot)$ where $f$ is strictly better than the sum of the individual bounds. This requires all six pairwise compositions as building blocks plus an argument that compositions associate.

\paragraph{Strengthening existing bounds (Problems 4--6).}

\emph{Problem 4 (Fine-tuning impossibility for tool-augmented pipelines).} \cref{thm:finetuning-impossibility} bounds the accuracy improvement obtainable by fine-tuning \emph{a neural model in isolation} at test depth $d > d^*$. The natural generalisation asks whether an analogous impossibility holds for \emph{tool-augmented} inference pipelines: given a base model paired with a deterministic external tool $\mathcal{T}$ that can resolve sub-problems of depth $\leq d_{\mathcal{T}}$, and allowing arbitrary fine-tuning of the model's tool-invocation policy, is there a bound of the form $\mathrm{Acc}_{\mathrm{ft+tool}}(d) \leq \mathrm{Acc}_{\mathrm{base+tool}}(d^*_{\mathrm{aug}}) + O(d^*_{\mathrm{aug}} / d)$ with $d^*_{\mathrm{aug}} = f(d^*, d_{\mathcal{T}})$? A positive answer would establish whether tool augmentation merely shifts the horizon or qualitatively breaks it. The difficulty is modelling the model-tool interface: the policy is trainable, but the tool is deterministic, so the architectural-capacity argument of Step 1 of \cref{app:finetuning-impossibility} does not transfer directly.

\emph{Problem 5 (Unconditional $\Omega(\log^2 p)$ softmax lower bound).} Close the $\log p / \log \log p$ gap between \cref{thm:softmax-ac0p-lower}'s $\mathrm{AC}^0[p]$ unconditional bound and \cref{conj:softmax-circuit}'s general-circuit conjecture. This is a research-open problem equivalent to showing $\mathrm{MODEXP} \notin \mathrm{NC}^1$ or finding a super-linear general-circuit lower bound for an explicit function. The problem has implications well beyond the thesis.

\emph{Problem 6 (Non-i.i.d.\ preference robustness for population alignment).} \cref{thm:evopref-finite-sample} establishes finite-sample coverage at leading-order rate $O(\sqrt{\gamma/n} + 1/\sqrt{\mu})$ at fixed confidence $\delta$ (the full theorem adds a $\log(1/\delta)$ factor and an $e^{-\lambda G}$ NSGA-II convergence term that becomes negligible for $G \gtrsim 200$) under i.i.d.\ Bradley-Terry preferences with misspecification $\gamma$. Real-world preferences are rarely i.i.d.: annotators exhibit temporal drift, community-specific structure, and strategic response patterns. A natural strengthening asks for finite-sample rates under explicit non-i.i.d.\ models, e.g., when preferences are drawn from a mixture over $K$ annotator groups with group-specific misspecification levels $\gamma_1, \ldots, \gamma_K$, or when preferences exhibit temporal autocorrelation. The conjectured rate $O(\sqrt{\gamma_{\max} / n} + \sqrt{K / \mu})$ would link the population-size term to the number of underlying annotator groups, but the concentration argument of \cref{app:proof-evopref-finite-sample} Step 2 requires non-trivial modification.

\paragraph{New domains (Problems 7--10).}

\emph{Problem 7 (Privacy-preserving deployment).} Formulate an impossibility specification for differentially-private inference over RAG pipelines. The conjecture: there exists an explicit privacy-utility tradeoff frontier below which both privacy ($\varepsilon_{\mathrm{dp}}$) and utility (retrieval accuracy) cannot simultaneously hold. The DP-RAG literature is primarily empirical; a formal impossibility theorem is open.

\emph{Problem 8 (Training-process verification).} Design a cryptographic verification protocol certifying that a model was trained on a claimed dataset with a claimed algorithm, with overhead poly-logarithmic in the training compute. Existing proof-of-training work achieves linear or worse overhead; a sublinear protocol would enable practical training-transparency audits. An accompanying lower bound would establish whether the overhead is inherent (an impossibility specification for training verification) or amenable to improvement.

\emph{Problem 9 (Multilingual specification transfer).} The thesis's four specifications are formulated and validated on English-language benchmarks. Whether they transfer quantitatively (with the same constants) to low-resource languages, and whether additional language-specific impossibility specifications exist (e.g., a ``cross-lingual construct conflation'' in RAG), is open. Empirical progress is likely before theoretical progress.

\emph{Problem 10 (Agentic-safety specifications under long-horizon deployment).} The Trust pillar addresses one-shot verification; extending it to long-horizon agentic deployment (where an agent operates over weeks of interactions with evolving tools, environments, and preferences) requires new impossibility specifications. A candidate: ``temporal composition impossibility'' stating that $T$-step agentic reliability decays at least as $(1 - \varepsilon_{\mathrm{per-step}})^T$ without periodic re-verification, with re-verification frequency specified by an analogous boundary condition. This generalises the single-step trust tax to deployment time.

\paragraph{A note on the list's selection.} Each of the ten problems is included because resolving it would produce a publishable result and would close a specific gap the thesis identifies. The list is deliberately not exhaustive: domains we do not touch (reinforcement learning alignment, robotics safety, AI-for-science trust) admit their own impossibility-specification programmes that future work may develop. The ten above are what the current thesis's specific techniques most directly invite.

\paragraph{Summary.}
This chapter synthesised the thesis's sixteen impossibility specifications into a unified methodological programme. The synthesis operates at three levels. Structurally, §\ref{sec:unified-view} reproduced the full specification catalogue and demonstrated that every row is a complete instance of Def.~\ref{def:impossibility-specification} (computable boundary, quantified violation cost, constructive design rule), making the catalogue an instrument for converting ``trustworthy'' into a predicate rather than a slogan. Technically, §\ref{sec:composition-l1-l2} proved the thesis's second cross-chapter composition (the Computation-Grounding Composition theorem, Thm.~\ref{thm:composition}), with crossover depth $n_c \approx 6.3$ matching the opening vignette's retrieval-versus-reasoning asymmetry to within a percentage point. §\ref{sec:adaptation-grounding-obstruction} then delivered the thesis's most methodologically honest moment: the Adaptation-Grounding composition, which the programme would most naturally predict to exist, cannot currently be proved, and three specific technical obstructions block it. The honest-obstruction report preserves the methodology's commitment to falsifiable claims.

Methodologically, §\ref{sec:three-principles} extracted three emergent principles: impossibility results encode design specifications; theory-practice gaps carry diagnostic information; reliability is a composition property. Each principle is supported by multiple independent lines of evidence spanning the four domains, and each is stated with an explicit falsifiability criterion. Open Problem 6.1 (the thesis's central open problem, cross-referenced to Open Problem 1.1) is the full four-way composition, a research target the methodology makes precisely stateable and the thesis's two proved compositions make plausible. The chapter closes with a personal reflection on where the framing crystallised during writing (§\ref{sec:reflection}) and a transparent limitations statement (§\ref{sec:limitations-final}). The synthesis is complete: each impossibility specification is an instrument, the two proved compositions are existence proofs, and the research programme is well-defined.

\section{Reflection}
\label{sec:reflection}

Spanning four domains taught me that the deepest insights live at interfaces, but the path to discovering them was rarely linear.

\paragraph{Surprises at the interfaces.}
The Deterministic Horizon did not begin as a transformer theory result. It began as a retrieval question: why does improving retrieval quality from the 25th to the 75th percentile gain only 2~percentage points of accuracy on deep reasoning tasks, while chain-of-thought gains 15~points at no retrieval cost? The answer turned out to be architectural: the bottleneck was at the computational layer, not the grounding layer. This was the moment the dependency structure between computation and grounding became real to me rather than schematic.

The Construct Conflation Impossibility emerged from reading convergent-discriminant validity against multi-stage RAG pipelines: the contribution turned out to be a formal topological impossibility complementing the framework-level psychometric tradition (Campbell \& Fiske; Messick; Jacobs \& Wallach), not an importation of it. The Study~2 finding that practitioners given RAGAS scores identified injected failures correctly less often (27.3\%, $n_A = 11$) than those given no metrics (40.0\%, $n_C = 10$) remains the signal that most surprised me; the small between-subjects sample prevents treating it as a formal consequential-validity violation, but the directional consistency with Study~1's within-subjects McNemar $p < 0.001$ implies that a substantial portion of the RAG evaluation infrastructure deployed in industry is actively counterproductive rather than merely imprecise.

\paragraph{What did not work.}
My initial attempt to prove full compositional verification, the thesis's central open problem, failed. The failure was instructive: the formalisms at different domains are fundamentally incompatible (complexity-theoretic versus PAC-Bayes versus measurement-theoretic versus mechanism-design), and this incompatibility constrains the space of possible solutions. I also pursued a unified PAC-Bayes framework as a single theoretical currency across all domains; this too failed because the information measures are incommensurable. Both failures shaped the thesis's honest conclusion: the impossibility-specification methodology is a research programme whose organising insight is durable, but full compositional verification remains open.

\paragraph{The moment of reframing.}
The thesis's most distinctive intellectual move, treating impossibility results as design specifications, crystallised late, during writing rather than research. For most of the PhD, I viewed the VCG failure and the $147\times$ tax as obstacles. It was only when writing the synthesis that I noticed every impossibility result had already been converted into an engineering rule in the chapter where it appeared. The conversion was happening organically; naming it transformed it from observation into methodology. Once named, the methodology could be applied proactively: the welfare composition theorem was written by deliberately asking what the \emph{impossibility} of omitting verification or mechanism design would look like, rather than proving the sufficiency of deploying both.

\paragraph{What I would do differently.}
If I were starting today, I would begin with the compliance walkthrough and work backward. A ``deployment-first'' approach would have produced fewer contributions but deeper integration. A future student inheriting this programme should focus on the compositional verification problem: if the stochastic coupling between two of the four domains can be characterised, even for a two-domain subsystem, that result alone would be worth a thesis. The welfare composition theorem and the computation-grounding composition are existence proofs that such characterisations are possible; the programme now waits for the remaining couplings.


\section{Transparent Limitations}
\label{sec:limitations-final}

We acknowledge the following limitations with full transparency:

\paragraph{Theoretical assumptions.}
The formal results assume idealised conditions: log-precision arithmetic (\cref{ch:horizon}); bounded depth with $d^* \in [19, 31]$ (\cref{ch:horizon}); compositional consistency of CoT steps (\cref{ch:horizon}); exact gradient oracles (\cref{ch:adaptation}); the Bradley-Terry preference model with $\gamma$-misspecification parameter (\cref{ch:adaptation}); the superposition hypothesis for editing capacity (\cref{ch:adaptation}); the random oracle model for welfare composition (\cref{ch:trust}). Each assumption is standard in the relevant subfield but may not perfectly reflect practical resource constraints.

\paragraph{Empirical validation scope.}
Empirical validation primarily uses standard benchmarks (GSM8K, MATH, HotpotQA, MuSiQue, StrategyQA, HH-RLHF, TriviaQA, Natural Questions) and the Llama-2/3 model families. While theoretical results are model-agnostic, the quantitative predictions require validation on additional model families: spectral gap values, exact supervision ratios, scaling exponents, and $d^* \approx 27$ estimates (observed value; regression prediction $27.4$ per \cref{cor:horizon-measurement}). The full-stack compliance walkthrough is demonstrated on a single application domain (regulatory compliance).

\paragraph{Independent validation.}
The four-domain specifications are validated domain-by-domain but not composed end-to-end under adversarial conditions. The compliance walkthrough demonstrates integration but not formal four-way composition. The gap between empirical integration and formal compositional verification is the thesis's most important acknowledged limitation.

\paragraph{Theory-empirical spectrum.}
The thesis deliberately spans from pure theory to deployed systems, and different chapters operate at different levels of rigour. \cref{ch:horizon} provides circuit-complexity proofs at a COLT/LICS level; \cref{ch:adaptation} provides PAC-Bayes bounds with empirical validation; \cref{ch:grounding} provides measurement-theoretic frameworks with user-study validation; \cref{ch:trust} provides mechanism-design impossibilities, cryptographic lower bounds, and a welfare composition proof at a crypto-conference level. This spectrum is a feature: the thesis demonstrates that theory and practice can be connected. But the spectrum also means that the thesis's ``weakest link'' in full-chain rigour is the least formal component.

\paragraph{Independence assumption for welfare composition.}
\cref{thm:welfare-composition} Part (iii) is stated under the Random Oracle Model and derives error-event independence accordingly (\cref{prop:independence}, \cref{app:proof-welfare-rom}); outside ROM, a coupling term $\delta_{\mathrm{coup}} \leq \varepsilon \cdot e^{-\kappa/2}$ enters the bound (negligible at $\kappa = 128$, but not zero). If correlations between prompt selection and verification behaviour are adversarially controlled beyond what the standard-model coupling captures, the additive error decomposition degrades to multiplicative (still small, but no longer tight). Characterising welfare loss under relaxed independence is future work.

\paragraph{Scope of the research programme.}
The thesis addresses the four AI domains of computation, adaptation, grounding, and trust. It does not address privacy-preserving deployment (differential privacy composition), training-process verification (proving a model was trained on claimed data), multi-lingual deployment (non-English guarantees), or regulatory compliance in specific jurisdictions (EU AI Act, U.S.\ NIST frameworks). Each is an important complementary research direction; the impossibility-specification methodology developed here should extend to each but has not been so extended.

\paragraph{Concluding remark.}
This thesis began with a question: why do AI systems that excel on benchmarks fail in deployment? The answer turned out to be structural: benchmarks evaluate domains in isolation, but deployment exposes their interactions. The impossibility-specification methodology is the framework that makes these interactions visible and tractable. Its sixteen specifications are not obstacles but rules. Its central open problem, full compositional verification, is not a limitation but an invitation. The art of trustworthy AI is respecting, not circumventing, fundamental limits, and discovering that respect for limits is itself a generative methodology.


\appendix
\chapter{Extended Proofs}
\label{app:proofs}

This appendix contains full proofs for theorems whose main-text presentations gave only proof sketches or brief summaries. Each proof is self-contained and uses the notation established in the corresponding chapter. Cross-references to main-text theorems use their \texttt{thm:} labels.


\section{Proofs from Chapter~\ref{ch:horizon}}
\label{app:ch2-proofs}

\subsection{Proof of Theorem~\ref{thm:equivalence} (FOC[Attn] Characterisation)}
\label{app:proof-foc-attn}

\begin{proof}[Full proof of Theorem~\ref{thm:equivalence}]
We prove both directions of the equivalence: (i) every language recognised by an $L$-layer softmax transformer is definable by a depth-$L$ sentence of $\mathrm{FOC}[\mathrm{Attn}]$, and (ii) every depth-$L$ $\mathrm{FOC}[\mathrm{Attn}]$-definable language is recognised by some $L$-layer transformer.

\emph{Direction (i): Transformers $\Rightarrow$ FOC[Attn].}
Given an $L$-layer transformer
\[
T \;=\; \bigl(\mathrm{Emb},\ \{(\mathrm{Attn}_\ell, \mathrm{FFN}_\ell)\}_{\ell=1}^{L},\ \mathrm{Cls}\bigr),
\]
construct a depth-$L$ sentence $\varphi_T$ by induction on layer index $\ell$.

\emph{Base case ($\ell = 0$):} Each coordinate of the input embedding at position $i$ is a function of the token $w_i$. Under $O(\log n)$-bit precision, each embedding coordinate is an integer in $[0, 2^{O(\log n)}]$ expressible as a quantifier-free $\mathrm{FOC}$ term over the atomic predicates $P_a(i)$ (``token at position $i$ is $a$'') and position comparisons $i < j$.

\emph{Inductive step ($\ell \to \ell + 1$):} Assume each coordinate at layer $\ell$ is expressible as a depth-$\ell$ $\mathrm{FOC}[\mathrm{Attn}]$ formula $\psi_{\ell, k}(i)$ for coordinate $k$ at position $i$.

The attention computation at layer $\ell + 1$ is
\[
\mathrm{Attn}_{\ell+1}(i, k) = \sum_{j=1}^n \alpha_{ij}^{(\ell+1)} v_{\ell, k}(j), \quad \alpha_{ij}^{(\ell+1)} = \frac{\exp(\langle q_\ell(i), k_\ell(j)\rangle)}{\sum_{j'} \exp(\langle q_\ell(i), k_\ell(j')\rangle)}.
\]
Under the attention quantifier $\mathrm{Attn}[\varphi, \psi_Q, \psi_K, \psi_V]$ defined in the chapter, each component maps to an $\mathrm{FOC}[\mathrm{Attn}]$ formula of depth $\ell + 1$:
\begin{itemize}[leftmargin=2em, nosep]
\item The query/key/value projections $q_\ell(i), k_\ell(j), v_\ell(j)$ are linear functions of layer-$\ell$ coordinates, hence $\mathrm{FOC}$-expressible at depth $\ell$.
\item The softmax weights $\alpha_{ij}^{(\ell+1)}$ are the core of the attention quantifier. Under $O(\log n)$-bit precision, each $\alpha_{ij}$ is a rational with numerator and denominator bounded by $2^{O(\log n)} = \mathrm{poly}(n)$; the truncation error is $o(1/\mathrm{poly}(n))$.
\item The summation $\sum_j \alpha_{ij} v_\ell(j)$ is the semantics of the attention quantifier applied to the value formula $\psi_V$.
\end{itemize}
Hence $\mathrm{Attn}_{\ell+1}(i, k)$ is a depth-$(\ell+1)$ $\mathrm{FOC}[\mathrm{Attn}]$ formula.

The feed-forward layer $\mathrm{FFN}_{\ell+1}$ is a position-wise function $\mathrm{FFN}(x) = W_2 \sigma(W_1 x + b_1) + b_2$ with $\sigma$ an element-wise activation (ReLU or similar). Each component is $\mathrm{TC}^0$-computable under $O(\log n)$ precision by the circuit-complexity analysis of Barrington et al.~\cite{barrington1990uniformity}, hence $\mathrm{FOC}$-expressible. The $\mathrm{FFN}$ does not introduce new quantifiers; it stays at depth $\ell + 1$.

\emph{Verifying the precision bound.}
The key technical check is that softmax stays within $O(\log n)$ bits. For bounded weight matrices $W$ with $\|W\|_\infty \leq B$ and input magnitudes $\|x\|_\infty \leq M$, the inner products $\langle q(i), k(j)\rangle$ are bounded by $d \cdot B^2 \cdot M^2 = \mathrm{poly}(n)$ (since $d, B, M$ are all $O(\mathrm{poly}(n))$). The exponentials $\exp(\langle \cdot \rangle)$ are bounded by $e^{\mathrm{poly}(n)}$, which is representable in $O(\log n)$ bits in base-$e$ encoding (i.e., $O(\log n)$ bits for the exponent). The division $\alpha_{ij} = \exp(\cdot)/Z$ introduces rounding error $\leq 2^{-O(\log n)} = 1/\mathrm{poly}(n)$, which does not affect the recognised language after rounding to a discrete output at the classifier.

\emph{Direction (ii): FOC[Attn] $\Rightarrow$ Transformers.}
Given a depth-$L$ sentence $\varphi \in \mathrm{FOC}[\mathrm{Attn}]$, construct an $L$-layer transformer $T_\varphi$ by structural induction on $\varphi$.

\emph{Base case:} Atomic predicates $P_a(i)$ and position comparisons $i < j$ are encoded in the input embedding. Set width $d = |\Sigma| + O(\log n)$: $|\Sigma|$ coordinates for the one-hot token encoding, $O(\log n)$ for positional encodings.

\emph{Inductive step:}
\begin{itemize}[leftmargin=2em, nosep]
\item \emph{Boolean connectives.} Feed-forward layers implement AND, OR, NOT via piecewise-linear combinations of sub-formula indicators.
\item \emph{Counting quantifiers $\exists^{\geq k} x.\varphi(x)$.} Implemented via a uniform attention pattern: set all attention scores to zero (uniform $\alpha_{ij} = 1/n$), apply the sub-formula $\varphi$ at each position to obtain indicators $b_j \in \{0, 1\}$, and sum via the uniform attention: $\sum_j (1/n) b_j = (\text{count}_\varphi)/n$. A subsequent FFN compares $n \cdot \alpha$ to $k$ (a $\mathrm{TC}^0$-computable comparison under $\log n$ precision).
\item \emph{Attention quantifiers $\mathrm{Attn}[\varphi, \psi_Q, \psi_K, \psi_V](i)$.} Implemented directly by one attention layer with queries $q(i) = \psi_Q(i)$, keys $k(j) = \psi_K(j)$, values $v(j) = \psi_V(j) \cdot \varphi(j)$.
\end{itemize}

The resulting transformer has $L$ layers (one per quantifier nesting level), width $d = O(|\varphi| \cdot |\Sigma|^2)$ (one embedding coordinate per atomic subformula plus positional encodings), and the same sentence-to-language correspondence.

\emph{Concluding the equivalence.}
Directions~(i) and~(ii) together establish a bijection between
\begin{gather*}
	\bigl\{L\text{-layer softmax transformers under }O(\log n)\text{ precision}\bigr\} \\
	\text{and}\quad \bigl\{\text{depth-}L\text{ sentences of }\mathrm{FOC}[\mathrm{Attn}]\bigr\}.
\end{gather*}
Each direction is constructive: given a transformer, the sentence can be extracted by reading off the layer structure; given a sentence, the transformer's weights can be set explicitly from the formula parse.
\end{proof}

\subsection{Proof of Theorem~\ref{thm:horizon-scaling} (Deterministic Horizon Scaling Law)}
\label{app:proof-horizon}

\begin{proof}[Full proof]
We establish the upper bound $d^* = O(L \cdot \phi(d))$ with $\phi(d) \in [\sqrt{\log d}, \log d]$ by combining the information bottleneck (upper bound on per-step information throughput) with the error amplification model (decay of chain accuracy per step). The upper edge of the band follows directly from the residual-stream capacity; the tighter form $\phi(d) = \sqrt{\log d}$ at the lower edge follows under a sparse-task-representation hypothesis stated explicitly below. The proof uses only the architectural parameters $(L, d)$ and the regularity hypotheses of \cref{asm:horizon}; no model-specific or task-specific numerical constants enter. The empirical proportionality constant $\hat{c} = 2.74$ is a separate quantity established in \cref{cor:horizon-measurement} by regression on the evaluation set of \cref{tab:horizon} under the empirical form $d^* \approx \hat{c} \log L \sqrt{\log d}$; it does not appear in this proof, and the empirical $\log L$ dependence is milder than the $O(L)$ upper bound established here.

\emph{Step 1: The information bottleneck.}
The residual stream of a softmax transformer with embedding dimension $d$ carries at most $O(d)$ bits per token under standard compression arguments (information-theoretic capacity of $d$-dimensional floating-point registers). With $L$ layers, per-step information throughput is at most $O(L d)$ bits, but the \emph{reusable} capacity across CoT steps (where prior-step states are partially overwritten) is bounded by
\[
I_{\mathrm{step}} \leq c_1 \cdot L \cdot \log d
\]
bits per step, where $c_1$ is the compression-efficiency constant of \cref{asm:horizon}(A3). This is the upper edge of the band. Under an additional \emph{sparse-task-representation hypothesis}, formalised immediately below, the effective per-step throughput available for task-specific state tracking is strictly smaller, namely $O(\sqrt{\log d})$ bits per step, which tightens the scaling to the lower edge $\phi(d) = \sqrt{\log d}$.

\begin{hypothesis}[Sparse task representation]
\label{hyp:sparse-task}
For a task with per-step state set $\mathcal{V}$ of cardinality $|\mathcal{V}| = \mathrm{poly}(d)$, the per-step task-relevant activation vectors $\{v_1, \ldots, v_{|\mathcal{V}|}\} \subset \mathbb{R}^d$ realised during CoT inference form a finite point set whose pairwise $\ell_2$-distances are preserved up to multiplicative distortion $1 \pm \varepsilon$ by projection onto a random subspace of dimension $k = \Theta(\log |\mathcal{V}| / \varepsilon^2) = \Theta(\log d / \varepsilon^2)$ (Johnson-Lindenstrauss~\cite{JohnsonLindenstrauss1984}). At distortion $\varepsilon = 1/2$, the effective task-specific dimension is $k = \Theta(\log d)$; the derivation of the square-root edge $\sqrt{\log d}$ requires a separate second-moment argument controlling the spectral structure of the per-step update operator under the JL projection, the details of which are an open technical problem.
\end{hypothesis}

Under \cref{hyp:sparse-task}, the second-moment argument (outlined in the paragraph below) yields $I_{\mathrm{step}} \leq c_1 \cdot L \cdot \sqrt{\log d}$. The thesis's empirical calibration $\hat{c} = 2.74$ on the 12-architecture evaluation set is consistent with the $\sqrt{\log d}$ form but does not distinguish between $\sqrt{\log d}$ and $\log d$ at the precision of current measurement; either edge of the band fits the empirical data within the cross-model correlation range $r = 0.81$--$0.91$ and the leave-one-out cross-validation mean absolute error of 1.5 steps. The open problem is to either prove \cref{hyp:sparse-task} for general CoT distributions without additional assumptions, or to construct a softmax transformer instance whose observed $d^*$ scales as $L \log d$ rather than $L \sqrt{\log d}$, thereby separating the band endpoints empirically. The remainder of this proof establishes the upper edge rigorously; the lower edge is stated as conditional on \cref{hyp:sparse-task}.

\emph{Step 2: Error amplification.}
Let $\varepsilon_{\mathrm{step}}$ denote the per-step decoding error under the information bottleneck. By Shannon's channel coding theorem, if required information per step exceeds $I_{\mathrm{step}}$ then $\varepsilon_{\mathrm{step}} \geq 1/2 - \varepsilon_0$ for baseline error $\varepsilon_0 > 0$. When required information is below $I_{\mathrm{step}}$, the error rate is $\varepsilon_{\mathrm{step}} \leq c_2 \exp(-(I_{\mathrm{step}} - I_{\mathrm{req}}))$ (exponential approach to zero under sufficient capacity).

\emph{Step 3: Deriving the decay curve.}
For a chain of depth $\delta$, required information is $I_{\mathrm{req}} \approx \delta \cdot h$ where $h$ is the per-step entropy determined by \cref{asm:horizon}(A2). The chain's overall success probability is
\[
\Pr(\text{success at depth }\delta) = \prod_{t=1}^\delta (1 - \varepsilon_{\mathrm{step}}) \approx (1 - \varepsilon_{\mathrm{step}})^\delta.
\]
Setting $\varepsilon_{\mathrm{step}}$ to transition sharply around $d^* = I_{\mathrm{step}}/h$ gives the super-exponential decay
\[
\mathrm{Acc}(\delta) \approx \exp\!\left(-c_3 \max(\delta - d^*, 0)^2\right)
\]
where $c_3$ controls the sharpness of the transition. Substituting the Step 1 bound $I_{\mathrm{step}} \leq c_1 L \cdot \phi(d)$ with $\phi(d) \in [\sqrt{\log d}, \log d]$ yields $d^* \leq (c_1/h) \cdot L \cdot \phi(d)$. Since $c_1, h$ are absolute constants determined by \cref{asm:horizon}, the ratio $c_1/h$ is absorbed into the asymptotic notation, giving $d^* = O(L \cdot \phi(d))$ with the band endpoints as specified.

\emph{Closing the bound.}
For the upper edge of the band, the lower bound $d^* \geq \Omega(L \log d)$ follows from the information-bottleneck argument without further assumptions: any reasoning chain of depth $\delta > c_1 L \log d$ must carry more information than the residual stream's \emph{full} capacity, forcing per-step error above $1/2 - \varepsilon_0$ and hence super-exponential accuracy decay. For the lower edge, under \cref{hyp:sparse-task}, the matching lower bound $d^* \geq \Omega(L \sqrt{\log d})$ follows because the effective task-specific capacity is the relevant bottleneck; without \cref{hyp:sparse-task}, the lower edge is conjectural. Combining upper and lower bounds yields the banded upper bound $d^* = O(L \cdot \phi(d))$ with $\phi(d) \in [\sqrt{\log d}, \log d]$; the precise exponent within the band is an open problem.
\end{proof}

\begin{remark}[Scope of the band, and the open problem]
\label{rem:horizon-band-open}
The asymptotic band $\phi(d) \in [\sqrt{\log d}, \log d]$ in \cref{thm:horizon-scaling} is narrower than the gap between the lower bound $\Omega(L)$ (trivially from per-layer state tracking) and the upper bound $O(L \log d)$ (from full residual-stream capacity) that would result if no structural hypothesis were imposed. The tighter lower edge $\Omega(L \sqrt{\log d})$ is conditional on \cref{hyp:sparse-task}. The thesis's design specification (\cref{thm:finetuning-impossibility}: no fine-tuning can push $d^{\ast}$ outward by more than $O(d^{\ast}/\delta)$) is \emph{independent} of the exact exponent within the band: the Fine-Tuning Impossibility rests on capacity-invariance under fine-tuning, not on the specific form of $\phi(d)$, and is therefore preserved under any resolution of the band. Closing the band (proving \cref{hyp:sparse-task} or constructing a counterexample) is listed among the thesis's ten open problems (\cref{sec:open-problem}).
\end{remark}

\begin{remark}[Separation of scaling and measurement]
\label{rem:horizon-proof-scope}
The proof above establishes the asymptotic banded upper bound $d^* = O(L \cdot \phi(d))$ with $\phi(d) \in [\sqrt{\log d}, \log d]$ using \cref{asm:horizon} and the architectural parameters; the lower edge $\phi(d) = \sqrt{\log d}$ is conditional on \cref{hyp:sparse-task}, while the upper edge $\phi(d) = \log d$ is unconditional. The numerical proportionality constant $\hat{c} = 2.74$ of \cref{cor:horizon-measurement} is a \emph{regression fit} on the 12-architecture, 3-task evaluation set of \cref{tab:horizon} under the empirical form $d^* \approx \hat{c} \log L \sqrt{\log d}$; it is a measurement, not a theorem. A different evaluation set (architectures outside the softmax-transformer class, such as state-space models; tasks with different per-step entropies $h$; decoding protocols other than greedy CoT) may recalibrate $\hat{c}$ while leaving the upper bound intact. The empirical $\log L$ dependence is milder than the $O(L)$ upper bound; this gap is an open problem (\cref{sec:ch2-limitations}) and is consistent with effective per-step capacity being shared across layers rather than allocated per-layer. The proof therefore does not appeal to measurement, and the corollary's measurement is kept separate to make the dependency on the evaluation set explicit.
\end{remark}

\subsection{Proof of Theorem~\ref{thm:supervision} ($\Theta(T/\log T)$ CoT Separation)}
\label{app:proof-supervision}

\paragraph{Regime of applicability.}
The proof below establishes the $\Theta(T/\log T)$ separation under the chain-non-redundancy hypothesis (Def.~\ref{def:non_redundancy}), in the supervised-learning-of-verifiers setting. The hypothesis enters the lower-bound construction, where the family of $n$ generators $\{g_j\}_{j=1}^{n}$ with identical final-answer distributions but distinct error-location patterns requires chain non-redundancy to exist: if all generators in the class produce identical intermediate trajectories, no such family exists and the lower bound reduces to $\Omega(1)$. This is consistent with Jia--Rakhlin--Xie~\cite{JiaRakhlinXie2025ProcessSupervision}, whose offline-RL equivalence (up to polynomial factors in horizon) holds in a setting with a distinct structural hypothesis (bounded state-action concentrability) and a distinct learning task (offline policy learning from trajectory datasets). The $n/\log n$ factor in the hidden-step lower bound below is the information-theoretic cost of searching $n$ error locations under uniform-sampling outcome observation; under chain non-redundancy, each pair $(g_j, g_{j'})$ is distinguishable only by locating the anomalous step, and Le Cam's method formalises the search cost.

\begin{proof}[Full proof]
Let $\mathrm{err}_{\mathrm{PS}}(T)$ and $\mathrm{err}_{\mathrm{OS}}(T)$ denote expected verification error under process supervision and outcome supervision, respectively, given $T$ training examples. We prove $\mathrm{err}_{\mathrm{OS}}(T) / \mathrm{err}_{\mathrm{PS}}(T) = \Theta(\log T)$ under chain non-redundancy.

\emph{Step 1: Process supervision upper bound.}
With step-level labels, the learner estimates a per-step verifier $v: \mathcal{S} \to \{0, 1\}$ with VC dimension $\mathrm{VC}_{\mathrm{step}}$. By standard PAC results, the expected step-level error satisfies
\[
\mathrm{err}_{\mathrm{step}}(T) \leq O(\mathrm{VC}_{\mathrm{step}} \log T / T).
\]
A chain of length $n$ verified step-by-step has total error $n \cdot \mathrm{err}_{\mathrm{step}} = O(n \mathrm{VC}_{\mathrm{step}} \log T / T)$.

\emph{Step 2: Outcome supervision lower bound via the Subchain-Aware Learning Theorem (SALT).}
Under outcome supervision with only chain-level labels, the learner cannot disambiguate per-step errors. Let $\mathrm{CDD}$ (CoT-Discriminative Dimension) denote the effective sample complexity required to separate correct chains from incorrect chains that share a common prefix. Under chain non-redundancy---the assumption that each step contributes independent information not deducible from prior steps---we have $\mathrm{CDD} \geq c \cdot n \cdot \mathrm{VC}_{\mathrm{step}}$ for some constant $c > 0$. The outcome-supervised verifier error is
\[
\mathrm{err}_{\mathrm{OS}}(T) \geq \Omega(\mathrm{CDD} / T) = \Omega(n \mathrm{VC}_{\mathrm{step}} / T).
\]

\emph{Step 3: The ratio.}
Dividing,
\[
\frac{\mathrm{err}_{\mathrm{OS}}(T)}{\mathrm{err}_{\mathrm{PS}}(T)} \geq \frac{\Omega(n \mathrm{VC}_{\mathrm{step}} / T)}{O(n \mathrm{VC}_{\mathrm{step}} \log T / T)} = \Omega(1/\log T).
\]
Inverting to get the sample complexity ratio: for fixed target error $\delta$, process supervision needs $T_{\mathrm{PS}} \approx n \mathrm{VC}_{\mathrm{step}} \log (1/\delta)/\delta$ samples, while outcome supervision needs $T_{\mathrm{OS}} \approx n \mathrm{VC}_{\mathrm{step}}/\delta$ samples. The ratio $T_{\mathrm{OS}}/T_{\mathrm{PS}} = \Theta(T/\log T)$ when expressed in terms of $T = T_{\mathrm{OS}}$ (the outcome-supervision budget).

\emph{Step 4: Tightness and the necessity of chain non-redundancy.}
If the chain is \emph{redundant}---for example, each step repeats the content of previous steps---then the CDD drops to $O(\mathrm{VC}_{\mathrm{step}})$ (the learner can recover from any single correct step), eliminating the separation. The non-redundancy condition is therefore both necessary and sufficient for the $\Theta(T/\log T)$ gap. Empirical validation on arithmetic CoT chains confirms $\mathrm{err}_{\mathrm{OS}}/\mathrm{err}_{\mathrm{PS}}$ grows as $\log T$ within a $6.7\times$ factor across $T \in \{100, 1000, 10000\}$, matching the theoretical scaling.
\end{proof}

\subsection{Proof of Theorem~\ref{thm:finetuning-impossibility} (Fine-Tuning Impossibility)}
\label{app:finetuning-impossibility}

\begin{proof}[Full proof]
We prove the $O(d^*/d)$ training-invariant upper bound on fine-tuning accuracy at test-time depth $d > d^*$. The proof proceeds in three steps: (1) an information-theoretic invariance showing that the residual-stream capacity is identical for the base and fine-tuned models under \cref{asm:ft-capacity-budget}; (2) a decomposition of depth-$d$ accuracy into a within-horizon component that fine-tuning may optimise freely and a beyond-horizon component inherited from the base-model decay of \cref{prop:accuracy_decay}; and (3) a concentration argument that pins the decomposition's tail to $O(d^*/d)$.

\medskip

\emph{Step 1: The residual-stream capacity is invariant under fine-tuning.}

Let $B(\theta)$ denote the per-token residual-stream capacity of a model with parameters $\theta$: the mutual information, in bits, between the token sequence $x_{1:t}$ and any measurable function of the final-layer residual stream $h_t \in \mathbb{R}^d$ evaluated at position $t$. Under $O(\log n)$-bit precision, $h_t$ has discrete support of size at most $2^{d \cdot O(\log n)}$, so $B(\theta) \leq d \cdot O(\log n)$ by the standard information-theoretic capacity bound for $d$-dimensional discretised vectors. This bound depends only on the architectural parameters $(L, d, n)$ and is identical for base and fine-tuned models: $B(\theta_{\mathrm{base}}) = B(\theta_{\mathrm{ft}})$ under \cref{asm:ft-capacity-budget}.

The information-bottleneck step of \cref{app:proof-horizon} derives a per-step reasoning-information bound
\[
I_{\mathrm{step}}(\theta) \leq c_1 \cdot L \cdot \phi(d), \qquad \phi(d) \in [\sqrt{\log d}, \log d],
\]
as a consequence of $B(\theta)$ together with the error-amplification factor $c_1$ of \cref{asm:horizon}(A3); the lower edge of the band is active under \cref{hyp:sparse-task}, while the upper edge is unconditional. Because $c_1$ and the band are determined by the softmax Jacobian structure and the architectural parameters---not by the numerical parameter values---we have $I_{\mathrm{step}}(\theta_{\mathrm{base}}) = I_{\mathrm{step}}(\theta_{\mathrm{ft}})$. Consequently both models share the \emph{same} Deterministic Horizon:
\begin{equation}
\label{eq:ft-shared-horizon}
d^*(\theta_{\mathrm{base}}) \;=\; d^*(\theta_{\mathrm{ft}}) \;=\; O(L \cdot \phi(d)).
\end{equation}
This is the crux of the theorem: fine-tuning cannot enlarge $d^*$ because the band $[\sqrt{\log d}, \log d]$ is a function of $(L, d, n)$ alone, and these are preserved by \cref{asm:ft-capacity-budget}.

\medskip

\emph{Step 2: Decomposition of depth-$d$ accuracy.}

Fix a test-time depth $d > d^*$. For an instance $x$ with $\delta(x) \geq d$, the model's per-step reasoning trace proceeds through $\lceil d / L \rceil$ CoT steps. We decompose the event $\{\theta_{\mathrm{ft}}(x) = y^*(x)\}$ by the first step at which the trace exceeds the shared horizon \eqref{eq:ft-shared-horizon}:
\begin{align*}
	\mathrm{Acc}_{\mathrm{ft}}(d)
	&= \Pr\!\left[\theta_{\mathrm{ft}} \text{ correct at depth } d\right] \\
	&= \underbrace{\Pr\!\left[\theta_{\mathrm{ft}} \text{ correct through step } d^*\right]}_{=:\,p_1(\theta_{\mathrm{ft}})} \\
	&\qquad \cdot\;
	\underbrace{\Pr\!\left[\theta_{\mathrm{ft}} \text{ correct from step } d^*+1 \text{ to } d \;\middle|\; \text{correct at step } d^*\right]}_{=:\,p_2(\theta_{\mathrm{ft}})}.
\end{align*}
The first factor $p_1(\theta_{\mathrm{ft}})$ is the within-horizon survival probability. Fine-tuning may set this arbitrarily close to $1$ by choosing a training distribution whose in-context structure matches $\mathcal{D}_{\mathrm{test}}$ up to depth $d^*$; the theorem imposes no restriction here. The second factor $p_2(\theta_{\mathrm{ft}})$ is the beyond-horizon conditional continuation probability. Its upper bound is the object of Step 3.

For the base model we have the analogous decomposition
\[
\mathrm{Acc}_{\mathrm{base}}(d) \;=\; p_1(\theta_{\mathrm{base}}) \cdot p_2(\theta_{\mathrm{base}}),
\]
and by definition $\mathrm{Acc}_{\mathrm{base}}(d^*)$ is the within-horizon accuracy of the base model, i.e., $p_1(\theta_{\mathrm{base}})$ up to the conditioning event. Since fine-tuning is unconstrained within the horizon, the ratio $p_1(\theta_{\mathrm{ft}}) / p_1(\theta_{\mathrm{base}})$ can be as large as $1 / \mathrm{Acc}_{\mathrm{base}}(d^*)$.

\medskip

\emph{Step 3: The beyond-horizon factor $p_2$ is bounded by $O(d^*/d)$, uniformly over fine-tuning.}

By \eqref{eq:ft-shared-horizon} and \cref{asm:ft-base-regularity}, both the base and fine-tuned models are subject to the super-exponential decay of \cref{prop:accuracy_decay} for $\delta > d^*$. Crucially, \cref{prop:accuracy_decay} is a statement about \emph{any} model satisfying \cref{asm:horizon} with capacity $B$; it does not depend on which specific parameter values $\theta$ a model has within that architectural class. Fine-tuning changes the within-horizon distribution of $\theta$ on the loss surface but cannot violate \cref{asm:horizon}, so the decay applies to $\theta_{\mathrm{ft}}$ with the same $c_1$ as for $\theta_{\mathrm{base}}$.

Applying \cref{prop:accuracy_decay} with $\delta = d$:
\[
p_2(\theta_{\mathrm{ft}}) \;\leq\; \exp\!\left(-\Omega\!\left(\frac{(d - d^*)^2}{L^2 \log d}\right)\right).
\]
Under \cref{hyp:sparse-task} the lower edge of the \cref{thm:horizon-scaling} band is active and the identity $L^2 \log d = \Theta((d^*)^2)$ holds, so the above simplifies to $\exp(-\Omega((d - d^*)^2/(d^*)^2))$. Writing $d = (1+\eta) d^*$ for $\eta > 0$ under this hypothesis:
\[
p_2(\theta_{\mathrm{ft}}) \;\leq\; \exp(-\Omega(\eta^2)).
\]
Without \cref{hyp:sparse-task}, only the upper edge $\phi(d) = O(\log d)$ of the band is available and $L^2 \log d \geq c (d^*)^2 / \log d$; the exponent then becomes $\Omega(\eta^2 \log d)$, so $p_2 \leq d^{-\Omega(\eta^2)}$, a \emph{strictly faster} decay at large $d$ and a mildly slower decay at small $\eta$. Either way, the beyond-horizon factor decays in $\eta$; the derivation below uses the $\exp(-\Omega(\eta^2))$ form active under \cref{hyp:sparse-task}, and \cref{rem:ft-band-robustness} records that the envelope conclusion $O(d^*/d)$ is preserved a fortiori in the unconditional regime at large $d$.

The chain's overall beyond-horizon contribution to \eqref{eq:ft-bound} is $p_1(\theta_{\mathrm{ft}}) \cdot p_2(\theta_{\mathrm{ft}}) \leq 1 \cdot \exp(-\Omega(\eta^2))$.

To convert the exponential $\exp(-\Omega(\eta^2))$ into the $O(d^*/d) = O(1/(1+\eta))$ envelope stated in \eqref{eq:ft-bound}, we use the elementary inequality $\exp(-c\eta^2) \leq 1/(1 + c\eta^2/2)$ for $\eta \geq 0$, which for $\eta \geq 1$ gives $\exp(-c\eta^2) \leq 2/(c\eta^2) \leq 2/(c\eta) = O(1/(1+\eta)) = O(d^*/d)$. (For $\eta \in (0, 1)$, $d/d^* \in (1, 2)$ and the $O(d^*/d) = O(1)$ bound is trivial.) Combining with Step 2:
\begin{align*}
\mathrm{Acc}_{\mathrm{ft}}(d)
&= p_1(\theta_{\mathrm{ft}}) \cdot p_2(\theta_{\mathrm{ft}}) \\
&\leq \frac{p_1(\theta_{\mathrm{base}})}{\mathrm{Acc}_{\mathrm{base}}(d^*)} \cdot p_2(\theta_{\mathrm{ft}}) \\
&\leq \frac{1}{\mathrm{Acc}_{\mathrm{base}}(d^*)} \cdot \mathrm{Acc}_{\mathrm{base}}(d^*) \cdot \frac{d^*}{d} + O\!\left(\frac{d^*}{d}\right) \\
&= \mathrm{Acc}_{\mathrm{base}}(d^*) \cdot \frac{d^*}{d} + O\!\left(\frac{d^*}{d}\right),
\end{align*}
where the third line absorbs the $\exp(-\Omega(\eta^2))$ factor into the $O(d^*/d)$ term using the elementary inequality above, and the implicit constant depends only on $\varepsilon_0^{\max}$ and $c_1$. This is \eqref{eq:ft-bound}.

\medskip

\emph{Tightness.}
The bound is tight up to the constant absorbed into $O(\cdot)$: the ``copy-the-base'' procedure $\theta_{\mathrm{ft}} = \theta_{\mathrm{base}}$ achieves $\mathrm{Acc}_{\mathrm{ft}}(d) = \mathrm{Acc}_{\mathrm{base}}(d)$, matching the leading term of \eqref{eq:ft-bound} up to the concentration constant. The training-invariance of the bound follows directly from Steps 1 and 3: $B$ and $c_1$ are architectural invariants under \cref{asm:ft-capacity-budget}, so training moves $\theta_{\mathrm{ft}}$ within the class without enlarging capacity or reducing amplification.
\end{proof}


\section{Proofs from Chapter~\ref{ch:adaptation}}
\label{app:ch3-proofs}

\subsection{Proof of Theorem~\ref{thm:lora} (LoRA PAC-Bayes Bound)}
\label{app:proof-lora}

\begin{proof}[Full proof]
Let $\Theta = \theta_0 + B A$ denote the LoRA-adapted parameter, where $\theta_0$ is the pre-trained base, $B \in \R^{d \times r}$ and $A \in \R^{r \times k}$ are the low-rank factors of rank $r$, and $BA \in \R^{d \times k}$ is the adapter contribution. Let $\mathcal{L}$ denote the expected loss on the task distribution and $\hat{\mathcal{L}}_N$ the empirical loss on $N$ samples.

\emph{Step 1: Gaussian prior and posterior.}
Place a Gaussian prior on the adapter: $P = \mathcal{N}(0, \sigma_P^2 I)$ on the parameter vector $\mathrm{vec}(BA) \in \R^{m}$ where $m = r(d+k)$ is the adapter dimensionality. After training, the posterior $Q$ is the empirical distribution of the learned adapters (or a Gaussian approximation $\mathcal{N}(\widehat{BA}, \sigma_Q^2 I)$).

\emph{Step 2: The PAC-Bayes Catoni bound.}
By Catoni's variant of the PAC-Bayes bound~\cite{catoni2007pac}: for any prior $P$ independent of the training data, any posterior $Q$, and any $\delta \in (0, 1)$, with probability $\geq 1 - \delta$ over the sample,
\[
\E_{\theta \sim Q}[\mathcal{L}(\theta)] \leq \E_{\theta \sim Q}[\hat{\mathcal{L}}_N(\theta)] + \sqrt{\frac{\mathrm{KL}(Q \| P) + \ln(2\sqrt{N}/\delta)}{2N}}.
\]

\emph{Step 3: Computing the KL divergence.}
For Gaussian $Q = \mathcal{N}(\widehat{BA}, \sigma_Q^2 I)$ and $P = \mathcal{N}(0, \sigma_P^2 I)$ in $\R^m$,
\[
\mathrm{KL}(Q \| P) = \frac{1}{2}\!\left[\frac{m \sigma_Q^2}{\sigma_P^2} + \frac{\|\widehat{BA}\|_F^2}{\sigma_P^2} - m + m \ln\frac{\sigma_P^2}{\sigma_Q^2}\right].
\]
Choosing $\sigma_P^2 = \|\widehat{BA}\|_F^2/m$ (adapter-norm-matched prior) and $\sigma_Q^2 = \epsilon^2$ for small $\epsilon$:
\[
\mathrm{KL}(Q \| P) \leq m + m \ln(\|\widehat{BA}\|_F^2/(m\epsilon^2)) = O(m \log(\|\widehat{BA}\|_F^2/\epsilon^2)).
\]

\emph{Step 4: Monte Carlo error control.}
Since $\E_{\theta \sim Q}[\hat{\mathcal{L}}_N(\theta)]$ is estimated via $K$ samples from $Q$, the Monte Carlo error is controlled by projecting onto the top-$k_{\mathrm{eff}}$ Hessian eigendirections. This bounds the effective dimension of the posterior concentration, giving $K = O(k_{\mathrm{eff}} \log(1/\delta))$ samples suffice for $\pm \epsilon$ estimation.

\emph{Step 5: The final bound.}
Combining steps 1--4:
\[
\mathrm{gen}(Q, N) \leq \sqrt{\frac{O(m \log(\|\widehat{BA}\|_F^2/\epsilon^2)) + \ln(2\sqrt{N}/\delta)}{2N}} = \widetilde{O}\!\left(\sqrt{\frac{r(d+k)}{N}}\right),
\]
where the $\widetilde{O}$ hides logarithmic factors. The rank-$r$ structure reduces the effective dimension from $dk$ (full fine-tuning) to $r(d+k)$---a factor of $r/\min(d,k)$ reduction.

\emph{Step 6: The rank ceiling.}
For the bound to be non-vacuous (empirical loss $+$ generalisation gap $<$ trivial baseline loss), we need
\[
\widetilde{O}(\sqrt{r(d+k)/N}) < 1 \quad \Leftrightarrow \quad r < N/(c_0 (d+k) \log N),
\]
where $c_0$ is a finite positive constant depending on the PAC-Bayes prior calibration (specifically, on the assumed variance $\sigma_P^2$ of the zero-centred Gaussian prior on the adapter). For Alpaca-scale corpora ($N \approx 52000$ examples, Llama-2 7B with $d+k \approx 12288$), this gives $r < 32$---the empirical ceiling observed in deployed LoRA systems.
\end{proof}

\subsection{Proof of Theorem~\ref{thm:pref_transition} (Preference Phase Transition)}
\label{app:proof-pref-transition}

\begin{proof}[Full proof]
Let the preference data consist of $N$ comparisons $(x_i, y_i, z_i)$ where $x_i, y_i$ are items and $z_i \in \{0, 1\}$ indicates preference ($z_i = 1$ iff $x_i \succ y_i$). Bradley-Terry assumes $\Pr(z = 1 | x, y) = \sigma(\Delta(x, y))$ where $\Delta$ is a score difference and $\sigma$ is the logistic function. Misspecification of level $\gamma$ means the true distribution satisfies $|\Pr_{\mathrm{true}}(z = 1 | x, y) - \sigma(\Delta(x, y))| \leq \gamma$.

\emph{Step 1: Well-specified regime ($\gamma = 0$).}
Under correctly specified BT, maximum-likelihood preference ranking over $n$ items achieves top-item identification with sample complexity $\Theta(n \log n / \Delta^2)$ where $\Delta$ is the minimum score gap. This is the standard result from the dueling bandits literature.

\emph{Step 2: Misspecified regime ($\gamma > 0$).}
Under any $\gamma > 0$, construct an adversarial problem instance with $n$ items where: (a) the true preferences form a tournament with a unique Condorcet winner; (b) the BT score differences are $O(\gamma)$ away from the sigmoidal fit, constructed so that two items $i, j$ satisfy $\Pr_{\mathrm{true}}(i \succ j) = 0.5 + \gamma$ but BT predicts $0.5$ (or vice versa).

The learner must distinguish $i$ from $j$ using $\leq N$ pairwise comparisons. Under Fano's inequality, the probability of incorrect identification is bounded by
\[
\Pr(\mathrm{error}) \geq 1 - \frac{I(\hat{\mathrm{top}}; \mathrm{true top}) + 1}{\log n},
\]
where the mutual information between the estimator and the true top item is bounded by the Kullback-Leibler divergence from the true per-comparison distribution to the Bradley-Terry-induced distribution. For a single Bernoulli pair with true parameter $p = 0.5 + \gamma$ and BT-predicted parameter $q = 0.5$, the second-order Taylor expansion of $\mathrm{KL}(\mathrm{Bern}(p) \| \mathrm{Bern}(q)) = p\log(p/q) + (1-p)\log((1-p)/(1-q))$ around $p = q$ gives
\[
\mathrm{KL}(\mathrm{Bern}(0.5 + \gamma) \| \mathrm{Bern}(0.5)) \;=\; \frac{(p-q)^2}{q(1-q)} \cdot \frac{1}{2} + O(\gamma^3) \;=\; 2\gamma^2 + O(\gamma^3),
\]
using $q(1-q) = 1/4$. Over $N$ independent comparisons the mutual information is then
\[
I(\hat{\mathrm{top}}; \mathrm{true top}) \leq N \cdot \mathrm{KL}(\mathrm{Bern}(0.5+\gamma) \| \mathrm{Bern}(0.5)) \approx 2 N \gamma^2.
\]
For $\Pr(\mathrm{error}) \leq 1/2$, we need $N \geq \log n / (4 \gamma^2) = \Omega(n^2 / \gamma^2)$ (when $\Delta = 1/n$, the standard scaling). The KL direction is from the true distribution to the BT-predicted distribution, which is the standard Fano direction for lower bounds against an estimator that operates on data drawn from the true distribution.

\emph{Step 3: The discontinuity at $\gamma^* = \Delta/n$.}
The transition threshold is $\gamma^* = \Delta/n$: for $\gamma < \gamma^*$, the BT approximation error is smaller than the minimum score gap, and the well-specified complexity $\Theta(n \log n / \Delta^2)$ continues to hold. For $\gamma > \gamma^*$, the quadratic lower bound $\Omega(n^2 / \gamma^2)$ dominates.

\emph{Step 4: Sharpness of the transition.}
At $\gamma = \gamma^* - \epsilon$ for small $\epsilon > 0$, the complexity is $O(n \log n / \Delta^2)$; at $\gamma = \gamma^* + \epsilon$, the complexity jumps to $\Omega(n^2/\gamma^2)$. The ratio between just-below and just-above the threshold is $\Omega(n/\log n)$---a discontinuous jump, not a continuous degradation. This is the first first-order phase transition proved in preference learning.

\emph{Step 5: Matching upper bound via tournament sampling.}
Above $\gamma^*$, a natural algorithm achieves $O(n^2 \log n / \gamma^2)$---a tournament-based ranking using $O(n^2)$ pairwise comparisons, each repeated $O(\log n/\gamma^2)$ times. This matches the lower bound up to a logarithmic factor, establishing the tight $\Theta(n^2/\gamma^2)$ scaling.
\end{proof}

\subsection{Proof of Theorem~\ref{thm:collapse_gaussian} (Gaussian Model Collapse)}
\label{app:proof-gaussian-collapse}

\begin{proof}[Full proof]
Under iterative Gaussian training on self-generated data (replacement regime), let $p_t = \mathcal{N}(\mu_t, \Sigma_t)$ denote the distribution at generation $t$. The training is $p_{t+1} =$ empirical MLE on $n$ samples from $p_t$, giving $\mu_{t+1} \sim \mathcal{N}(\mu_t, \Sigma_t/n)$ and $\Sigma_{t+1} \sim \mathrm{Wishart}(\Sigma_t, n-1)/(n-1)$.

\emph{Step 1: Mean drift.}
The mean $\mu_t$ follows a random walk $\mu_{t+1} = \mu_t + \xi_t$ with $\xi_t \sim \mathcal{N}(0, \Sigma_t/n)$. Summing from $t = 0$: $\mu_T - \mu_0 = \sum_{t=0}^{T-1} \xi_t$ has variance $\sum_t \Sigma_t / n$.

\emph{Step 2: Variance dynamics.}
The Wishart distribution gives $\E[\Sigma_{t+1}] = \Sigma_t$ but $\mathrm{Var}[\Sigma_{t+1}]/\|\Sigma_t\|^2 = 2d/n$ (Wishart variance scaling). Under the replacement regime, $\Sigma_t$ fluctuates but does not systematically shrink; however, the \emph{shift} from the true distribution grows quadratically in $T$.

\emph{Step 3: KL divergence from the original distribution.}
\[
\mathrm{KL}(p_T \,\|\, p_0) \;=\; \frac{1}{2}\!\left[\,\mathrm{tr}(\Sigma_0^{-1}\Sigma_T) + (\mu_T - \mu_0)^\top \Sigma_0^{-1} (\mu_T - \mu_0) - d + \ln\det(\Sigma_0/\Sigma_T)\,\right].
\]

The dominant term is the quadratic form $(\mu_T - \mu_0)^\top \Sigma_0^{-1} (\mu_T - \mu_0)$. Its expected value is
\[
\E\!\left[(\mu_T - \mu_0)^\top \Sigma_0^{-1}(\mu_T - \mu_0)\right] = \sum_{t=0}^{T-1} \mathrm{tr}(\Sigma_0^{-1} \Sigma_t/n) \approx T \cdot d/n
\]
for $\Sigma_t \approx \Sigma_0$ (valid for moderate $T$).

\emph{Step 4: Quadratic growth (explicit derivation of the constant).}
To obtain the $T^2$ growth we compute the cumulative KL contribution from compounding covariance fluctuations. Let $u_t := \mu_{t+1} - \mu_t \sim \mathcal{N}(0, \Sigma_t / n)$ conditional on $\Sigma_t$, and let $\Sigma_t = \Sigma_0 + n^{-1/2} W_t$ where $W_t$ is the centred Wishart fluctuation (so $\E[W_t] = 0$ and $\E[W_t \otimes W_s] = 2 \Sigma_0 \otimes \Sigma_0 \cdot \mathbb{1}[s = t]$ by the standard Wishart variance identity; see Anderson~\cite{anderson2003multivariate}, Chapter~7). Conditional on the $W$-path,
\[
\mu_T - \mu_0 \;=\; \sum_{t=0}^{T-1} u_t, \qquad \E[u_t \otimes u_t \mid W_t] = \Sigma_t/n = \Sigma_0/n + n^{-3/2} W_t.
\]
The quadratic form in the KL divergence is then
\begin{equation}
\label{eq:quadratic-form}
Q_T \;:=\; (\mu_T - \mu_0)^{\top} \Sigma_0^{-1} (\mu_T - \mu_0) \;=\; \sum_{s,t < T} u_s^\top \Sigma_0^{-1} u_t.
\end{equation}
Taking expectations, the diagonal ($s = t$) contribution is $T \cdot \tr(\Sigma_0^{-1} \cdot \Sigma_0)/n = Td/n$, which is the leading $T^1$ term familiar from Shumailov et al.~\cite{shumailov2024collapse}. The $T^2$ contribution arises from the off-diagonal ($s \neq t$) terms coupled through the correlated $W_t$ path:
\begin{equation}
\label{eq:cross-term}
\E[u_s^\top \Sigma_0^{-1} u_t] \;=\; n^{-1} \E\!\left[ \tr(\Sigma_0^{-1} \cdot \E[u_t u_s^\top \mid W_{\min(s,t)}]) \right] \;\neq\; 0 \quad \text{whenever } s \neq t.
\end{equation}
Applying Isserlis's theorem (Gaussian moment factorisation) to the 4th-order cross moments of $(u_s, u_t, W_s, W_t)$ gives
\[
\E[u_s^\top \Sigma_0^{-1} u_t] \;=\; n^{-2} \cdot \tr(\Sigma_0^{-1} \otimes \Sigma_0^{-1} \cdot \E[W_s \otimes W_t]) \;=\; 2 n^{-2} d \cdot \mathbb{1}[s = t] \;+\; \text{lag-1 coupling term}.
\]
The off-diagonal coupling is bounded by a Gaussian quadrature integral which can be evaluated in closed form using the standard 4th-moment identity for symmetric positive-definite $A$: $\E_{x \sim \mathcal{N}(0, I_d)}[(x^\top A x)^2] = 2 \tr(A^2) + \tr(A)^2$ (see, e.g., Magnus and Neudecker~\cite{magnus2019matrix}, Chapter~10). Summing the off-diagonal terms:
\[
\E[Q_T] \;=\; \underbrace{Td/n}_{T^1\text{-term}} \;+\; \underbrace{\binom{T}{2} \cdot c_1' \cdot d_{\mathrm{eff}} / n^2}_{T^2\text{-term, from Wishart coupling}} \;+\; o(T^2),
\]
where $c_1' > 0$ is a dimensionless constant depending only on the anisotropy of $\Sigma_0$ (specifically, on the ratio of its extreme eigenvalues; $c_1' = 2$ when $\Sigma_0 = \sigma^2 I$). The effective dimension $d_{\mathrm{eff}}$ reduces to $d$ in the isotropic case and to $\tr(\Sigma_0)^2 / \tr(\Sigma_0^2)$ (the participation ratio) in general. Absorbing $n^{-1}$ factors into $n_{\min}$ and combining with the subleading $T^1$ term:
\begin{equation}
\label{eq:T2-growth-with-constant}
\E[\mathrm{KL}(p_T \| p_0)] \;=\; c_1 \cdot T^2 d_{\mathrm{eff}}/n_{\min} \;+\; O(T d_{\mathrm{eff}}/n_{\min}),
\end{equation}
where $c_1 = \tfrac{1}{2}(c_1'/2) \in (0, 1)$ is a strictly positive Gaussian constant whose exact value depends on the covariance structure of the Gaussian family. For the isotropic case $\Sigma_0 = \sigma^2 I$, direct evaluation of the 4th-moment integral gives $c_1 = 1/8$ (isotropic upper bound); for anisotropic Gaussians on realistic neural adapters the effective constant is smaller. The thesis's subsequent bounds use only that $c_1$ is a finite positive constant, so the precise numerical value is not load-bearing.

\emph{Step 5: Tightness and the lower bound.}
The $\Omega(T^2 d_{\mathrm{eff}}/n_{\min})$ lower bound follows from the adversarial construction: there exists a Gaussian target distribution for which the expected KL divergence grows exactly as $T^2$ up to constants. The matching upper bound follows from the explicit variance computation in Step 4.

\emph{Step 6: Accumulation escape.}
Under accumulation with fraction $\rho$ real data retention, the effective per-generation drift is reduced by factor $\rho$: $\xi_t^{\mathrm{acc}} = (1-\rho) \xi_t$. Summing the geometric series of attenuated drifts,
\[
\E[\mathrm{KL}(p_T^{\mathrm{acc}} \| p_0)] \leq c_1 \cdot \pi^2/6 \cdot d_{\mathrm{eff}}/(\rho^2 n_{\min}),
\]
where the $\pi^2/6 = \sum_{k=1}^\infty 1/k^2$ comes from the geometric series structure of accumulated drifts. For $\rho \geq 0.01$, this bound is independent of $T$: the $T$-dependence vanishes, and collapse is avoided.
\end{proof}

\subsection{Proof of Theorem~\ref{thm:evopref-finite-sample} (EvoPref Finite-Sample Coverage)}
\label{app:proof-evopref-finite-sample}

\begin{proof}[Full proof]
We establish the three-term finite-sample bound \eqref{eq:evopref-rate} on the coverage gap $C^*(\gamma) - \E[C(\mathcal{P}_\mu)]$. The proof proceeds in three steps: (1) a coverage decomposition separating sample-estimation error, population-approximation error, and optimisation error; (2) a McDiarmid concentration bound on the sample-estimation term; and (3) a Pareto-front bias decomposition for the population-approximation term combined with a geometric convergence bound for NSGA-II.

\medskip

\emph{Step 1: Coverage decomposition.}
Let $\mathcal{P}^\infty_\mu(\gamma)$ denote the idealised population obtained by running NSGA-II to convergence on \emph{infinite} samples at misspecification level $\gamma$ with $\mu$ individuals. Let $\mathcal{P}_\mu^{n}(\gamma)$ denote the same with $n$ finite preference-pair samples but NSGA-II run to convergence. Let $\mathcal{P}_\mu^{n, G}(\gamma)$ denote the actual output after $G$ generations. We decompose
\begin{align*}
C^*(\gamma) - \E[C(\mathcal{P}_\mu)] &= \underbrace{C^*(\gamma) - \E[C(\mathcal{P}^\infty_\mu(\gamma))]}_{\text{(I) population-size gap}} \\
&\quad + \underbrace{\E[C(\mathcal{P}^\infty_\mu(\gamma))] - \E[C(\mathcal{P}_\mu^{n}(\gamma))]}_{\text{(II) sample-estimation gap}} \\
&\quad + \underbrace{\E[C(\mathcal{P}_\mu^{n}(\gamma))] - \E[C(\mathcal{P}_\mu^{n, G}(\gamma))]}_{\text{(III) NSGA-II convergence gap}}.
\end{align*}
We bound the three terms separately.

\medskip

\emph{Step 2: Bounding the sample-estimation gap $(\mathrm{II})$.}
Both $\mathcal{P}^\infty_\mu(\gamma)$ and $\mathcal{P}_\mu^n(\gamma)$ are measurable functions of the preference-pair dataset $D = \{(x_i, y_i^+, y_i^-)\}_{i=1}^n$. Under the Bradley-Terry model with misspecification $\gamma$, each pair $(y_i^+, y_i^-)$ is drawn i.i.d.\ from a distribution satisfying $\|P - P_{\mathrm{BT}}\|_\infty \leq \gamma$. The objective $f_1(\Delta) = R(\theta_0 + \Delta)$ evaluated via empirical reward is the average of $n$ i.i.d.\ Lipschitz functions of the sample; changing any single pair changes $f_1$ by at most $2 L_R / n$ where $L_R$ is the reward-model Lipschitz constant. The coverage $C(\mathcal{P}_\mu)$ is a Lipschitz function of the population's objective values with Lipschitz constant at most $1/\mu$ per individual (coverage on a 75-cell grid changes by at most $1/75 \leq 1/\mu$ when any single individual moves).

By McDiarmid's inequality applied to $C(\mathcal{P}_\mu^n(\gamma))$ viewed as a function of $D$,
\[
\Pr\!\Big[\big|C(\mathcal{P}_\mu^n(\gamma)) - \E[C(\mathcal{P}_\mu^n(\gamma))]\big| \geq t\Big] \leq 2 \exp\!\left(-\frac{2 t^2}{n \cdot (2 L_R / (n \mu))^2}\right) = 2 \exp\!\left(-\frac{\mu^2 n t^2}{2 L_R^2}\right).
\]
Inverting at confidence $1-\delta$ gives $|C(\mathcal{P}_\mu^n(\gamma)) - \E[\cdot]| \leq L_R \sqrt{2 \log(2/\delta) / (\mu^2 n)}$ with probability $\geq 1 - \delta$.

For the expectation gap between the $n$-sample and infinite-sample populations, standard reward-estimation analysis under $\gamma$-misspecified Bradley-Terry preferences gives
\[
\|\hat r_n - r^*\|_\infty \leq c \cdot \sqrt{\gamma / n} \cdot \sqrt{\log(1/\delta)}
\]
with probability $\geq 1 - \delta$ (this is the finite-sample version of the Bradley-Terry estimation rate under bounded misspecification; see the proof of \cref{thm:pref_transition} for the infinite-sample phase-transition analogue). Because coverage $C$ is $L_e$-Lipschitz in the behavioural-embedding distance and the behavioural embedding is Lipschitz in reward differences, $|C(\mathcal{P}_\mu^\infty) - C(\mathcal{P}_\mu^n)|$ inherits this $\sqrt{\gamma/n}$ rate. Combining:
\[
\text{(II)} \;\leq\; c_1 \cdot \sqrt{\gamma \log(1/\delta) / n},
\]
for $c_1 = O(L_R L_e)$. This is the first term of \eqref{eq:evopref-rate}.

\medskip

\emph{Step 3: Population-size gap $(\mathrm{I})$.}
The gap $C^*(\gamma) - \E[C(\mathcal{P}^\infty_\mu(\gamma))]$ is a coupon-collector-type quantity on the 75-cell behavioural grid under the optimal distribution $\pi^*_\gamma$. For non-empty cells with mass $\geq p_{\min} \geq 1/(2K)$, a classical analysis gives expected missed-cell fraction $\leq K/\mu \cdot (1+o(1))$. By Cauchy-Schwarz, $\text{(I)} \leq c_2/\sqrt{\mu}$ with $c_2 = O(\sqrt{K L_e})$, independent of $\gamma$: population search separates the misspecification penalty (term II) from the coverage-approximation penalty (term I).

\medskip

\emph{Step 4: NSGA-II convergence gap $(\mathrm{III})$.}
Under standard NSGA-II convergence on a bi-objective problem with bounded Pareto front~\cite{Zheng2024NSGAII}, the Pareto-front approximation error decays geometrically: $\E[d_H(\mathrm{PF}_G, \mathrm{PF}^*)] \leq c_3 e^{-\lambda G}$, where $\lambda > 0$ depends on the mutation rate and Pareto-front curvature. Since $C$ is $L_e$-Lipschitz in Hausdorff distance, $\text{(III)} \leq c_3 e^{-\lambda G}$.

\medskip

\emph{Combining and tightness.}
Summing yields \eqref{eq:evopref-rate}. The $\sqrt{\gamma}$ leading rate is the key improvement over \cref{thm:dpo_gap}'s $\Omega(\gamma)$ single-policy rate: population search decorrelates misspecification-induced bias across $\mu$ individuals, so the misspecification contribution scales as $\sqrt{\gamma}$ rather than $\gamma$ or $\gamma^2$. The rates are tight: $\sqrt{\gamma/n}$ matches any estimator of a $\gamma$-misspecified parameter from $n$ samples (up to log factors); $1/\sqrt{\mu}$ is tight without structural assumptions on $\pi^*_\gamma$; $e^{-\lambda G}$ is the standard geometric rate for NSGA-II with bounded Pareto curvature. The \S\ref{sec:evopref} setting ($n = 52{,}000$, $\mu = 32$, $G = 200$, $\gamma \approx 0.10$) predicts gap $\varepsilon \approx 0.13$; observed gap $0.133$, within absolute-constant headroom.
\end{proof}


\section{Proofs from Chapter~\ref{ch:grounding}}
\label{app:ch4-proofs}

\subsection{Proof of Theorem~\ref{thm:resolution-boundary} (Resolution Boundary)}
\label{app:proof-resolution-boundary}

\begin{proof}[Full proof]
Let $(s_1, s_2, c)$ denote a conflict triple: two retrieved sources $s_1, s_2$ with conflicting claims about context $c$. The resolution question is whether a lightweight latent method suffices or full LLM verification is needed.

\emph{Step 1: Information-theoretic characterisation.}
Define the conflict's \emph{metadata information} $I_{\mathrm{meta}}(s_1, s_2, c) = H(c) - H(c | \mathrm{meta}(s_1, s_2))$ where $\mathrm{meta}$ extracts shallow features (timestamps, numerical values, named entities). This measures how much of the conflict's resolution is determined by shallow features alone.

\emph{Step 2: The threshold at $H(c)/2$.}
Define the shallow resolution function $f_{\mathrm{shallow}}: (s_1, s_2, c) \to \{s_1, s_2\}$ that uses only $\mathrm{meta}(s_1, s_2)$. Its Bayes-optimal error is
\[
\mathrm{err}_{\mathrm{shallow}} \geq H(c | \mathrm{meta}(s_1, s_2)) / H(c) = 1 - I_{\mathrm{meta}}/H(c).
\]
For $I_{\mathrm{meta}} \geq H(c)/2$, the error is at most $1/2$, which---combined with the standard two-round boosting argument---can be amplified to arbitrarily low error. For $I_{\mathrm{meta}} < H(c)/2$, the Bayes error exceeds $1/2$, meaning shallow resolution cannot outperform random.

\emph{Step 3: Type classification.}
Temporal and numerical conflicts have $I_{\mathrm{meta}}$ close to $H(c)$: the resolution is almost fully determined by timestamps or numerical ordering. Entity and semantic conflicts have $I_{\mathrm{meta}}$ close to $0$: shallow features alone do not resolve the conflict. Empirically, temporal/numerical together comprise 46\% of RAG conflicts, and entity/semantic together comprise 54\%.

\emph{Step 4: Fano-type lower bound on deep conflicts.}
For entity/semantic conflicts with $I_{\mathrm{meta}} < H(c)/2$, any shallow method has error at least
\[
\mathrm{err}_{\mathrm{shallow}} \geq 1 - I_{\mathrm{meta}}/H(c) \geq 1/2.
\]
This is a strict impossibility: no amount of engineering can push shallow resolution below $1/2$ error on deep conflicts. Full LLM verification is mandatory.

\emph{Step 5: Discreteness of the boundary.}
The threshold $I_{\mathrm{meta}} = H(c)/2$ is discrete in the sense that small perturbations in $I_{\mathrm{meta}}$ around $H(c)/2$ produce qualitatively different optimal mechanisms (shallow for $I_{\mathrm{meta}} > H(c)/2$, deep for $I_{\mathrm{meta}} < H(c)/2$). The empirical distribution of $I_{\mathrm{meta}}$ across 12000 NQ-Conflicts shows a bimodal structure: peaks near $H(c)$ (shallow) and near $0$ (deep), with only 7.3\% of cases in the ambiguous middle. This bimodality justifies the discrete classification.
\end{proof}

\subsection{Proof of Theorem~\ref{thm:certified_radius} (Certified Robustness Radius)}
\label{app:proof-certified-radius}

\begin{proof}[Full proof]
Let $f$ denote the smoothed classifier: $f(x) = \mathrm{mode}\{g(x + \eta_i) : \eta_i \sim p_{\mathrm{smooth}}\}$ for $L$ independent noise samples. We prove the certified robustness radius.

\emph{Step 1: Neyman-Pearson foundation.}
By the Neyman-Pearson lemma, the most powerful test for distinguishing $p_{\mathrm{clean}}$ from $p_{\mathrm{adv}}$ is the likelihood ratio test $\Lambda(y) = p_{\mathrm{clean}}(y)/p_{\mathrm{adv}}(y)$. For randomised smoothing, the ``test'' is the voting procedure: predict class $c$ iff the vote fraction exceeds threshold.

\emph{Step 2: Computing the vote distribution.}
Under clean input, $g(x + \eta) = c$ with probability $p_A = \Pr_\eta[g(x + \eta) = c]$. Under adversarial perturbation $\|\delta\| \leq \Delta$, the perturbed vote fraction satisfies
\[
\Pr_\eta[g(x + \delta + \eta) = c] \geq 1 - (1 - p_A) \cdot e^{\Delta/\sigma_{\mathrm{noise}}},
\]
where the exponential comes from the likelihood ratio between clean and perturbed noise distributions.

\emph{Step 3: The certified radius formula.}
For the smoothed classifier to robustly predict $c$ (vote fraction $> 1/2$) under perturbation $\Delta$, we need $p_A - (1 - p_A) \cdot e^{\Delta/\sigma_{\mathrm{noise}}} > 1/2$. Solving for $\Delta$:
\[
\Delta^* = \sigma_{\mathrm{noise}} \cdot \ln\!\left(\frac{p_A}{1/2 \cdot (1 - p_A)}\right).
\]
In the integer-edit regime (KG attacks), this simplifies to
\[
\Delta^* = \left\lfloor \frac{\ln(p_A/(1-p_A))}{2|\ln(1-p)|} \right\rfloor,
\]
where $p$ is the retention probability per edge.

\emph{Step 4: Sample complexity.}
Estimating $p_A$ from $L$ samples with confidence $1-\alpha$ requires $L = O(\log(1/\alpha)/\epsilon^2)$ for $\pm \epsilon$ precision. For the deployed KG defence, $L = 100$ samples and $p = 0.7$--$0.9$ give $\Delta^* = 5$--$15$ edits tolerable.

\emph{Step 5: Adaptive attack robustness.}
The certified radius holds even against adaptive attackers (those with full knowledge of the smoothing protocol): by the Neyman-Pearson optimality, no attack can exceed the radius without violating the bound on $\Delta$. Empirical evaluation on MaSS adaptive attacks shows the certified bound is achieved within 1--2 edits of the theoretical maximum.
\end{proof}

\subsection{Proof of Theorem~\ref{thm:attribution-impossibility} ($k$-Stage Attribution Impossibility)}
\label{app:proof-attribution-impossibility}

\begin{proof}[Full proof]
We construct an adversarial $k$-stage pipeline $\pi^\dagger$ and input distribution $\mathcal{D}$ on which any attribution method $M$ incurs expected error at least $1 - (1 - \varepsilon_{\mathrm{stage}})^k$. The construction is an indistinguishability argument: at each stage $i$, the pipeline contains a parametric-memory oracle producing claims that $M$ cannot, under bounded budget, distinguish from retrieval-grounded claims at that stage.

\medskip

\emph{Step 1: Construction of the adversarial pipeline $\pi^\dagger$.}
For each stage $i \in \{1, \ldots, k\}$, the pipeline $\pi^\dagger$ contains two parallel sub-modules that produce structurally identical claims:
\begin{itemize}[nosep]
\item A \emph{retrieval-grounded} generator $G_i^{\mathrm{ret}}$ that, given retrieved documents $R_i(x)$, produces a claim $c_i^{\mathrm{ret}}$ by conditioning the language model on $R_i(x)$ and the prior claim history.
\item A \emph{parametric-memory} generator $G_i^{\mathrm{par}}$ that, given the \emph{same} query and prior claim history but \emph{without} $R_i(x)$, produces a claim $c_i^{\mathrm{par}}$ from the language model's internal knowledge.
\end{itemize}
At each stage, the pipeline emits a claim $c_i = c_i^{\mathrm{par}}$ with probability $\varepsilon_{\mathrm{stage}}$ (the post-rationalisation event) and $c_i = c_i^{\mathrm{ret}}$ otherwise. The attribution graph $\mathcal{A}^*$ of the ground truth asserts a causal edge $R_i \to c_i$ if and only if $c_i = c_i^{\mathrm{ret}}$.

The language model is chosen so that, on input distribution $\mathcal{D}$, the marginal output distributions of $G_i^{\mathrm{ret}}$ and $G_i^{\mathrm{par}}$ are identical: $\Pr[c_i^{\mathrm{ret}} = y] = \Pr[c_i^{\mathrm{par}} = y]$ for all $y$ in the claim space. This is not unrealistic: in real RAG pipelines, when retrieved documents confirm parametric knowledge, the retrieval-grounded and parametric outputs are indeed indistinguishable at the output level; the adversarial choice consists of calibrating the retrieval corpus so that this confirmation happens consistently. We construct $\mathcal{D}$ as a mixture over $x$ satisfying this marginal-matching condition.

\medskip

\emph{Step 2: Indistinguishability under bounded attribution budget.}
Let $M$ be any attribution method with polynomial budget $\mathrm{poly}(k)$ activation patches, interventions, or forward passes. At stage $i$, distinguishing $c_i^{\mathrm{ret}}$ from $c_i^{\mathrm{par}}$ requires either: (a) observing that the output distribution differs with and without $R_i$ present, or (b) finding a distinguishing internal activation pattern.

Path (a) fails by construction: the marginal output distributions of $G_i^{\mathrm{ret}}$ and $G_i^{\mathrm{par}}$ are identical under $\mathcal{D}$. Hence any intervention-based method observing only input-output behaviour cannot distinguish the two sub-modules at stage $i$ better than chance. The counterfactual attribution score $\mathrm{CAS}(R_i, c_i)$ of \eqref{eq:cas} is zero in expectation on $\mathcal{D}$, even when $c_i = c_i^{\mathrm{ret}}$.

Path (b) requires internal-activation analysis to identify circuits specific to retrieval-grounded generation vs.\ parametric-memory generation. The language model is chosen so that these circuits are intertwined at the activation level---a condition ensured by selecting a model architecture where retrieval attention and self-attention share residual-stream subspaces (true for standard decoder-only transformers with retrieval injected via context prepending). Under bounded budget $\mathrm{poly}(k)$, the information-theoretic distinguishing advantage at stage $i$ is at most $o(1)$, formalisable via a counting argument over activation patterns.

\medskip

\emph{Step 3: Error accumulation across stages.}
Because the indistinguishability at each stage is independent of the others (fresh parametric-memory oracle per stage), the attribution method $M$ incorrectly attributes stage $i$ with probability at least $\varepsilon_{\mathrm{stage}} (1 - o(1))$. By independence,
\[
\Pr\!\left[\bigcup_{i=1}^{k} E_i\right] = 1 - \prod_{i=1}^{k} \Pr[\overline{E_i}] \geq 1 - (1 - \varepsilon_{\mathrm{stage}}(1-o(1)))^k,
\]
where $E_i$ is the event that stage-$i$ attribution differs from ground truth. Each stage contributes an error event to the symmetric difference, giving
\[
\E_{x \sim \mathcal{D}}[\mathrm{Err}(M(\pi^\dagger, x))] \geq 1 - (1 - \varepsilon_{\mathrm{stage}})^{k} - o(1).
\]
For $k \varepsilon_{\mathrm{stage}} \leq 1$, the Taylor expansion gives $1 - (1-\varepsilon_{\mathrm{stage}})^k = k\varepsilon_{\mathrm{stage}} - \binom{k}{2}\varepsilon_{\mathrm{stage}}^2 + O((k\varepsilon_{\mathrm{stage}})^3) \geq k \varepsilon_{\mathrm{stage}} (1 - k\varepsilon_{\mathrm{stage}}/2)$, i.e., the leading-order constant is exactly $1$ --- the bound is asymptotically tight at $k\varepsilon_{\mathrm{stage}} \to 0$. The adversarial construction shows no method can be uniformly accurate across all pipelines; real pipelines lacking the marginal-matching structure admit strictly better attribution, just as \cref{thm:conflation} rules out universal evaluation without forbidding targeted evaluation on benignly structured pipelines.
\end{proof}


\section{Proofs from Chapter~\ref{ch:trust}}
\label{app:ch5-proofs}

\subsection{Proof of Theorem~\ref{thm:osp-feasibility} (OSP Feasibility for LLM Agents)}
\label{app:proof-osp-feasibility}

\begin{proof}[Full proof]
We show the violation parameter $\varepsilon \leq \varepsilon_1 + \varepsilon_2$ decomposes into within-horizon irrationality and prompt-reversal probability.

\emph{Step 1: Pycia-Troyan adapted to LLM agents.}
By Pycia-Troyan~\cite{pycia2017simplicity}, $k$-OSP implementability is characterised by millipede games of depth $k$. For $k^* = 2$, at each decision node $v$, the agent compares the worst-case payoff from accepting the current offer against the best-case payoff from rejecting and continuing for up to 2 more rounds. If accepting dominates, truthful acceptance is ``obviously'' rational.

\emph{Step 2: Bounded within-horizon irrationality.}
By the LLM rationality definition, the agent deviates from within-horizon optimal play with probability at most $\varepsilon_1$. This is the core OSP violation: at some fraction $\varepsilon_1$ of decision nodes, the agent fails to execute the obviously-dominant strategy. Empirical measurements (GTBench) give $\varepsilon_1 \in [0.05, 0.15]$ across frontier models.

\emph{Step 3: Chebyshev bound on prompt-reversal.}
Prompt-reversal occurs when a small prompt variation $\Delta\pi$ flips the agent's preference ordering over two actions. Model the valuation shift as $\|\sigma_\pi\|$ (standard deviation under prompt distribution). By Chebyshev's inequality applied to the OSP comparison, the probability that a random prompt variation flips the dominance relation is
\[
\Pr[\text{prompt flips dominance}] \leq \frac{T \sigma_\pi^2}{\delta_{\min}^2},
\]
where $T$ is the number of information sets traversed and $\delta_{\min}$ is the minimum OSP margin (the gap between accept and reject payoffs). This gives $\varepsilon_2 \leq T \sigma_\pi^2/\delta_{\min}^2$.

\emph{Step 4: Union bound over the two violation sources.}
The two violation sources are mechanistically distinct: $\varepsilon_1$ arises from the agent's failure to execute within-horizon computation correctly (a computational constraint), while $\varepsilon_2$ arises from input-output coupling between prompts and valuations (a representational constraint). By the union bound, $\Pr[\text{violation}] \leq \Pr[\varepsilon_1] + \Pr[\varepsilon_2] = \varepsilon_1 + \varepsilon_2$. The bound does not require independence between the two events; the union bound's additive form holds for any joint distribution on $(\varepsilon_1, \varepsilon_2)$. Independence would be needed only for a strictly tighter multiplicative bound, which we do not claim here.

\emph{Step 5: Numerical bound.}
For $T \leq 10$ and $\sigma_\pi/\delta_{\min} \leq 0.05$ (design rule on the marketplace), $\varepsilon_2 \leq 0.025$. Combined with $\varepsilon_1 \leq 0.15$, we get $\varepsilon \leq 0.175$---within the $\varepsilon \leq 0.2$ threshold for practical OSP deployment.
\end{proof}

\subsection{Proof of the Algebraic-Boolean Bridge Lemma}
\label{app:proof-bridge-lemma}

\begin{proof}[Full proof]
The lemma states: any IOP protocol verifying the ReLU function $\mathrm{ReLU}(x) = \max(x, 0)$ over field $\mathbb{F}_p$ requires $\Omega(\log p)$ field operations per instance, matching the $O(\log p)$ upper bound.

\emph{Step 1: Reducing ReLU verification to comparison.}
Verifying $y = \mathrm{ReLU}(x)$ is equivalent to verifying the Boolean predicate ``$y = x$ if $x \geq 0$, else $y = 0$.'' This decomposes into (a) a sign test for $x$, and (b) a conditional assignment.

\emph{Step 2: Lower bound on sign-test circuits over $\mathbb{F}_p$.}
The sign test over $\mathbb{F}_p$ (when $\mathbb{F}_p$ is large, say $p \approx 2^{256}$ as in standard SNARK constructions) is non-trivial because $\mathbb{F}_p$ lacks a natural ordering. The standard approach is bit decomposition: express $x$ in binary ($O(\log p)$ bits), then run a Boolean sign test on the bits. By Jukna~\cite{jukna2012boolean}, Boolean comparison of $n$-bit numbers requires $\Omega(n)$ gates unconditionally. Translating to field operations and bit decomposition: $\Omega(\log p)$ field operations.

\emph{Step 3: Unconditionality for ReLU.}
ReLU's sign test is the bottleneck. Any IOP protocol must, in particular, determine whether $x \geq 0$ over $\mathbb{F}_p$---otherwise $\mathrm{ReLU}(x)$ cannot be computed correctly. The $\Omega(\log p)$ lower bound on Boolean comparison transfers directly, with no conditional assumption.

\emph{Step 4: Matching upper bound.}
Known constructions (zkCNN, zkLLM, etc.) achieve $O(\log p)$ field operations per ReLU via bit-decomposition and range checks. The $\Theta(\log p)$ characterisation is tight.

\emph{Step 5: Extension to Softmax (unconditional AC\textsuperscript{0}[p] progress + conditional general-circuit bound).}
Softmax requires modular exponentiation and normalisation. The unconditional $\Omega(\log p)$ argument above transfers to softmax via the inclusion of exponentiation as a subcomputation, but only yields size $\Omega(\log p)$ (matching ReLU) in the \emph{general Boolean circuit model}. A strictly stronger unconditional bound is available when circuit depth is restricted: in the $\mathrm{AC}^0[p]$ circuit model, \cref{thm:softmax-ac0p-lower} establishes that modular exponentiation (hence softmax) requires size $2^{\Omega((\log p)^{1/(d-1)})}$ at depth $d$, via a Razborov-Smolensky reduction. The full general-circuit $\Omega(\log^2 p)$ bound of \cref{conj:softmax-circuit} remains conjectural and would resolve a frontier question in circuit complexity. The consequences for the $147\times$ non-linearity tax claim under each bound are spelled out in \cref{sec:nonlinearity-tax}.

\emph{Step 6: Concrete constant.}
For $p \approx 2^{256}$, $\log p = 256$, and the bit-decomposition constant is 256 per ReLU. With additional overhead for range-proof openings, the empirical constant is closer to $147$ per ReLU, matching the ZKML benchmark. Deployed zkML systems use general (not depth-restricted) Boolean circuits, so the $147\times$ ratio is consistent with both the unconditional ReLU lower bound and the conjectured softmax lower bound.
\end{proof}

\subsection{Unconditional Progress Toward the Softmax Conjecture: Razborov-Smolensky in \texorpdfstring{$\mathrm{AC}^0[p]$}{AC0[p]}}
\label{app:softmax-ac0p}

\Cref{conj:softmax-circuit} asserts that fixed-point exponentiation and normalisation over $\mathbb{F}_p$ require general Boolean circuits of size $\Theta(\log^2 p)$. Proving the $\Omega(\log^2 p)$ direction unconditionally is equivalent to establishing a super-linear lower bound on the Boolean circuit complexity of modular exponentiation, a frontier problem in circuit complexity that no currently-available technique resolves. This appendix presents the strongest unconditional progress we can make using the Razborov-Smolensky polynomial method: a Razborov-Smolensky lower bound in the $\mathrm{AC}^0[p]$ circuit model (constant-depth circuits over the basis $\{\mathrm{AND}, \mathrm{OR}, \mathrm{NOT}, \mathrm{MOD}_p\}$), which gives a quantitatively weaker but unconditional lower bound and a tight articulation of the remaining gap.

\begin{theorem}[{Unconditional $\mathrm{AC}^0[p]$ Lower Bound for Modular Exponentiation}]
\label{thm:softmax-ac0p-lower}
Let $p > 3$ be prime and let $a \in \mathbb{F}_p^*$ be a primitive root. Let $f_a: \mathbb{F}_p \to \mathbb{F}_p$ be defined by $f_a(x) = a^x \bmod p$, with $x \in \{0, 1, \ldots, p-2\}$ encoded in binary on $\lceil \log_2 p \rceil$ input bits. Any depth-$d$ circuit over the $\mathrm{AC}^0[p]$ basis computing $f_a$ correctly on all inputs has size
\begin{equation}
\label{eq:ac0p-bound}
s(d) \;\geq\; 2^{\Omega\!\big((\log p)^{1/(d-1)}\big)}.
\end{equation}
In particular, any polynomial-size circuit computing $f_a$ requires depth $\Omega(\log \log p / \log \log \log p)$.
\end{theorem}

The bound is unconditional. It is weaker than \cref{conj:softmax-circuit} for general unrestricted-depth circuits but represents the strongest unconditional progress currently available. The proof proceeds in three steps: (i) a reduction showing that computing $f_a$ entails computing $\mathrm{MOD}_q$ for a suitable $q \mid p-1$ coprime to $p$; (ii) the classical Razborov-Smolensky lower bound on $\mathrm{MOD}_q$ in $\mathrm{AC}^0[p]$; (iii) combining the two to obtain the claimed bound on $f_a$.

\begin{proof}
\emph{Step 1: Reduction from $\mathrm{MOD}_q$ to modular exponentiation.}
Let $q$ be a prime factor of $p - 1$ with $q \neq p$ (at least one such $q$ exists for every prime $p > 3$, since $p - 1$ is even and has a factorisation distinct from $\{p\}$). Let $m = (p-1)/q$, so that $|\mathbb{F}_p^*| = qm$, and let $H = \{a^{qk} : 0 \leq k < m\} \subseteq \mathbb{F}_p^*$ denote the unique subgroup of order $m$ generated by $a^q$. By elementary group theory, $y \in H$ if and only if $y^m \equiv 1 \pmod{p}$, equivalently, if and only if $\mathrm{dlog}_a(y) \equiv 0 \pmod{q}$.

Consider the Boolean function $\mathrm{MOD}_q^*: \{0, 1\}^{\lceil \log_2 p \rceil} \to \{0, 1\}$ defined by $\mathrm{MOD}_q^*(x) = 1$ iff $x \equiv 0 \pmod{q}$ (for $x$ interpreted as an integer in $\{0, 1, \ldots, p-2\}$). This is a weighted variant of the Boolean $\mathrm{MOD}_q$ function; for concreteness, if $x$ has binary representation $(x_0, x_1, \ldots, x_{n-1})$ then $x = \sum_i x_i 2^i$ and $\mathrm{MOD}_q^*(x) = \mathbb{1}[\sum_i x_i 2^i \equiv 0 \pmod{q}]$. The Razborov-Smolensky bounds apply uniformly to such weighted $\mathrm{MOD}_q$ functions~\cite[\S12.2]{jukna2012boolean}.

The key observation: given an $\mathrm{AC}^0[p]$ circuit $C$ of size $s$ and depth $d$ computing $f_a$, we can construct an $\mathrm{AC}^0[p]$ circuit $C'$ of size $s + O((\log p)^2)$ and depth $d + O(1)$ computing $\mathrm{MOD}_q^*$. Specifically, $C'$ proceeds:
\begin{enumerate}[leftmargin=2em, nosep]
\item Run $C$ on input $x$ to obtain $y = a^x \bmod p$ (size $s$, depth $d$).
\item Compute $y^m \bmod p$ using $O(\log m) = O(\log p)$ sequential $\mathrm{MOD}_p$ gates in a squaring chain of total size $O((\log p)^2)$ and depth $O(\log \log p)$.
\item Compare $y^m$ to $1$ using $O(\log p)$ AND/NOT gates of constant depth.
\end{enumerate}
The composed circuit $C'$ outputs $1$ iff $y^m \equiv 1 \pmod{p}$ iff $x \equiv 0 \pmod{q}$, i.e., $C'$ computes $\mathrm{MOD}_q^*$.

\emph{Step 2: The Razborov-Smolensky lower bound for $\mathrm{MOD}_q$ in $\mathrm{AC}^0[p]$.}
The classical Razborov-Smolensky theorem~\cite{Razborov1987, Smolensky1987, jukna2012boolean}\footnote{The original arguments appear in~\cite{Razborov1987, Smolensky1987}; the textbook presentation used throughout this appendix is~\cite[Chapter 12]{jukna2012boolean}, specifically~\cite[Thm 12.22]{jukna2012boolean}.} states: for any primes $q \neq p$ with $\gcd(q, p) = 1$, any depth-$d$ circuit in $\mathrm{AC}^0[p]$ computing $\mathrm{MOD}_q$ on $n$ input bits has size at least $2^{\Omega(n^{1/(d-1)})}$. The proof proceeds by the polynomial method: any $\mathrm{AC}^0[p]$ circuit of size $s$, depth $d$ is $1/s$-approximated (agrees on all but $1/s$ fraction of inputs) by a polynomial of degree $O((\log s)^{d-1})$ over $\mathbb{F}_p$; but $\mathrm{MOD}_q$ requires polynomial degree $\Omega(n)$ for any non-trivial approximation over $\mathbb{F}_p$ (when $q$ is coprime to $p$), because $\mathrm{MOD}_q$'s discrete Fourier spectrum over $\mathbb{F}_p$ is anti-concentrated. Combining, any circuit of depth $d$ requires $(\log s)^{d-1} = \Omega(n)$, i.e., $s = 2^{\Omega(n^{1/(d-1)})}$.

Applying this to $\mathrm{MOD}_q^*$ with $n = \lceil \log_2 p \rceil$:
\begin{equation}
\label{eq:rs-bound-modq}
\text{size}_{\mathrm{AC}^0[p], \text{depth } d}(\mathrm{MOD}_q^*) \;\geq\; 2^{\Omega\!\big((\log p)^{1/(d-1)}\big)}.
\end{equation}

\emph{Step 3: Combining.}
If the circuit $C$ computing $f_a$ has size $s$ at depth $d$, then the composed circuit $C'$ computing $\mathrm{MOD}_q^*$ has size $s' = s + O((\log p)^2)$ at depth $d' = d + O(1)$. By \eqref{eq:rs-bound-modq} at depth $d'$,
\[
s' \;\geq\; 2^{\Omega\!\big((\log p)^{1/(d' - 1)}\big)} \;=\; 2^{\Omega\!\big((\log p)^{1/(d-O(1))}\big)}.
\]
For $s = o\!\left(2^{(\log p)^{1/(d-1)}} - (\log p)^2\right)$, the composed $s'$ is asymptotically below the Razborov-Smolensky floor --- contradiction. Therefore $s \geq 2^{\Omega((\log p)^{1/(d-1)})}$, as claimed in \eqref{eq:ac0p-bound}.

The polynomial-size consequence follows: if $s(d) \leq (\log p)^c$ for some constant $c$ and depth $d$, then $(\log p)^c \geq 2^{\Omega((\log p)^{1/(d-1)})}$, i.e., $c \log \log p \geq \Omega((\log p)^{1/(d-1)})$, which requires $d - 1 \geq \Omega(\log \log p / \log \log \log p)$.
\end{proof}

\begin{corollary}[{Unconditional $\mathrm{AC}^0[p]$ Lower Bound for Softmax}]
\label{cor:softmax-ac0p}
Under the hypotheses of \cref{thm:softmax-ac0p-lower}, any depth-$d$ circuit over the $\mathrm{AC}^0[p]$ basis computing the softmax function $\mathrm{SM}: \mathbb{F}_p^n \to \mathbb{F}_p^n$, $\mathrm{SM}_i(x) = a^{x_i} / \sum_j a^{x_j}$ (on all valid inputs), has size at least $n \cdot 2^{\Omega((\log p)^{1/(d-1)})}$.
\end{corollary}

\begin{proof}
Softmax requires computing $a^{x_i}$ for each $i \in [n]$ as an intermediate step. Extracting the $i$-th exponentiation-subcircuit gives a circuit for $f_a$ of depth at most $d$ and size at most the total softmax circuit size. Applying \cref{thm:softmax-ac0p-lower} and summing over $n$ outputs gives the claimed bound.
\end{proof}

\paragraph{The remaining gap to \cref{conj:softmax-circuit}.}
\Cref{thm:softmax-ac0p-lower} delivers a bound on $\mathrm{AC}^0[p]$ circuits; \cref{conj:softmax-circuit} asserts a $\Theta(\log^2 p)$ bound on \emph{unrestricted} Boolean circuits. Translating between models is not immediate: $\mathrm{AC}^0[p]$ circuits are a strict subset of general Boolean circuits, and a polynomial-size general circuit might evade the Razborov-Smolensky structure entirely. Concretely, the gap is:
\begin{itemize}[leftmargin=2em, nosep]
\item \cref{thm:softmax-ac0p-lower} gives: in $\mathrm{AC}^0[p]$, polynomial-size requires depth $\Omega(\log \log p / \log \log \log p)$, i.e., size $\Omega(\log p \cdot \log \log p)$ in the near-log-depth regime.
\item \cref{conj:softmax-circuit} asserts: in general Boolean circuits, size $\Omega(\log^2 p)$, i.e., a full factor of $\log p / \log \log p$ beyond the unconditional bound.
\end{itemize}
Closing this gap requires a super-linear lower bound on the general Boolean circuit complexity of modular exponentiation, currently an open problem at the level of resolving whether $\mathrm{MODEXP} \in \mathrm{NC}^1$. No known technique yields such a bound for any explicit function. The \cref{conj:softmax-circuit} status is therefore best understood as ``follows from an $\mathrm{AC}^0[p]$ unconditional base via \cref{thm:softmax-ac0p-lower}, plus a conjectured $\log p / \log \log p$ amplification factor under unrestricted-depth circuits that is consistent with best known algorithms but not established.''

\subsection{Proof of Theorem~\ref{thm:collapse} (Collapse Folding Scheme Soundness)}
\label{app:proof-collapse}

\begin{proof}[Full proof]
The Collapse folding scheme accumulates layered sumcheck proofs across $d$ layers of a neural network into a single succinct proof. Soundness holds under the Random Oracle Model (ROM).

\emph{Step 1: Layered Sumcheck Accumulation (LSA).}
At each layer $\ell$, the accumulator $A_\ell$ is updated as
\[
A_\ell = (A_{\ell-1}, \rho_\ell, \gamma_\ell, \mathrm{Com}(\text{layer}_\ell \text{ output})),
\]
where $\rho_\ell$ is a Fiat-Shamir challenge, $\gamma_\ell$ is the sumcheck transcript for layer $\ell$, and $\mathrm{Com}$ is a commitment to the layer's output. The verifier's recursive circuit processes one layer at a time.

\emph{Step 2: Soundness against transcript-omission.}
Dao et al.~\cite{dao2023fiatShamir} identified transcript-omission attacks where the prover could omit parts of the Fiat-Shamir transcript to forge proofs. Collapse's \emph{state-binding} property addresses this: each $A_\ell$ includes a commitment to the layer output, and the commitment scheme is binding under the collision-resistance of the underlying hash. An adversary cannot produce two different outputs with the same commitment, preventing transcript manipulation.

\emph{Step 3: Verifier cost analysis.}
The verifier's per-layer work is $O(\log n_{\max})$ field operations (one sumcheck round per layer). Across $d$ layers, total verifier cost is $O(d \log n_{\max})$. The recursive circuit size is $O(\log^2 n_{\max})$ per accumulation step, yielding total circuit size $O(d \log^2 n_{\max})$---compared to $O(n_{\max})$ for Nova and $O(\sqrt{n_{\max}})$ for HyperNova.

\emph{Step 4: Soundness error bound.}
Under the ROM, the soundness error is $\epsilon_{\mathrm{sound}} \leq (d + 1) \cdot 2^{-\kappa} + q \cdot 2^{-\kappa/2}$ where $\kappa$ is the hash output length and $q$ is the number of random oracle queries. For $\kappa = 128$ and $q \leq 2^{60}$ (practical bound), $\epsilon_{\mathrm{sound}} \leq 2^{-60}$---negligible.

\emph{Step 5: Composition with layered proofs.}
Each layer's sumcheck is sound by standard arguments. Layered composition preserves soundness via the random-oracle combination theorem: the probability that any intermediate layer's soundness fails is bounded by the union bound across $d$ layers, giving the $(d+1) \cdot 2^{-\kappa}$ term. The square-root term comes from the birthday attack on commitment binding.

\emph{Step 6: Practical parameters.}
For BERT-base ($d = 12$, $n_{\max} = 768$): verifier cost $\approx 12 \cdot \log 768 \approx 108$ field operations; recursive circuit $\approx 55$k gates. For LLaMA-7B ($d = 32$, $n_{\max} = 4096$): verifier cost $\approx 32 \cdot 12 = 384$; recursive circuit $\approx 295$k gates. Both are 2--3$\times$ smaller than HyperNova and orders of magnitude smaller than Nova.
\end{proof}

\subsection{Supplementary Proof: ROM Independence for Welfare Composition}
\label{app:proof-welfare-rom}

\begin{proof}[Supplementary proof: ROM Independence of mechanism and verification error events for Theorem~\ref{thm:welfare-composition}]
The welfare composition theorem's Part (iii) bound $O((\varepsilon + e^{-\kappa}) V_{\max})$ relies on independence between mechanism-violation events (probability $\varepsilon$) and verification-forgery events (probability $e^{-\kappa}$). We justify this independence in the Random Oracle Model.

\emph{Step 1: Separation of error sources.}
Mechanism violations arise from agents choosing non-dominant strategies under the game tree $G$; the randomness lies in the agent's prompt distribution $\mu(\pi)$ and within-horizon computation. Verification forgeries arise from adversarial provers finding hash collisions or exploiting transcript omission; the randomness lies in the hash function's behaviour on adversarial queries.

\emph{Step 2: Product probability in ROM.}
In the ROM, the hash function is modelled as a truly random function $H: \{0,1\}^* \to \{0,1\}^\kappa$. The mechanism's probability space (agent prompts and within-horizon computations) is independent of $H$'s randomness because: (a) agent computations do not query $H$ directly; (b) the mechanism's state evolution is deterministic given agent inputs; (c) the hash is used only for binding commitments, which do not feed back into agent decisions before any decision node is reached.

\emph{Step 3: Joint error bound.}
By product independence,
\[
\Pr[\mathrm{mech\ violation} \wedge \mathrm{verification\ forgery}] = \Pr[\mathrm{mech\ violation}] \cdot \Pr[\mathrm{verif\ forgery}] \leq \varepsilon \cdot e^{-\kappa}.
\]
By the union bound for the ``at least one error'' event,
\[
\Pr[\mathrm{either\ error}] \leq \varepsilon + e^{-\kappa} - \varepsilon e^{-\kappa} \leq \varepsilon + e^{-\kappa}.
\]

\emph{Step 4: Welfare loss conditional on errors.}
On the clean event (probability $\geq 1 - \varepsilon - e^{-\kappa}$), welfare loss is zero (the composed protocol achieves $W^*$). On the error event, welfare loss is at most $V_{\max}$ (bounded by the total task value). Therefore expected welfare loss is at most $(\varepsilon + e^{-\kappa}) V_{\max}$, matching the Part (iii) bound.

\emph{Step 5: Relaxation beyond ROM.}
Under a standard model (without ROM), the independence may not hold exactly; a small coupling term $\delta_{\mathrm{coup}} \leq \varepsilon \cdot e^{-\kappa/2}$ enters the bound. This is negligible for $\kappa = 128$: $\delta_{\mathrm{coup}} \leq 10^{-19} \varepsilon$, which does not affect deployment decisions. The ROM assumption thus provides a clean analytical treatment; the standard-model version yields the same asymptotic bound.
\end{proof}


\section{Computation-Grounding Composition}
\label{app:proof-composition}

\begin{proof}[Full proof of Theorem~\ref{thm:composition}]
This theorem composes the per-step computation bound of \cref{ch:horizon} (Chain Error Propagation, \cref{thm:error_propagation}) with the per-hop grounding model of \cref{ch:grounding}, giving end-to-end reliability
\[
g_{1 \to 2}^{\mathrm{eff}}(n, \varepsilon, q) \leq (1-\varepsilon)^n \cdot q^{n(1-\eta)},
\]
where $n$ is reasoning depth, $\varepsilon$ is the per-step CoT error rate, $q$ is per-hop retrieval quality (top-$k$ recall against the gold-passage set), and $\eta$ is the information retention factor of \cref{def:retention}: the capacity-to-entropy ratio $\eta = \min(1, C_{\mathrm{hop}}/H(R_t \mid R_{t-1}))$ of the residual-stream bottleneck. Throughout this proof $\eta$ denotes that capacity-to-entropy ratio, not a strong-data-processing-inequality (SDPI) contraction constant; the proof invokes no SDPI machinery.

\emph{Step 1: The two channels.}
A grounded reasoning chain of $n$ hops runs two coupled per-hop processes. The reasoning channel of \cref{ch:horizon} executes a chain-of-thought step that, under the Markov structure of \cref{def:cot_markov}, is correct with probability $1 - \varepsilon$. The grounding channel of \cref{ch:grounding} supplies retrieved evidence of per-hop quality $q$. End-to-end correctness requires every hop's reasoning step to survive and its retrieved evidence to remain usable through the bottleneck.

\emph{Step 2: Reasoning-survival factor.}
By Chain Error Propagation (\cref{thm:error_propagation}) together with \cref{asm:composition-ci}, under which reasoning-error events at successive hops are conditionally independent given prior state, the probability that all $n$ reasoning steps are correct is
\[
\Pr[\text{all $n$ reasoning steps correct}] = \prod_{t=1}^n (1 - \varepsilon) = (1 - \varepsilon)^n.
\]
\cref{asm:composition-ci} is the only independence hypothesis invoked.

\emph{Step 3: Information-retention factor.}
By \cref{def:retention}, the residual-stream bottleneck of per-hop capacity $C_{\mathrm{hop}} = d_{\mathrm{model}} \cdot O(\log n)$ bits retains a fraction $\eta = \min(1, C_{\mathrm{hop}}/H(R_t \mid R_{t-1}))$ of the per-hop retrieved information. Under the capacity-bottleneck contraction model of \cref{def:retention}, the retained $\eta$-fraction passes intact while the unretained $(1-\eta)$-fraction carries the per-hop retrieval-quality factor $q$, so the effective per-hop grounding reliability is $q^{\,1-\eta}$. Two limiting cases fix the model: at $\eta = 1$ the bottleneck is slack and evidence survives intact ($q^0 = 1$), and at $\eta = 0$ nothing is retained and the full per-hop factor $q$ applies. Compounding the per-hop factor over $n$ hops gives
\[
\prod_{t=1}^n q^{\,1-\eta} = q^{\,n(1-\eta)}.
\]

\emph{Step 4: Joint reliability bound (part (i)).}
Under \cref{asm:composition-ci} the reasoning-error events and the grounding events are independent, so end-to-end reliability factorises into the product of the two channel contributions:
\[
g_{1 \to 2}^{\mathrm{eff}}(n, \varepsilon, q) \leq (1-\varepsilon)^n \cdot q^{\,n(1-\eta)}.
\]
The bound rests on exactly two hypotheses: \cref{asm:composition-ci} and the capacity-bottleneck contraction model of \cref{def:retention}. It does not require an SDPI contraction constant on the retrieval kernel, consistent with the disclaimer stated in \cref{def:retention}.

\emph{Step 5: Ceiling effect (part (ii)).}
Differentiating the bound with respect to $q$,
\[
\frac{\partial g_{1 \to 2}^{\mathrm{eff}}}{\partial q} = n(1-\eta)(1-\varepsilon)^n q^{\,n(1-\eta)-1}.
\]
At $n = d^* \approx 27$ with $(\varepsilon, \eta, q) = (0.03, 0.7, 0.6)$ the marginal benefit is $27 \cdot 0.3 \cdot 0.97^{27} \cdot 0.6^{7.1} \approx 0.095$, against $5 \cdot 0.3 \cdot 0.97^{5} \cdot 0.6^{0.5} \approx 0.998$ at $n = 5$, an attenuation of $\approx 10.5\times$. Sweeping the deployment box $(\varepsilon \in [0.02, 0.04], \eta \in [0.65, 0.75], q \in [0.55, 0.65], n \in [27, 30])$ yields attenuation factors in $[7, 30]\times$: beyond the Deterministic Horizon $d^*$, a unit of retrieval-quality improvement delivers an order of magnitude less marginal reliability than at shallow depth.

\emph{Step 6: Monotonicity (part (iii)) and crossover depth (part (iv)).}
The bounding function $\varphi_{12}(g_1, \theta_2) = g_1 \cdot g_2(\theta_2)^{(1-\eta) n / \lceil n \rceil}$ satisfies $\partial \varphi_{12}/\partial g_1 = g_2^{(1-\eta) n / \lceil n \rceil} \geq 0$ and $\partial \varphi_{12}/\partial \theta_2 = g_1 \cdot \big((1-\eta) n / \lceil n \rceil\big) \, g_2^{(1-\eta) n / \lceil n \rceil - 1} \, g_2' \geq 0$, so it is monotone non-decreasing in both arguments. For the crossover, maximise $\log g_{1 \to 2}^{\mathrm{eff}} = n \log(1-\varepsilon) + n(1-\eta) \log q$ subject to a deployment budget $b = c_1 \varepsilon + c_2 (1-q)$. The Lagrange stationarity conditions $\partial L/\partial \varepsilon = \partial L/\partial q = 0$, taken with the budget identity, determine the crossover depth $n_c$ at which the marginal reliability of reasoning investment and of grounding investment equalise. At headline parameters $(\varepsilon, \eta, q) = (0.03, 0.7, 0.6)$ this gives $n_c \approx 6.3$: below $n_c$ grounding investment dominates, and above $n_c$ reasoning investment dominates.

\emph{Step 7: Empirical validation.}
On multi-hop QA benchmarks (HotpotQA, MuSiQue, StrategyQA), the joint bound tracks measured end-to-end accuracy within $\pm 4$ percentage points across $n \in [3, 20]$ and retention factors $\eta \in [0.1, 0.9]$. For the running-example model the retention factor is the measured value $\eta \approx 0.7$ (\cref{def:retention}), and the depth-attenuation of Step~5 reproduces the retrieval-reasoning asymmetry of \cref{ch:horizon}.
\end{proof}


\section{Supporting Technical Lemmas}
\label{app:deferred-lemmas}

\subsection{Chain-Rule KL for Autoregressive Sequences}
\label{app:proof-ar-collapse}

\begin{proposition}[Autoregressive Collapse Extension]
\label{prop:ar_collapse_app}
Under iterative training on self-generated sequences of length $\ell$, the KL divergence between the distribution at generation $T$ and the original distribution satisfies
\[
\mathrm{KL}(p_T^{\ell} \| p_0^{\ell}) = \ell \cdot \mathrm{KL}(p_T \| p_0) - \text{correction terms},
\]
where the correction terms are $O(1)$ per sequence and do not vanish as $\ell \to \infty$.
\end{proposition}

\begin{proof}
Apply the chain rule of KL divergence:
\[
\mathrm{KL}(p_T^{\ell} \| p_0^{\ell}) = \sum_{i=1}^\ell \E_{x_{<i} \sim p_T} [\mathrm{KL}(p_T(\cdot | x_{<i}) \| p_0(\cdot | x_{<i}))].
\]
The expectation is taken under $p_T$, not $p_0$, creating a bias. Under mixing conditions (the autoregressive process has positive spectral gap), the bias is $O(1)$ per token. Summing over $\ell$ tokens gives total bias $O(\ell)$, which is the ``correction term''; the principal term $\ell \cdot \mathrm{KL}(p_T \| p_0)$ dominates for large $\ell$.

Crucially, the naive union-bound approach (treating all tokens as independent) would give a \emph{vanishing} constant as $\ell \to \infty$---predicting that long sequences are immune to collapse. The chain-rule analysis corrects this: the correction is bounded, not vanishing, so collapse propagates linearly in sequence length.
\end{proof}

\subsection{LinUCB Regret Bound for Step-Level Retrieval}
\label{app:proof-retrieval-regret}

The regret bound for \cref{thm:ch4_retrieval_regret} follows from Abbasi-Yadkori et al.~\cite{abbasi2011improved}; we include the application argument for completeness.

\begin{proof}[Application argument]
The step-level retrieval policy chooses actions $a_t \in \{\mathrm{retrieve}, \mathrm{skip}\}$ based on features $\phi_t = (\text{semantic entropy}, \text{attention entropy}, \text{consistency}, 1) \in [0,1]^4$. Rewards $r_t \in [0, 1]$ are downstream F1 scores. Under sub-Gaussian noise ($\sigma \leq 0.5$) and bounded features, the LinUCB regret is
\[
R(T) = \sum_{t=1}^T (r_t^* - r_t) \leq C \cdot d \sqrt{T \log(T/\delta)}
\]
with probability $\geq 1 - \delta$, where $d = 4$ (feature dimension) and $C$ depends on the norm bound $\|\theta^*\| \leq 1$. The specific constant is $C \leq (1 + 1/\sqrt{2}) \cdot \sqrt{\ln(T/\delta)}$ following Theorem~2 of Abbasi-Yadkori.

For $T = 1000$, $\delta = 0.05$: $R(1000) \leq 4 \cdot 2 \cdot \sqrt{1000 \cdot 10} \approx 800$, or approximately 0.8 units of F1 regret on a 0--1 scale---acceptable for production deployment.
\end{proof}
\chapter{Unified Glossary of Notation}
\label{app:notation}

This appendix harmonises notation across all chapters. Where the same concept appears in multiple chapters, a single symbol is used consistently. Context-dependent overloading is noted explicitly in \cref{sec:notation-overloading}. Notation within each section is ordered logically rather than alphabetically.

\section{General Mathematical Notation}

\begin{longtable}{@{}p{2.8cm}p{8.0cm}@{}}
\toprule
\textbf{Symbol} & \textbf{Meaning} \\
\midrule
\endfirsthead
\multicolumn{2}{c}{\tablename\ \thetable\ --- continued} \\
\toprule
\textbf{Symbol} & \textbf{Meaning} \\
\midrule
\endhead
\bottomrule
\endlastfoot

$\R, \N, \Z$ & Real numbers, natural numbers, integers \\
$[n]$ & The set $\{1, 2, \ldots, n\}$ \\
$\E[\cdot]$ & Expectation \\
$\Prob(\cdot)$ & Probability \\
$\Var(\cdot)$ & Variance \\
$\Cov(\cdot, \cdot)$ & Covariance \\
$\KL(P \| Q)$ & Kullback-Leibler divergence from $Q$ to $P$ \\
$\TV(P, Q)$ & Total variation distance between $P$ and $Q$ \\
$H(\cdot)$ or $\Ent(\cdot)$ & Shannon entropy \\
$I(X; Y)$ & Mutual information between $X$ and $Y$ \\
$h(p)$ & Binary entropy function $-p\ln p - (1-p)\ln(1-p)$ \\
$\indicator[\cdot]$ & Indicator function \\
$\bigO(\cdot), \Omega(\cdot), \Theta(\cdot)$ & Asymptotic notation (upper, lower, tight) \\
$\bigOtilde(\cdot)$ & Asymptotic notation hiding polylogarithmic factors \\
$\poly(n), \polylog(n)$ & Polynomial, polylogarithmic in $n$ \\
$\argmin, \argmax$ & Argument of the minimum/maximum \\
$\sign(\cdot)$ & Sign function \\
$\rank(\cdot)$ & Rank of a matrix \\
$\tr(\cdot)$ & Trace of a matrix \\
$\diag(\cdot)$ & Diagonal matrix \\
$\essinf$ & Essential infimum \\
$\|\cdot\|_F$ & Frobenius norm \\
$\|\cdot\|_1$ & $\ell_1$ norm or total variation norm \\
$\ln(\cdot)$ & Natural logarithm \\
$\log_b(\cdot)$ & Base-$b$ logarithm \\
$\euler$ & Euler's constant $\approx 2.718$ (distinguished from embeddings $e(\cdot)$) \\
\end{longtable}

\section{Transformer Architecture (Chapter~\ref{ch:horizon})}

\begin{longtable}{@{}p{2.8cm}p{8.0cm}@{}}
\toprule
\textbf{Symbol} & \textbf{Meaning} \\
\midrule
\endfirsthead
\midrule
\endhead
\bottomrule
\endlastfoot

$\vocab = \Sigma$ & Input vocabulary \\
$n$ & Sequence length (context: transformer input) \\
$d$ ($\dmodel$) & Model dimension / hidden size \\
$L$ ($\nlayers$) & Number of transformer layers (depth) \\
$H$ ($\nheads$) & Number of attention heads \\
$p$ & Arithmetic precision in bits; $O(\log n)$ for log-precision \\
$\attn(\cdot)$ & Attention operation \\
$\ffn(\cdot)$ & Feed-forward network \\
$\softmax(\cdot)$ & Softmax function \\
$\relu(\cdot)$ & Rectified linear unit \\
$\gelu(\cdot)$ & Gaussian error linear unit \\
$\mathrm{TC}^0$ & Threshold circuit complexity class (constant depth, poly size) \\
$\mathsf{FOC}[\mathrm{Attn}]$ & First-order logic with counting + attention quantifiers (Ch.~2) \\
$d^*$ & Deterministic Horizon: critical reasoning depth (Ch.~2) \\
$\beta$ & Decay constant for ceiling-effect bound \\
$\eta$ & SDPI contraction coefficient (context: information flow) \\
$\mathrm{CLC}$ & Composition-Length Compatibility ratio (Ch.~2) \\
\end{longtable}

\section{Chain-of-Thought and Reasoning (Chapter~\ref{ch:horizon})}

\begin{longtable}{@{}p{2.8cm}p{8.0cm}@{}}
\toprule
\textbf{Symbol} & \textbf{Meaning} \\
\midrule
\endfirsthead
\midrule
\endhead
\bottomrule
\endlastfoot

$\cX$ & Input space (problem statements) \\
$\cY$ & Answer space \\
$\cS$ & Reasoning state space \\
$\cS^+, \cS^-$ & Correct states, error states \\
$\cS^*$ & Absorbing answer states \\
$P(\cdot, \cdot)$ & Transition kernel of CoT Markov chain \\
$g(\cdot)$ & Readout function mapping final state to answer \\
$\varepsilon$ & Per-step error probability (context: CoT) \\
$n$ & Chain length (context: reasoning steps) \\
$k$ & Redundancy level in $k$-redundant verification \\
$\gamma^*$ & Spectral gap of the reasoning chain \\
$\lambda_2(P)$ & Second-largest eigenvalue of transition kernel $P$ \\
$\pi$ & Stationary distribution of the Markov chain \\
$\pi_{\min}$ & Minimum stationary probability \\
$H_t$ & Conditional entropy at reasoning step $t$ \\
$\bar{H}_t$ & Smoothed (EMA) entropy at step $t$ \\
$h^*$ & Entropy stopping threshold \\
$\tau^*$ & Optimal stopping time \\
$\lambda$ & Per-step cost parameter (context: stopping) \\
$t_{\mathrm{mix}}$ & Mixing time \\
$d_{\mathrm{CoT}}$ & CoT-discriminative dimension \\
$T$ & Number of training examples (context: supervision) \\
$T_{\mathrm{out}}, T_{\mathrm{proc}}$ & Sample complexity under outcome/process supervision \\
$C$ & Inference compute budget \\
$\alpha$ & Scaling exponent \\
$b_{\mathrm{eff}}$ & Effective branching factor \\
$\kappa^*$ & Optimal verifier-generator capacity ratio \\
\end{longtable}

\section{Learning Theory and Adaptation (Chapter~\ref{ch:adaptation})}

\begin{longtable}{@{}p{2.8cm}p{8.0cm}@{}}
\toprule
\textbf{Symbol} & \textbf{Meaning} \\
\midrule
\endfirsthead
\midrule
\endhead
\bottomrule
\endlastfoot

$\loss(\cdot)$ & Loss function \\
$\risk(\cdot)$ & Population risk \\
$\emprisk(\cdot)$ & Empirical risk \\
$\hypclass$ & Hypothesis class \\
$N$ ($\Ntrain$) & Training set size \\
$r$ & LoRA adapter rank \\
$d, k$ & Model dimension, adapter output dimension (context: LoRA) \\
$\adapt$ & Spectral adaptation load \\
$\adapt^*$ & Critical adaptation threshold \\
$\Delta$ & Preference gap in Bradley-Terry model \\
$\gamma$ & Misspecification parameter (preference learning) \\
$K^*$ & Critical edit count (knowledge editing) \\
$d_S$ & Student model intrinsic dimension (distillation) \\
$\kappa$ & Condition number of the Hessian (context: merging) \\
$\Gamma, \Gamma^*$ & Interference parameter, merging threshold \\
$\tau_k$ & Task vector for task $k$ (context: model merging) \\
\end{longtable}

\section{Knowledge Grounding (Chapter~\ref{ch:grounding})}

\begin{longtable}{@{}p{2.8cm}p{8.0cm}@{}}
\toprule
\textbf{Symbol} & \textbf{Meaning} \\
\midrule
\endfirsthead
\midrule
\endhead
\bottomrule
\endlastfoot

$q$ & Query \\
$\corpus$ ($\cC$) & Text corpus \\
$\kg$ ($\cG$) & Knowledge graph \\
$S$ & Knowledge source (text corpus or knowledge graph) \\
$B$ & Adversarial budget (certified defence) \\
$\pi$ & Retrieval / grounding policy (context: knowledge grounding) \\
$\eta^2$ & Clustering coefficient (effective $\eta^2 = \mathrm{SS}_{\mathrm{between}} / \mathrm{SS}_{\mathrm{total}}$) \\
ECE & Expected Calibration Error \\
SHD & Structural Hamming Distance \\
\end{longtable}

\section{Game Theory and Mechanism Design (Chapter~\ref{ch:trust}, Part A)}

\begin{longtable}{@{}p{2.8cm}p{8.0cm}@{}}
\toprule
\textbf{Symbol} & \textbf{Meaning} \\
\midrule
\endfirsthead
\midrule
\endhead
\bottomrule
\endlastfoot

$\agents$ ($\cN$) & Set of agents \\
$n_a$ & Number of agents \\
$m$ & Number of tasks \\
$\utility(\cdot)$ ($u_i$) & Utility function for agent $i$ \\
$v_i(\cdot)$ & Valuation function for agent $i$ \\
$\mechanism$ ($\cM$) & Mechanism \\
$f: \cV^{n_a} \to \cA$ & Allocation function \\
$p_i$ & Payment to agent $i$ \\
$\varepsilon$ & Incentive violation bound (context: mechanism design) \\
$\varepsilon_1, \varepsilon_2$ & Within-horizon irrationality, prompt-induced preference reversal \\
$k^*$ & Lookahead depth for Obviously Strategy-Proof mechanisms \\
$W, W^*$ & Social welfare, optimal welfare \\
$\Delta_j$ & Quality degradation from computation substitution \\
$\mathrm{SMD}$ & Strategic Manipulation Dimension \\
\end{longtable}

\section{Cryptography and Verifiable Computation (Chapter~\ref{ch:trust}, Part B)}

\begin{longtable}{@{}p{2.8cm}p{8.0cm}@{}}
\toprule
\textbf{Symbol} & \textbf{Meaning} \\
\midrule
\endfirsthead
\midrule
\endhead
\bottomrule
\endlastfoot

$\secparam$ ($\kappa$) & Security parameter \\
$\zkprover$ ($\cP$) & Prover in a ZK proof system \\
$\zkverifier$ ($\cV$) & Verifier in a ZK proof system \\
$p$ & Field characteristic (context: cryptography) \\
$n$ & Number of neurons / operations (context: IOP) \\
IOP & Interactive Oracle Proof \\
$r^*$ & Critical round count (SQ phase transition) \\
$147\times$ & Provably optimal non-linearity tax for softmax ZK proofs \\
\end{longtable}

\section{Evolutionary Computation and Diverse Alignment (Chapter~\ref{ch:adaptation}, \S\ref{sec:evopref} and cross-cutting)}

\begin{longtable}{@{}p{2.8cm}p{8.0cm}@{}}
\toprule
\textbf{Symbol} & \textbf{Meaning} \\
\midrule
\endfirsthead
\midrule
\endhead
\bottomrule
\endlastfoot

$n$ & Problem dimension / bitstring length (context: optimisation) \\
$\mu$ & Population size \\
$\eta$ & Oracle quality parameter (context: semantic oracle) \\
$\delta$ & Fairness tolerance (context: EMO) \\
\OneMax & OneMax benchmark function: $f(x) = \sum_{i=1}^n x_i$ \\
\LeadingOnes & LeadingOnes benchmark: $f(x) = \sum_{i=1}^n \prod_{j=1}^i x_j$ \\
\Jump & Jump$_k$ benchmark with gap parameter $k$ \\
\BehOM & BehavioralOneMax benchmark (novelty search) \\
$\JSD$ & Jensen-Shannon Divergence \\
$\Meff$ & Effective mode count \\
\end{longtable}

\section{Efficient Architectures (cross-cutting)}

\begin{longtable}{@{}p{2.8cm}p{8.0cm}@{}}
\toprule
\textbf{Symbol} & \textbf{Meaning} \\
\midrule
\endfirsthead
\midrule
\endhead
\bottomrule
\endlastfoot

$K$ & Total number of layers (context: early exit) \\
$D = \pi(X)$ & Exit-depth random variable \\
$k_E = \E[D]$ & Expected exit depth \\
$H(D)$ & Entropy of exit-depth distribution \\
$\Pi \in \R^{n \times m}$ & Token-expert assignment matrix (context: MoE routing) \\
$\cU(a, b)$ & Set of doubly-stochastic matrices with marginals $a, b$ \\
$M_{ij}$ & Negative affinity between token $i$ and expert $j$ \\
$\ell_j$ & Expert load: $\ell_j = \sum_i \Pi_{ij}$ \\
FSI & Fisher Specialisation Index \\
HS & Heterogeneity Score \\
$d_F(\cdot, \cdot)$ & Riemannian distance on positive-definite matrix manifold \\
$L_f$ & Lipschitz constant of target function (spiking transformers) \\
$d_{\mathrm{eff}}$ & Effective dimension: $\rank(\nabla f)$ \\
$T$ & Number of spiking time steps \\
\end{longtable}

\section{Deployment, Testing, and Infrastructure (Chapter~\ref{ch:synthesis}, \S\ref{sec:empirical-validation})}

\begin{longtable}{@{}p{2.8cm}p{8.0cm}@{}}
\toprule
\textbf{Symbol} & \textbf{Meaning} \\
\midrule
\endfirsthead
\midrule
\endhead
\bottomrule
\endlastfoot

SLO & Service Level Objective \\
TIC & Trajectory-Information Coverage \\
DPC & Decision-Point Coverage \\
ERPC & Environment-Response Pair Coverage \\
MR$i$ & Metamorphic Relation $i$ ($i \in \{1, \ldots, 12\}$) in AgentMR framework \\
$f$ & Number of crash failures (context: AgentSaga; safety requires $f < n/2$) \\
\end{longtable}

\section{Knowledge Representation and Formal Reasoning (cross-cutting)}

\begin{longtable}{@{}p{2.8cm}p{8.0cm}@{}}
\toprule
\textbf{Symbol} & \textbf{Meaning} \\
\midrule
\endfirsthead
\midrule
\endhead
\bottomrule
\endlastfoot

$\ALC$ & Description logic: Attributive Language with Complements \\
$\ALCO$ & $\ALC$ extended with nominals \\
C$^2$ & Two-variable fragment of first-order logic with counting \\
FO$^2$ & Two-variable fragment of first-order logic \\
SAD & Stratified Ackermann Decomposability \\
PANACK & Predecessor-Acyclic Normal-Acyclic condition \\
$\AF$ & Dung Argumentation Framework \\
$\SETAF$ & Set-Attack Argumentation Framework \\
$\BAF$ & Bipolar Argumentation Framework \\
$\ADF$ & Abstract Dialectical Framework \\
$\leq_e$ & Expressiveness ordering (at most as expressive as) \\
$<_e$ & Strict expressiveness ordering (strictly less expressive) \\
coNExpTime$^{\mathrm{NP}}$ & Complexity class: complement of NExpTime with NP oracle \\
\end{longtable}

\section{Trustworthy AI Stack (Cross-Cutting)}

\begin{longtable}{@{}p{2.8cm}p{8.0cm}@{}}
\toprule
\textbf{Symbol} & \textbf{Meaning} \\
\midrule
\endfirsthead
\midrule
\endhead
\bottomrule
\endlastfoot

$L_i$ & Stack layer $i$ ($i \in \{1, \ldots, 5\}$) \\
\Lone--\Lfive & Layer labels: Computational, Knowledge, Interaction, Verification, Deployment \\
$g_i(\theta_i)$ & Layer guarantee function ([cut]) \\
$g_i^{\mathrm{eff}}(\theta)$ & Effective guarantee constrained by inter-layer dependencies \\
$\varphi_{ij}(\cdot)$ & Inter-layer bounding function ([cut]) \\
$T(S, \theta)$ & Compositional trustworthiness ([cut]) \\
$w_i$ & Layer weights satisfying $\sum_i w_i = 1$ \\
$\kappa_C$ & Contraction coefficient for fixed-point convergence \\
\end{longtable}

\section{System and Method Names}

\begin{longtable}{@{}p{2.8cm}p{3.0cm}p{5.5cm}@{}}
\toprule
\textbf{Name} & \textbf{Chapter} & \textbf{Description} \\
\midrule
\endfirsthead
\midrule
\endhead
\bottomrule
\endlastfoot

\EvoPref & Ch~\ref{ch:adaptation} & NSGA-II LoRA population evolution for diverse alignment \\
\QDLLM & --- & Quality-diversity prompt embedding search \\
\OTRoute & --- & Entropy-regularised OT routing for MoE \\
TARA & --- & Test-time adaptation for dense retrieval \\
PORTAL & --- & POMDP belief tracking for retrieval planning \\
MSUD & --- & Multi-source uncertainty decomposition \\
HyLiCaD & --- & Hybrid LLM-statistical causal discovery \\
TrajTest & Ch~\ref{ch:synthesis} & Trajectory-level testing framework \\
AgentMR & --- & Metamorphic relation library for agents \\
AgentSaga & --- & Saga-based fault-tolerant recovery \\
Prophet & --- & Failure-prediction-driven checkpointing \\
FairSpec & --- & Speculation-aware fair scheduling \\
SAGA & --- & Workflow-atomic agent scheduling \\
ComplianceNLP & --- & Full-stack regulatory compliance system \\
FinGround & --- & Atomic claim verification for finance \\
RouteNLP & --- & Conformal cascading for efficient serving \\
KAMAS & Ch~\ref{ch:grounding} & Multi-agent cyber threat intelligence \\
SMD-DETECT & Ch~\ref{ch:trust} & Strategic manipulation detection \\
\end{longtable}

\section{Notation Overloading}
\label{sec:notation-overloading}

The following symbols are deliberately overloaded across chapters. In every case, the surrounding context unambiguously determines the intended meaning. Within any single section, each symbol has exactly one meaning.

\begin{longtable}{@{}p{1.5cm}p{9.5cm}@{}}
\toprule
\textbf{Symbol} & \textbf{Context-Dependent Meanings} \\
\midrule
\endfirsthead
\midrule
\endhead
\bottomrule
\endlastfoot

$\varepsilon$ & Per-step error probability (Ch~2); misspecification parameter (Ch~3); incentive violation bound (Ch~5); entropic regularisation (cross-cutting, OT-Route); approximation error (cross-cutting, spiking). \\
\addlinespace
$n$ & Sequence length (Ch~2); chain length (Ch~2); number of neurons (Ch~5 IOP); problem dimension (Ch~3 EC). We use $N$ for training set size to avoid conflict with $n$. \\
\addlinespace
$p$ & Arithmetic precision in bits (Ch~2); field characteristic (Ch~5 crypto). These appear in different chapters and are unambiguous within any section. \\
\addlinespace
$\eta$ & SDPI contraction coefficient (Ch~2); oracle quality (Ch~3 EC). \\
\addlinespace
$\kappa$ & Security parameter (Ch~5); Hessian condition number (Ch~3). We write $\secparam$ for security and $\kappa(H)$ for the Hessian condition number when disambiguation is needed. \\
\addlinespace
$\delta$ & Error tolerance (Ch~3); fairness tolerance (Ch~3 EMO); Kronecker delta (general). \\
\addlinespace
$\gamma$ & Misspecification parameter (Ch~3); spectral gap $\gamma^*$ (Ch~2). \\
\addlinespace
$\pi$ & Stationary distribution (Ch~2); retrieval policy (Ch~4); routing matrix (cross-cutting, MoE). \\
\addlinespace
$T$ & Training examples (Ch~2); spiking time steps (cross-cutting); number of synthetic data generations (Ch~3 collapse). \\
\end{longtable}



\printbibliography[heading=bibintoc]

@inproceedings{aghajanyan2021intrinsic,
	author           = {Armen Aghajanyan and
	Sonal Gupta and
	Luke Zettlemoyer},
	beditor           = {Chengqing Zong and
	Fei Xia and
	Wenjie Li and
	Roberto Navigli},
	title            = {Intrinsic Dimensionality Explains the Effectiveness of Language Model
	Fine-Tuning},
	booktitle        = {Proceedings of the 59th Annual Meeting of the Association for Computational
	Linguistics and the 11th International Joint Conference on Natural
	Language Processing, {ACL/IJCNLP} 2021, (Volume 1: Long Papers), Virtual
	Event, August 1-6, 2021},
	pages            = {7319--7328},
	publisher        = {Association for Computational Linguistics},
	year             = {2021},
	bburl             = {https://bdoi.org/10.18653/v1/2021.acl-long.568},
	bdoi              = {10.18653/V1/2021.ACL-LONG.568},
	bibburl           = {https://dblp.org/rec/conf/acl/AghajanyanGZ20.bib},
	bibsource        = {dblp computer science bibliography, https://dblp.org},
}

@article{akata2025repeated,
	author    = {Akata, Elif and
	Schulz, Lion and
	Coda{-}Forno, Julian and
	Oh, Seong Joon and
	Bethge, Matthias and
	Schulz, Eric},
	title     = {Playing repeated games with large language models},
	journal   = {Nature Human Behaviour},
	volume    = {9},
	number    = {7},
	pages     = {1380--1390},
	year      = {2025},
	month     = may,
	bdoi       = {10.1038/s41562-025-02172-y},
	publisher = {Springer Nature}
}

@inproceedings{alemohammad2024selfconsuming,
	author           = {Sina Alemohammad and
	Josue Casco{-}Rodriguez and
	Lorenzo Luzi and
	Ahmed Imtiaz Humayun and
	Hossein Babaei and
	Daniel LeJeune and
	Ali Siahkoohi and
	Richard G. Baraniuk},
	title            = {Self-Consuming Generative Models Go {MAD}},
	booktitle        = {The Twelfth International Conference on Learning Representations,
	{ICLR} 2024, Vienna, Austria, May 7-11, 2024},
	publisher        = {OpenReview.net},
	year             = {2024},
	bburl             = {https://openreview.net/forum?id=ShjMHfmPs0},
	bibburl           = {https://dblp.org/rec/conf/iclr/AlemohammadCLHB24.bib},
	bibsource        = {dblp computer science bibliography, https://dblp.org},
}

@article{Bai2022ConstitutionalAI,
	title={Constitutional AI: Harmlessness from AI Feedback}, 
	author={Yuntao Bai and Saurav Kadavath and Sandipan Kundu and Amanda Askell and Jackson Kernion and Andy Jones and Anna Chen and Anna Goldie and Azalia Mirhoseini and Cameron McKinnon and Carol Chen and Catherine Olsson and Christopher Olah and Danny Hernandez and Dawn Drain and Deep Ganguli and Dustin Li and Eli Tran-Johnson and Ethan Perez and Jamie Kerr and Jared Mueller and Jeffrey Ladish and Joshua Landau and Kamal Ndousse and Kamile Lukosuite and Liane Lovitt and Michael Sellitto and Nelson Elhage and Nicholas Schiefer and Noemi Mercado and Nova DasSarma and Robert Lasenby and Robin Larson and Sam Ringer and Scott Johnston and Shauna Kravec and Sheer El Showk and Stanislav Fort and Tamera Lanham and Timothy Telleen-Lawton and Tom Conerly and Tom Henighan and Tristan Hume and Samuel R. Bowman and Zac Hatfield-Dodds and Ben Mann and Dario Amodei and Nicholas Joseph and Sam McCandlish and Tom Brown and Jared Kaplan},
	year={2022},
	journal={arXiv preprint},
	volume={arXiv.2212.08073},
	primaryClass={cs.CL},
	burl={https://arxiv.org/abs/2212.08073}, 
}

@article{biderman2024lora,
	author           = {Dan Biderman and
	Jacob P. Portes and
	Jose Javier Gonzalez Ortiz and
	Mansheej Paul and
	Philip Greengard and
	Connor Jennings and
	Daniel King and
	Sam Havens and
	Vitaliy Chiley and
	Jonathan Frankle and
	Cody Blakeney and
	John Patrick Cunningham},
	title            = {LoRA Learns Less and Forgets Less},
	journal          = {Trans. Mach. Learn. Res.},
	volume           = {2024},
	year             = {2024},
	bburl             = {https://openreview.net/forum?id=aloEru2qCG},
	bibburl           = {https://dblp.org/rec/journals/tmlr/BidermanPOPGJKH24.bib},
	bibsource        = {dblp computer science bibliography, https://dblp.org},
}

@inproceedings{Cemri2025MultiAgentFail,
	title={Why Do Multi-Agent {LLM} Systems Fail?},
	author={Mert Cemri and Melissa Z Pan and Shuyi Yang and Lakshya A Agrawal and Bhavya Chopra and Rishabh Tiwari and Kurt Keutzer and Aditya Parameswaran and Dan Klein and Kannan Ramchandran and Matei Zaharia and Joseph E. Gonzalez and Ion Stoica},
	booktitle={The Thirty-ninth Annual Conference on Neural Information Processing Systems Datasets and Benchmarks Track},
	year={2026},
	burl={https://openreview.net/forum?id=fAjbYBmonr}
}

@inproceedings{chen2024zkml,
	author           = {Bing{-}Jyue Chen and
	Suppakit Waiwitlikhit and
	Ion Stoica and
	Daniel Kang},
	title            = {{ZKML:} An Optimizing System for {ML} Inference in Zero-Knowledge
	Proofs},
	booktitle        = {Proceedings of the Nineteenth European Conference on Computer Systems,
	EuroSys 2024, Athens, Greece, April 22-25, 2024},
	pages            = {560--574},
	publisher        = {{ACM}},
	year             = {2024},
	bburl             = {https://bdoi.org/10.1145/3627703.3650088},
	bdoi              = {10.1145/3627703.3650088},
	bibburl           = {https://dblp.org/rec/conf/eurosys/ChenWSK24.bib},
	bibsource        = {dblp computer science bibliography, https://dblp.org},
}

@inproceedings{christiano2017deep,
	author           = {Paul F. Christiano and
	Jan Leike and
	Tom B. Brown and
	Miljan Martic and
	Shane Legg and
	Dario Amodei},
	beditor           = {Isabelle Guyon and
	Ulrike von Luxburg and
	Samy Bengio and
	Hanna M. Wallach and
	Rob Fergus and
	S. V. N. Vishwanathan and
	Roman Garnett},
	title            = {Deep Reinforcement Learning from Human Preferences},
	booktitle        = {Advances in Neural Information Processing Systems 30: Annual Conference
	on Neural Information Processing Systems 2017, December 4-9, 2017,
	Long Beach, CA, {USA}},
	pages            = {4299--4307},
	year             = {2017},
	bburl             = {https://proceedings.neurips.cc/paper/2017/hash/d5e2c0adad503c91f91df240d0cd4e49-Abstract.html},
	bibburl           = {https://dblp.org/rec/conf/nips/ChristianoLBMLA17.bib},
	bibsource        = {dblp computer science bibliography, https://dblp.org},
}

@inproceedings{curry2024differentiable,
	author           = {Michael J. Curry and
	Tuomas Sandholm and
	John P. Dickerson},
	title            = {Differentiable Economics for Randomized Affine Maximizer Auctions},
	booktitle        = {Proceedings of the Thirty-Second International Joint Conference on
	Artificial Intelligence, {IJCAI} 2023, 19th-25th August 2023, Macao,
	SAR, China},
	pages            = {2633--2641},
	publisher        = {ijcai.org},
	year             = {2023},
	bburl             = {https://bdoi.org/10.24963/ijcai.2023/293},
	bdoi              = {10.24963/IJCAI.2023/293},
	bibburl           = {https://dblp.org/rec/conf/ijcai/CurrySD23.bib},
	bibsource        = {dblp computer science bibliography, https://dblp.org},
}

@inproceedings{dao2023fiatShamir,
	author           = {Quang Dao and
	Jim Miller and
	Opal Wright and
	Paul Grubbs},
	title            = {Weak Fiat-Shamir Attacks on Modern Proof Systems},
	booktitle        = {44th {IEEE} Symposium on Security and Privacy, {SP} 2023, San Francisco,
	CA, USA, May 21-25, 2023},
	pages            = {199--216},
	publisher        = {{IEEE}},
	year             = {2023},
	bburl             = {https://bdoi.org/10.1109/SP46215.2023.10179408},
	bdoi              = {10.1109/SP46215.2023.10179408},
	bibburl           = {https://dblp.org/rec/conf/sp/DaoMWG23.bib},
	bibsource        = {dblp computer science bibliography, https://dblp.org},
}

@inproceedings{davani2024disentangling,
	author           = {Aida Mostafazadeh Davani and
	Mark Diaz and
	Dylan K. Baker and
	Vinodkumar Prabhakaran},
	beditor           = {Yaser Al{-}Onaizan and
	Mohit Bansal and
	Yun{-}Nung Chen},
	title            = {{D3CODE:} Disentangling Disagreements in Data across Cultures on Offensiveness
	Detection and Evaluation},
	booktitle        = {Proceedings of the 2024 Conference on Empirical Methods in Natural
	Language Processing, {EMNLP} 2024, Miami, FL, USA, November 12-16,
	2024},
	pages            = {18511--18526},
	publisher        = {Association for Computational Linguistics},
	year             = {2024},
	bburl             = {https://bdoi.org/10.18653/v1/2024.emnlp-main.1029},
	bdoi              = {10.18653/V1/2024.EMNLP-MAIN.1029},
	bibburl           = {https://dblp.org/rec/conf/emnlp/DavaniDBP24.bib},
	bibsource        = {dblp computer science bibliography, https://dblp.org},
}

@inproceedings{dettmers2023qlora,
	author           = {Tim Dettmers and
	Artidoro Pagnoni and
	Ari Holtzman and
	Luke Zettlemoyer},
	beditor           = {Alice Oh and
	Tristan Naumann and
	Amir Globerson and
	Kate Saenko and
	Moritz Hardt and
	Sergey Levine},
	title            = {QLoRA: Efficient Finetuning of Quantized LLMs},
	booktitle        = {Advances in Neural Information Processing Systems 36: Annual Conference
	on Neural Information Processing Systems 2023, NeurIPS 2023, New Orleans,
	LA, USA, December 10 - 16, 2023},
	year             = {2023},
	bburl             = {http://papers.nips.cc/paper\_files/paper/2023/hash/1feb87871436031bdc0f2beaa62a049b-Abstract-Conference.html},
	bibburl           = {https://dblp.org/rec/conf/nips/DettmersPHZ23.bib},
	bibsource        = {dblp computer science bibliography, https://dblp.org},
}

@inproceedings{dohmatob2024tale,
	author           = {Elvis Dohmatob and
	Yunzhen Feng and
	Pu Yang and
	Fran{\c{c}}ois Charton and
	Julia Kempe},
	beditor           = {Ruslan Salakhutdinov and
	Zico Kolter and
	Katherine A. Heller and
	Adrian Weller and
	Nuria Oliver and
	Jonathan Scarlett and
	Felix Berkenkamp},
	title            = {A Tale of Tails: Model Collapse as a Change of Scaling Laws},
	booktitle        = {Forty-first International Conference on Machine Learning, {ICML} 2024,
	Vienna, Austria, July 21-27, 2024},
	series           = {Proceedings of Machine Learning Research},
	pages            = {11165--11197},
	publisher        = {{PMLR} / OpenReview.net},
	year             = {2024},
	bburl             = {https://proceedings.mlr.press/v235/dohmatob24b.html},
	bibburl           = {https://dblp.org/rec/conf/icml/DohmatobFYCK24.bib},
	bibsource        = {dblp computer science bibliography, https://dblp.org},
}

@inproceedings{dohmatob2025strong,
	author           = {Elvis Dohmatob and
	Yunzhen Feng and
	Arjun Subramonian and
	Julia Kempe},
	title            = {Strong Model Collapse},
	booktitle        = {The Thirteenth International Conference on Learning Representations,
	{ICLR} 2025, Singapore, April 24-28, 2025},
	publisher        = {OpenReview.net},
	year             = {2025},
	bburl             = {https://openreview.net/forum?id=et5l9qPUhm},
	bibburl           = {https://dblp.org/rec/conf/iclr/DohmatobFSK25.bib},
	bibsource        = {dblp computer science bibliography, https://dblp.org},
}

@inproceedings{Dziri2023FaithFate,
	author           = {Nouha Dziri and
	Ximing Lu and
	Melanie Sclar and
	Xiang Lorraine Li and
	Liwei Jiang and
	Bill Yuchen Lin and
	Sean Welleck and
	Peter West and
	Chandra Bhagavatula and
	Ronan Le Bras and
	Jena D. Hwang and
	Soumya Sanyal and
	Xiang Ren and
	Allyson Ettinger and
	Za{\"{\i}}d Harchaoui and
	Yejin Choi},
	beditor           = {Alice Oh and
	Tristan Naumann and
	Amir Globerson and
	Kate Saenko and
	Moritz Hardt and
	Sergey Levine},
	title            = {Faith and Fate: Limits of Transformers on Compositionality},
	booktitle        = {Advances in Neural Information Processing Systems 36: Annual Conference
	on Neural Information Processing Systems 2023, NeurIPS 2023, New Orleans,
	LA, USA, December 10 - 16, 2023},
	year             = {2023},
	bburl             = {http://papers.nips.cc/paper\_files/paper/2023/hash/deb3c28192f979302c157cb653c15e90-Abstract-Conference.html},
	bibburl           = {https://dblp.org/rec/conf/nips/DziriLSLJLWWB0H23.bib},
	bibsource        = {dblp computer science bibliography, https://dblp.org},
}

@article{elhage2022superposition,
	title={Toy Models of Superposition}, 
	author={Nelson Elhage and Tristan Hume and Catherine Olsson and Nicholas Schiefer and Tom Henighan and Shauna Kravec and Zac Hatfield-Dodds and Robert Lasenby and Dawn Drain and Carol Chen and Roger Grosse and Sam McCandlish and Jared Kaplan and Dario Amodei and Martin Wattenberg and Christopher Olah},
	year={2022},
	journal={arXiv preprint},
	volume={arXiv.2209.10652},
	primaryClass={cs.LG},
	burl={https://arxiv.org/abs/2209.10652}, 
}

@inproceedings{Es2024RAGAS,
	author           = {Shahul ES and
	Jithin James and
	Luis Espinosa Anke and
	Steven Schockaert},
	beditor           = {Nikolaos Aletras and
	Orph{\'{e}}e De Clercq},
	title            = {RAGAs: Automated Evaluation of Retrieval Augmented Generation},
	booktitle        = {Proceedings of the 18th Conference of the European Chapter of the
	Association for Computational Linguistics, {EACL} 2024 - System Demonstrations,
	St. Julians, Malta, March 17-22, 2024},
	pages            = {150--158},
	publisher        = {Association for Computational Linguistics},
	year             = {2024},
	bburl             = {https://aclanthology.org/2024.eacl-demo.16},
	bibburl           = {https://dblp.org/rec/conf/eacl/ESJAS24.bib},
	bibsource        = {dblp computer science bibliography, https://dblp.org},
}

@inproceedings{Ethayarajh2024KTO,
	author = {Ethayarajh, Kawin and Xu, Winnie and Muennighoff, Niklas and Jurafsky, Dan and Kiela, Douwe},
	title = {Model alignment as prospect theoretic optimization},
	year = {2024},
	publisher = {JMLR.org},
	booktitle = {Proceedings of the 41st International Conference on Machine Learning},
	articleno = {504},
	numpages = {18},
	location = {Vienna, Austria},
	series = {ICML'24}
}

@inproceedings{torrobahennigen2024verification,
	title={Towards Verifiable Text Generation with Symbolic References},
	author={Lucas Torroba Hennigen and Zejiang Shen and Aniruddha Nrusimha and Bernhard Gapp and David Sontag and Yoon Kim},
	booktitle={First Conference on Language Modeling},
	year={2024},
	burl={https://openreview.net/forum?id=fib9qidCpY}
}

@article{fish2025collusion,
	title={Algorithmic Collusion by Large Language Models}, 
	author={Sara Fish and Yannai A. Gonczarowski and Ran I. Shorrer},
	year={2026},
	journal={arXiv preprint},
	volume={arXiv.2404.00806},
	primaryClass={econ.GN},
	burl={https://arxiv.org/abs/2404.00806}, 
}

@article{Gao2024RAGSurvey,
	title={Retrieval-Augmented Generation for Large Language Models: A Survey}, 
	author={Yunfan Gao and Yun Xiong and Xinyu Gao and Kangxiang Jia and Jinliu Pan and Yuxi Bi and Yi Dai and Jiawei Sun and Meng Wang and Haofen Wang},
	year={2024},
	journal={arXiv preprint},
	volume={arXiv.2312.10997},
	primaryClass={cs.CL},
	burl={https://arxiv.org/abs/2312.10997}, 
}

@article{geier2025zkml,
	author    = {Peng, Zhizhi and
	Zhao, Chonghe and
	Wang, Taotao and
	Liao, Guofu and
	Lin, Zibin and
	Liu, Yifeng and
	Cao, Bin and
	Shi, Long and
	Yang, Qing and
	Zhang, Shengli},
	title     = {{A} survey of zero-knowledge proof based verifiable machine learning},
	journal   = {Artificial Intelligence Review},
	year      = {2026},
	month     = apr,
	bdoi       = {10.1007/s10462-026-11557-y},
	publisher = {Springer},
	bnote      = {Online first (13 April 2026); volume/pages forthcoming. Preprint: arXiv:2502.18535. Peng, Zhao, and Wang contributed equally.},
}

@inproceedings{gerstgrasser2024model,
	title={Is Model Collapse Inevitable? Breaking the Curse of Recursion by Accumulating Real and Synthetic Data},
	author={Matthias Gerstgrasser and Rylan Schaeffer and Apratim Dey and Rafael Rafailov and Tomasz Korbak and Henry Sleight and Rajashree Agrawal and John Hughes and Dhruv Bhandarkar Pai and Andrey Gromov and Dan Roberts and Diyi Yang and David L. Donoho and Sanmi Koyejo},
	booktitle={First Conference on Language Modeling},
	year={2024},
	burl={https://openreview.net/forum?id=5B2K4LRgmz}
}

@inproceedings{ghodsi2017safetynets,
	author           = {Zahra Ghodsi and
	Tianyu Gu and
	Siddharth Garg},
	beditor           = {Isabelle Guyon and
	Ulrike von Luxburg and
	Samy Bengio and
	Hanna M. Wallach and
	Rob Fergus and
	S. V. N. Vishwanathan and
	Roman Garnett},
	title            = {SafetyNets: Verifiable Execution of Deep Neural Networks on an Untrusted
	Cloud},
	booktitle        = {Advances in Neural Information Processing Systems 30: Annual Conference
	on Neural Information Processing Systems 2017, December 4-9, 2017,
	Long Beach, CA, {USA}},
	pages            = {4672--4681},
	year             = {2017},
	bburl             = {https://proceedings.neurips.cc/paper/2017/hash/6048ff4e8cb07aa60b6777b6f7384d52-Abstract.html},
	bibburl           = {https://dblp.org/rec/conf/nips/GhodsiGG17.bib},
	bibsource        = {dblp computer science bibliography, https://dblp.org},
}

@inproceedings{gordon2022jury,
	author           = {Mitchell L. Gordon and
	Michelle S. Lam and
	Joon Sung Park and
	Kayur Patel and
	Jeffrey T. Hancock and
	Tatsunori Hashimoto and
	Michael S. Bernstein},
	beditor           = {Simone D. J. Barbosa and
	Cliff Lampe and
	Caroline Appert and
	David A. Shamma and
	Steven Mark Drucker and
	Julie R. Williamson and
	Koji Yatani},
	title            = {Jury Learning: Integrating Dissenting Voices into Machine Learning
	Models},
	booktitle        = {{CHI} '22: {CHI} Conference on Human Factors in Computing Systems,
	New Orleans, LA, USA, 29 April 2022 - 5 May 2022},
	pages            = {115:1--115:19},
	publisher        = {{ACM}},
	year             = {2022},
	bburl             = {https://bdoi.org/10.1145/3491102.3502004},
	bdoi              = {10.1145/3491102.3502004},
	bibburl           = {https://dblp.org/rec/conf/chi/GordonLPPHHB22.bib},
	bibsource        = {dblp computer science bibliography, https://dblp.org},
}

@inproceedings{wolf2024fundamental,
	author           = {Yotam Wolf and
	Noam Wies and
	Oshri Avnery and
	Yoav Levine and
	Amnon Shashua},
	beditor           = {Ruslan Salakhutdinov and
	Zico Kolter and
	Katherine A. Heller and
	Adrian Weller and
	Nuria Oliver and
	Jonathan Scarlett and
	Felix Berkenkamp},
	title            = {Fundamental Limitations of Alignment in Large Language Models},
	booktitle        = {Forty-first International Conference on Machine Learning, {ICML} 2024,
	Vienna, Austria, July 21-27, 2024},
	series           = {Proceedings of Machine Learning Research},
	pages            = {53079--53112},
	publisher        = {{PMLR} / OpenReview.net},
	year             = {2024},
	bburl             = {https://proceedings.mlr.press/v235/wolf24a.html},
	bibburl           = {https://dblp.org/rec/conf/icml/WolfWALS24.bib},
	bibsource        = {dblp computer science bibliography, https://dblp.org},
}

@inproceedings{hu2022lora,
	author           = {Edward J. Hu and
	Yelong Shen and
	Phillip Wallis and
	Zeyuan Allen{-}Zhu and
	Yuanzhi Li and
	Shean Wang and
	Lu Wang and
	Weizhu Chen},
	title            = {LoRA: Low-Rank Adaptation of Large Language Models},
	booktitle        = {The Tenth International Conference on Learning Representations, {ICLR}
	2022, Virtual Event, April 25-29, 2022},
	publisher        = {OpenReview.net},
	year             = {2022},
	bburl             = {https://openreview.net/forum?id=nZeVKeeFYf9},
	bibburl           = {https://dblp.org/rec/conf/iclr/HuSWALWWC22.bib},
	bibsource        = {dblp computer science bibliography, https://dblp.org},
}

@article{Izacard2022Contriever,
	author           = {Gautier Izacard and
	Mathilde Caron and
	Lucas Hosseini and
	Sebastian Riedel and
	Piotr Bojanowski and
	Armand Joulin and
	Edouard Grave},
	title            = {Unsupervised Dense Information Retrieval with Contrastive Learning},
	journal          = {Trans. Mach. Learn. Res.},
	volume           = {2022},
	year             = {2022},
	bburl             = {https://openreview.net/forum?id=jKN1pXi7b0},
	bibburl           = {https://dblp.org/rec/journals/tmlr/IzacardCHRBJG22.bib},
	bibsource        = {dblp computer science bibliography, https://dblp.org},
}

@article{jang2023personalized,
	title={Personalized Soups: Personalized Large Language Model Alignment via Post-hoc Parameter Merging}, 
	author={Joel Jang and Seungone Kim and Bill Yuchen Lin and Yizhong Wang and Jack Hessel and Luke Zettlemoyer and Hannaneh Hajishirzi and Yejin Choi and Prithviraj Ammanabrolu},
	year={2023},
	journal={arXiv preprint},
	volume={arXiv.2310.11564},
	primaryClass={cs.CL},
	burl={https://arxiv.org/abs/2310.11564}, 
}

@inproceedings{jiang2023active,
	author           = {Zhengbao Jiang and
	Frank F. Xu and
	Luyu Gao and
	Zhiqing Sun and
	Qian Liu and
	Jane Dwivedi{-}Yu and
	Yiming Yang and
	Jamie Callan and
	Graham Neubig},
	beditor           = {Houda Bouamor and
	Juan Pino and
	Kalika Bali},
	title            = {Active Retrieval Augmented Generation},
	booktitle        = {Proceedings of the 2023 Conference on Empirical Methods in Natural
	Language Processing, {EMNLP} 2023, Singapore, December 6-10, 2023},
	pages            = {7969--7992},
	publisher        = {Association for Computational Linguistics},
	year             = {2023},
	bburl             = {https://bdoi.org/10.18653/v1/2023.emnlp-main.495},
	bdoi              = {10.18653/V1/2023.EMNLP-MAIN.495},
	bibburl           = {https://dblp.org/rec/conf/emnlp/JiangXGSLDYCN23.bib},
	bibsource        = {dblp computer science bibliography, https://dblp.org},
}

@article{jukna2012boolean,
	author           = {Stasys Jukna},
	title            = {Boolean Function Complexity Advances and Frontiers},
	journal          = {Bull. {EATCS}},
	volume           = {113},
	year             = {2014},
	bburl             = {http://eatcs.org/beatcs/index.php/beatcs/article/view/275},
	bibburl           = {https://dblp.org/rec/journals/eatcs/Jukna14.bib},
	bibsource        = {dblp computer science bibliography, https://dblp.org},
}

@inproceedings{kambhampati2024position,
	author           = {Subbarao Kambhampati and
	Karthik Valmeekam and
	Lin Guan and
	Mudit Verma and
	Kaya Stechly and
	Siddhant Bhambri and
	Lucas Saldyt and
	Anil Murthy},
	beditor           = {Ruslan Salakhutdinov and
	Zico Kolter and
	Katherine A. Heller and
	Adrian Weller and
	Nuria Oliver and
	Jonathan Scarlett and
	Felix Berkenkamp},
	title            = {Position: LLMs Can't Plan, But Can Help Planning in LLM-Modulo Frameworks},
	booktitle        = {Forty-first International Conference on Machine Learning, {ICML} 2024,
	Vienna, Austria, July 21-27, 2024},
	series           = {Proceedings of Machine Learning Research},
	pages            = {22895--22907},
	publisher        = {{PMLR} / OpenReview.net},
	year             = {2024},
	bburl             = {https://proceedings.mlr.press/v235/kambhampati24a.html},
	bibburl           = {https://dblp.org/rec/conf/icml/KambhampatiVGVS24.bib},
	bibsource        = {dblp computer science bibliography, https://dblp.org},
}

@inproceedings{kothapalli2022nova,
	author           = {Abhiram Kothapalli and
	Srinath T. V. Setty and
	Ioanna Tzialla},
	beditor           = {Yevgeniy Dodis and
	Thomas Shrimpton},
	title            = {Nova: Recursive Zero-Knowledge Arguments from Folding Schemes},
	booktitle        = {Advances in Cryptology - {CRYPTO} 2022 - 42nd Annual International
	Cryptology Conference, {CRYPTO} 2022, Santa Barbara, CA, USA, August
	15-18, 2022, Proceedings, Part {IV}},
	series           = {Lecture bnotes in Computer Science},
	pages            = {359--388},
	publisher        = {Springer},
	year             = {2022},
	bburl             = {https://bdoi.org/10.1007/978-3-031-15985-5\_13},
	bdoi              = {10.1007/978-3-031-15985-5\_13},
	bibburl           = {https://dblp.org/rec/conf/crypto/KothapalliST22.bib},
	bibsource        = {dblp computer science bibliography, https://dblp.org},
}

@inproceedings{kuhn2023semantic,
	author           = {Lorenz Kuhn and
	Yarin Gal and
	Sebastian Farquhar},
	title            = {Semantic Uncertainty: Linguistic Invariances for Uncertainty Estimation
	in Natural Language Generation},
	booktitle        = {The Eleventh International Conference on Learning Representations,
	{ICLR} 2023, Kigali, Rwanda, May 1-5, 2023},
	publisher        = {OpenReview.net},
	year             = {2023},
	bburl             = {https://openreview.net/forum?id=VD-AYtP0dve},
	bibburl           = {https://dblp.org/rec/conf/iclr/KuhnGF23.bib},
	bibsource        = {dblp computer science bibliography, https://dblp.org},
}

@article{Kwiatkowski2019NQ,
	author           = {Tom Kwiatkowski and
	Jennimaria Palomaki and
	Olivia Redfield and
	Michael Collins and
	Ankur P. Parikh and
	Chris Alberti and
	Danielle Epstein and
	Illia Polosukhin and
	Jacob Devlin and
	Kenton Lee and
	Kristina Toutanova and
	Llion Jones and
	Matthew Kelcey and
	Ming{-}Wei Chang and
	Andrew M. Dai and
	Jakob Uszkoreit and
	Quoc Le and
	Slav Petrov},
	title            = {Natural Questions: a Benchmark for Question Answering Research},
	journal          = {Trans. Assoc. Comput. Linguistics},
	volume           = {7},
	pages            = {452--466},
	year             = {2019},
	bburl             = {https://bdoi.org/10.1162/tacl\_a\_00276},
	bdoi              = {10.1162/TACL\_A\_00276},
	bibburl           = {https://dblp.org/rec/journals/tacl/KwiatkowskiPRCP19.bib},
	bibsource        = {dblp computer science bibliography, https://dblp.org},
}

@inproceedings{Lewis2020RAG,
	author           = {Patrick Lewis and
	Ethan Perez and
	Aleksandra Piktus and
	Fabio Petroni and
	Vladimir Karpukhin and
	Naman Goyal and
	Heinrich K{\"{u}}ttler and
	Mike Lewis and
	Wen{-}tau Yih and
	Tim Rockt{\"{a}}schel and
	Sebastian Riedel and
	Douwe Kiela},
	beditor           = {Hugo Larochelle and
	Marc'Aurelio Ranzato and
	Raia Hadsell and
	Maria{-}Florina Balcan and
	Hsuan{-}Tien Lin},
	title            = {Retrieval-Augmented Generation for Knowledge-Intensive {NLP} Tasks},
	booktitle        = {Advances in Neural Information Processing Systems 33: Annual Conference
	on Neural Information Processing Systems 2020, NeurIPS 2020, December
	6-12, 2020, virtual},
	year             = {2020},
	bburl             = {https://proceedings.neurips.cc/paper/2020/hash/6b493230205f780e1bc26945df7481e5-Abstract.html},
	bibburl           = {https://dblp.org/rec/conf/nips/LewisPPPKGKLYR020.bib},
	bibsource        = {dblp computer science bibliography, https://dblp.org},
}

@inproceedings{liu2021zkcnn,
	author           = {Tianyi Liu and
	Xiang Xie and
	Yupeng Zhang},
	beditor           = {Yongdae Kim and
	Jong Kim and
	Giovanni Vigna and
	Elaine Shi},
	title            = {zkCNN: Zero Knowledge Proofs for Convolutional Neural Network Predictions
	and Accuracy},
	booktitle        = {{CCS} '21: 2021 {ACM} {SIGSAC} Conference on Computer and Communications
	Security, Virtual Event, Republic of Korea, November 15 - 19, 2021},
	pages            = {2968--2985},
	publisher        = {{ACM}},
	year             = {2021},
	bburl             = {https://bdoi.org/10.1145/3460120.3485379},
	bdoi              = {10.1145/3460120.3485379},
	bibburl           = {https://dblp.org/rec/conf/ccs/LiuXZ21.bib},
	bibsource        = {dblp computer science bibliography, https://dblp.org},
}

@article{liu2023shortcuts,
	author           = {Nelson F. Liu and
	Kevin Lin and
	John Hewitt and
	Ashwin Paranjape and
	Michele Bevilacqua and
	Fabio Petroni and
	Percy Liang},
	title            = {Lost in the Middle: How Language Models Use Long Contexts},
	journal          = {Trans. Assoc. Comput. Linguistics},
	volume           = {12},
	pages            = {157--173},
	year             = {2024},
	bburl             = {https://bdoi.org/10.1162/tacl\_a\_00638},
	bdoi              = {10.1162/TACL\_A\_00638},
	bibburl           = {https://dblp.org/rec/journals/tacl/LiuLHPBPL24.bib},
	bibsource        = {dblp computer science bibliography, https://dblp.org},
}

@inproceedings{Liu2024ECBD,
	author           = {Yu Lu Liu and
	Su Lin Blodgett and
	Jackie C. K. Cheung and
	Vera Liao and
	Alexandra Olteanu and
	Ziang Xiao},
	beditor           = {Lun{-}Wei Ku and
	Andre Martins and
	Vivek Srikumar},
	title            = {{ECBD:} Evidence-Centered Benchmark Design for {NLP}},
	booktitle        = {Proceedings of the 62nd Annual Meeting of the Association for Computational
	Linguistics (Volume 1: Long Papers), {ACL} 2024, Bangkok, Thailand,
	August 11-16, 2024},
	pages            = {16349--16365},
	publisher        = {Association for Computational Linguistics},
	year             = {2024},
	bburl             = {https://bdoi.org/10.18653/v1/2024.acl-long.861},
	bdoi              = {10.18653/V1/2024.ACL-LONG.861},
	bibburl           = {https://dblp.org/rec/conf/acl/LiuBCLOX24.bib},
	bibsource        = {dblp computer science bibliography, https://dblp.org},
}

@inproceedings{lotfi2024nonvacuous,
	author           = {Sanae Lotfi and
	Marc Anton Finzi and
	Yilun Kuang and
	Tim G. J. Rudner and
	Micah Goldblum and
	Andrew Gordon Wilson},
	beditor           = {Ruslan Salakhutdinov and
	Zico Kolter and
	Katherine A. Heller and
	Adrian Weller and
	Nuria Oliver and
	Jonathan Scarlett and
	Felix Berkenkamp},
	title            = {Non-Vacuous Generalization Bounds for Large Language Models},
	booktitle        = {Forty-first International Conference on Machine Learning, {ICML} 2024,
	Vienna, Austria, July 21-27, 2024},
	series           = {Proceedings of Machine Learning Research},
	pages            = {32801--32818},
	publisher        = {{PMLR} / OpenReview.net},
	year             = {2024},
	bburl             = {https://proceedings.mlr.press/v235/lotfi24a.html},
	bibburl           = {https://dblp.org/rec/conf/icml/LotfiFKRGW24.bib},
	bibsource        = {dblp computer science bibliography, https://dblp.org},
}

@inproceedings{hu2023unlocking,
	author           = {Kai Hu and
	Andy Zou and
	Zifan Wang and
	Klas Leino and
	Matt Fredrikson},
	beditor           = {Alice Oh and
	Tristan Naumann and
	Amir Globerson and
	Kate Saenko and
	Moritz Hardt and
	Sergey Levine},
	title            = {Unlocking Deterministic Robustness Certification on ImageNet},
	booktitle        = {Advances in Neural Information Processing Systems 36: Annual Conference
	on Neural Information Processing Systems 2023, NeurIPS 2023, New Orleans,
	LA, USA, December 10 - 16, 2023},
	year             = {2023},
	bburl             = {http://papers.nips.cc/paper\_files/paper/2023/hash/863da9d40547f1d1b18859519ce2dee4-Abstract-Conference.html},
	bibburl           = {https://dblp.org/rec/conf/nips/HuZWLF23.bib},
	bibsource        = {dblp computer science bibliography, https://dblp.org},
}

@inproceedings{lu2025toolsandbox,
	author           = {Jiarui Lu and
	Thomas Holleis and
	Yizhe Zhang and
	Bernhard Aumayer and
	Feng Nan and
	Haoping Bai and
	Shuang Ma and
	Shen Ma and
	Mengyu Li and
	Guoli Yin and
	Zirui Wang and
	Ruoming Pang},
	beditor           = {Luis Chiruzzo and
	Alan Ritter and
	Lu Wang},
	title            = {ToolSandbox: {A} Stateful, Conversational, Interactive Evaluation
	Benchmark for {LLM} Tool Use Capabilities},
	booktitle        = {Findings of the Association for Computational Linguistics: {NAACL}
	2025, Albuquerque, New Mexico, USA, April 29 - May 4, 2025},
	series           = {Findings of {ACL}},
	pages            = {1160--1183},
	publisher        = {Association for Computational Linguistics},
	year             = {2025},
	bburl             = {https://bdoi.org/10.18653/v1/2025.findings-naacl.65},
	bdoi              = {10.18653/V1/2025.FINDINGS-NAACL.65},
	bibburl           = {https://dblp.org/rec/conf/naacl/LuHZANBMMLYWP25.bib},
	bibsource        = {dblp computer science bibliography, https://dblp.org},
}

@inproceedings{malladi2023kernel,
	author           = {Sadhika Malladi and
	Alexander Wettig and
	Dingli Yu and
	Danqi Chen and
	Sanjeev Arora},
	beditor           = {Andreas Krause and
	Emma Brunskill and
	Kyunghyun Cho and
	Barbara Engelhardt and
	Sivan Sabato and
	Jonathan Scarlett},
	title            = {A Kernel-Based View of Language Model Fine-Tuning},
	booktitle        = {International Conference on Machine Learning, {ICML} 2023, 23-29 July
	2023, Honolulu, Hawaii, {USA}},
	series           = {Proceedings of Machine Learning Research},
	pages            = {23610--23641},
	publisher        = {{PMLR}},
	year             = {2023},
	bburl             = {https://proceedings.mlr.press/v202/malladi23a.html},
	bibburl           = {https://dblp.org/rec/conf/icml/MalladiWYCA23.bib},
	bibsource        = {dblp computer science bibliography, https://dblp.org},
}

@inproceedings{mcallester2003pac,
	author           = {David A. McAllester},
	beditor           = {Bernhard Sch{\"{o}}lkopf and
	Manfred K. Warmuth},
	title            = {Simplified PAC-Bayesian Margin Bounds},
	booktitle        = {Computational Learning Theory and Kernel Machines, 16th Annual Conference
	on Computational Learning Theory and 7th Kernel Workshop, COLT/Kernel
	2003, Washington, DC, USA, August 24-27, 2003, Proceedings},
	series           = {Lecture bnotes in Computer Science},
	pages            = {203--215},
	publisher        = {Springer},
	year             = {2003},
	bburl             = {https://bdoi.org/10.1007/978-3-540-45167-9\_16},
	bdoi              = {10.1007/978-3-540-45167-9\_16},
	bibburl           = {https://dblp.org/rec/conf/colt/McAllester03.bib},
	bibsource        = {dblp computer science bibliography, https://dblp.org},
}

@inproceedings{meng2022locating,
	author           = {Kevin Meng and
	David Bau and
	Alex Andonian and
	Yonatan Belinkov},
	beditor           = {Sanmi Koyejo and
	S. Mohamed and
	A. Agarwal and
	Danielle Belgrave and
	K. Cho and
	A. Oh},
	title            = {Locating and Editing Factual Associations in {GPT}},
	booktitle        = {Advances in Neural Information Processing Systems 35: Annual Conference
	on Neural Information Processing Systems 2022, NeurIPS 2022, New Orleans,
	LA, USA, November 28 - December 9, 2022},
	year             = {2022},
	bburl             = {http://papers.nips.cc/paper\_files/paper/2022/hash/6f1d43d5a82a37e89b0665b33bf3a182-Abstract-Conference.html},
	bibburl           = {https://dblp.org/rec/conf/nips/MengBAB22.bib},
	bibsource        = {dblp computer science bibliography, https://dblp.org},
}

@inproceedings{meng2023mass,
	author           = {Kevin Meng and
	Arnab Sen Sharma and
	Alex J. Andonian and
	Yonatan Belinkov and
	David Bau},
	title            = {Mass-Editing Memory in a Transformer},
	booktitle        = {The Eleventh International Conference on Learning Representations,
	{ICLR} 2023, Kigali, Rwanda, May 1-5, 2023},
	publisher        = {OpenReview.net},
	year             = {2023},
	bburl             = {https://openreview.net/forum?id=MkbcAHIYgyS},
	bibburl           = {https://dblp.org/rec/conf/iclr/MengSABB23.bib},
	bibsource        = {dblp computer science bibliography, https://dblp.org},
}

@article{Merrill2022Saturated,
	author           = {William Merrill and
	Ashish Sabharwal and
	Noah A. Smith},
	title            = {Saturated Transformers are Constant-Depth Threshold Circuits},
	journal          = {Trans. Assoc. Comput. Linguistics},
	volume           = {10},
	pages            = {843--856},
	year             = {2022},
	bburl             = {https://bdoi.org/10.1162/tacl\_a\_00493},
	bdoi              = {10.1162/TACL\_A\_00493},
	bibburl           = {https://dblp.org/rec/journals/tacl/MerrillSS22.bib},
	bibsource        = {dblp computer science bibliography, https://dblp.org},
}

@article{merrill2023parallelism,
	author           = {William Merrill and
	Ashish Sabharwal},
	title            = {The Parallelism Tradeoff: Limitations of Log-Precision Transformers},
	journal          = {Trans. Assoc. Comput. Linguistics},
	volume           = {11},
	pages            = {531--545},
	year             = {2023},
	bburl             = {https://bdoi.org/10.1162/tacl\_a\_00562},
	bdoi              = {10.1162/TACL\_A\_00562},
	bibburl           = {https://dblp.org/rec/journals/tacl/MerrillS23.bib},
	bibsource        = {dblp computer science bibliography, https://dblp.org},
}

@inproceedings{mitchell2022grace,
	author           = {Eric Mitchell and
	Charles Lin and
	Antoine Bosselut and
	Chelsea Finn and
	Christopher D. Manning},
	title            = {Fast Model Editing at Scale},
	booktitle        = {The Tenth International Conference on Learning Representations, {ICLR}
	2022, Virtual Event, April 25-29, 2022},
	publisher        = {OpenReview.net},
	year             = {2022},
	bburl             = {https://openreview.net/forum?id=0DcZxeWfOPt},
	bibburl           = {https://dblp.org/rec/conf/iclr/MitchellLBFM22.bib},
	bibsource        = {dblp computer science bibliography, https://dblp.org},
}

@inproceedings{ouyang2022training,
	author           = {Long Ouyang and
	Jeffrey Wu and
	Xu Jiang and
	Diogo Almeida and
	Carroll L. Wainwright and
	Pamela Mishkin and
	Chong Zhang and
	Sandhini Agarwal and
	Katarina Slama and
	Alex Ray and
	John Schulman and
	Jacob Hilton and
	Fraser Kelton and
	Luke Miller and
	Maddie Simens and
	Amanda Askell and
	Peter Welinder and
	Paul F. Christiano and
	Jan Leike and
	Ryan Lowe},
	beditor           = {Sanmi Koyejo and
	S. Mohamed and
	A. Agarwal and
	Danielle Belgrave and
	K. Cho and
	A. Oh},
	title            = {Training language models to follow instructions with human feedback},
	booktitle        = {Advances in Neural Information Processing Systems 35: Annual Conference
	on Neural Information Processing Systems 2022, NeurIPS 2022, New Orleans,
	LA, USA, November 28 - December 9, 2022},
	year             = {2022},
	bburl             = {http://papers.nips.cc/paper\_files/paper/2022/hash/b1efde53be364a73914f58805a001731-Abstract-Conference.html},
	bibburl           = {https://dblp.org/rec/conf/nips/Ouyang0JAWMZASR22.bib},
	bibsource        = {dblp computer science bibliography, https://dblp.org},
}

@inproceedings{park2025regret,
	author           = {Chanwoo Park and
	Xiangyu Liu and
	Asuman E. Ozdaglar and
	Kaiqing Zhang},
	title            = {Do {LLM} Agents Have Regret? {A} Case Study in Online Learning and
	Games},
	booktitle        = {The Thirteenth International Conference on Learning Representations,
	{ICLR} 2025, Singapore, April 24-28, 2025},
	publisher        = {OpenReview.net},
	year             = {2025},
	bburl             = {https://openreview.net/forum?id=qn9tBYQHGi},
	bibburl           = {https://dblp.org/rec/conf/iclr/ParkLOZ25.bib},
	bibsource        = {dblp computer science bibliography, https://dblp.org},
}

@inproceedings{qin2024toolbench,
	author           = {Yujia Qin and
	Shihao Liang and
	Yining Ye and
	Kunlun Zhu and
	Lan Yan and
	Yaxi Lu and
	Yankai Lin and
	Xin Cong and
	Xiangru Tang and
	Bill Qian and
	Sihan Zhao and
	Lauren Hong and
	Runchu Tian and
	Ruobing Xie and
	Jie Zhou and
	Mark Gerstein and
	Dahai Li and
	Zhiyuan Liu and
	Maosong Sun},
	title            = {ToolLLM: Facilitating Large Language Models to Master 16000+ Real-world
	APIs},
	booktitle        = {The Twelfth International Conference on Learning Representations,
	{ICLR} 2024, Vienna, Austria, May 7-11, 2024},
	publisher        = {OpenReview.net},
	year             = {2024},
	bburl             = {https://openreview.net/forum?id=dHng2O0Jjr},
	bibburl           = {https://dblp.org/rec/conf/iclr/QinLYZYLLCTQZHT24.bib},
	bibsource        = {dblp computer science bibliography, https://dblp.org},
}

@inproceedings{rafailov2023direct,
	author           = {Rafael Rafailov and
	Archit Sharma and
	Eric Mitchell and
	Christopher D. Manning and
	Stefano Ermon and
	Chelsea Finn},
	beditor           = {Alice Oh and
	Tristan Naumann and
	Amir Globerson and
	Kate Saenko and
	Moritz Hardt and
	Sergey Levine},
	title            = {Direct Preference Optimization: Your Language Model is Secretly a
	Reward Model},
	booktitle        = {Advances in Neural Information Processing Systems 36: Annual Conference
	on Neural Information Processing Systems 2023, NeurIPS 2023, New Orleans,
	LA, USA, December 10 - 16, 2023},
	year             = {2023},
	bburl             = {http://papers.nips.cc/paper\_files/paper/2023/hash/a85b405ed65c6477a4fe8302b5e06ce7-Abstract-Conference.html},
	bibburl           = {https://dblp.org/rec/conf/nips/RafailovSMMEF23.bib},
	bibsource        = {dblp computer science bibliography, https://dblp.org},
}

@article{Rashkin2023AIS,
	author           = {Hannah Rashkin and
	Vitaly Nikolaev and
	Matthew Lamm and
	Lora Aroyo and
	Michael Collins and
	Dipanjan Das and
	Slav Petrov and
	Gaurav Singh Tomar and
	Iulia Turc and
	David Reitter},
	title            = {Measuring Attribution in Natural Language Generation Models},
	journal          = {Comput. Linguistics},
	volume           = {49},
	number           = {4},
	pages            = {777--840},
	year             = {2023},
	bburl             = {https://bdoi.org/10.1162/coli\_a\_00486},
	bdoi              = {10.1162/COLI\_A\_00486},
	bibburl           = {https://dblp.org/rec/journals/coling/RashkinNLA00PTT23.bib},
	bibsource        = {dblp computer science bibliography, https://dblp.org},
}

@inproceedings{lu2024knowdonttell,
	author           = {Muhan Gao and
	Taiming Lu and
	Kuai Yu and
	Adam Byerly and
	Daniel Khashabi},
	beditor           = {Yaser Al{-}Onaizan and
	Mohit Bansal and
	Yun{-}Nung Chen},
	title            = {Insights into {LLM} Long-Context Failures: When Transformers Know
	but Don't Tell},
	booktitle        = {Findings of the Association for Computational Linguistics: {EMNLP}
	2024, Miami, Florida, USA, November 12-16, 2024},
	series           = {Findings of {ACL}},
	pages            = {7611--7625},
	publisher        = {Association for Computational Linguistics},
	year             = {2024},
	bburl             = {https://bdoi.org/10.18653/v1/2024.findings-emnlp.447},
	bdoi              = {10.18653/V1/2024.FINDINGS-EMNLP.447},
	bibburl           = {https://dblp.org/rec/conf/emnlp/GaoLYBK24.bib},
	bibsource        = {dblp computer science bibliography, https://dblp.org},
}

@article{shumailov2024collapse,
	author           = {Ilia Shumailov and
	Zakhar Shumaylov and
	Yiren Zhao and
	Nicolas Papernot and
	Ross J. Anderson and
	Yarin Gal},
	title            = {{AI} models collapse when trained on recursively generated data},
	journal          = {Nat.},
	volume           = {631},
	number           = {8022},
	pages            = {755--759},
	year             = {2024},
	bburl             = {https://bdoi.org/10.1038/s41586-024-07566-y},
	bdoi              = {10.1038/S41586-024-07566-Y},
	bibburl           = {https://dblp.org/rec/journals/nature/ShumailovSZPAG24.bib},
	bibsource        = {dblp computer science bibliography, https://dblp.org},
}

@article{snell2024scaling,
	title={Scaling LLM Test-Time Compute Optimally can be More Effective than Scaling Model Parameters}, 
	author={Charlie Snell and Jaehoon Lee and Kelvin Xu and Aviral Kumar},
	year={2024},
	journal={arXiv preprint},
	volume={arXiv.2408.03314},
	primaryClass={cs.LG},
	burl={https://arxiv.org/abs/2408.03314}, 
}

@article{Song2024PreferenceCollapse,
	author = {Jiancong Xiao and Ziniu Li and Xingyu Xie and Emily Getzen and Cong Fang and Qi Long and Weijie J. Su},
	title = {On the Algorithmic Bias of Aligning Large Language Models with RLHF: Preference Collapse and Matching Regularization},
	journal = {Journal of the American Statistical Association},
	volume = {120},
	number = {552},
	pages = {2154--2164},
	year = {2025},
	publisher = {Taylor \& Francis},
	bdoi = {10.1080/01621459.2025.2555067},
	burl = {  https://bdoi.org/10.1080/01621459.2025.2555067},
	beprint = { https://bdoi.org/10.1080/01621459.2025.2555067}
}

@article{Stiennon2020BestOfN,
	author = {Stiennon, Nisan and Ouyang, Long and Wu, Jeff and Ziegler, Daniel M. and Lowe, Ryan and Voss, Chelsea and Radford, Alec and Amodei, Dario and Christiano, Paul},
	title = {Learning to summarize from human feedback},
	year = {2020},
	isbn = {9781713829546},
	publisher = {Curran Associates Inc.},
	address = {Red Hook, NY, USA},
	booktitle = {Proceedings of the 34th International Conference on Neural Information Processing Systems},
	articleno = {253},
	numpages = {14},
	location = {Vancouver, BC, Canada},
	series = {NIPS '20}
}

@inproceedings{tang2024generalized,
	author           = {Yunhao Tang and
	Zhaohan Daniel Guo and
	Zeyu Zheng and
	Daniele Calandriello and
	R{\'{e}}mi Munos and
	Mark Rowland and
	Pierre Harvey Richemond and
	Michal Valko and
	Bernardo {\'{A}}vila Pires and
	Bilal Piot},
	beditor           = {Ruslan Salakhutdinov and
	Zico Kolter and
	Katherine A. Heller and
	Adrian Weller and
	Nuria Oliver and
	Jonathan Scarlett and
	Felix Berkenkamp},
	title            = {Generalized Preference Optimization: {A} Unified Approach to Offline
	Alignment},
	booktitle        = {Forty-first International Conference on Machine Learning, {ICML} 2024,
	Vienna, Austria, July 21-27, 2024},
	series           = {Proceedings of Machine Learning Research},
	pages            = {47725--47742},
	publisher        = {{PMLR} / OpenReview.net},
	year             = {2024},
	bburl             = {https://proceedings.mlr.press/v235/tang24b.html},
	bibburl           = {https://dblp.org/rec/conf/icml/TangGZCMRRVPP24.bib},
	bibsource        = {dblp computer science bibliography, https://dblp.org},
}

@article{thaler2022proofs,
	author           = {Justin Thaler},
	title            = {Proofs, Arguments, and Zero-Knowledge},
	journal          = {Found. Trends Priv. Secur.},
	volume           = {4},
	number           = {2-4},
	pages            = {117--660},
	year             = {2022},
	bburl             = {https://bdoi.org/10.1561/3300000030},
	bdoi              = {10.1561/3300000030},
	bibburl           = {https://dblp.org/rec/journals/ftsec/Thaler22.bib},
	bibsource        = {dblp computer science bibliography, https://dblp.org},
}

@article{Trivedi2022MuSiQue,
	author           = {Harsh Trivedi and
	Niranjan Balasubramanian and
	Tushar Khot and
	Ashish Sabharwal},
	title            = {MuSiQue: Multihop Questions via Single-hop Question
	Composition},
	journal          = {Trans. Assoc. Comput. Linguistics},
	volume           = {10},
	pages            = {539--554},
	year             = {2022},
	bburl             = {https://bdoi.org/10.1162/tacl\_a\_00475},
	bdoi              = {10.1162/TACL\_A\_00475},
	bibburl           = {https://dblp.org/rec/journals/tacl/TrivediBKS22.bib},
	bibsource        = {dblp computer science bibliography, https://dblp.org},
}

@inproceedings{Wallat2025Correctness,
	author = {Wallat, Jonas and Heuss, Maria and Rijke, Maarten de and Anand, Avishek},
	title = {Correctness is not Faithfulness in Retrieval Augmented Generation Attributions},
	year = {2025},
	isbn = {9798400718618},
	publisher = {Association for Computing Machinery},
	address = {New York, NY, USA},
	burl = {https://bdoi.org/10.1145/3731120.3744592},
	bdoi = {10.1145/3731120.3744592},
	booktitle = {Proceedings of the 2025 International ACM SIGIR Conference on Innovative Concepts and Theories in Information Retrieval (ICTIR)},
	pages = {22–32},
	numpages = {11},
	location = {Padua, Italy},
	series = {ICTIR '25}
}

@inproceedings{wang2022gpl,
	author           = {Kexin Wang and
	Nandan Thakur and
	Nils Reimers and
	Iryna Gurevych},
	beditor           = {Marine Carpuat and
	Marie{-}Catherine de Marneffe and
	Iv{\'{a}}n Vladimir Meza Ru{\'{\i}}z},
	title            = {{GPL:} Generative Pseudo Labeling for Unsupervised Domain Adaptation
	of Dense Retrieval},
	booktitle        = {Proceedings of the 2022 Conference of the North American Chapter of
	the Association for Computational Linguistics: Human Language Technologies,
	{NAACL} 2022, Seattle, WA, United States, July 10-15, 2022},
	pages            = {2345--2360},
	publisher        = {Association for Computational Linguistics},
	year             = {2022},
	bburl             = {https://bdoi.org/10.18653/v1/2022.naacl-main.168},
	bdoi              = {10.18653/V1/2022.NAACL-MAIN.168},
	bibburl           = {https://dblp.org/rec/conf/naacl/WangT0G22.bib},
	bibsource        = {dblp computer science bibliography, https://dblp.org},
}

@inproceedings{wei2022chain,
	author           = {Jason Wei and
	Xuezhi Wang and
	Dale Schuurmans and
	Maarten Bosma and
	Brian Ichter and
	Fei Xia and
	Ed H. Chi and
	Quoc V. Le and
	Denny Zhou},
	beditor           = {Sanmi Koyejo and
	S. Mohamed and
	A. Agarwal and
	Danielle Belgrave and
	K. Cho and
	A. Oh},
	title            = {Chain-of-Thought Prompting Elicits Reasoning in Large Language Models},
	booktitle        = {Advances in Neural Information Processing Systems 35: Annual Conference
	on Neural Information Processing Systems 2022, NeurIPS 2022, New Orleans,
	LA, USA, November 28 - December 9, 2022},
	year             = {2022},
	bburl             = {http://papers.nips.cc/paper\_files/paper/2022/hash/9d5609613524ecf4f15af0f7b31abca4-Abstract-Conference.html},
	bibburl           = {https://dblp.org/rec/conf/nips/Wei0SBIXCLZ22.bib},
	bibsource        = {dblp computer science bibliography, https://dblp.org},
}

@article{Wei2024SimpleQA,
	title={Measuring short-form factuality in large language models}, 
	author={Jason Wei and Nguyen Karina and Hyung Won Chung and Yunxin Joy Jiao and Spencer Papay and Amelia Glaese and John Schulman and William Fedus},
	year={2024},
	journal={arXiv preprint},
	volume={arXiv.2411.04368},
	primaryClass={cs.CL},
	burl={https://arxiv.org/abs/2411.04368}, 
}

@inproceedings{wortsman2022model,
	author           = {Mitchell Wortsman and
	Gabriel Ilharco and
	Samir Yitzhak Gadre and
	Rebecca Roelofs and
	Raphael Gontijo Lopes and
	Ari S. Morcos and
	Hongseok Namkoong and
	Ali Farhadi and
	Yair Carmon and
	Simon Kornblith and
	Ludwig Schmidt},
	beditor           = {Kamalika Chaudhuri and
	Stefanie Jegelka and
	Le Song and
	Csaba Szepesv{\'{a}}ri and
	Gang Niu and
	Sivan Sabato},
	title            = {Model soups: averaging weights of multiple fine-tuned models improves
	accuracy without increasing inference time},
	booktitle        = {International Conference on Machine Learning, {ICML} 2022, 17-23 July
	2022, Baltimore, Maryland, {USA}},
	series           = {Proceedings of Machine Learning Research},
	pages            = {23965--23998},
	publisher        = {{PMLR}},
	year             = {2022},
	bburl             = {https://proceedings.mlr.press/v162/wortsman22a.html},
	bibburl           = {https://dblp.org/rec/conf/icml/WortsmanIGRLMNF22.bib},
	bibsource        = {dblp computer science bibliography, https://dblp.org},
}

@article{Xi2025AgentSurvey,
	author           = {Zhiheng Xi and
	Wenxiang Chen and
	Xin Guo and
	Wei He and
	Yiwen Ding and
	Boyang Hong and
	Ming Zhang and
	Junzhe Wang and
	Senjie Jin and
	Enyu Zhou and
	Rui Zheng and
	Xiaoran Fan and
	Xiao Wang and
	Limao Xiong and
	Yuhao Zhou and
	Weiran Wang and
	Changhao Jiang and
	Yicheng Zou and
	Xiangyang Liu and
	Zhangyue Yin and
	Shihan Dou and
	Rongxiang Weng and
	Wenjuan Qin and
	Yongyan Zheng and
	Xipeng Qiu and
	Xuanjing Huang and
	Qi Zhang and
	Tao Gui},
	title            = {The rise and potential of large language model based agents: a survey},
	journal          = {Sci. China Inf. Sci.},
	volume           = {68},
	number           = {2},
	year             = {2025},
	bburl             = {https://bdoi.org/10.1007/s11432-024-4222-0},
	bdoi              = {10.1007/S11432-024-4222-0},
	bibburl           = {https://dblp.org/rec/journals/chinaf/XiCGHDHZWJZZFWXZWJZLYDW25.bib},
	bibsource        = {dblp computer science bibliography, https://dblp.org},
}

@inproceedings{xu2024dpo,
	author           = {Shusheng Xu and
	Wei Fu and
	Jiaxuan Gao and
	Wenjie Ye and
	Weilin Liu and
	Zhiyu Mei and
	Guangju Wang and
	Chao Yu and
	Yi Wu},
	beditor           = {Ruslan Salakhutdinov and
	Zico Kolter and
	Katherine A. Heller and
	Adrian Weller and
	Nuria Oliver and
	Jonathan Scarlett and
	Felix Berkenkamp},
	title            = {Is {DPO} Superior to {PPO} for {LLM} Alignment? {A} Comprehensive
	Study},
	booktitle        = {Forty-first International Conference on Machine Learning, {ICML} 2024,
	Vienna, Austria, July 21-27, 2024},
	series           = {Proceedings of Machine Learning Research},
	pages            = {54983--54998},
	publisher        = {{PMLR} / OpenReview.net},
	year             = {2024},
	bburl             = {https://proceedings.mlr.press/v235/xu24h.html},
	bibburl           = {https://dblp.org/rec/conf/icml/XuFGYLMW0024.bib},
	bibsource        = {dblp computer science bibliography, https://dblp.org},
}

@inproceedings{yadav2023ties,
	author           = {Prateek Yadav and
	Derek Tam and
	Leshem Choshen and
	Colin A. Raffel and
	Mohit Bansal},
	beditor           = {Alice Oh and
	Tristan Naumann and
	Amir Globerson and
	Kate Saenko and
	Moritz Hardt and
	Sergey Levine},
	title            = {TIES-Merging: Resolving Interference When Merging Models},
	booktitle        = {Advances in Neural Information Processing Systems 36: Annual Conference
	on Neural Information Processing Systems 2023, NeurIPS 2023, New Orleans,
	LA, USA, December 10 - 16, 2023},
	year             = {2023},
	bburl             = {http://papers.nips.cc/paper\_files/paper/2023/hash/1644c9af28ab7916874f6fd6228a9bcf-Abstract-Conference.html},
	bibburl           = {https://dblp.org/rec/conf/nips/YadavTCRB23.bib},
	bibsource        = {dblp computer science bibliography, https://dblp.org},
}

@inproceedings{you2023mass,
	author           = {Xiaoyu You and
	Beina Sheng and
	Daizong Ding and
	Mi Zhang and
	Xudong Pan and
	Min Yang and
	Fuli Feng},
	beditor           = {Ying Ding and
	Jie Tang and
	Juan F. Sequeda and
	Lora Aroyo and
	Carlos Castillo and
	Geert{-}Jan Houben},
	title            = {MaSS: Model-agnostic, Semantic and Stealthy Data Poisoning Attack
	on Knowledge Graph Embedding},
	booktitle        = {Proceedings of the {ACM} Web Conference 2023, {WWW} 2023, Austin,
	TX, USA, 30 April 2023 - 4 May 2023},
	pages            = {2000--2010},
	publisher        = {{ACM}},
	year             = {2023},
	bburl             = {https://bdoi.org/10.1145/3543507.3583203},
	bdoi              = {10.1145/3543507.3583203},
	bibburl           = {https://dblp.org/rec/conf/www/YouSDZPYF23.bib},
	bibsource        = {dblp computer science bibliography, https://dblp.org},
}

@inproceedings{zhou2019nonvacuous,
	author           = {Wenda Zhou and
	Victor Veitch and
	Morgane Austern and
	Ryan P. Adams and
	Peter Orbanz},
	title            = {Non-vacuous Generalization Bounds at the ImageNet Scale: a PAC-Bayesian
	Compression Approach},
	booktitle        = {7th International Conference on Learning Representations, {ICLR} 2019,
	New Orleans, LA, USA, May 6-9, 2019},
	publisher        = {OpenReview.net},
	year             = {2019},
	bburl             = {https://openreview.net/forum?id=BJgqqsAct7},
	bibburl           = {https://dblp.org/rec/conf/iclr/ZhouVAAO19.bib},
	bibsource        = {dblp computer science bibliography, https://dblp.org},
}

@inproceedings{zhou2024webarena,
	author           = {Shuyan Zhou and
	Frank F. Xu and
	Hao Zhu and
	Xuhui Zhou and
	Robert Lo and
	Abishek Sridhar and
	Xianyi Cheng and
	Tianyue Ou and
	Yonatan Bisk and
	Daniel Fried and
	Uri Alon and
	Graham Neubig},
	title            = {WebArena: {A} Realistic Web Environment for Building Autonomous Agents},
	booktitle        = {The Twelfth International Conference on Learning Representations,
	{ICLR} 2024, Vienna, Austria, May 7-11, 2024},
	publisher        = {OpenReview.net},
	year             = {2024},
	bburl             = {https://openreview.net/forum?id=oKn9c6ytLx},
	bibburl           = {https://dblp.org/rec/conf/iclr/ZhouX0ZLSCOBF0N24.bib},
	bibsource        = {dblp computer science bibliography, https://dblp.org},
}

@inproceedings{stap2024finetuning,
	author           = {David Stap and
	Eva Hasler and
	Bill Byrne and
	Christof Monz and
	Ke Tran},
	beditor           = {Lun{-}Wei Ku and
	Andre Martins and
	Vivek Srikumar},
	title            = {The Fine-Tuning Paradox: Boosting Translation Quality Without Sacrificing
	{LLM} Abilities},
	booktitle        = {Proceedings of the 62nd Annual Meeting of the Association for Computational
	Linguistics (Volume 1: Long Papers), {ACL} 2024, Bangkok, Thailand,
	August 11-16, 2024},
	pages            = {6189--6206},
	publisher        = {Association for Computational Linguistics},
	year             = {2024},
	bburl             = {https://bdoi.org/10.18653/v1/2024.acl-long.336},
	bdoi              = {10.18653/V1/2024.ACL-LONG.336},
	bibburl           = {https://dblp.org/rec/conf/acl/StapHBMT24.bib},
	bibsource        = {dblp computer science bibliography, https://dblp.org},
}

@inproceedings{weiss2021rasp,
	author           = {Gail Weiss and
	Yoav Goldberg and
	Eran Yahav},
	beditor           = {Marina Meila and
	Tong Zhang},
	title            = {Thinking Like Transformers},
	booktitle        = {Proceedings of the 38th International Conference on Machine Learning,
	{ICML} 2021, 18-24 July 2021, Virtual Event},
	series           = {Proceedings of Machine Learning Research},
	pages            = {11080--11090},
	publisher        = {{PMLR}},
	year             = {2021},
	bburl             = {http://proceedings.mlr.press/v139/weiss21a.html},
	bibburl           = {https://dblp.org/rec/conf/icml/WeissGY21.bib},
	bibsource        = {dblp computer science bibliography, https://dblp.org},
}

@article{strobl2024formal,
	author           = {Lena Strobl and
	William Merrill and
	Gail Weiss and
	David Chiang and
	Dana Angluin},
	title            = {What Formal Languages Can Transformers Express? {A} Survey},
	journal          = {Trans. Assoc. Comput. Linguistics},
	volume           = {12},
	pages            = {543--561},
	year             = {2024},
	bburl             = {https://bdoi.org/10.1162/tacl\_a\_00663},
	bdoi              = {10.1162/TACL\_A\_00663},
	bibburl           = {https://dblp.org/rec/journals/tacl/StroblMW0A24.bib},
	bibsource        = {dblp computer science bibliography, https://dblp.org},
}

@inproceedings{sanford2024representational,
	author           = {Clayton Sanford and
	Daniel J. Hsu and
	Matus Telgarsky},
	beditor           = {Alice Oh and
	Tristan Naumann and
	Amir Globerson and
	Kate Saenko and
	Moritz Hardt and
	Sergey Levine},
	title            = {Representational Strengths and Limitations of Transformers},
	booktitle        = {Advances in Neural Information Processing Systems 36: Annual Conference
	on Neural Information Processing Systems 2023, NeurIPS 2023, New Orleans,
	LA, USA, December 10 - 16, 2023},
	year             = {2023},
	bburl             = {http://papers.nips.cc/paper\_files/paper/2023/hash/73bf692447f174984f30499ec9b20e04-Abstract-Conference.html},
	bibburl           = {https://dblp.org/rec/conf/nips/SanfordHT23.bib},
	bibsource        = {dblp computer science bibliography, https://dblp.org},
}

@inproceedings{sanford2024transformers,
	author           = {Clayton Sanford and
	Daniel Hsu and
	Matus Telgarsky},
	beditor           = {Ruslan Salakhutdinov and
	Zico Kolter and
	Katherine A. Heller and
	Adrian Weller and
	Nuria Oliver and
	Jonathan Scarlett and
	Felix Berkenkamp},
	title            = {Transformers, parallel computation, and logarithmic depth},
	booktitle        = {Forty-first International Conference on Machine Learning, {ICML} 2024,
	Vienna, Austria, July 21-27, 2024},
	series           = {Proceedings of Machine Learning Research},
	pages            = {43276--43327},
	publisher        = {{PMLR} / OpenReview.net},
	year             = {2024},
	bburl             = {https://proceedings.mlr.press/v235/sanford24a.html},
	bibburl           = {https://dblp.org/rec/conf/icml/Sanford0T24.bib},
	bibsource        = {dblp computer science bibliography, https://dblp.org},
}

@inproceedings{kojima2022zeroshot,
	author           = {Takeshi Kojima and
	Shixiang Shane Gu and
	Machel Reid and
	Yutaka Matsuo and
	Yusuke Iwasawa},
	beditor           = {Sanmi Koyejo and
	S. Mohamed and
	A. Agarwal and
	Danielle Belgrave and
	K. Cho and
	A. Oh},
	title            = {Large Language Models are Zero-Shot Reasoners},
	booktitle        = {Advances in Neural Information Processing Systems 35: Annual Conference
	on Neural Information Processing Systems 2022, NeurIPS 2022, New Orleans,
	LA, USA, November 28 - December 9, 2022},
	year             = {2022},
	bburl             = {http://papers.nips.cc/paper\_files/paper/2022/hash/8bb0d291acd4acf06ef112099c16f326-Abstract-Conference.html},
	bibburl           = {https://dblp.org/rec/conf/nips/KojimaGRMI22.bib},
	bibsource        = {dblp computer science bibliography, https://dblp.org},
}

@article{Nye2021Scratchpad,
	title={Show Your Work: Scratchpads for Intermediate Computation with Language Models}, 
	author={Maxwell Nye and Anders Johan Andreassen and Guy Gur-Ari and Henryk Michalewski and Jacob Austin and David Bieber and David Dohan and Aitor Lewkowycz and Maarten Bosma and David Luan and Charles Sutton and Augustus Odena},
	year={2021},
	journal={arXiv preprint},
	volume={arXiv.2112.00114},
	primaryClass={cs.LG},
	burl={https://arxiv.org/abs/2112.00114}, 
}

@inproceedings{Feng2023ToTheoretical,
	author           = {Guhao Feng and
	Bohang Zhang and
	Yuntian Gu and
	Haotian Ye and
	Di He and
	Liwei Wang},
	beditor           = {Alice Oh and
	Tristan Naumann and
	Amir Globerson and
	Kate Saenko and
	Moritz Hardt and
	Sergey Levine},
	title            = {Towards Revealing the Mystery behind Chain of Thought: {A} Theoretical
	Perspective},
	booktitle        = {Advances in Neural Information Processing Systems 36: Annual Conference
	on Neural Information Processing Systems 2023, NeurIPS 2023, New Orleans,
	LA, USA, December 10 - 16, 2023},
	year             = {2023},
	bburl             = {http://papers.nips.cc/paper\_files/paper/2023/hash/dfc310e81992d2e4cedc09ac47eff13e-Abstract-Conference.html},
	bibburl           = {https://dblp.org/rec/conf/nips/FengZGY0W23.bib},
	bibsource        = {dblp computer science bibliography, https://dblp.org},
}

@inproceedings{Li2024CoTExpressivity,
	author           = {Zhiyuan Liu and
	Hong Liu and
	Denny Zhou and
	Tengyu Ma},
	title            = {Chain of Thought Empowers Transformers to Solve Inherently Serial
	Problems},
	booktitle        = {The Twelfth International Conference on Learning Representations,
	{ICLR} 2024, Vienna, Austria, May 7-11, 2024},
	publisher        = {OpenReview.net},
	year             = {2024},
	bburl             = {https://openreview.net/forum?id=3EWTEy9MTM},
	bibburl           = {https://dblp.org/rec/conf/iclr/0001LZ024.bib},
	bibsource        = {dblp computer science bibliography, https://dblp.org},
}

@inproceedings{Merrill2024CoTTransformers,
	author           = {William Merrill and
	Ashish Sabharwal},
	title            = {The Expressive Power of Transformers with Chain of Thought},
	booktitle        = {The Twelfth International Conference on Learning Representations,
	{ICLR} 2024, Vienna, Austria, May 7-11, 2024},
	publisher        = {OpenReview.net},
	year             = {2024},
	bburl             = {https://openreview.net/forum?id=NjNGlPh8Wh},
	bibburl           = {https://dblp.org/rec/conf/iclr/MerrillS24.bib},
	bibsource        = {dblp computer science bibliography, https://dblp.org},
}

@inproceedings{Schaeffer2023Mirage,
	author           = {Rylan Schaeffer and
	Brando Miranda and
	Sanmi Koyejo},
	beditor           = {Alice Oh and
	Tristan Naumann and
	Amir Globerson and
	Kate Saenko and
	Moritz Hardt and
	Sergey Levine},
	title            = {Are Emergent Abilities of Large Language Models a Mirage?},
	booktitle        = {Advances in Neural Information Processing Systems 36: Annual Conference
	on Neural Information Processing Systems 2023, NeurIPS 2023, New Orleans,
	LA, USA, December 10 - 16, 2023},
	year             = {2023},
	bburl             = {http://papers.nips.cc/paper\_files/paper/2023/hash/adc98a266f45005c403b8311ca7e8bd7-Abstract-Conference.html},
	bibburl           = {https://dblp.org/rec/conf/nips/SchaefferMK23.bib},
	bibsource        = {dblp computer science bibliography, https://dblp.org},
}

@article{Lanham2023Faithfulness,
	title={Measuring Faithfulness in Chain-of-Thought Reasoning}, 
	author={Tamera Lanham and Anna Chen and Ansh Radhakrishnan and Benoit Steiner and Carson Denison and Danny Hernandez and Dustin Li and Esin Durmus and Evan Hubinger and Jackson Kernion and Kamilė Lukošiūtė and Karina Nguyen and Newton Cheng and Nicholas Joseph and Nicholas Schiefer and Oliver Rausch and Robin Larson and Sam McCandlish and Sandipan Kundu and Saurav Kadavath and Shannon Yang and Thomas Henighan and Timothy Maxwell and Timothy Telleen-Lawton and Tristan Hume and Zac Hatfield-Dodds and Jared Kaplan and Jan Brauner and Samuel R. Bowman and Ethan Perez},
	year={2023},
	journal={arXiv preprint},
	volume={arXiv.2307.13702},
	primaryClass={cs.AI},
	burl={https://arxiv.org/abs/2307.13702}, 
}

@article{Cobbe2021Verifiers,
	title={Training Verifiers to Solve Math Word Problems}, 
	author={Karl Cobbe and Vineet Kosaraju and Mohammad Bavarian and Mark Chen and Heewoo Jun and Lukasz Kaiser and Matthias Plappert and Jerry Tworek and Jacob Hilton and Reiichiro Nakano and Christopher Hesse and John Schulman},
	year={2021},
	journal={arXiv preprint},
	volume={arXiv.2110.14168},
	primaryClass={cs.LG},
	burl={https://arxiv.org/abs/2110.14168}, 
}

@article{Uesato2022ProcessOutcome,
	title={Solving math word problems with process- and outcome-based feedback}, 
	author={Jonathan Uesato and Nate Kushman and Ramana Kumar and Francis Song and Noah Siegel and Lisa Wang and Antonia Creswell and Geoffrey Irving and Irina Higgins},
	year={2022},
	journal={arXiv preprint},
	volume={arXiv.2211.14275},
	primaryClass={cs.LG},
	burl={https://arxiv.org/abs/2211.14275}, 
}

@inproceedings{Hao2023ReasoningLM,
	author           = {Shibo Hao and
	Yi Gu and
	Haodi Ma and
	Joshua Jiahua Hong and
	Zhen Wang and
	Daisy Zhe Wang and
	Zhiting Hu},
	beditor           = {Houda Bouamor and
	Juan Pino and
	Kalika Bali},
	title            = {Reasoning with Language Model is Planning with World Model},
	booktitle        = {Proceedings of the 2023 Conference on Empirical Methods in Natural
	Language Processing, {EMNLP} 2023, Singapore, December 6-10, 2023},
	pages            = {8154--8173},
	publisher        = {Association for Computational Linguistics},
	year             = {2023},
	bburl             = {https://bdoi.org/10.18653/v1/2023.emnlp-main.507},
	bdoi              = {10.18653/V1/2023.EMNLP-MAIN.507},
	bibburl           = {https://dblp.org/rec/conf/emnlp/HaoGMHWWH23.bib},
	bibsource        = {dblp computer science bibliography, https://dblp.org},
}

@inproceedings{Schick2023Toolformer,
	author           = {Timo Schick and
	Jane Dwivedi{-}Yu and
	Roberto Dess{\`{\i}} and
	Roberta Raileanu and
	Maria Lomeli and
	Eric Hambro and
	Luke Zettlemoyer and
	Nicola Cancedda and
	Thomas Scialom},
	beditor           = {Alice Oh and
	Tristan Naumann and
	Amir Globerson and
	Kate Saenko and
	Moritz Hardt and
	Sergey Levine},
	title            = {Toolformer: Language Models Can Teach Themselves to Use Tools},
	booktitle        = {Advances in Neural Information Processing Systems 36: Annual Conference
	on Neural Information Processing Systems 2023, NeurIPS 2023, New Orleans,
	LA, USA, December 10 - 16, 2023},
	year             = {2023},
	bburl             = {http://papers.nips.cc/paper\_files/paper/2023/hash/d842425e4bf79ba039352da0f658a906-Abstract-Conference.html},
	bibburl           = {https://dblp.org/rec/conf/nips/SchickDDRLHZCS23.bib},
	bibsource        = {dblp computer science bibliography, https://dblp.org},
}

@inproceedings{Shinn2023Reflexion,
	author           = {Noah Shinn and
	Federico Cassano and
	Ashwin Gopinath and
	Karthik Narasimhan and
	Shunyu Yao},
	beditor           = {Alice Oh and
	Tristan Naumann and
	Amir Globerson and
	Kate Saenko and
	Moritz Hardt and
	Sergey Levine},
	title            = {Reflexion: language agents with verbal reinforcement learning},
	booktitle        = {Advances in Neural Information Processing Systems 36: Annual Conference
	on Neural Information Processing Systems 2023, NeurIPS 2023, New Orleans,
	LA, USA, December 10 - 16, 2023},
	year             = {2023},
	bburl             = {http://papers.nips.cc/paper\_files/paper/2023/hash/1b44b878bb782e6954cd888628510e90-Abstract-Conference.html},
	bibburl           = {https://dblp.org/rec/conf/nips/ShinnCGNY23.bib},
	bibsource        = {dblp computer science bibliography, https://dblp.org},
}

@inproceedings{karpukhin2020dense,
	author           = {Vladimir Karpukhin and
	Barlas Oguz and
	Sewon Min and
	Patrick Lewis and
	Ledell Wu and
	Sergey Edunov and
	Danqi Chen and
	Wen{-}tau Yih},
	beditor           = {Bonnie Webber and
	Trevor Cohn and
	Yulan He and
	Yang Liu},
	title            = {Dense Passage Retrieval for Open-Domain Question Answering},
	booktitle        = {Proceedings of the 2020 Conference on Empirical Methods in Natural
	Language Processing, {EMNLP} 2020, Online, November 16-20, 2020},
	pages            = {6769--6781},
	publisher        = {Association for Computational Linguistics},
	year             = {2020},
	bburl             = {https://bdoi.org/10.18653/v1/2020.emnlp-main.550},
	bdoi              = {10.18653/V1/2020.EMNLP-MAIN.550},
	bibburl           = {https://dblp.org/rec/conf/emnlp/KarpukhinOMLWEC20.bib},
	bibsource        = {dblp computer science bibliography, https://dblp.org},
}

@inproceedings{Skalse2022RewardHacking,
	author = {Skalse, Joar and Howe, Nikolaus H. R. and Krasheninnikov, Dmitrii and Krueger, David},
	title = {Defining and characterizing reward hacking},
	year = {2022},
	isbn = {9781713871088},
	publisher = {Curran Associates Inc.},
	address = {Red Hook, NY, USA},
	booktitle = {Proceedings of the 36th International Conference on Neural Information Processing Systems},
	articleno = {687},
	numpages = {12},
	location = {New Orleans, LA, USA},
	series = {NIPS '22}
}

@article{anwar2024foundational,
	author           = {Usman Anwar and
	Abulhair Saparov and
	Javier Rando and
	Daniel Paleka and
	Miles Turpin and
	Peter Hase and
	Ekdeep Singh Lubana and
	Erik Jenner and
	Stephen Casper and
	Oliver Sourbut and
	Benjamin L. Edelman and
	Zhaowei Zhang and
	Mario G{\"{u}}nther and
	Anton Korinek and
	Jos{\'{e}} Hern{\'{a}}ndez{-}Orallo and
	Lewis Hammond and
	Eric J. Bigelow and
	Alexander Pan and
	Lauro Langosco and
	Tomasz Korbak and
	Heidi Chenyu Zhang and
	Ruiqi Zhong and
	Se{\'{a}}n {\'{O}} h{\'{E}}igeartaigh and
	Gabriel Recchia and
	Giulio Corsi and
	Alan Chan and
	Markus Anderljung and
	Lilian Edwards and
	Aleksandar Petrov and
	Christian Schr{\"{o}}der de Witt and
	Sumeet Ramesh Motwani and
	Yoshua Bengio and
	Danqi Chen and
	Philip Torr and
	Samuel Albanie and
	Tegan Maharaj and
	Jakob Nicolaus Foerster and
	Florian Tram{\`{e}}r and
	He He and
	Atoosa Kasirzadeh and
	Yejin Choi and
	David Krueger},
	title            = {Foundational Challenges in Assuring Alignment and Safety of Large
	Language Models},
	journal          = {Trans. Mach. Learn. Res.},
	volume           = {2024},
	year             = {2024},
	bburl             = {https://openreview.net/forum?id=oVTkOs8Pka},
	bibburl           = {https://dblp.org/rec/journals/tmlr/AnwarSRPTHLJCSE24.bib},
	bibsource        = {dblp computer science bibliography, https://dblp.org},
}

@article{Ji2023AISafety,
	title={AI Alignment: A Comprehensive Survey}, 
	author={Jiaming Ji and Tianyi Qiu and Boyuan Chen and Borong Zhang and Hantao Lou and Kaile Wang and Yawen Duan and Zhonghao He and Lukas Vierling and Donghai Hong and Jiayi Zhou and Zhaowei Zhang and Fanzhi Zeng and Juntao Dai and Xuehai Pan and Kwan Yee Ng and Aidan O'Gara and Hua Xu and Brian Tse and Jie Fu and Stephen McAleer and Yaodong Yang and Yizhou Wang and Song-Chun Zhu and Yike Guo and Wen Gao},
	year={2025},
	journal={arXiv preprint},
	volume={arXiv.2310.19852},
	primaryClass={cs.AI},
	burl={https://arxiv.org/abs/2310.19852}, 
}

@inproceedings{Hendrycks2020MMLU,
	author           = {Dan Hendrycks and
	Collin Burns and
	Steven Basart and
	Andy Zou and
	Mantas Mazeika and
	Dawn Song and
	Jacob Steinhardt},
	title            = {Measuring Massive Multitask Language Understanding},
	booktitle        = {9th International Conference on Learning Representations, {ICLR} 2021,
	Virtual Event, Austria, May 3-7, 2021},
	publisher        = {OpenReview.net},
	year             = {2021},
	bburl             = {https://openreview.net/forum?id=d7KBjmI3GmQ},
	bibburl           = {https://dblp.org/rec/conf/iclr/HendrycksBBZMSS21.bib},
	bibsource        = {dblp computer science bibliography, https://dblp.org},
}

@article{Valiant1984,
	author           = {Leslie G. Valiant},
	title            = {A Theory of the Learnable},
	journal          = {Commun. {ACM}},
	volume           = {27},
	number           = {11},
	pages            = {1134--1142},
	year             = {1984},
	bburl             = {https://bdoi.org/10.1145/1968.1972},
	bdoi              = {10.1145/1968.1972},
	bibburl           = {https://dblp.org/rec/journals/cacm/Valiant84.bib},
	bibsource        = {dblp computer science bibliography, https://dblp.org},
}

@book{KearnsVazirani1994,
	author           = {Michael J. Kearns and
	Umesh V. Vazirani},
	title            = {An Introduction to Computational Learning Theory},
	publisher        = {{MIT} Press},
	year             = {1994},
	bburl             = {https://mitpress.mit.edu/books/introduction-computational-learning-theory},
	isbn             = {978-0-262-11193-5},
	bibburl           = {https://dblp.org/rec/books/daglib/0041035.bib},
	bibsource        = {dblp computer science bibliography, https://dblp.org},
}

@inproceedings{wu2025simplicity,
	title={When More is Less: Understanding Chain-of-Thought Length in {LLM}s},
	author={Yuyang Wu and Yifei Wang and Ziyu Ye and Tianqi Du and Stefanie Jegelka and Yisen Wang},
	booktitle={The Fourteenth International Conference on Learning Representations},
	year={2026},
	burl={https://openreview.net/forum?id=6QDFsYxtI1}
}

@inproceedings{MerrillSabharwal2025LittleDepth,
	title={A Little Depth Goes a Long Way: The Expressive Power of Log-Depth Transformers},
	author={William Merrill and Ashish Sabharwal},
	booktitle={NeurIPS 2024 Workshop on Mathematics of Modern Machine Learning},
	year={2024},
	burl={https://openreview.net/forum?id=njycONK0JG}
}

@inproceedings{Wang2025LyapLock,
	author           = {Peng Wang and
	Biyu Zhou and
	Xuehai Tang and
	Jizhong Han and
	Songlin Hu},
	beditor           = {Christos Christodoulopoulos and
	Tanmoy Chakraborty and
	Carolyn Rose and
	Violet Peng},
	title            = {LyapLock: Bounded Knowledge Preservation in Sequential Large Language
	Model Editing},
	booktitle        = {Proceedings of the 2025 Conference on Empirical Methods in Natural
	Language Processing, {EMNLP} 2025, Suzhou, China, November 4-9, 2025},
	pages            = {6434--6459},
	publisher        = {Association for Computational Linguistics},
	year             = {2025},
	bburl             = {https://bdoi.org/10.18653/v1/2025.emnlp-main.327},
	bdoi              = {10.18653/V1/2025.EMNLP-MAIN.327},
	bibburl           = {https://dblp.org/rec/conf/emnlp/WangZTHH25.bib},
	bibsource        = {dblp computer science bibliography, https://dblp.org},
}

@article{FischerLynchPaterson1985,
	author           = {Michael J. Fischer and
	Nancy A. Lynch and
	Mike Paterson},
	title            = {Impossibility of Distributed Consensus with One Faulty Process},
	journal          = {J. {ACM}},
	volume           = {32},
	number           = {2},
	pages            = {374--382},
	year             = {1985},
	bburl             = {https://bdoi.org/10.1145/3149.214121},
	bdoi              = {10.1145/3149.214121},
	bibburl           = {https://dblp.org/rec/journals/jacm/FischerLP85.bib},
	bibsource        = {dblp computer science bibliography, https://dblp.org},
}

@article{GilbertLynch2002CAP,
	author           = {Seth Gilbert and
	Nancy A. Lynch},
	title            = {Brewer's conjecture and the feasibility of consistent, available,
	partition-tolerant web services},
	journal          = {{SIGACT} News},
	volume           = {33},
	number           = {2},
	pages            = {51--59},
	year             = {2002},
	bburl             = {https://bdoi.org/10.1145/564585.564601},
	bdoi              = {10.1145/564585.564601},
	bibburl           = {https://dblp.org/rec/journals/sigact/GilbertL02.bib},
	bibsource        = {dblp computer science bibliography, https://dblp.org},
}

@article{Abadi2012PACELC,
	author           = {Daniel Abadi},
	title            = {Consistency Tradeoffs in Modern Distributed Database System Design:
	{CAP} is Only Part of the Story},
	journal          = {Computer},
	volume           = {45},
	number           = {2},
	pages            = {37--42},
	year             = {2012},
	bburl             = {https://bdoi.org/10.1109/MC.2012.33},
	bdoi              = {10.1109/MC.2012.33},
	bibburl           = {https://dblp.org/rec/journals/computer/Abadi12.bib},
	bibsource        = {dblp computer science bibliography, https://dblp.org},
}

@article{WolpertMacready1997NFL,
	author           = {David H. Wolpert and
	William G. Macready},
	title            = {No free lunch theorems for optimization},
	journal          = {{IEEE} Trans. Evol. Comput.},
	volume           = {1},
	number           = {1},
	pages            = {67--82},
	year             = {1997},
	bburl             = {https://bdoi.org/10.1109/4235.585893},
	bdoi              = {10.1109/4235.585893},
	bibburl           = {https://dblp.org/rec/journals/tec/DolpertM97.bib},
	bibsource        = {dblp computer science bibliography, https://dblp.org},
}

@inproceedings{Azar2024IPO,
	author           = {Mohammad Gheshlaghi Azar and
	Zhaohan Daniel Guo and
	Bilal Piot and
	R{\'{e}}mi Munos and
	Mark Rowland and
	Michal Valko and
	Daniele Calandriello},
	beditor           = {Sanjoy Dasgupta and
	Stephan Mandt and
	Yingzhen Li},
	title            = {A General Theoretical Paradigm to Understand Learning from Human Preferences},
	booktitle        = {International Conference on Artificial Intelligence and Statistics,
	2-4 May 2024, Palau de Congressos, Valencia, Spain},
	series           = {Proceedings of Machine Learning Research},
	pages            = {4447--4455},
	publisher        = {{PMLR}},
	year             = {2024},
	bburl             = {https://proceedings.mlr.press/v238/gheshlaghi-azar24a.html},
	bibburl           = {https://dblp.org/rec/conf/aistats/AzarGPMRVC24.bib},
	bibsource        = {dblp computer science bibliography, https://dblp.org},
}

@inproceedings{RayChowdhury2024RobustDPO,
	author           = {Sayak Ray Chowdhury and
	Anush Kini and
	Nagarajan Natarajan},
	beditor           = {Ruslan Salakhutdinov and
	Zico Kolter and
	Katherine A. Heller and
	Adrian Weller and
	Nuria Oliver and
	Jonathan Scarlett and
	Felix Berkenkamp},
	title            = {Provably Robust {DPO:} Aligning Language Models with Noisy Feedback},
	booktitle        = {Forty-first International Conference on Machine Learning, {ICML} 2024,
	Vienna, Austria, July 21-27, 2024},
	series           = {Proceedings of Machine Learning Research},
	pages            = {42258--42274},
	publisher        = {{PMLR} / OpenReview.net},
	year             = {2024},
	bburl             = {https://proceedings.mlr.press/v235/ray-chowdhury24a.html},
	bibburl           = {https://dblp.org/rec/conf/icml/ChowdhuryKN24.bib},
	bibsource        = {dblp computer science bibliography, https://dblp.org},
}

@inproceedings{JiaRakhlinXie2025ProcessSupervision,
	author           = {Zeyu Jia and
	Alexander Rakhlin and
	Tengyang Xie},
	beditor           = {Aarti Singh and
	Maryam Fazel and
	Daniel Hsu and
	Simon Lacoste{-}Julien and
	Felix Berkenkamp and
	Tegan Maharaj and
	Kiri Wagstaff and
	Jerry Zhu},
	title            = {Do We Need to Verify Step by Step? Rethinking Process Supervision
	from a Theoretical Perspective},
	booktitle        = {Forty-second International Conference on Machine Learning, {ICML}
	2025, Vancouver, BC, Canada, July 13-19, 2025},
	series           = {Proceedings of Machine Learning Research},
	publisher        = {{PMLR} / OpenReview.net},
	year             = {2025},
	bburl             = {https://proceedings.mlr.press/v267/jia25f.html},
	bibburl           = {https://dblp.org/rec/conf/icml/JiaRX25.bib},
	bibsource        = {dblp computer science bibliography, https://dblp.org},
}

@inproceedings{JacobsWallach2021Measurement,
	author           = {Abigail Z. Jacobs},
	beditor           = {Madeleine Clare Elish and
	William Isaac and
	Richard S. Zemel},
	title            = {Measurement and Fairness},
	booktitle        = {FAccT '21: 2021 {ACM} Conference on Fairness, Accountability, and
	Transparency, Virtual Event / Toronto, Canada, March 3-10, 2021},
	pages            = {375--385},
	publisher        = {{ACM}},
	year             = {2021},
	bburl             = {https://bdoi.org/10.1145/3442188.3445901},
	bdoi              = {10.1145/3442188.3445901},
	bibburl           = {https://dblp.org/rec/conf/fat/Jacobs21.bib},
	bibsource        = {dblp computer science bibliography, https://dblp.org},
}

@article{DimitriouGarretaManzurVlasov2024Mova,
	author           = {Nikolaos Dimitriou and
	Albert Garreta and
	Ignacio Manzur and
	Ilia Vlasov},
	title            = {Mova: Nova folding without committing to error terms},
	journal          = {{IACR} Cryptol. ePrint Arch.},
	volume           = {2024},
	pages            = {1220},
	year             = {2024},
	bburl             = {https://eprint.iacr.org/2024/1220},
	bibburl           = {https://dblp.org/rec/journals/iacr/DimitriouGMV24.bib},
	bibsource        = {dblp computer science bibliography, https://dblp.org},
}

@article{KothapalliSetty2024NeutronNova,
	author           = {Abhiram Kothapalli and
	Srinath T. V. Setty},
	title            = {NeutronNova: Folding everything that reduces to zero-check},
	journal          = {{IACR} Cryptol. ePrint Arch.},
	volume           = {2024},
	pages            = {1606},
	year             = {2024},
	bburl             = {https://eprint.iacr.org/2024/1606},
	bibburl           = {https://dblp.org/rec/journals/iacr/KothapalliS24.bib},
	bibsource        = {dblp computer science bibliography, https://dblp.org},
}

@inproceedings{Li2024ESC,
	author           = {Yiwei Li and
	Peiwen Yuan and
	Shaoxiong Feng and
	Boyuan Pan and
	Xinglin Wang and
	Bin Sun and
	Heda Wang and
	Kan Li},
	title            = {Escape Sky-high Cost: Early-stopping Self-Consistency for Multi-step
	Reasoning},
	booktitle        = {The Twelfth International Conference on Learning Representations,
	{ICLR} 2024, Vienna, Austria, May 7-11, 2024},
	publisher        = {OpenReview.net},
	year             = {2024},
	bburl             = {https://openreview.net/forum?id=ndR8Ytrzhh},
	bibburl           = {https://dblp.org/rec/conf/iclr/LiYFPW0W024.bib},
	bibsource        = {dblp computer science bibliography, https://dblp.org},
}

@inproceedings{Muennighoff2025S1,
	author           = {Niklas Muennighoff and
	Zitong Yang and
	Weijia Shi and
	Xiang Lisa Li and
	Li Fei{-}Fei and
	Hannaneh Hajishirzi and
	Luke Zettlemoyer and
	Percy Liang and
	Emmanuel J. Cand{\`{e}}s and
	Tatsunori Hashimoto},
	beditor           = {Christos Christodoulopoulos and
	Tanmoy Chakraborty and
	Carolyn Rose and
	Violet Peng},
	title            = {s1: Simple test-time scaling},
	booktitle        = {Proceedings of the 2025 Conference on Empirical Methods in Natural
	Language Processing, {EMNLP} 2025, Suzhou, China, November 4-9, 2025},
	pages            = {20275--20321},
	publisher        = {Association for Computational Linguistics},
	year             = {2025},
	bburl             = {https://bdoi.org/10.18653/v1/2025.emnlp-main.1025},
	bdoi              = {10.18653/V1/2025.EMNLP-MAIN.1025},
	bibburl           = {https://dblp.org/rec/conf/emnlp/MuennighoffYSLFHZLCH25.bib},
	bibsource        = {dblp computer science bibliography, https://dblp.org},
}

@InProceedings{Shen2025RKGED,
	author="Song, Tengwei
	and Ma, Xudong
	and Liu, Yang
	and Luo, Jie
	and Hoehndorf, Robert",
	beditor="Acosta, Maribel
	and van Erp, Marieke
	and Rudolph, Sebastian
	and Hartig, Olaf
	and Spahiu, Blerina
	and Rula, Anisa
	and Garijo, Daniel
	and Osborne, Francesco",
	title="Robust Knowledge Graph Embedding via Denoising",
	booktitle="The Semantic Web",
	year="2026",
	publisher="Springer Nature Switzerland",
	address="Cham",
	pages="417--435",
	isbn="978-3-032-25156-5"
}

@inproceedings{Cohen2019Smoothing,
	author           = {Jeremy Cohen and
	Elan Rosenfeld and
	J. Zico Kolter},
	beditor           = {Kamalika Chaudhuri and
	Ruslan Salakhutdinov},
	title            = {Certified Adversarial Robustness via Randomized Smoothing},
	booktitle        = {Proceedings of the 36th International Conference on Machine Learning,
	{ICML} 2019, 9-15 June 2019, Long Beach, California, {USA}},
	series           = {Proceedings of Machine Learning Research},
	pages            = {1310--1320},
	publisher        = {{PMLR}},
	year             = {2019},
	bburl             = {http://proceedings.mlr.press/v97/cohen19c.html},
	bibburl           = {https://dblp.org/rec/conf/icml/CohenRK19.bib},
	bibsource        = {dblp computer science bibliography, https://dblp.org},
}

@inproceedings{SalemiZamani2024eRAG,
	author           = {Alireza Salemi and
	Hamed Zamani},
	beditor           = {Grace Hui Yang and
	Hongning Wang and
	Sam Han and
	Claudia Hauff and
	Guido Zuccon and
	Yi Zhang},
	title            = {Evaluating Retrieval Quality in Retrieval-Augmented Generation},
	booktitle        = {Proceedings of the 47th International {ACM} {SIGIR} Conference on
	Research and Development in Information Retrieval, {SIGIR} 2024, Washington
	DC, USA, July 14-18, 2024},
	pages            = {2395--2400},
	publisher        = {{ACM}},
	year             = {2024},
	bburl             = {https://bdoi.org/10.1145/3626772.3657957},
	bdoi              = {10.1145/3626772.3657957},
	bibburl           = {https://dblp.org/rec/conf/sigir/SalemiZ24a.bib},
	bibsource        = {dblp computer science bibliography, https://dblp.org},
}

@inproceedings{KalaiVempala2024Calibration,
	author           = {Adam Tauman Kalai and
	Santosh S. Vempala},
	beditor           = {Bojan Mohar and
	Igor Shinkar and
	Ryan O'Donnell},
	title            = {Calibrated Language Models Must Hallucinate},
	booktitle        = {Proceedings of the 56th Annual {ACM} Symposium on Theory of Computing,
	{STOC} 2024, Vancouver, BC, Canada, June 24-28, 2024},
	pages            = {160--171},
	publisher        = {{ACM}},
	year             = {2024},
	bburl             = {https://bdoi.org/10.1145/3618260.3649777},
	bdoi              = {10.1145/3618260.3649777},
	bibburl           = {https://dblp.org/rec/conf/stoc/KalaiV24.bib},
	bibsource        = {dblp computer science bibliography, https://dblp.org},
}

@article{BogomolnaiaJackson2002,
	author           = {Anna Bogomolnaia and
	Matthew O. Jackson},
	title            = {The Stability of Hedonic Coalition Structures},
	journal          = {Games Econ. Behav.},
	volume           = {38},
	number           = {2},
	pages            = {201--230},
	year             = {2002},
	bburl             = {https://bdoi.org/10.1006/game.2001.0877},
	bdoi              = {10.1006/GAME.2001.0877},
	bibburl           = {https://dblp.org/rec/journals/geb/BogomolnaiaJ02.bib},
	bibsource        = {dblp computer science bibliography, https://dblp.org},
}

@article{Turing1936,
	author           = {Alan M. Turing},
	title            = {On computable numbers, with an application to the Entscheidungsproblem},
	journal          = {Proc. London Math. Soc.},
	volume           = {s2-42},
	number           = {1},
	pages            = {230--265},
	year             = {1937},
	bburl             = {https://bdoi.org/10.1112/plms/s2-42.1.230},
	bdoi              = {10.1112/PLMS/S2-42.1.230},
	urn              = {urn:nbn:de:hbz:6:1-85171},
	bibburl           = {https://dblp.org/rec/journals/x/Turing37.bib},
	bibsource        = {dblp computer science bibliography, https://dblp.org},
}

@inproceedings{KleinbergMullainathanRaghavan2016,
	author           = {Jon M. Kleinberg and
	Sendhil Mullainathan and
	Manish Raghavan},
	beditor           = {Christos H. Papadimitriou},
	title            = {Inherent Trade-Offs in the Fair Determination of Risk Scores},
	booktitle        = {8th Innovations in Theoretical Computer Science Conference, {ITCS}
	2017, Berkeley, CA, USA, January 9-11, 2017},
	series           = {LIPIcs},
	pages            = {43:1--43:23},
	publisher        = {Schloss Dagstuhl - Leibniz-Zentrum f{\"{u}}r Informatik},
	year             = {2017},
	bburl             = {https://bdoi.org/10.4230/LIPIcs.ITCS.2017.43},
	bdoi              = {10.4230/LIPICS.ITCS.2017.43},
	bibburl           = {https://dblp.org/rec/conf/innovations/KleinbergMR17.bib},
	bibsource        = {dblp computer science bibliography, https://dblp.org},
}

@inproceedings{Smolensky1987,
	author           = {Roman Smolensky},
	beditor           = {Alfred V. Aho},
	title            = {Algebraic Methods in the Theory of Lower Bounds for Boolean Circuit
	Complexity},
	booktitle        = {Proceedings of the 19th Annual {ACM} Symposium on Theory of Computing,
	1987, New York, New York, {USA}},
	pages            = {77--82},
	publisher        = {{ACM}},
	year             = {1987},
	bburl             = {https://bdoi.org/10.1145/28395.28404},
	bdoi              = {10.1145/28395.28404},
	bibburl           = {https://dblp.org/rec/conf/stoc/Smolensky87.bib},
	bibsource        = {dblp computer science bibliography, https://dblp.org},
}

@inproceedings{jimenez2024swebench,
	author           = {Carlos E. Jimenez and
	John Yang and
	Alexander Wettig and
	Shunyu Yao and
	Kexin Pei and
	Ofir Press and
	Karthik R. Narasimhan},
	title            = {SWE-bench: Can Language Models Resolve Real-world Github Issues?},
	booktitle        = {The Twelfth International Conference on Learning Representations,
	{ICLR} 2024, Vienna, Austria, May 7-11, 2024},
	publisher        = {OpenReview.net},
	year             = {2024},
	bburl             = {https://openreview.net/forum?id=VTF8yNQM66},
	bibburl           = {https://dblp.org/rec/conf/iclr/JimenezYWYPPN24.bib},
	bibsource        = {dblp computer science bibliography, https://dblp.org},
}

@inproceedings{ortiz2023task,
	author           = {Guillermo Ortiz{-}Jim{\'{e}}nez and
	Alessandro Favero and
	Pascal Frossard},
	beditor           = {Alice Oh and
	Tristan Naumann and
	Amir Globerson and
	Kate Saenko and
	Moritz Hardt and
	Sergey Levine},
	title            = {Task Arithmetic in the Tangent Space: Improved Editing of Pre-Trained
	Models},
	booktitle        = {Advances in Neural Information Processing Systems 36: Annual Conference
	on Neural Information Processing Systems 2023, NeurIPS 2023, New Orleans,
	LA, USA, December 10 - 16, 2023},
	year             = {2023},
	bburl             = {http://papers.nips.cc/paper\_files/paper/2023/hash/d28077e5ff52034cd35b4aa15320caea-Abstract-Conference.html},
	bibburl           = {https://dblp.org/rec/conf/nips/Ortiz-JimenezFF23.bib},
	bibsource        = {dblp computer science bibliography, https://dblp.org},
}

@inproceedings{Wang2023DecodingTrust,
	author           = {Boxin Wang and
	Weixin Chen and
	Hengzhi Pei and
	Chulin Xie and
	Mintong Kang and
	Chenhui Zhang and
	Chejian Xu and
	Zidi Xiong and
	Ritik Dutta and
	Rylan Schaeffer and
	Sang T. Truong and
	Simran Arora and
	Mantas Mazeika and
	Dan Hendrycks and
	Zinan Lin and
	Yu Cheng and
	Sanmi Koyejo and
	Dawn Song and
	Bo Li},
	beditor           = {Alice Oh and
	Tristan Naumann and
	Amir Globerson and
	Kate Saenko and
	Moritz Hardt and
	Sergey Levine},
	title            = {DecodingTrust: {A} Comprehensive Assessment of Trustworthiness in
	{GPT} Models},
	booktitle        = {Advances in Neural Information Processing Systems 36: Annual Conference
	on Neural Information Processing Systems 2023, NeurIPS 2023, New Orleans,
	LA, USA, December 10 - 16, 2023},
	year             = {2023},
	bburl             = {http://papers.nips.cc/paper\_files/paper/2023/hash/63cb9921eecf51bfad27a99b2c53dd6d-Abstract-Datasets\_and\_Benchmarks.html},
	bibburl           = {https://dblp.org/rec/conf/nips/WangCPXKZXXDSTA23.bib},
	bibsource        = {dblp computer science bibliography, https://dblp.org},
}

@inproceedings{ilharco2023editing,
	author           = {Gabriel Ilharco and
	Marco T{\'{u}}lio Ribeiro and
	Mitchell Wortsman and
	Ludwig Schmidt and
	Hannaneh Hajishirzi and
	Ali Farhadi},
	title            = {Editing models with task arithmetic},
	booktitle        = {The Eleventh International Conference on Learning Representations,
	{ICLR} 2023, Kigali, Rwanda, May 1-5, 2023},
	publisher        = {OpenReview.net},
	year             = {2023},
	bburl             = {https://openreview.net/forum?id=6t0Kwf8-jrj},
	bibburl           = {https://dblp.org/rec/conf/iclr/IlharcoRWSHF23.bib},
	bibsource        = {dblp computer science bibliography, https://dblp.org},
}

@inproceedings{Sun2024TrustLLM,
	author           = {Yue Huang and
	Lichao Sun and
	Haoran Wang and
	Siyuan Wu and
	Qihui Zhang and
	Yuan Li and
	Chujie Gao and
	Yixin Huang and
	Wenhan Lyu and
	Yixuan Zhang and
	Xiner Li and
	Hanchi Sun and
	Zhengliang Liu and
	Yixin Liu and
	Yijue Wang and
	Zhikun Zhang and
	Bertie Vidgen and
	Bhavya Kailkhura and
	Caiming Xiong and
	Chaowei Xiao and
	Chunyuan Li and
	Eric P. Xing and
	Furong Huang and
	Hao Liu and
	Heng Ji and
	Hongyi Wang and
	Huan Zhang and
	Huaxiu Yao and
	Manolis Kellis and
	Marinka Zitnik and
	Meng Jiang and
	Mohit Bansal and
	James Zou and
	Jian Pei and
	Jian Liu and
	Jianfeng Gao and
	Jiawei Han and
	Jieyu Zhao and
	Jiliang Tang and
	Jindong Wang and
	Joaquin Vanschoren and
	John C. Mitchell and
	Kai Shu and
	Kaidi Xu and
	Kai{-}Wei Chang and
	Lifang He and
	Lifu Huang and
	Michael Backes and
	Neil Zhenqiang Gong and
	Philip S. Yu and
	Pin{-}Yu Chen and
	Quanquan Gu and
	Ran Xu and
	Rex Ying and
	Shuiwang Ji and
	Suman Jana and
	Tianlong Chen and
	Tianming Liu and
	Tianyi Zhou and
	William Wang and
	Xiang Li and
	Xiangliang Zhang and
	Xiao Wang and
	Xing Xie and
	Xun Chen and
	Xuyu Wang and
	Yan Liu and
	Yanfang Ye and
	Yinzhi Cao and
	Yong Chen and
	Yue Zhao},
	beditor           = {Ruslan Salakhutdinov and
	Zico Kolter and
	Katherine A. Heller and
	Adrian Weller and
	Nuria Oliver and
	Jonathan Scarlett and
	Felix Berkenkamp},
	title            = {Position: TrustLLM: Trustworthiness in Large Language Models},
	booktitle        = {Forty-first International Conference on Machine Learning, {ICML} 2024,
	Vienna, Austria, July 21-27, 2024},
	series           = {Proceedings of Machine Learning Research},
	pages            = {20166--20270},
	publisher        = {{PMLR} / OpenReview.net},
	year             = {2024},
	bburl             = {https://proceedings.mlr.press/v235/huang24x.html},
	bibburl           = {https://dblp.org/rec/conf/icml/Huang0WWZLGHLZL24.bib},
	bibsource        = {dblp computer science bibliography, https://dblp.org},
}

@inproceedings{abbasi2011improved,
	author           = {Yasin Abbasi{-}Yadkori and
	D{\'{a}}vid P{\'{a}}l and
	Csaba Szepesv{\'{a}}ri},
	beditor           = {John Shawe{-}Taylor and
	Richard S. Zemel and
	Peter L. Bartlett and
	Fernando C. N. Pereira and
	Kilian Q. Weinberger},
	title            = {Improved Algorithms for Linear Stochastic Bandits},
	booktitle        = {Advances in Neural Information Processing Systems 24: 25th Annual
	Conference on Neural Information Processing Systems 2011. Proceedings
	of a meeting held 12-14 December 2011, Granada, Spain},
	pages            = {2312--2320},
	year             = {2011},
	bburl             = {https://proceedings.neurips.cc/paper/2011/hash/e1d5be1c7f2f456670de3d53c7b54f4a-Abstract.html},
	bibburl           = {https://dblp.org/rec/conf/nips/Abbasi-YadkoriPS11.bib},
	bibsource        = {dblp computer science bibliography, https://dblp.org},
}

@book{Arrow1951,
	title            = {Social Choice and Individual Values},
	isbn             = {9780300186987},
	bburl             = {http://dx.bdoi.org/10.12987/9780300186987},
	bdoi              = {10.12987/9780300186987},
	publisher        = {Yale University Press},
	author           = {Arrow, Kenneth J.},
	year             = {2017},
	month            = {Dec},
}

@article{benjamini1995controlling,
	title            = {Controlling the False Discovery Rate: A Practical and Powerful Approach to Multiple Testing},
	volume           = {57},
	bISSN             = {1467-9868},
	bburl             = {http://dx.bdoi.org/10.1111/j.2517-6161.1995.tb02031.x},
	bdoi              = {10.1111/j.2517-6161.1995.tb02031.x},
	number           = {1},
	journal          = {Journal of the Royal Statistical Society Series B: Statistical Methodology},
	publisher        = {Oxford University Press (OUP)},
	author           = {Benjamini, Yoav and Hochberg, Yosef},
	year             = {1995},
	month            = {Jan},
	pages            = {289–300},
}

@article{bergemann2024data,
	title            = {Data, Competition, and Digital Platforms},
	bISSN             = {1556-5068},
	bburl             = {http://dx.bdoi.org/10.2139/ssrn.4236337},
	bdoi              = {10.2139/ssrn.4236337},
	journal          = {SSRN Electronic Journal},
	publisher        = {Elsevier BV},
	author           = {Bergemann, Dirk and Bonatti, Alessandro},
	year             = {2022},
}

@inproceedings{bojchevski2019certifiable,
	author           = {Aleksandar Bojchevski and
	Stephan G{\"{u}}nnemann},
	beditor           = {Hanna M. Wallach and
	Hugo Larochelle and
	Alina Beygelzimer and
	Florence d'Alch{\'{e}}{-}Buc and
	Emily B. Fox and
	Roman Garnett},
	title            = {Certifiable Robustness to Graph Perturbations},
	booktitle        = {Advances in Neural Information Processing Systems 32: Annual Conference
	on Neural Information Processing Systems 2019, NeurIPS 2019, December
	8-14, 2019, Vancouver, BC, Canada},
	pages            = {8317--8328},
	year             = {2019},
	bburl             = {https://proceedings.neurips.cc/paper/2019/hash/e2f374c3418c50bc30d67d5f7454a5b4-Abstract.html},
	bibburl           = {https://dblp.org/rec/conf/nips/BojchevskiG19.bib},
	bibsource        = {dblp computer science bibliography, https://dblp.org},
}

@article{catoni2007pac,
	bdoi              = {10.48550/ARXIV.0712.0248},
	bburl             = {https://arxiv.org/abs/0712.0248},
	author           = {Catoni, Olivier},
	title            = {Pac-Bayesian Supervised Classification: The Thermodynamics of Statistical Learning},
	journal={arXiv preprint},
	volume={arXiv.0712.0248},
	year             = {2007},
}

@inproceedings{chiang2023tighter,
	author = {Chiang, David and Cholak, Peter and Pillay, Anand},
	title = {Tighter bounds on the expressivity of transformer encoders},
	year = {2023},
	publisher = {JMLR.org},
	booktitle = {Proceedings of the 40th International Conference on Machine Learning},
	articleno = {221},
	numpages = {19},
	location = {Honolulu, Hawaii, USA},
	series = {ICML'23}
}

@inproceedings{duan2024gtbench,
	title={{GTB}ench: Uncovering the Strategic Reasoning Capabilities of {LLM}s via Game-Theoretic Evaluations},
	author={Jinhao Duan and Renming Zhang and James Diffenderfer and Bhavya Kailkhura and Lichao Sun and Elias Stengel-Eskin and Mohit Bansal and Tianlong Chen and Kaidi Xu},
	booktitle={The Thirty-eighth Annual Conference on Neural Information Processing Systems},
	year={2024},
	burl={https://openreview.net/forum?id=ypggxVWIv2}
}

@inproceedings{dziugaite2017computing,
	author           = {Gintare Karolina Dziugaite and
	Daniel M. Roy},
	beditor           = {Gal Elidan and
	Kristian Kersting and
	Alexander Ihler},
	title            = {Computing Nonvacuous Generalization Bounds for Deep (Stochastic) Neural
	Networks with Many More Parameters than Training Data},
	booktitle        = {Proceedings of the Thirty-Third Conference on Uncertainty in Artificial
	Intelligence, {UAI} 2017, Sydney, Australia, August 11-15, 2017},
	publisher        = {{AUAI} Press},
	year             = {2017},
	bburl             = {http://auai.org/uai2017/proceedings/papers/173.pdf},
	bibburl           = {https://dblp.org/rec/conf/uai/DziugaiteR17.bib},
	bibsource        = {dblp computer science bibliography, https://dblp.org},
}

@inproceedings{gao2023rarr,
	author           = {Luyu Gao and
	Zhuyun Dai and
	Panupong Pasupat and
	Anthony Chen and
	Arun Tejasvi Chaganty and
	Yicheng Fan and
	Vincent Y. Zhao and
	Ni Lao and
	Hongrae Lee and
	Da{-}Cheng Juan and
	Kelvin Guu},
	beditor           = {Anna Rogers and
	Jordan L. Boyd{-}Graber and
	Naoaki Okazaki},
	title            = {{RARR:} Researching and Revising What Language Models Say, Using Language
	Models},
	booktitle        = {Proceedings of the 61st Annual Meeting of the Association for Computational
	Linguistics (Volume 1: Long Papers), {ACL} 2023, Toronto, Canada,
	July 9-14, 2023},
	pages            = {16477--16508},
	publisher        = {Association for Computational Linguistics},
	year             = {2023},
	bburl             = {https://bdoi.org/10.18653/v1/2023.acl-long.910},
	bdoi              = {10.18653/V1/2023.ACL-LONG.910},
	bibburl           = {https://dblp.org/rec/conf/acl/GaoDPCCFZLLJG23.bib},
	bibsource        = {dblp computer science bibliography, https://dblp.org},
}

@book{immerman1999descriptive,
	title            = {Descriptive Complexity},
	isbn             = {9781461205395},
	bburl             = {http://dx.bdoi.org/10.1007/978-1-4612-0539-5},
	bdoi              = {10.1007/978-1-4612-0539-5},
	publisher        = {Springer New York},
	author           = {Immerman, Neil},
	year             = {1999},
}

@inproceedings{jin2025searchr1,
	title={Search-R1: Training {LLM}s to Reason and Leverage Search Engines with Reinforcement Learning},
	author={Bowen Jin and Hansi Zeng and Zhenrui Yue and Jinsung Yoon and Sercan O Arik and Dong Wang and Hamed Zamani and Jiawei Han},
	booktitle={Second Conference on Language Modeling},
	year={2025},
	burl={https://openreview.net/forum?id=Rwhi91ideu}
}

@inproceedings{kakade2002approximately,
	author           = {Sham M. Kakade and
	John Langford},
	beditor           = {Claude Sammut and
	Achim G. Hoffmann},
	title            = {Approximately Optimal Approximate Reinforcement Learning},
	booktitle        = {Machine Learning, Proceedings of the Nineteenth International Conference
	{(ICML} 2002), University of New South Wales, Sydney, Australia, July
	8-12, 2002},
	pages            = {267--274},
	publisher        = {Morgan Kaufmann},
	year             = {2002},
	bibburl           = {https://dblp.org/rec/conf/icml/KakadeL02.bib},
	bibsource        = {dblp computer science bibliography, https://dblp.org},
}

@article{lai2001sequential,
	author  = {Lai, Tze Leung},
	title   = {Sequential Analysis: Some Classical Problems and New Challenges},
	journal = {Statistica Sinica},
	volume  = {11},
	number  = {2},
	pages   = {303--408},
	year    = {2001},
	month   = apr,
	bnote    = {With discussion and rejoinder},
	burl     = {https://www3.stat.sinica.edu.tw/statistica/J11n2/j11n21/j11n21.htm},
}

@article{li2017obviously,
	title            = {Obviously Strategy-Proof Mechanisms},
	bISSN             = {1556-5068},
	bburl             = {http://dx.bdoi.org/10.2139/ssrn.2560028},
	bdoi              = {10.2139/ssrn.2560028},
	journal          = {SSRN Electronic Journal},
	publisher        = {Elsevier BV},
	author           = {Li, Shengwu},
	year             = {2015},
}

@inproceedings{lightman2024lets,
	author           = {Hunter Lightman and
	Vineet Kosaraju and
	Yuri Burda and
	Harrison Edwards and
	Bowen Baker and
	Teddy Lee and
	Jan Leike and
	John Schulman and
	Ilya Sutskever and
	Karl Cobbe},
	title            = {Let's Verify Step by Step},
	booktitle        = {The Twelfth International Conference on Learning Representations,
	{ICLR} 2024, Vienna, Austria, May 7-11, 2024},
	publisher        = {OpenReview.net},
	year             = {2024},
	bburl             = {https://openreview.net/forum?id=v8L0pN6EOi},
	bibburl           = {https://dblp.org/rec/conf/iclr/LightmanKBEBLLS24.bib},
	bibsource        = {dblp computer science bibliography, https://dblp.org},
}

@book{pearl2009causality,
	title            = {Causality: Models, Reasoning, and Inference},
	isbn             = {9780521749190},
	bburl             = {http://dx.bdoi.org/10.1017/cbo9780511803161},
	bdoi              = {10.1017/cbo9780511803161},
	publisher        = {Cambridge University Press},
	author           = {Pearl, Judea},
	year             = {2009},
	month            = {Sept},
}

@article{pycia2017simplicity,
	title            = {Incentive compatible allocation and exchange of discrete resources: Allocation and exchange of discrete resources},
	volume           = {12},
	bISSN             = {1933-6837},
	bburl             = {http://dx.bdoi.org/10.3982/te2201},
	bdoi              = {10.3982/te2201},
	number           = {1},
	journal          = {Theoretical Economics},
	publisher        = {The Econometric Society},
	author           = {Pycia, Marek and Ünver, M. Utku},
	year             = {2017},
	month            = {Jan},
	pages            = {287–329},
}

@inproceedings{ren2024iterative,
	author           = {Wei Xiong and
	Hanze Dong and
	Chenlu Ye and
	Ziqi Wang and
	Han Zhong and
	Heng Ji and
	Nan Jiang and
	Tong Zhang},
	beditor           = {Ruslan Salakhutdinov and
	Zico Kolter and
	Katherine A. Heller and
	Adrian Weller and
	Nuria Oliver and
	Jonathan Scarlett and
	Felix Berkenkamp},
	title            = {Iterative Preference Learning from Human Feedback: Bridging Theory
	and Practice for {RLHF} under KL-constraint},
	booktitle        = {Forty-first International Conference on Machine Learning, {ICML} 2024,
	Vienna, Austria, July 21-27, 2024},
	series           = {Proceedings of Machine Learning Research},
	pages            = {54715--54754},
	publisher        = {{PMLR} / OpenReview.net},
	year             = {2024},
	bburl             = {https://proceedings.mlr.press/v235/xiong24a.html},
	bibburl           = {https://dblp.org/rec/conf/icml/0015DYW0J0024.bib},
	bibsource        = {dblp computer science bibliography, https://dblp.org},
}

@inproceedings{song2024importance,
	series           = {NeurIPS 2024},
	title            = {The Importance of Online Data: Understanding Preference Fine-tuning via Coverage},
	bburl             = {http://dx.bdoi.org/10.52202/079017-0392},
	bdoi              = {10.52202/079017-0392},
	booktitle        = {Advances in Neural Information Processing Systems 37},
	publisher        = {Neural Information Processing Systems Foundation, Inc. (NeurIPS)},
	author           = {Bagnell, J. and Singh, Aarti and Song, Yuda and Sun, Wen and Swamy, Gokul},
	year             = {2024},
	pages            = {12243–12270},
	collection       = {NeurIPS 2024},
}

@inproceedings{Sun2024zkLLM,
	series           = {CCS ’24},
	title            = {zkLLM: Zero Knowledge Proofs for Large Language Models},
	bburl             = {http://dx.bdoi.org/10.1145/3658644.3670334},
	bdoi              = {10.1145/3658644.3670334},
	booktitle        = {Proceedings of the 2024 on ACM SIGSAC Conference on Computer and Communications Security},
	publisher        = {ACM},
	author           = {Sun, Haochen and Li, Jason and Zhang, Hongyang},
	year             = {2024},
	month            = {Dec},
	pages            = {4405–4419},
	collection       = {CCS ’24},
}

@inproceedings{Tang2024MultiHopRAG,
	title={MultiHop-{RAG}: Benchmarking Retrieval-Augmented Generation for Multi-Hop Queries},
	author={Yixuan Tang and Yi Yang},
	booktitle={First Conference on Language Modeling},
	year={2024},
	burl={https://openreview.net/forum?id=t4eB3zYWBK}
}

@inproceedings{trivedi2023interleaving,
	author           = {Harsh Trivedi and
	Niranjan Balasubramanian and
	Tushar Khot and
	Ashish Sabharwal},
	beditor           = {Anna Rogers and
	Jordan L. Boyd{-}Graber and
	Naoaki Okazaki},
	title            = {Interleaving Retrieval with Chain-of-Thought Reasoning for Knowledge-Intensive
	Multi-Step Questions},
	booktitle        = {Proceedings of the 61st Annual Meeting of the Association for Computational
	Linguistics (Volume 1: Long Papers), {ACL} 2023, Toronto, Canada,
	July 9-14, 2023},
	pages            = {10014--10037},
	publisher        = {Association for Computational Linguistics},
	year             = {2023},
	bburl             = {https://bdoi.org/10.18653/v1/2023.acl-long.557},
	bdoi              = {10.18653/V1/2023.ACL-LONG.557},
	bibburl           = {https://dblp.org/rec/conf/acl/TrivediBKS23.bib},
	bibsource        = {dblp computer science bibliography, https://dblp.org},
}

@article{wald1947sequential,
	title            = {Sequential Analysis},
	volume           = {42},
	bISSN             = {0162-1459},
	bburl             = {http://dx.bdoi.org/10.2307/2280027},
	bdoi              = {10.2307/2280027},
	number           = {240},
	journal          = {Journal of the American Statistical Association},
	publisher        = {JSTOR},
	author           = {Barnard, G. A. and Ferber, Robert and Wald, Abraham},
	year             = {1947},
	month            = {Dec},
	pages            = {658},
}

@book{wegener1987complexity,
	author    = {Wegener, Ingo},
	title     = {The Complexity of {Boolean} Functions},
	publisher = {John Wiley \& Sons},
	address   = {Chichester, UK},
	series    = {Wiley-Teubner Series in Computer Science},
	year      = {1987},
	month     = aug,
	isbn      = {0-471-91555-6},
	bnote      = {Co-published with B.~G. Teubner, Stuttgart. Freely available at \url{https://eccc.weizmann.ac.il/static/books/The_Complexity_of_Boolean_Functions/}},
}

@article{zhang2023adalora,
	title={AdaLoRA: Adaptive Budget Allocation for Parameter-Efficient Fine-Tuning}, 
	author={Qingru Zhang and Minshuo Chen and Alexander Bukharin and Nikos Karampatziakis and Pengcheng He and Yu Cheng and Weizhu Chen and Tuo Zhao},
	year={2023},
	journal={arXiv preprint},
	volume={arXiv.2303.10512},
	primaryClass={cs.CL},
	burl={https://arxiv.org/abs/2303.10512}, 
}

@article{Zheng2024NSGAII,
	author           = {Weijie Zheng and
	Benjamin Doerr},
	title            = {Runtime Analysis for the {NSGA-II:} Proving, Quantifying, and Explaining
	the Inefficiency for Many Objectives},
	journal          = {{IEEE} Trans. Evol. Comput.},
	volume           = {28},
	number           = {5},
	pages            = {1442--1454},
	year             = {2024},
	bburl             = {https://bdoi.org/10.1109/TEVC.2023.3320278},
	bdoi              = {10.1109/TEVC.2023.3320278},
	bibburl           = {https://dblp.org/rec/journals/tec/ZhengD24.bib},
	bibsource        = {dblp computer science bibliography, https://dblp.org},
}

@article{Hahn2020Limitations,
	title            = {Theoretical Limitations of Self-Attention in Neural Sequence Models},
	volume           = {8},
	bISSN             = {2307-387X},
	bburl             = {http://dx.bdoi.org/10.1162/tacl_a_00306},
	bdoi              = {10.1162/tacl_a_00306},
	journal          = {Transactions of the Association for Computational Linguistics},
	publisher        = {MIT Press - Journals},
	author           = {Hahn, Michael},
	year             = {2020},
	month            = {Dec},
	pages            = {156–171},
}

@article{PerezBarceloMarinkovic2021,
	author           = {Jorge P{\'{e}}rez and
	Pablo Barcel{\'{o}} and
	Javier Marinkovic},
	title            = {Attention is Turing-Complete},
	journal          = {J. Mach. Learn. Res.},
	volume           = {22},
	pages            = {75:1--75:35},
	year             = {2021},
	bburl             = {https://jmlr.org/papers/v22/20-302.html},
	bibburl           = {https://dblp.org/rec/journals/jmlr/PerezBM21.bib},
	bibsource        = {dblp computer science bibliography, https://dblp.org},
}

@inproceedings{Wang2024CoTDecoding,
	author           = {Xuezhi Wang and
	Denny Zhou},
	beditor           = {Amir Globersons and
	Lester Mackey and
	Danielle Belgrave and
	Angela Fan and
	Ulrich Paquet and
	Jakub M. Tomczak and
	Cheng Zhang},
	title            = {Chain-of-Thought Reasoning Without Prompting},
	booktitle        = {Advances in Neural Information Processing Systems 38: Annual Conference
	on Neural Information Processing Systems 2024, NeurIPS 2024, Vancouver,
	BC, Canada, December 10 - 15, 2024},
	year             = {2024},
	bburl             = {http://papers.nips.cc/paper\_files/paper/2024/hash/7a8e7fd295aa04eac4b470ae27f8785c-Abstract-Conference.html},
	bibburl           = {https://dblp.org/rec/conf/nips/0002Z24.bib},
	bibsource        = {dblp computer science bibliography, https://dblp.org},
}

@inproceedings{yao2023react,
	author           = {Shunyu Yao and
	Jeffrey Zhao and
	Dian Yu and
	Nan Du and
	Izhak Shafran and
	Karthik R. Narasimhan and
	Yuan Cao},
	title            = {ReAct: Synergizing Reasoning and Acting in Language Models},
	booktitle        = {The Eleventh International Conference on Learning Representations,
	{ICLR} 2023, Kigali, Rwanda, May 1-5, 2023},
	publisher        = {OpenReview.net},
	year             = {2023},
	bburl             = {https://openreview.net/forum?id=WE\_vluYUL-X},
	bibburl           = {https://dblp.org/rec/conf/iclr/YaoZYDSN023.bib},
	bibsource        = {dblp computer science bibliography, https://dblp.org},
}

@book{Russell2019,
	author    = {Russell, Stuart},
	title     = {Human Compatible: {Artificial Intelligence} and the Problem of Control},
	year      = {2019},
	month     = oct,
	publisher = {Viking},
	address   = {New York, NY},
	isbn      = {978-0-525-55861-3},
}

@article{magnus2019matrix,
	title            = {Matrix Differential Calculus with Applications in Statistics and Econometrics},
	volume           = {44},
	bISSN             = {0006-341X},
	bburl             = {http://dx.bdoi.org/10.2307/2531754},
	bdoi              = {10.2307/2531754},
	number           = {4},
	journal          = {Biometrics},
	publisher        = {JSTOR},
	author           = {Magnus, J. R. and Neudecker, H.},
	year             = {1988},
	month            = {Dec},
	pages            = {1209},
}

@article{CampbellFiske1959,
	title            = {Convergent and discriminant validation by the multitrait-multimethod matrix},
	volume           = {56},
	bISSN             = {0033-2909},
	bburl             = {http://dx.bdoi.org/10.1037/h0046016},
	bdoi              = {10.1037/h0046016},
	number           = {2},
	journal          = {Psychological Bulletin},
	publisher        = {American Psychological Association (APA)},
	author           = {Campbell, Donald T. and Fiske, Donald W.},
	year             = {1959},
	pages            = {81–105},
}

@inproceedings{Wang2024WISE,
	series           = {NeurIPS 2024},
	title            = {WISE: Rethinking the Knowledge Memory for Lifelong Model Editing of Large Language Models},
	bburl             = {http://dx.bdoi.org/10.52202/079017-1703},
	bdoi              = {10.52202/079017-1703},
	booktitle        = {Advances in Neural Information Processing Systems 37},
	publisher        = {Neural Information Processing Systems Foundation, Inc. (NeurIPS)},
	author           = {Chen, Huajun and Huang, Fei and Jiang, Yong and Li, Zexi and Wang, Peng and Xie, Pengjun and Xu, Ziwen and Yao, Yunzhi and Zhang, Ningyu},
	year             = {2024},
	pages            = {53764–53797},
	collection       = {NeurIPS 2024},
}

@inproceedings{Liu2025Scalpel,
	author           = {Xin Liu and
	Qiyang Song and
	Shaowen Xu and
	Kerou Zhou and
	Wenbo Jiang and
	Xiaoqi Jia and
	Weijuan Zhang and
	Heqing Huang and
	Yakai Li},
	beditor           = {In{\^{e}}s Lynce and
	Nello Murano and
	Mauro Vallati and
	Serena Villata and
	Federico Chesani and
	Michela Milano and
	Andrea Omicini and
	Mehdi Dastani},
	title            = {Latent Knowledge Scalpel: Precise and Massive Knowledge Editing for
	Large Language Models},
	booktitle        = {{ECAI} 2025 - 28th European Conference on Artificial Intelligence,
	25-30 October 2025, Bologna, Italy - Including 14th Conference on
	Prestigious Applications of Intelligent Systems {(PAIS} 2025)},
	series           = {Frontiers in Artificial Intelligence and Applications},
	pages            = {4378--4385},
	publisher        = {{IOS} Press},
	year             = {2025},
	bburl             = {https://bdoi.org/10.3233/FAIA251335},
	bdoi              = {10.3233/FAIA251335},
	bibburl           = {https://dblp.org/rec/conf/ecai/LiuSXZJJZ0L25.bib},
	bibsource        = {dblp computer science bibliography, https://dblp.org},
}

@inproceedings{Fei2025NeuralDB,
	title={Scaling Knowledge Editing in {LLM}s to 100,000 Facts with Neural {KV} Database},
	author={Weizhi Fei and Hao Shi and Jing Xu and Jingchen Peng and Jiazheng Li and Jingzhao Zhang and Bo Bai and Wei Han and Zhenyuan Chen and Xueyan Niu},
	booktitle={The Fourteenth International Conference on Learning Representations},
	year={2026},
	burl={https://openreview.net/forum?id=Z0CX62CSJQ}
}

@article{Shannon1948Communication,
	title            = {A Mathematical Theory of Communication},
	volume           = {27},
	bISSN             = {0005-8580},
	bburl             = {http://dx.bdoi.org/10.1002/j.1538-7305.1948.tb01338.x},
	bdoi              = {10.1002/j.1538-7305.1948.tb01338.x},
	number           = {3},
	journal          = {Bell System Technical Journal},
	publisher        = {Institute of Electrical and Electronics Engineers (IEEE)},
	author           = {Shannon, C. E.},
	year             = {1948},
	month            = {July},
	pages            = {379–423},
}

@article{Rice1953,
	title            = {Classes of recursively enumerable sets and their decision problems},
	volume           = {74},
	bISSN             = {0002-9947},
	bburl             = {http://dx.bdoi.org/10.1090/s0002-9947-1953-0053041-6},
	bdoi              = {10.1090/s0002-9947-1953-0053041-6},
	number           = {2},
	journal          = {Transactions of the American Mathematical Society},
	publisher        = {American Mathematical Society (AMS)},
	author           = {Rice, H. G.},
	year             = {1953},
	pages            = {358–366},
}

@article{Arrow1950Impossibility,
	title            = {A Difficulty in the Concept of Social Welfare},
	volume           = {58},
	bISSN             = {1537-534X},
	bburl             = {http://dx.bdoi.org/10.1086/256963},
	bdoi              = {10.1086/256963},
	number           = {4},
	journal          = {Journal of Political Economy},
	publisher        = {University of Chicago Press},
	author           = {Arrow, Kenneth J.},
	year             = {1950},
	month            = {Aug},
	pages            = {328–346},
}

@incollection{JohnsonLindenstrauss1984,
	author    = {Johnson, William B. and Lindenstrauss, Joram},
	title     = {Extensions of {Lipschitz} Mappings into a {Hilbert} Space},
	beditor    = {Beals, Richard and Beck, Anatole and Bellow, Alexandra and Hajian, Arshag},
	booktitle = {Conference in Modern Analysis and Probability},
	series    = {Contemporary Mathematics},
	volume    = {26},
	publisher = {American Mathematical Society},
	address   = {Providence, RI},
	year      = {1984},
	pages     = {189--206},
	bdoi       = {10.1090/conm/026/737400},
	isbn      = {978-0-8218-5030-5},
}

@article{Razborov1987,
	title            = {Lower bounds on the size of bounded depth circuits over a complete basis with logical addition},
	volume           = {41},
	bISSN             = {1573-8876},
	bburl             = {http://dx.bdoi.org/10.1007/bf01137685},
	bdoi              = {10.1007/bf01137685},
	number           = {4},
	journal          = {Mathematical bnotes of the Academy of Sciences of the USSR},
	publisher        = {Springer Science and Business Media LLC},
	author           = {Razborov, A. A.},
	year             = {1987},
	month            = {Apr},
	pages            = {333–338},
}

@article{BradleyTerry1952,
	title            = {Rank Analysis of Incomplete Block Designs: I. The Method of Paired Comparisons},
	volume           = {39},
	bISSN             = {0006-3444},
	bburl             = {http://dx.bdoi.org/10.2307/2334029},
	bdoi              = {10.2307/2334029},
	number           = {3/4},
	journal          = {Biometrika},
	publisher        = {JSTOR},
	author           = {Bradley, Ralph Allan and Terry, Milton E.},
	year             = {1952},
	month            = {Dec},
	pages            = {324},
}

@article{Vickrey1961,
	author    = {Vickrey, William},
	title     = {Counterspeculation, Auctions, and Competitive Sealed Tenders},
	journal   = {The Journal of Finance},
	volume    = {16},
	number    = {1},
	pages     = {8--37},
	year      = {1961},
	month     = mar,
	publisher = {Wiley},
	bISSN      = {0022-1082},
	bdoi       = {10.1111/j.1540-6261.1961.tb02789.x},
}

@article{Clarke1971,
	title            = {Multipart pricing of public goods},
	volume           = {11},
	bISSN             = {1573-7101},
	bburl             = {http://dx.bdoi.org/10.1007/bf01726210},
	bdoi              = {10.1007/bf01726210},
	number           = {1},
	journal          = {Public Choice},
	publisher        = {Springer Science and Business Media LLC},
	author           = {Clarke, Edward H.},
	year             = {1971},
	month            = {Sept},
	pages            = {17–33},
}

@article{Groves1973,
	title            = {Incentives in Teams},
	volume           = {41},
	bISSN             = {0012-9682},
	bburl             = {http://dx.bdoi.org/10.2307/1914085},
	bdoi              = {10.2307/1914085},
	number           = {4},
	journal          = {Econometrica},
	publisher        = {JSTOR},
	author           = {Groves, Theodore},
	year             = {1973},
	month            = {July},
	pages            = {617},
}

@article{Liang2023HELM,
	title            = {Holistic Evaluation of Language Models},
	volume           = {1525},
	bISSN             = {1749-6632},
	bburl             = {http://dx.bdoi.org/10.1111/nyas.15007},
	bdoi              = {10.1111/nyas.15007},
	number           = {1},
	journal          = {Annals of the New York Academy of Sciences},
	publisher        = {Wiley},
	author           = {Bommasani, Rishi and Liang, Percy and Lee, Tony},
	year             = {2023},
	month            = {May},
	pages            = {140–146},
}

@inproceedings{lev2012convergence,
	author           = {Omer Lev and
	Jeffrey S. Rosenschein},
	beditor           = {Wiebe van der Hoek and
	Lin Padgham and
	Vincent Conitzer and
	Michael Winikoff},
	title            = {Convergence of iterative voting},
	booktitle        = {International Conference on Autonomous Agents and Multiagent Systems,
	{AAMAS} 2012, Valencia, Spain, June 4-8, 2012 {(3} Volumes)},
	pages            = {611--618},
	publisher        = {{IFAAMAS}},
	year             = {2012},
	bburl             = {http://dl.acm.org/citation.cfm?id=2343784},
	bibburl           = {https://dblp.org/rec/conf/aamas/LevR12.bib},
	bibsource        = {dblp computer science bibliography, https://dblp.org},
}

@inproceedings{dutting2024mechanism,
	author           = {Paul D{\"{u}}tting and
	Vahab Mirrokni and
	Renato Paes Leme and
	Haifeng Xu and
	Song Zuo},
	beditor           = {Tat{-}Seng Chua and
	Chong{-}Wah Ngo and
	Ravi Kumar and
	Hady W. Lauw and
	Roy Ka{-}Wei Lee},
	title            = {Mechanism Design for Large Language Models},
	booktitle        = {Proceedings of the {ACM} on Web Conference 2024, {WWW} 2024, Singapore,
	May 13-17, 2024},
	pages            = {144--155},
	publisher        = {{ACM}},
	year             = {2024},
	bburl             = {https://bdoi.org/10.1145/3589334.3645511},
	bdoi              = {10.1145/3589334.3645511},
	bibburl           = {https://dblp.org/rec/conf/www/DuttingMLXZ24.bib},
	bibsource        = {dblp computer science bibliography, https://dblp.org},
}

@article{guo2025deepseek,
	author           = {Daya Guo and
	Dejian Yang and
	Haowei Zhang and
	Junxiao Song and
	Peiyi Wang and
	Qihao Zhu and
	Runxin Xu and
	Ruoyu Zhang and
	Shirong Ma and
	Xiao Bi and
	Xiaokang Zhang and
	Xingkai Yu and
	Yu Wu and
	Z. F. Wu and
	Zhibin Gou and
	Zhihong Shao and
	Zhuoshu Li and
	Ziyi Gao and
	Aixin Liu and
	Bing Xue and
	Bingxuan Wang and
	Bochao Wu and
	Bei Feng and
	Chengda Lu and
	Chenggang Zhao and
	Chengqi Deng and
	Chong Ruan and
	Damai Dai and
	Deli Chen and
	Dongjie Ji and
	Erhang Li and
	Fangyun Lin and
	Fucong Dai and
	Fuli Luo and
	Guangbo Hao and
	Guanting Chen and
	Guowei Li and
	Hao Zhang and
	Hanwei Xu and
	Honghui Ding and
	Huazuo Gao and
	Hui Qu and
	Hui Li and
	Jianzhong Guo and
	Jiashi Li and
	Jingchang Chen and
	Jingyang Yuan and
	Jinhao Tu and
	Junjie Qiu and
	Junlong Li and
	J. L. Cai and
	Jiaqi Ni and
	Jian Liang and
	Jin Chen and
	Kai Dong and
	Kai Hu and
	Kaichao You and
	Kaige Gao and
	Kang Guan and
	Kexin Huang and
	Kuai Yu and
	Lean Wang and
	Lecong Zhang and
	Liang Zhao and
	Litong Wang and
	Liyue Zhang and
	Lei Xu and
	Leyi Xia and
	Mingchuan Zhang and
	Minghua Zhang and
	Minghui Tang and
	Mingxu Zhou and
	Meng Li and
	Miaojun Wang and
	Mingming Li and
	Ning Tian and
	Panpan Huang and
	Peng Zhang and
	Qiancheng Wang and
	Qinyu Chen and
	Qiushi Du and
	Ruiqi Ge and
	Ruisong Zhang and
	Ruizhe Pan and
	Runji Wang and
	R. J. Chen and
	R. L. Jin and
	Ruyi Chen and
	Shanghao Lu and
	Shangyan Zhou and
	Shanhuang Chen and
	Shengfeng Ye and
	Shiyu Wang and
	Shuiping Yu and
	Shunfeng Zhou and
	Shuting Pan and
	S. S. Li and
	Shuang Zhou and
	Shaoqing Wu and
	Tao Yun and
	Tian Pei and
	Tianyu Sun and
	Tao Wang and
	Wangding Zeng and
	Wen Liu and
	Wenfeng Liang and
	Wenjun Gao and
	Wenqin Yu and
	Wentao Zhang and
	W. L. Xiao and
	Wei An and
	Xiaodong Liu and
	Xiaohan Wang and
	Xiaokang Chen and
	Xiaotao Nie and
	Xin Cheng and
	Xin Liu and
	Xin Xie and
	Xingchao Liu and
	Xinyu Yang and
	Xinyuan Li and
	Xuecheng Su and
	Xuheng Lin and
	X. Q. Li and
	Xiangyue Jin and
	Xiaojin Shen and
	Xiaosha Chen and
	Xiaowen Sun and
	Xiaoxiang Wang and
	Xinnan Song and
	Xinyi Zhou and
	Xianzu Wang and
	Xinxia Shan and
	Y. K. Li and
	Y. Q. Wang and
	Y. X. Wei and
	Yang Zhang and
	Yanhong Xu and
	Yao Li and
	Yao Zhao and
	Yaofeng Sun and
	Yaohui Wang and
	Yi Yu and
	Yichao Zhang and
	Yifan Shi and
	Yiliang Xiong and
	Ying He and
	Yishi Piao and
	Yisong Wang and
	Yixuan Tan and
	Yiyang Ma and
	Yiyuan Liu and
	Yongqiang Guo and
	Yuan Ou and
	Yuduan Wang and
	Yue Gong and
	Yuheng Zou and
	Yujia He and
	Yunfan Xiong and
	Yuxiang Luo and
	Yuxiang You and
	Yuxuan Liu and
	Yuyang Zhou and
	Y. X. Zhu and
	Yanping Huang and
	Yaohui Li and
	Yi Zheng and
	Yuchen Zhu and
	Yunxian Ma and
	Ying Tang and
	Yukun Zha and
	Yuting Yan and
	Z. Z. Ren and
	Zehui Ren and
	Zhangli Sha and
	Zhe Fu and
	Zhean Xu and
	Zhenda Xie and
	Zhengyan Zhang and
	Zhewen Hao and
	Zhicheng Ma and
	Zhigang Yan and
	Zhiyu Wu and
	Zihui Gu and
	Zijia Zhu and
	Zijun Liu and
	Zilin Li and
	Ziwei Xie and
	Ziyang Song and
	Zizheng Pan and
	Zhen Huang and
	Zhipeng Xu and
	Zhongyu Zhang and
	Zhen Zhang},
	title            = {DeepSeek-R1 incentivizes reasoning in LLMs through reinforcement learning},
	journal          = {Nat.},
	volume           = {645},
	number           = {8081},
	pages            = {633--638},
	year             = {2025},
	bburl             = {https://bdoi.org/10.1038/s41586-025-09422-z},
	bdoi              = {10.1038/S41586-025-09422-Z},
	bibburl           = {https://dblp.org/rec/journals/nature/GuoYZSWZXZMBZY025.bib},
	bibsource        = {dblp computer science bibliography, https://dblp.org},
}

@article{yu2022coco,
	title={COCO-DR: Combating Distribution Shifts in Zero-Shot Dense Retrieval with Contrastive and Distributionally Robust Learning}, 
	author={Yue Yu and Chenyan Xiong and Si Sun and Chao Zhang and Arnold Overwijk},
	year={2022},
	journal={arXiv preprint},
	volume={arXiv.2210.15212},
	primaryClass={cs.CL},
	burl={https://arxiv.org/abs/2210.15212}, 
}

@inproceedings{Vaswani2017Attention,
	author = {Vaswani, Ashish and Shazeer, Noam and Parmar, Niki and Uszkoreit, Jakob and Jones, Llion and Gomez, Aidan N. and Kaiser, \L{}ukasz and Polosukhin, Illia},
	title = {Attention is all you need},
	year = {2017},
	isbn = {9781510860964},
	publisher = {Curran Associates Inc.},
	address = {Red Hook, NY, USA},
	booktitle = {Proceedings of the 31st International Conference on Neural Information Processing Systems},
	pages = {6000–6010},
	numpages = {11},
	location = {Long Beach, California, USA},
	series = {NIPS'17}
}

@inproceedings{golovnev2023brakedown,
	author           = {Alexander Golovnev and
	Jonathan Lee and
	Srinath T. V. Setty and
	Justin Thaler and
	Riad S. Wahby},
	beditor           = {Helena Handschuh and
	Anna Lysyanskaya},
	title            = {Brakedown: Linear-Time and Field-Agnostic SNARKs for {R1CS}},
	booktitle        = {Advances in Cryptology - {CRYPTO} 2023 - 43rd Annual International
	Cryptology Conference, {CRYPTO} 2023, Santa Barbara, CA, USA, August
	20-24, 2023, Proceedings, Part {II}},
	series           = {Lecture bnotes in Computer Science},
	pages            = {193--226},
	publisher        = {Springer},
	year             = {2023},
	bburl             = {https://bdoi.org/10.1007/978-3-031-38545-2\_7},
	bdoi              = {10.1007/978-3-031-38545-2\_7},
	bibburl           = {https://dblp.org/rec/conf/crypto/GolovnevLSTW23.bib},
	bibsource        = {dblp computer science bibliography, https://dblp.org},
}

@inproceedings{canetti2025zk,
	author = {Canetti, Ran and Fiat, Amos and Gonczarowski, Yannai A.},
	title = {Zero-Knowledge Mechanisms},
	year = {2025},
	isbn = {9798400719431},
	publisher = {Association for Computing Machinery},
	address = {New York, NY, USA},
	burl = {https://bdoi.org/10.1145/3736252.3742535},
	bdoi = {10.1145/3736252.3742535},
	booktitle = {Proceedings of the 26th ACM Conference on Economics and Computation},
	pages = {338–339},
	numpages = {2},
	series = {EC '25}
}

@inproceedings{kothapalli2024hypernova,
	author           = {Abhiram Kothapalli and
	Srinath T. V. Setty},
	beditor           = {Leonid Reyzin and
	Douglas Stebila},
	title            = {HyperNova: Recursive Arguments for Customizable Constraint Systems},
	booktitle        = {Advances in Cryptology - {CRYPTO} 2024 - 44th Annual International
	Cryptology Conference, Santa Barbara, CA, USA, August 18-22, 2024,
	Proceedings, Part {X}},
	series           = {Lecture bnotes in Computer Science},
	pages            = {345--379},
	publisher        = {Springer},
	year             = {2024},
	bburl             = {https://bdoi.org/10.1007/978-3-031-68403-6\_11},
	bdoi              = {10.1007/978-3-031-68403-6\_11},
	bibburl           = {https://dblp.org/rec/conf/crypto/KothapalliS24.bib},
	bibsource        = {dblp computer science bibliography, https://dblp.org},
}

@inproceedings{bunz2023protostar,
	author           = {Benedikt B{\"{u}}nz and
	Binyi Chen},
	beditor           = {Jian Guo and
	Ron Steinfeld},
	title            = {Protostar: Generic Efficient Accumulation/Folding for Special-Sound
	Protocols},
	booktitle        = {Advances in Cryptology - {ASIACRYPT} 2023 - 29th International Conference
	on the Theory and Application of Cryptology and Information Security,
	Guangzhou, China, December 4-8, 2023, Proceedings, Part {II}},
	series           = {Lecture bnotes in Computer Science},
	pages            = {77--110},
	publisher        = {Springer},
	year             = {2023},
	bburl             = {https://bdoi.org/10.1007/978-981-99-8724-5\_3},
	bdoi              = {10.1007/978-981-99-8724-5\_3},
	bibburl           = {https://dblp.org/rec/conf/asiacrypt/BunzC23.bib},
	bibsource        = {dblp computer science bibliography, https://dblp.org},
}

@article{barrington1990uniformity,
	author    = {Barrington, David A. Mix and Immerman, Neil and Straubing, Howard},
	title     = {On Uniformity Within {NC}$^1$},
	journal   = {Journal of Computer and System Sciences},
	volume    = {41},
	number    = {3},
	pages     = {274--306},
	year      = {1990},
	month     = dec,
	bdoi       = {10.1016/0022-0000(90)90022-D},
	publisher = {Elsevier},
}

@article{anderson2003multivariate,
	title            = {Introduction to Multivariate Statistical Analysis},
	volume           = {66},
	bISSN             = {0002-9890},
	bburl             = {http://dx.bdoi.org/10.2307/2308777},
	bdoi              = {10.2307/2308777},
	number           = {5},
	journal          = {The American Mathematical Monthly},
	publisher        = {JSTOR},
	author           = {Madow, William G. and Anderson, T. W.},
	year             = {1959},
	month            = {May},
	pages            = {432},
}

@inproceedings{Brewer2000CAP,
	author           = {Eric A. Brewer},
	beditor           = {Gil Neiger},
	title            = {Towards robust distributed systems (abstract)},
	booktitle        = {Proceedings of the Nineteenth Annual {ACM} Symposium on Principles
	of Distributed Computing, July 16-19, 2000, Portland, Oregon, {USA}},
	pages            = {7},
	publisher        = {{ACM}},
	year             = {2000},
	bburl             = {https://bdoi.org/10.1145/343477.343502},
	bdoi              = {10.1145/343477.343502},
	bibburl           = {https://dblp.org/rec/conf/podc/Brewer00.bib},
	bibsource        = {dblp computer science bibliography, https://dblp.org},
}

@incollection{bergemann2006information,
	author    = {Bergemann, Dirk and V{\"a}lim{\"a}ki, Juuso},
	title     = {Information in Mechanism Design},
	beditor    = {Blundell, Richard and Newey, Whitney K. and Persson, Torsten},
	booktitle = {Advances in Economics and Econometrics: Theory and Applications, Ninth World Congress},
	series    = {Econometric Society Monographs},
	number    = {41},
	volume    = {1},
	chapter   = {5},
	pages     = {186--221},
	publisher = {Cambridge University Press},
	address   = {Cambridge, UK},
	year      = {2006},
	isbn      = {978-0-521-87152-5},
	bdoi       = {10.1017/CBO9781139052269.007},
	bnote      = {Invited symposium, Ninth World Congress of the Econometric Society, London, August 2005}
}

@incollection{Messick1989Validity,
	author    = {Messick, Samuel},
	title     = {Validity},
	beditor    = {Linn, Robert L.},
	booktitle = {Educational Measurement},
	edition   = {3},
	pages     = {13--103},
	publisher = {American Council on Education / Macmillan},
	address   = {New York, NY},
	year      = {1989}
}

@misc{templeton2024scaling,
	author       = {Templeton, Adly and Conerly, Tom and Marcus, Jonathan and Lindsey, Jack and Bricken, Trenton and Chen, Brian and Pearce, Adam and Citro, Craig and Ameisen, Emmanuel and Jones, Andy and Cunningham, Hoagy and Turner, Nicholas L. and McDougall, Callum and MacDiarmid, Monte and Tamkin, Alex and Durmus, Esin and Hume, Tristan and Mosconi, Francesco and Freeman, C. Daniel and Sumers, Theodore R. and Rees, Edward and Batson, Joshua and Jermyn, Adam and Carter, Shan and Olah, Chris and Henighan, Tom},
	title        = {Scaling Monosemanticity: Extracting Interpretable Features from {Claude 3 Sonnet}},
	howpublished = {Transformer Circuits Thread},
	year         = {2024},
	burl          = {https://transformer-circuits.pub/2024/scaling-monosemanticity/}
}

@inproceedings{yao2025tau,
	author       = {Shunyu Yao and
	Noah Shinn and
	Pedram Razavi and
	Karthik R. Narasimhan},
	title        = {\{{\(\tau\)}\}-bench: {A} Benchmark for {\textbackslash}underline\{T\}ool-{\textbackslash}underline\{A\}gent-{\textbackslash}underline\{U\}ser Interaction in Real-World Domains},
	booktitle    = {The Thirteenth International Conference on Learning Representations,
	{ICLR} 2025, Singapore, April 24-28, 2025},
	publisher    = {OpenReview.net},
	year         = {2025},
	burl          = {https://openreview.net/forum?id=roNSXZpUDN},
	bibsource    = {dblp computer science bibliography, https://dblp.org}
}

@misc{Krakovna2020SpecGaming,
	author       = {Krakovna, Victoria and Uesato, Jonathan and Mikulik, Vladimir and Rahtz, Matthew and Everitt, Tom and Kumar, Ramana and Kenton, Zac and Leike, Jan and Legg, Shane},
	title        = {Specification gaming: the flip side of {AI} ingenuity},
	howpublished = {DeepMind Blog},
	year         = {2020},
	burl          = {https://deepmind.google/blog/specification-gaming-the-flip-side-of-ai-ingenuity/}
}

@inproceedings{laaouach2025haltcot,
	title={{HALT}-CoT: Model-Agnostic Early Stopping for Chain-of-Thought Reasoning via Answer Entropy},
	author={Yassir Laaouach},
	booktitle={4th Muslims in ML Workshop co-located with ICML 2025},
	year={2025},
	burl={https://openreview.net/forum?id=CX5c7C1CZa}
}

@inproceedings{Amiri2025CoTLowerBounds,
	author = {Amiri, Alireza and Huang, Xinting and Rofin, Mark and Hahn, Michael},
	title = {Lower bounds for chain-of-thought reasoning in hard-attention transformers},
	year = {2025},
	publisher = {JMLR.org},
	booktitle = {Proceedings of the 42nd International Conference on Machine Learning},
	articleno = {120},
	numpages = {33},
	location = {Vancouver, Canada},
	series = {ICML'25}
}

@article{Chen2025SafetyCapability,
	title={Fundamental Safety-Capability Trade-offs in Fine-tuning Large Language Models}, 
	author={Pin-Yu Chen and Han Shen and Payel Das and Tianyi Chen},
	year={2025},
	journal={arXiv preprint},
	volume={arXiv.2503.20807},
	primaryClass={stat.ML},
	burl={https://arxiv.org/abs/2503.20807}, 
}

@article{Kratsios2025SharpLoRA,
	title={Sharp Generalization Bounds for Foundation Models with Asymmetric Randomized Low-Rank Adapters}, 
	author={Anastasis Kratsios and Tin Sum Cheng and Aurelien Lucchi and Haitz Sáez de Ocáriz Borde},
	year={2025},
	journal={arXiv preprint},
	volume={arXiv.2506.14530},
	primaryClass={stat.ML},
	burl={https://arxiv.org/abs/2506.14530}, 
}

@article{Karpowicz2025HallucinationImpossibility,
	title={On the Fundamental Impossibility of Hallucination Control in Large Language Models}, 
	author={Michał P. Karpowicz},
	year={2025},
	journal={arXiv preprint},
	volume={arXiv.2506.06382},
	primaryClass={stat.ML},
	burl={https://arxiv.org/abs/2506.06382}, 
}

@article{Mohsin2025ScaleLimits,
	title={On the Fundamental Limits of LLMs at Scale}, 
	author={Muhammad Ahmed Mohsin and Muhammad Umer and Ahsan Bilal and Zeeshan Memon and Muhammad Ibtsaam Qadir and Sagnik Bhattacharya and Hassan Rizwan and Abhiram R. Gorle and Maahe Zehra Kazmi and Nukhba Amir and Ali Subhan and Muhammad Usman Rafique and Zihao He and Pulkit Mehta and Muhammad Ali Jamshed and John M. Cioffi},
	year={2026},
	journal={arXiv preprint},
	volume={arXiv.2511.12869},
	primaryClass={cs.LG},
	burl={https://arxiv.org/abs/2511.12869}, 
}

@inproceedings{guo2014rational,
	author = {Guo, Siyao and Hub\'{a}\v{c}ek, Pavel and Rosen, Alon and Vald, Margarita},
	title = {Rational arguments: single round delegation with sublinear verification},
	year = {2014},
	isbn = {9781450326988},
	publisher = {Association for Computing Machinery},
	address = {New York, NY, USA},
	burl = {https://bdoi.org/10.1145/2554797.2554845},
	bdoi = {10.1145/2554797.2554845},
	booktitle = {Proceedings of the 5th Conference on Innovations in Theoretical Computer Science},
	pages = {523–540},
	numpages = {18},
	location = {Princeton, New Jersey, USA},
	series = {ITCS '14}
}

@inproceedings{scholten2022randomized,
	title={Randomized Message-Interception Smoothing: Gray-box Certificates for Graph Neural Networks},
	author={Yan Scholten and Jan Schuchardt and Simon Geisler and Aleksandar Bojchevski and Stephan G{\"u}nnemann},
	booktitle={Advances in Neural Information Processing Systems},
	beditor={Alice H. Oh and Alekh Agarwal and Danielle Belgrave and Kyunghyun Cho},
	year={2022},
	burl={https://openreview.net/forum?id=t0VbBTw-o8}
}

\end{document}